\documentclass[10pt,journal,compsoc]{IEEEtran}
\ifCLASSOPTIONcompsoc
  \usepackage[nocompress]{cite}
\else
  \usepackage{cite}
\fi
\ifCLASSINFOpdf
\else
\fi
\usepackage{url}

\usepackage{multirow}
\usepackage[table,xcdraw]{xcolor}

\usepackage{caption}
\usepackage{graphicx} 
\usepackage{amsmath}
\usepackage{amssymb}
\usepackage{multirow}
\usepackage{subfigure}
\usepackage{makecell}
\usepackage{booktabs}
\usepackage{amsmath}
\usepackage{adjustbox}
\usepackage{float}
\usepackage[colorlinks=true,citecolor=black,linkcolor=black]{hyperref}
\usepackage{lscape}
\usepackage[nameinlink]{cleveref}
\Crefname{figure}{Fig.}{Fig.}
\Crefname{table}{Tab.}{Tab.}
\Crefname{section}{Sec.}{Sec.}

\usepackage{xr}

\newcommand{\tabref}[1]{Tab.~\ref{#1}}

\newcommand{\figref}[1]{Fig.~\ref{#1}}
\newcommand{\secref}[1]{Sec.~\ref{#1}}

\newcommand{\eg}{\textit{e.g.}}

\begin{document}
%
\title{A Study on the Generality of Neural Network Structures for Monocular Depth Estimation}
%
%
%
%

\author{Jinwoo Bae${^\dag}$, Kyumin Hwang and Sunghoon Im${^*}$ 
\IEEEcompsocitemizethanks{
\IEEEcompsocthanksitem ${^\dag}$ Work done in DGIST.
\IEEEcompsocthanksitem ${^*}$ Corresponding author.
\IEEEcompsocthanksitem J. Bae is with Hyundai Motor Group, Seoul, Republic of Korea. jinwoobae@hyundai.com
\IEEEcompsocthanksitem K. Hwang and S. Im are with the Department of Electrical Engineering and Computer Science, DGIST, Daegu, 42988, Republic of Korea. \{kyumin, sunghoonim\}@dgist.ac.kr}}

%
%

\markboth{IEEE TRANSACTIONS ON PATTERN ANALYSIS AND MACHINE INTELLIGENCE, VOL. XX, NO. XX, 2023}%
{}
%



\IEEEtitleabstractindextext{%
\begin{abstract}
Monocular depth estimation has been widely studied, and significant improvements in performance have been recently reported.
However, most previous works are evaluated on a few benchmark datasets, such as KITTI datasets, and none of the works provide an in-depth analysis of the generalization performance of monocular depth estimation.
In this paper, we deeply investigate the various backbone networks (\eg CNN and Transformer models) toward the generalization of monocular depth estimation.
First, we evaluate state-of-the-art models on both in-distribution and out-of-distribution datasets, which have never been seen during network training.
Then, we investigate the internal properties of the representations from the intermediate layers of CNN-/Transformer-based models using synthetic texture-shifted datasets.
Through extensive experiments, we observe that the Transformers exhibit a strong shape-bias rather than CNNs, which have a strong texture-bias.
We also discover that texture-biased models exhibit worse generalization performance for monocular depth estimation than shape-biased models.
We demonstrate that similar aspects are observed in real-world driving datasets captured under diverse environments.
Lastly, we conduct a dense ablation study with various backbone networks which are utilized in modern strategies. 
The experiments demonstrate that the intrinsic locality of the CNNs and the self-attention of the Transformers induce texture-bias and shape-bias, respectively.

\end{abstract}

}

\maketitle

\IEEEdisplaynontitleabstractindextext

%
\IEEEpeerreviewmaketitle

\section{Introduction}

Monocular depth estimation (MDE) is widely utilized for spatial recognition technologies such as autonomous driving \cite{xue2020toward,cheng2022physical,de2021deep} or AR/VR \cite{benavides2022phonedepth,wu2022toward} because of its portability and cost-effectiveness. Various MDE processes have achieved remarkable progress over the past decade.
Most previous works \cite{godard2017unsupervised,fu2018deep,godard2019digging,guizilini20203d,bhat2021adabins,yang2021transformer,lyu2020hr,li2022depthformer} concentrate on boosting performance using limited benchmark datasets, especially the KITTI dataset~\cite{geiger2013vision}. 
However, these works have not provided a deep investigation into what MDE networks have learned. This means that they cannot guarantee the model's behavior is correct.
To examine the interpretability of MDE networks, one previous work \cite{hu2019visualization} employs a target network for MDE model analysis. Another approach \cite{dijk2019neural} uses a synthetic dataset, which contains the changes in image contents (e.g., camera pose). However, universality questions  about other networks still remain because of the fixed specific network \cite{godard2017unsupervised} in certain experimental conditions. 

\begin{figure}[t!]
    \centering
    \resizebox{\columnwidth}{!}{
    \includegraphics{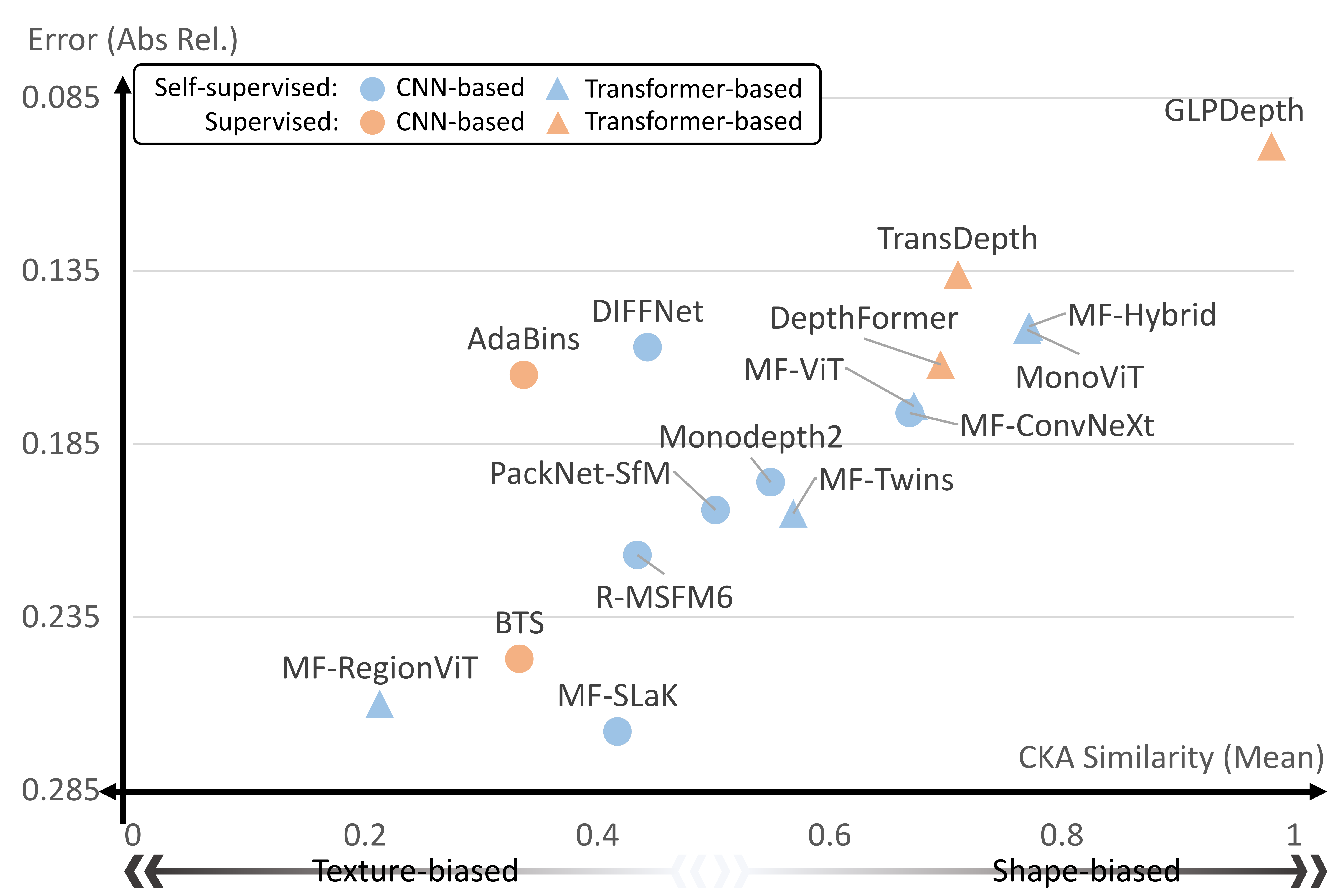}
    }
    \caption{\textbf{Analysis on the generality of state-of-the-art models and modern network structures.} The $x$-axis is the CKA similarity indicating whether the network is shape biased or texture biased.
    The $y$-axis shows Absolute Relative error where a lower number is better performance. We use the synthetic texture-shifted datasets described in \secref{texture-shifted-exp}.}
    \label{figure_concept}
    \vspace{-0.5cm}
\end{figure}

Recently, several works \cite{selvaraju2017grad,geirhos2018imagenet,islam2021shape,tuli2021convolutional} aim to analyze interpretability, taking  inspiration from convolutional neural networks (CNNs) whose designs are based on human visual processes \cite{lindsay2021convolutional,de2021humans}.
Then, how can humans efficiently extract and recognize important information from complex scenes? Compared to other cues like texture or color, the biological vision system considers the object's shape to be the single most important visual cue \cite{landau1988importance}. This allows humans, including little kids, to easily distinguish an object from a line drawing or a silhouette image. Many researchers believe that CNN would also behave similarly \cite{hubel1959receptive,fukushima1988neocognitron,kriegeskorte2015deep}.
However, in contrast to belief, recent studies \cite{geirhos2018imagenet,morrison2021exploring,tuli2021convolutional,naseer2021intriguing} have discovered that CNNs are heavily predisposed to recognize textures rather than shapes. CNN-based models accurately assign labels to images even when the shapes of the structures are disturbed \cite{gatys2017texture,brendel2019approximating}. On the other hand, CNN models are unable to accurately predict labels in a texture-removed image with a well-preserved shape \cite{ballester2016performance}. 
Transformers \cite{vaswani2017attention} achieve outstanding performance on various computer vision tasks \cite{carion2020end,xie2021segformer,guo2021sotr} and have attracted much attention. Moreover, many works \cite{morrison2021exploring,zhang2021delving,gu2022vision,bhojanapalli2021understanding} reveal that Transformers have a strong shape bias notwithstanding the lack of a spatial locality, compared to CNNs. Due to its strong shape bias, Transformer is considered more robust than a CNN and more similar to human cognitive processes \cite{paul2022vision}.

Then, how does this observation affect the MDE? We hypothesize two things. 
First, the network's generality will differ depending on the texture-/shape-bias. To verify the generality, we evaluate state-of-the-art MDE models trained on KITTI \cite{geiger2013vision} using five public depth datasets (SUN3D \cite{xiao2013sun3d}, RGBD \cite{silberman2012indoor}, MVS \cite{ummenhofer2017demon}, Scenes11 \cite{ummenhofer2017demon}, and ETH3D \cite{schops2017multi}). We also conduct experiments on six different texture-shifted datasets, including three synthetic texture-shifted datasets (Watercolor, Pencil-sketch, Style-transfer) and three real-world texture-shifted datasets (Oxford Robotcar \cite{maddern20171}, Rainy Cityscapes \cite{hu2019depth}, Foggy Cityscapes \cite{sakaridis2018semantic}).
Through these experiments, we confirm that texture-bias is vulnerable to generalization while shape-bias shows robust generalization performance.
Second, the texture-/shape-bias are related to the intrinsic properties of CNN and Transformer structures. The modern Transformer-based model \cite{chen2021regionvit,chu2021twins} imitates the intrinsic locality inductive bias of the CNNs. The modern CNN-based model \cite{liu2022convnet,liu2022more} is designed to mimic the self-attention of the Transformer. 
We finally conduct ablation studies on the generalization performance, and the texture-/shape-bias of various modern backbone structures that originate from a specific design for each CNN and Transformer (\eg, locality and self-attention) such as RegionViT~\cite{chen2021regionvit}, Twins~\cite{chu2021twins}, ConvNeXt~\cite{liu2022convnet} and SLaK~\cite{liu2022more}.


This paper extends our previous work in \cite{bae2022deep} which proposes a new self-supervised monocular depth estimation network adopting Transformers~\cite{vaswani2017attention}. The Transformer-based network shows the generalization performance on various environments on depth estimation. While the short version \cite{bae2022deep} only addresses self-supervised MDE models, the extended version deals with full MDE models, including both self-supervised and supervised methods. Moreover, we deeply analyze the reason why the proposed network achieves better-generalized performance rather than CNN-based models by comparing the performance, and intermediate-layer feature similarity \cite{kornblith2019similarity,raghu2021vision} on various texture-shifted datasets. 
As shown in \figref{figure_concept}, we observe that the Transformer structure has more shape-biased properties than the CNN structure, which has texture-biased properties.
It enables the Transformer-based models to achieve better generalization performance than the CNN-based models. 
We also experiment extensively on modern backbone architectures (\eg, ConvNeXt \cite{liu2022convnet}, RegionViT \cite{chen2021regionvit}) to investigate the origin of the texture-/shape-biased properties.
Through these extensive experiments, we observe that the intrinsic locality of CNNs induces texture-biased characteristics, while the self-attention mechanism, the base layer of Transformers, induces shape-biased properties. 

\section{Related Work}

\begin{figure*}[t!] 
\begin{center}
\includegraphics[width=0.9\textwidth]{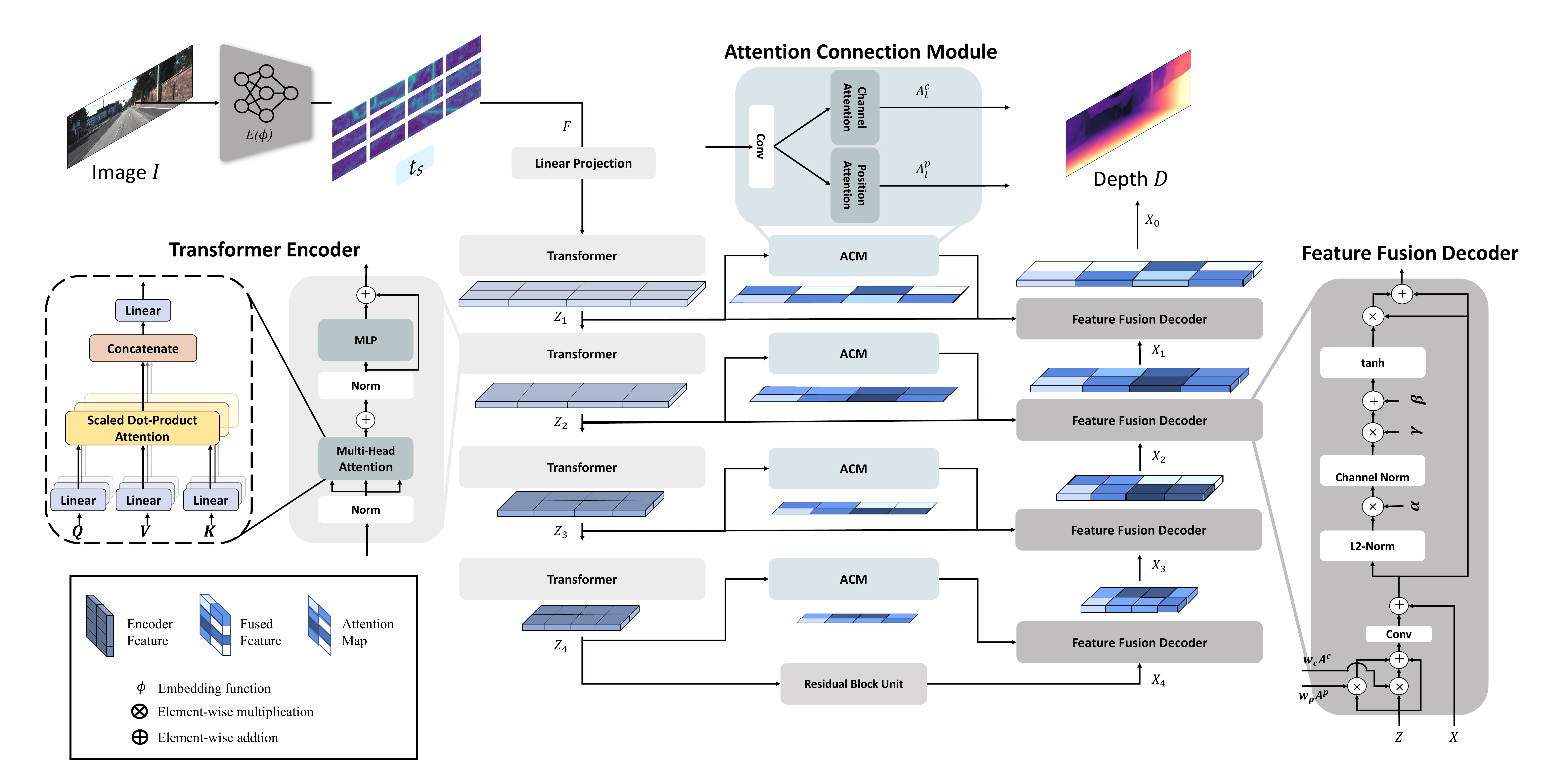}
\end{center}
\vspace{-3mm}
\caption{\textbf{Overall Architecture of MonoFormer (MF-ours).} }
\label{figure_network_overview}
\end{figure*} 

\subsection{Self-supervised monocular depth estimation}
Self-supervised depth estimation methods \cite{zhou2017unsupervised,godard2019digging,guizilini20203d,lyu2020hr,klingner2020self,xiong2021self} simultaneously train depth and motion network by imposing photometric consistency loss between the target and source images warped by the predicted depth and motion.
SfMLearner \cite{zhou2017unsupervised} first proposes a depth and ego-motion estimation pipeline without the ground truth depth and motion.
Monodepth2 \cite{godard2019digging} presents a minimum reprojection loss to handle occlusions, a full-resolution multi-scale sampling method to reduce visual artifacts, and an auto-masking loss to ignore outlier pixels. 
PackNet-SfM \cite{guizilini20203d} introduces packing and unpacking blocks that leveraged 3D convolutions to learn the dense appearance and geometric information in real-time. 
HR-Depth \cite{lyu2020hr} analyzes the reason for the inaccurate depth prediction in large gradient regions and designed a skip connection to extract representative features in high resolution.

\subsection{Supervised monocular depth estimation}
Supervised methods \cite{lee2019big,song2021monocular,kim2022global,li2022depthformer,bhat2021adabins} use a ground truth depth acquired from RGB-D cameras or LiDAR sensors for supervision in training. 
They also estimate depth maps given a single image as input.
BTS \cite{lee2019big} adopts a local planar guidance layer to densely encoded features to preserve local detail and create depth map sharpness at multi-stages in the decoder. 
AdaBins \cite{bhat2021adabins} estimate the depth by linear combinations of bin centers that are adaptively decided per image. The bin building block divides the depth range of the image into bins.
LapDepth \cite{song2021monocular} employs a Laplacian pyramid at the multi-level upscaling encoder to preserve local detail, such as a boundary. It also trains stably by utilizing weight standardization.
GLPDepth \cite{kim2022global} proposes a hierarchical transformer encoder to capture the global context of images and a selective feature fusion module to connect multi-scale local features and global context information at the decoder. The feature fusion module helps the decoder become more powerful, even if the decoder is lightweight.
DepthFormer \cite{li2022depthformer} proposes leveraging the transformer's effective attention mechanism and the spatial inductive bias of the CNN to capture long-range correlation. It also uses a hierarchical aggregation and heterogeneous interaction module to enhance the affinity of the network.

\subsection{Vision Transformers}
Recently, Transformers \cite{vaswani2017attention} has shown promise for solving computer vision tasks such as image classification \cite{dosovitskiy2020image,touvron2021training}, object detection \cite{carion2020end}, and dense prediction \cite{zheng2021rethinking,ranftl2021vision,yang2021transformer,guizilini2022multi}. 
ViT \cite{dosovitskiy2020image} employs a Transformers architecture on fixed-size image patches for image classification for the first time. DeiT \cite{touvron2021training} utilizes Knowledge Distillation on ViT architecture, showing good performance only with the ImageNet dataset.
DETR \cite{carion2020end} proposes the direct set prediction approach, which simplifies the object detection pipeline, based on a CNN-Transformer network and bipartite matching.
Some works \cite{ranftl2021vision,yang2021transformer} have employed Transformers for monocular depth estimation in a supervised manner. 
DPT \cite{ranftl2021vision} introduces a dense prediction using a Transformer as the basic computational building block of the encoder. These works show  generalized performance, but they require a large number of training datasets captured in diverse environments with ground truth depth maps. 
TransDepth \cite{yang2021transformer} utilizes multi-scale information to capture local level details. Previous works lack studies on whether models behave as intended on another domain dataset.
The work \cite{ruhkamp2021attention} aggregates the attention map between a single frame and cross frames to refine the attraction map to improve performance. The works \cite{yang2021transformer,ruhkamp2021attention} only focus on improving performance on benchmark datasets.

\subsection{Modern Architecture}
Despite the tremendous success of Transformers for vision tasks, a Transformer requires a large model and data size to achieve state-of-the-art performance. 
Recently, many works \cite{liu2021swin,chen2021regionvit,chu2021twins,liu2022convnet,liu2022more} utilize local attention on the Transformer to alleviate problems.
The Swin Transformer \cite{liu2021swin} designs a pyramid structure network differently from a vision transformer \cite{dosovitskiy2020image}, which is an isotropic structure. It achieves state-of-the-art performance in classification tasks with local attention using the sliding window strategy.
Twins \cite{chu2021twins} employs global attention for the non-overlapping region and local attention to perform better under limited conditions.
RegionViT \cite{chen2021regionvit} shows state-of-the-art performance on several visual tasks using regional-to-local attention, which alleviates the weakness of the standard attention mechanism.
ConvNeXt \cite{liu2022convnet} modernizes convolution neural layers such as depth-wise convolution and utilizes techniques such as Patchify stem and achieves competitive performance with the Transformer.
SLaK \cite{liu2022more} attempts to design the network with an extremely large kernel size (\eg, $51\times51$) to leverage the sparsity that is observed from the human visual system. SLaK \cite{liu2022more} repeats the Prune-and-Grow step in training to optimize the sparse kernels.

\section{Method}
\begin{figure*}
    \centering
    \resizebox{\linewidth}{!}{
    \begin{tabular}{ccccc}
     \includegraphics[]{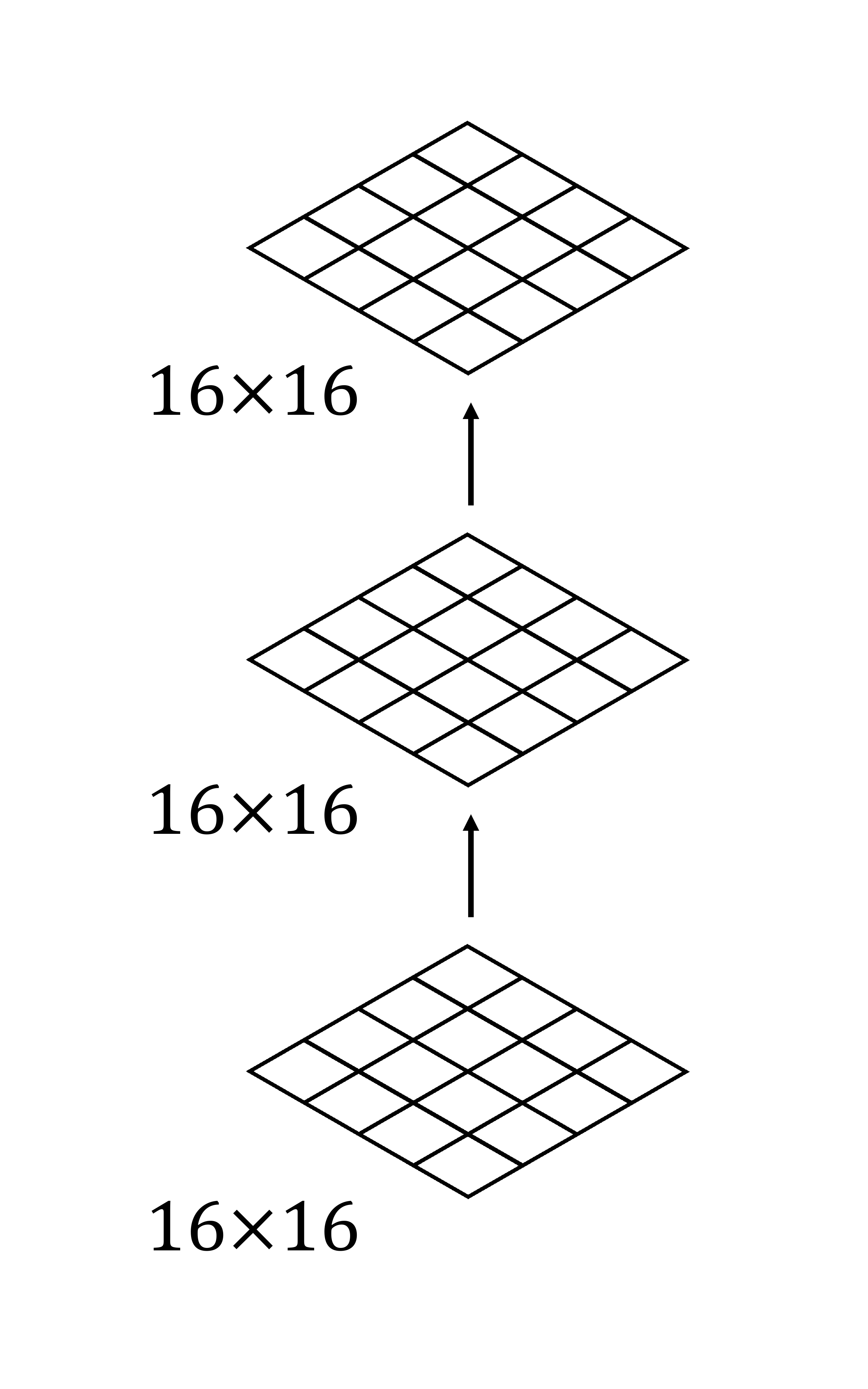}&
     \includegraphics[]{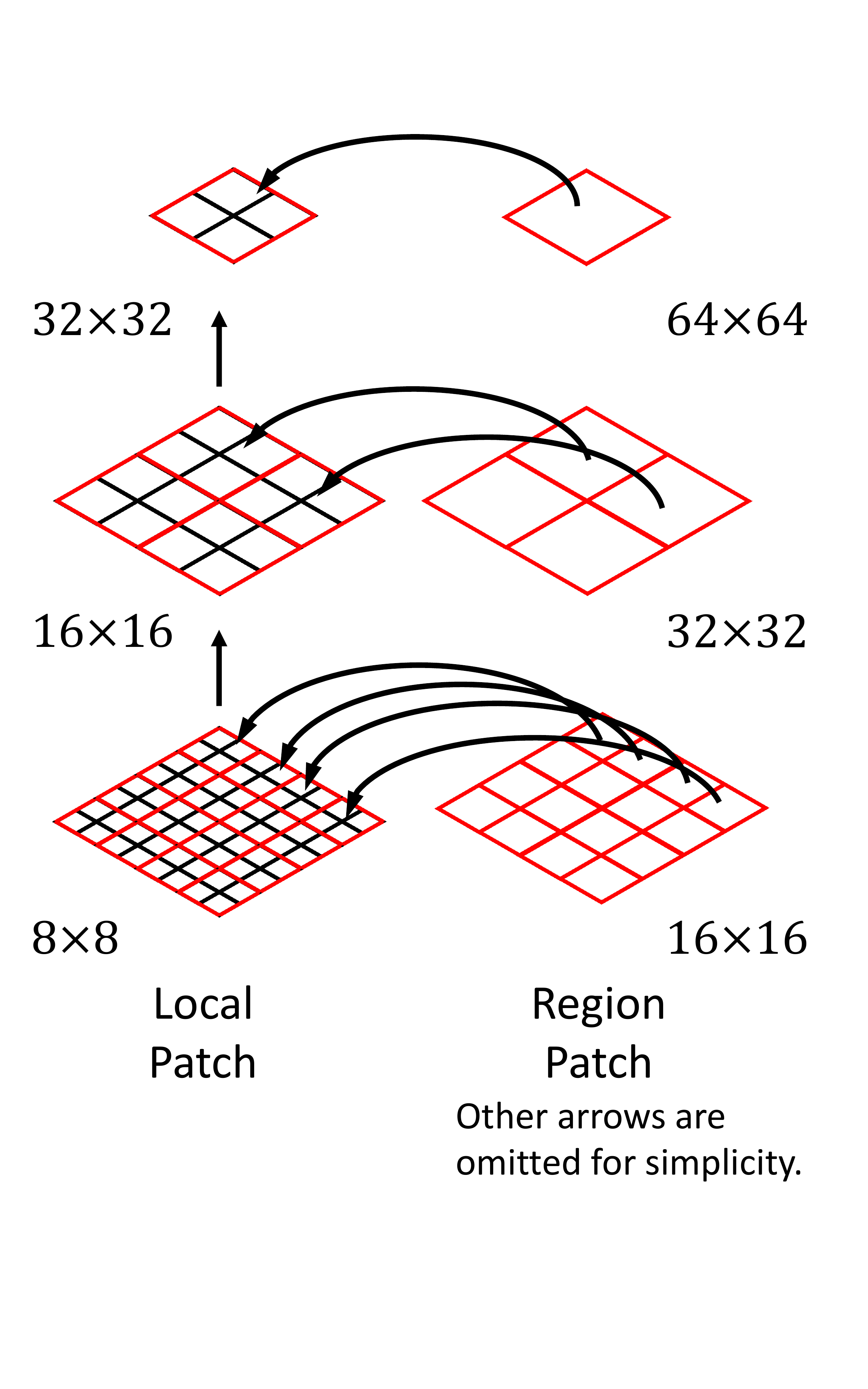}& 
     \includegraphics[]{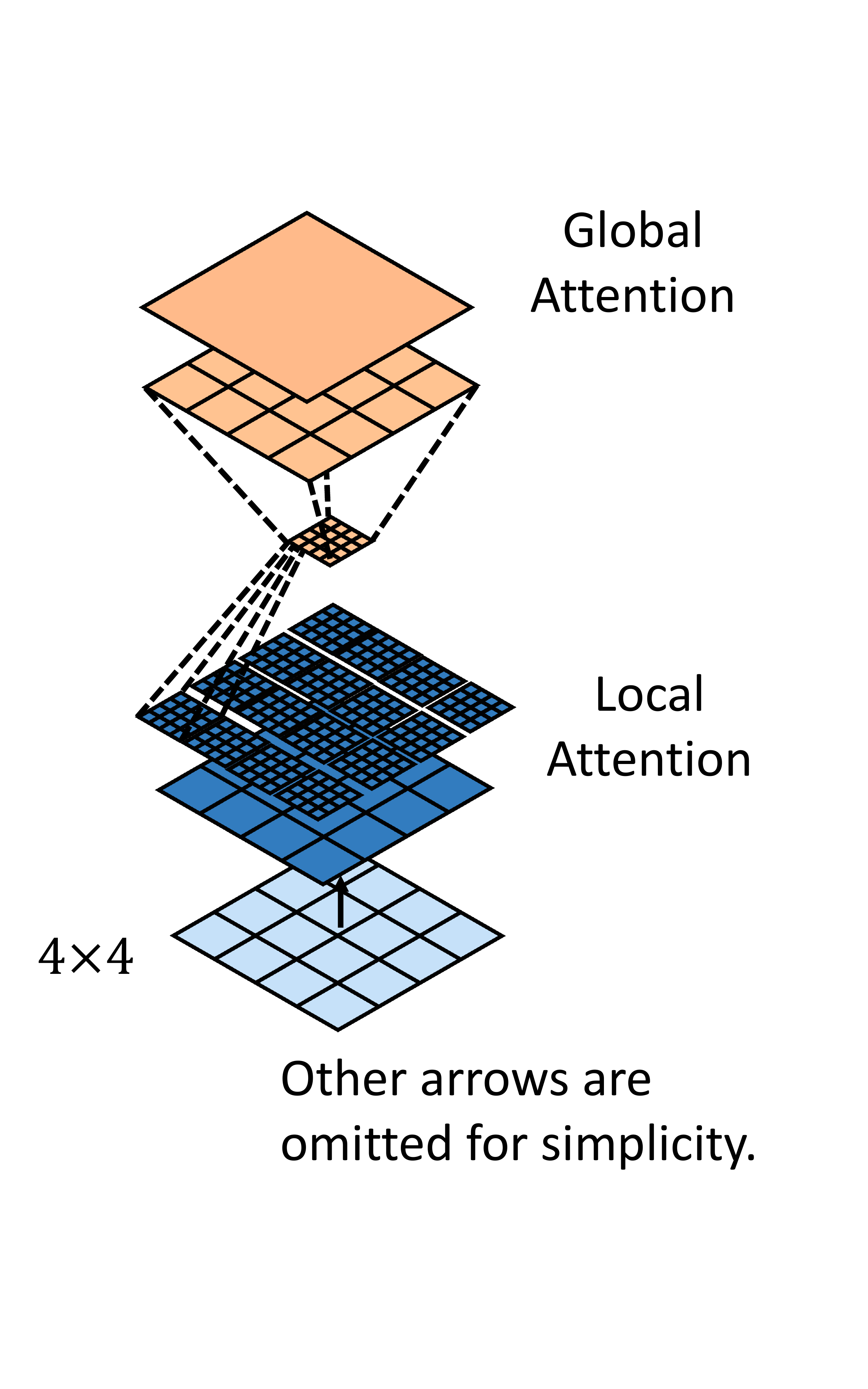}& 
     \includegraphics[]{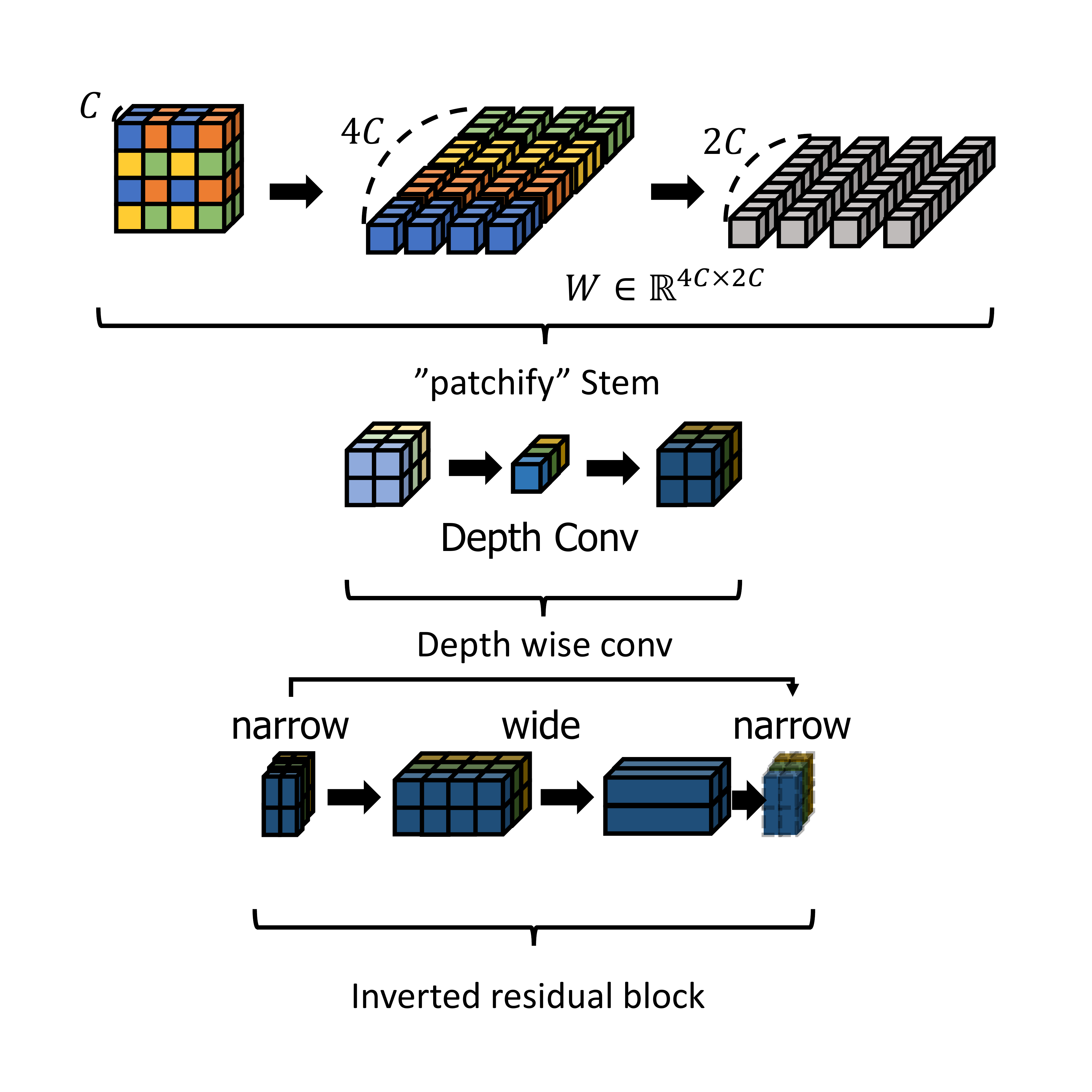}& 
     \includegraphics[]{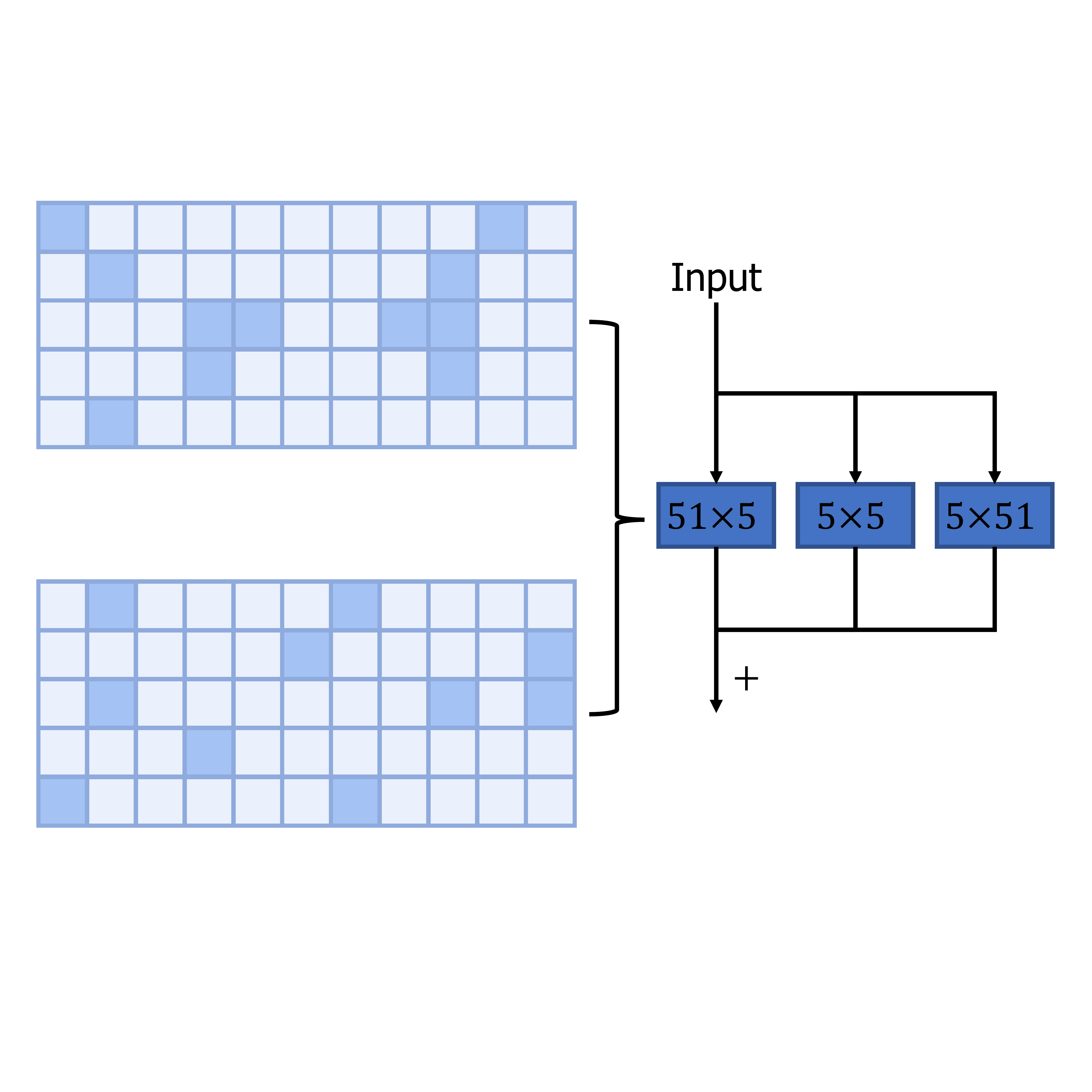}\\
     \fontsize{70}{50} \selectfont ViT~\cite{dosovitskiy2020image} & 
     \fontsize{70}{50} \selectfont RegionViT~\cite{chen2021regionvit}  & 
     \fontsize{70}{50} \selectfont Twins~\cite{chu2021twins}  & 
     \fontsize{70}{50} \selectfont ConvNeXt~\cite{liu2022convnet}  &
     \fontsize{70}{50} \selectfont SLaK~\cite{liu2022more} 
    \end{tabular}
    }
    \caption{\textbf{Illustrations of various modern architectures.}}
    \label{figure_modern_summary}
    \vspace{-0.3cm}
\end{figure*}

This section describes MonoFormer, which is an encoder-decoder structure with a multi-level feature fusion module for self-supervised monocular depth estimation described in \secref{sec:encoder}. 
An Attention Connection Module (ACM) learns the channel and position attentions in \secref{sec:ACM}. A Feature Fusion Decoder (FFD) adaptively fuses the encoder features with the attention maps in \figref{figure_network_overview}.

\subsection{Transformer-based Encoder}
\label{sec:encoder}
MonoFormer \cite{bae2022deep} is composed of a CNN and Transformer for an image encoder. 
The encoder employs ResNet50 \cite{he2016deep} as the CNN backbone ($E(\phi)$ in \figref{figure_network_overview}), and $L$ number of Transformers. MonoFormer sets the $L$ to 4.
The encoder is used to extract a feature map $F \in \mathbb{R}^{C \times H \times W}$ from an input image $I$, and the map is divided into $N$ $(= \frac{H}{16} \times \frac{W}{16})$ number of patches $p_{n} \in \mathbb{R}^{C\times 16 \times 16}$, which is utilized as the input of the first Transformer layer.
Following the work \cite{ranftl2021vision}, MonoFormer additionally use a special token $t_s$.
MonoFormer input the patch tokens $p_{n},~n \in\{{1,...,N}\}$ and the special token $t_s$ with a learnable linear projection layer $E$ as follows:
\begin{equation}
\small
    Z_0 = [t_s;~p_1E;~p_2E;~...~;~p_{N}E],
\end{equation}
where $Z_0$ is the latent embedding vector.
The CNN-Transformer encoder comprises a Multi-head Self-Attention (MSA) layer, a Multi-Layer Perceptron (MLP) layer, and Layer Norm (LN) layers. The MLP is built with GELU non-linearity \cite{hendrycks2016gaussian}. The LN is used before each block, and residual connections are used after every block.
Self-Attention (SA) at each layer $l \in \{1,...,L\}$ is processed with the learnable parameters $W^m_Q , W^m_K , W^m_V \in \mathbb{R}^{C \times d}$ of \{query, key, value\} weight matrices, given the embedding vector $Z_l \in \mathbb{R}^{N \times C}$ as follows:
\begin{equation}
\small
\begin{gathered}
        \text{SA}^m_{l-1} = \text{softmax}\big(\frac{Q^m_{l-1} (K^m_{l-1})^\text{T}}{\sqrt{d}} \big)V^m_{l-1},~m \in \{1,...,M\},\\
    Q^m_{l-1} = Z_{l-1}W^m_Q,~K^m_{l-1} = Z_{l-1}W^m_K,~V^m_{l-1} = Z_{l-1}W^m_V,
\end{gathered}
\end{equation}
where $M$ and $d$ are the number of SA blocks and the dimension of the self-attention block, which is the same as the dimension of the weight matrices, respectively. 
The Multi-head Self-Attention (MSA) consists of the $M$ number of SA blocks with the learnable parameters of weight matrices $W \in \mathbb{R}^{Md \times C }$ as follows:
\begin{equation}
\small
\begin{gathered}
\text{MSA}_{l-1} = Z_{l-1} +  \text{concat}(\text{SA}^1_{l-1};~\text{SA}^2_{l-1};~ \dots ;~\text{SA}^M_{l-1} )W,\\
Z_{l} = \text{MLP}(\text{LN}(\text{MSA}_{l-1}))+\text{MSA}_{l-1}.
\end{gathered}
\end{equation}
This Transformer layer is repeated $L$ times with unique learnable parameters. The outputs of the Transformers $\{Z_1,...,Z_L\}$ are utilized as the input of the following layers ACM and FFD.
The hybrid encoder can be replaced with the other backbones.
We evaluate all the performance of our depth estimation networks with various backbones by changing the encoder to the modern backbone structures, such as ViT~\cite{dosovitskiy2020image}, RegionViT~\cite{chen2021regionvit}, Twins~\cite{chu2021twins}, ConvNeXt~\cite{liu2022convnet} and SLaK~\cite{liu2022more} as shown in \figref{figure_modern_summary}.

\subsection{Attention Connection Module (ACM)}
\label{sec:ACM}
The skip connection module of MonoFormer, ACM, that extracts global context attention and a semantic presentation from the given features $Z_{l},~l\in\{1,...,L\}$. 
The skip connection, widely utilized for dense prediction tasks  \cite{ronneberger2015u}, helps keep the fine detail by directly transferring the spatial information from the encoder to the decoder. 
However, because of its simplicity, it is challenging for the naive skip connection method to preserve local detail like object boundaries \cite{zhou2018unet++}. 
To address the issue, the ACM is proposed that extracts attention weight from the spatial and channel domains inspired by \cite{fu2019dual}.
It consists of position attention, channel attention modules, and a fusion block that gathers important information from two attentions.
The position attention module produces a position attention map $A^p_{l} \in \mathbb{R}^{C \times N}$ as follows:
\begin{equation}
\small
    A^p_{l} = \text{softmax}({Q}^p_{l} ({K}^p_{l})^\text{T})V^p_{l},
\end{equation}
where ${Q}^p_l, {K}^p_l$ and ${V}^p_l$ are the query, key, and value matrices computed by passing $Z_{l}$ through a single convolutional layer.
The channel attention module directly calculate the channel attention map $ A^c_l \in \mathbb{R}^{C \times N}$ by computing the gram matrix of $Z_l$ as follows: 
\begin{equation}
\small
    A^c_l = \text{softmax}(Z_lZ_l^{\text{T}}).
\end{equation}
The position attention map $A^p_l$ and channel attention map $A^c_l$ enhance the feature representation by capturing long-range context and exploiting the inter-dependencies between each channel map, respectively. 
These two attention maps are utilized in the following section, which highlights the importance of the features.

\subsection{Feature Fusion Decoder (FFD)}
\label{sec:FFD}
The FFD gets the encoder features $Z_l$, the attention maps $A^p_l,~A^c_l$, and the output feature $X_L$ of the last Transformer layer passed through a Residual convolutional layer. 
The features $X_{L-l+1},~l \in \{1,...,L\}$ are fused through the decoder with a single Convolutional layer (Conv) and Channel Normalization (CN) with learnable parameters $\alpha, \beta$ and $\gamma$ as follows:
\begin{equation}
\small
\begin{gathered}
    X_{L-l} = \hat{X}_{L-l}[1+\tanh(\gamma(\text{CN}(\alpha||\hat{X}_{L-l}{||}_2  + \beta )],
    \\
    \hat{X}_{L-l} = \text{Conv}(w_pA^p_lZ_l + w_cA^c_lZ_l + Z_l) + X_{L-l+1},
\end{gathered}
\end{equation}
where $w_{p}$ and $w_c$ are the learnable parameters that determine the importance of the position and channel attentions \cite{zhang2019self}.
The parameter $\alpha$ works so that each channel can learn about each other individually, and $\gamma$ and $\beta$ control the activation channel-wisely following the work in \cite{yang2020gated}. 
Through this process, the FFD can assemble the local detailed semantic representation and the global context from the fused features to produce the fine depth map.

\subsection{Training loss and implementation detail}
We train both depth and motion networks using photometric consistency (L1 loss and SSIM loss) and edge-aware smoothness losses following the best practices of self-supervised monocular depth estimation \cite{zhou2017unsupervised,godard2019digging,guizilini20203d}.
We set the weight for SSIM, L1  photometric, and smoothness losses as $0.85$, $0.15$ and $0.001$, respectively. 
We use 7 convolution layers for 6-DoF camera pose estimation following the work in \cite{zhou2017unsupervised}.
We implement our framework on PyTorch and train it on 4 Titan RTX GPUs. We use the Adam optimizer \cite{kingma2014adam} with $\beta_1 = 0.9$ and $\beta_2 = 0.999$. Our model is trained for 50 epochs with a batch size of 8. The learning rates for depth and pose network are $2 \times 10^{-5}$ and $5 \times 10^{-4}$, respectively. \\ The MF-(ConvNeXt/SLaK/RegionViT/Twins/ViT) models to experiment with the change of encoder in Monoformer were all trained with the same conditions as Monoformer by importing the ImageNet\cite{deng2009imagenet} pre-trained weights. \\
In addition, to analyze the tendency of different supervised methods, we used the pre-trained supervised models~\cite{lee2019big, bhat2021adabins, yang2021transformer, li2022depthformer, kim2022global}, which are provided by authors, in our experiments. Almost supervised models are optimized by scale-invariant log loss \cite{eigen2014depth}. The detail of  supervised method's loss is described in  \cite{eigen2014depth}. \\ The code is available: \textit{https://github.com/sjg02122/MonoFormer.} 

\begin{figure*}[t]
\newcommand\w{110}
\newcommand\h{60}
\newcommand\ww{3cm}
\newcommand\wh{0.1cm}
\begin{subfigure}
\centering
\resizebox{\textwidth}{!}{%
\begin{tabular}{ccccc}
 \includegraphics[]{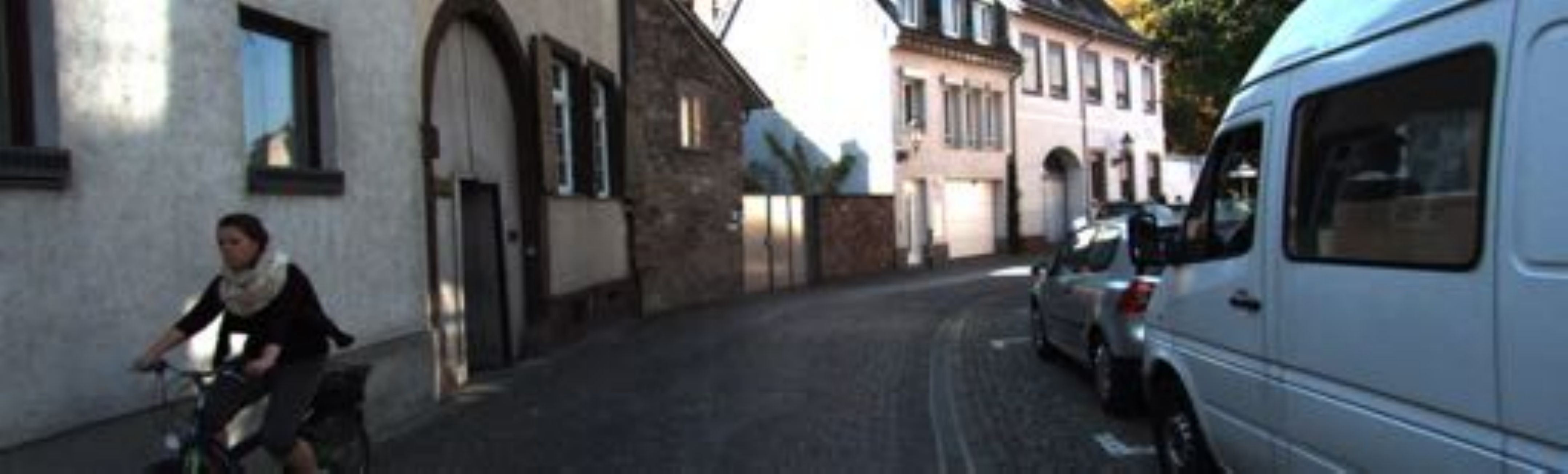}&
 \includegraphics[]{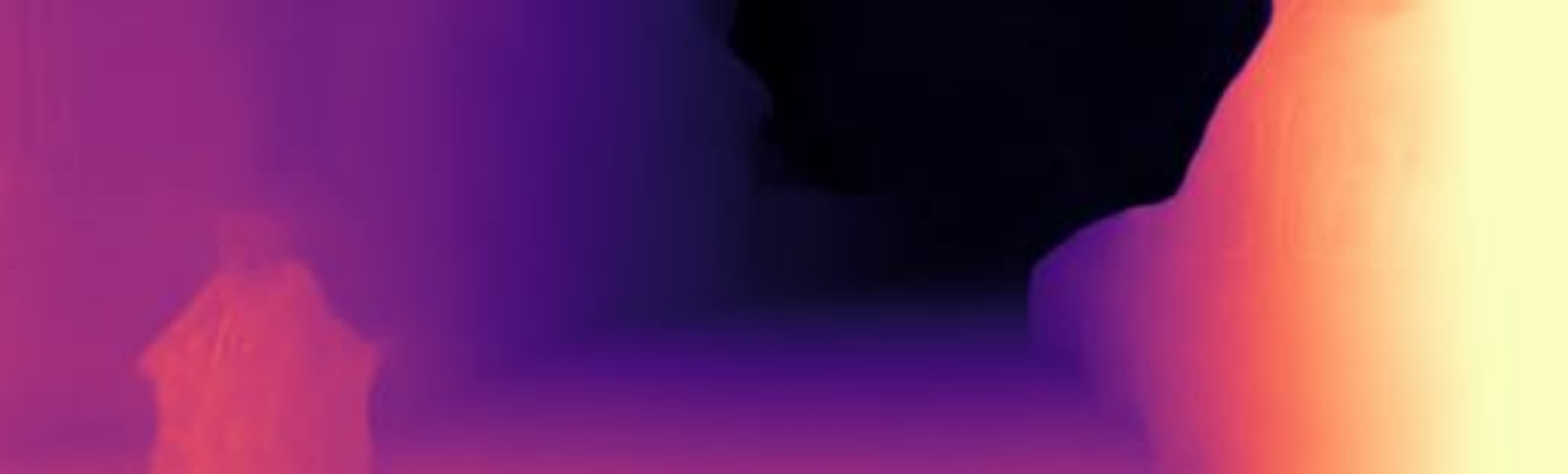}&  
 \includegraphics[]{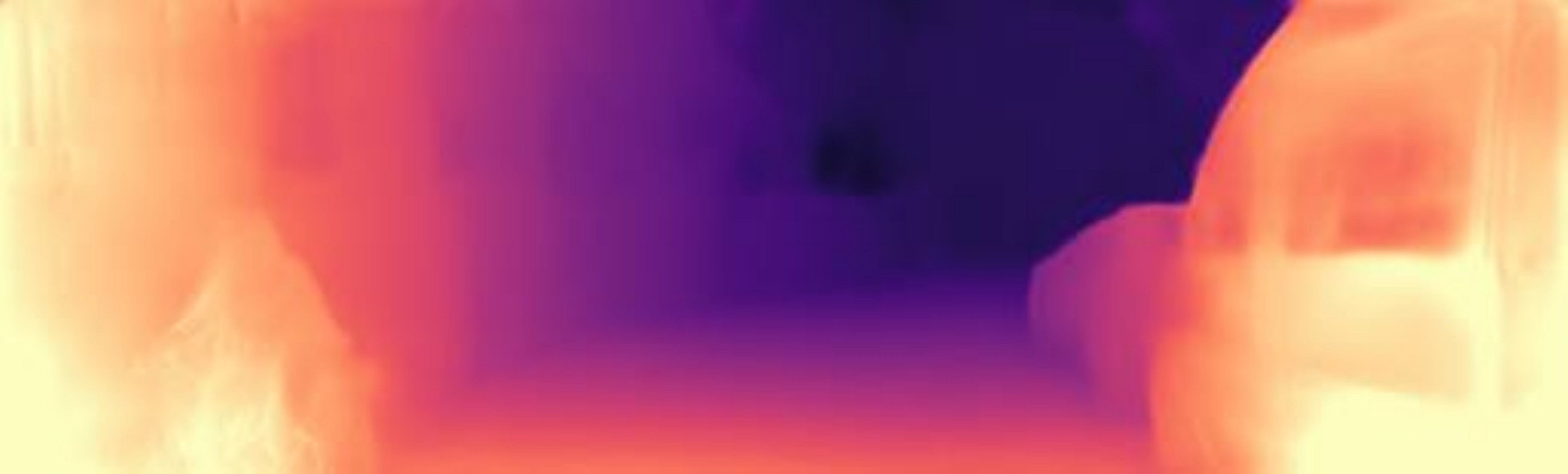}&  
 \includegraphics[]{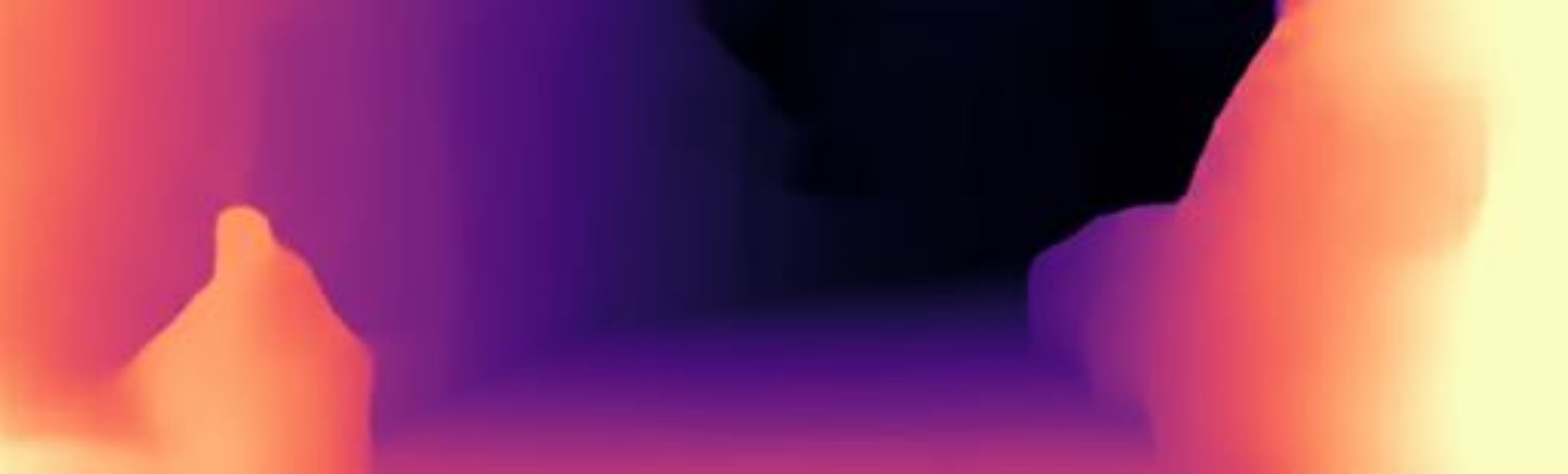}& 
 \includegraphics[]{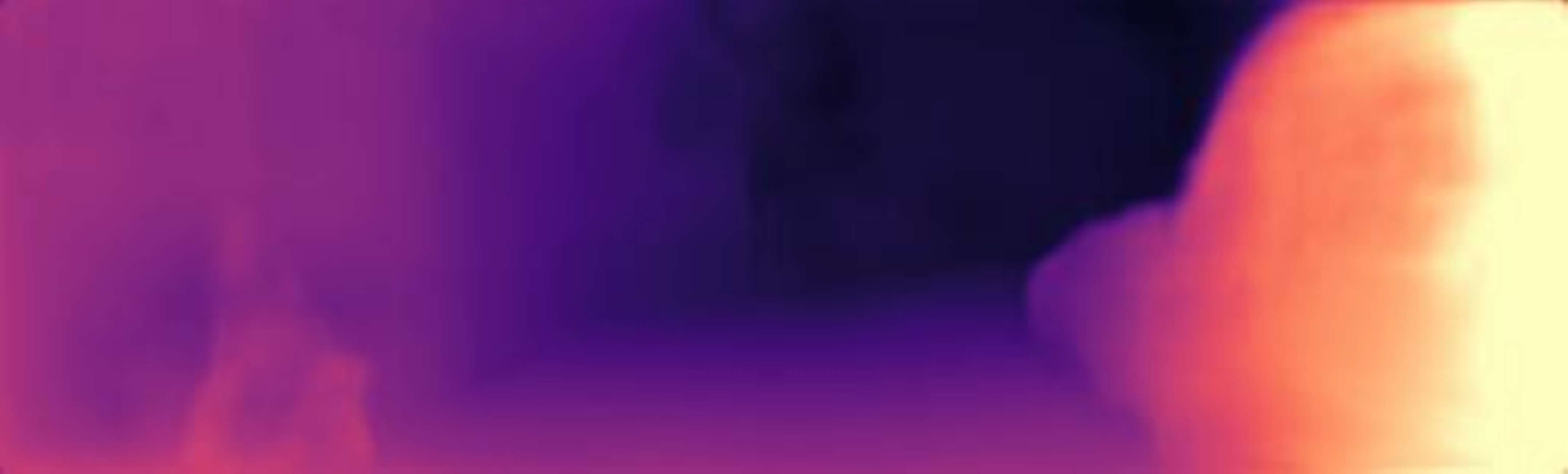}\\
 \fontsize{\w}{\h} \selectfont Input images&  
 \fontsize{\w}{\h} \selectfont Monodepth2&  
 \fontsize{\w}{\h} \selectfont PackNet-SfM&  
 \fontsize{\w}{\h} \selectfont R-MSFM6&
 \fontsize{\w}{\h} \selectfont MF-ConvNeXt\\
 \includegraphics[]{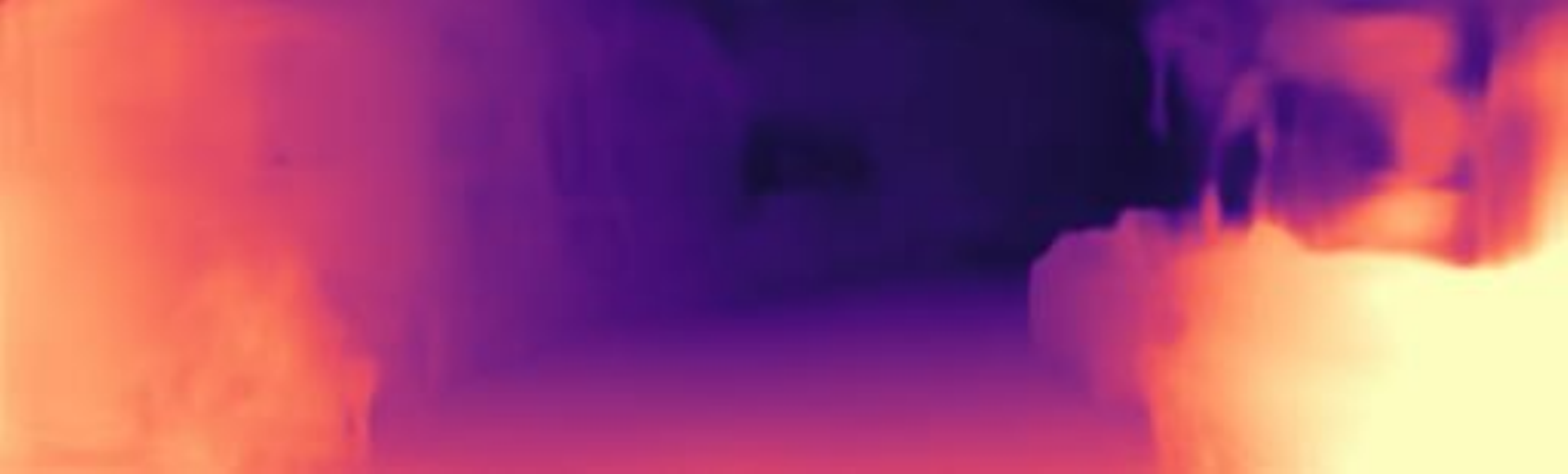}&  
 \includegraphics[]{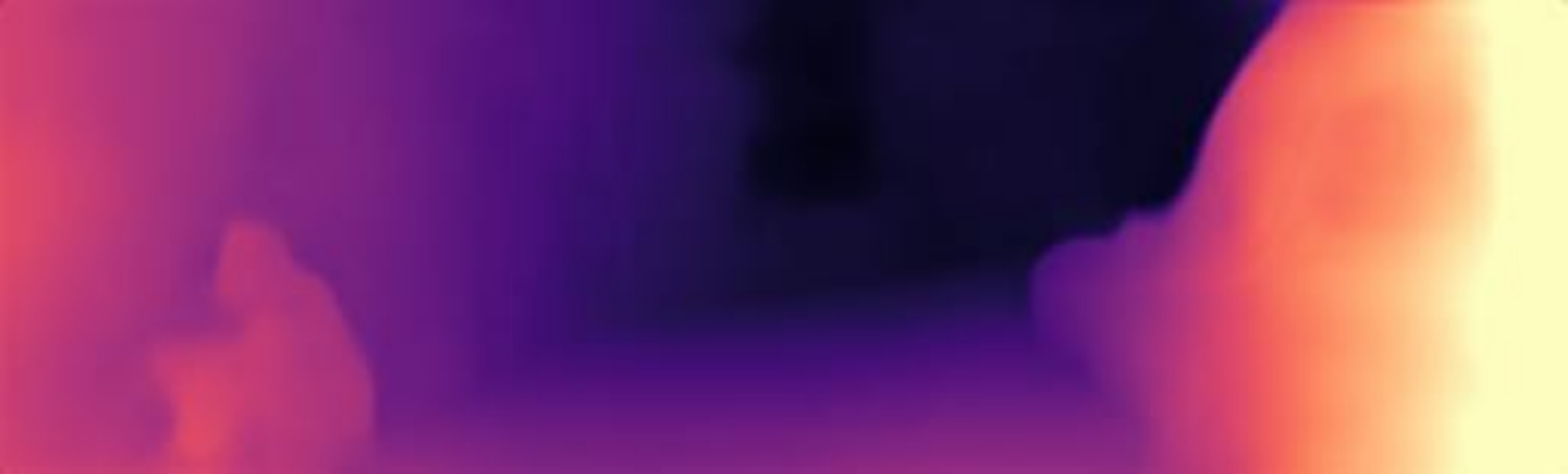}&  
 \includegraphics[]{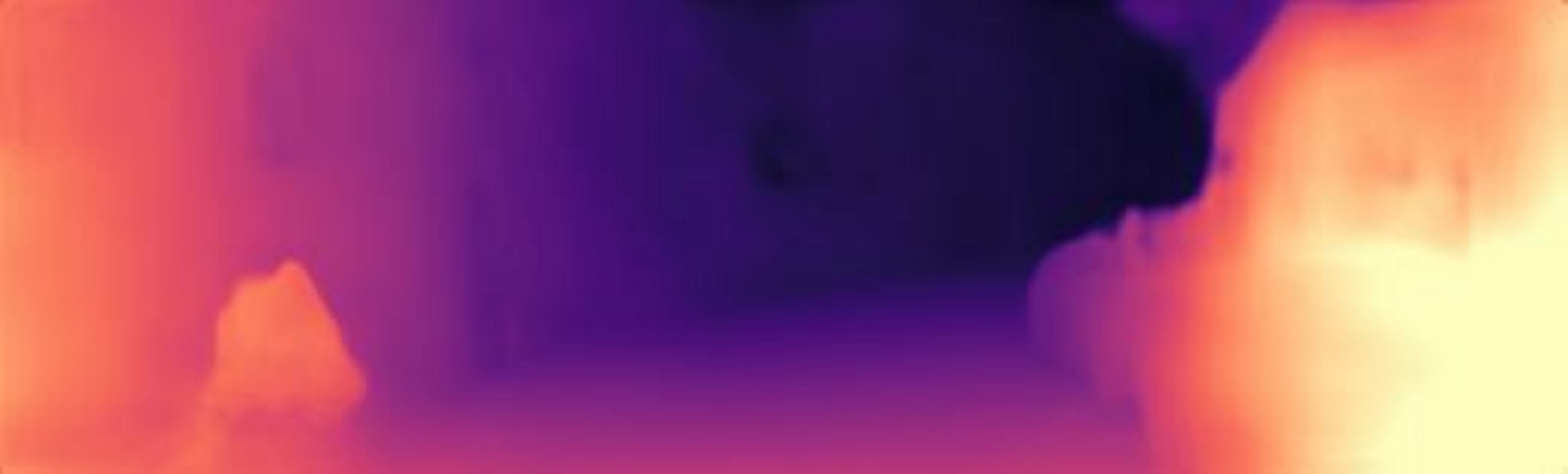}&  
 \includegraphics[]{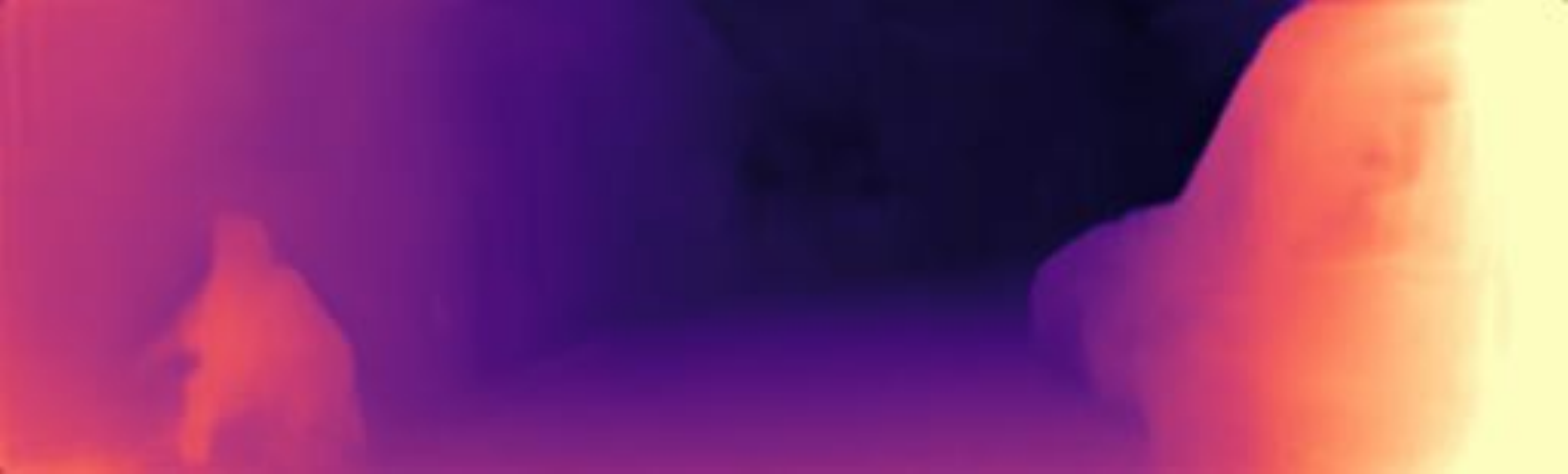}&
 \includegraphics[]{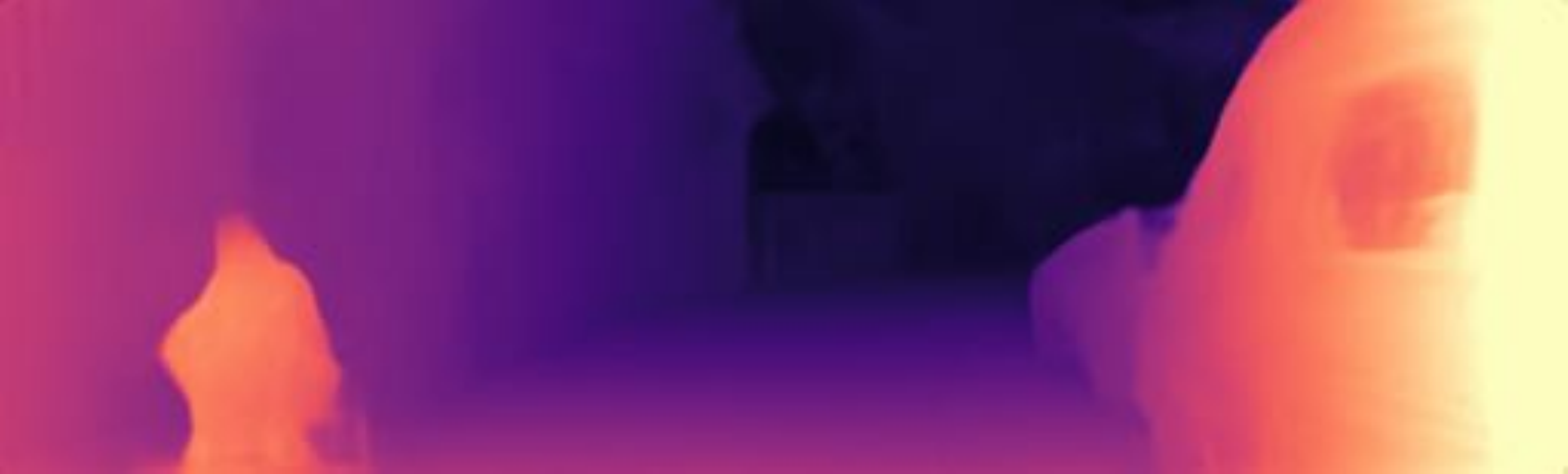}\\
 \fontsize{\w}{\h} \selectfont MF-SLaK&   
 \fontsize{\w}{\h} \selectfont MF-ViT &  
 \fontsize{\w}{\h} \selectfont MF-RegionViT&  
  \fontsize{\w}{\h} \selectfont MF-Twins&
 \fontsize{\w}{\h} \selectfont MF-Ours  \\
 \includegraphics[]{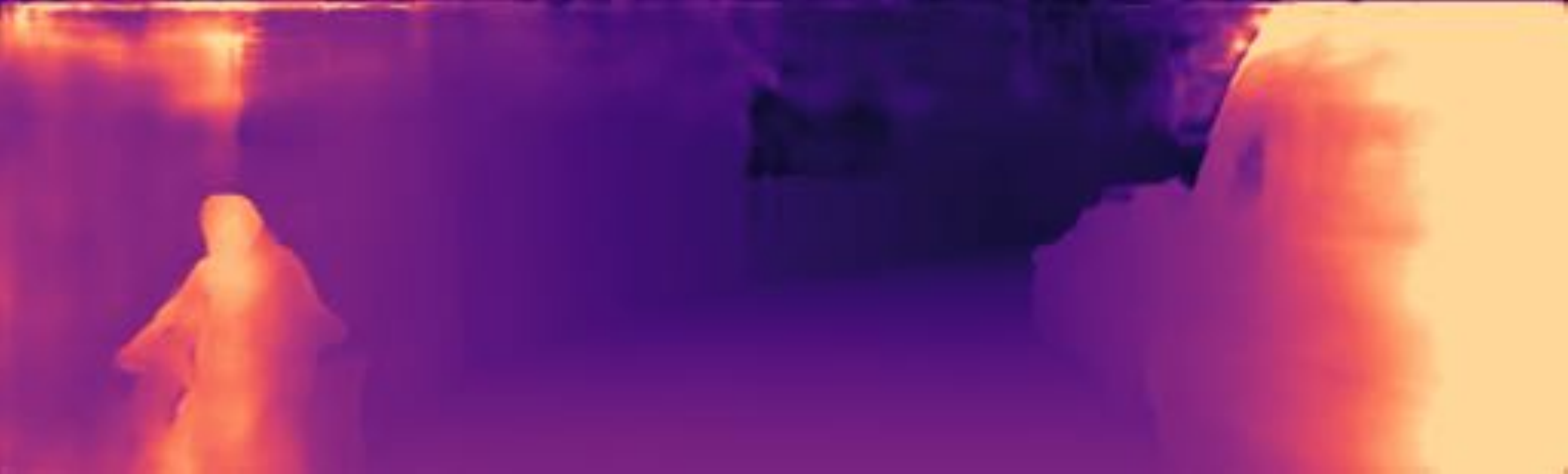}&  
 \includegraphics[]{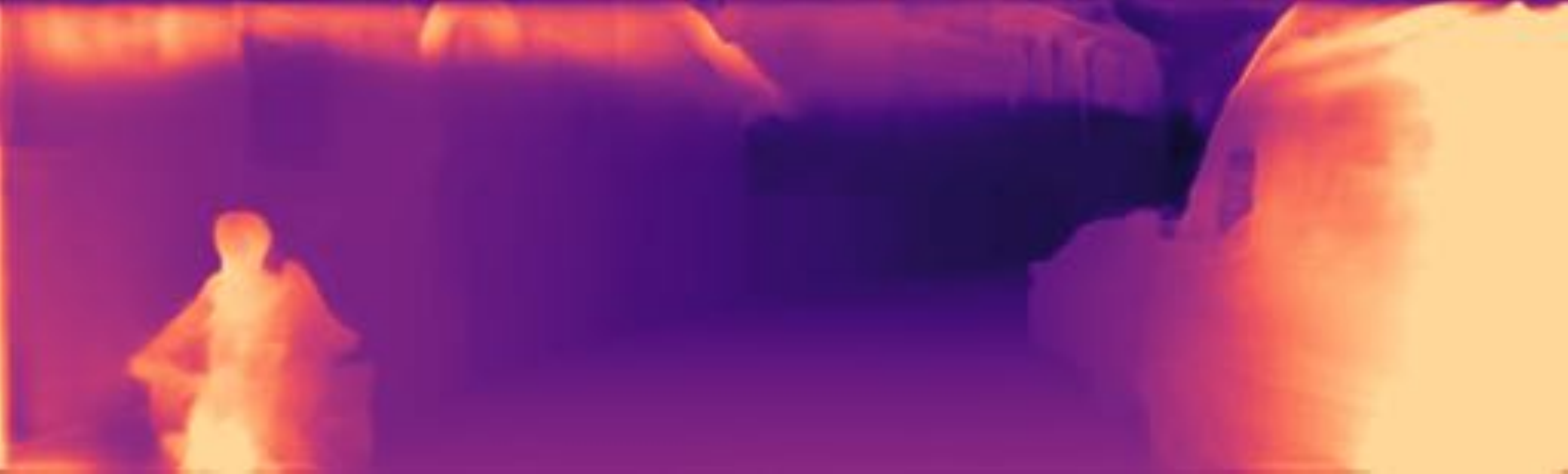}&    
 \includegraphics[]{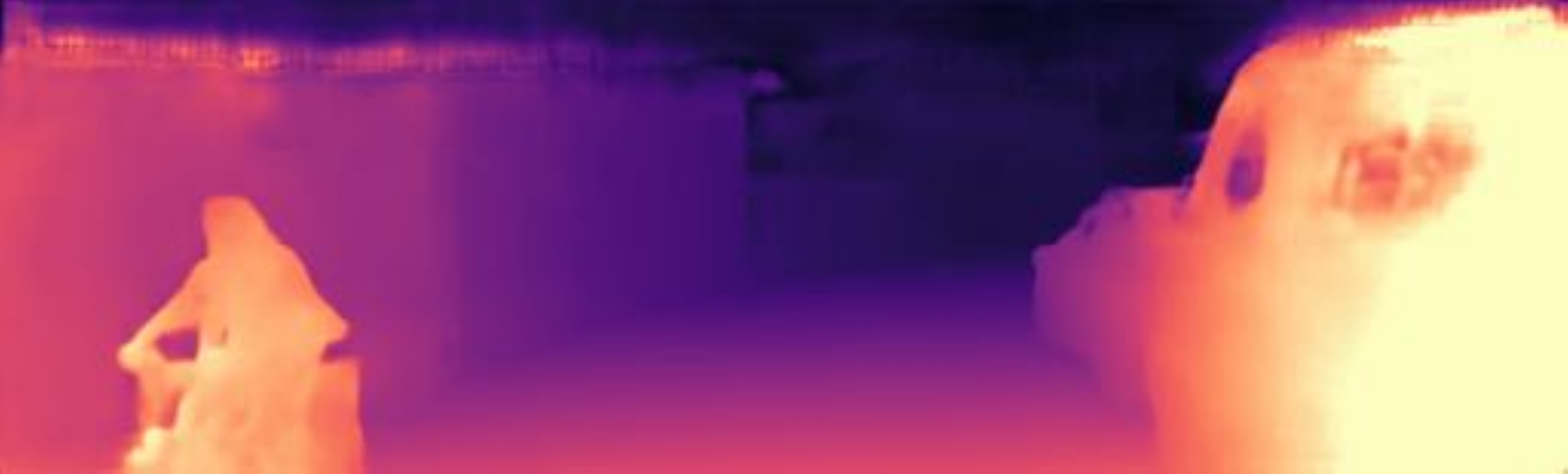}& 
 \includegraphics[]{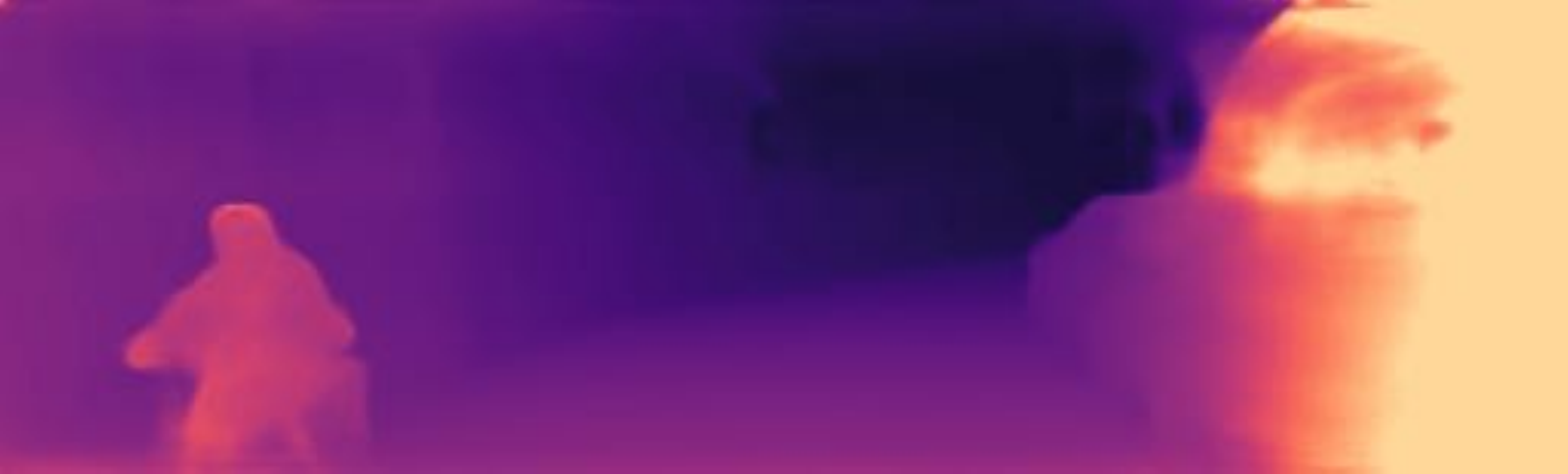}&    
 \includegraphics[]{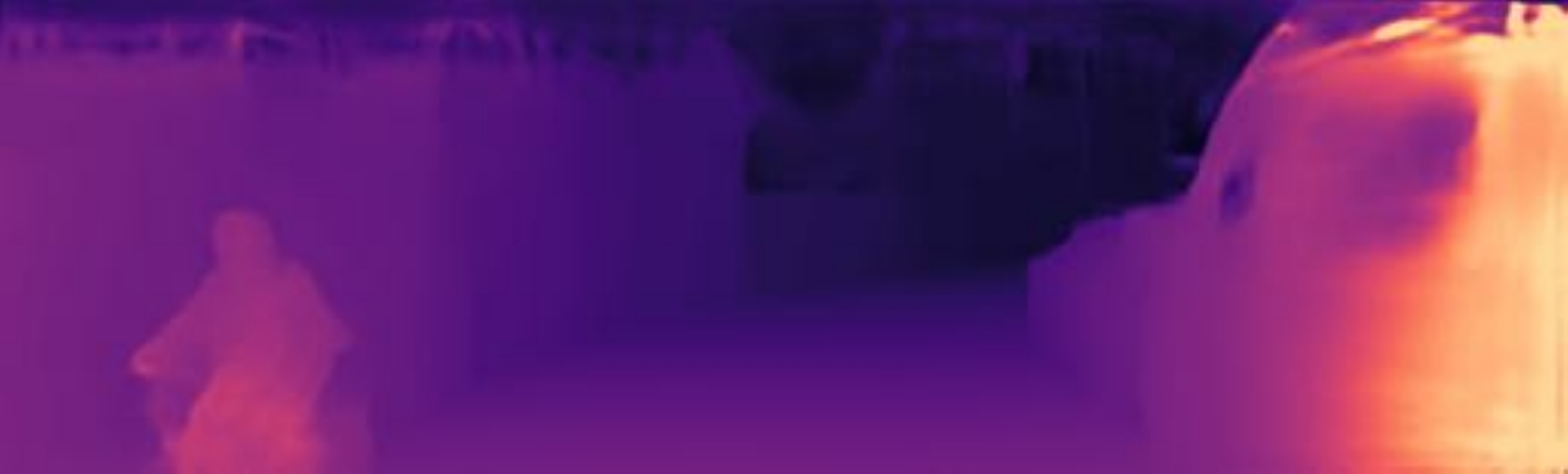}\\
 \fontsize{\w}{\h} \selectfont BTS &  
 \fontsize{\w}{\h} \selectfont AdaBins&  
 \fontsize{\w}{\h} \selectfont TransDepth&  
 \fontsize{\w}{\h} \selectfont DepthFormer&   
  \fontsize{\w}{\h} \selectfont GLPDepth 
\end{tabular}%
}
\end{subfigure}
 \vspace{-2mm}
\begin{subfigure}
\centering
\resizebox{\linewidth}{!}{%
\begin{tabular}{ccccc}
 \includegraphics[width=\ww,height=\wh]{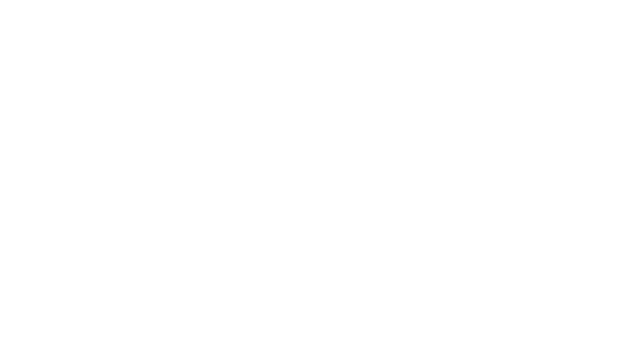}&
 \includegraphics[width=\ww,height=\wh]{figure/white.PNG}&
 \includegraphics[width=\ww,height=\wh]{figure/white.PNG}&
 \includegraphics[width=\ww,height=\wh]{figure/white.PNG}&
 \includegraphics[width=\ww,height=\wh]{figure/white.PNG}
\end{tabular}%
}
\end{subfigure}
\begin{subfigure}
\centering
\resizebox{\textwidth}{!}{%
\begin{tabular}{ccccc}
 \includegraphics[]{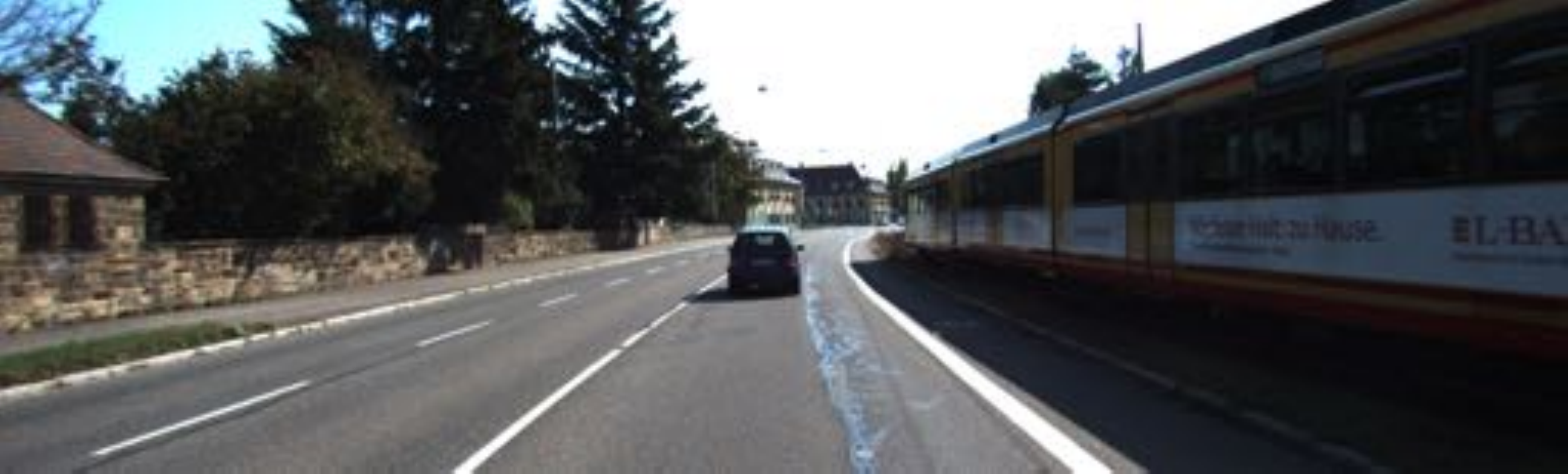}& 
 \includegraphics[]{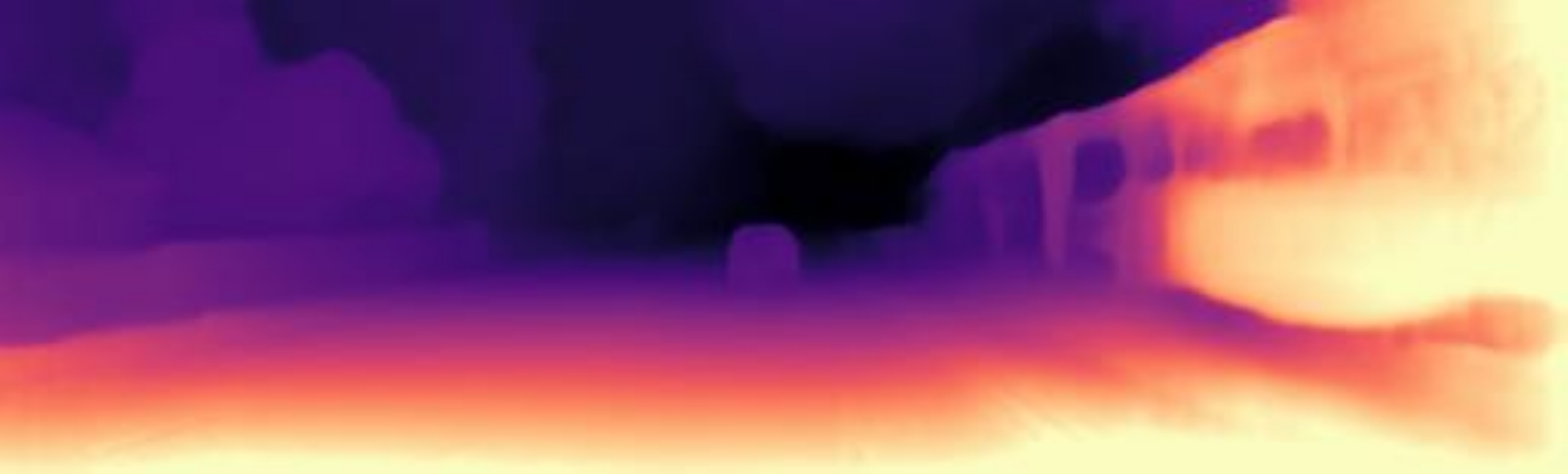}&  
 \includegraphics[]{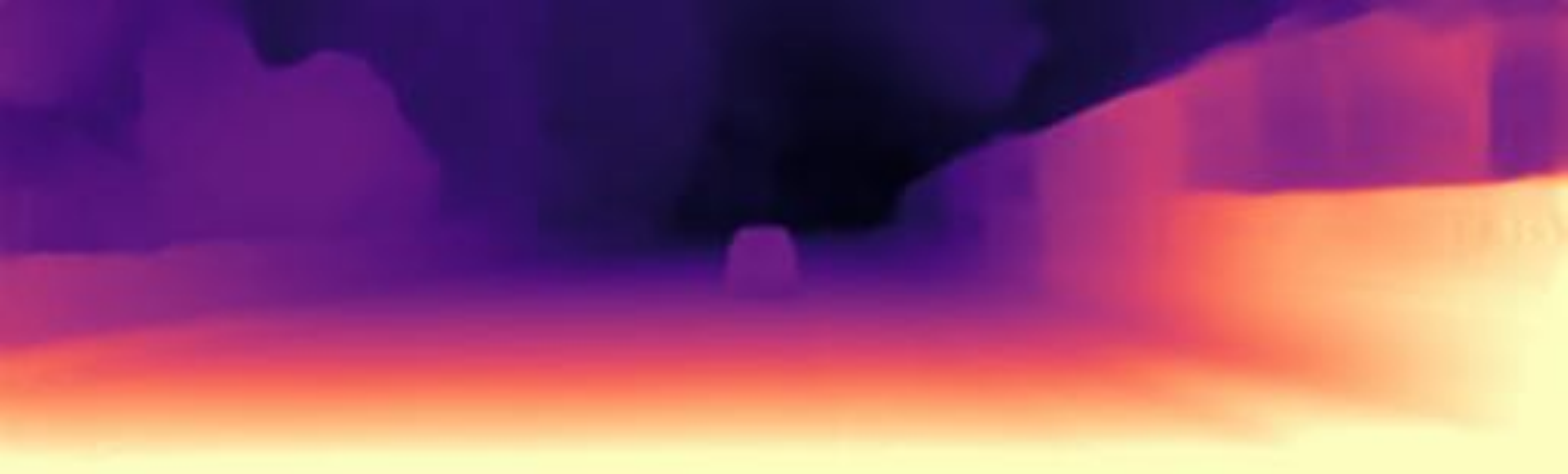}&  
 \includegraphics[]{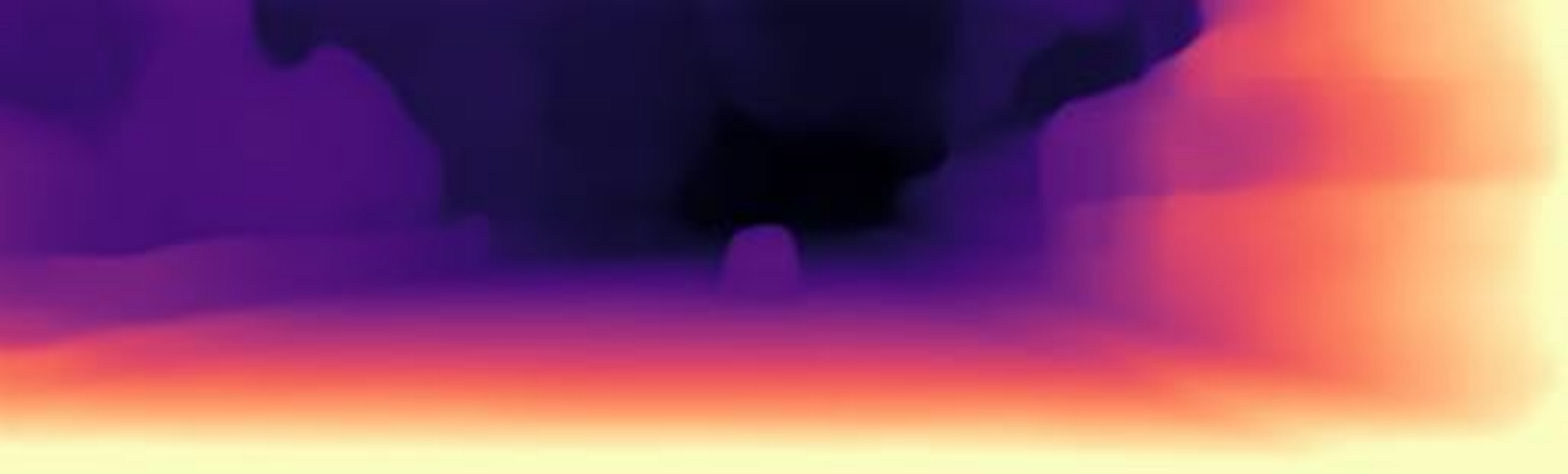}&
 \includegraphics[]{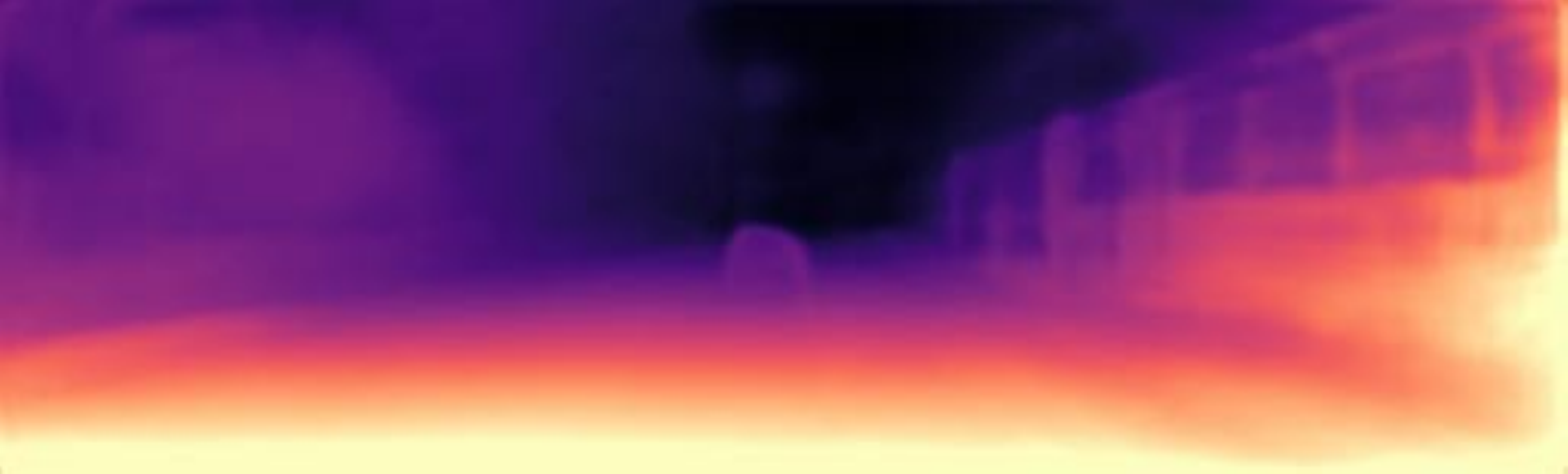}\\
 \fontsize{\w}{\h} \selectfont Input images&   
 \fontsize{\w}{\h} \selectfont Monodepth2&  
 \fontsize{\w}{\h} \selectfont PackNet-SfM&  
 \fontsize{\w}{\h} \selectfont R-MSFM6&
 \fontsize{\w}{\h} \selectfont MF-ConvNeXt\\
 \includegraphics[]{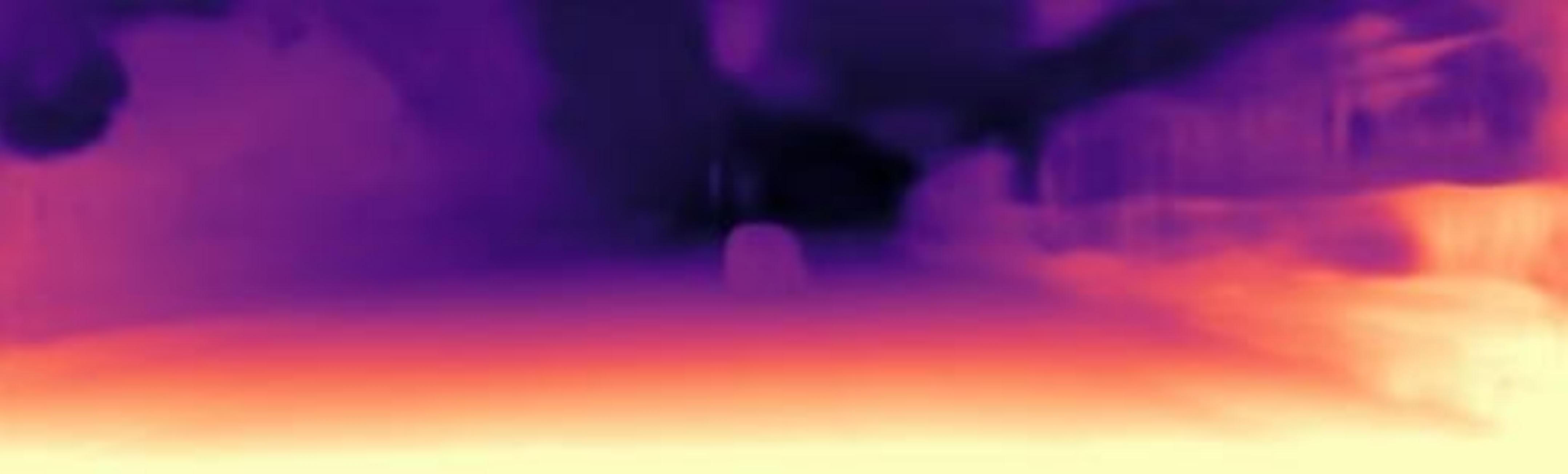}&  
 \includegraphics[]{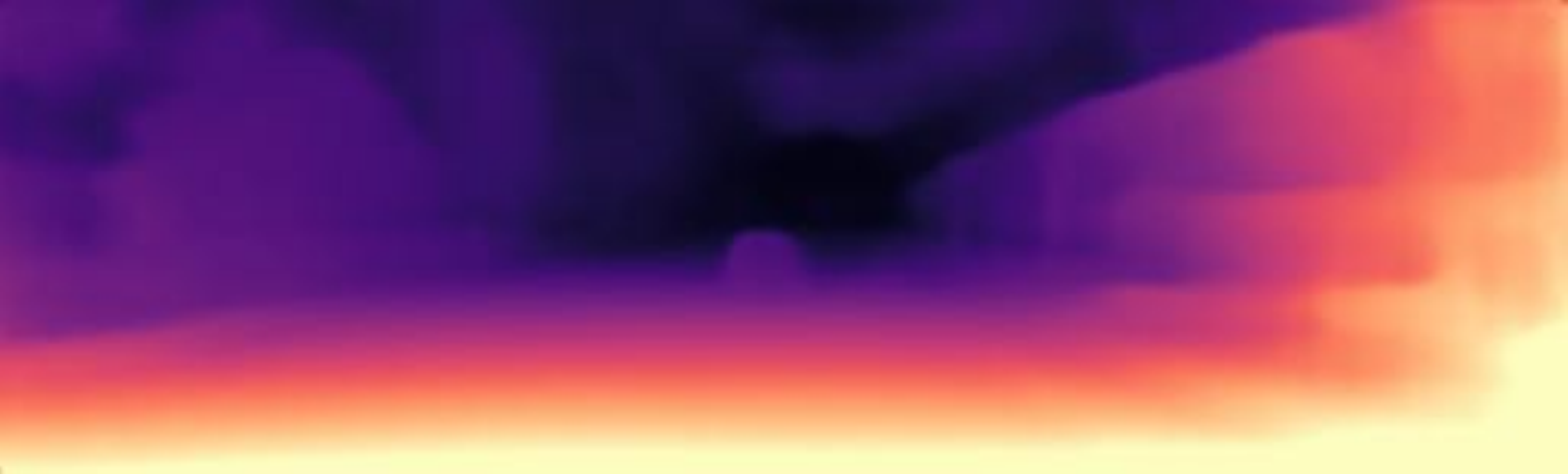}&  
 \includegraphics[]{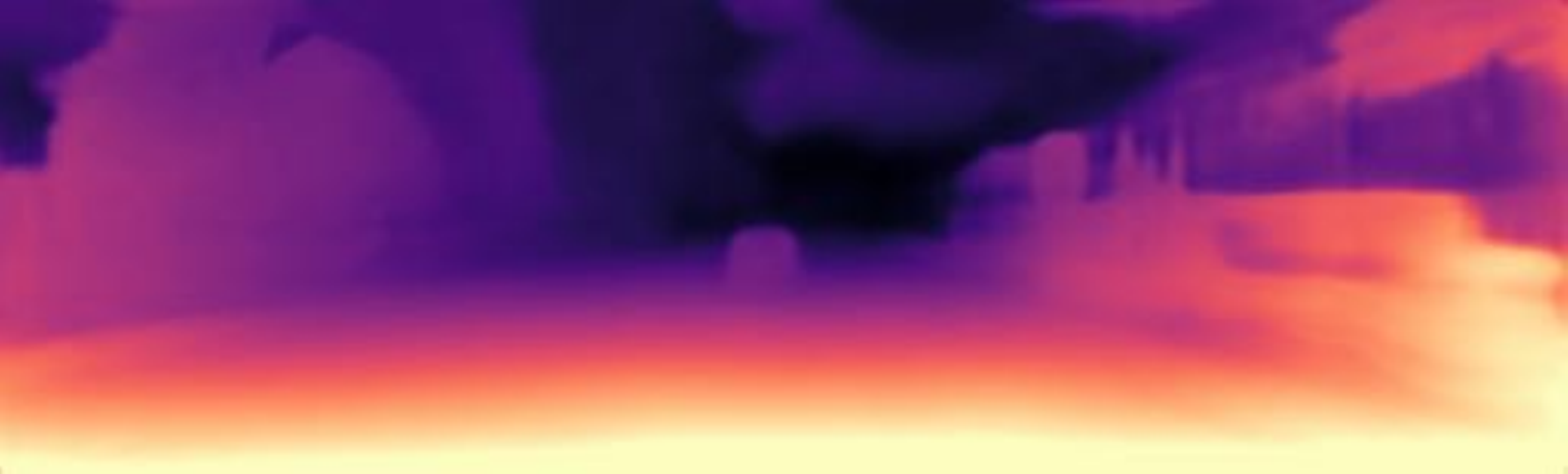}&  
 \includegraphics[]{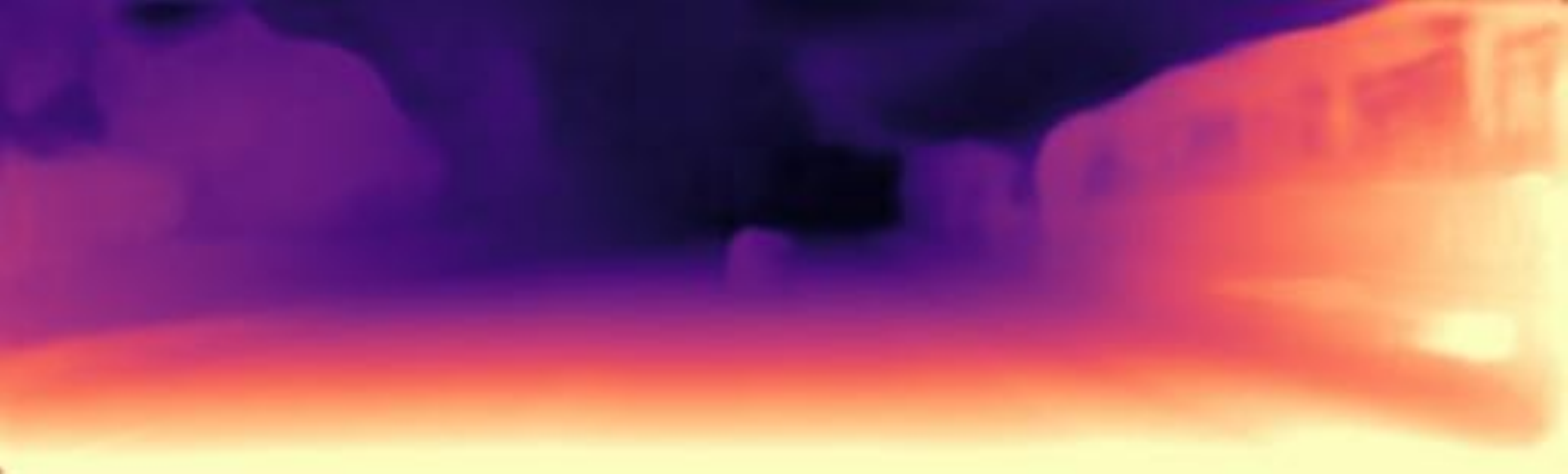}&
 \includegraphics[]{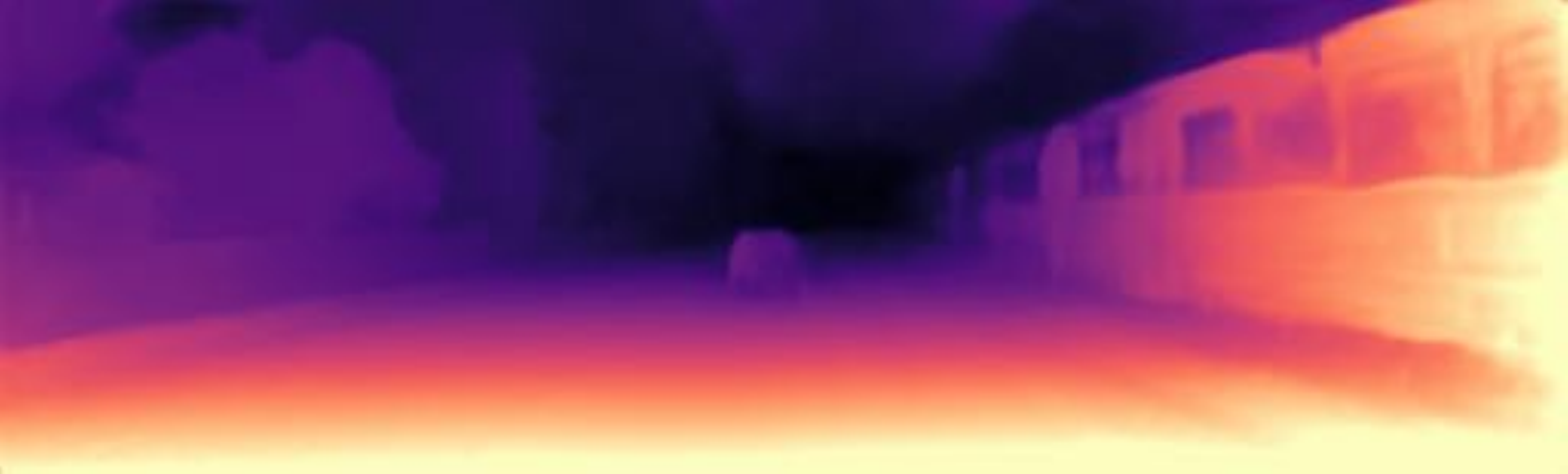}  \\
 \fontsize{\w}{\h} \selectfont MF-SLaK&   
 \fontsize{\w}{\h} \selectfont MF-ViT &  
 \fontsize{\w}{\h} \selectfont MF-RegionViT&  
  \fontsize{\w}{\h} \selectfont MF-Twins&
 \fontsize{\w}{\h} \selectfont MF-Ours  \\
 \includegraphics[]{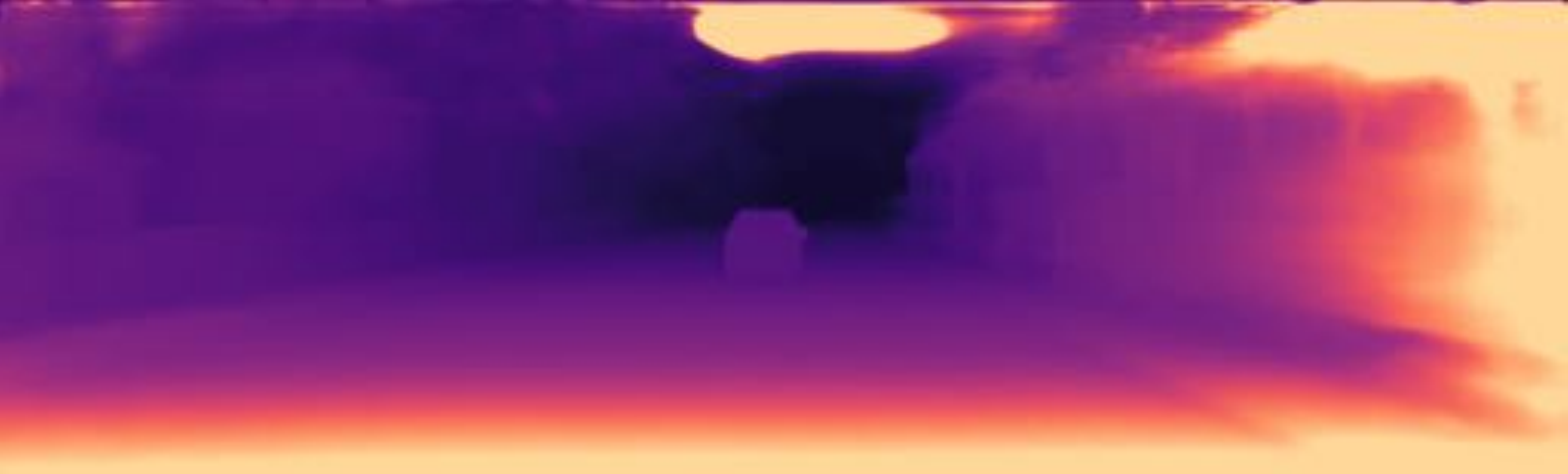}&  
 \includegraphics[]{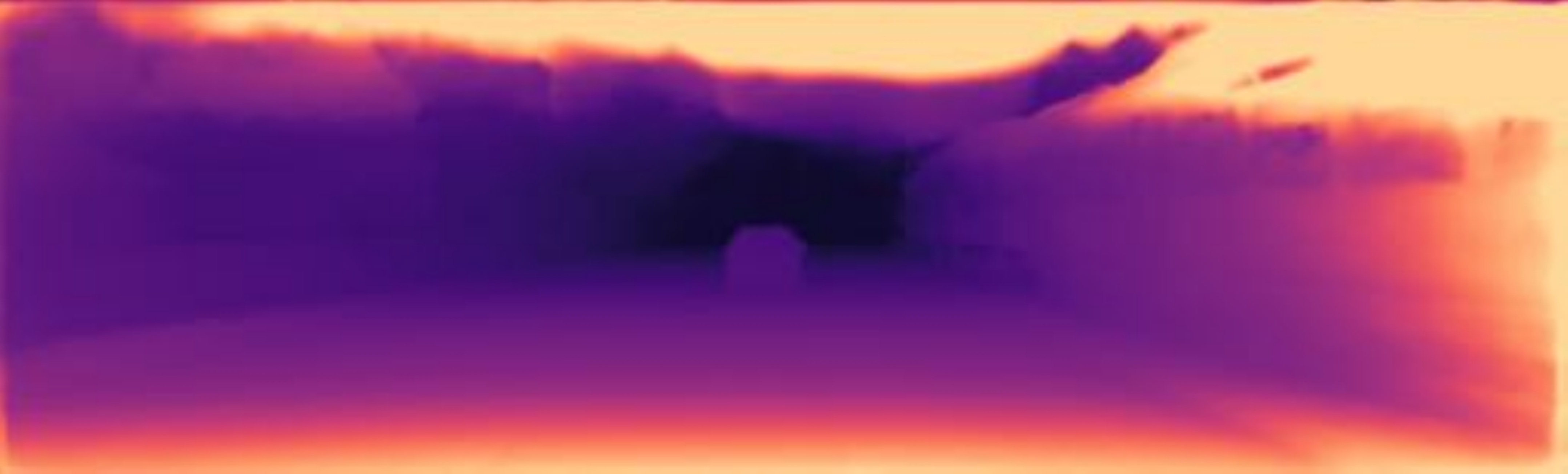}&    
 \includegraphics[]{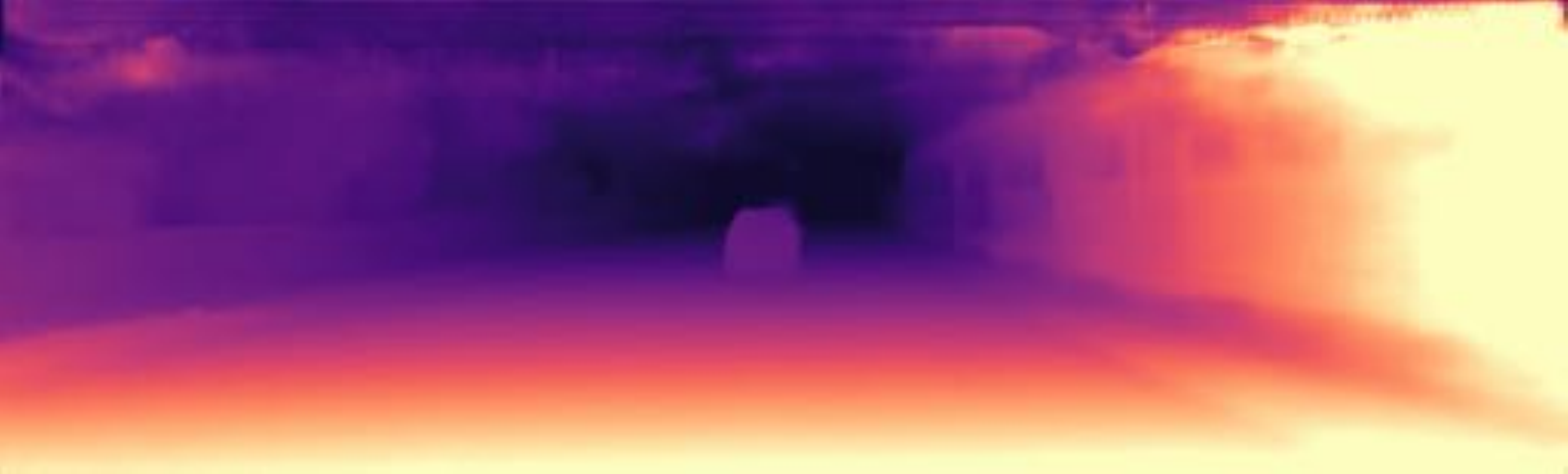}&  
 \includegraphics[]{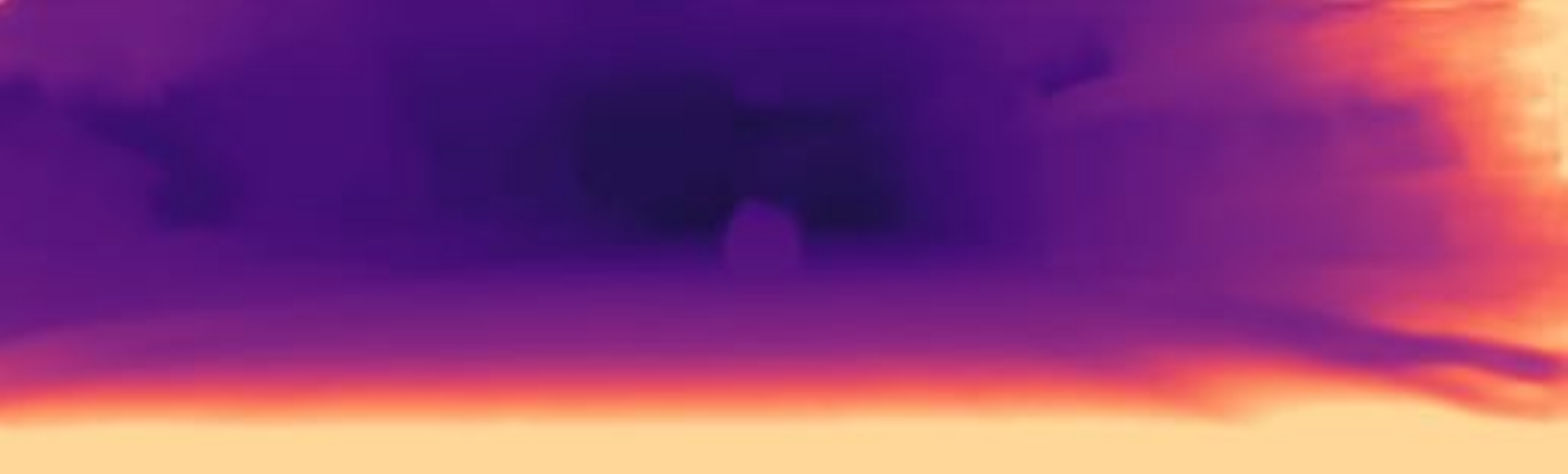}&    
 \includegraphics[]{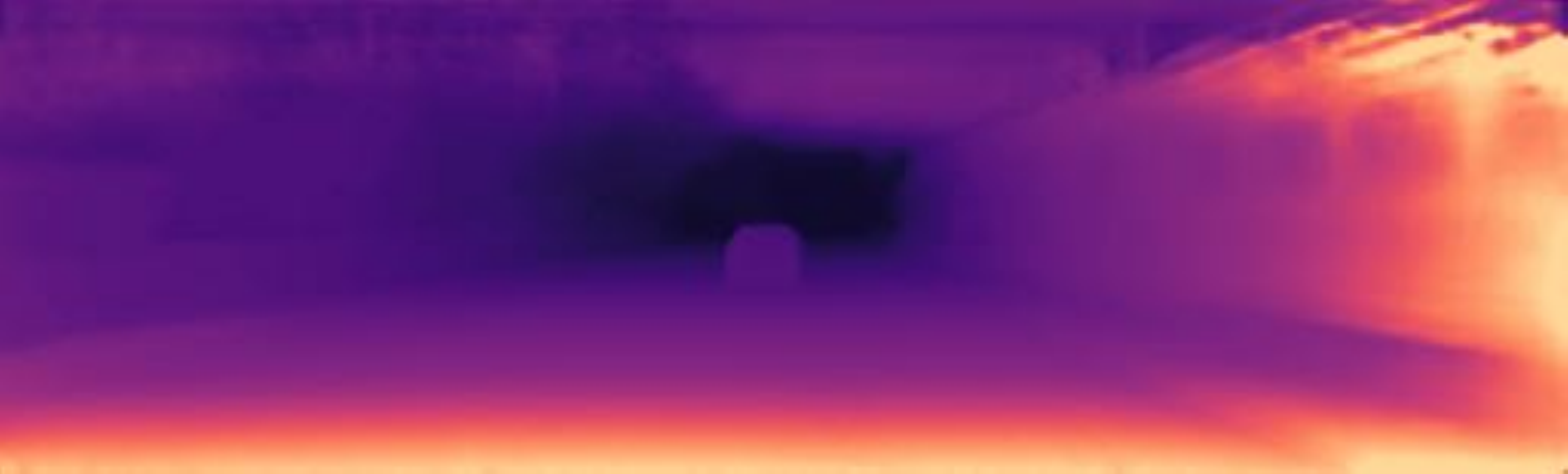}\\
 \fontsize{\w}{\h} \selectfont BTS &  
 \fontsize{\w}{\h} \selectfont AdaBins&  
 \fontsize{\w}{\h} \selectfont TransDepth&  
 \fontsize{\w}{\h} \selectfont DepthFormer&   
  \fontsize{\w}{\h} \selectfont GLPDepth 
\end{tabular}%
}
\end{subfigure}
 \vspace{-2mm}
\caption{\textbf{Depth map results on the in-distribution dataset.} MonoFormer (MF-Ours) is from our previous work~\cite{bae2022deep}.}
 \vspace{-2mm}
\label{figure_result_kitti}
\end{figure*}

\section{EXPERIMENTAL RESULTS}
\externaldocument{7_supp.tex}

\vspace{-0.25mm}
We conduct extensive experiments to investigate the generalization performance of various network structures and the effect of texture-/shape-bias for monocular depth estimation.
First, we evaluate state-of-the-art KITTI-trained models, including various modern backbone architectures on the KITTI dataset and various depth benchmark datasets. 
Evaluations are conducted on the KITTI, an in-distribution dataset in \secref{compriason_kitti}, and the other depth datasets, out-of-distribution datasets in \secref{genealization-exp}.
We also conduct an ablation study on the MonoFormer in \secref{ablation}.
Then, we investigate the texture-bias and shape-bias of the MDE models using both the self-supervised and supervised methods with texture-shifted datasets in \secref{texture-shifted-exp}.
We analyze the depth estimation performance and the feature representation of each model on the texture-shifted datasets, which are synthetically generated.
Lastly, we demonstrate the intrinsic properties of the CNN-based and Transformer-based networks are observed in the real-world texture-shifted datasets captured from different driving environments, such as the weather changes and illumination changes in \secref{practical-texture-exp}. \\
To provide clearly and highly visible quantitative evaluation, we utilizes absolute relative error (Abs Rel) and the inlier ratio (Accuracy metric 1, $\delta < 1.25$) metric to provide figures that can represent the trend of the whole table at main manuscript. Additionally, we provide figures and tables for all evaluation metrics in the appendix.
\begin{table}[t]
\resizebox{\columnwidth}{!}{
\Large
\begin{tabular}{llll}  
 \hline
 Models & Supervision & & Methods \\ \hline
\textit{CNN-based} & Self-supervised & & Monodepth2~\cite{godard2019digging}, PackNet-SfM~\cite{guizilini20203d}, R-MSFM6~\cite{zhou2021r}, DIFFNet~\cite{zhou2021diffnet},\\ 
 & & &  MF-(ConvNeXt~\cite{liu2022convnet}, SLaK~\cite{liu2022more})\\ \cline{2-4}
& Supervised & & BTS~\cite{lee2019big}, AdaBins~\cite{bhat2021adabins} \\ \hline
\textit{Transformer-based} & Self-supervised & & MF-(ViT~\cite{dosovitskiy2020image}, RegionViT~\cite{chen2021regionvit}, Twins~\cite{chu2021twins}, Ours~\cite{bae2022deep}), MonoViT~\cite{Zhao2022monovit} \\  \cline{2-4}
& Supervised & &  TransDepth~\cite{yang2021transformer}, DepthFormer~\cite{li2022depthformer}, GLPDepth~\cite{kim2022global} \\  \hline
\end{tabular}}
\vspace{-2mm}
\caption{\textbf{Taxonomy of MDE methods w.r.t backbones.} }
\vspace{-3.5mm}
\label{taxonomy}
\end{table}

\vspace{-1.5mm}

\begin{table}[t!]
\centering
\resizebox{\columnwidth}{!}{%
\begin{tabular}{cccccccccc}
\hline \hline
\multirow{2}{*}{Method} & \multirow{2}{*}{Train} & \multirow{2}{*}{Backbone} & \multicolumn{4}{c}{Lower is better $\downarrow$}                               & \multicolumn{3}{c}{Higher is better $\uparrow$}                               \\ \cline{4-10} 
                       &                           &                          &
                       Abs Rel        & Sq Rel         & RMSE           & RMSElog        & $\delta <1.25$  & $\delta < 1.25^2$ & $\delta < 1.25^3$ \\ \hline
Monodepth2      & M      & CNN                       & 0.115          & 0.903          & 4.863          & 0.193          & 0.877          & 0.959                   & 0.981                   \\
PackNet-SfM         & M  & CNN                       & 0.111          & 0.785 & 4.601          & 0.189          & 0.878          & 0.960                   & \textbf{0.982}          \\
R-MSFM6             & M   & CNN                       & 0.112          & 0.806          & 4.704          & 0.191          & 0.878          & 0.960                   & 0.981                   \\ 

MF-ConvNeXt & M& CNN &  0.111   & \textbf{0.760}   & \textbf{4.533}   &  0.184   &  0.878   &  0.961   & \textbf{0.982}   \\
MF-SLaK & M& CNN &  0.117   &  0.866   &  4.811   &  0.199   &  0.866   &  0.947   &  0.976   \\
MF-ViT      & M & Trans               & 0.118          & 0.942          & 4.840          & 0.193          & 0.873          & 0.956                   & 0.981                   \\
MF-RegionViT & M& Trans &  0.113   &  0.893   &  4.756   &  0.193   &  0.875   &  0.954   &  0.979   \\
MF-Twins & M& Trans &  0.125   &  1.309   &  4.973   &  0.197   &  0.866   &  0.948   &  0.974   \\ 
MF-Ours    & M & Trans               & \textbf{0.104} & 0.846          & 4.580 & \textbf{0.183} & \textbf{0.891} & \textbf{0.962}          & \textbf{0.982}          \\ \hline \hline
BTS          & D    & CNN                       & 0.061          & 0.250          & 2.803          & 0.098          & 0.954          & 0.992                   & 0.998                   \\
AdaBins         & D    & CNN                       & 0.058          & 0.190          & 2.360          & 0.088          & 0.965          & 0.995                   & \textbf{0.999}                   \\ 
TransDepth         & D   & Trans                    & 0.064          & 0.252 & 2.755          & 0.098          & 0.956          & 0.994                   & \textbf{0.999}          \\
DepthFormer          & D      & Trans                       & \textbf{0.052}          & \textbf{0.156}          & \textbf{2.133}          & \textbf{0.079}          & \textbf{0.974}          & \textbf{0.997}                   & \textbf{0.999}                   \\ 
GLPDepth              & D  & Trans                       & 0.059          & 0.187          & \textbf{2.133}          & \textbf{0.079}          & \textbf{0.974}          & \textbf{0.997}                   & \textbf{0.999}                   \\ \hline \hline
\end{tabular}%
}
\vspace{-2mm}
\caption{\textbf{Quantitative results on the in-distribution dataset.} M is Monocular images, and D is GT depth. Trans means Transformer. \textbf{Bold} is the best performance.}
\label{table_resutl_kitti}
\vspace{-2mm}
\end{table}

\begin{table}[t!]
\centering
\resizebox{\columnwidth}{!}{%
\begin{tabular}{cccccccccc}
\hline \hline
\multirow{2}{*}{Method} & \multirow{2}{*}{Train} & \multirow{2}{*}{Backbone} & \multicolumn{4}{c}{Lower is better $\downarrow$}                               & \multicolumn{3}{c}{Higher is better $\uparrow$}                               \\ \cline{4-10} 
                       &                           &                          &
                       Abs Rel        & Sq Rel         & RMSE           & RMSElog        & $\delta <1.25$  & $\delta < 1.25^2$ & $\delta < 1.25^3$ \\ \hline
Monodepth2      & M & CNN   & 0.090 & 0.545 & 3.942 & 0.137 & 0.914 & 0.983 & 0.995 \\
PackNet-SfM     & M & CNN   & 0.078 & 0.420 & 3.485 & 0.121 & 0.931 & 0.986 & 0.996 \\
DIFFNet         & M & CNN   & 0.076 & 0.412 & 3.494 & 0.119 & 0.935 & 0.988 & 0.996 \\ 
MF-ConvNeXt     & M & CNN   & 0.079	& 0.411	& 3.483	& 0.121	& 0.929	& 0.987	& \textbf{0.997} \\
MonoViT         & M & Trans & 0.075 & \textbf{0.389} & \textbf{3.419} & 0.115 & 0.938 & \textbf{0.989} & \textbf{0.997} \\
MF-ViT          & M & Trans & 0.086	& 0.520	& 3.818	& 0.131	& 0.919	& 0.983	& 0.995 \\
MF-Ours         & M & Trans & \textbf{0.071}	& 0.436	& 3.457	& \textbf{0.113}	& \textbf{0.939}	& 0.987	& 0.996 \\ \hline \hline
\end{tabular}%
}
\vspace{-2mm}
\caption{\textbf{Quantitative results on the KITTI Depth Prediction Evaluation dataset~\cite{uhrig2017improvedkitti} (in-distribution dataset).} M is Monocular images. Trans means Transformer. \textbf{Bold} is the best performance.}
\label{table_resutl_improved_kitti}
\vspace{-4mm}
\end{table}

\subsection{Competitive methods and evaluation setups}
\label{setup}
We conduct extensive experiments to compare the performances of the monocular depth estimation methods.
For self-supervised setting, we compare our work, called MF-Ours~\cite{bae2022deep}, with state-of-the-art methods, Monodepth2 {\cite{godard2019digging}}, PackNet-SfM {\cite{guizilini20203d}}, R-MSFM {\cite{zhou2021r}}, DIFFNet{\cite{zhou2021diffnet}} and MonoViT{\cite{Zhao2022monovit}}. 
We also evaluate state-of-the-art supervised methods, BTS~\cite{lee2019big}, AdaBins~\cite{bhat2021adabins}, TransDepth~\cite{yang2021transformer}, DepthFormer~\cite{li2022depthformer}, and GLPDepth~\cite{kim2022global}.
We note that some Transformer-based models, such as MF-Ours, TransDepth, and DepthFormer, use both CNNs and Transformers for their encoders.
We also analyze the performance of various modern architectures such as RegionViT~\cite{chen2021regionvit}, Twins~\cite{chu2021twins}, ConvNeXt~\cite{liu2022convnet} and SLaK~\cite{liu2022more}.
We replace the encoder of MF-Ours with these modern backbones, whose names are defined as MF-(backbone).
Its taxonomy is described in \tabref{taxonomy}.
We illustrate these basic network structures of the modern CNN and Transformer architectures in \figref{figure_modern_summary}.
Transformer-based models \cite{chen2021regionvit,chu2021twins} add the locality by using local attention to compensate for the shortcomings of the ViT (\eg, necessity of large dataset). 
CNN-based architectures \cite{liu2022convnet,liu2022more} aim to extract global information while utilizing the intrinsic locality inductive bias. They also imitate the self-attention of the Transformer using improved strategies such as large kernel size, patchify stem \cite{liu2022convnet}, and Layer Norm \cite{ba2016layer}.

We train all networks using the KITTI Eigen split \cite{geiger2013vision,eigen2015predicting} consisting of 39,810 training and 4,424 validation data whose size is $640 \times 192$. All experiments in this paper are conducted with this KITTI-trained model whose results are reported in \Cref{compriason_kitti,ablation,genealization-exp,texture-shifted-exp,practical-texture-exp}.
Following a previous work~\cite{guizilini20203d}, we remove around 5\% of the total data for training to address the infinite-depth problems that commonly occur in dynamic scenes. We use typical error and accuracy metrics for depth, absolute relative (Abs Rel), square relative (Sq Rel), root-mean-square-error (RMSE), its log (RMSElog), and the ratio of inliers following the work \cite{guizilini20203d}.
\subsection{Evaluation on the in-distribution dataset}
\label{compriason_kitti}
We conduct the evaluations with an in-distribution dataset, where the KITTI-trained models are evaluated on 697 KITTI test data. 
The qualitative results in \figref{figure_result_kitti} show that our self-supervised method precisely preserves object boundaries. This indicates that the encoder captures global context and informative local features and transfers them to the decoder for pixel-wise prediction. 
The quantitative results in \tabref{table_resutl_kitti} show that MF-Ours outperforms all competitive methods.
We also evaluate the performance of models with the other backbones, MF-(ConvNeXt/SLaK/RegionViT/Twins/ViT). 
We employ only the encoder part of our method as a backbone without changing other parts in this experiment.
Supervised methods outperform all self-supervised methods regardless of backbones.
MF-Ours and DepthFormer achieve the best performance among the self-supervised methods and supervised methods on the in-distribution datasets, respectively.
Overall, the performance gap among all the competitive methods is marginal for the in-distribution dataset, but the results in ~\tabref{table_resutl_improved_kitti}, a quantitative evaluation of the improved KITTI, also show that the Transformer-based models generally outperform the CNN-based models.

\begin{table}[t]
    \centering
    \resizebox{\columnwidth}{!}{
    \begin{tabular}{llllll}
    \hline
    & Abs Rel $\downarrow$ & Sq Rel $\downarrow$ & RMSE$\downarrow$ & RMSE$_{log}$ $\downarrow$ & $\delta <1.25 $ $\uparrow$ \\ \hline
    baseline     &   0.113 & 0.899  & 4.783& 0.189 & 0.882      \\
    +ACM         &   0.113 & 0.879  & 4.820& 0.189 & 0.879      \\
    +FFD         &   0.112 & 0.860  & 4.803& 0.186 & 0.879       \\
    +ACM\&FFD &   \textbf{0.104}  & \textbf{0.846} & \textbf{4.580} & \textbf{0.183} & \textbf{0.891}       \\ \hline
    \end{tabular}}
    \vspace{-2mm}
    \caption{\textbf{Ablation study on ACM and FFD.}}
    \label{table_result_modules}
    \vspace{-3mm}
\end{table}

\begin{figure}[t] 
    \newcommand\w{4cm}
    \centering
    \begin{tabular}{c@{\hspace{1mm}}c@{\hspace{1mm}}}
    \includegraphics[width=\w]{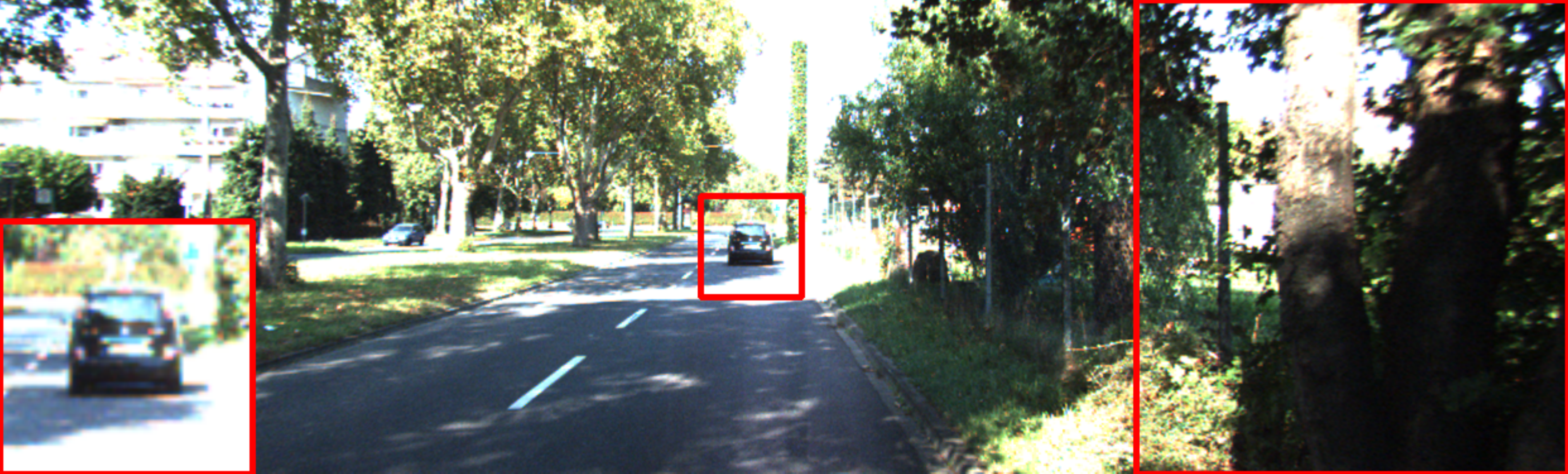} &
    \includegraphics[width=\w]{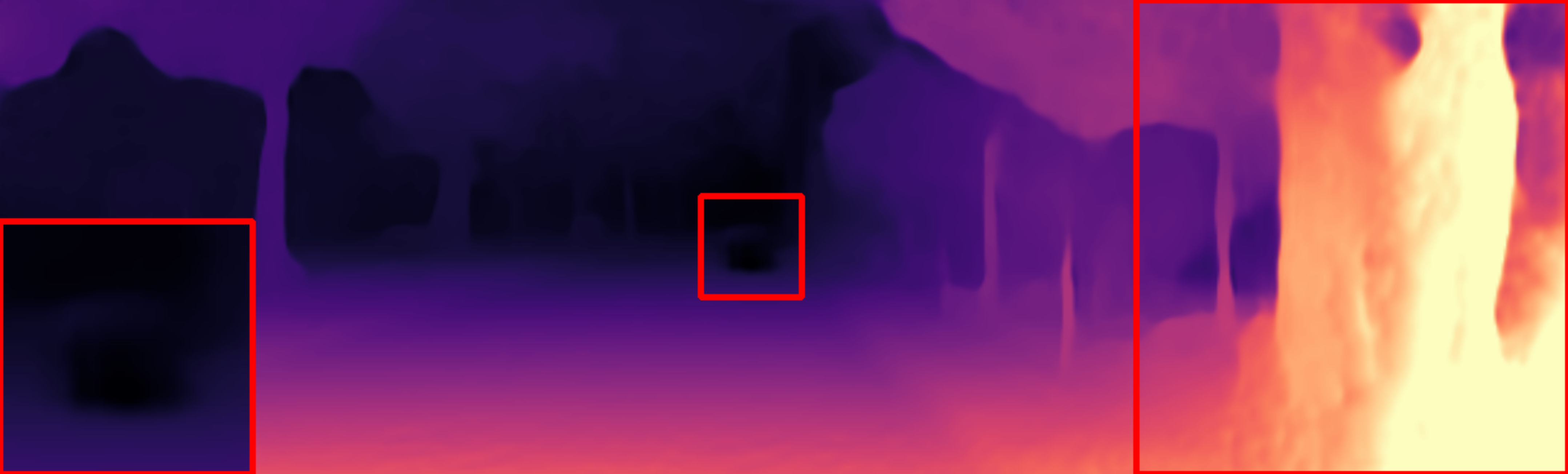}  \\
    \small{(a) Input image}  &
    \small{(b) Results with ACM}  \\
    \includegraphics[width=\w]{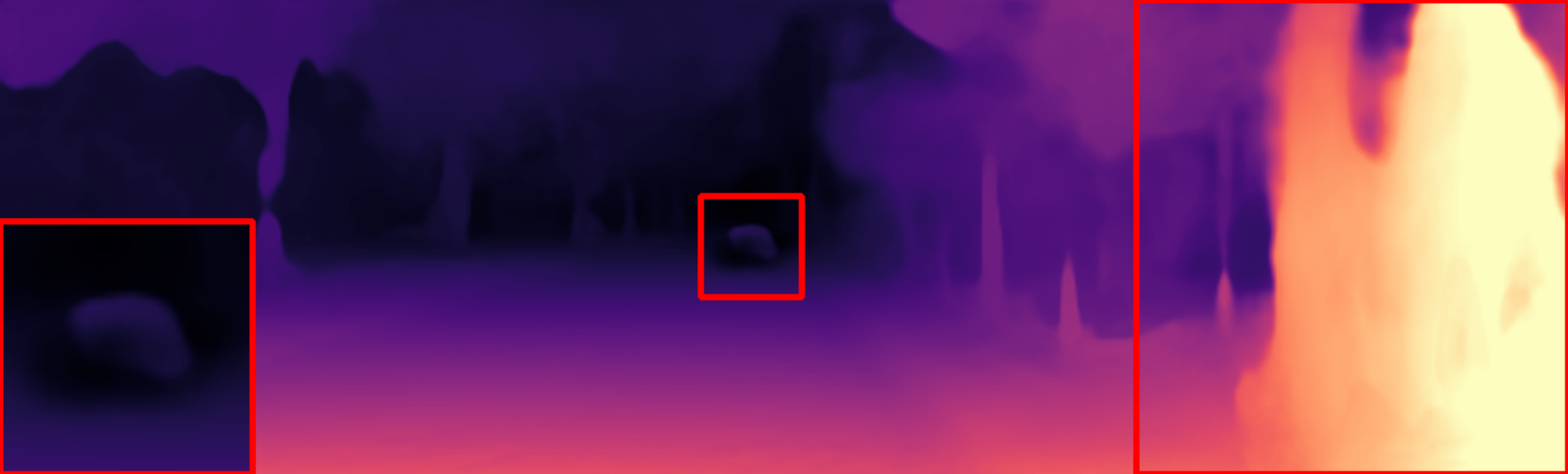}  & \includegraphics[width=\w]{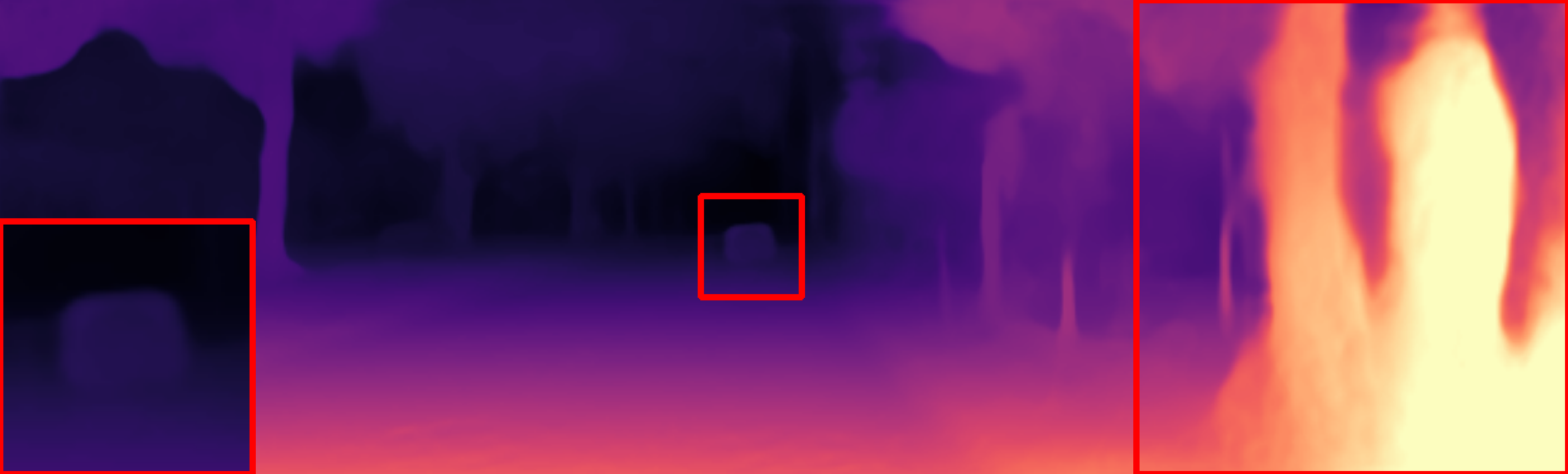}  \\
    \small{(c) Results with FFD}  &
    \small{(d) Results with ACM and FFD}
    \end{tabular}
    \vspace{-2mm}
    \captionof{figure}{\textbf{Results of MF-Ours with/without ACM and FFD.}}
    \label{figure_ablation_modules}
    \vspace{-2mm}
\end{figure}

\subsection{Ablation study}
\label{ablation}
In this section, we conduct the ablation study on MF-Ours proposed in our previous work~\cite{bae2022deep}. 
We use the same models and datasets for the experiments used in \secref{compriason_kitti}.
\subsubsection{Effectiveness of the proposed modules.}
We conduct an ablation study to demonstrate the effectiveness of the proposed modules, ACM and FFD in \tabref{table_result_modules}.
The models with only the ACM module or FFD module marginally improve the depth estimation performance due to the absence of proper attention map fusions. On the other hand, our MonoFormer with both ACM and FFD significantly improves the performance. The results show the proposed model achieves the best performance in all measurements. The qualitative comparison in \figref{figure_ablation_modules} shows that the model with both ACM and FFD keeps clearer object boundaries, even a small car in far depth.

\subsubsection{Visualization of attention maps.}
\begin{figure}[t!]
    \centering
    \newcommand\w{10cm}
    \resizebox{\columnwidth}{!}{
    \begin{tabular}{cc}
    \includegraphics[width=\w]{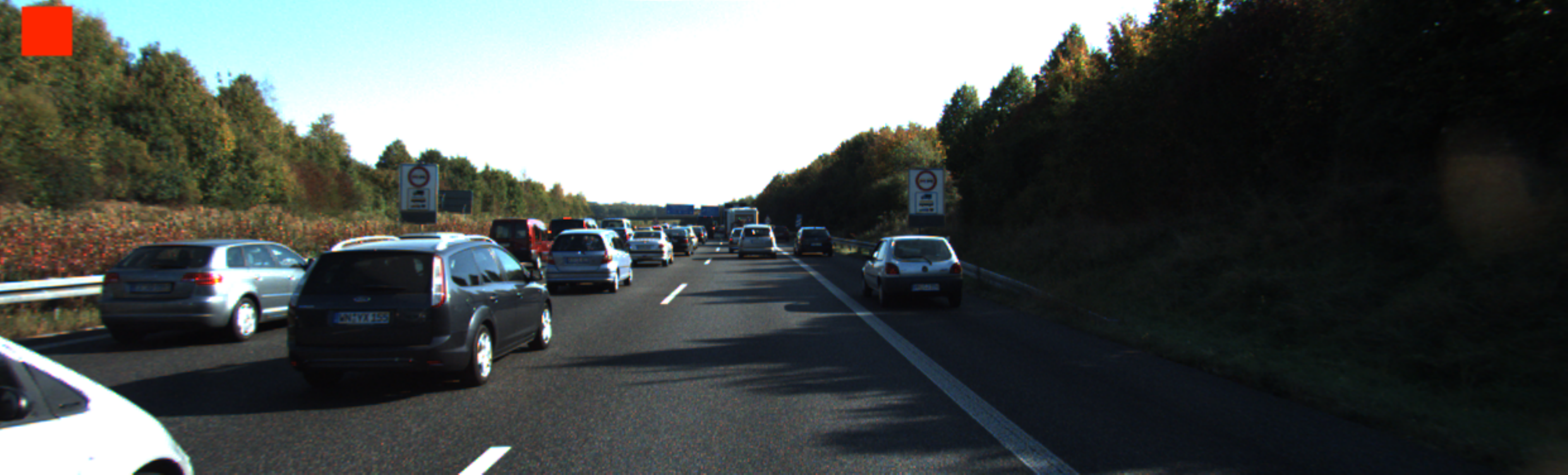}&\includegraphics[width=\w]{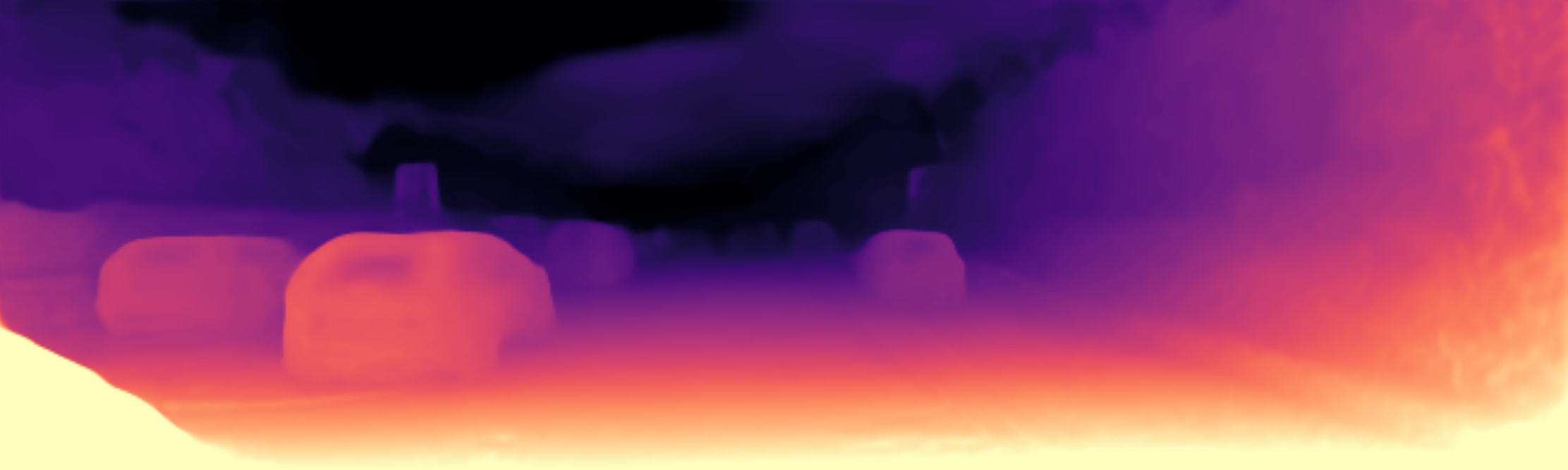}            \\
    \fontsize{20}{15} \selectfont Input image           &  \fontsize{20}{15} \selectfont Output Depth\\
    \includegraphics[width=\w]{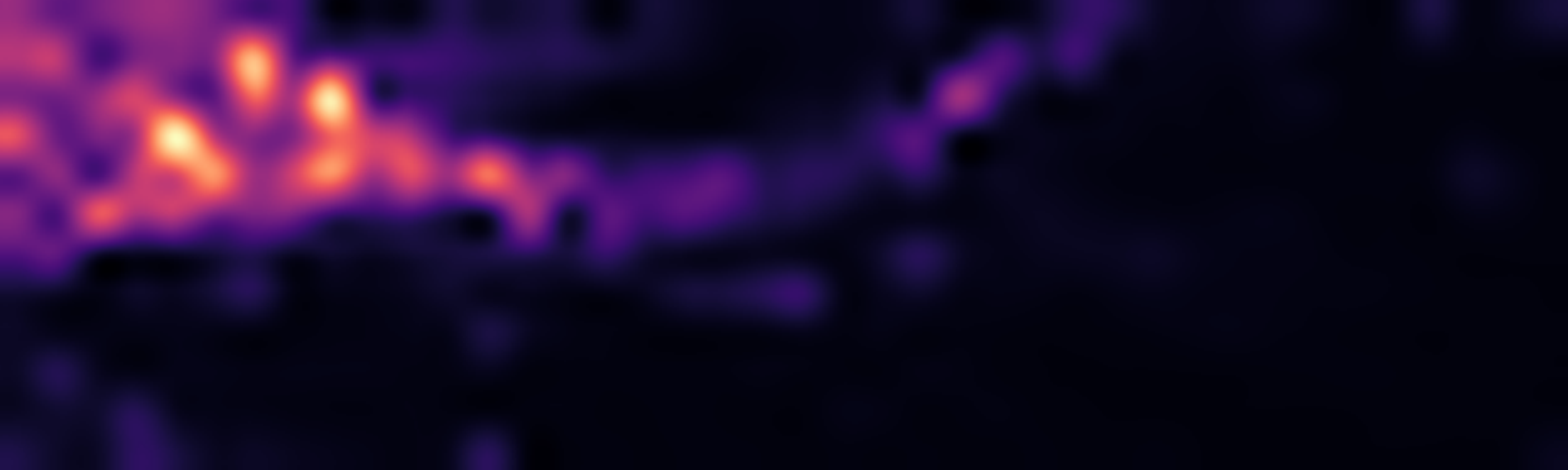}           &   \includegraphics[width=\w]{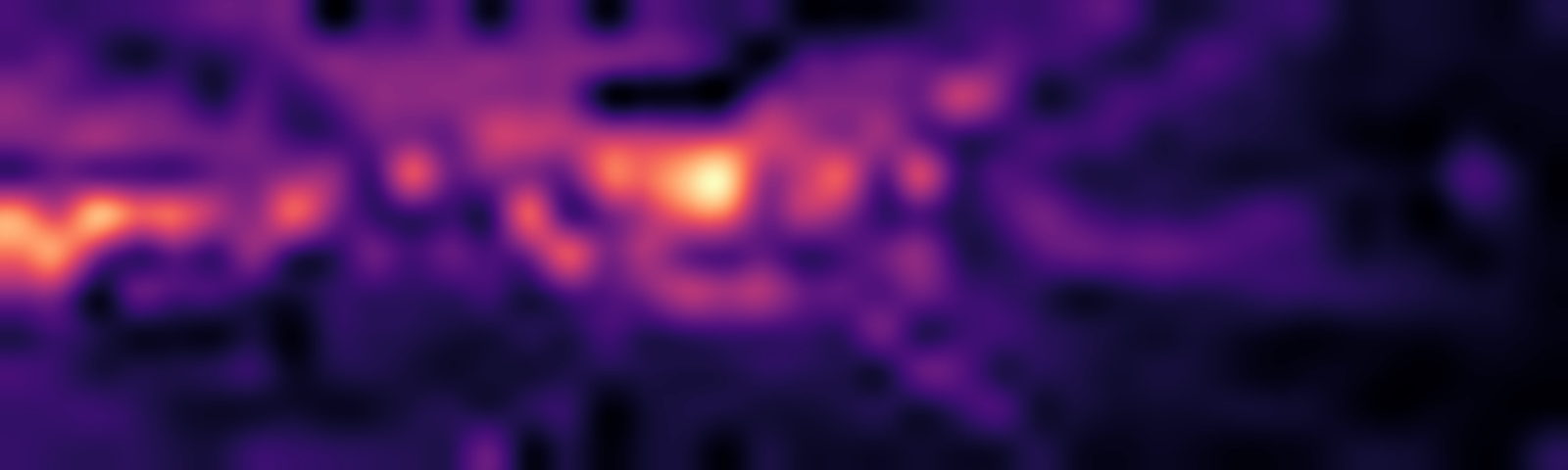}         \\
    \multicolumn{2}{c}{\fontsize{20}{15} \selectfont (a) Attention map of CNN-Transformer encoder } \\
    \includegraphics[width=\w]{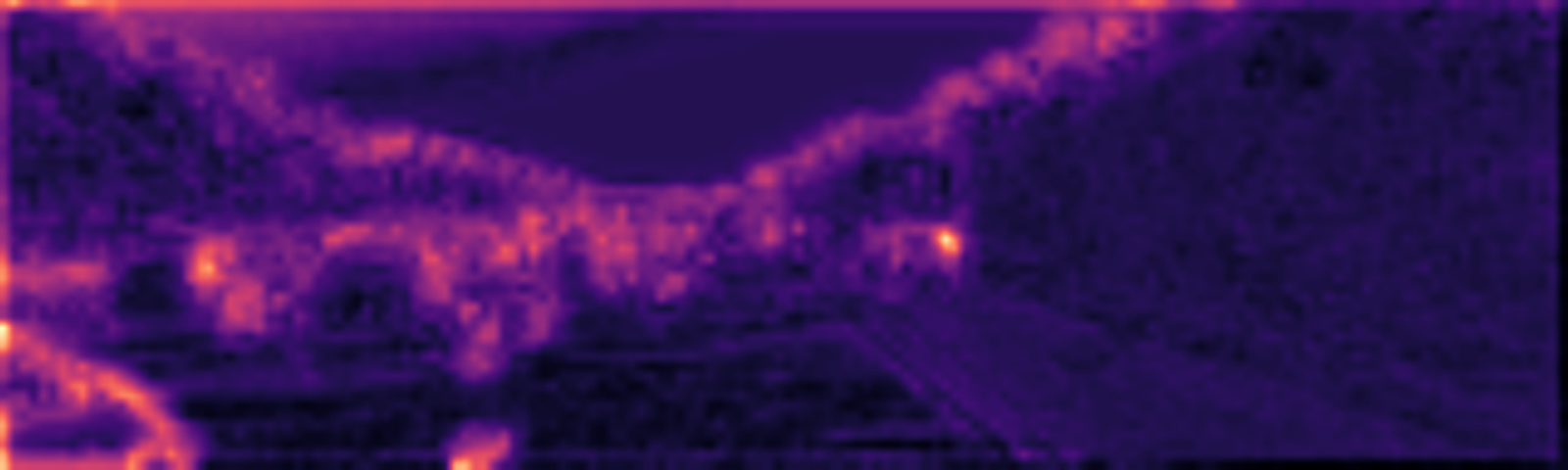}&\includegraphics[width=\w]{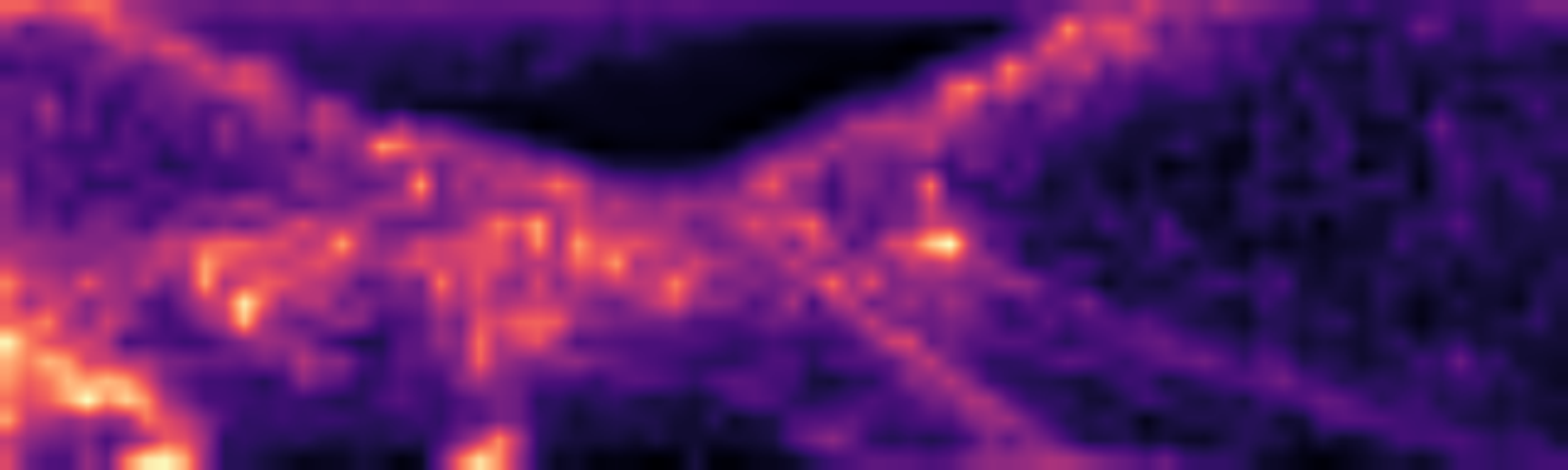}            \\
    \multicolumn{2}{c}{\fontsize{20}{15} \selectfont (b) Attention map of ACM} \\
    \includegraphics[width=\w]{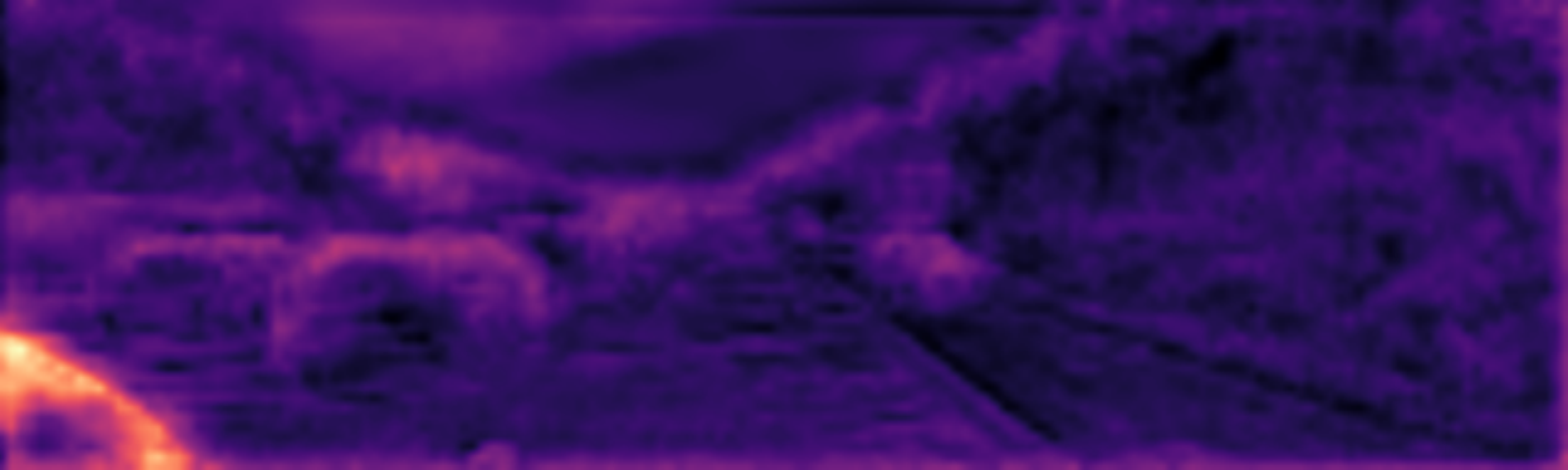}           &\includegraphics[width=\w]{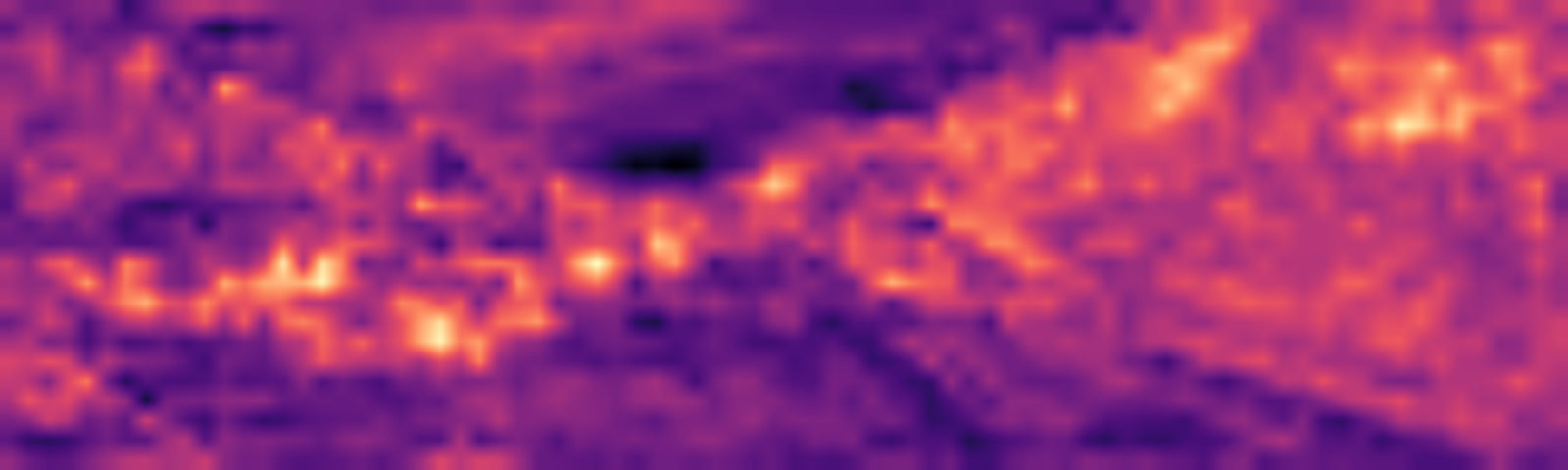}            \\
    \multicolumn{2}{c}{\fontsize{20}{15} \selectfont (c) Feature map of FFD}
    \end{tabular}}
    \vspace{-2mm}
    \captionof{figure}{\textbf{Visualization of attention and feature maps}. The left column from the second row is the maps from shallow layers, whereas the right is the maps from deep layers.}
    \label{figure_monoformer_attention}
    \vspace{-2mm}
\end{figure}

We visualize the attention maps from the lower to higher layers of Transformers. As shown in \figref{figure_monoformer_attention}, the encoder in the shallow layer extracts local region features. The deeper the layer, the more global shape contexts are extracted. Another observation is that ACM captures more detailed attention at different depths of the encoder features. FFD enhances the encoder features by fusing them with the attention map from ACM. The fused feature captures features from coarse to fine details. These experiments show that our model is capable of accurate pixel-wise prediction as it secures adequate local details. 

\begin{table}[t]
    \centering
    \resizebox{\columnwidth}{!}{
    \begin{tabular}{cccccc}
    \hline
    $\#$ of layers & Abs Rel $\downarrow$ & RMSE $\downarrow$ &  & $\delta < 1.25$ $\uparrow$ & $\delta < 1.25^2 $ $\uparrow$ \\ \hline
    $L=2$ & 0.148 & 5.327  &   &0.810  & 0.939  \\
    $L=3$ & 0.112 & 4.745&  & 0.881  & \textbf{0.962} \\
    $L=4$ & \textbf{0.104} & \textbf{4.580} &  & 0.873 & \textbf{0.962} \\
    $L=5$ & 0.111 & 4.692 & &  \textbf{0.884} & \textbf{0.962} \\
    \hline
    \end{tabular}}
    \vspace{-2mm}
    \caption{\textbf{Ablation study on the number of layers.}}
    \label{table_abliation_L}
    \vspace{-3mm}
\end{table}

\vspace{-0.5mm}

\begin{figure*}[h]
\centering
\begin{tabular}{cccc}
\hspace{-5mm}
\includegraphics[width=0.33\textwidth]{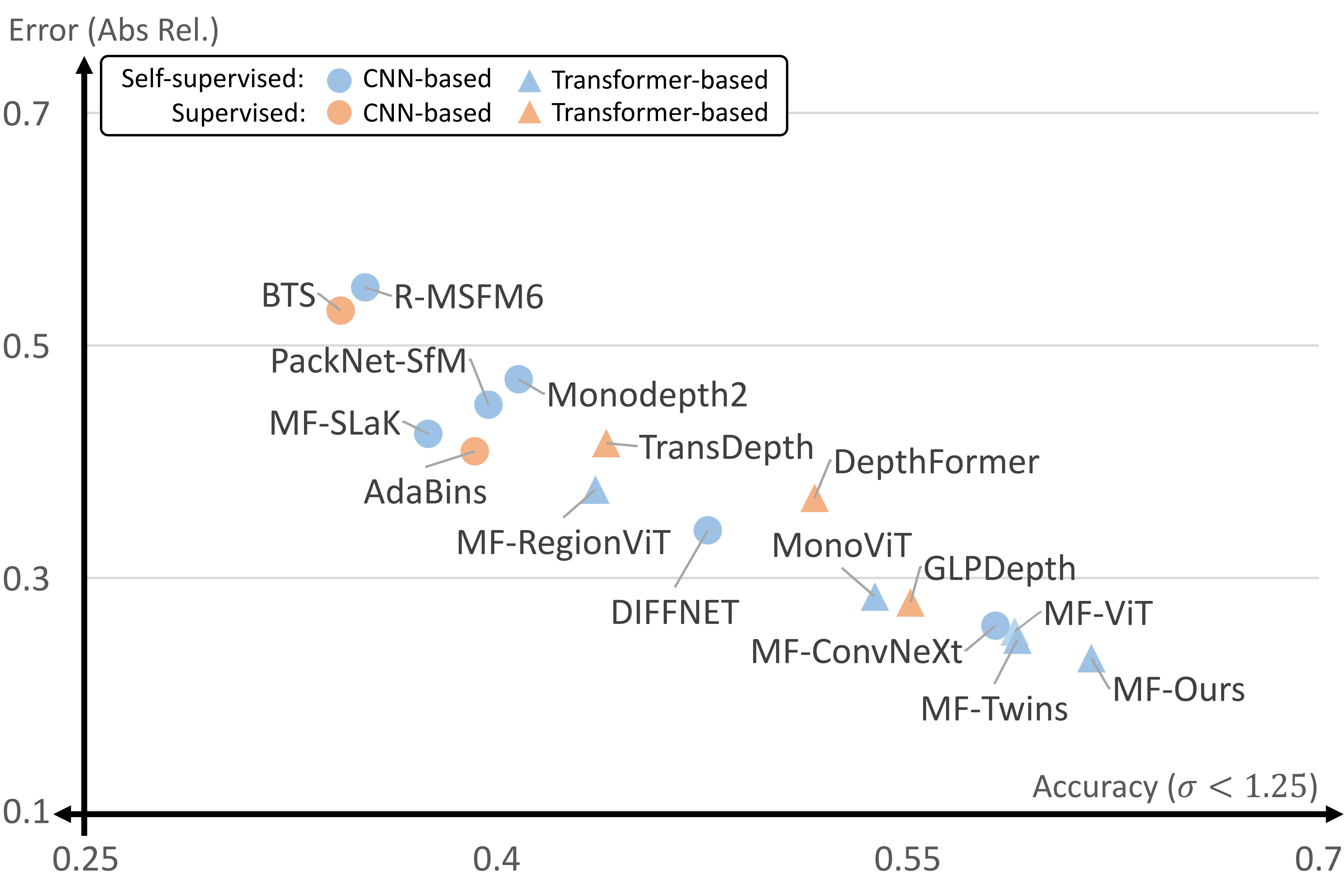} &
\includegraphics[width=0.33\textwidth]{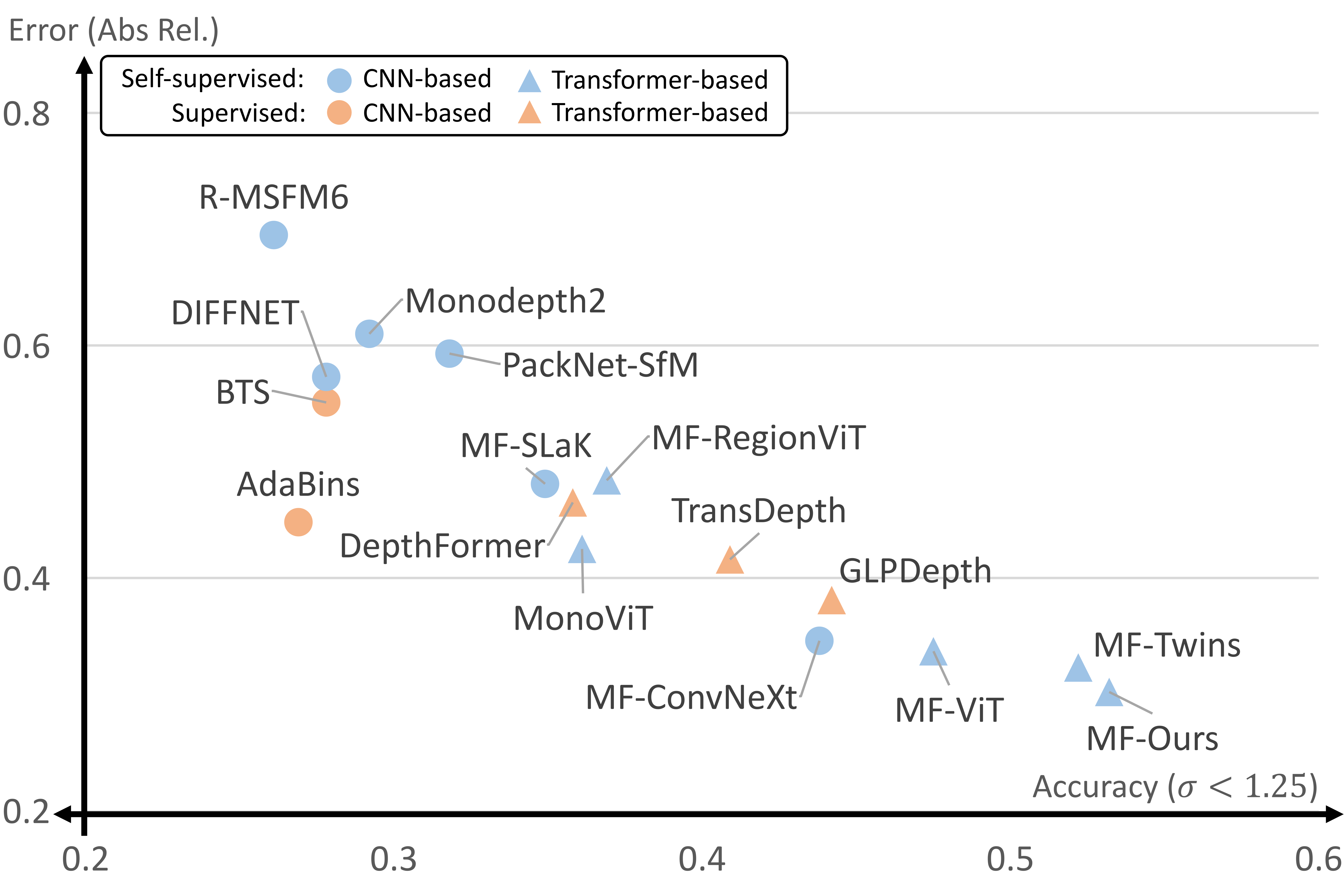} &
\includegraphics[width=0.33\textwidth]{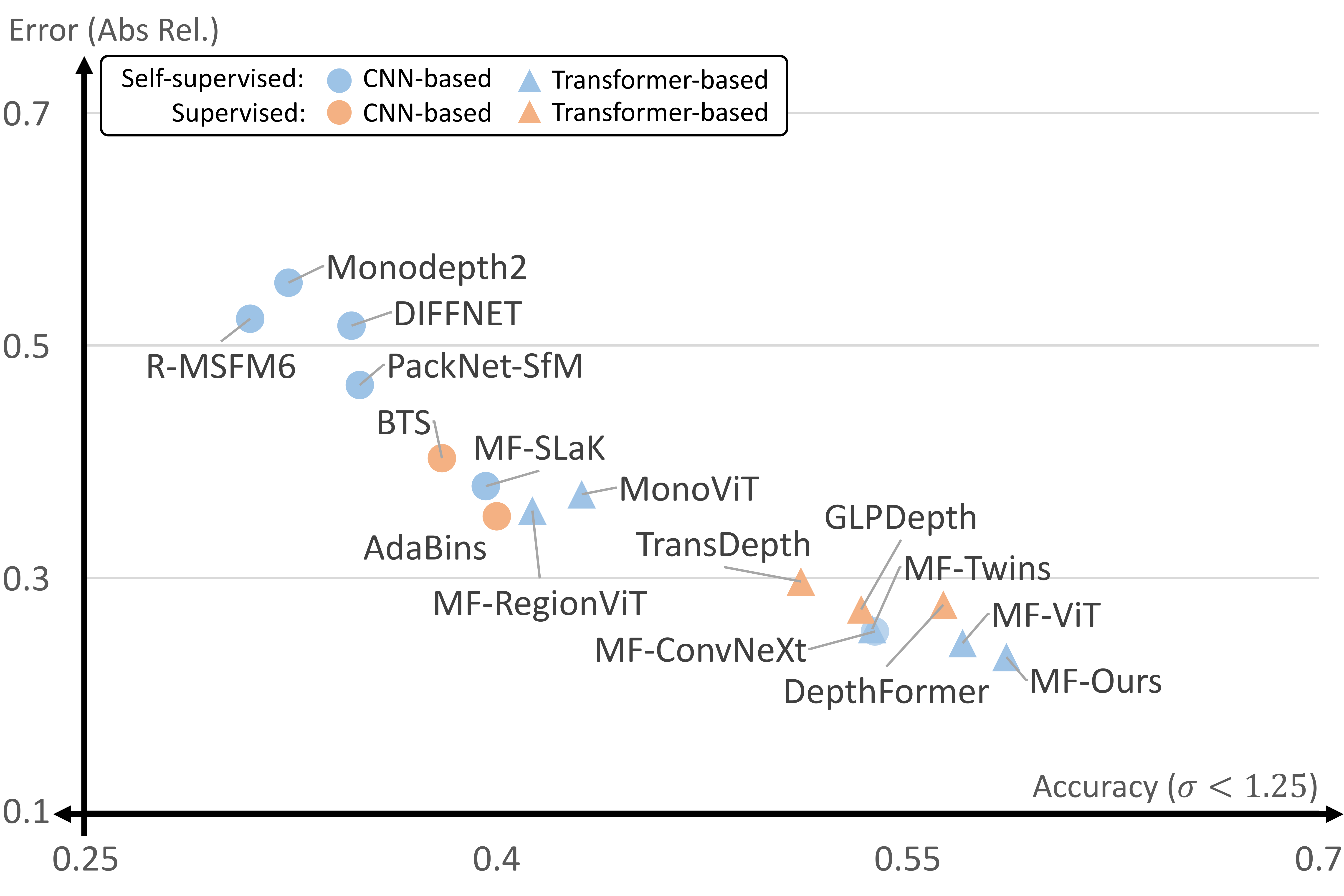} \\
(a) MVS  & (b) RGBD & (c) SUN3D \\ 
\vspace{-1mm}
\end{tabular}
\begin{tabular}{cccc}
\includegraphics[width=0.33\textwidth]{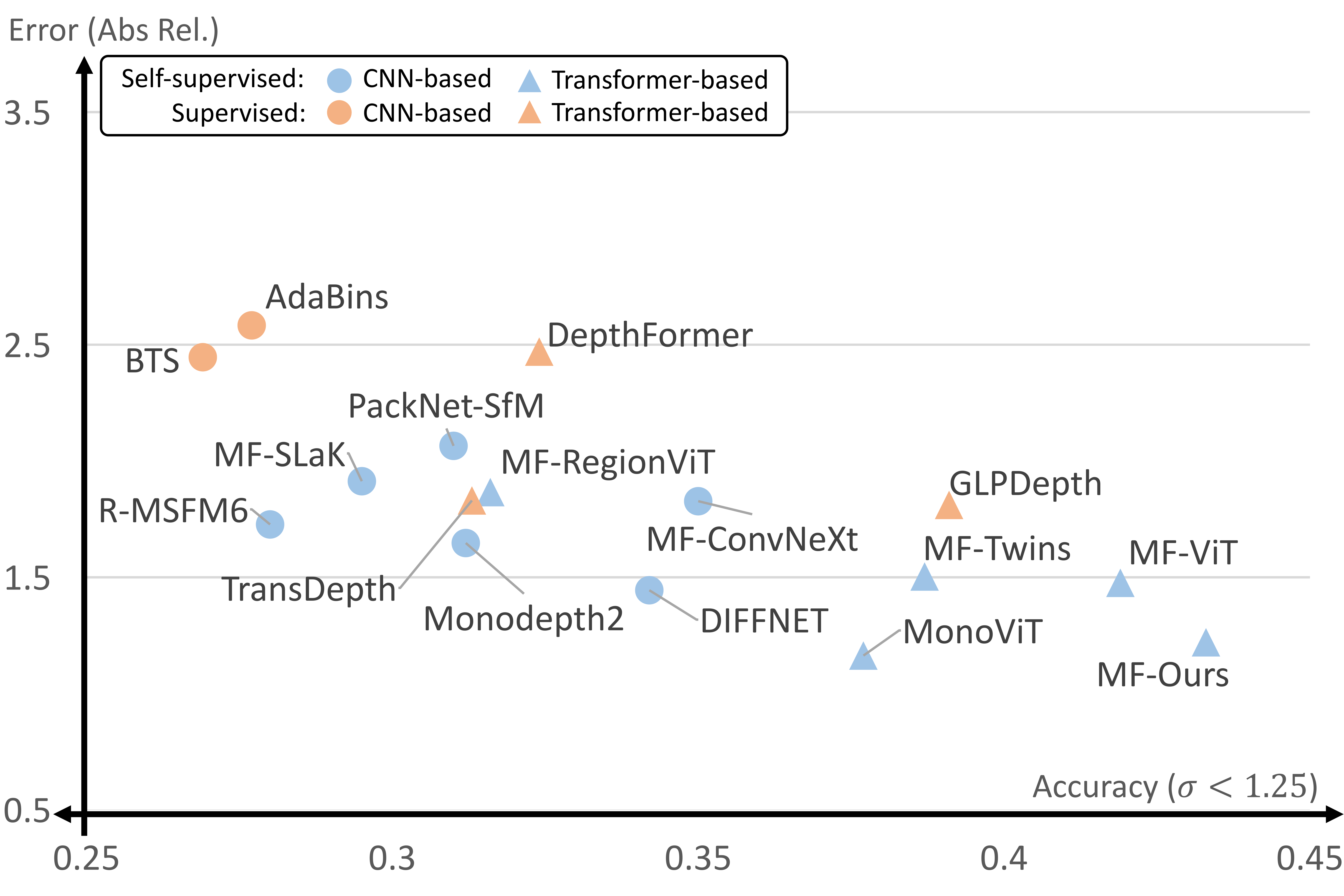} &
\includegraphics[width=0.33\textwidth]{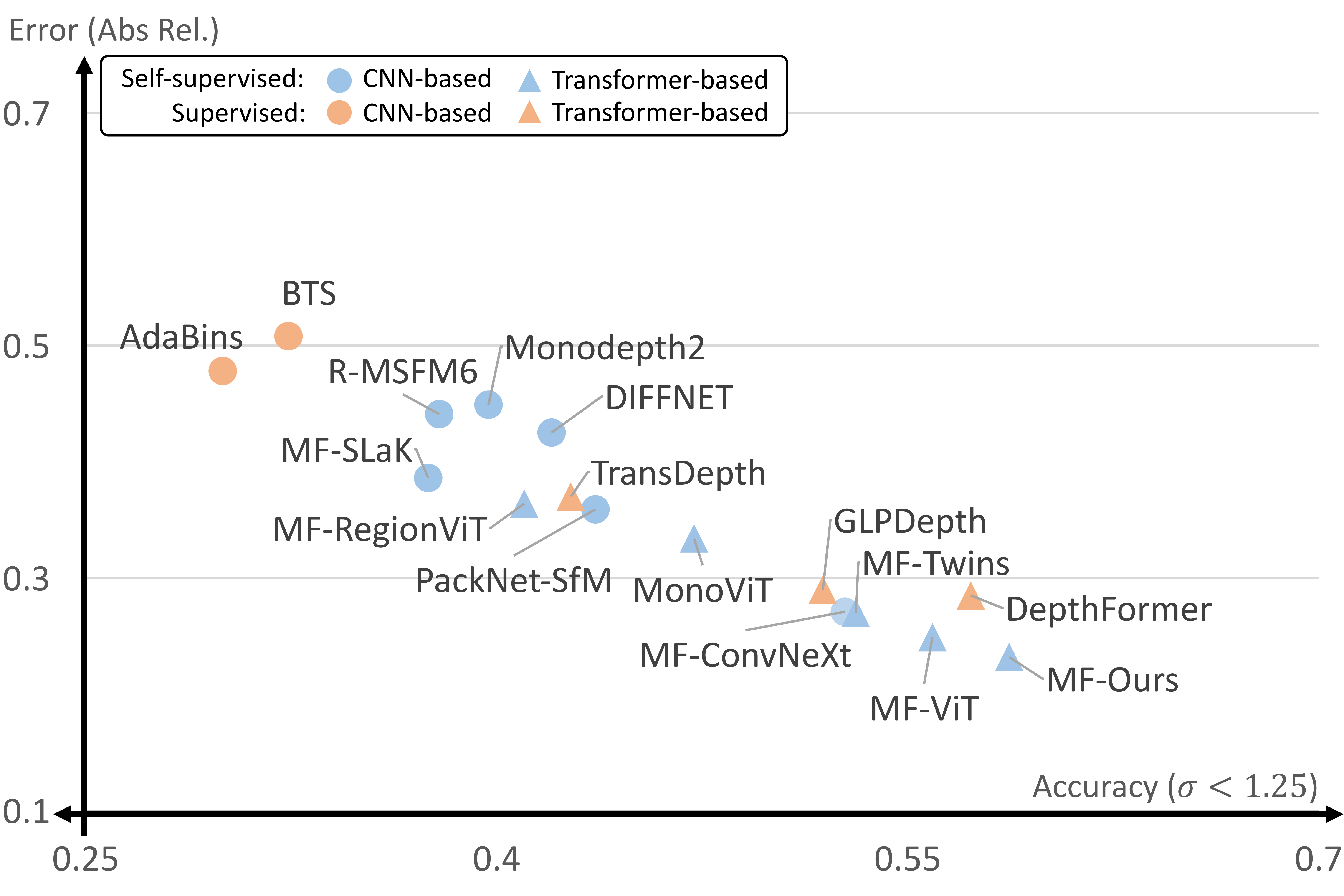} \\
(d) Scenes11   & (e) ETH3D \\
\end{tabular}
\caption{ \textbf{Depth Performance Comparison on the out-of-distribution (RGBD, SUN3D, MVS, ETH3D, and Scenes11) datasets.} The $x$-axis is the accuracy under threshold $\delta < 1.25$. The $y$-axis shows absolute relative error (Abs Rel).}
\label{table_result_mvs}
\end{figure*}

\subsubsection{The number of encoder and decoder layers.}
We compare the performance of MF-Ours according to the number of encoder and decoder layers in \tabref{table_abliation_L}. 
Each layer has ACM and FFD modules. 
We find out that the model with four transformer layers achieves the best performance. 
The results are slightly degraded with the MF-Ours with 3 or 5 layers. 
Therefore, we set $L$ as four for MF-Ours in all experiments in this paper.

\begin{figure*}[p]
\centering
\newcommand\iw{80cm}
\newcommand\ih{30cm}
\newcommand\w{200}
\newcommand\h{180}
\newcommand\textw{120}
\newcommand\texth{200}
\resizebox{\linewidth}{!}{%
\begin{tabular}{ccccccccc}
\multicolumn{1}{c}{\fontsize{\w}{\h} \selectfont RGBD } & 
\multicolumn{1}{c}{\fontsize{\w}{\h} \selectfont SUN3D } & 
\multicolumn{1}{c}{\fontsize{\w}{\h} \selectfont MVS } & 
\multicolumn{1}{c}{\fontsize{\w}{\h} \selectfont ETH3D } & 
\multicolumn{1}{c}{\fontsize{\w}{\h} \selectfont Scenes11 } \\
\vspace{30mm}\\
\rotatebox[origin=c]{90}{\fontsize{\textw}{\texth}\selectfont Input Images\hspace{-310mm}}\hspace{10mm} 
\includegraphics[width=\iw,height=\ih]{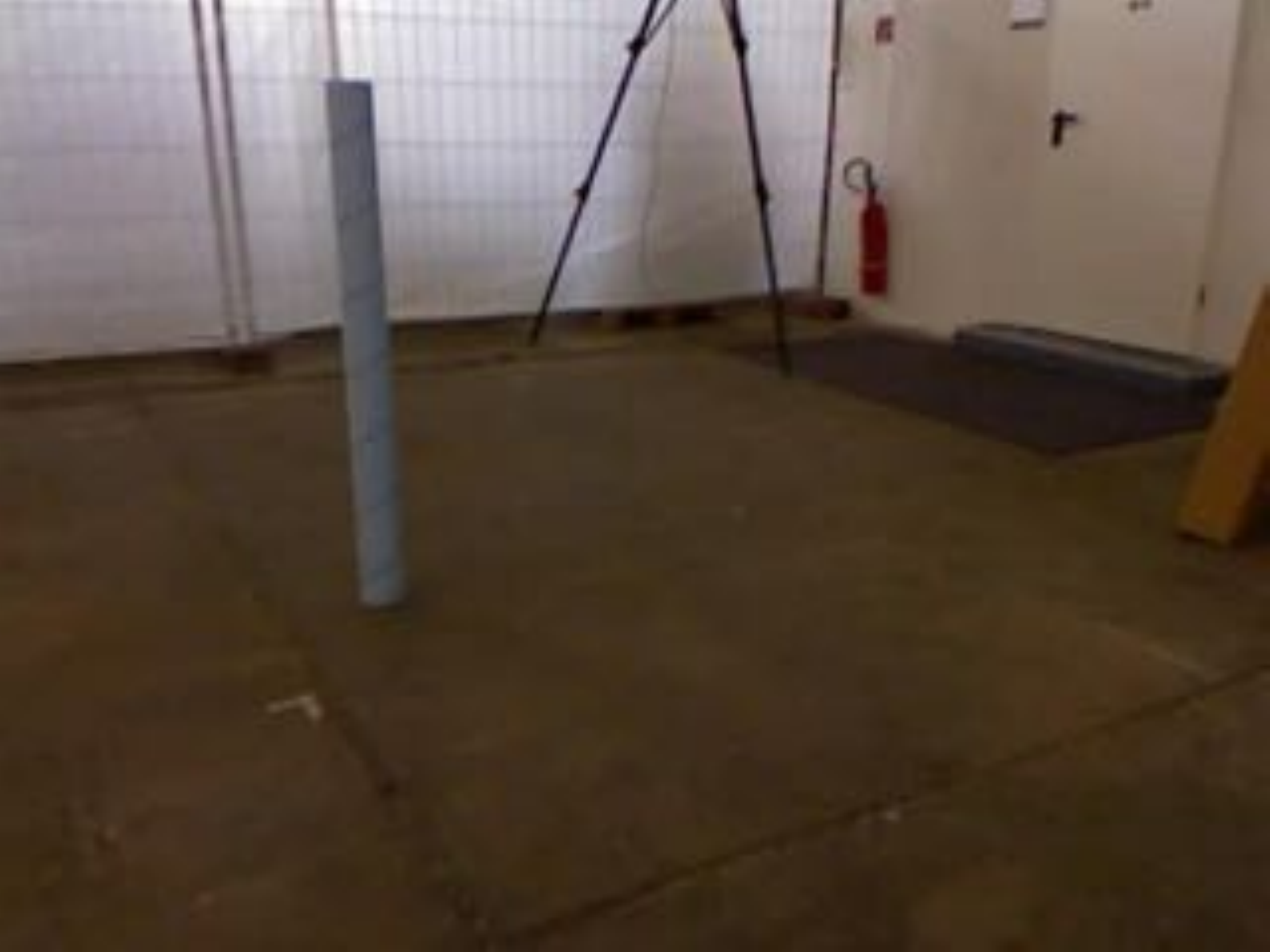} \qquad\qquad\quad &    
\includegraphics[width=\iw,height=\ih]{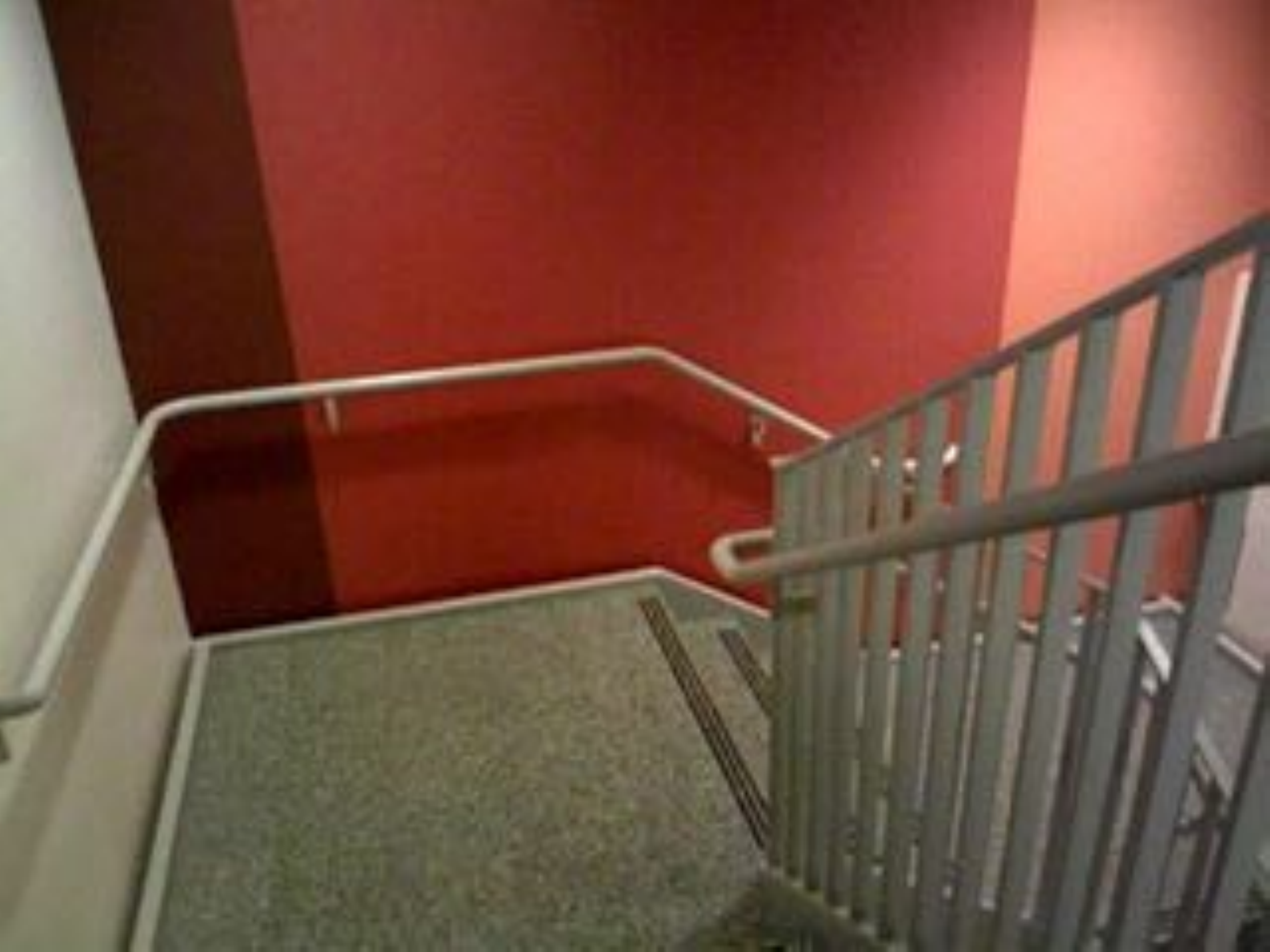} \qquad\qquad\quad &   
\includegraphics[width=\iw,height=\ih]{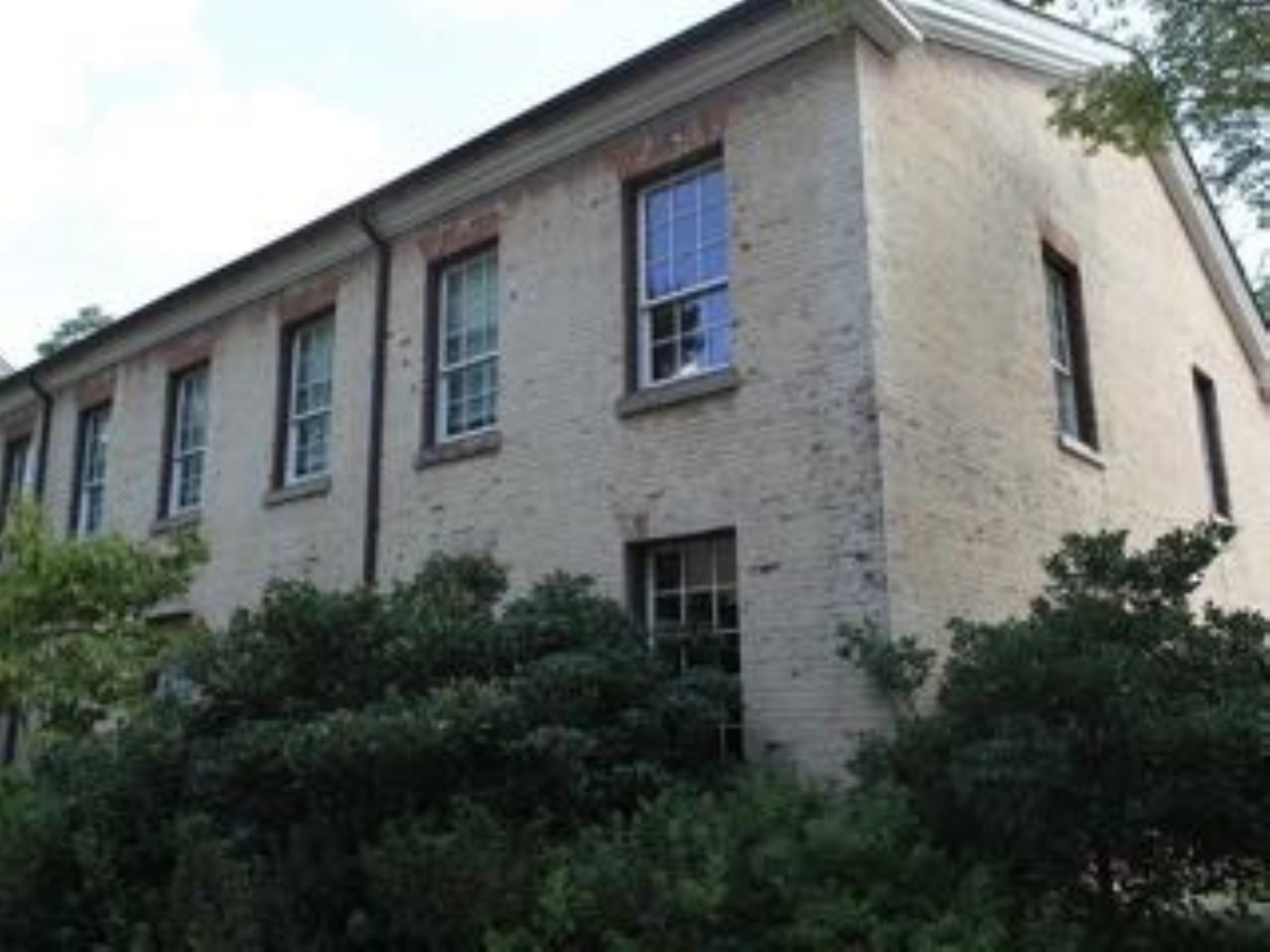} \qquad\qquad\quad &   
\includegraphics[width=\iw,height=\ih]{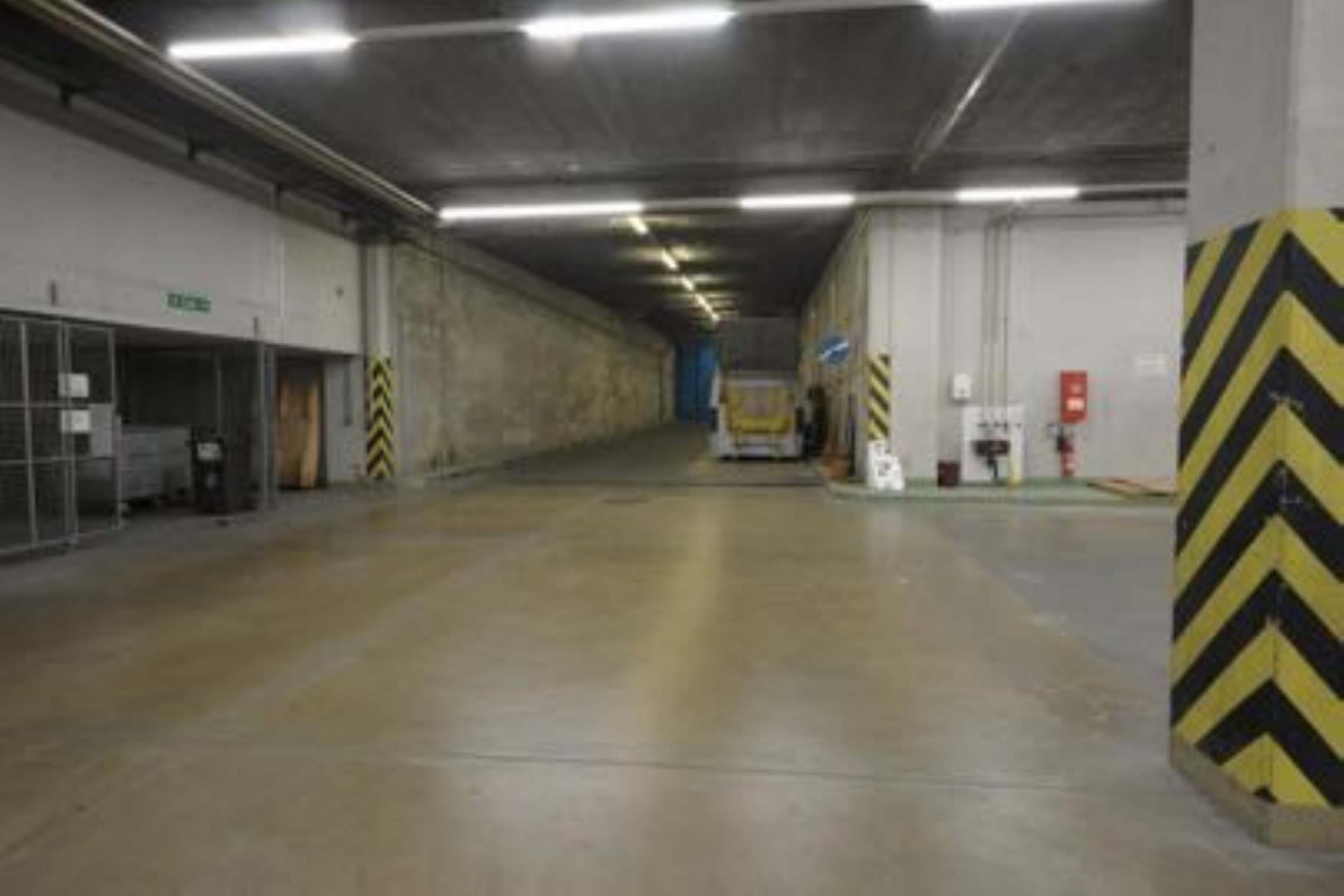} \qquad\qquad\quad &   
\includegraphics[width=\iw,height=\ih]{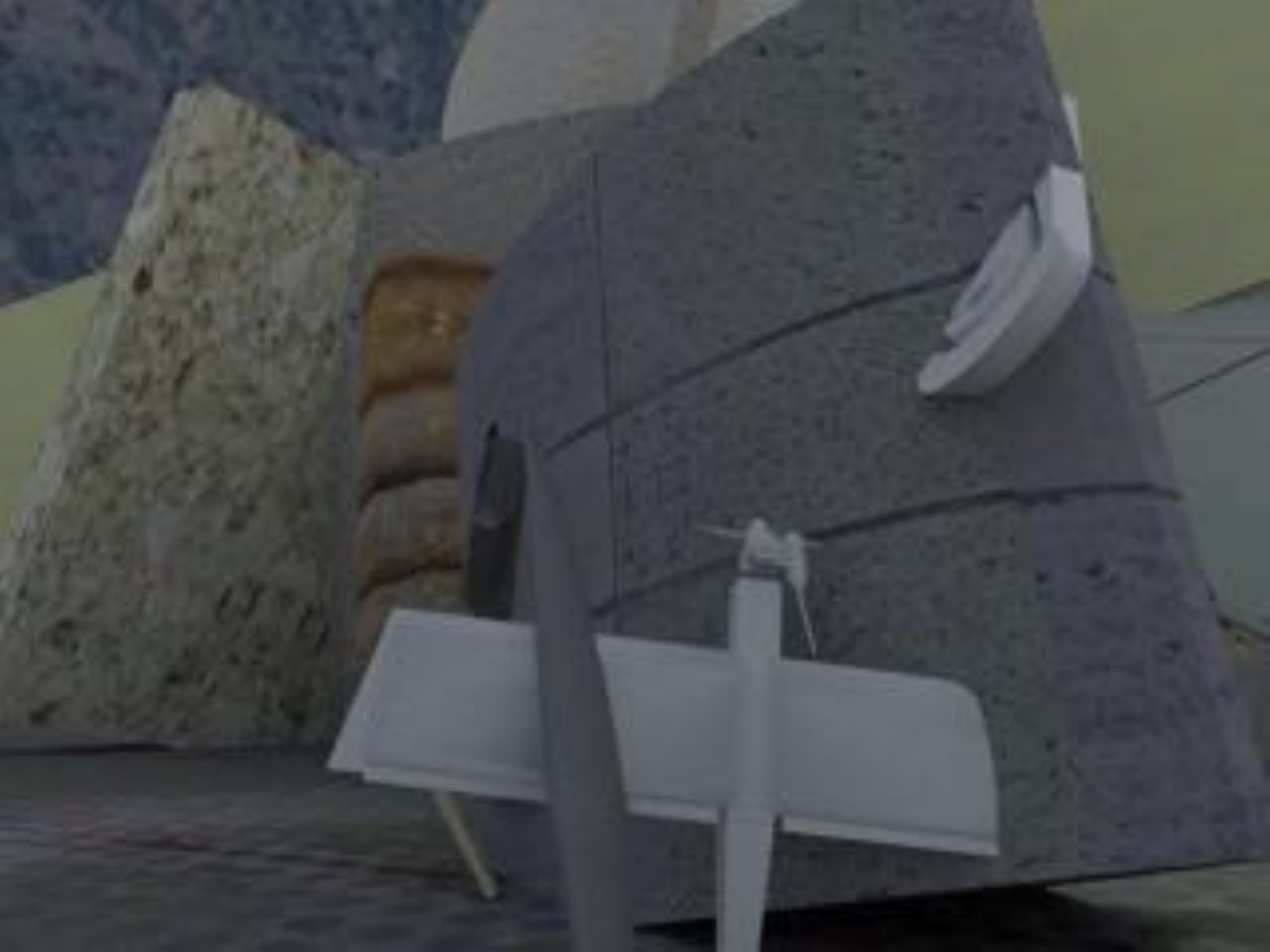}\\ 
\vspace{10mm}\\
\rotatebox[origin=c]{90}{\fontsize{\textw}{\texth} \selectfont Groundtruth\hspace{-250mm}}\hspace{25mm}
\includegraphics[width=\iw,height=\ih]{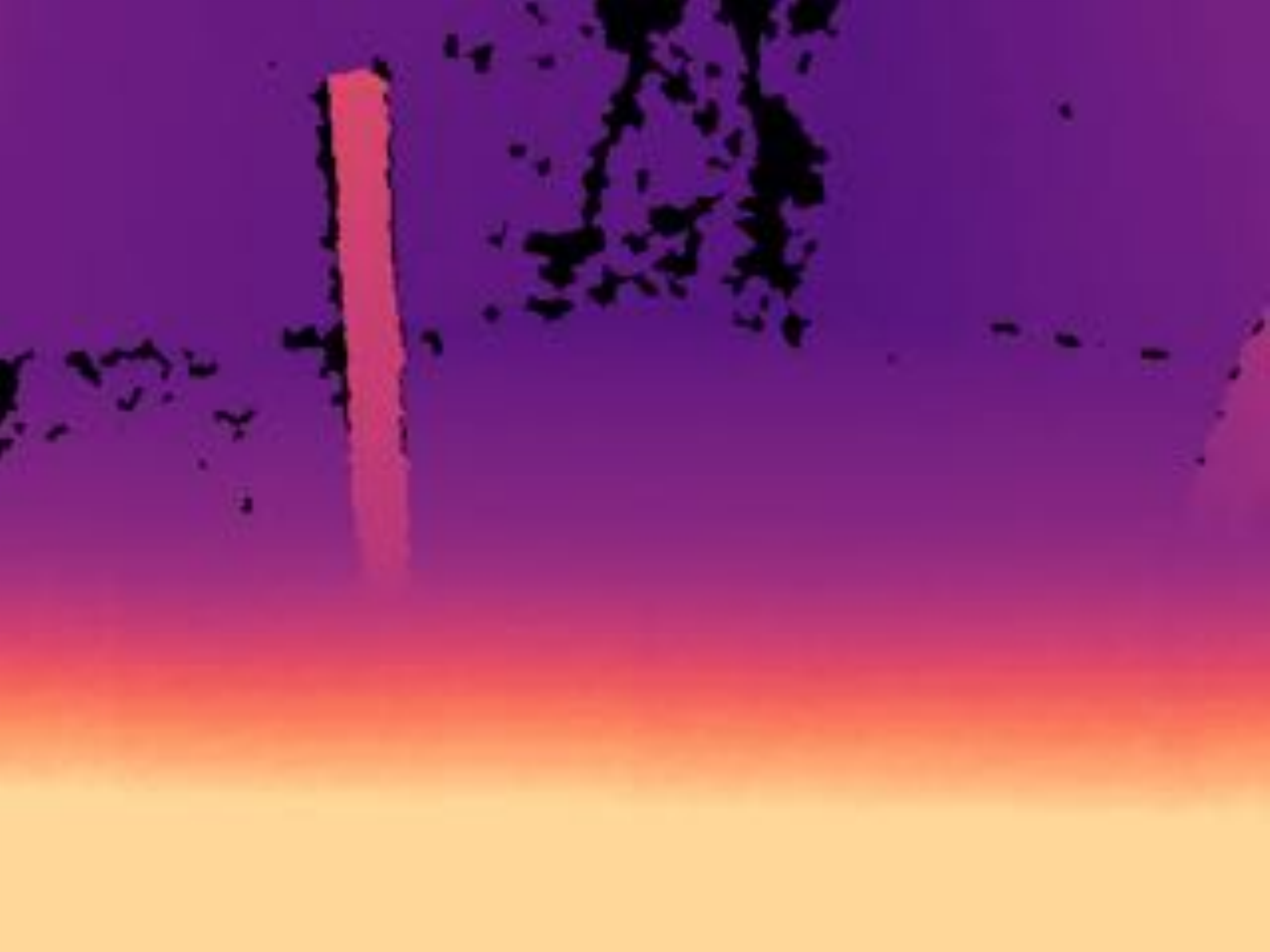} \qquad\qquad\quad &    
\includegraphics[width=\iw,height=\ih]{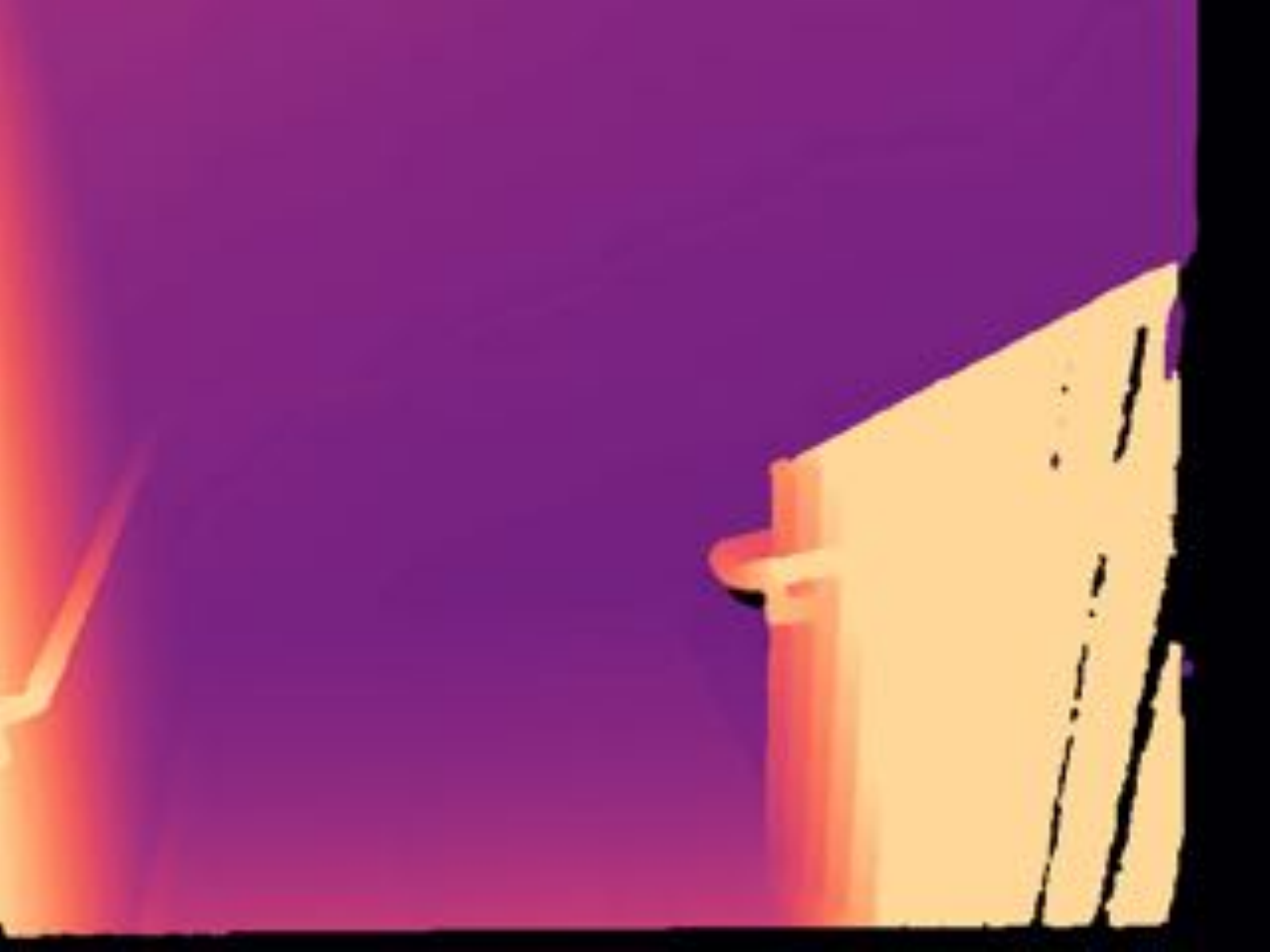} \qquad\qquad\quad &    
\includegraphics[width=\iw,height=\ih]{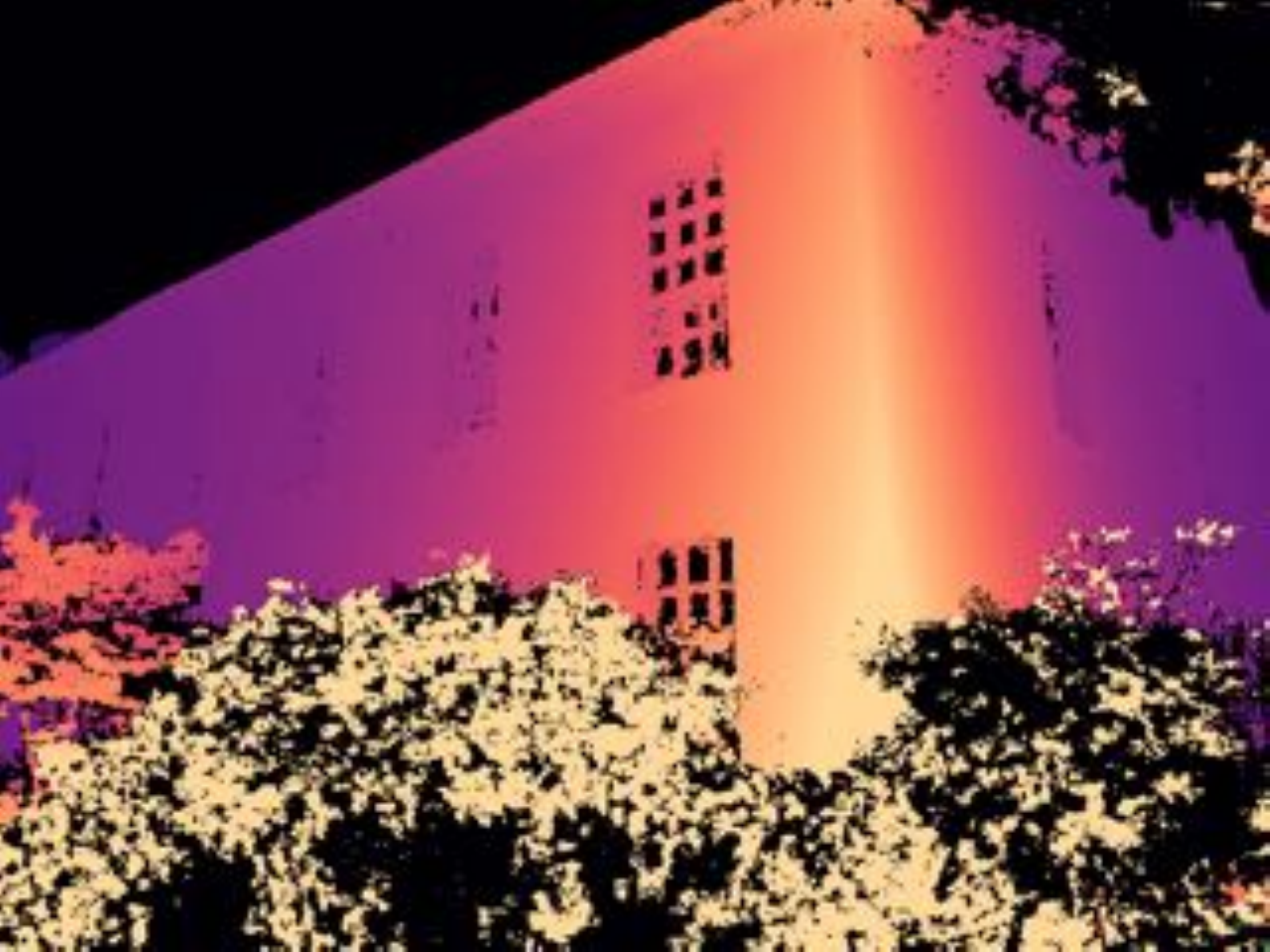} \qquad\qquad\quad &    
\includegraphics[width=\iw,height=\ih]{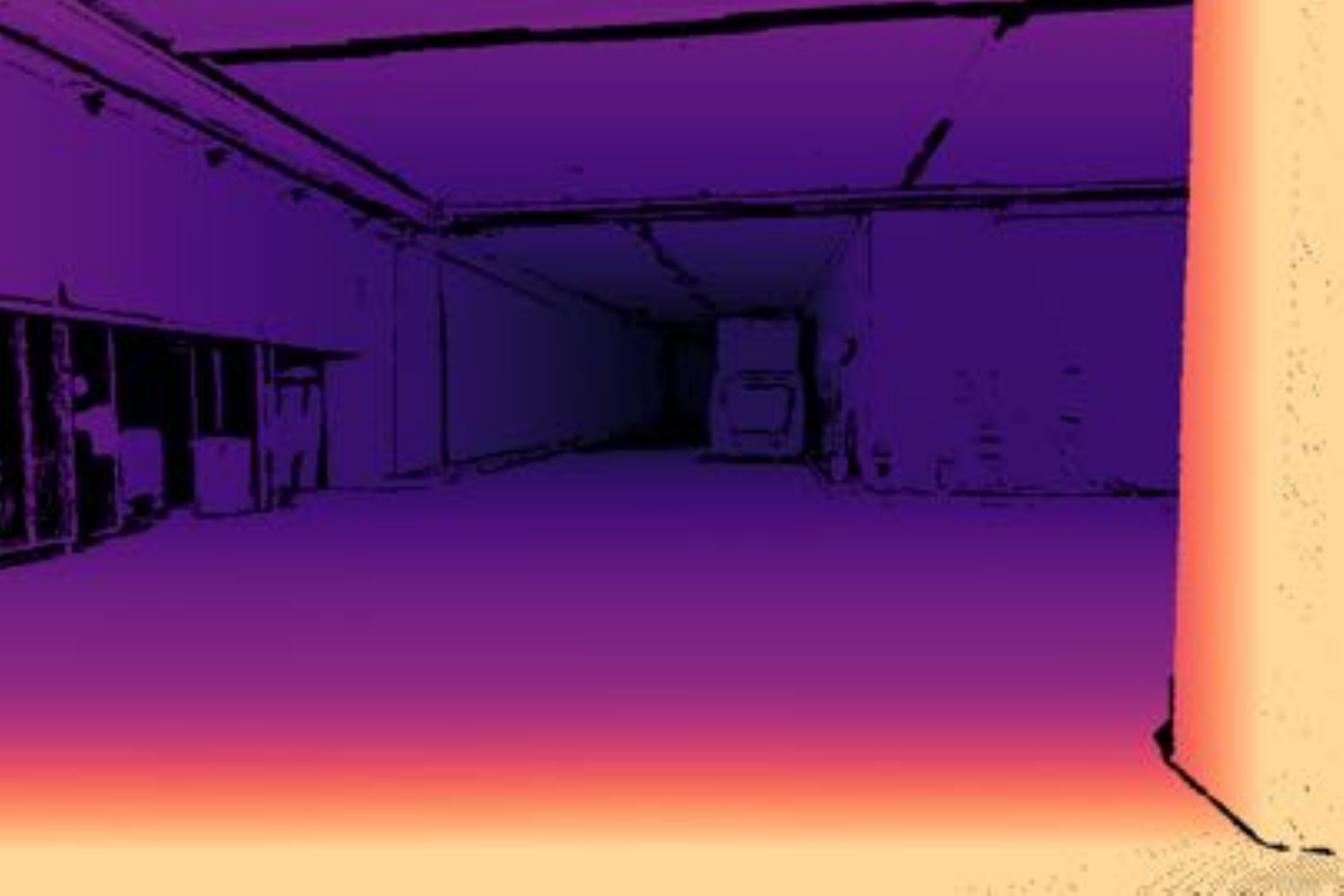} \qquad\qquad\quad &    
\includegraphics[width=\iw,height=\ih]{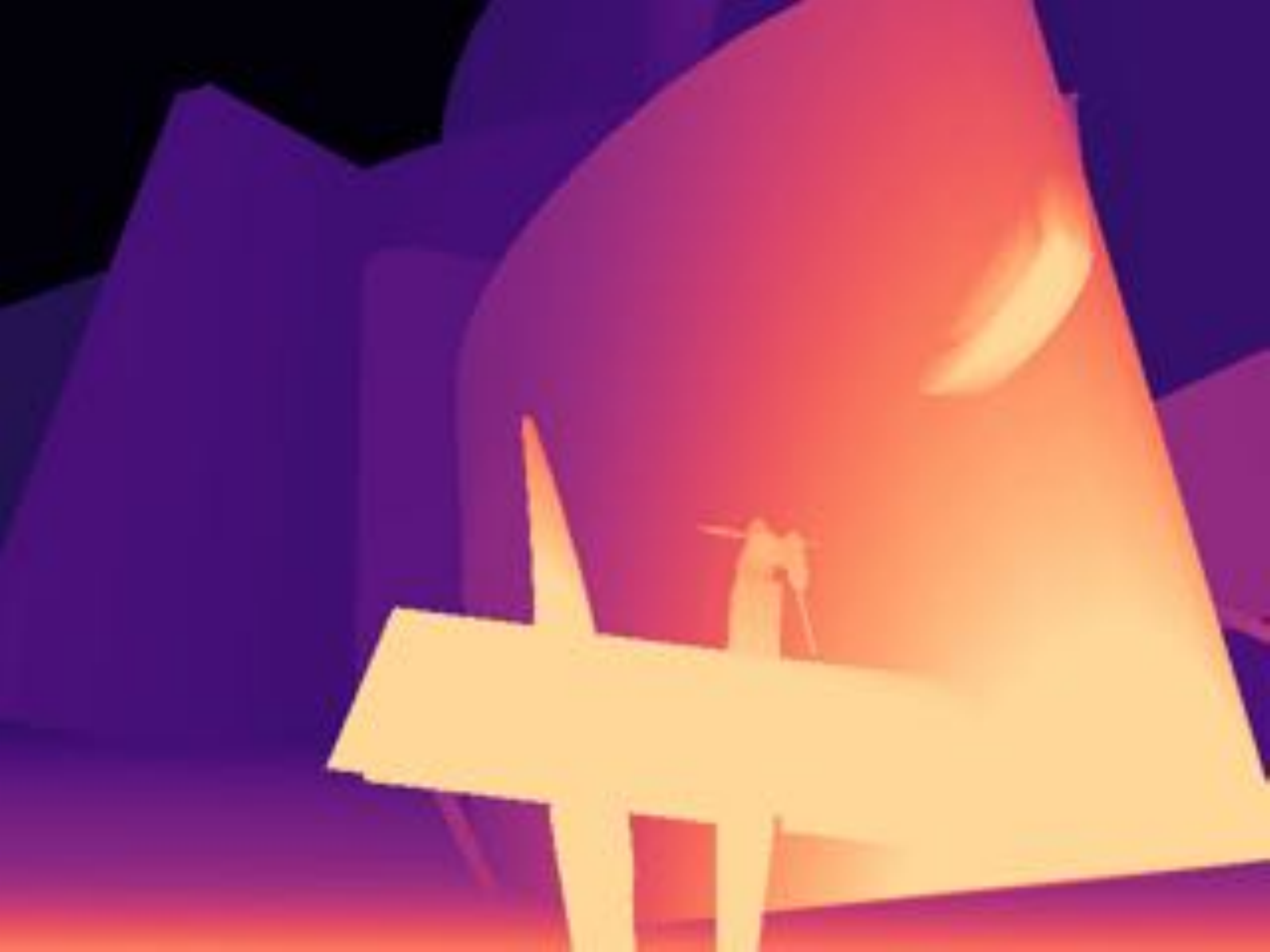}\\

\vspace{30mm}\\
\\\cmidrule{1-5}
\vspace{30mm}\\
\rotatebox[origin=c]{90}{\fontsize{\textw}{\texth} \selectfont Monodepth2\hspace{-270mm}}\hspace{10mm}
\includegraphics[width=\iw,height=\ih]{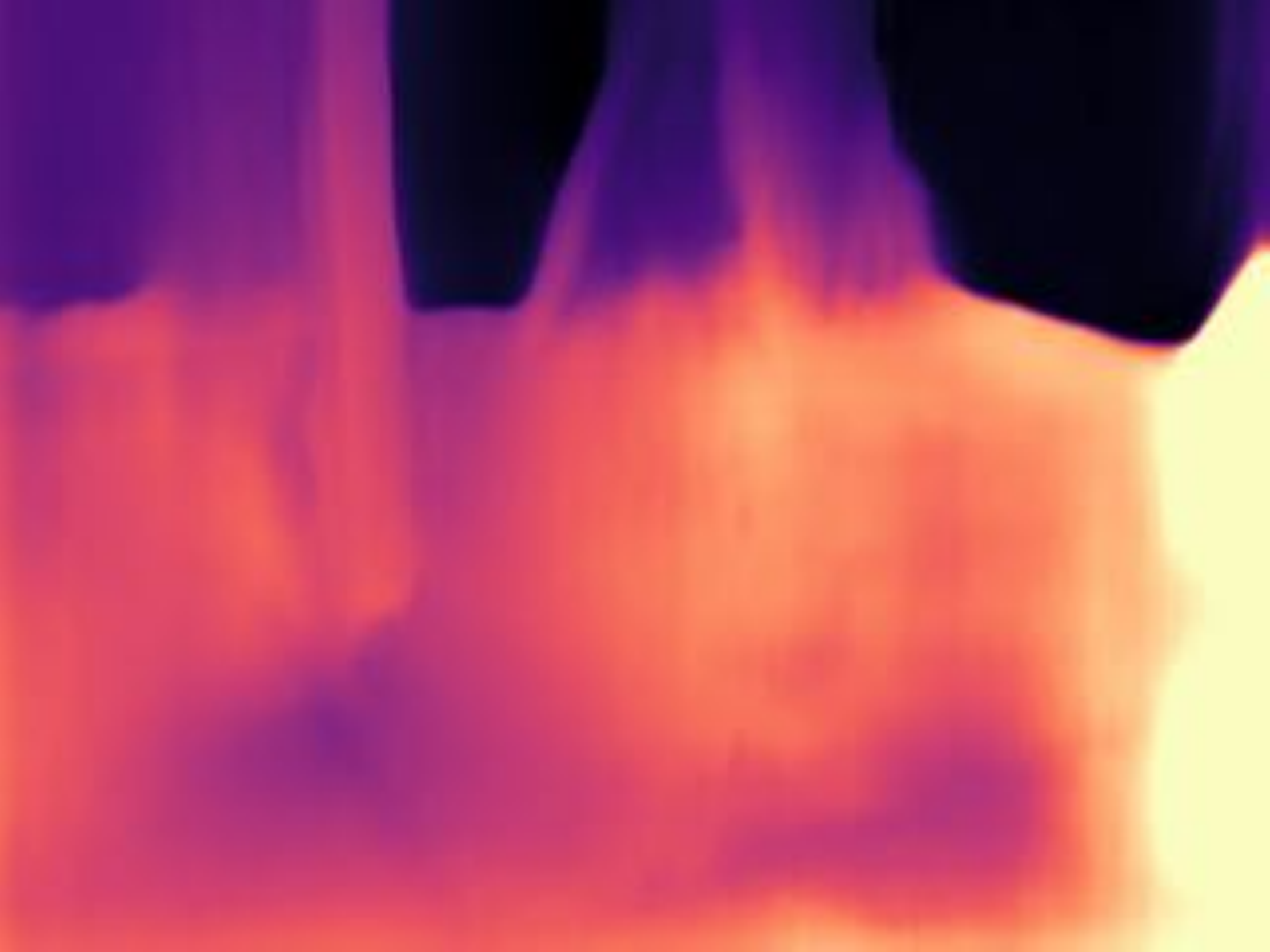} \qquad\qquad\quad & 
\includegraphics[width=\iw,height=\ih]{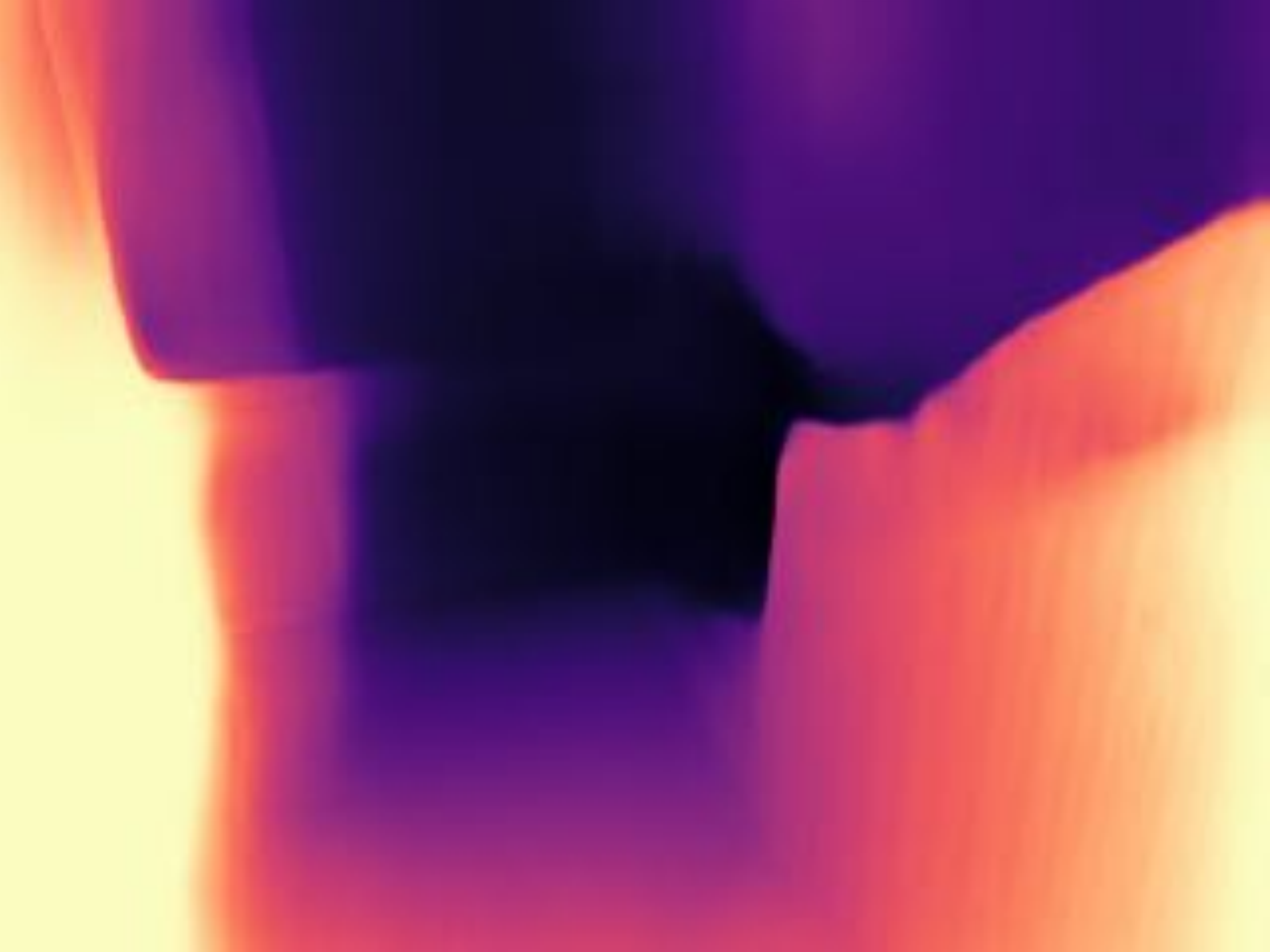} \qquad\qquad\quad & 
\includegraphics[width=\iw,height=\ih]{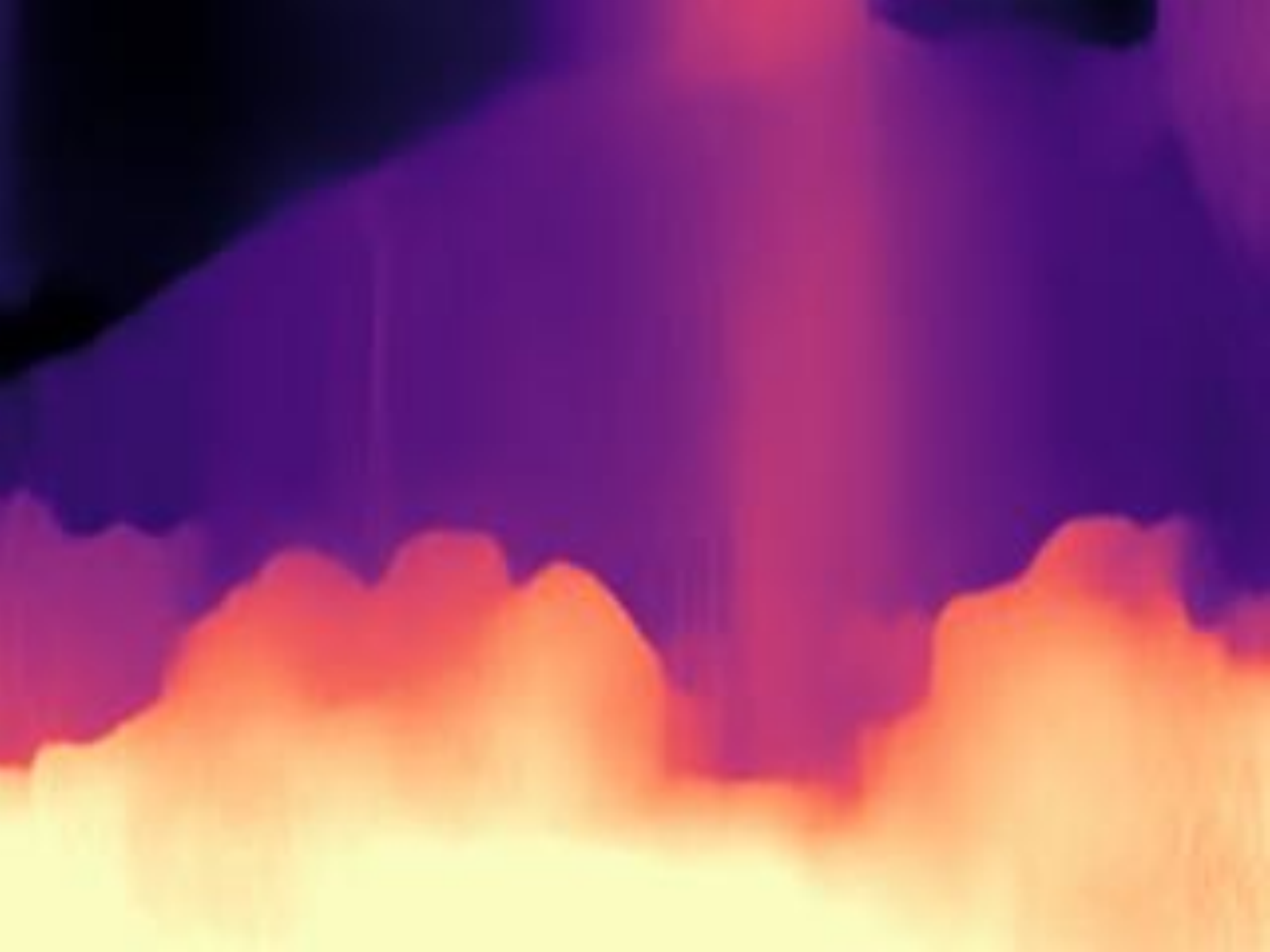} \qquad\qquad\quad &  
\includegraphics[width=\iw,height=\ih]{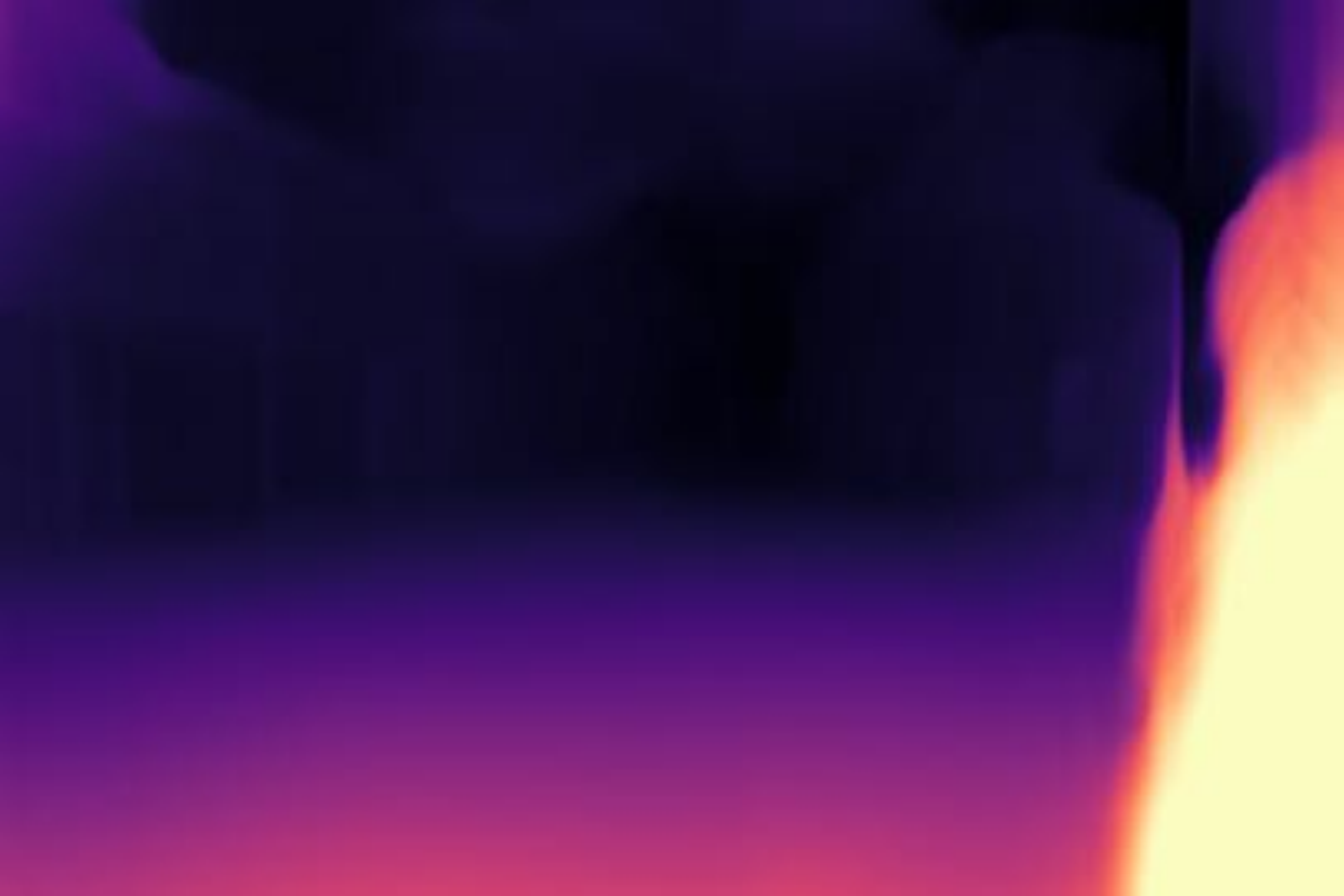} \qquad\qquad\quad &
\includegraphics[width=\iw,height=\ih]{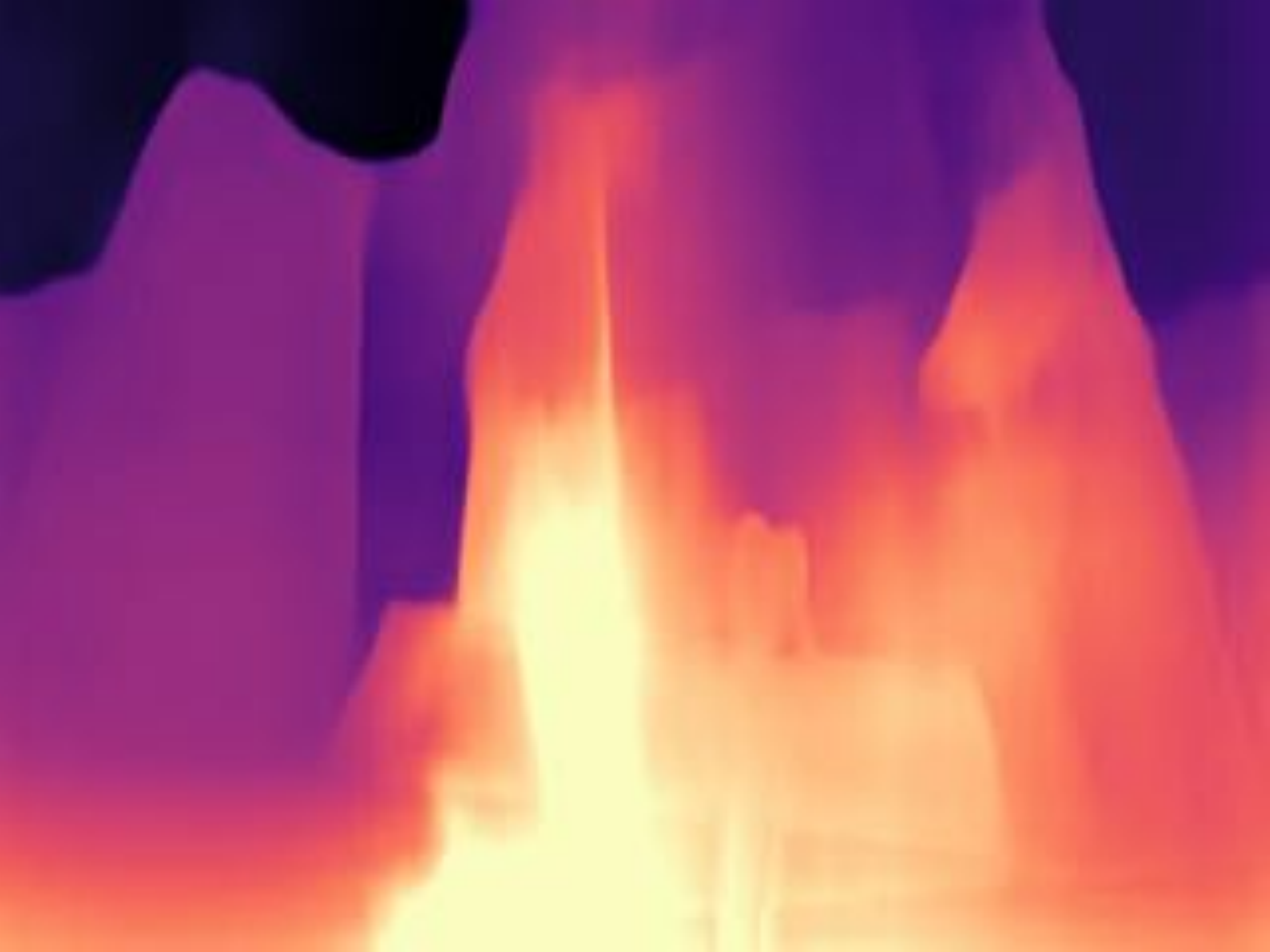}\\

\vspace{10mm}\\
\rotatebox[origin=c]{90}{\fontsize{\textw}{\texth} \selectfont PackNet-SfM\hspace{-270mm}}\hspace{25mm}
\includegraphics[width=\iw,height=\ih]{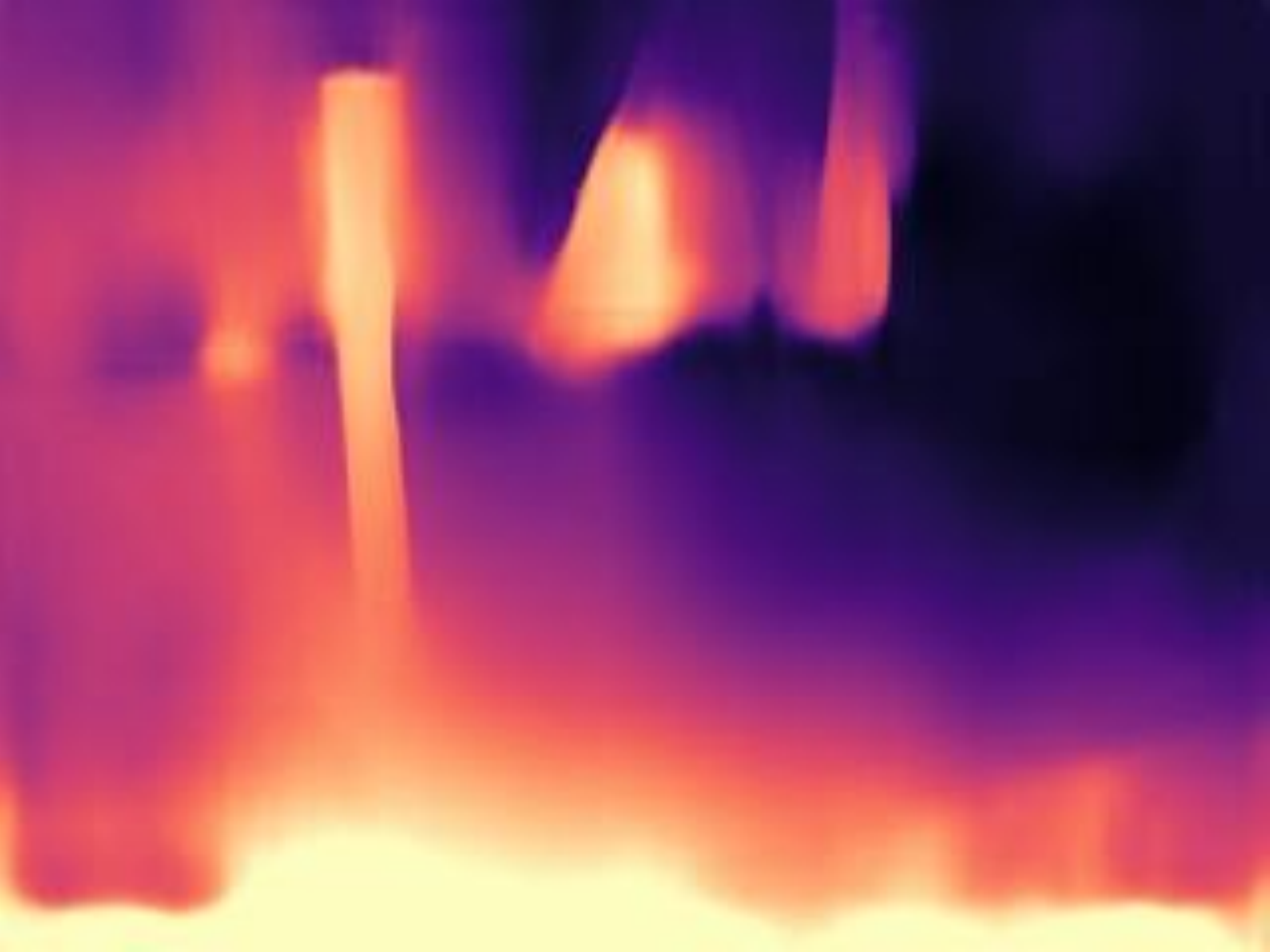} \qquad\qquad\quad &  
\includegraphics[width=\iw,height=\ih]{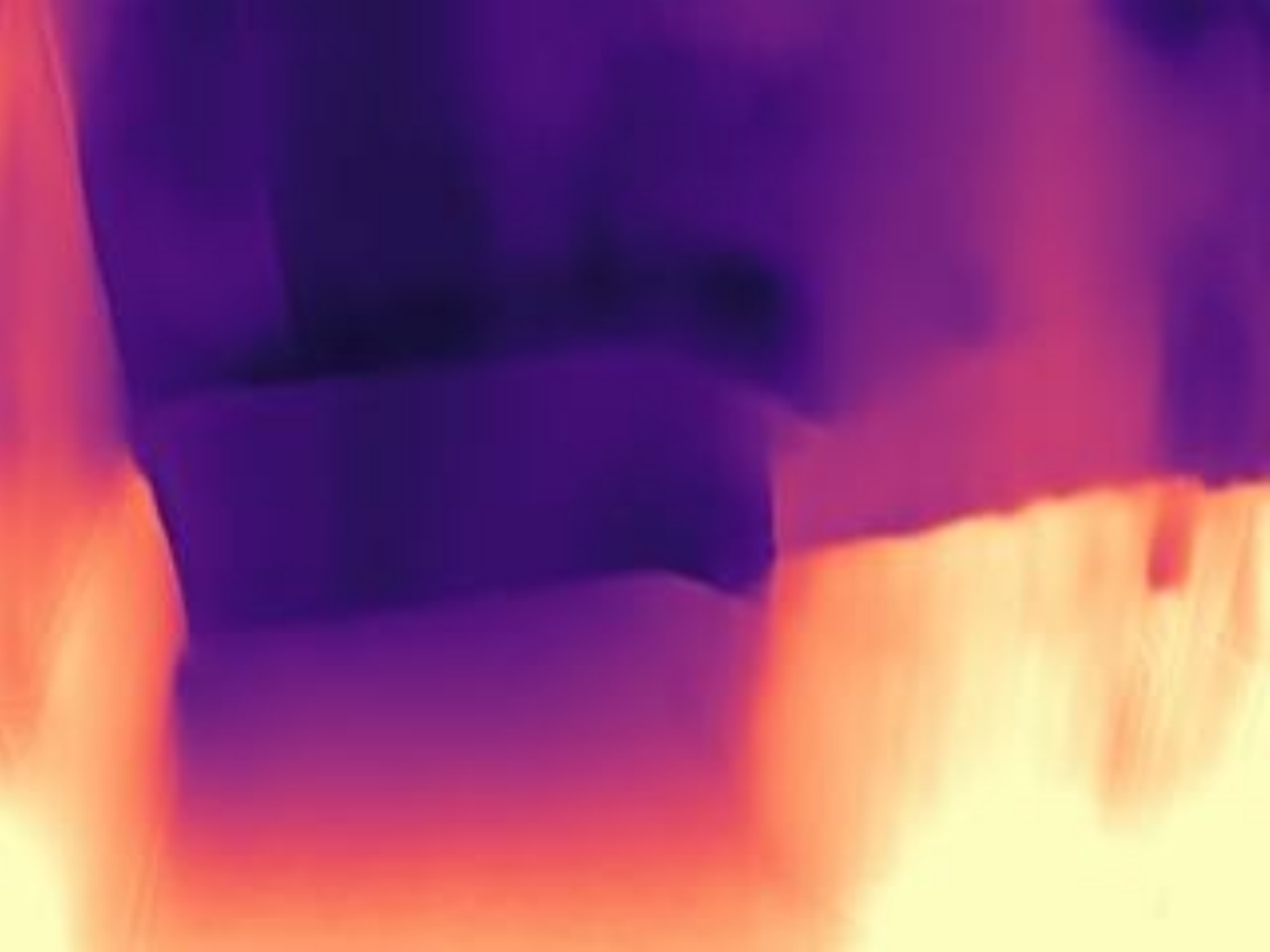} \qquad\qquad\quad &  
\includegraphics[width=\iw,height=\ih]{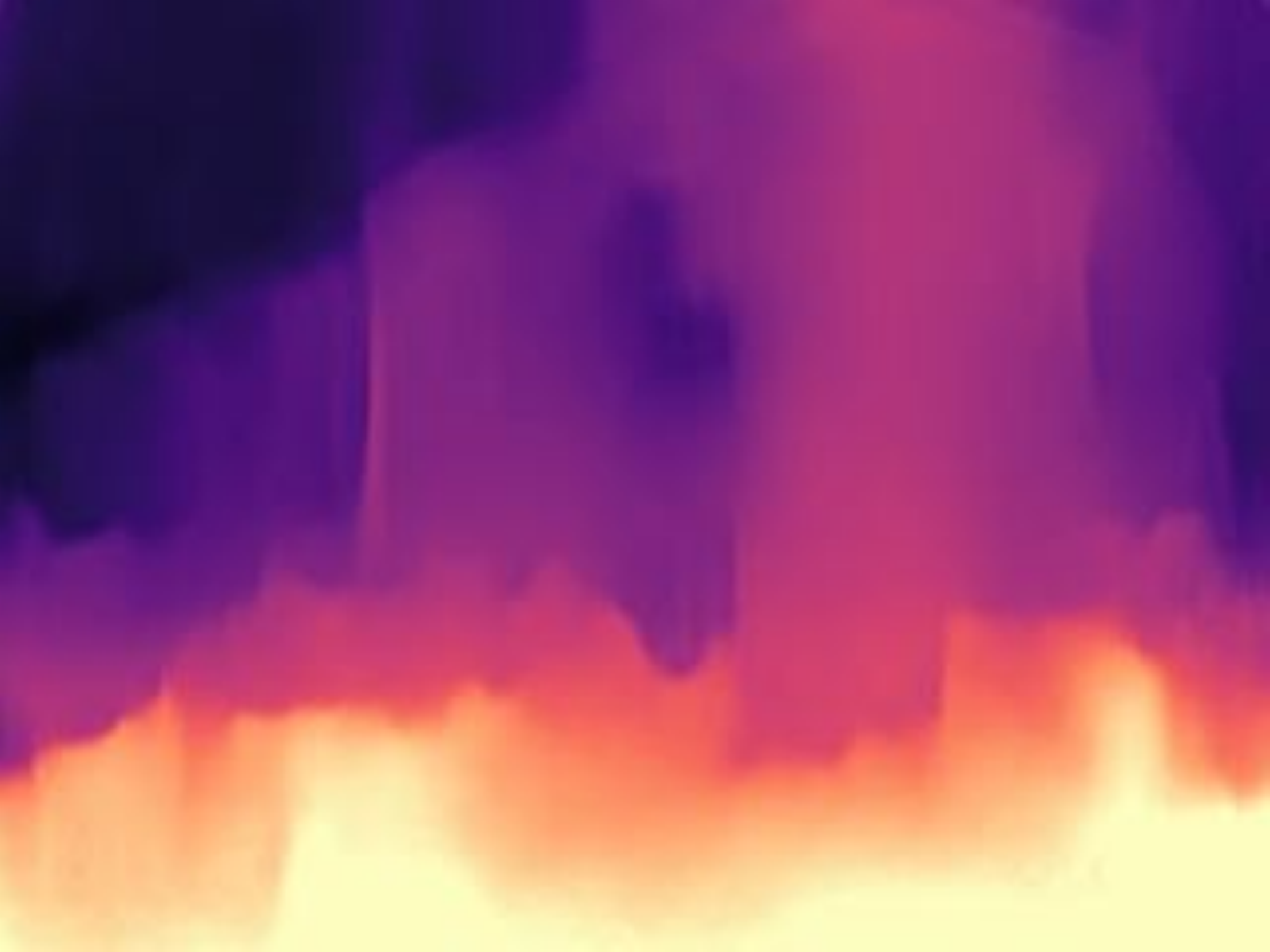} \qquad\qquad\quad &   
\includegraphics[width=\iw,height=\ih]{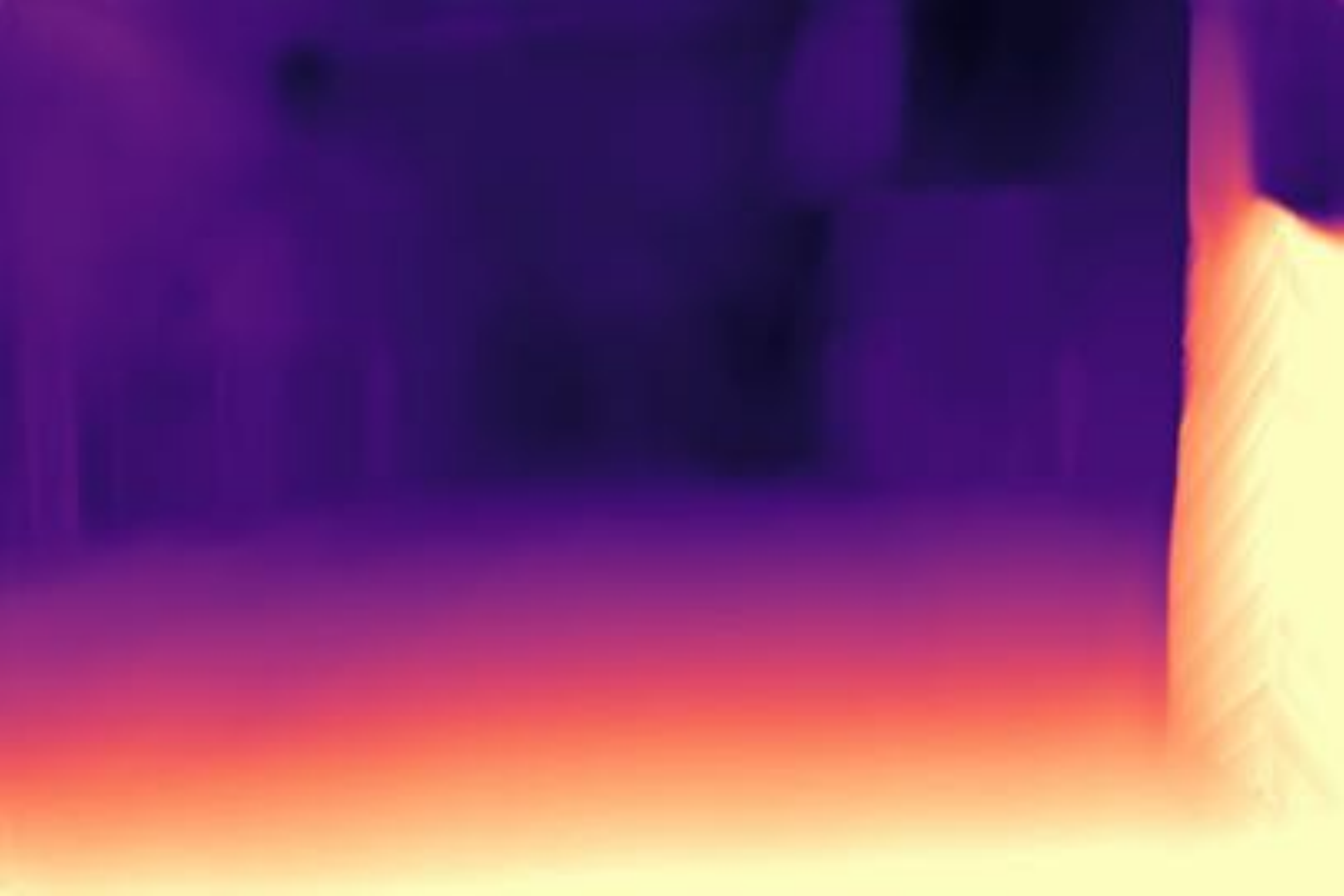} \qquad\qquad\quad & 
\includegraphics[width=\iw,height=\ih]{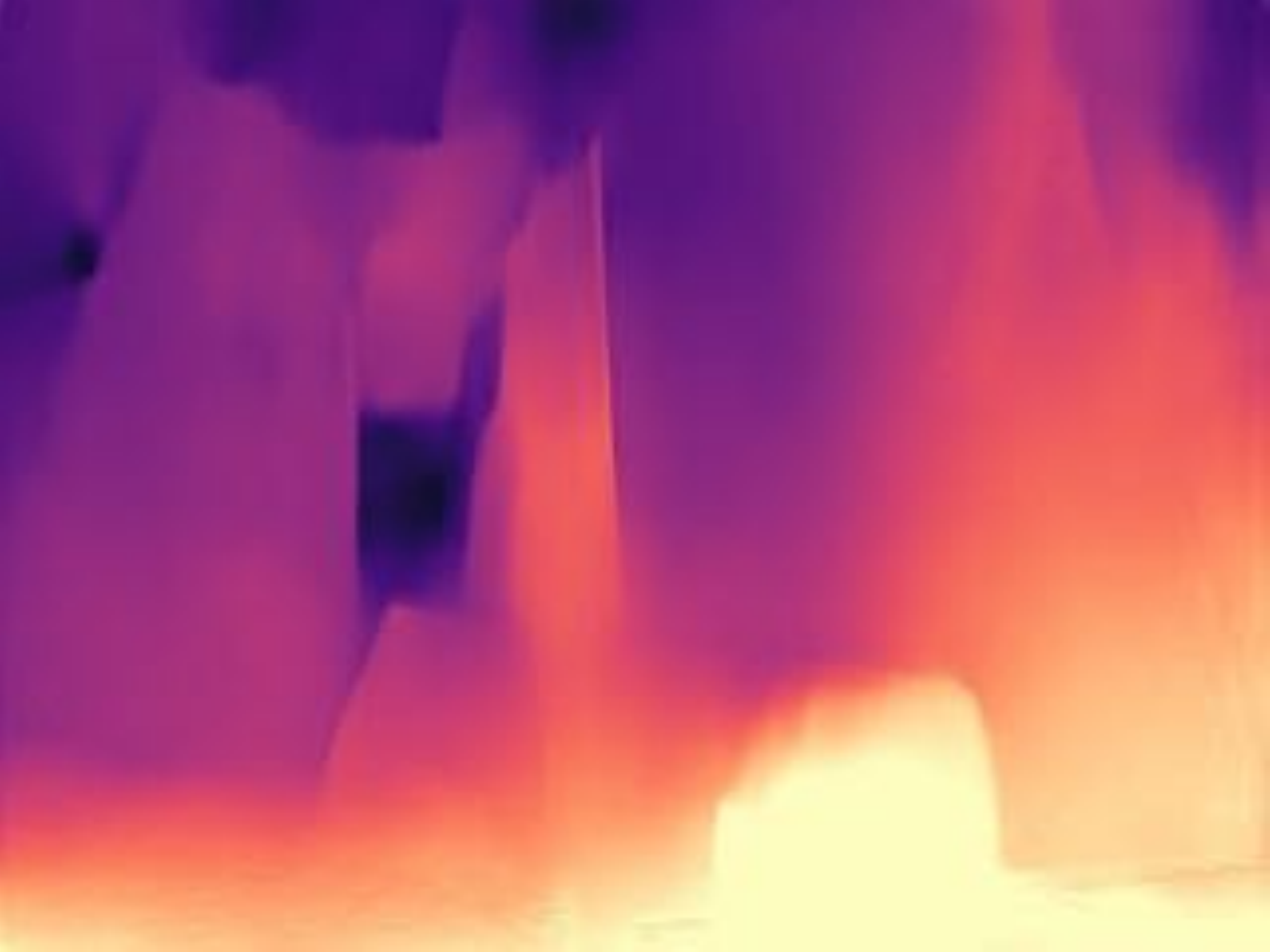} \\
\vspace{10mm}\\
\rotatebox[origin=c]{90}{\fontsize{\textw}{\texth} \selectfont R-MSFM6\hspace{-270mm}}\hspace{24mm}
\includegraphics[width=\iw,height=\ih]{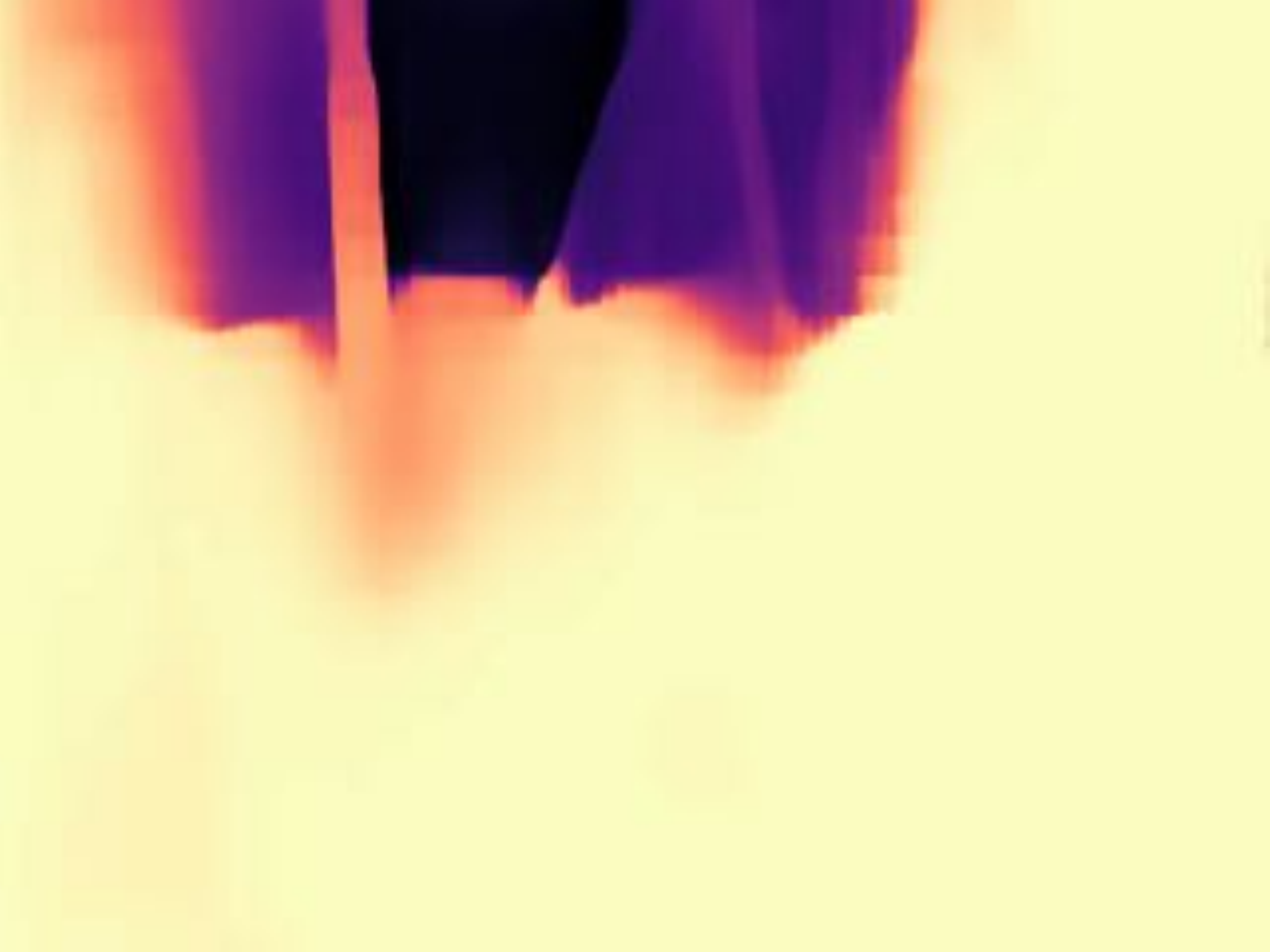} \qquad\qquad\quad & 
\includegraphics[width=\iw,height=\ih]{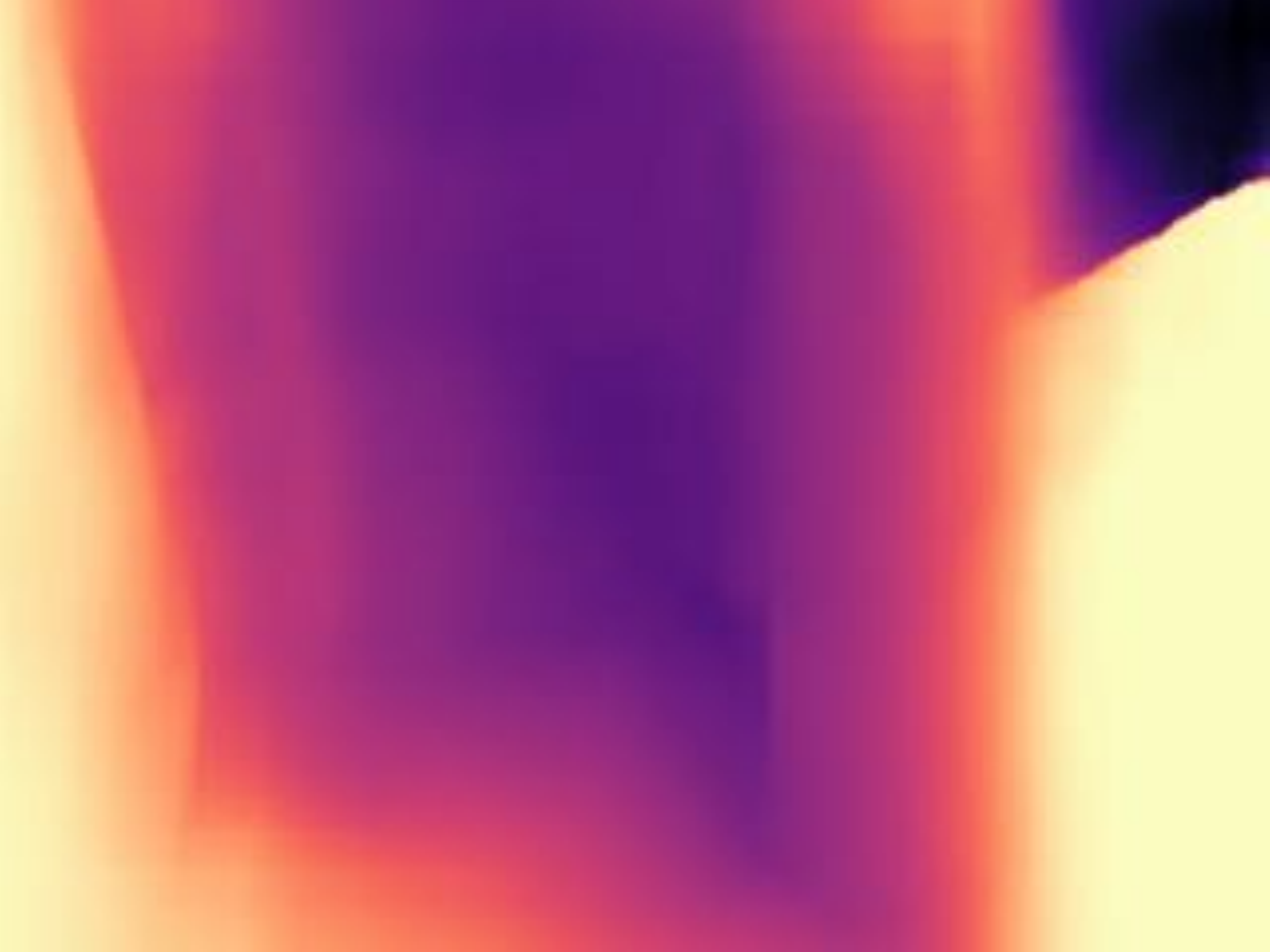} \qquad\qquad\quad & 
\includegraphics[width=\iw,height=\ih]{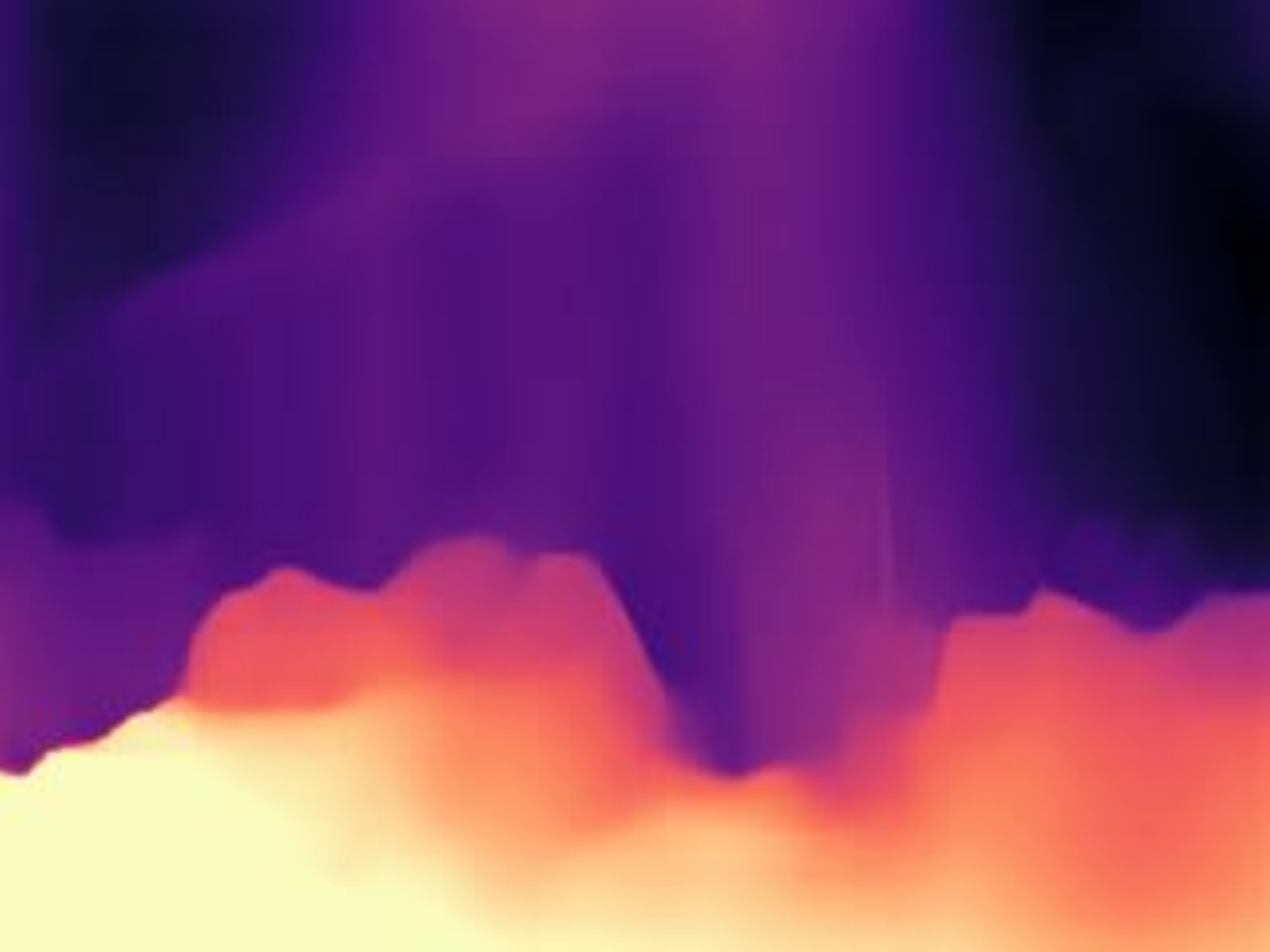} \qquad\qquad\quad &  
\includegraphics[width=\iw,height=\ih]{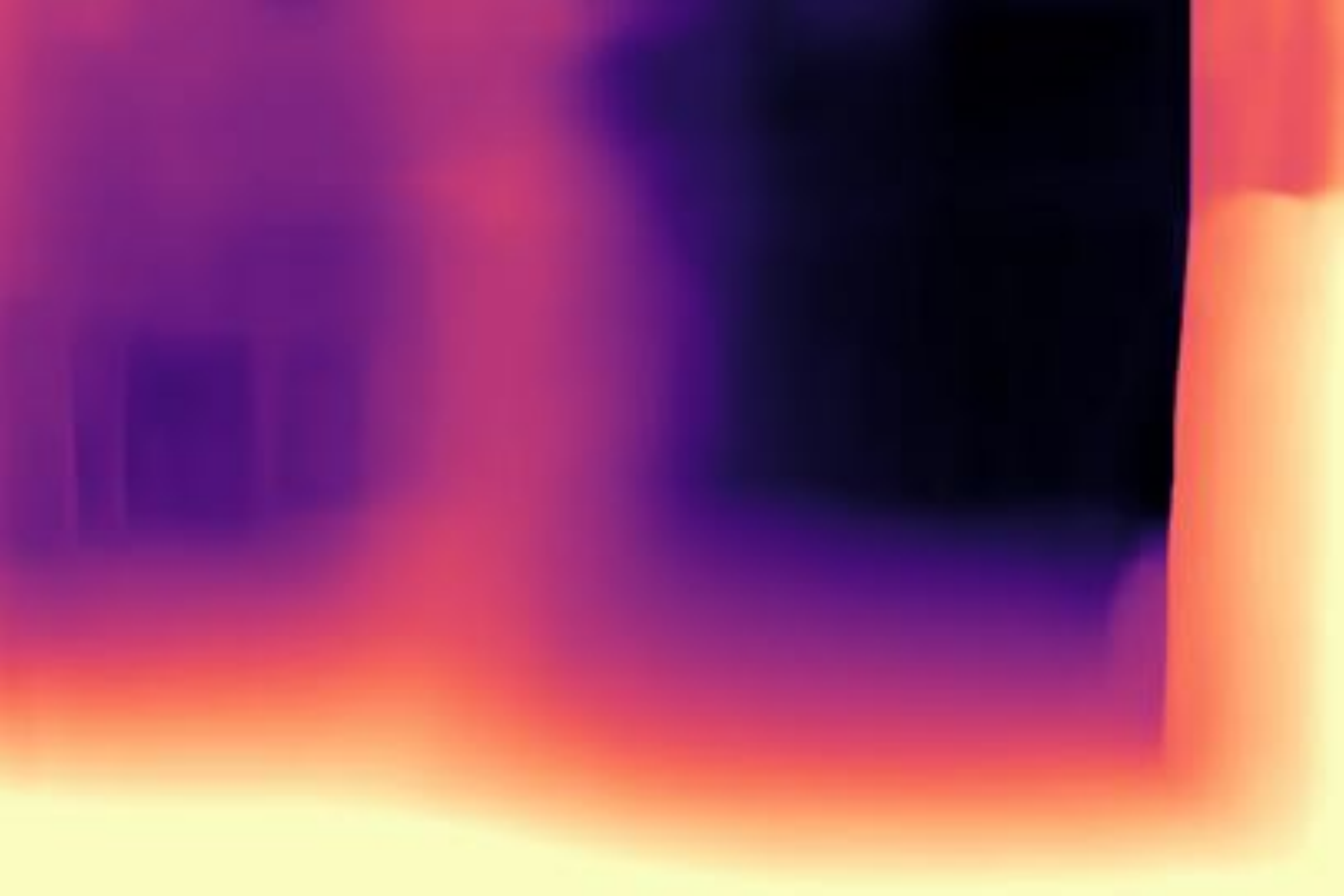} \qquad\qquad\quad &
\includegraphics[width=\iw,height=\ih]{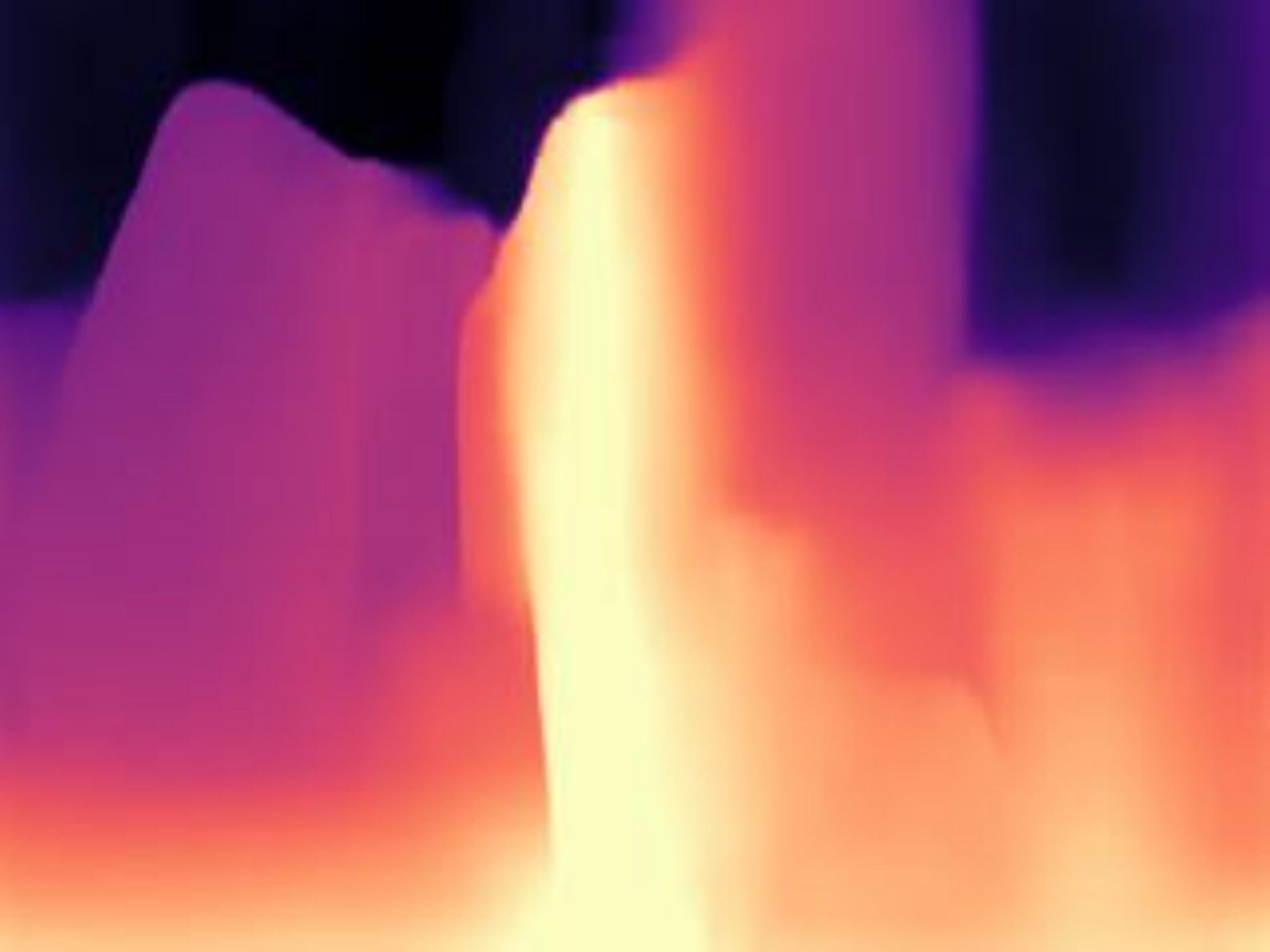} \\
\vspace{10mm}\\
\rotatebox[origin=c]{90}{\fontsize{\textw}{\texth} \selectfont MF-ConvNeXt\hspace{-260mm}}\hspace{24mm}
\includegraphics[width=\iw,height=\ih]{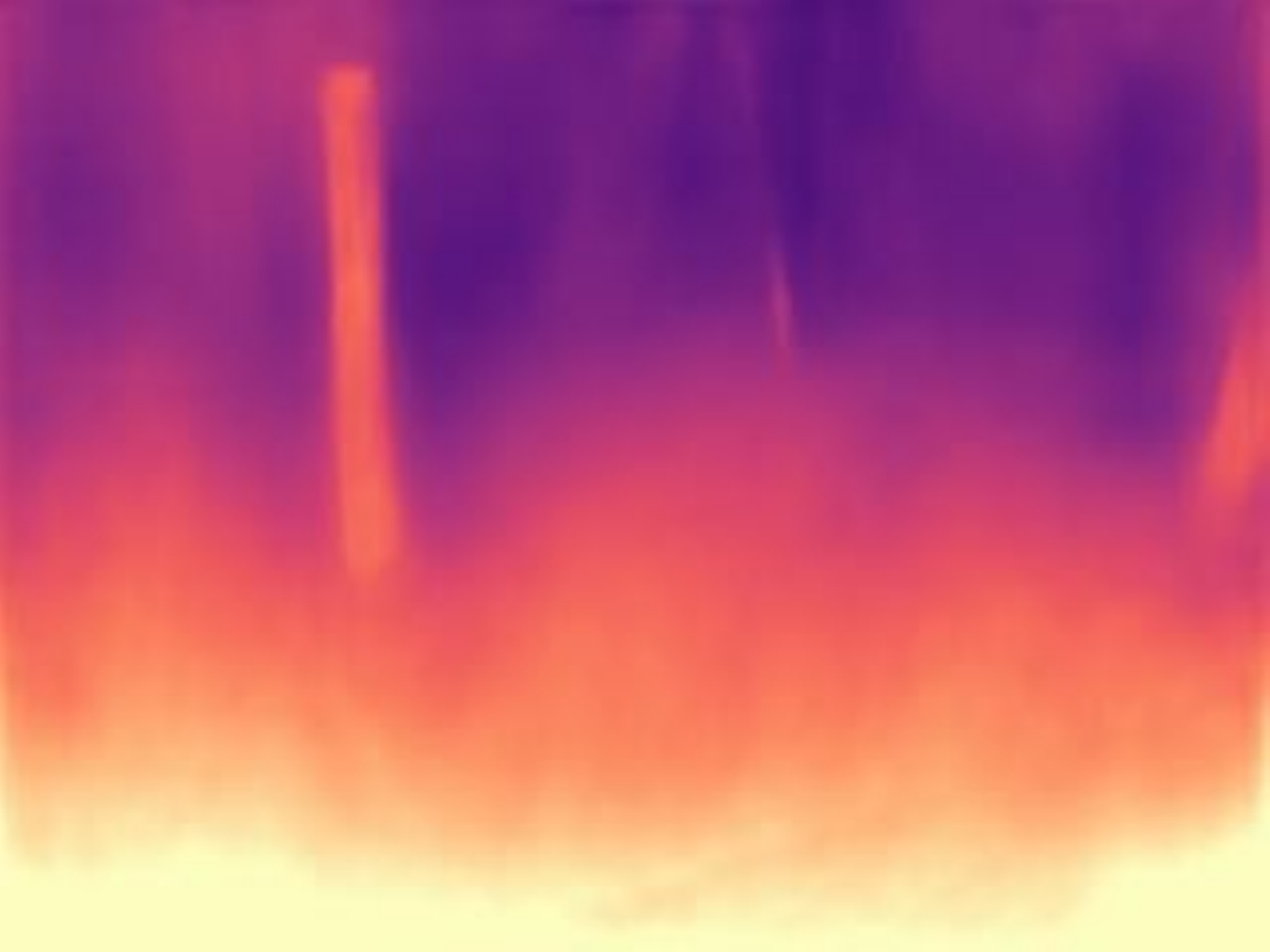} \qquad\qquad\quad &  
\includegraphics[width=\iw,height=\ih]{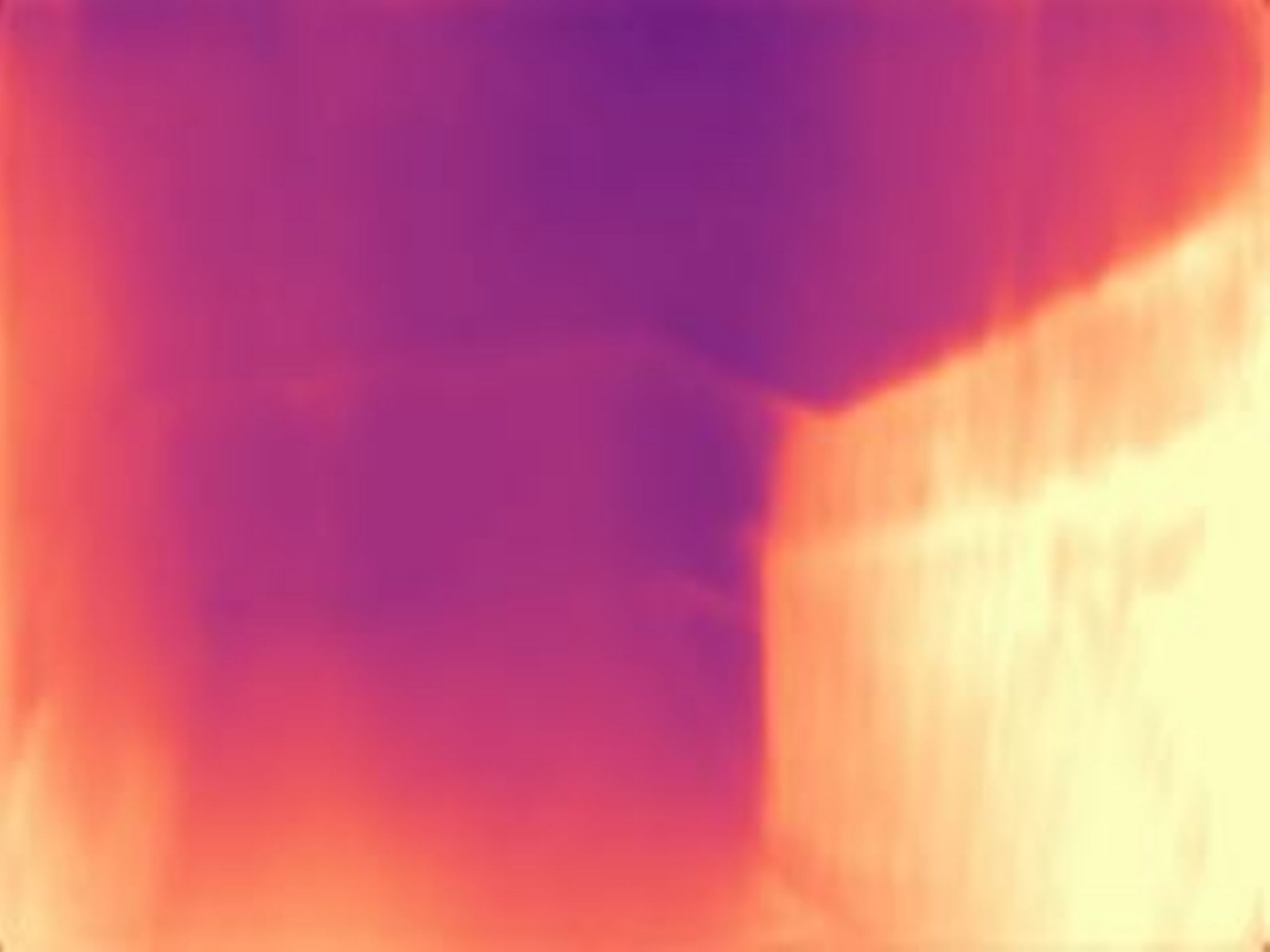} \qquad\qquad\quad & 
\includegraphics[width=\iw,height=\ih]{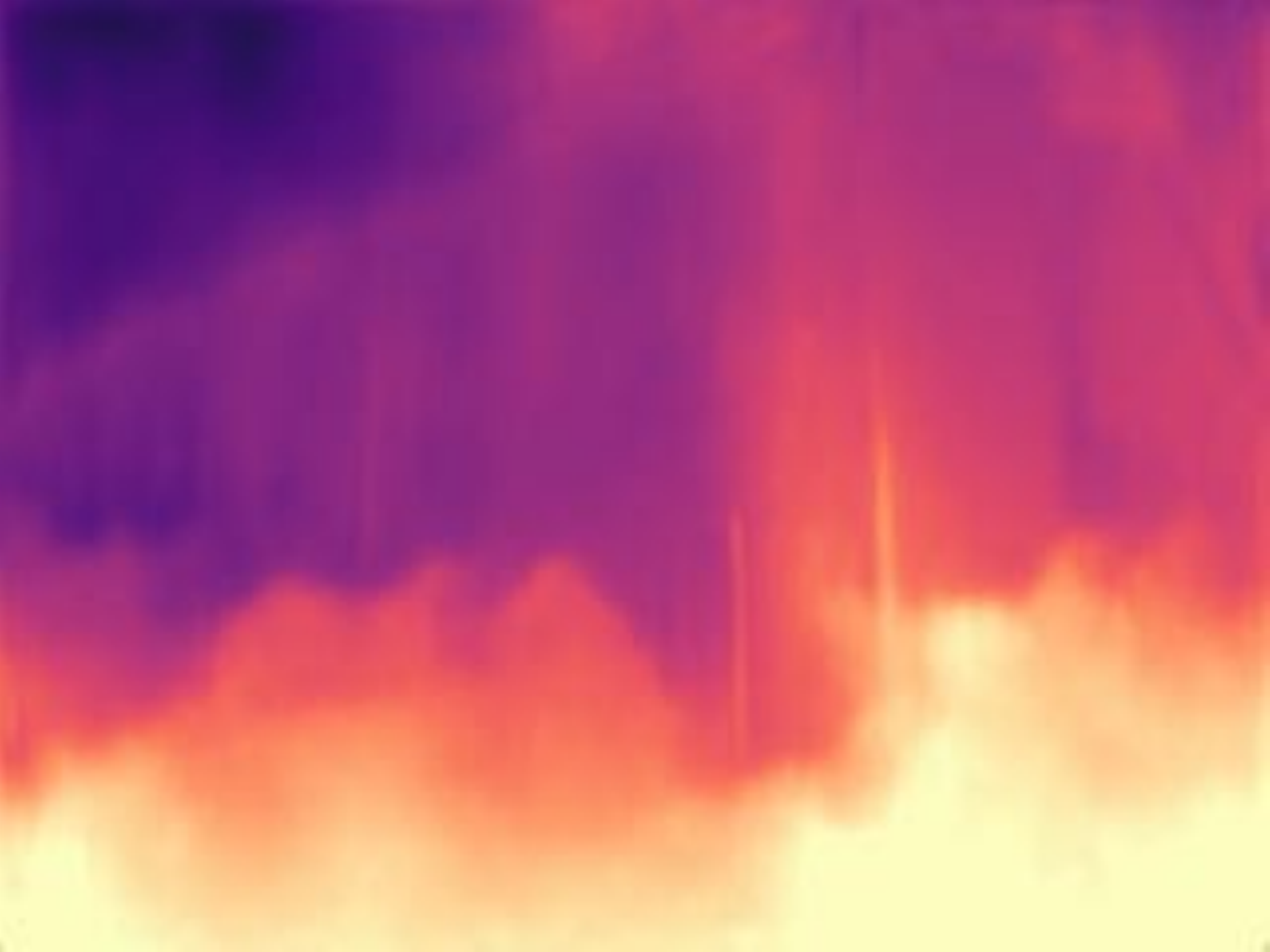} \qquad\qquad\quad &   
\includegraphics[width=\iw,height=\ih]{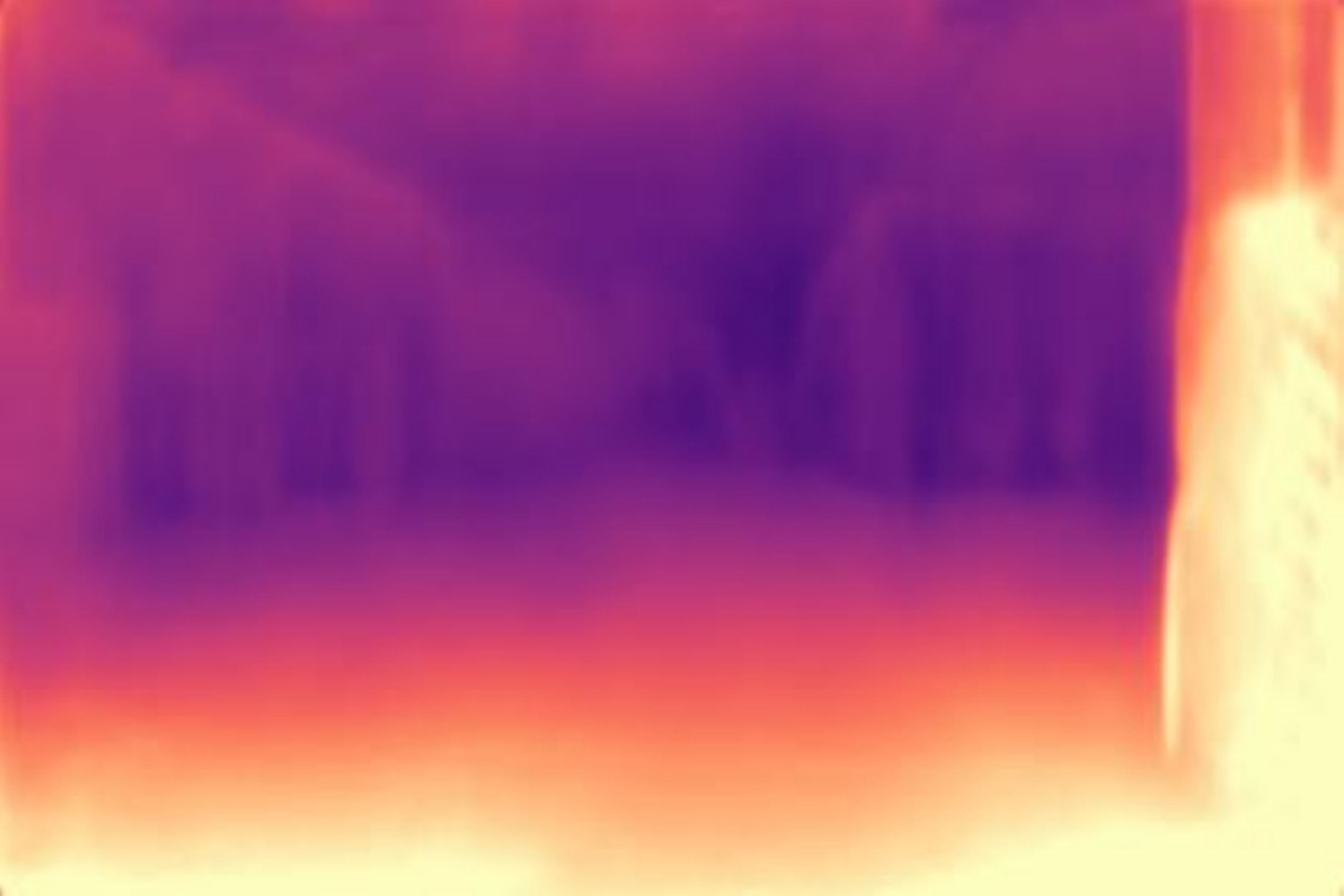} \qquad\qquad\quad & 
\includegraphics[width=\iw,height=\ih]{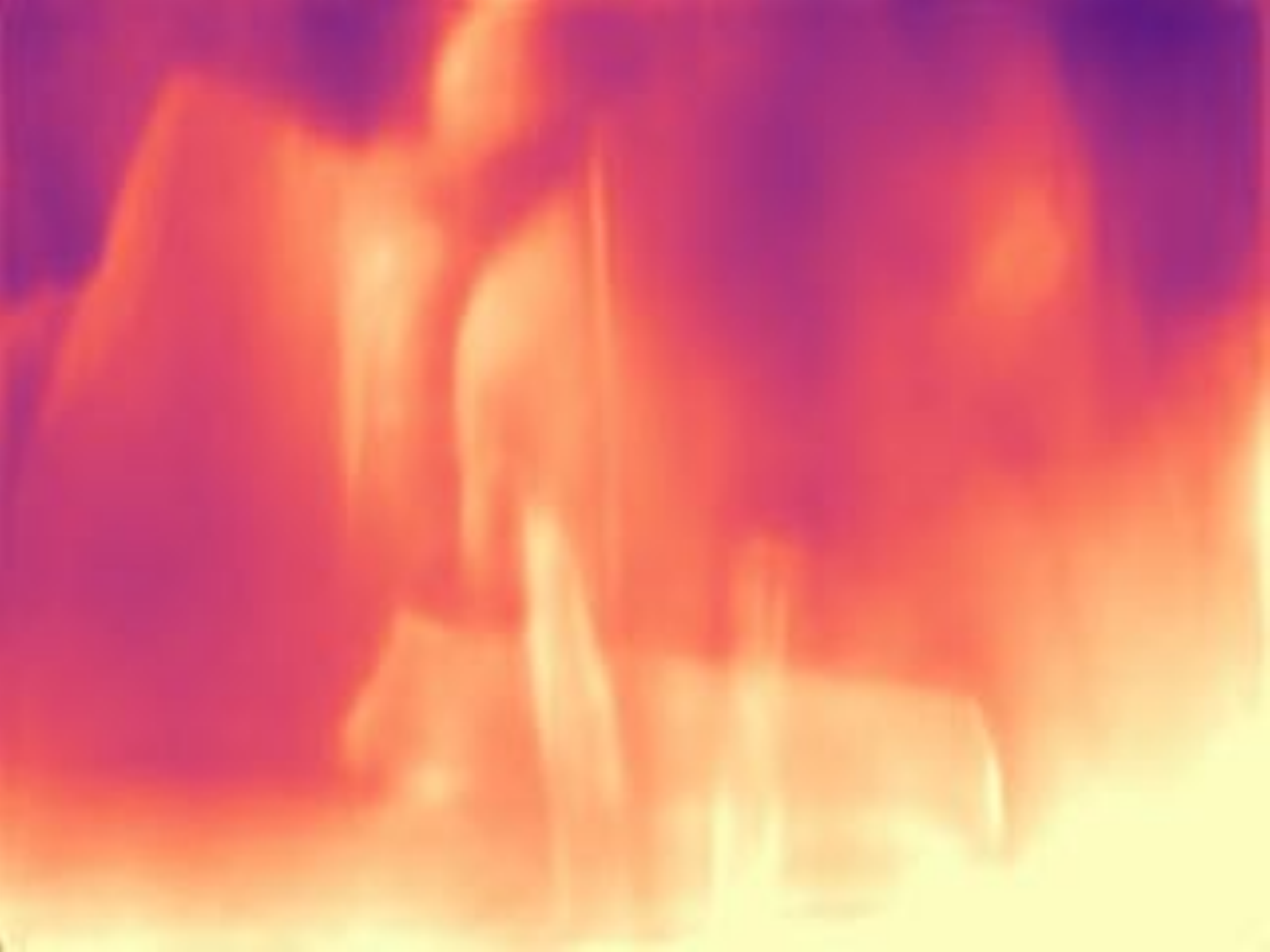} \\
\vspace{10mm}\\
\rotatebox[origin=c]{90}{\fontsize{\textw}{\texth} \selectfont MF-SLaK\hspace{-270mm}}\hspace{24mm}
\includegraphics[width=\iw,height=\ih]{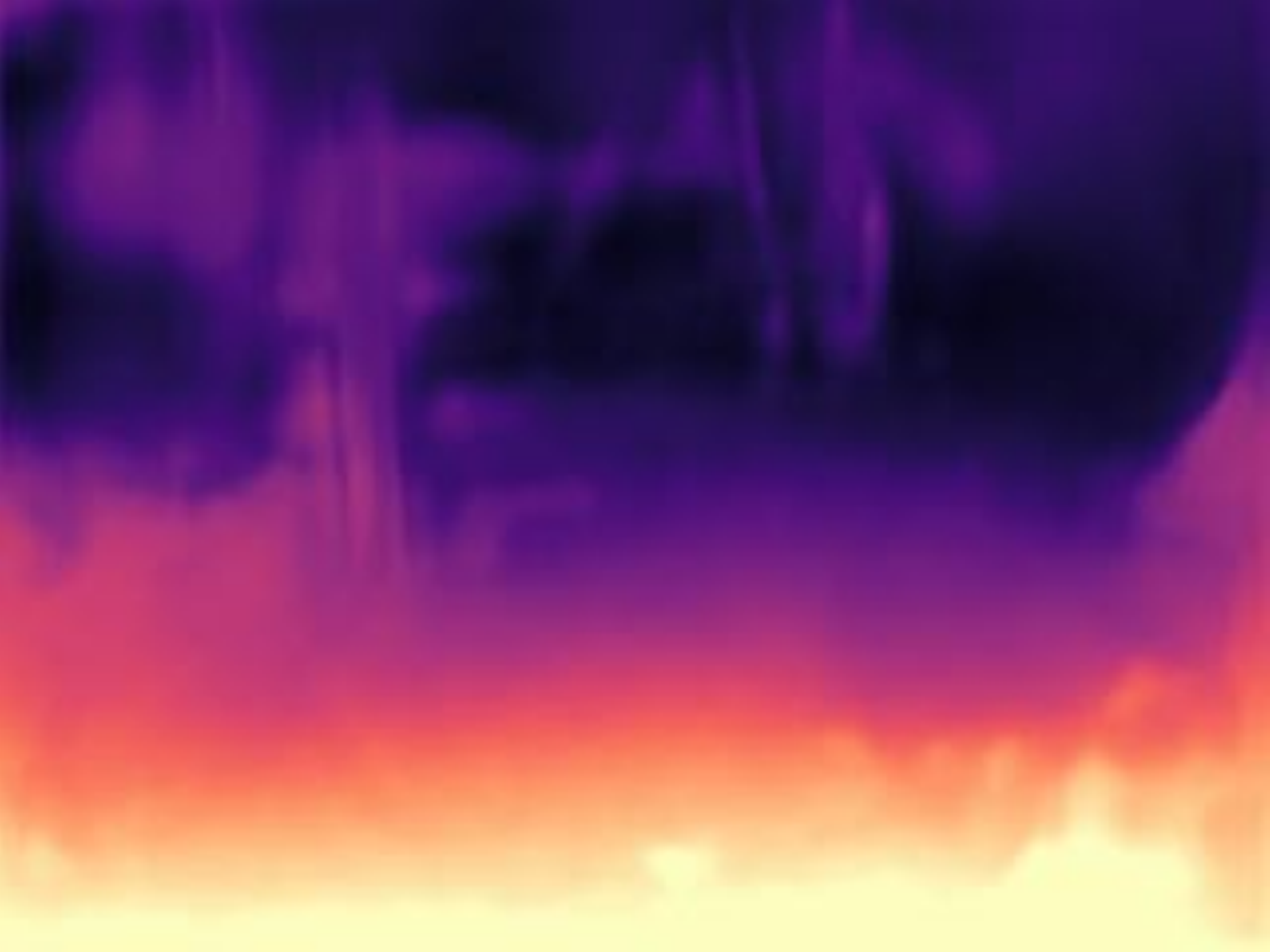} \qquad\qquad\quad & 
\includegraphics[width=\iw,height=\ih]{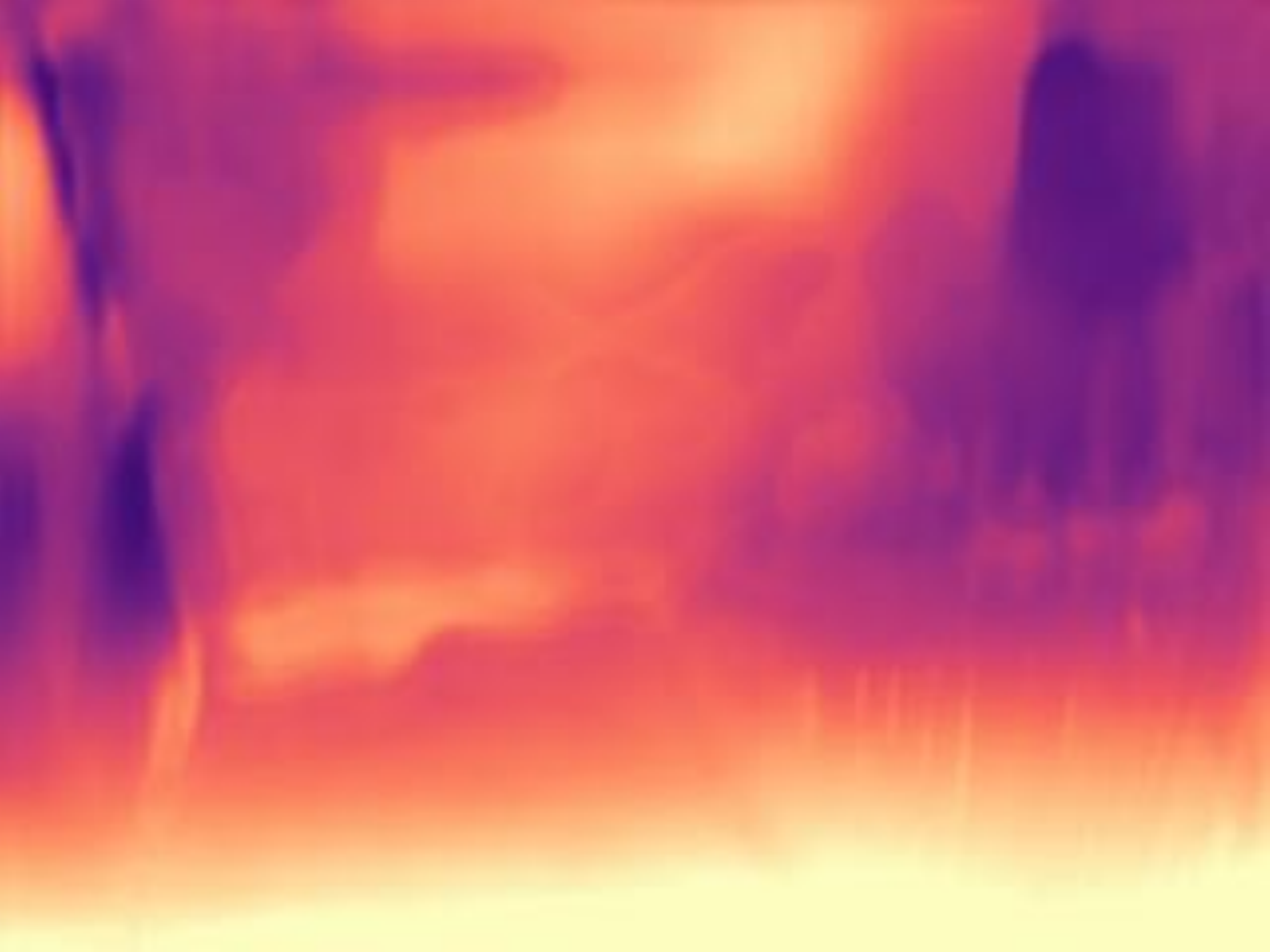} \qquad\qquad\quad & 
\includegraphics[width=\iw,height=\ih]{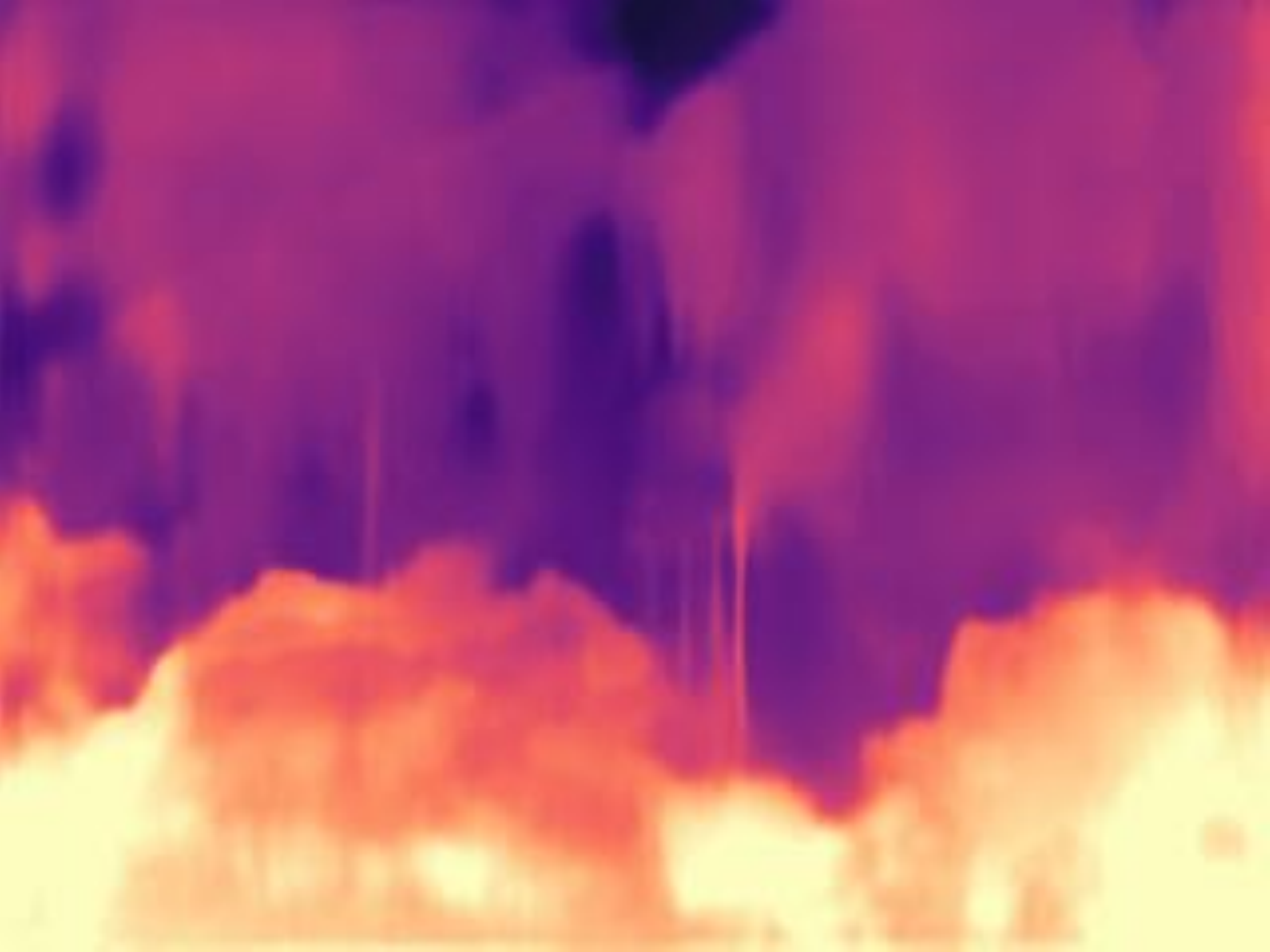} \qquad\qquad\quad &  
\includegraphics[width=\iw,height=\ih]{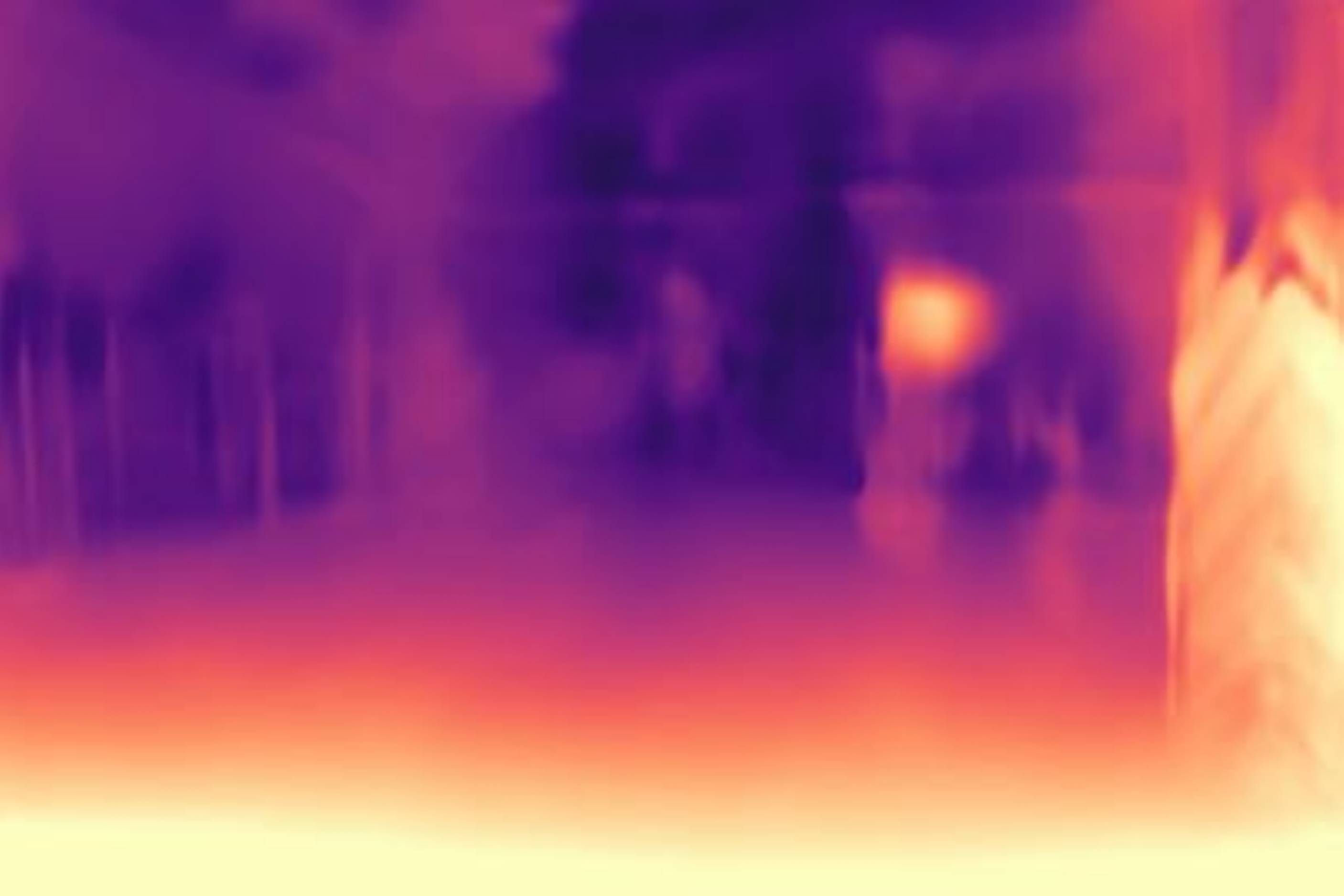} \qquad\qquad\quad &
\includegraphics[width=\iw,height=\ih]{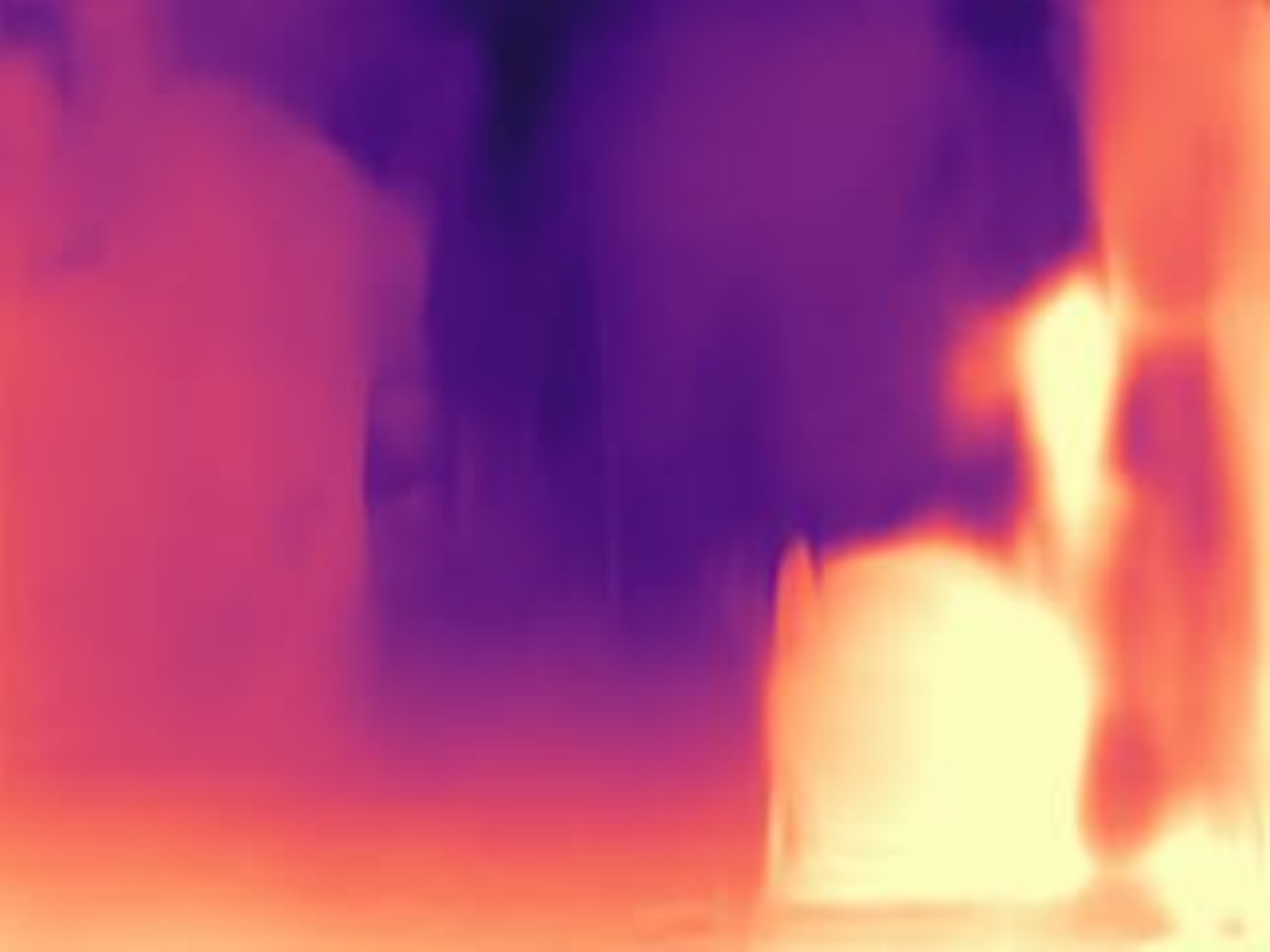} \\
\vspace{30mm}\\
\multicolumn{5}{c}{\fontsize{\w}{\h} \selectfont (a) Self-supervised CNN-based methods } & 
\vspace{30mm}\\
\rotatebox[origin=c]{90}{\fontsize{\textw}{\texth} \selectfont MF-ViT\hspace{-270mm}}\hspace{24mm}
\includegraphics[width=\iw,height=\ih]{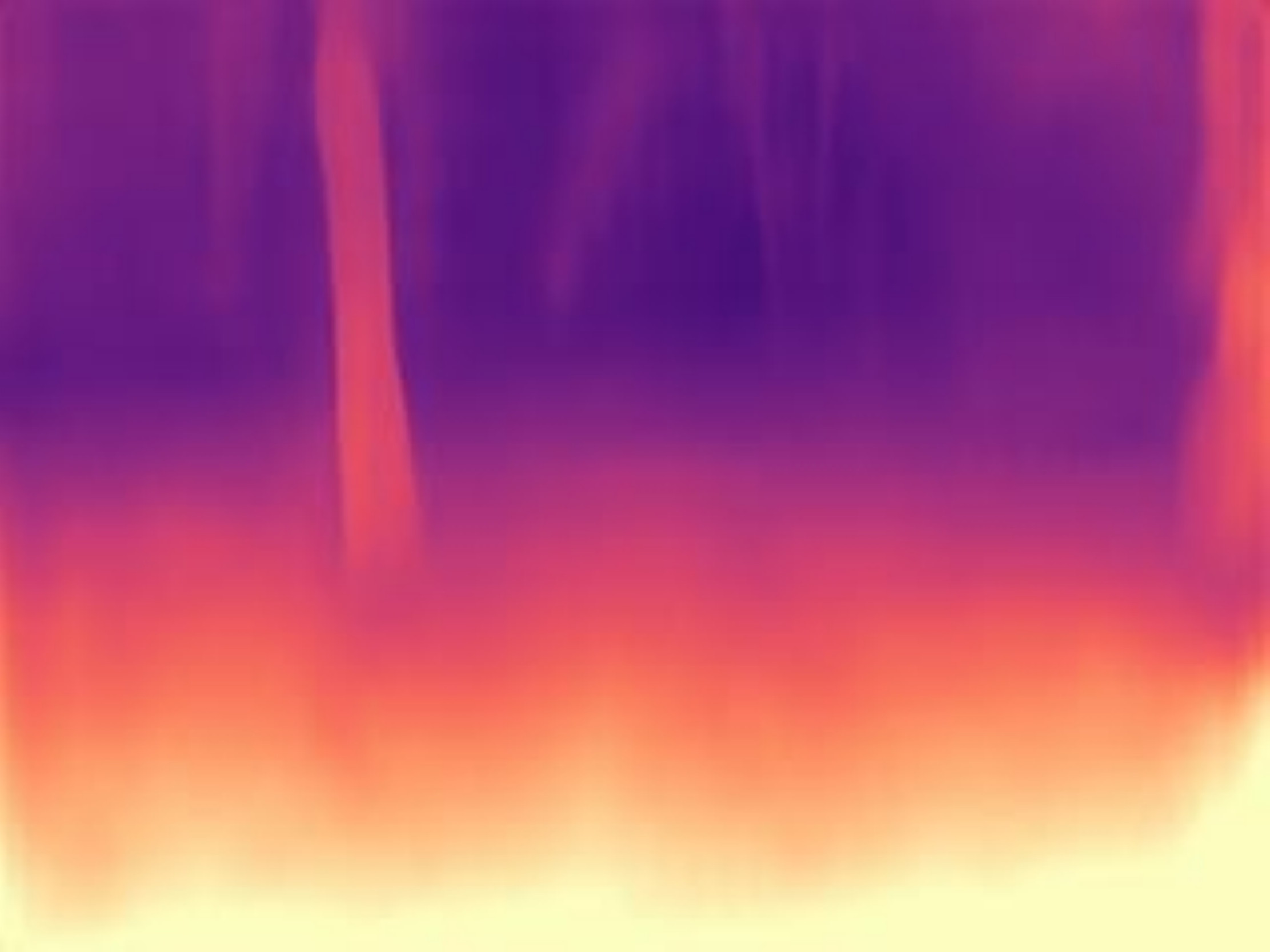} \qquad\qquad\quad & 
\includegraphics[width=\iw,height=\ih]{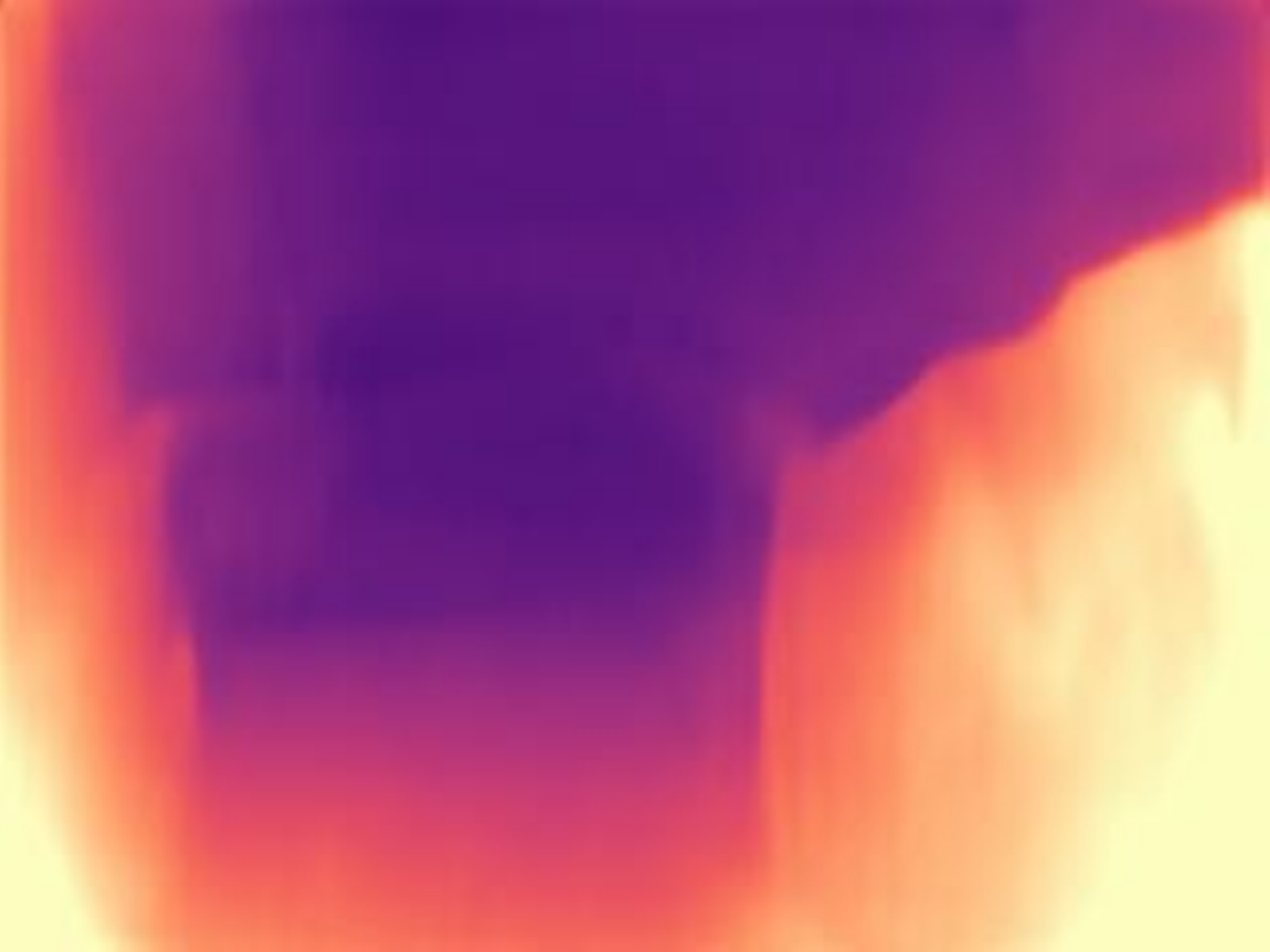} \qquad\qquad\quad & 
\includegraphics[width=\iw,height=\ih]{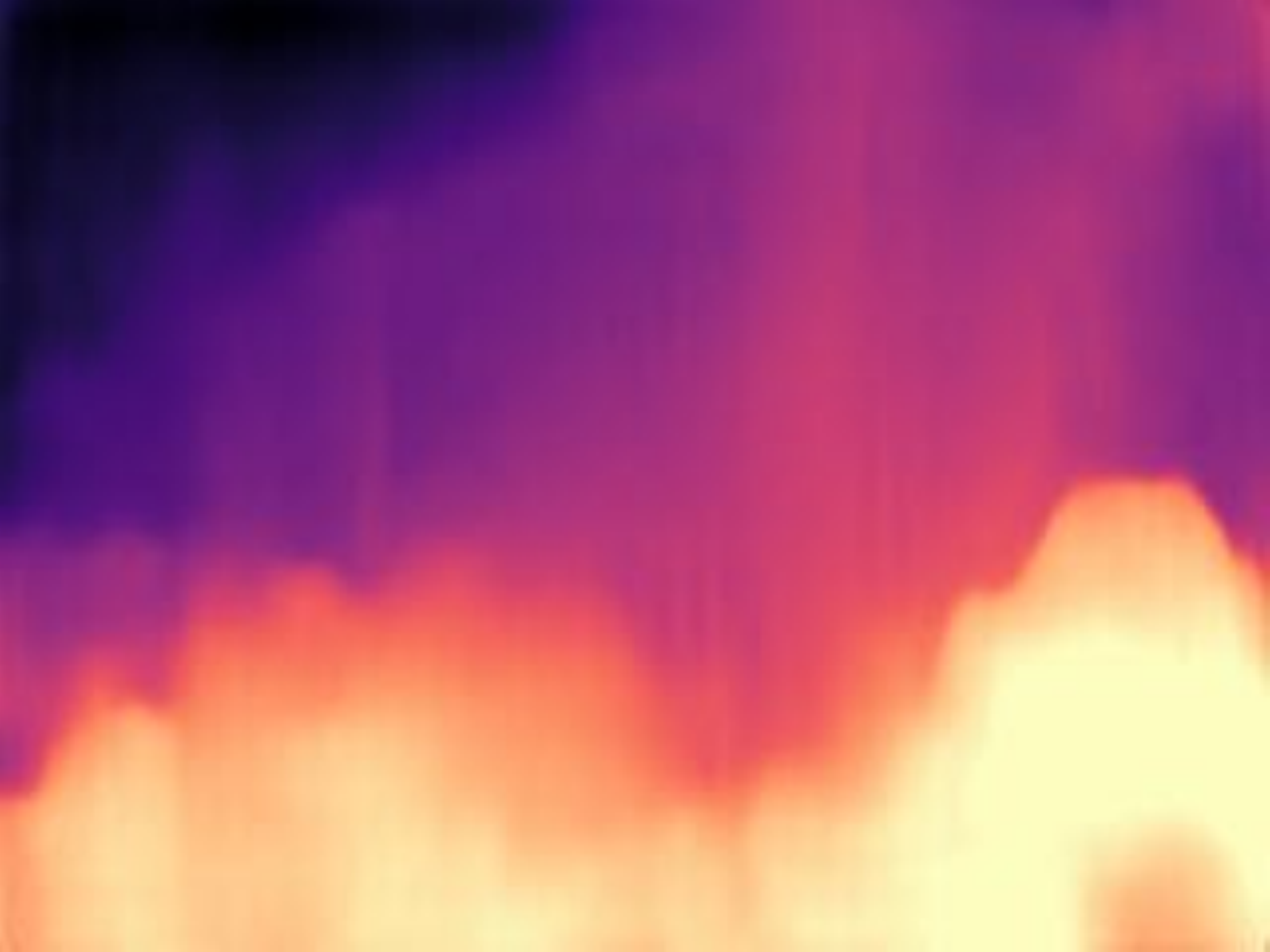} \qquad\qquad\quad &  
\includegraphics[width=\iw,height=\ih]{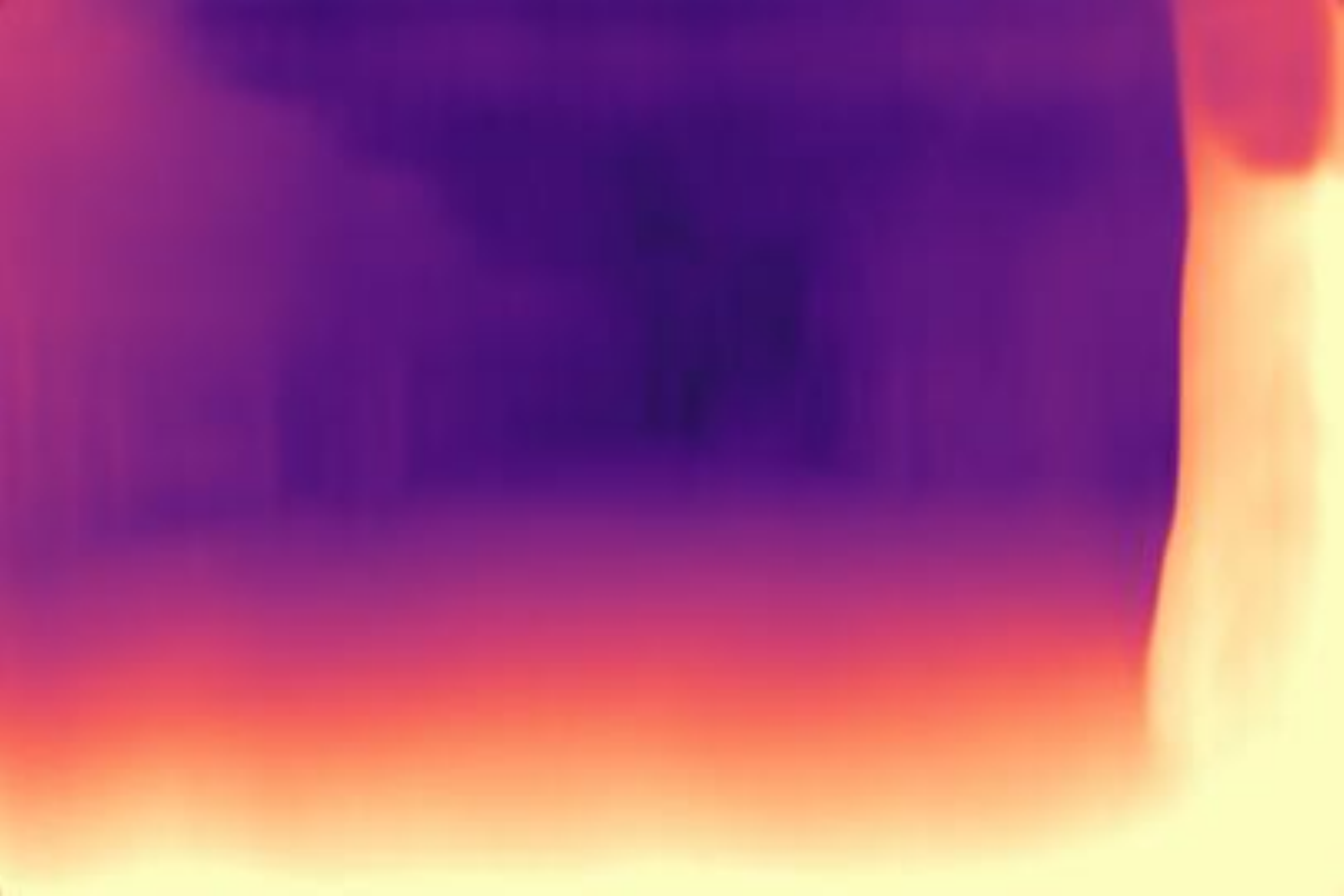} \qquad\qquad\quad &
\includegraphics[width=\iw,height=\ih]{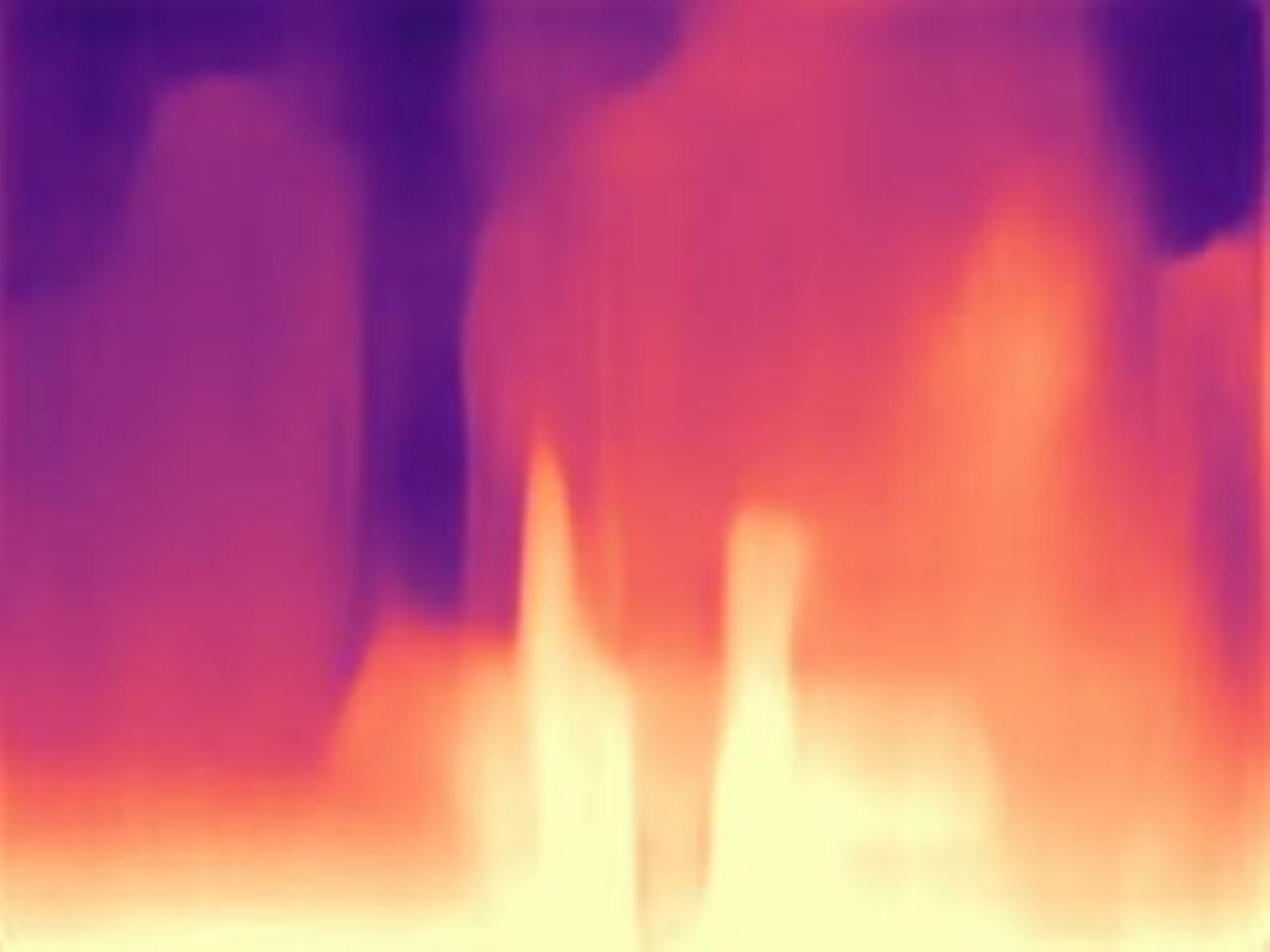} \\
\vspace{10mm}\\
\rotatebox[origin=c]{90}{\fontsize{\textw}{\texth} \selectfont MF-RegionViT\hspace{-270mm}}\hspace{24mm}
\includegraphics[width=\iw,height=\ih]{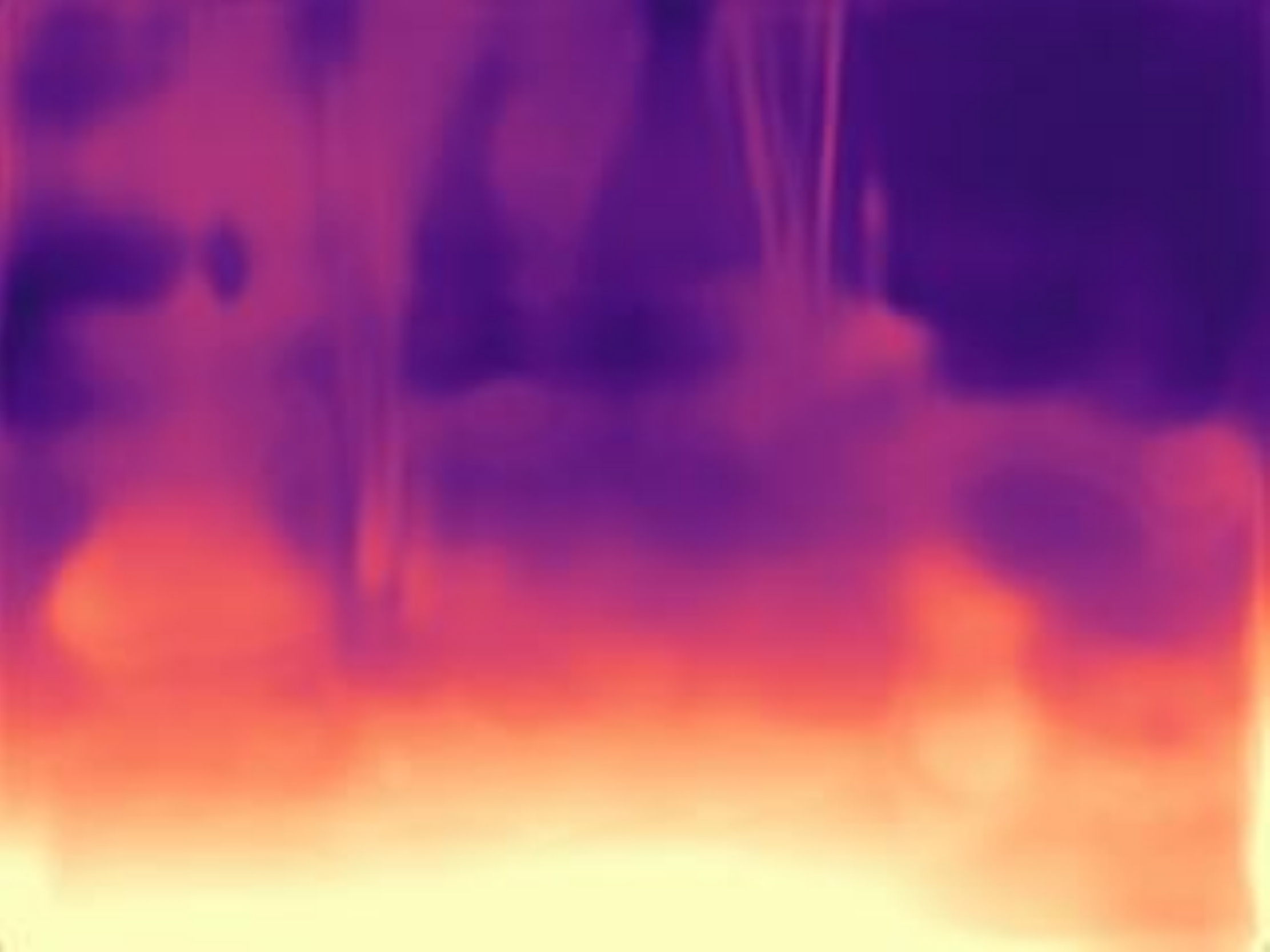} \qquad\qquad\quad & 
\includegraphics[width=\iw,height=\ih]{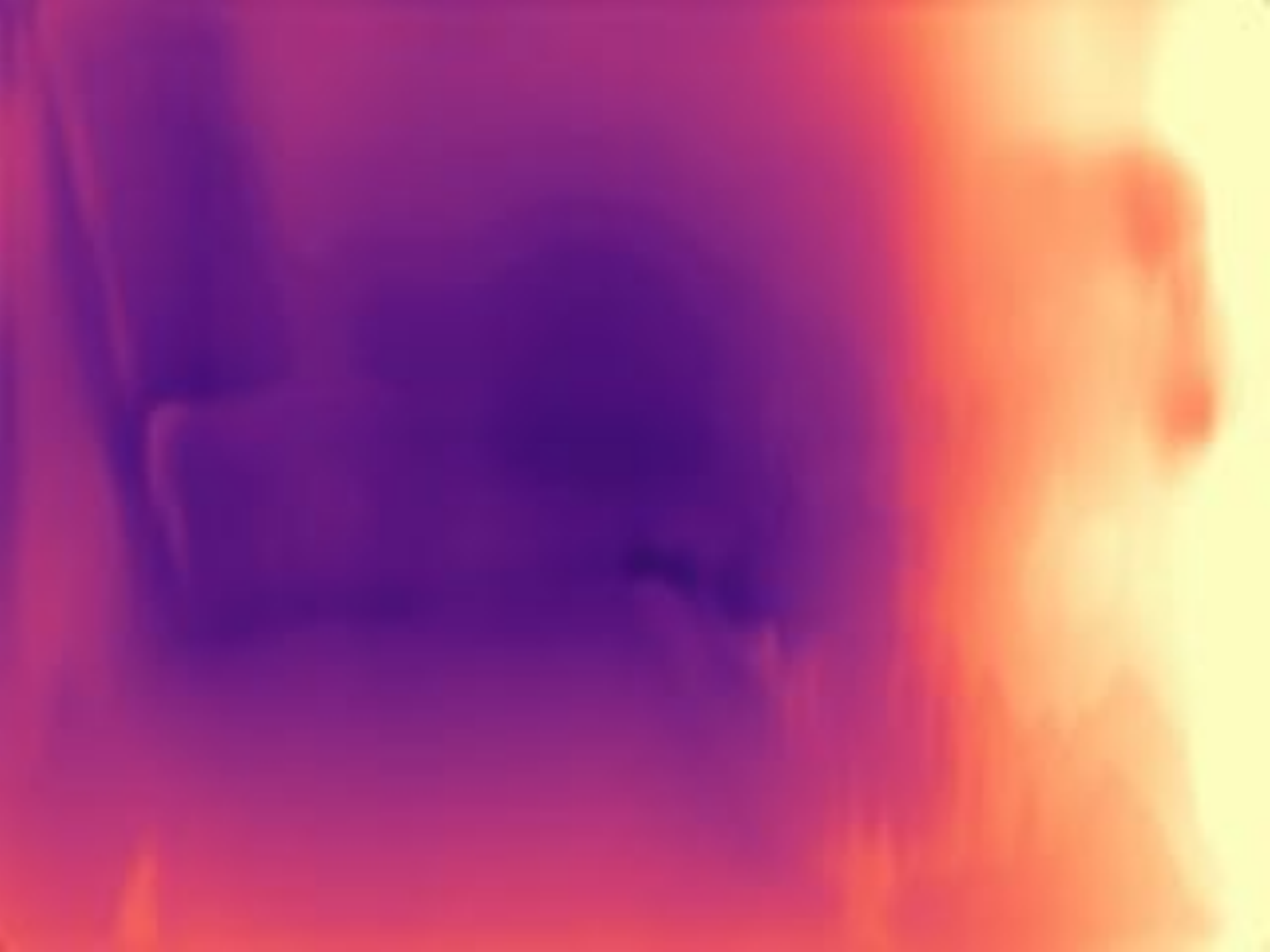} \qquad\qquad\quad & 
\includegraphics[width=\iw,height=\ih]{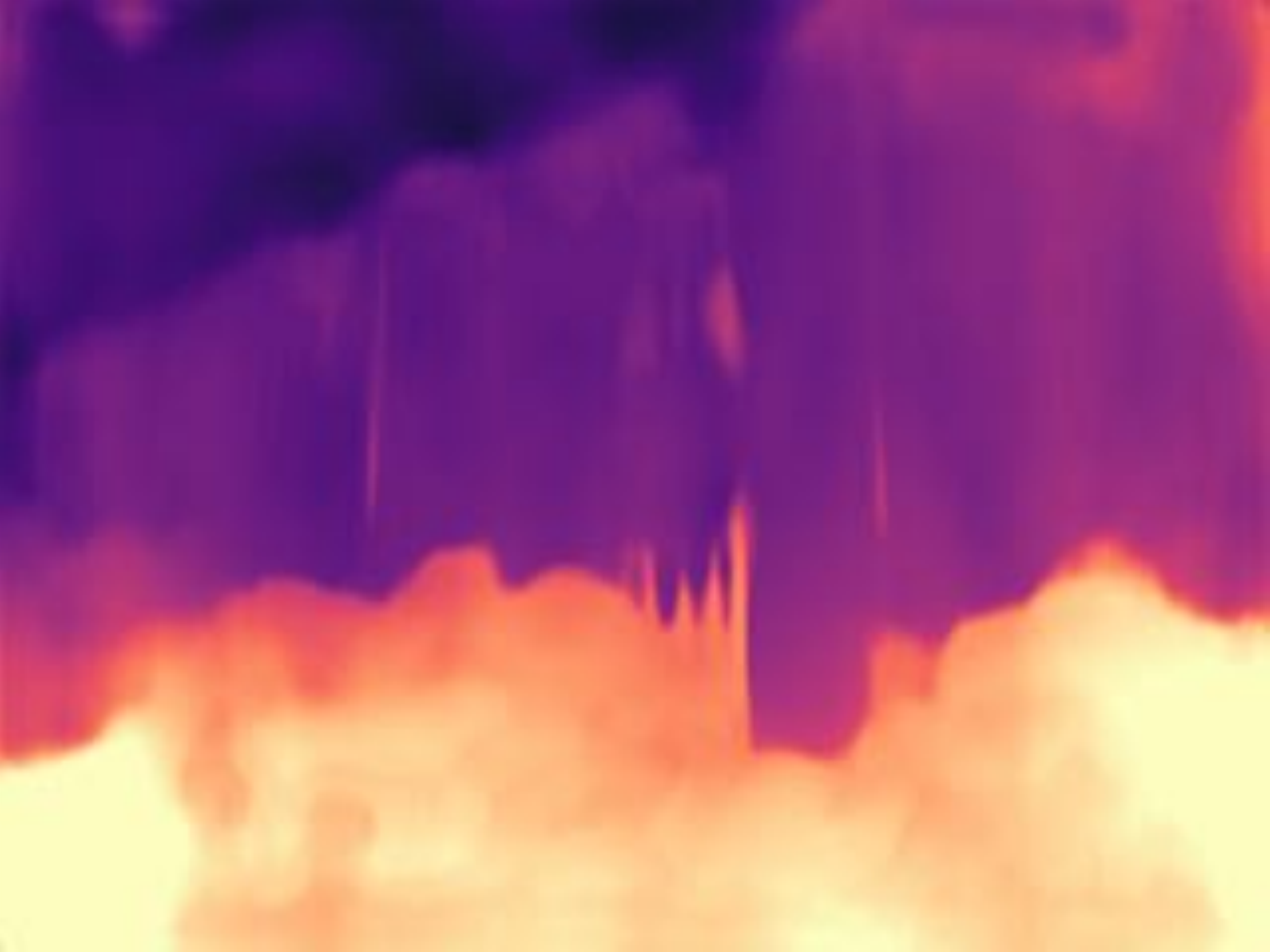} \qquad\qquad\quad &  
\includegraphics[width=\iw,height=\ih]{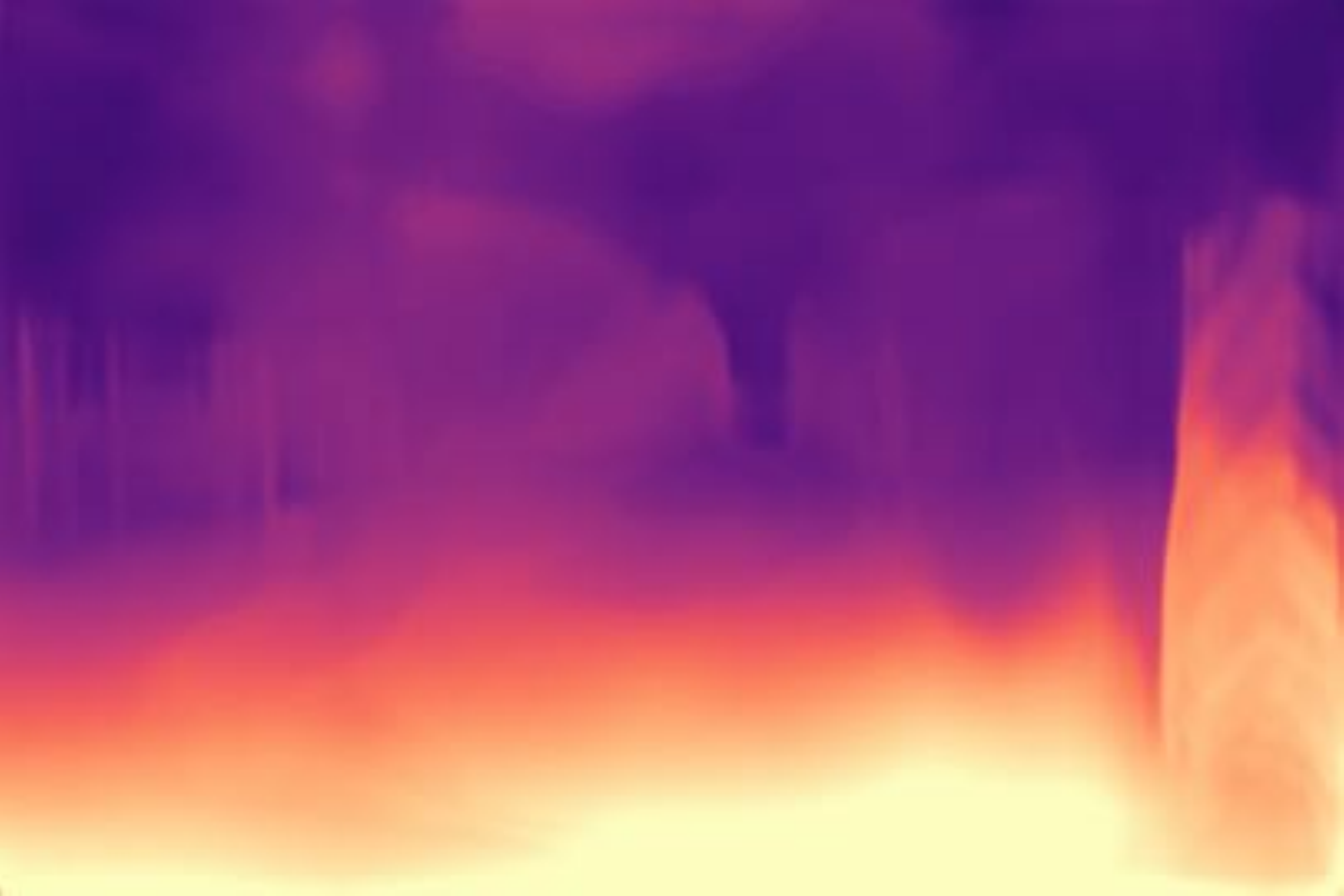}  \qquad\qquad\quad &
\includegraphics[width=\iw,height=\ih]{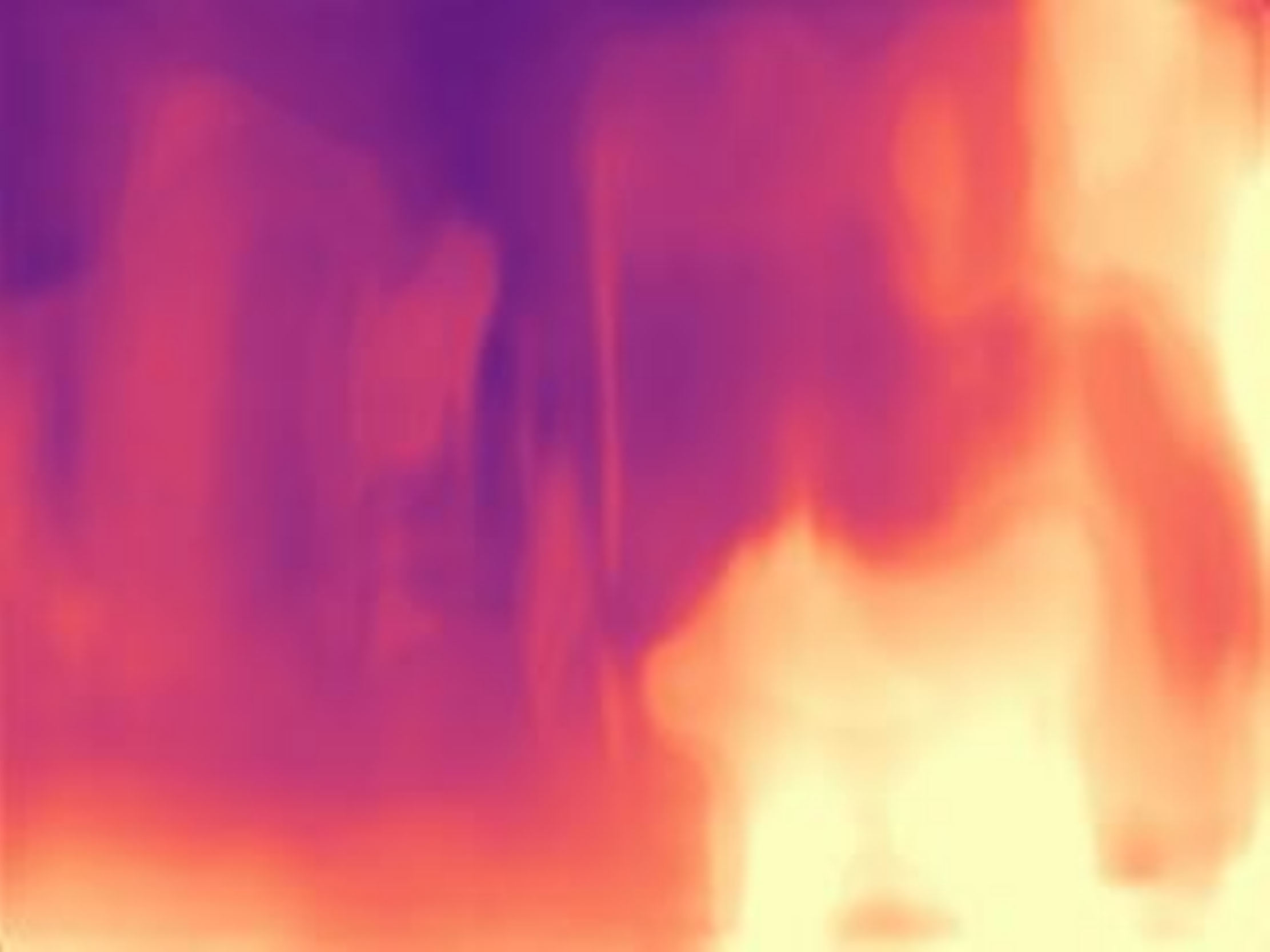} \\
\vspace{10mm}\\
\rotatebox[origin=c]{90}{\fontsize{\textw}{\texth} \selectfont MF-Twins\hspace{-270mm}}\hspace{24mm}
\includegraphics[width=\iw,height=\ih]{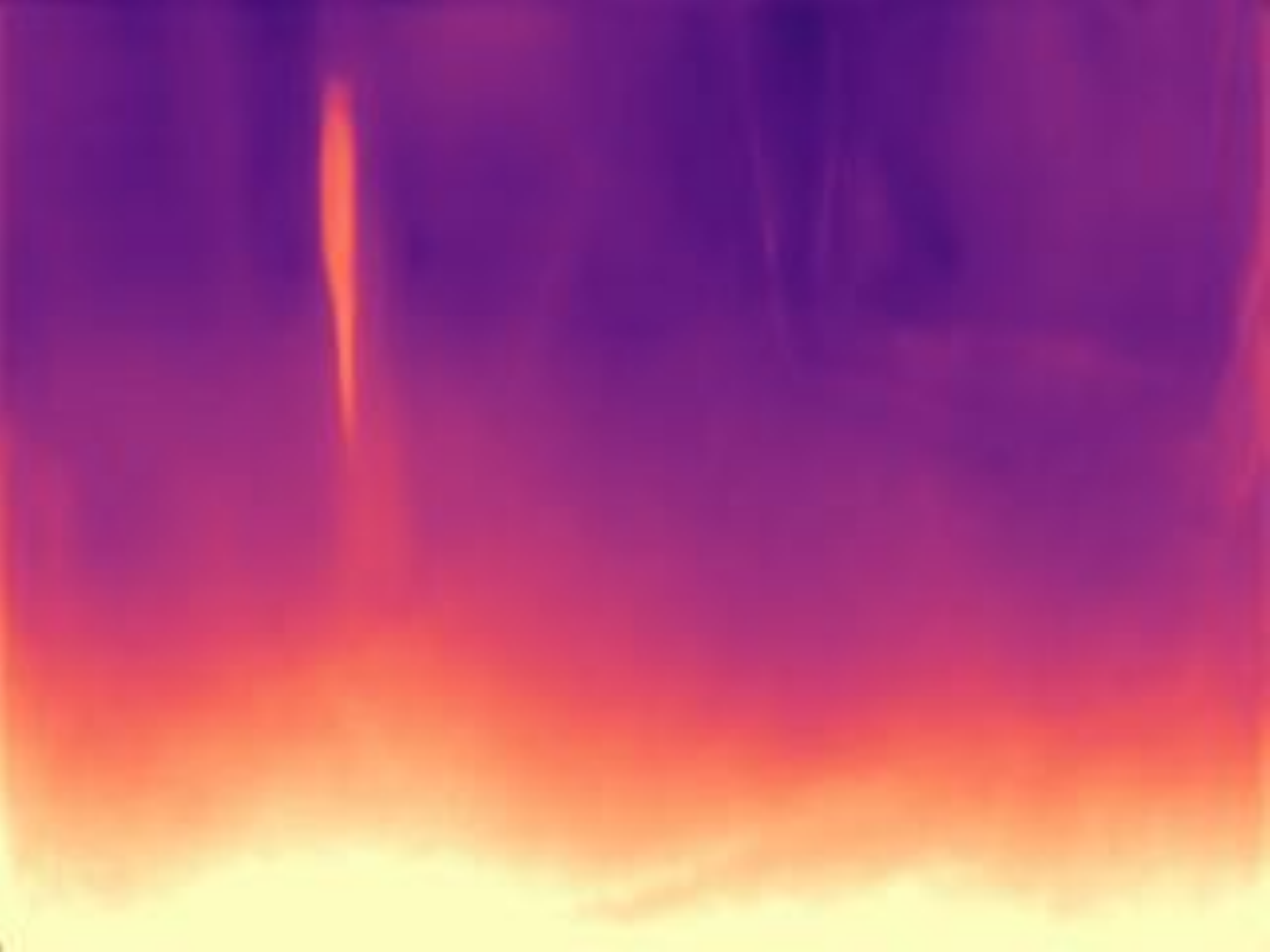} \qquad\qquad\quad & 
\includegraphics[width=\iw,height=\ih]{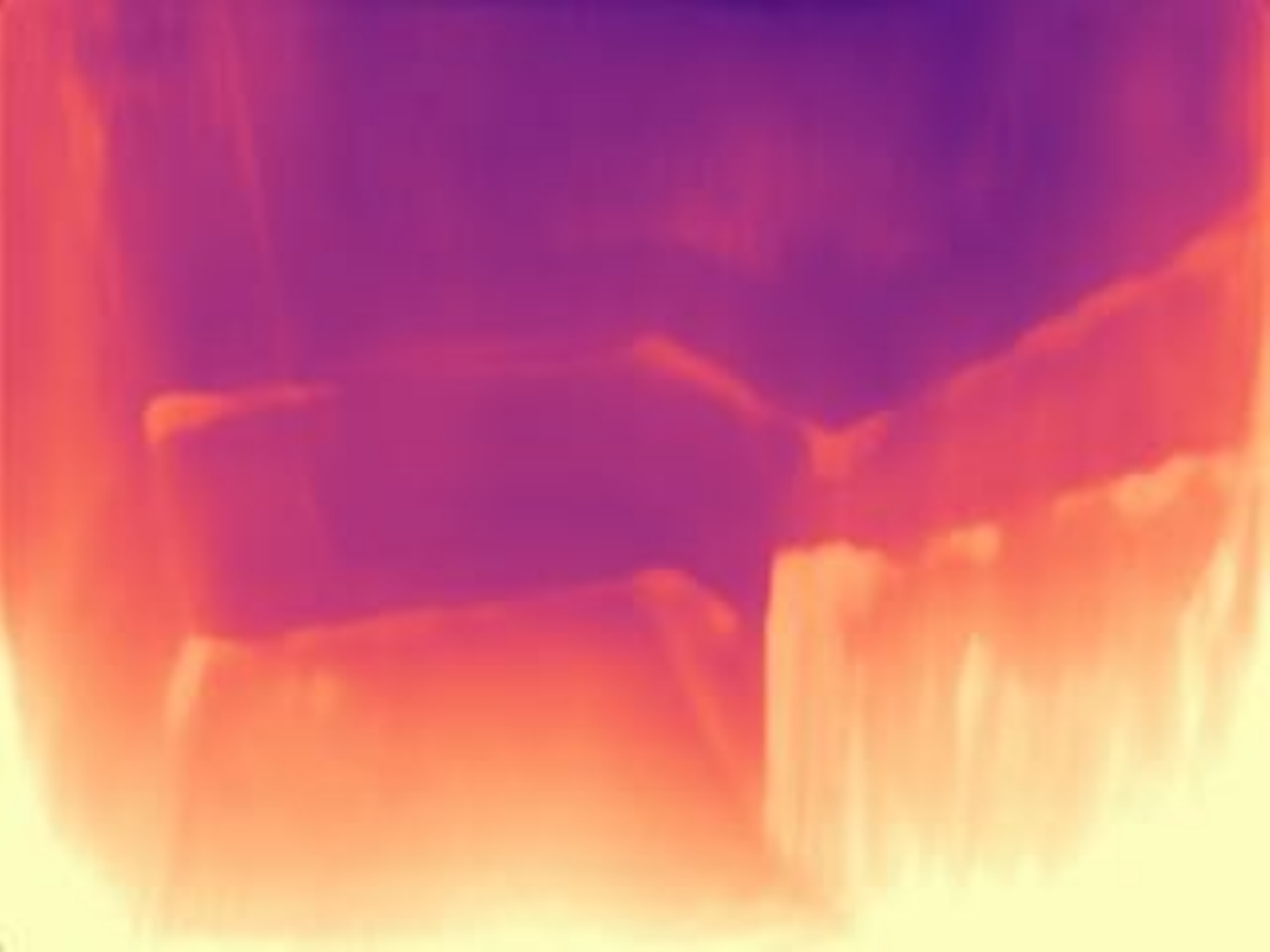} \qquad\qquad\quad & 
\includegraphics[width=\iw,height=\ih]{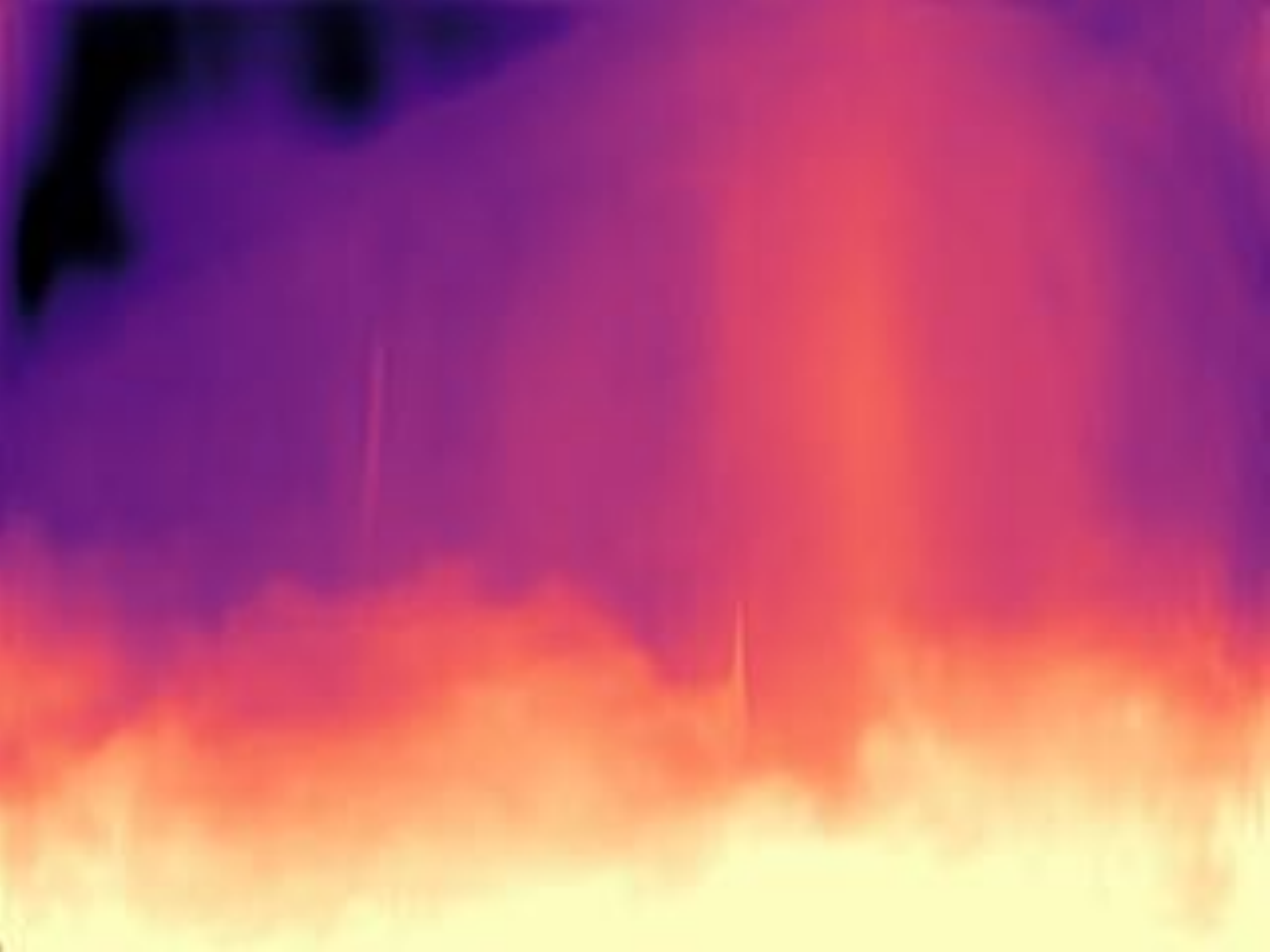} \qquad\qquad\quad &  
\includegraphics[width=\iw,height=\ih]{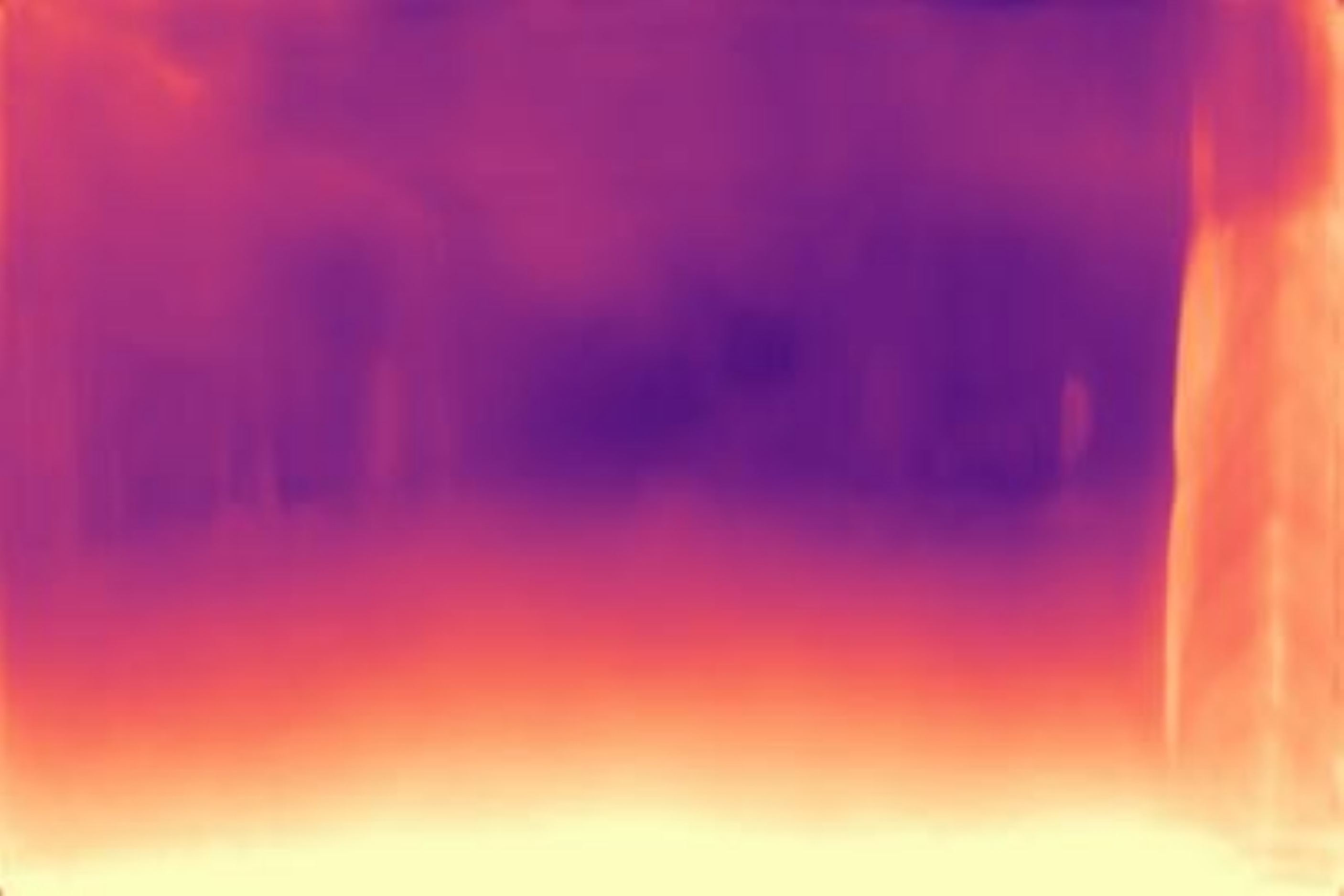} \qquad\qquad\quad &
\includegraphics[width=\iw,height=\ih]{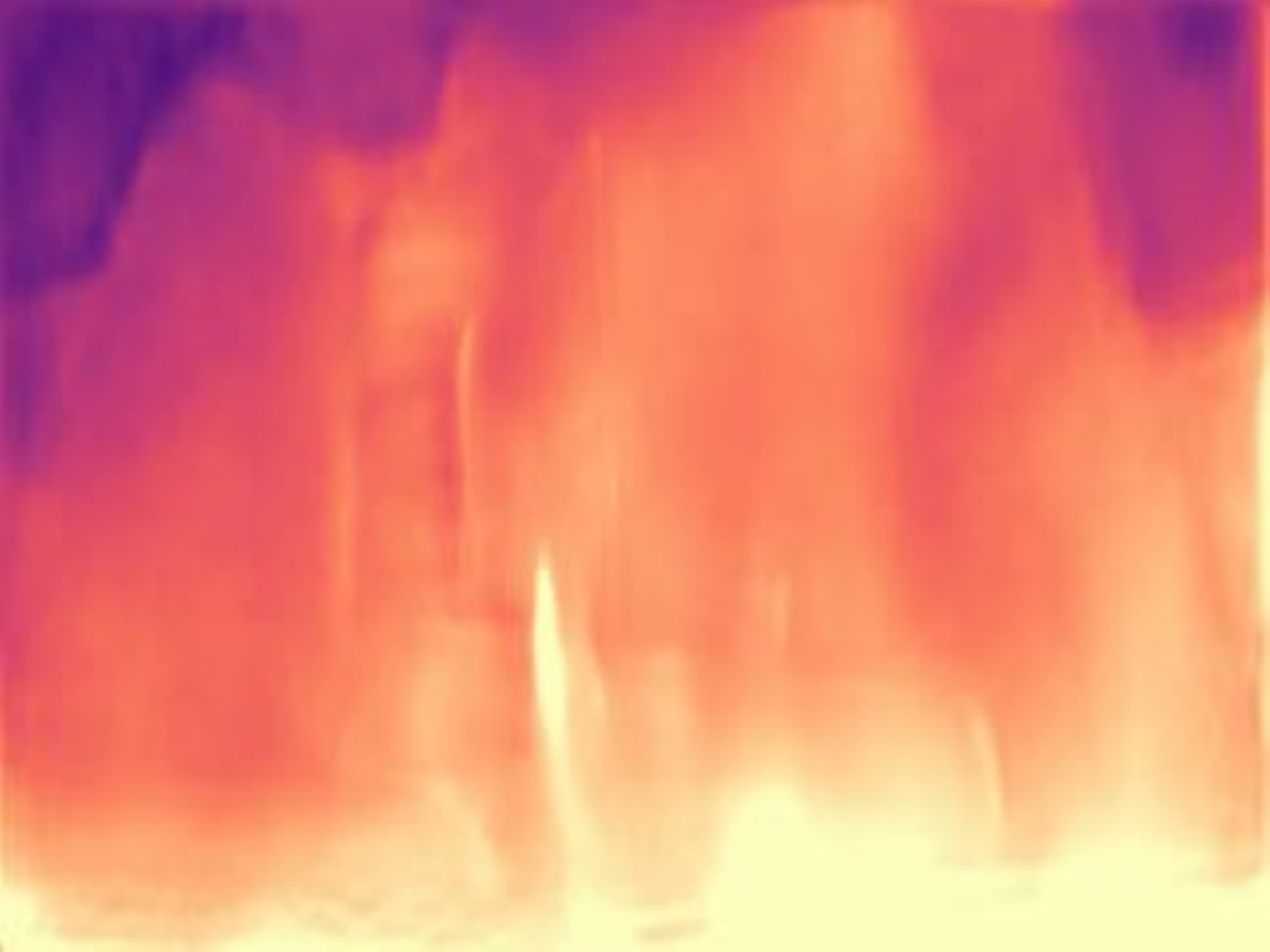} \\
\vspace{10mm}\\
\rotatebox[origin=c]{90}{\fontsize{\textw}{\texth} \selectfont MF-Ours\hspace{-270mm}}\hspace{24mm}
\includegraphics[width=\iw,height=\ih]{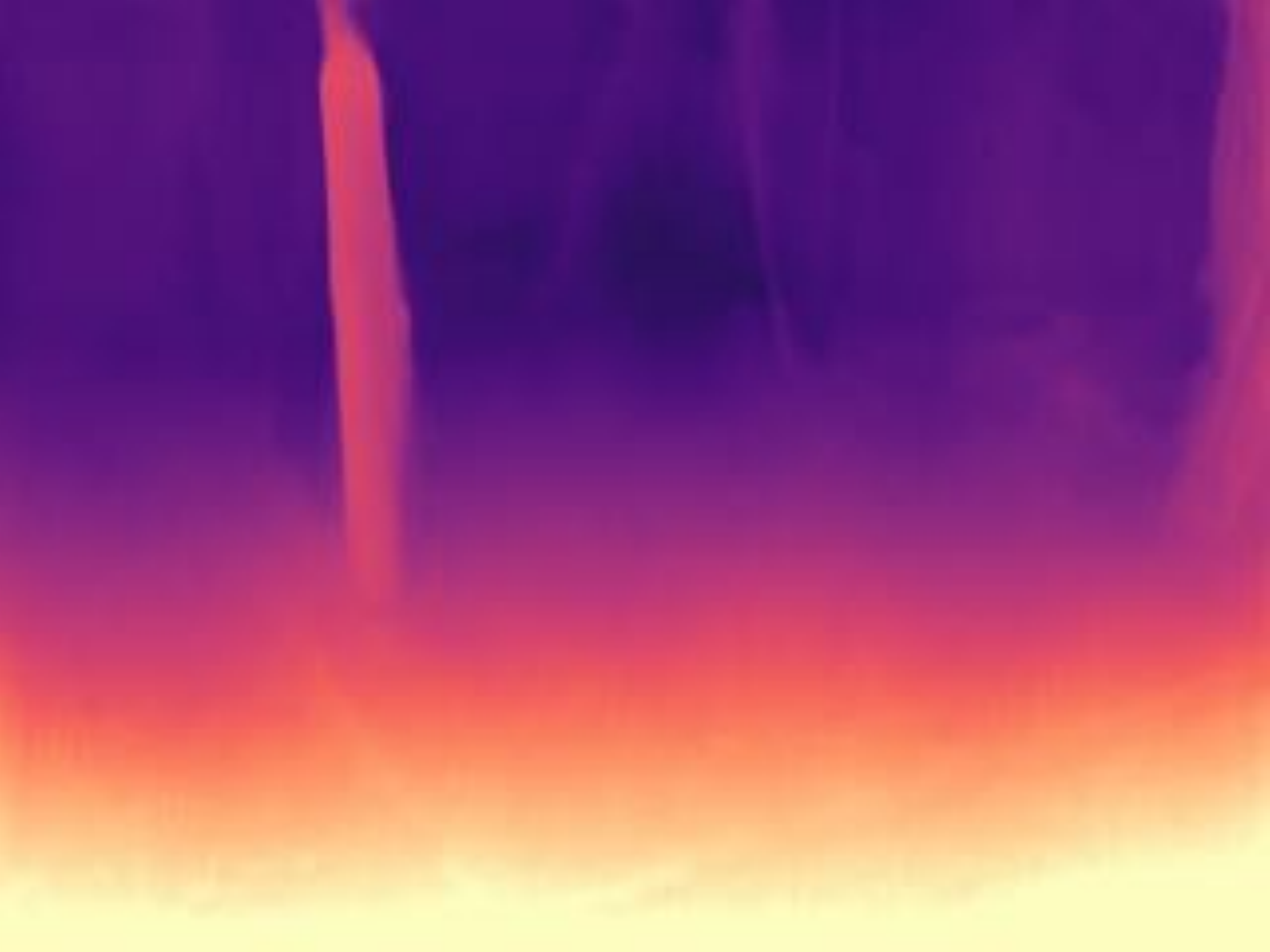} \qquad\qquad\quad & 
\includegraphics[width=\iw,height=\ih]{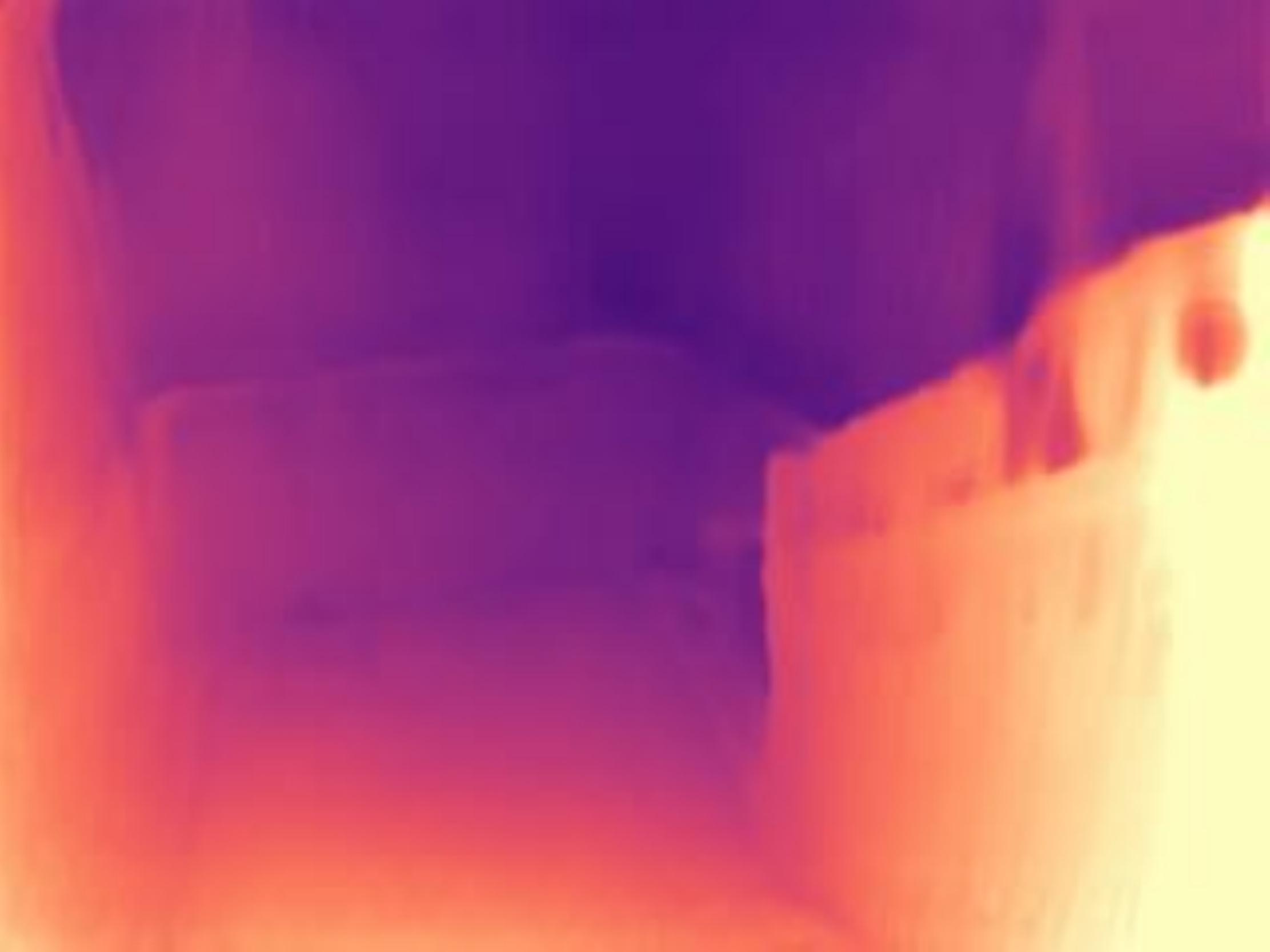} \qquad\qquad\quad & 
\includegraphics[width=\iw,height=\ih]{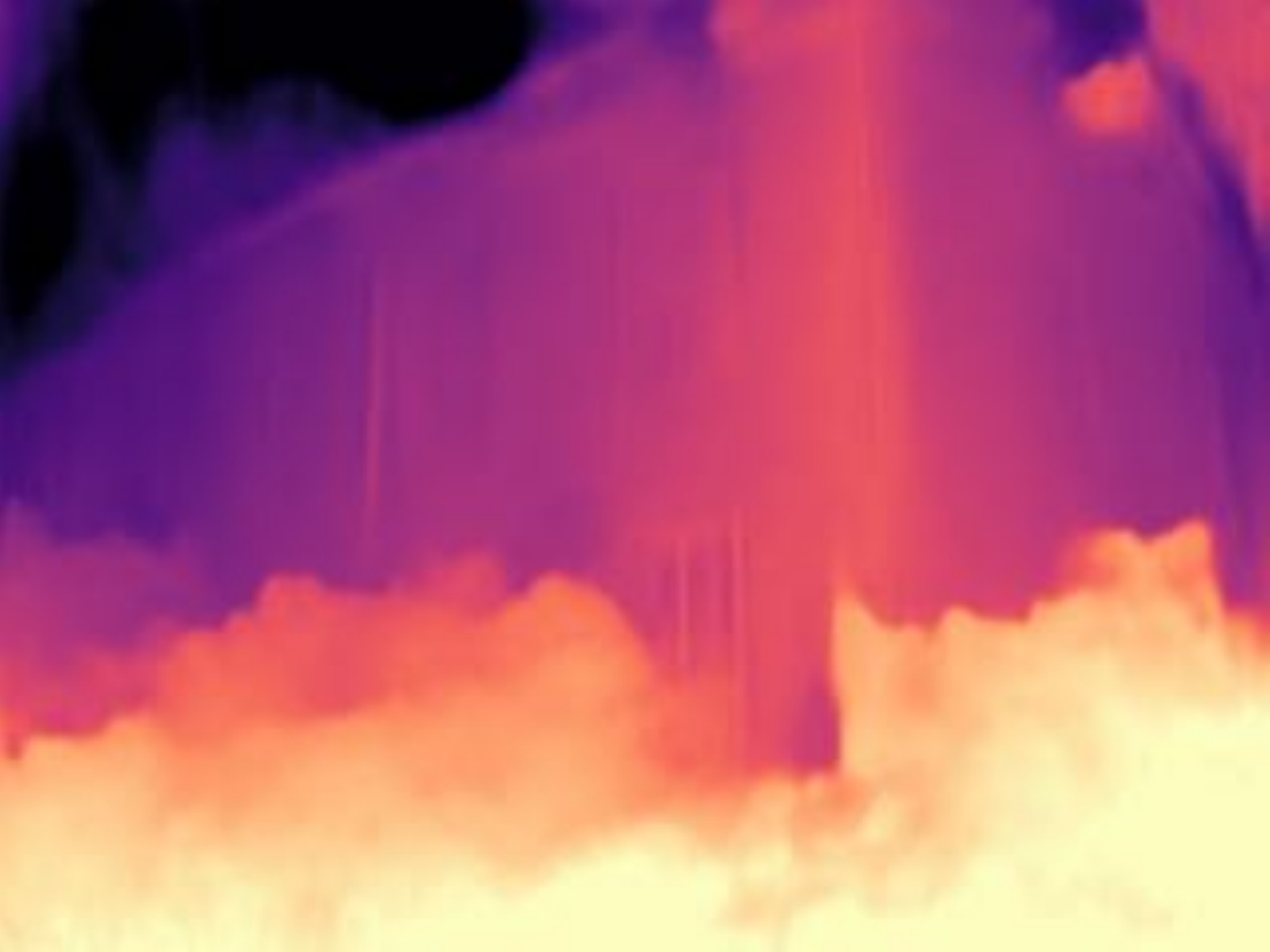} \qquad\qquad\quad &  
\includegraphics[width=\iw,height=\ih]{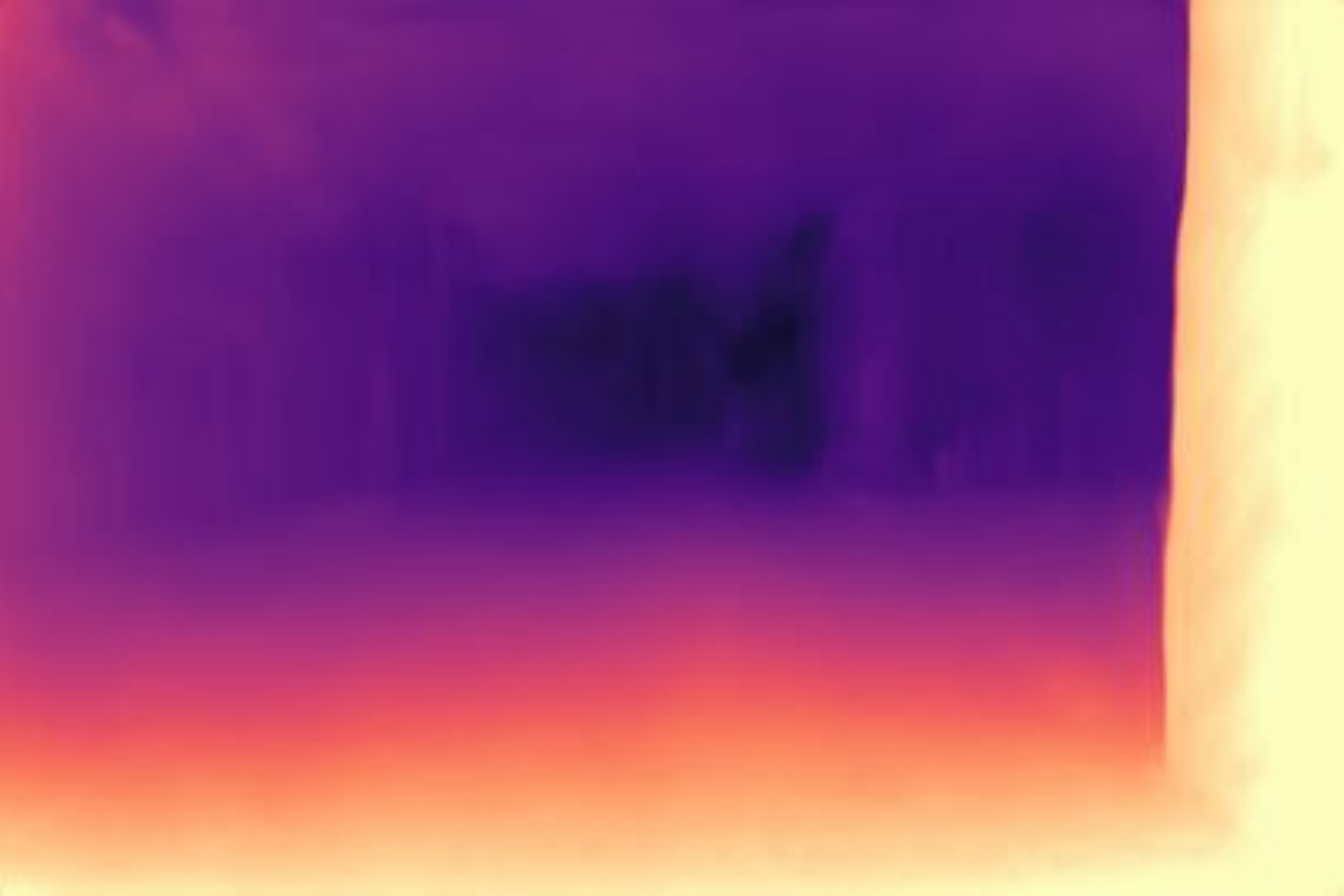} \qquad\qquad\quad &
\includegraphics[width=\iw,height=\ih]{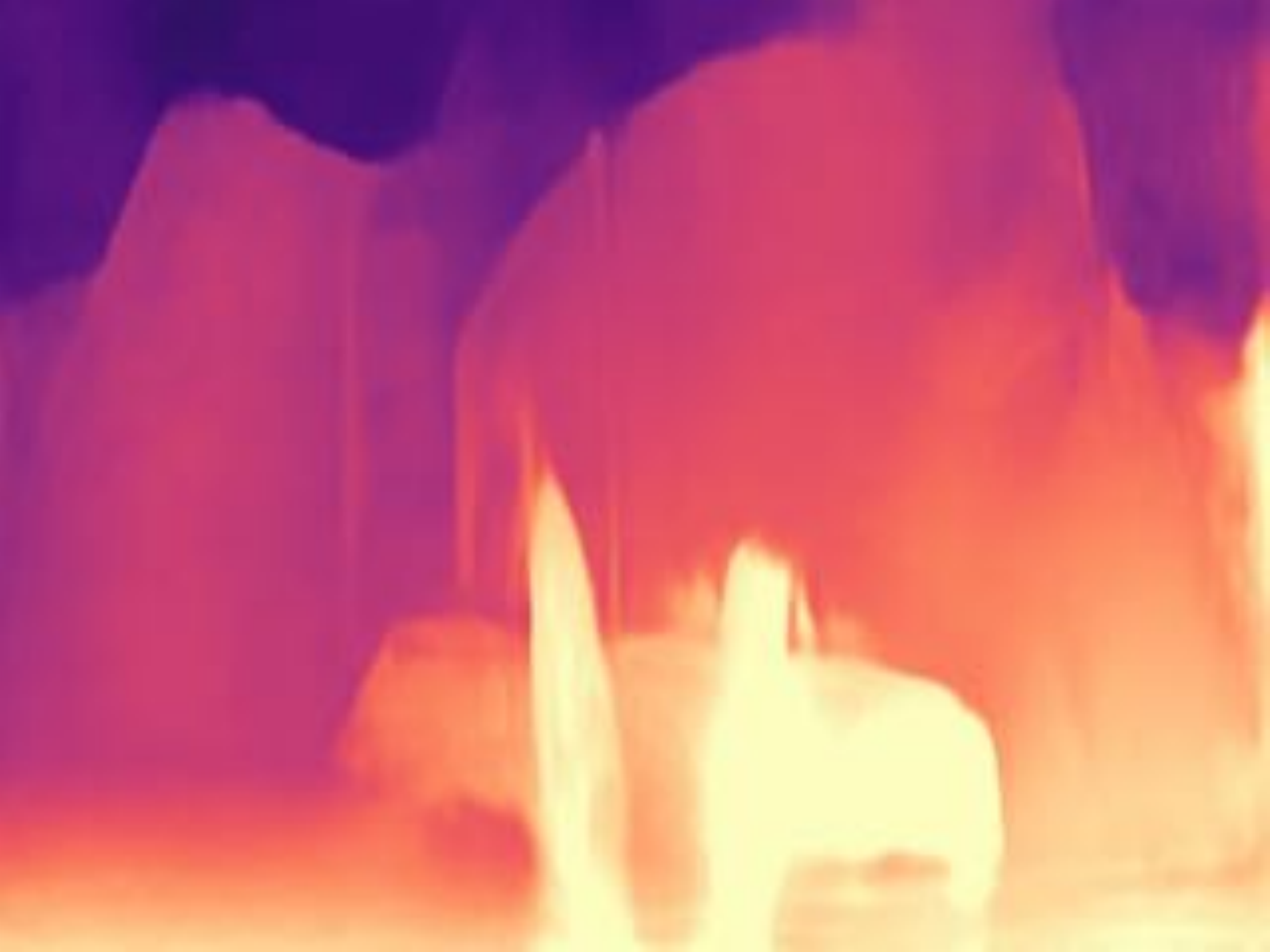} \\
\vspace{30mm}\\
\multicolumn{5}{c}{\fontsize{\w}{\h} \selectfont (b) Self-supervised Transformer-based methods } & 
\vspace{30mm}\\
\rotatebox[origin=c]{90}{\fontsize{\textw}{\texth} \selectfont BTS\hspace{-270mm}}\hspace{24mm}
\includegraphics[width=\iw,height=\ih]{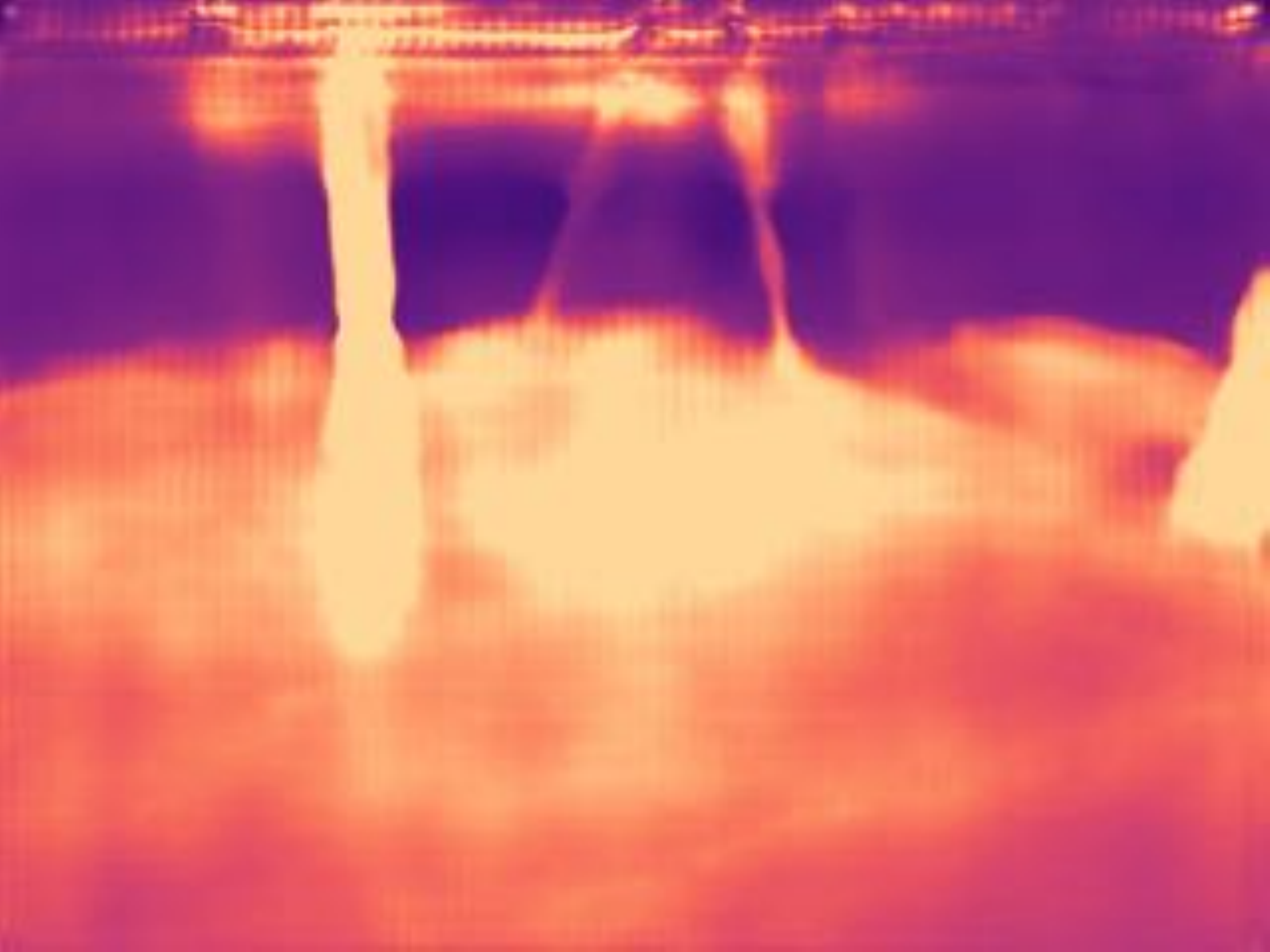} \qquad\qquad\quad & 
\includegraphics[width=\iw,height=\ih]{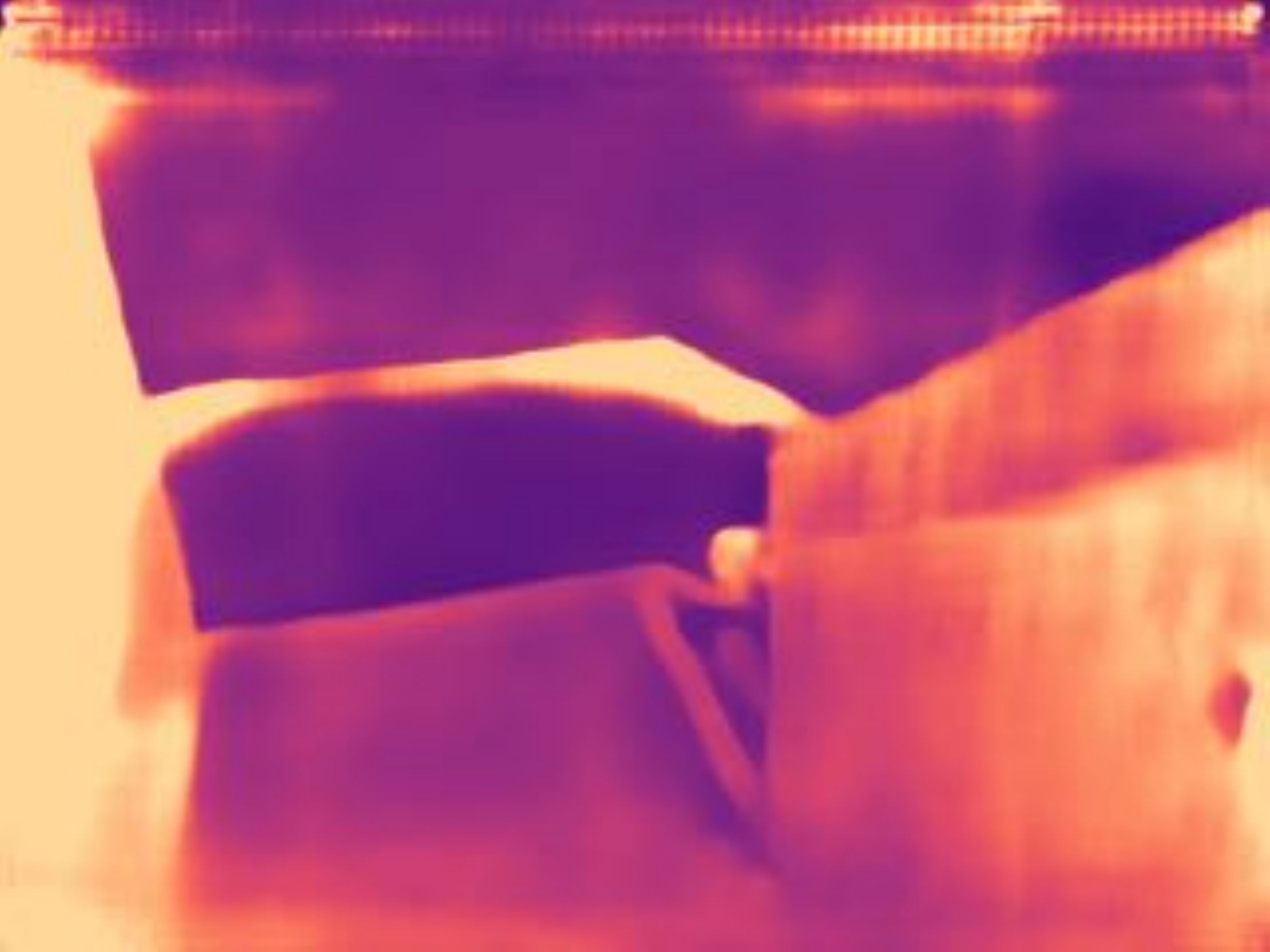} \qquad\qquad\quad & 
\includegraphics[width=\iw,height=\ih]{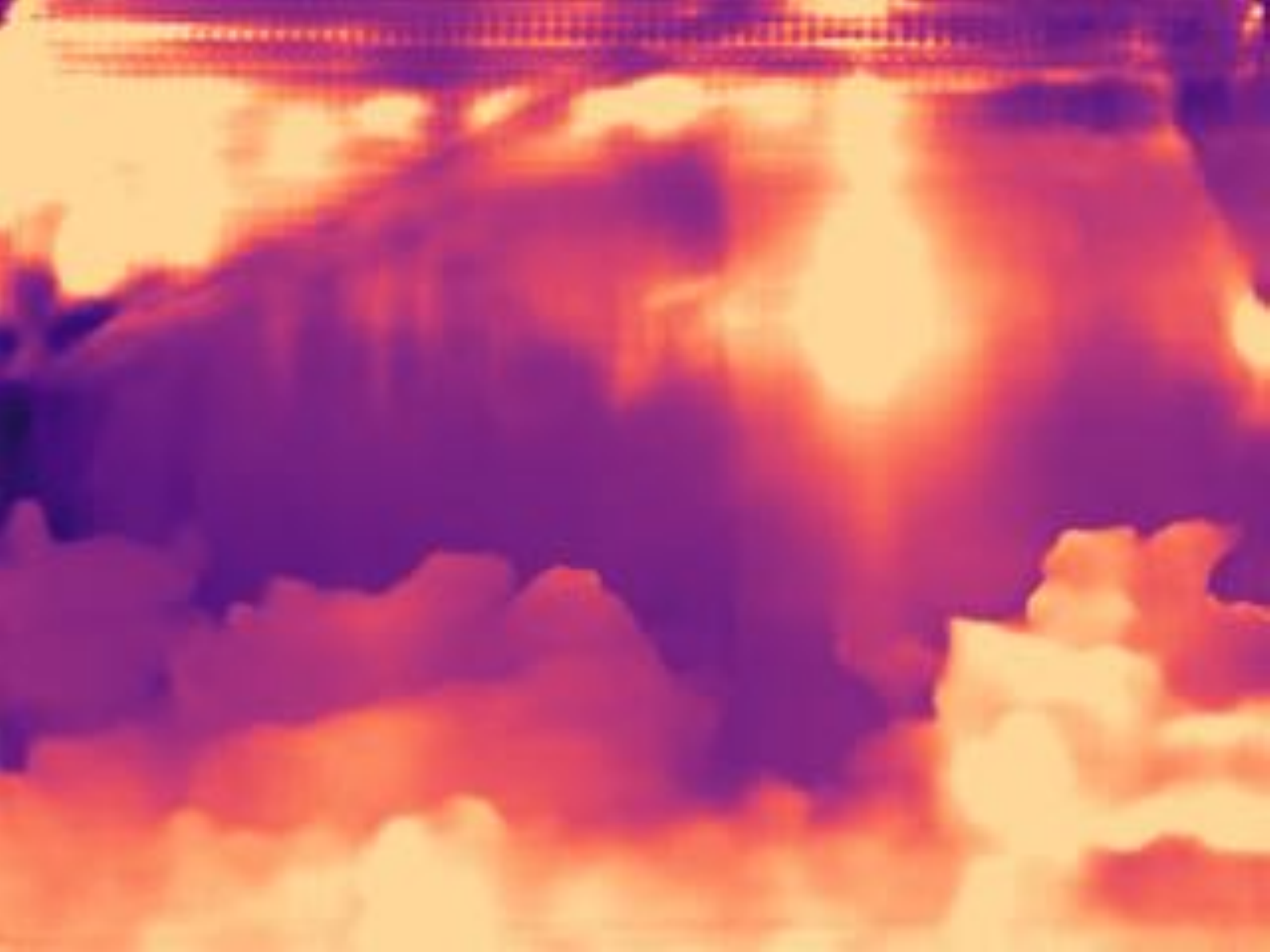} \qquad\qquad\quad &  
\includegraphics[width=\iw,height=\ih]{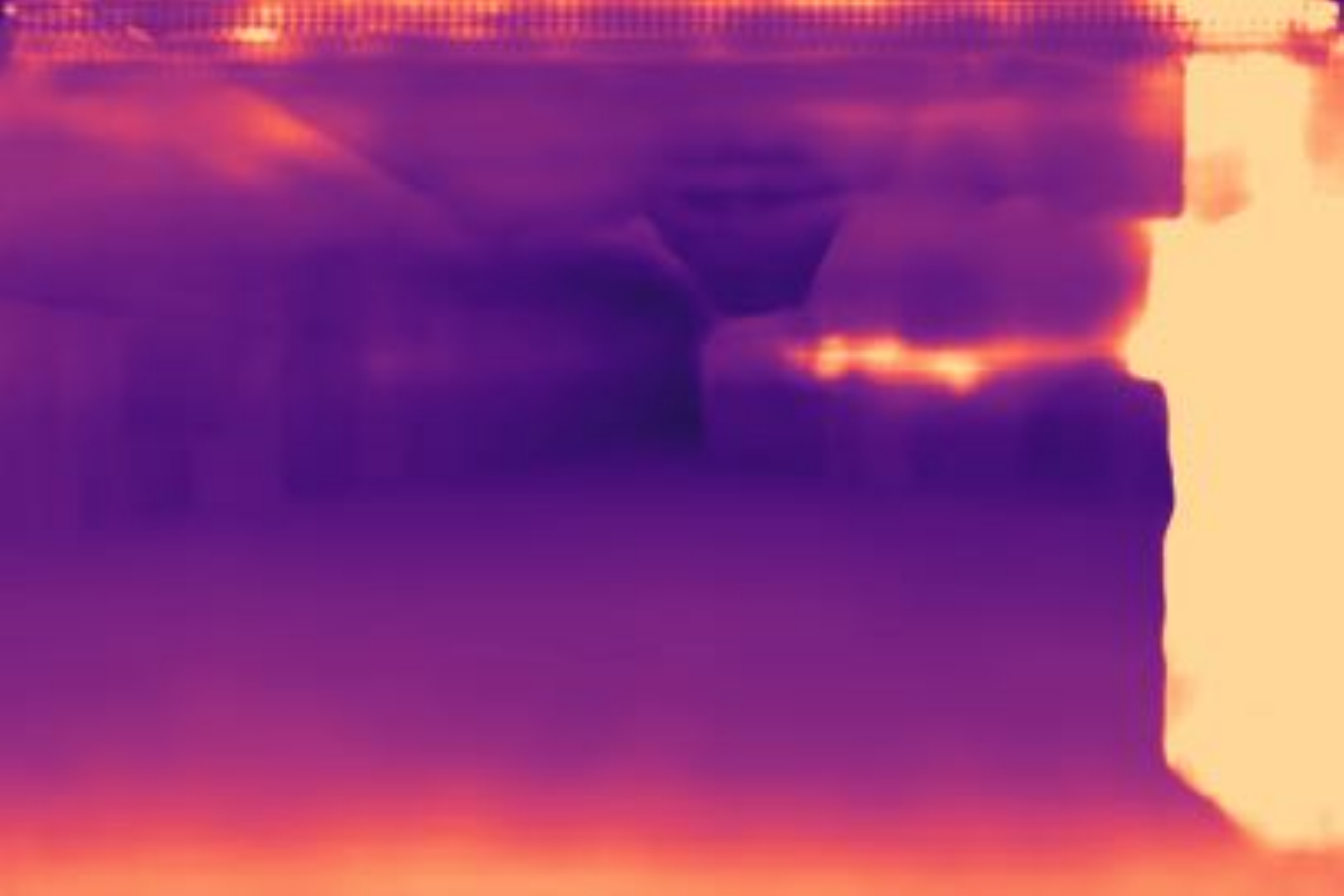} \qquad\qquad\quad &
\includegraphics[width=\iw,height=\ih]{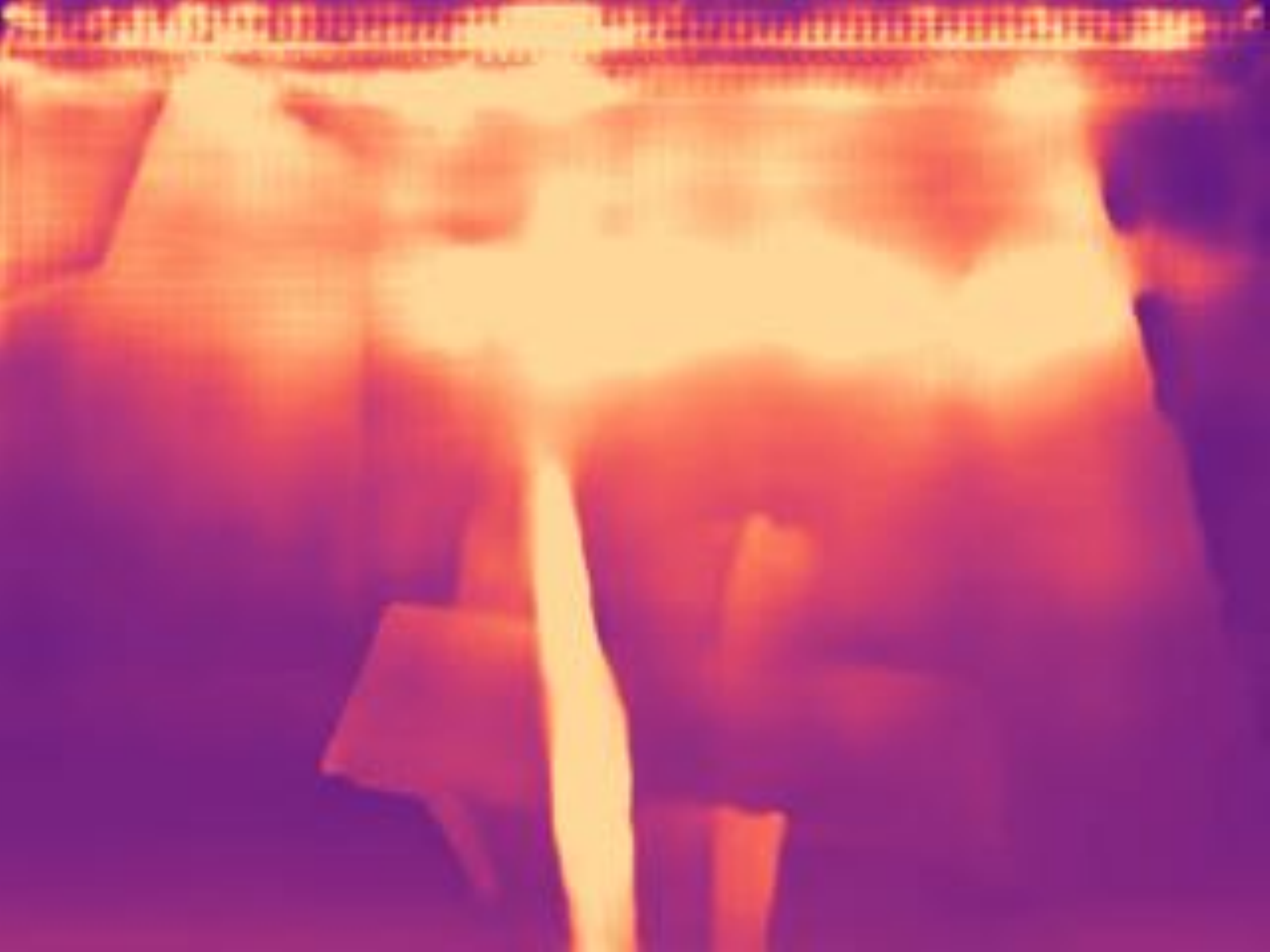} \\
\vspace{10mm}\\
\rotatebox[origin=c]{90}{\fontsize{\textw}{\texth} \selectfont Adabins\hspace{-270mm}}\hspace{24mm}
\includegraphics[width=\iw,height=\ih]{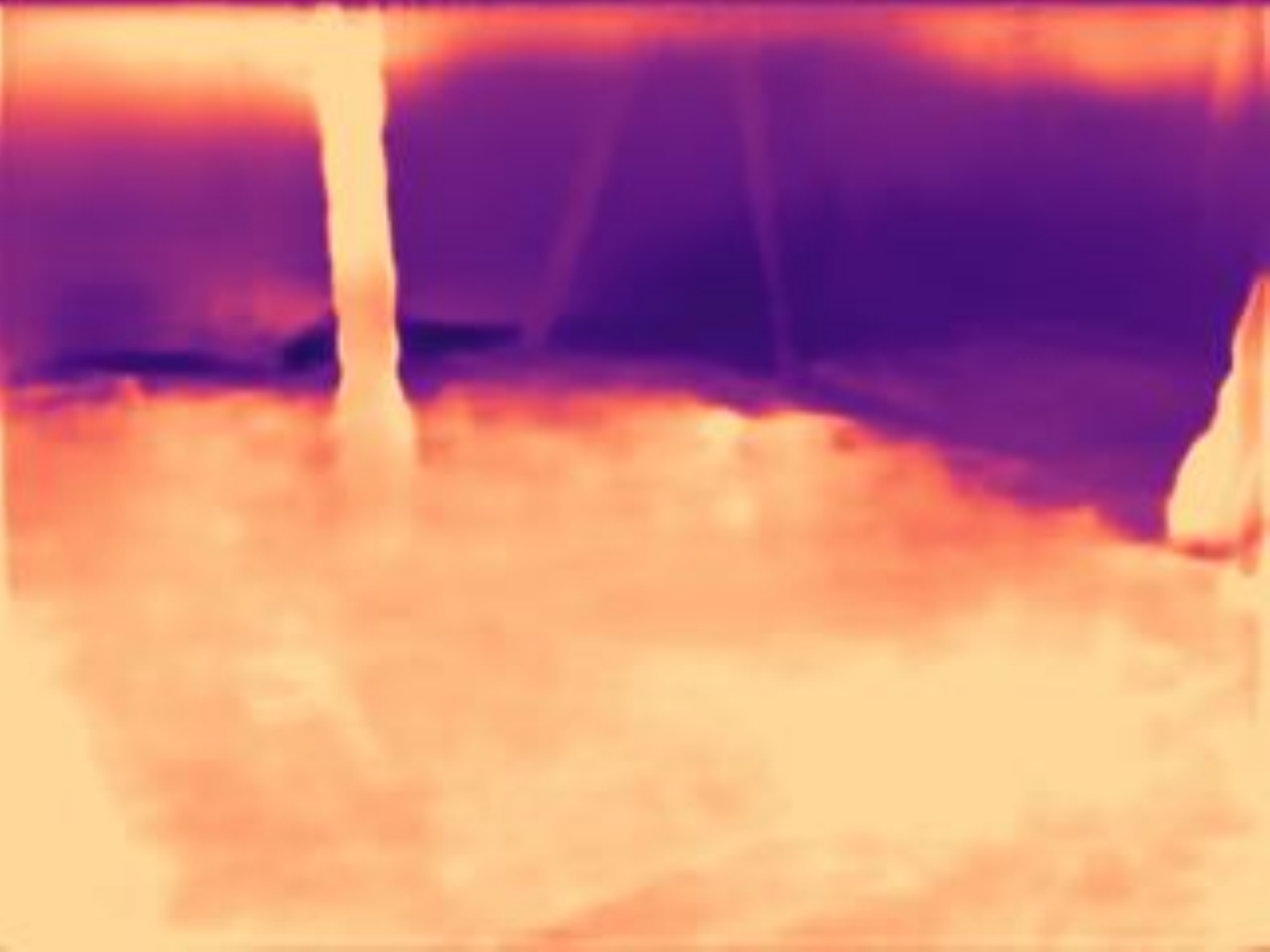} \qquad\qquad\quad & 
\includegraphics[width=\iw,height=\ih]{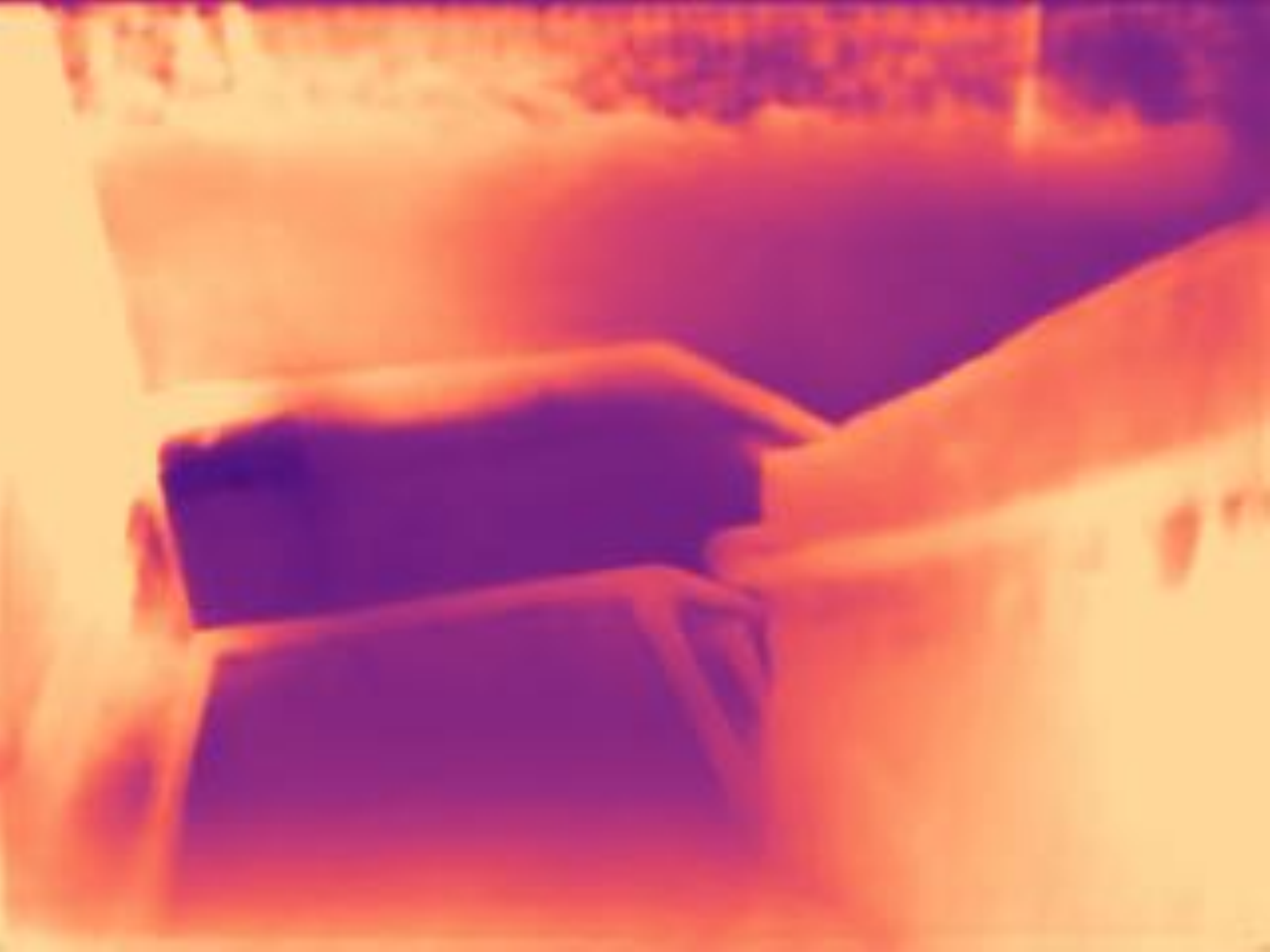} \qquad\qquad\quad & 
\includegraphics[width=\iw,height=\ih]{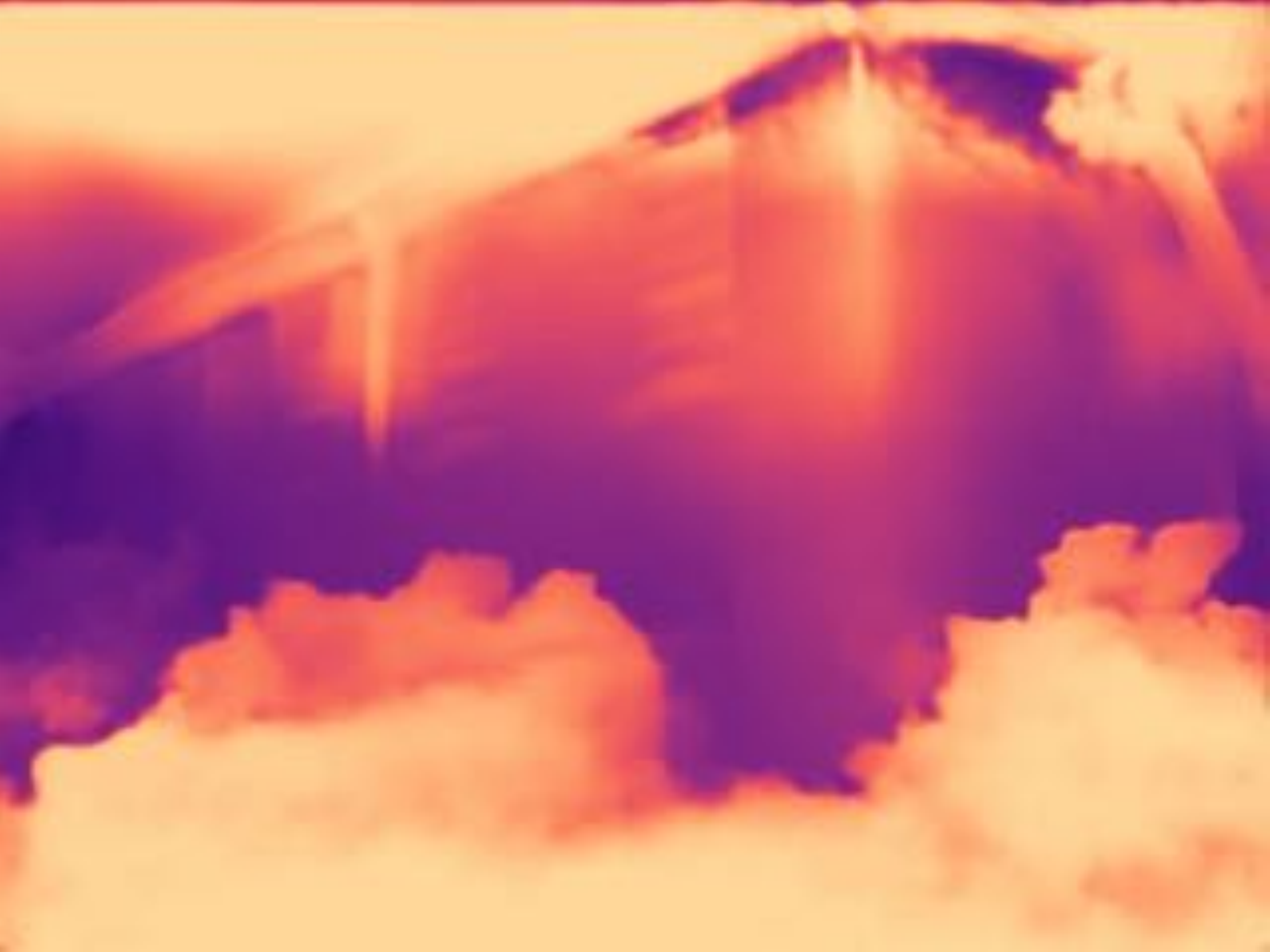} \qquad\qquad\quad &  
\includegraphics[width=\iw,height=\ih]{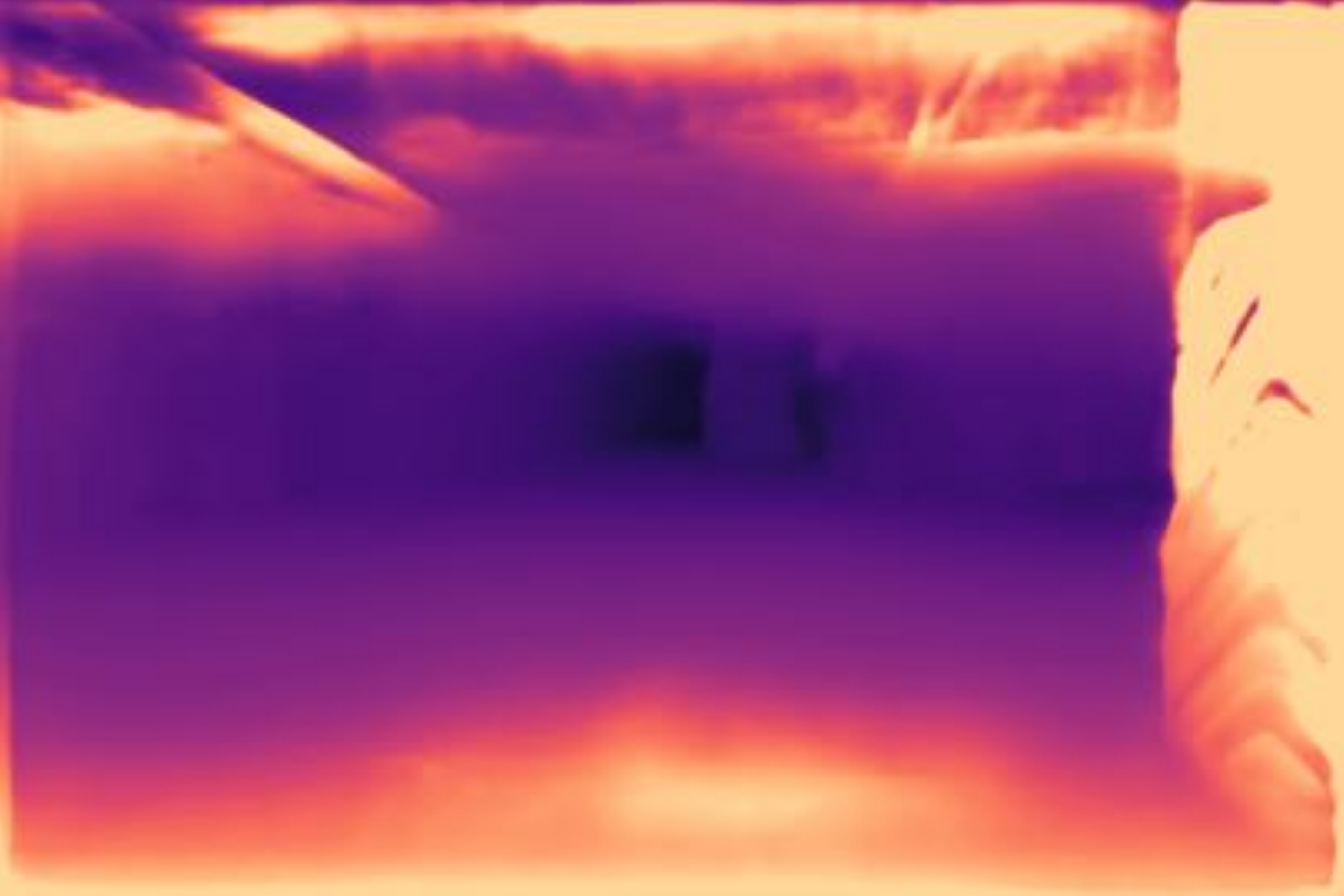} \qquad\qquad\quad &
\includegraphics[width=\iw,height=\ih]{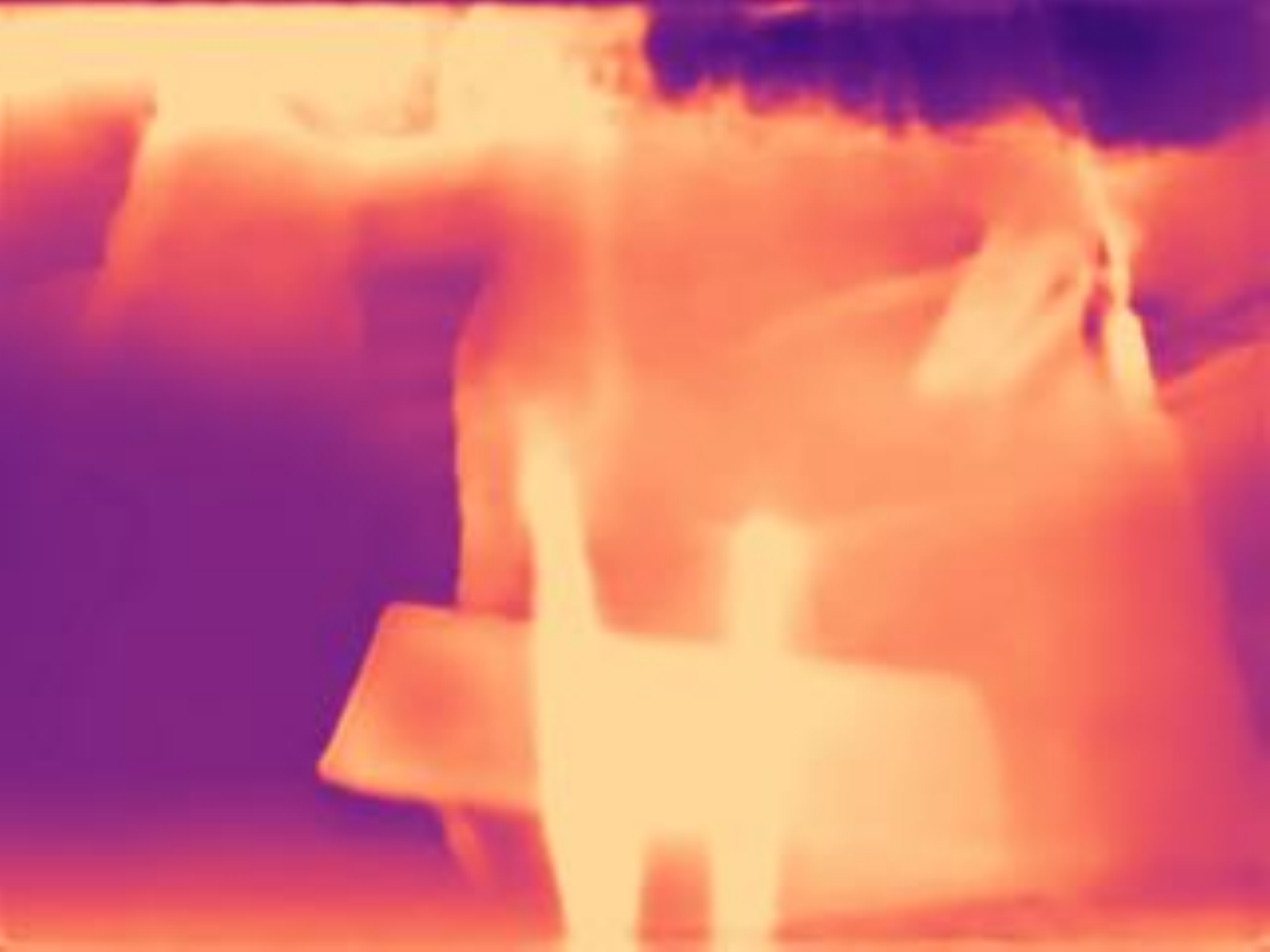}\\
\vspace{30mm}\\
\multicolumn{5}{c}{\fontsize{\w}{\h} \selectfont (c) Supervised CNN-based methods } & 
\vspace{30mm}\\
\rotatebox[origin=c]{90}{\fontsize{\textw}{\texth} \selectfont TransDepth\hspace{-270mm}}\hspace{24mm}
\includegraphics[width=\iw,height=\ih]{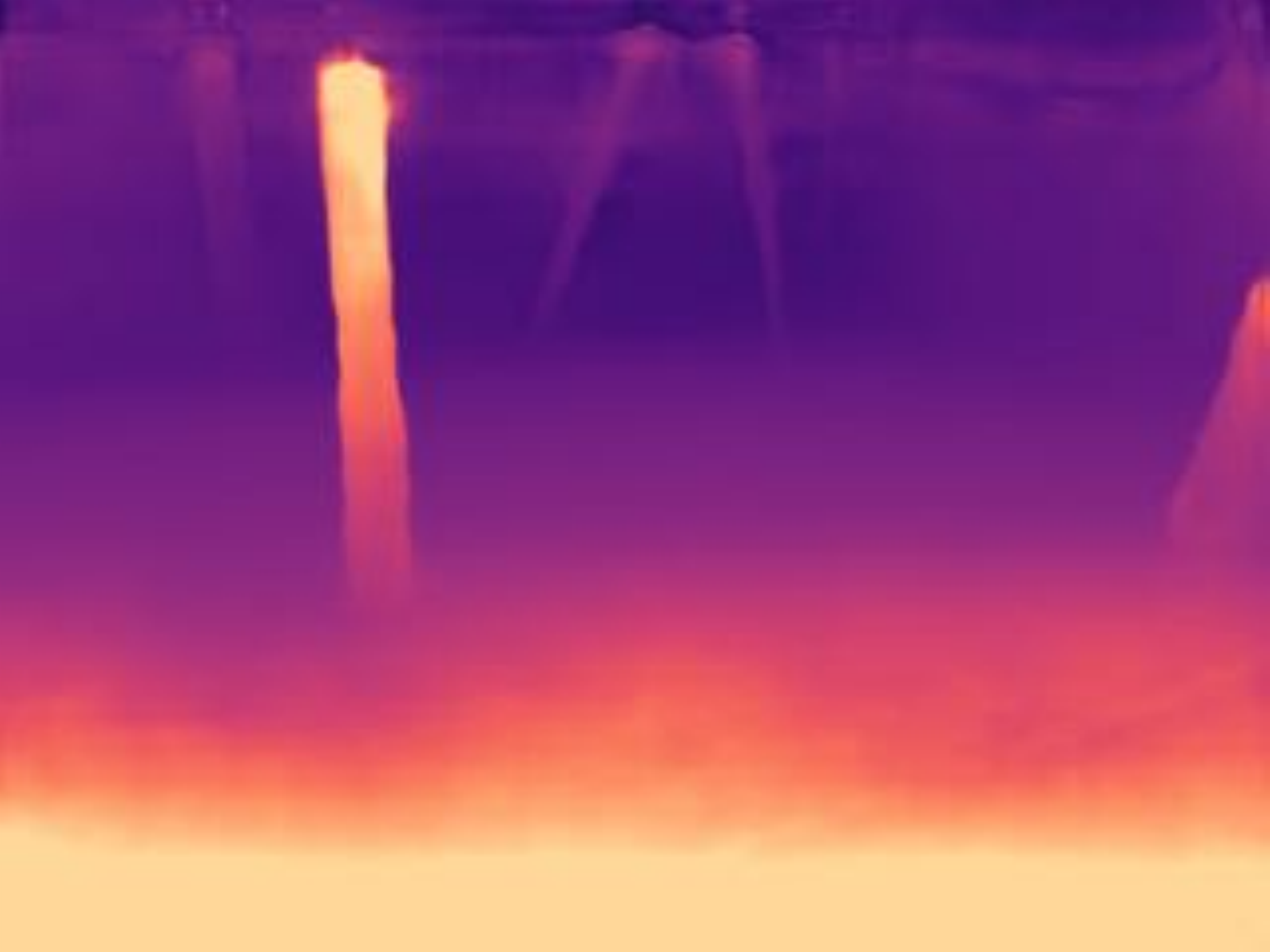} \qquad\qquad\quad & 
\includegraphics[width=\iw,height=\ih]{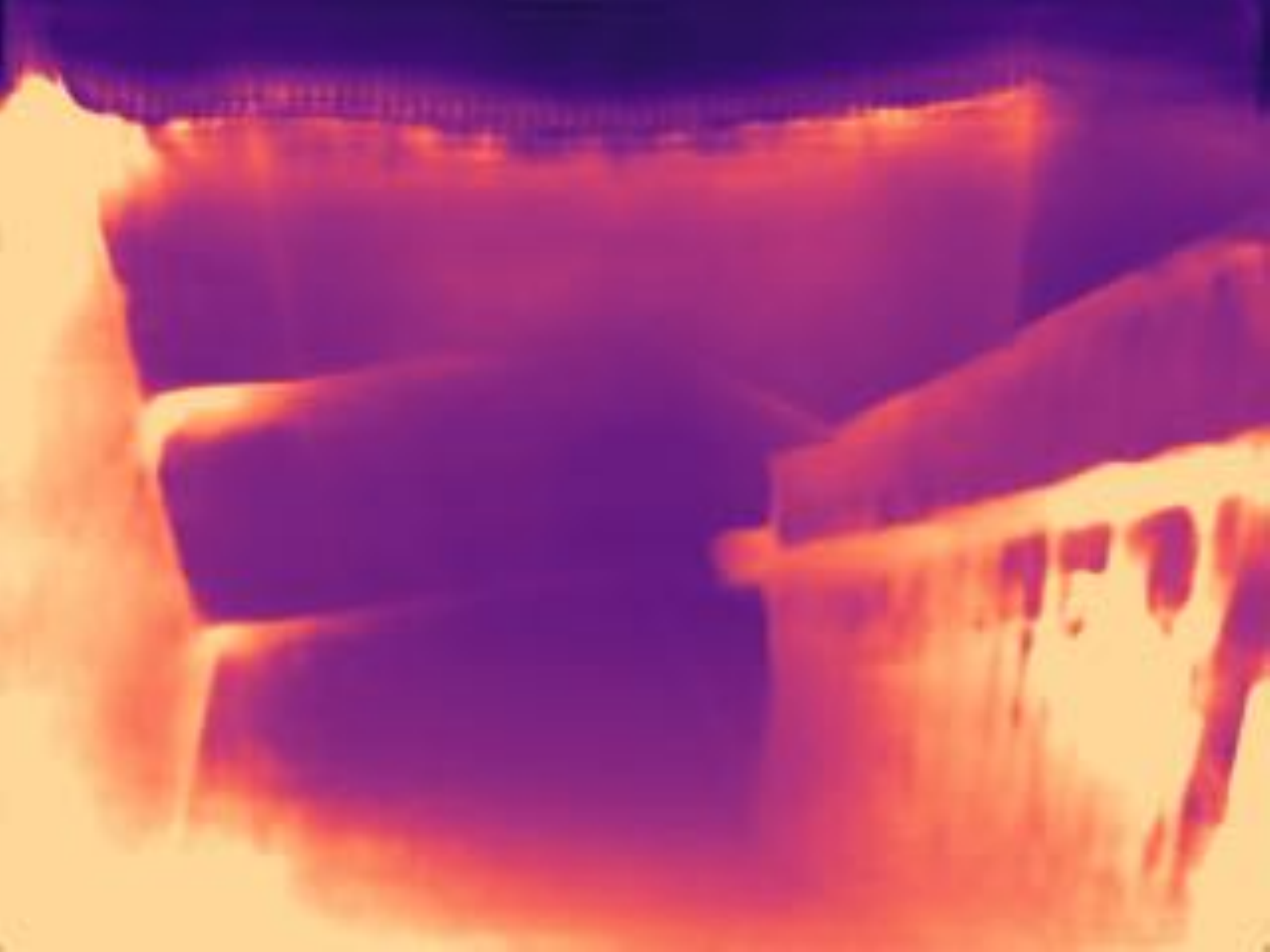} \qquad\qquad\quad & 
\includegraphics[width=\iw,height=\ih]{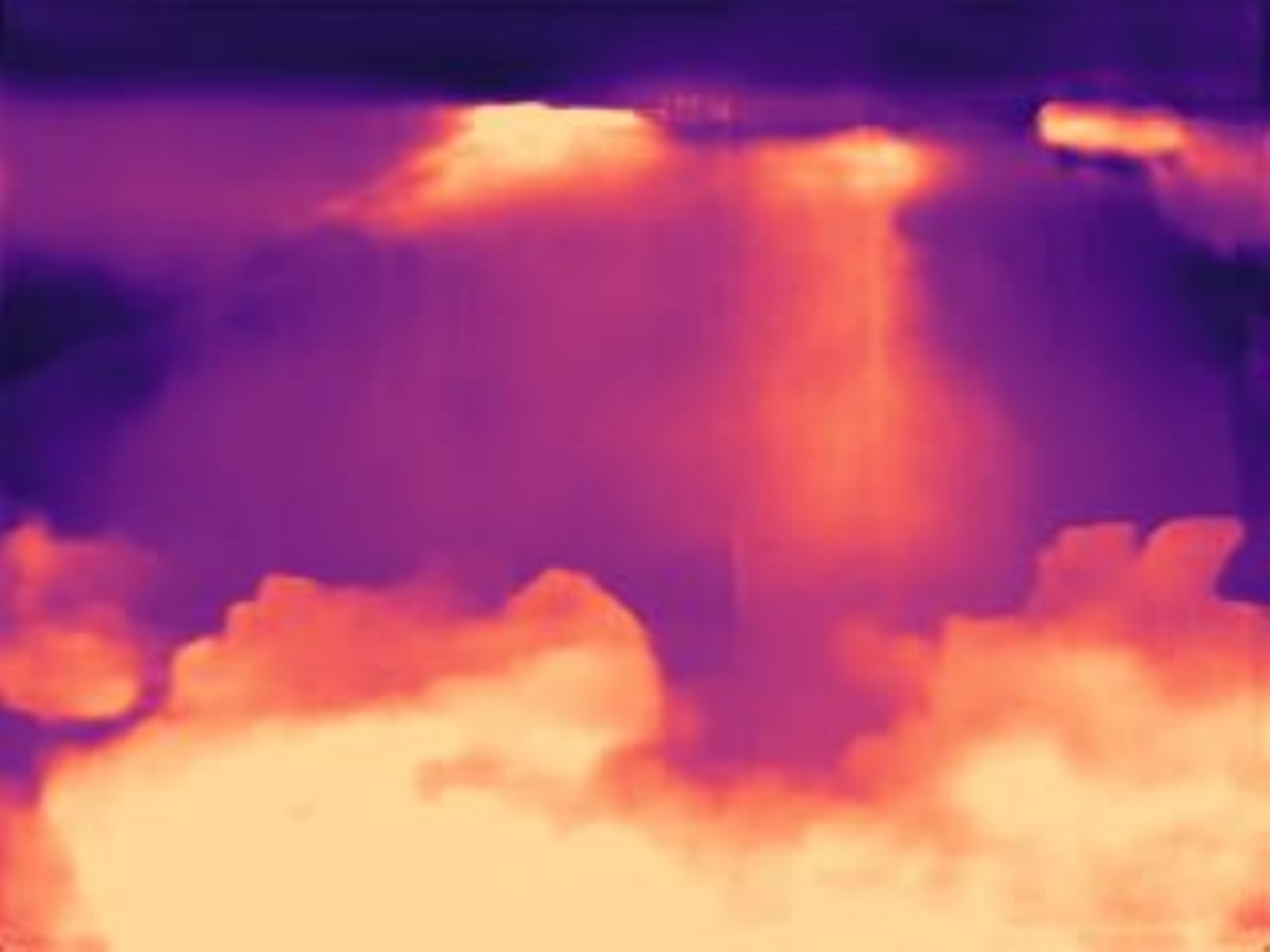} \qquad\qquad\quad &  
\includegraphics[width=\iw,height=\ih]{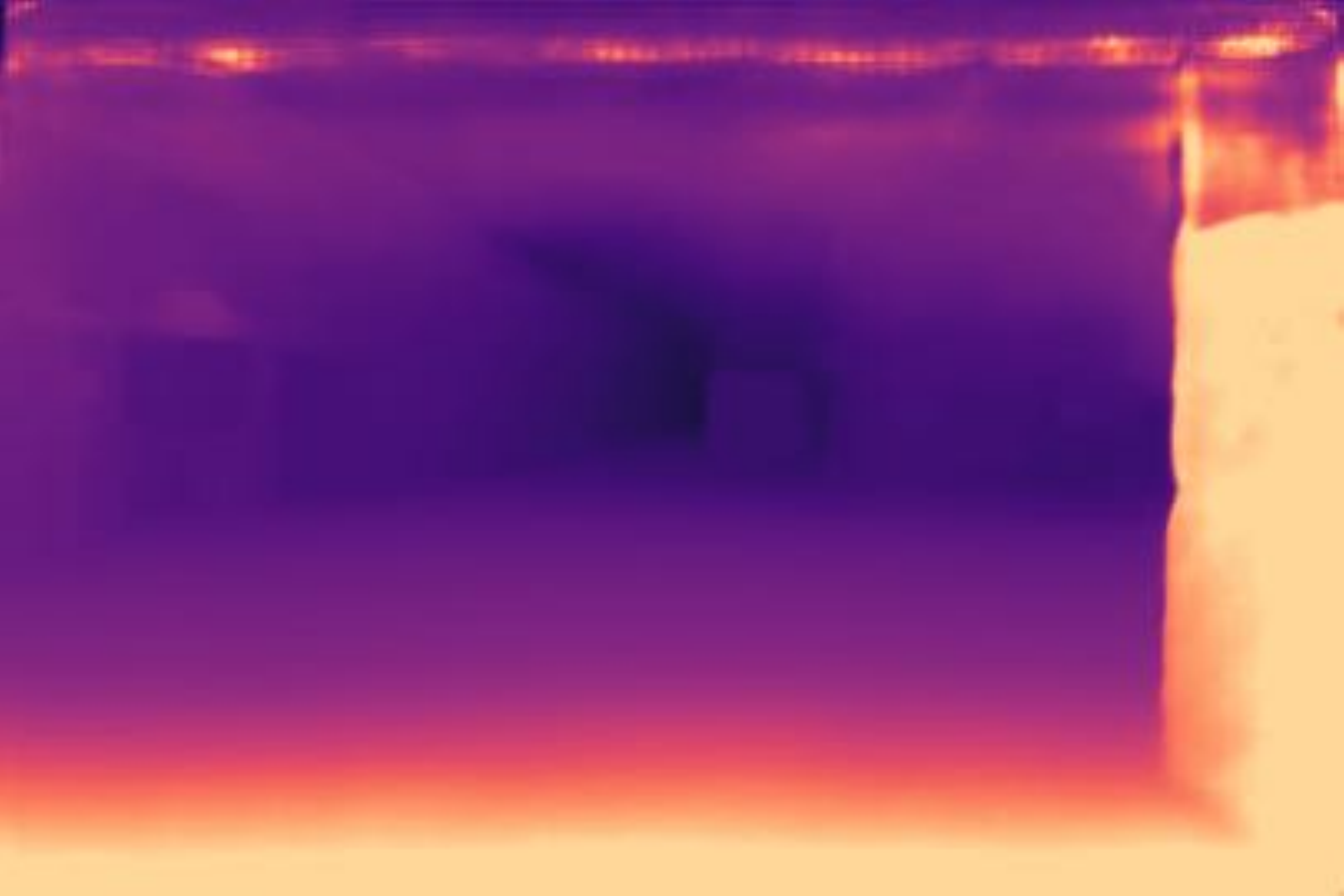} \qquad\qquad\quad &
\includegraphics[width=\iw,height=\ih]{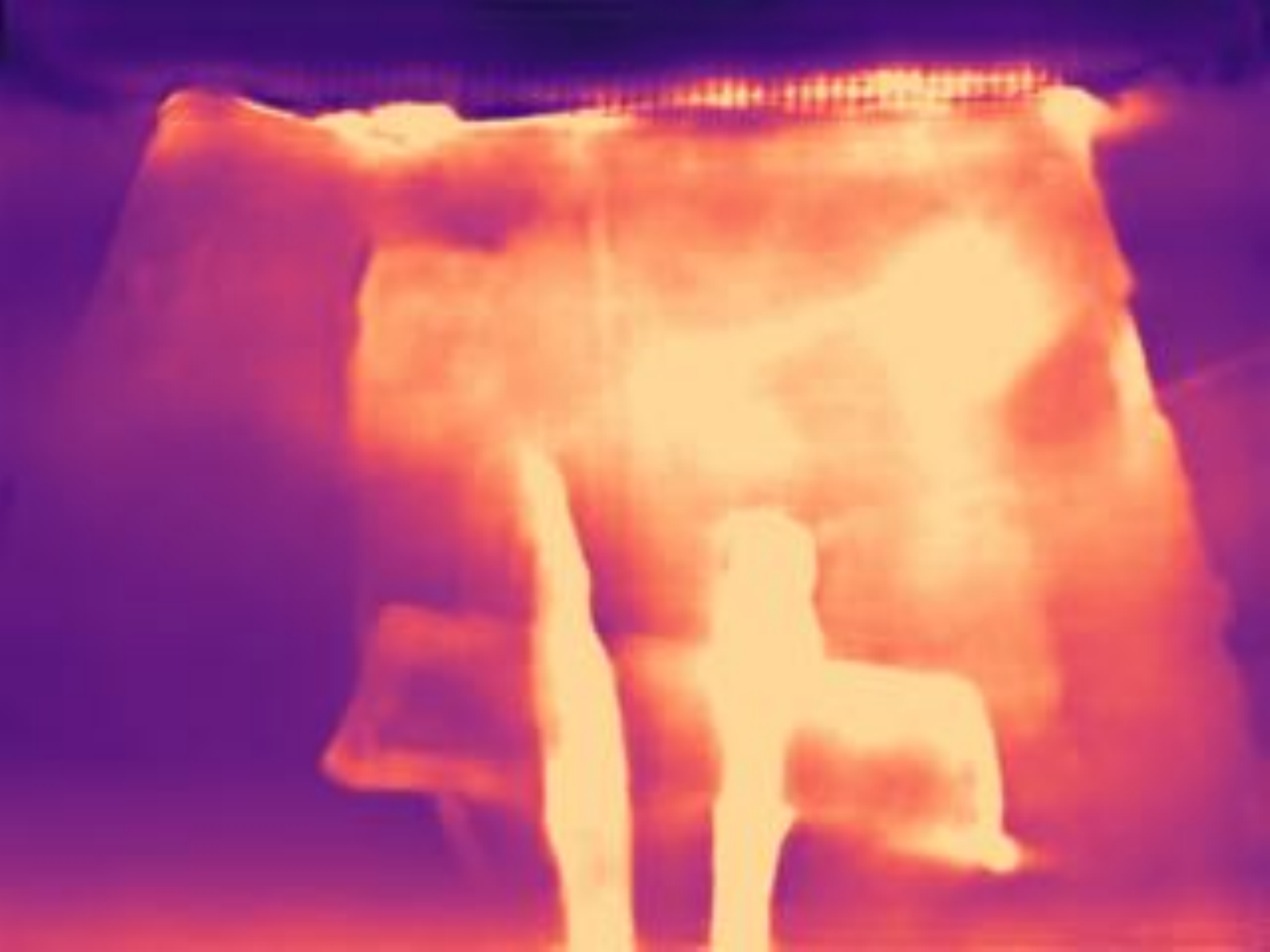} \\
\vspace{10mm}\\
\rotatebox[origin=c]{90}{\fontsize{\textw}{\texth} \selectfont DepthFormer\hspace{-270mm}}\hspace{24mm}
\includegraphics[width=\iw,height=\ih]{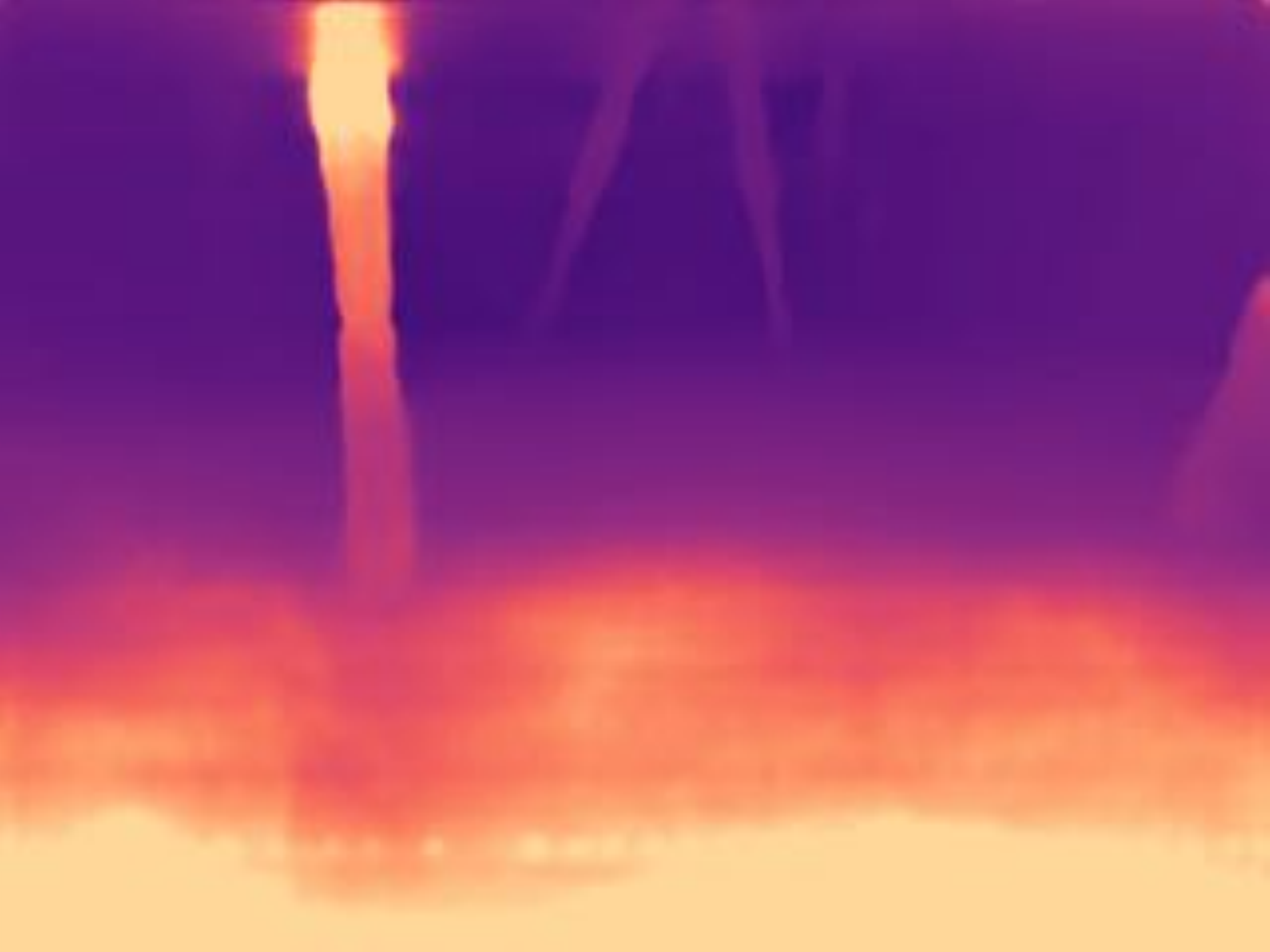} \qquad\qquad\quad & 
\includegraphics[width=\iw,height=\ih]{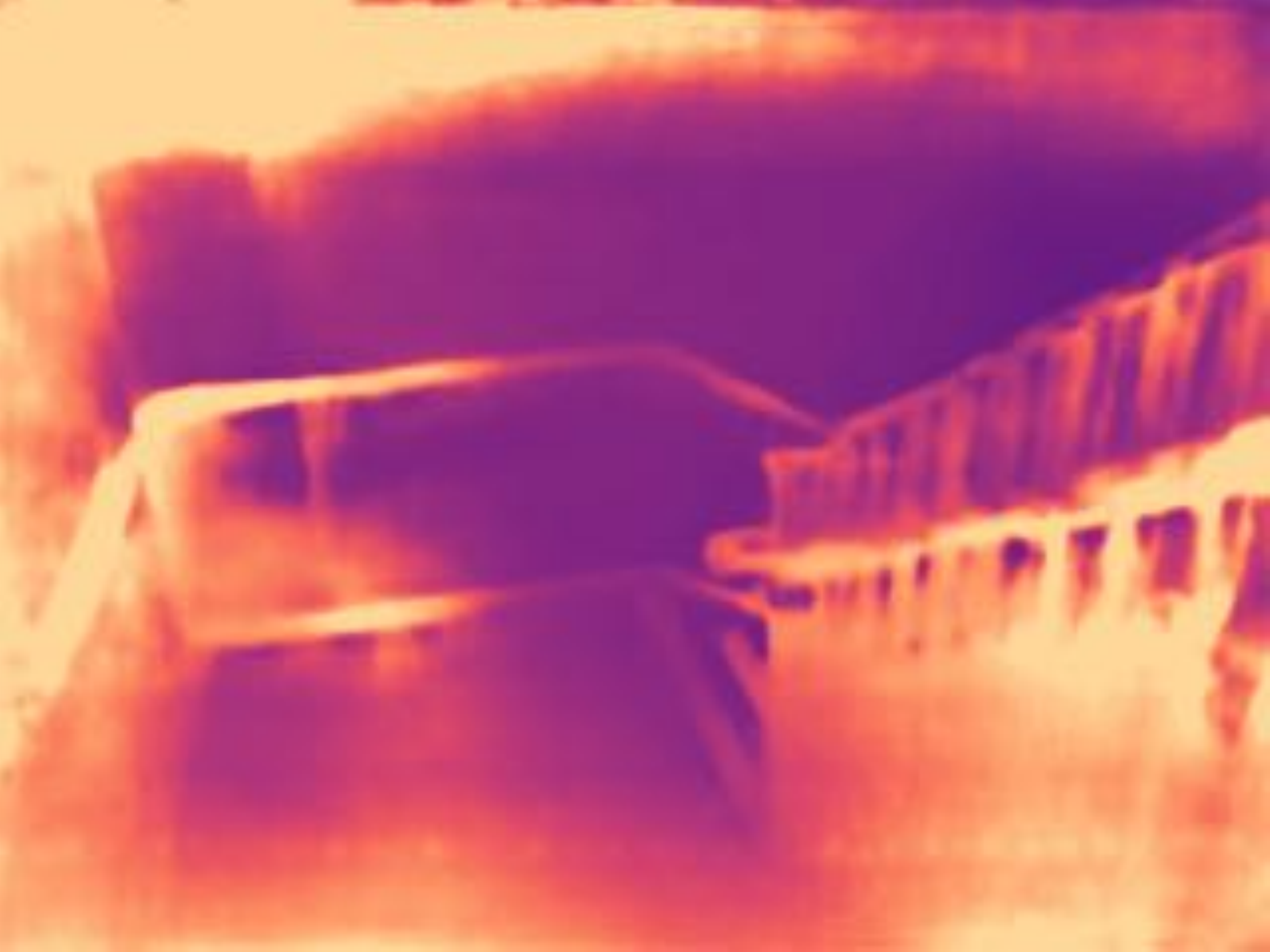} \qquad\qquad\quad & 
\includegraphics[width=\iw,height=\ih]{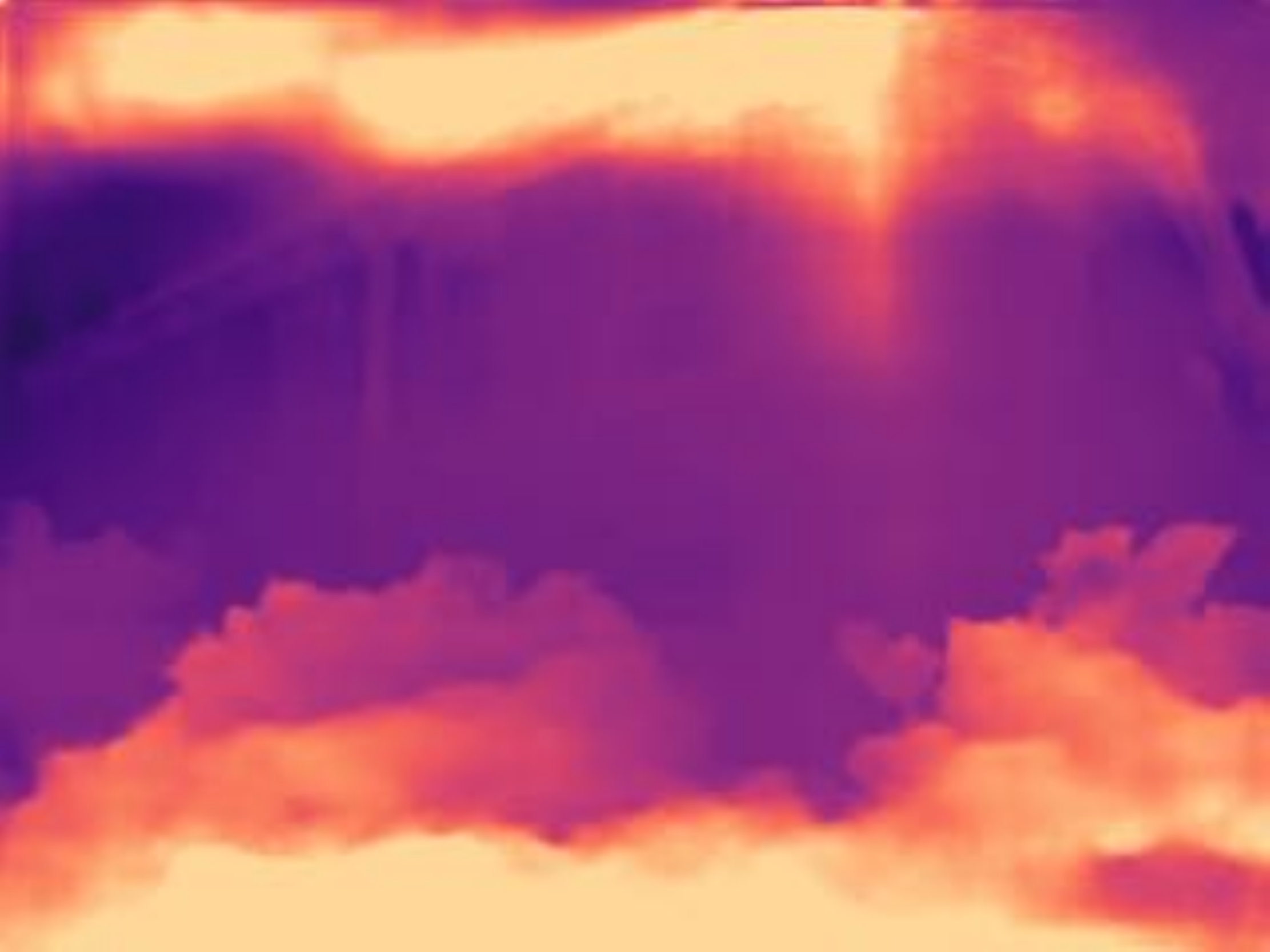} \qquad\qquad\quad &  
\includegraphics[width=\iw,height=\ih]{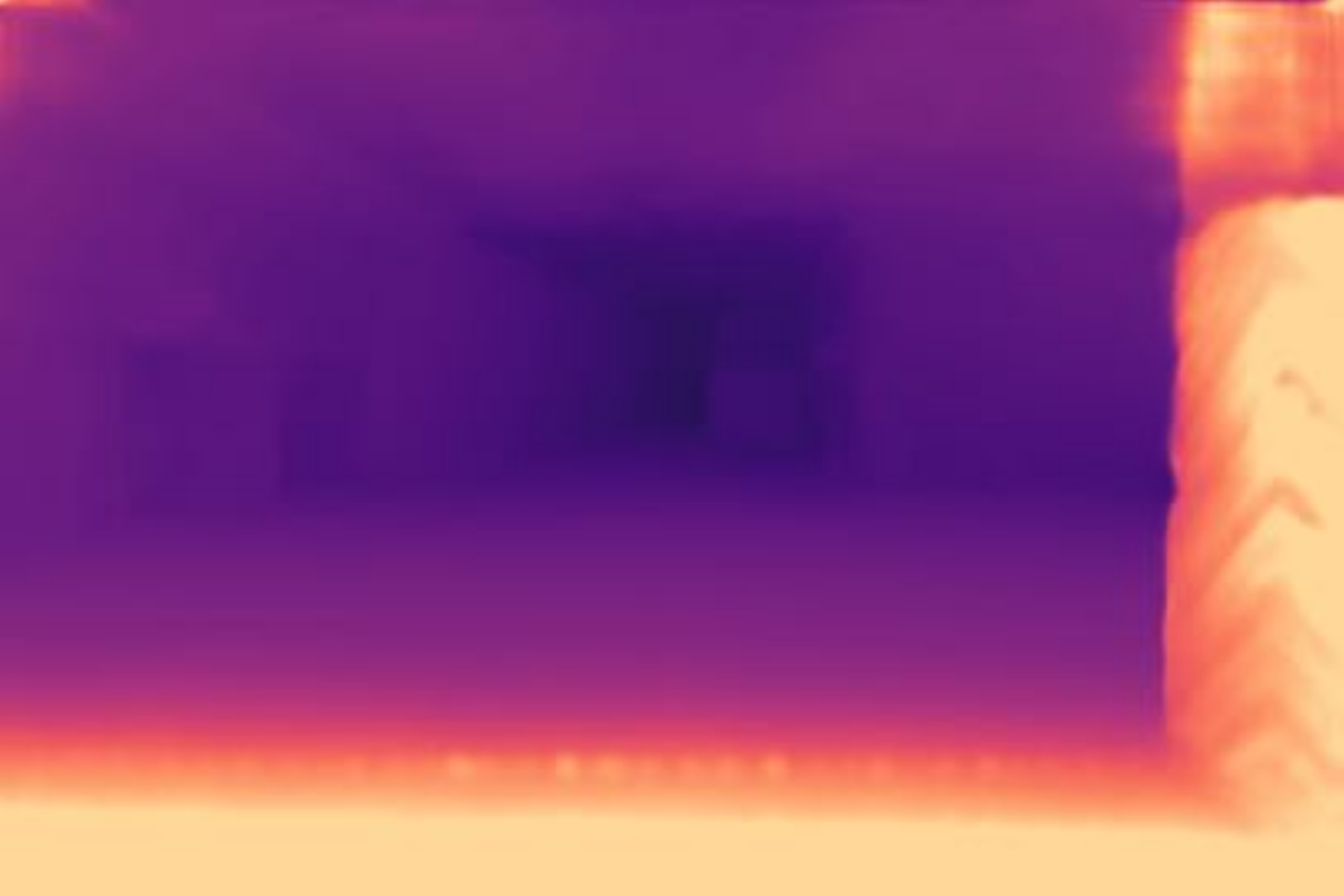} \qquad\qquad\quad &
\includegraphics[width=\iw,height=\ih]{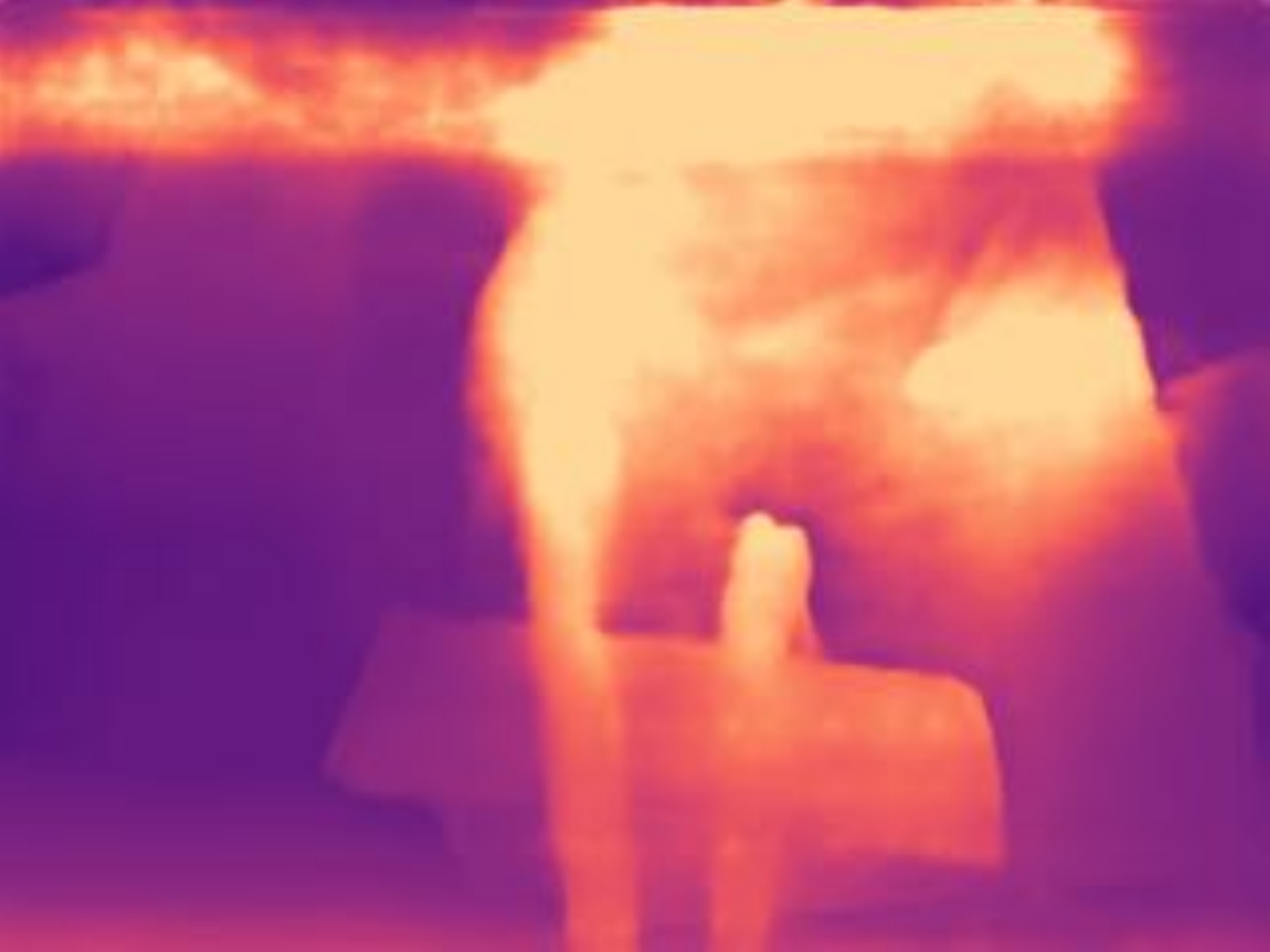} \\
\vspace{10mm}\\
\rotatebox[origin=c]{90}{\fontsize{\textw}{\texth} \selectfont GLPDepth\hspace{-270mm}}\hspace{24mm}
\includegraphics[width=\iw,height=\ih]{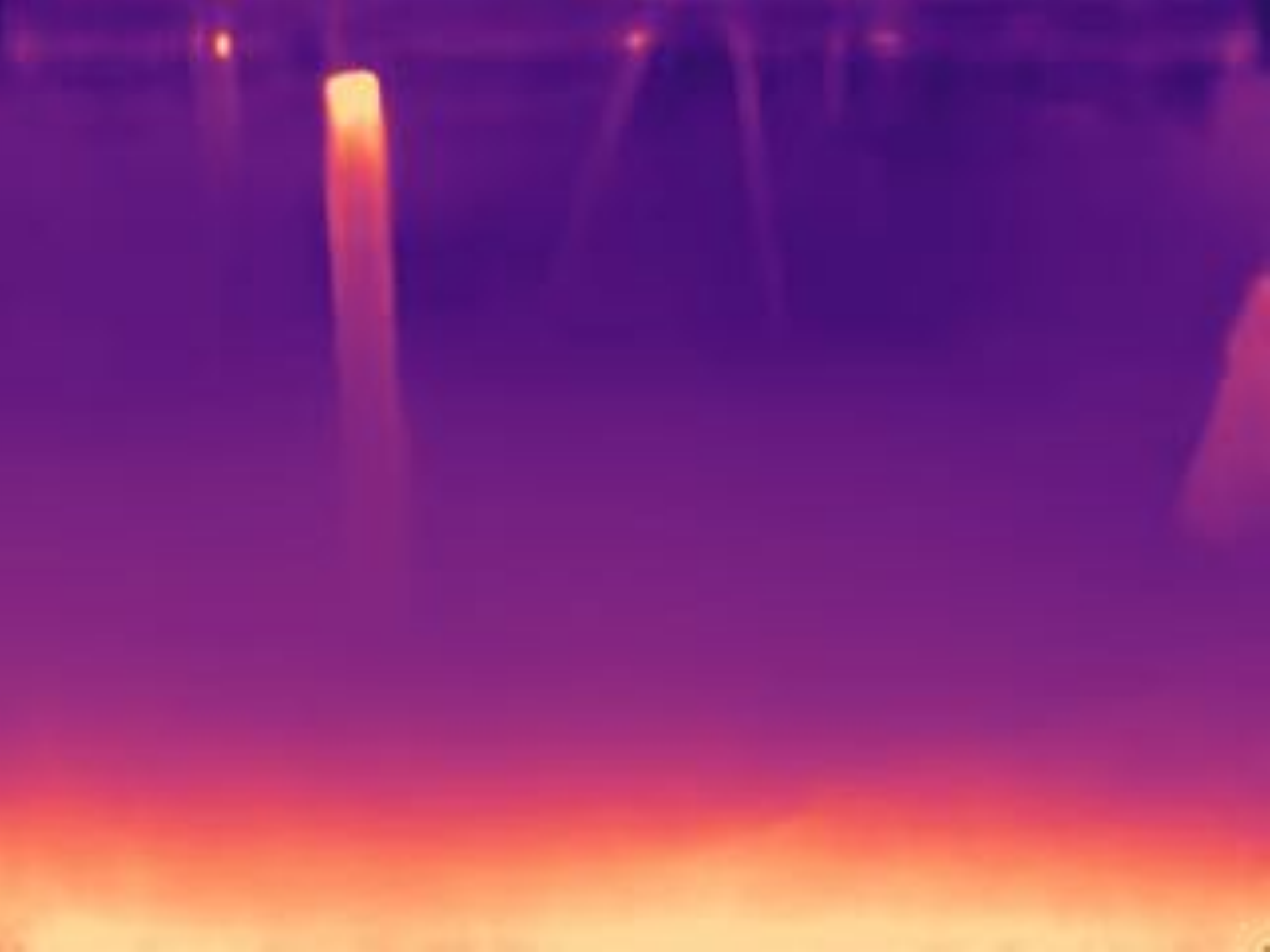} \qquad\qquad\quad & 
\includegraphics[width=\iw,height=\ih]{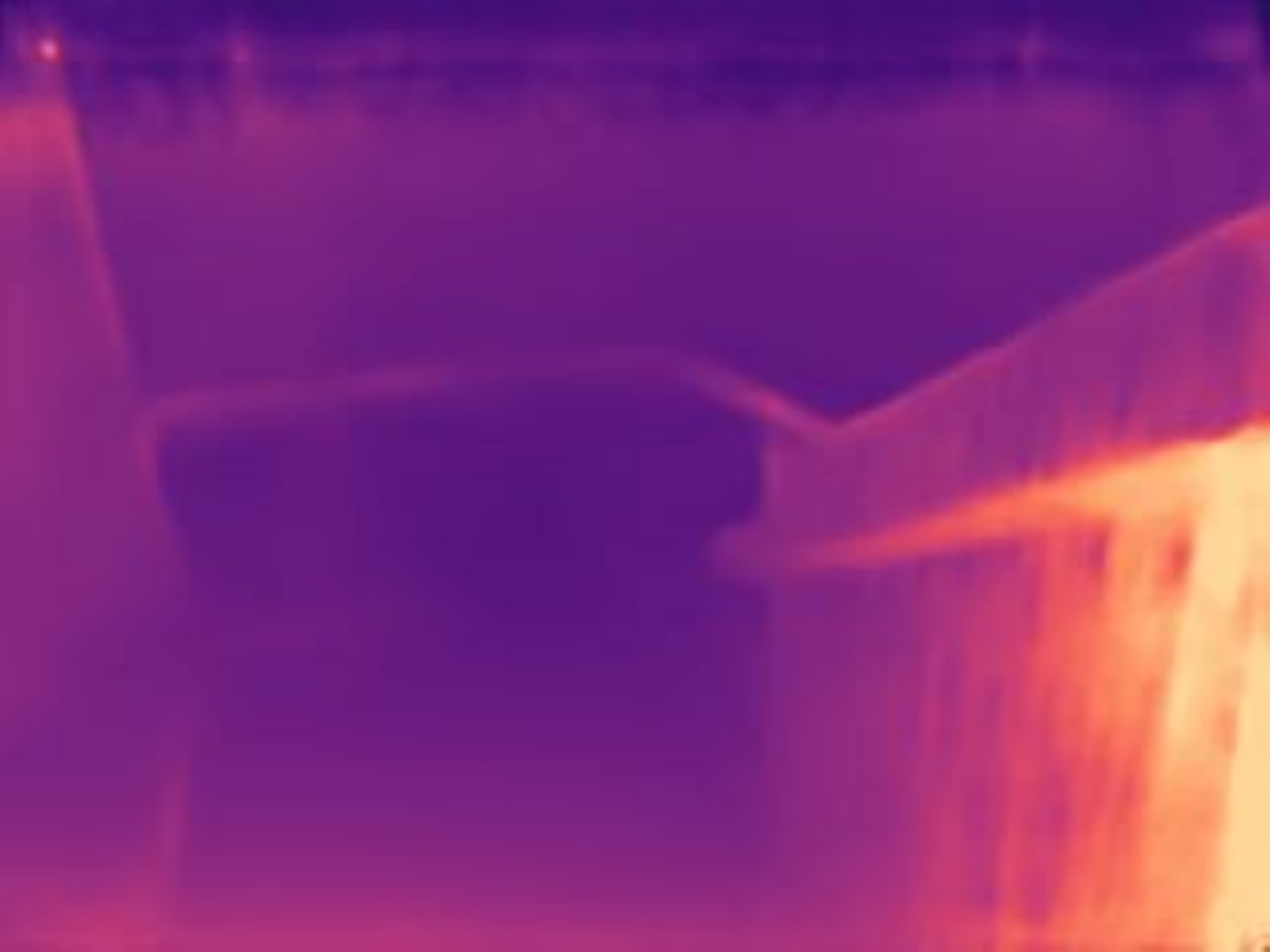} \qquad\qquad\quad & 
\includegraphics[width=\iw,height=\ih]{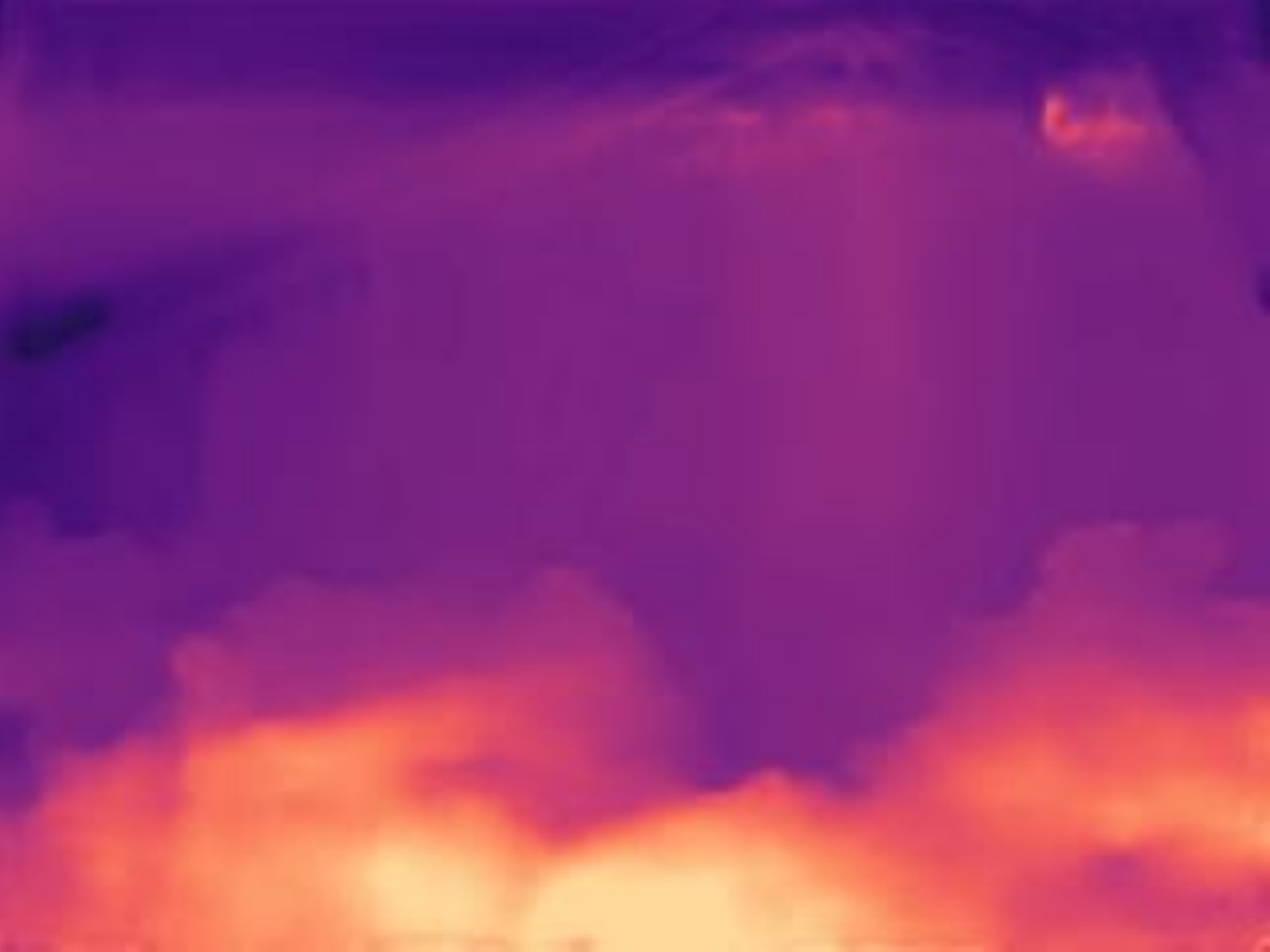} \qquad\qquad\quad &  
\includegraphics[width=\iw,height=\ih]{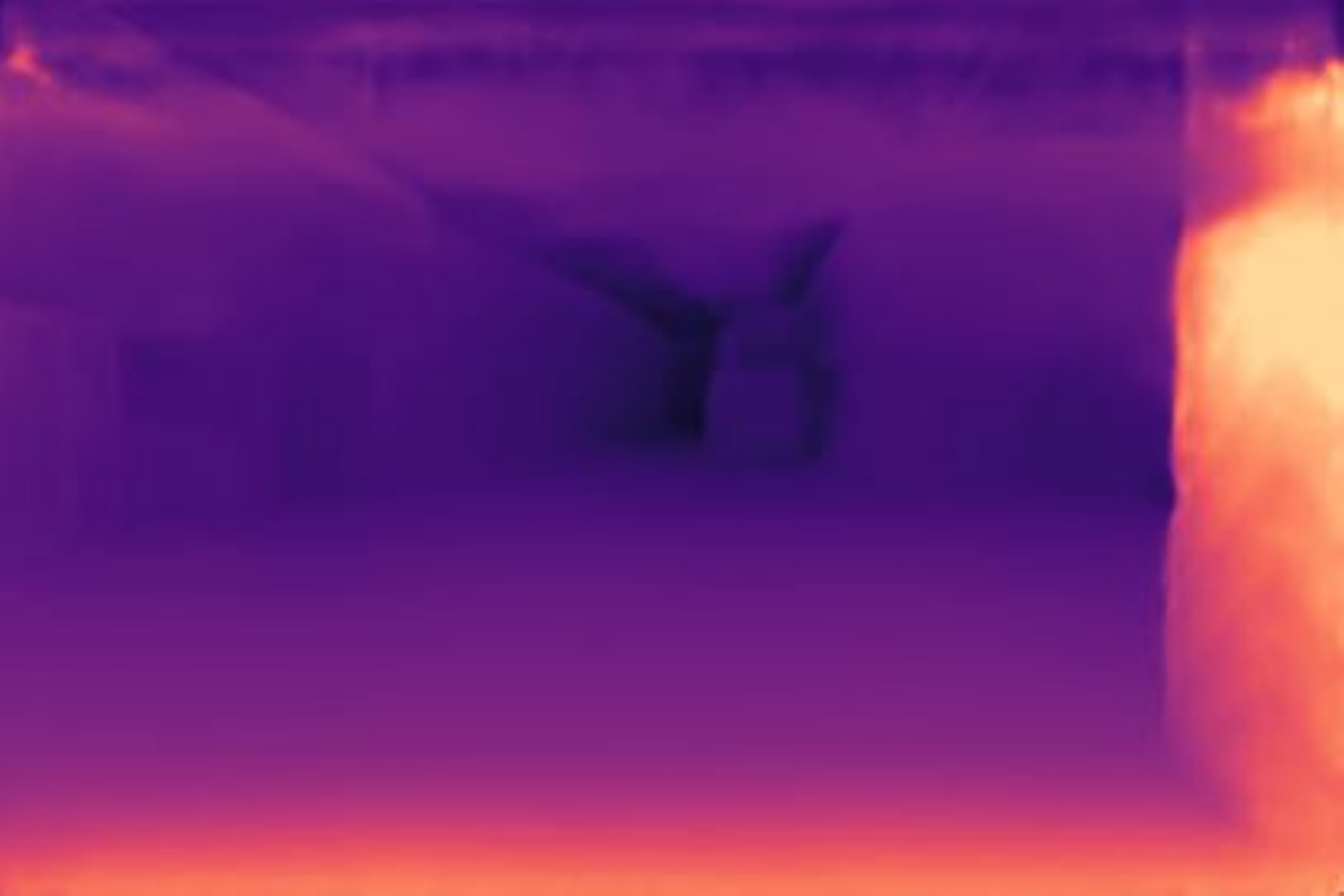} \qquad\qquad\quad &
\includegraphics[width=\iw,height=\ih]{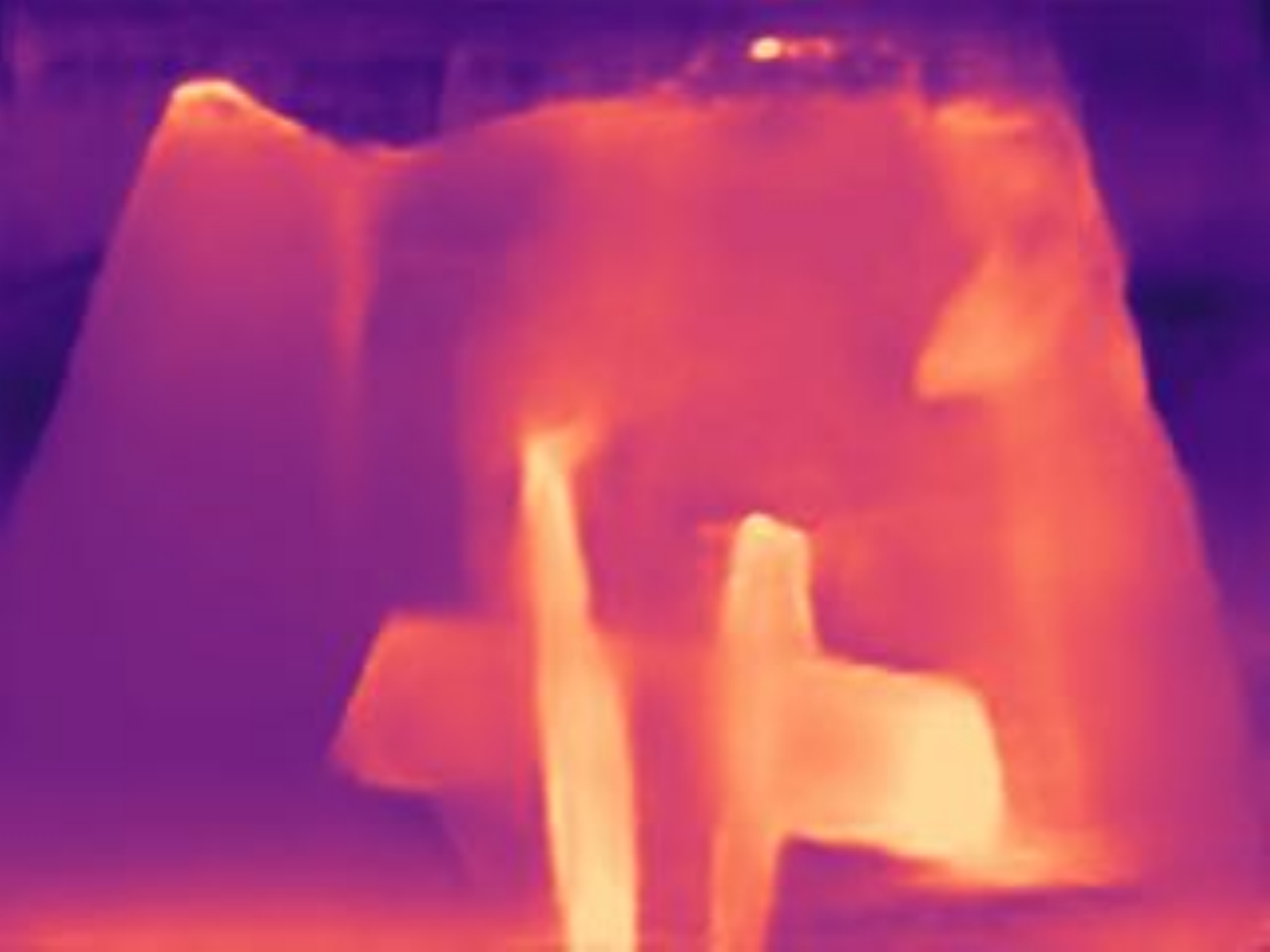} \\
\vspace{30mm}\\
\multicolumn{5}{c}{\fontsize{\w}{\h} \selectfont (d) Supervised Transformer-based methods } & 
\vspace{30mm}\\
\end{tabular}%
}
\caption{\textbf{Depth map results on out-of-distribution (RGBD, SUN3D, MVS, ETH3D, and Scenes11) datasets.}}
\label{figure_result_demons}
\end{figure*}

\subsection{Evaluation on out-of-distribution datasets}
\label{genealization-exp}
We compare the generalization performance of all the competitive models using public depth datasets captured at common indoor environments, including office workspaces, meeting rooms, and kitchen areas (RGBD \cite{sturm2012benchmark}, SUN3D \cite{xiao2013sun3d}), man-made indoor and outdoor environments (MVS \cite{ummenhofer2017demon}, ETH3D \cite{schops2017multi}), and synthetic scenes from graphics tools (Scenes11 \cite{ummenhofer2017demon}).
Interestingly, both qualitative and quantitative results in \figref{table_result_mvs}, \figref{figure_result_demons}, \tabref{table_result_indoor}\footnote[1]{Note that these \tabref{table_result_indoor} and \tabref{table_result_inoutdoor} are in the appendix} and \tabref{table_result_inoutdoor}\footnotemark[1] show that Transformer-based models better generalization performance even in the out-of-distribution datasets that have never been seen during network training.
Meanwhile, CNN-based models predict unreliable depth maps that have lost the detail of object boundaries and produce significant errors in texture-less regions, such as the wall of a building. 
The experiments demonstrate that Transformers are more robust than CNNs for environmental changes. 
Another interesting observation is that MF-ConvNeXt generally outperforms all the other CNN-based models and produces depth results comparable to other Transformer-based models. On the other hand, MF-RegionViT fails to estimate depth accurately, even though Transformer-based.
A detailed analysis of why Transformers and MF-ConvNeXt show better generalization performance is provided in the following section.

\subsection{Analysis of texture-/shape-bias on state-of-the-art methods and various backbone networks}
\label{texture-shifted-exp}

In this section, we verify the intrinsic properties of CNNs and Transformers that lead to the robustness of depth estimation to environmental changes.
We hypothesize that CNNs and transformers identify texture and shape information as key visual cues for depth estimation, respectively, inspired by the work~\cite{tuli2021convolutional,naseer2021intriguing}.
Thus, we synthetically generate the texture-shifted datasets in \secref{data-generation}.
Then, we validate the texture-/shape-bias of the model by comparing the performance of the competitive methods on the generated datasets in \secref{artifical-texture-exp}. 
Finally, we analyze the internal representation of each neural network structure by measuring centered kernel alignment (CKA) in \secref{feature-representation}.

\begin{figure*}[t!]
\centering
\newcommand\w{240}
\newcommand\h{220}
\newcommand\iw{80cm}
\newcommand\ih{35cm}
\newcommand\textw{120}
\newcommand\texth{200}
\resizebox{\linewidth}{!}{%
\begin{tabular}{ccccccc}
\multicolumn{2}{c}{\fontsize{\w}{\h} \selectfont Watercolor } & 
\multicolumn{2}{c}{\fontsize{\w}{\h} \selectfont Pencil-sketch } & 
\multicolumn{2}{c}{\fontsize{\w}{\h} \selectfont Style-transfer }  \\
\vspace{30mm}\\
\rotatebox[origin=c]{90}{\fontsize{\textw}{\texth}\selectfont Input Images\hspace{-330mm}}\hspace{10mm} 
\includegraphics[width=\iw,height=\ih]{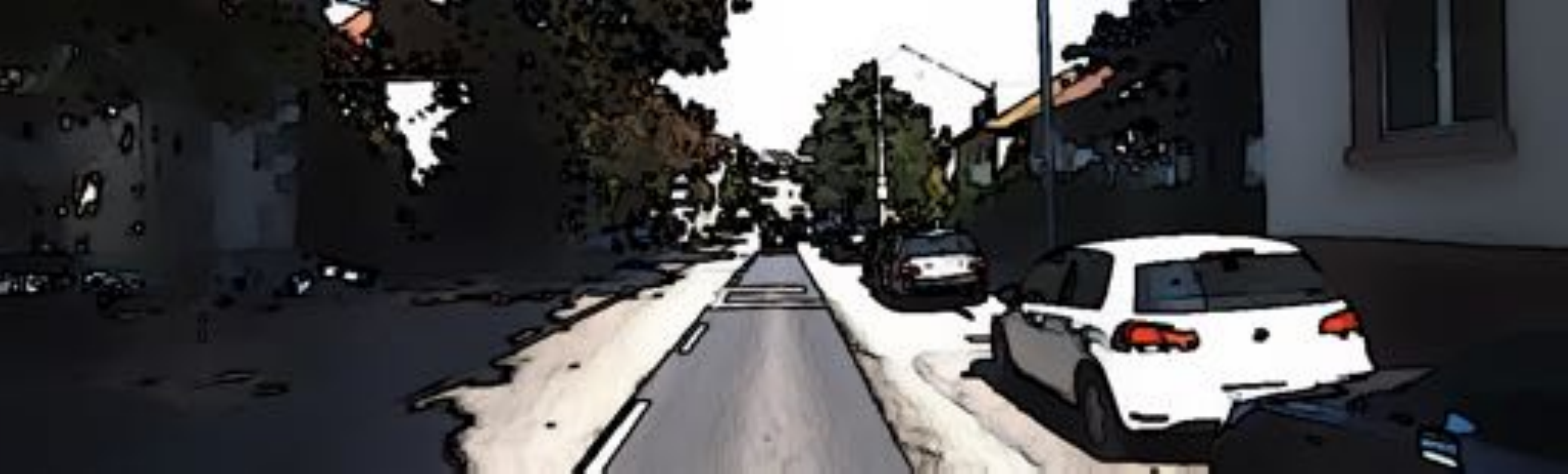} \qquad\qquad\quad &  
\includegraphics[width=\iw,height=\ih]{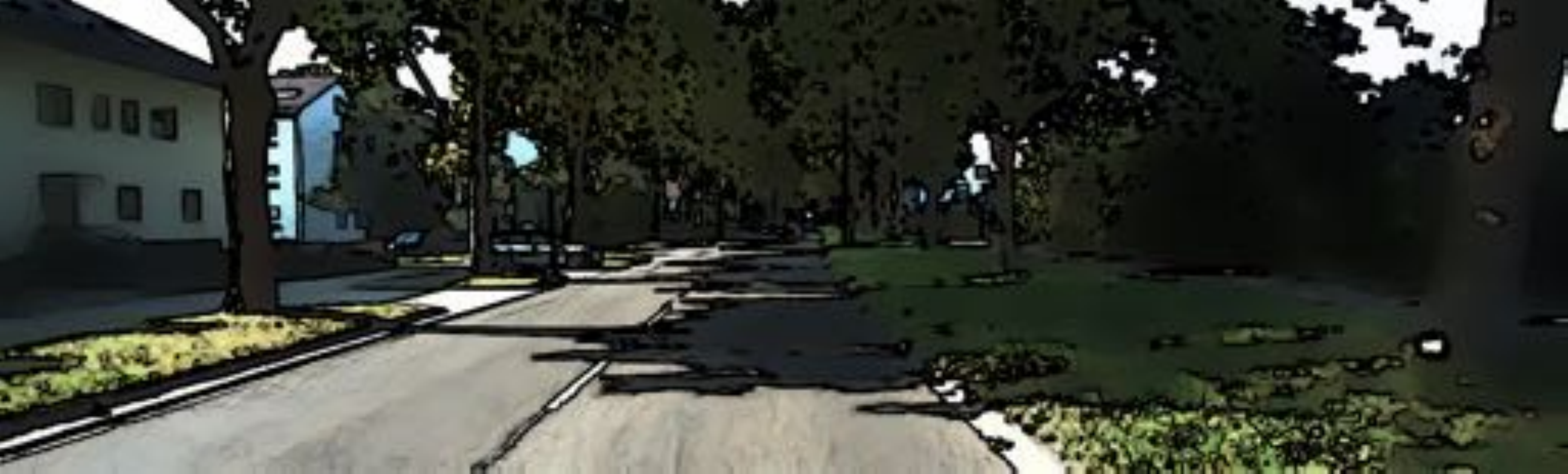} \qquad\qquad\quad &
\includegraphics[width=\iw,height=\ih]{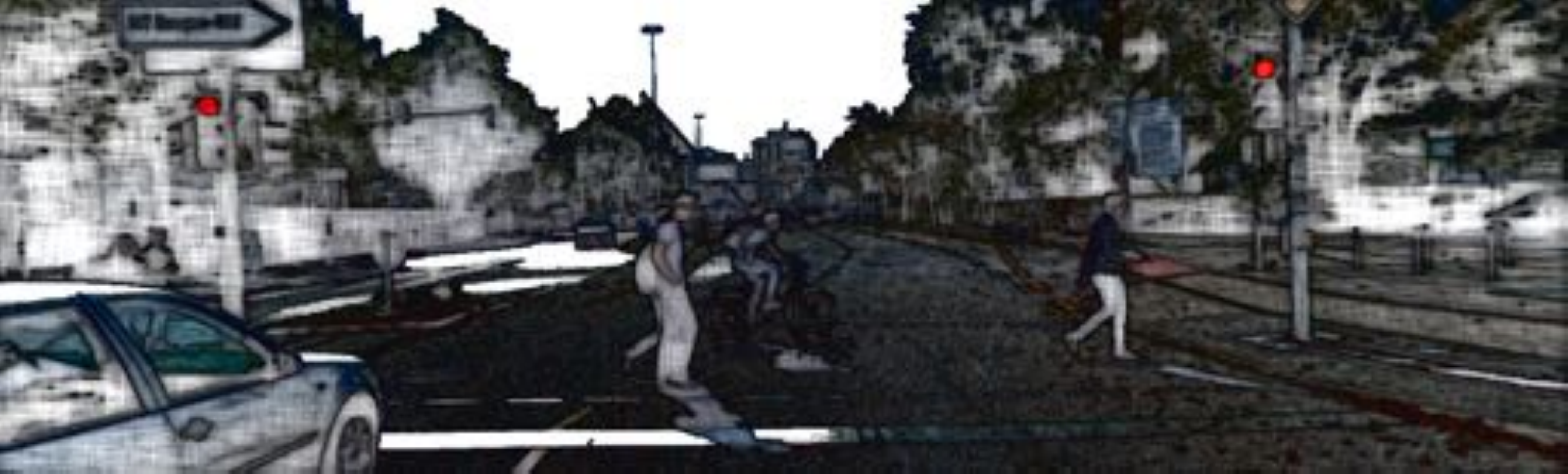} \qquad\qquad\quad &  
\includegraphics[width=\iw,height=\ih]{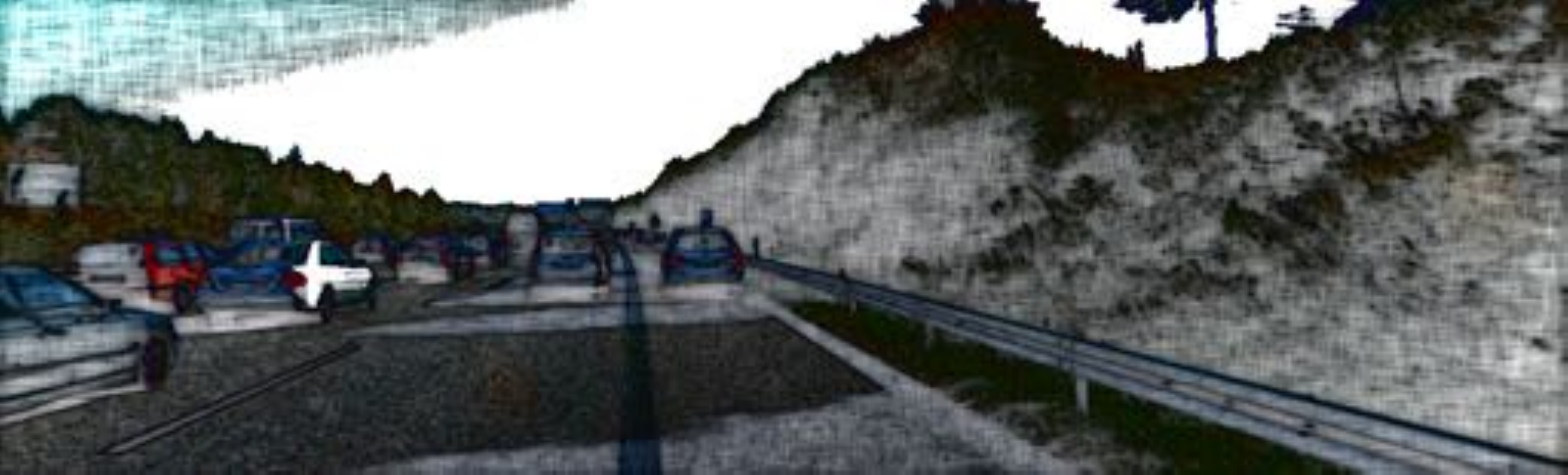} \qquad\qquad\quad &  
\includegraphics[width=\iw,height=\ih]{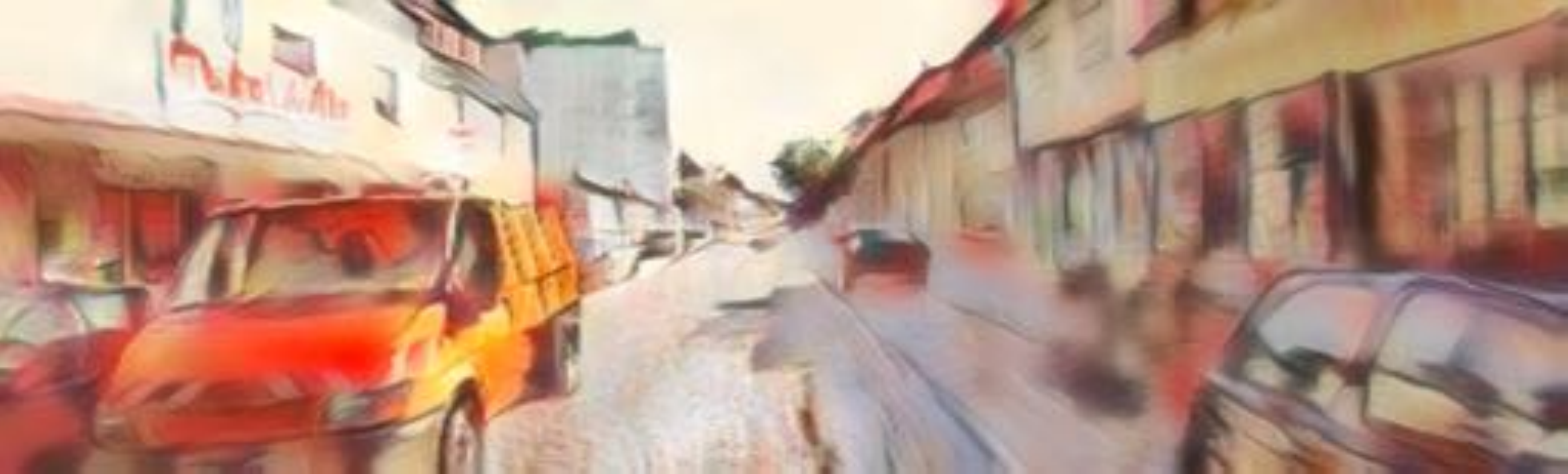} \qquad\qquad\quad &  
\includegraphics[width=\iw,height=\ih]{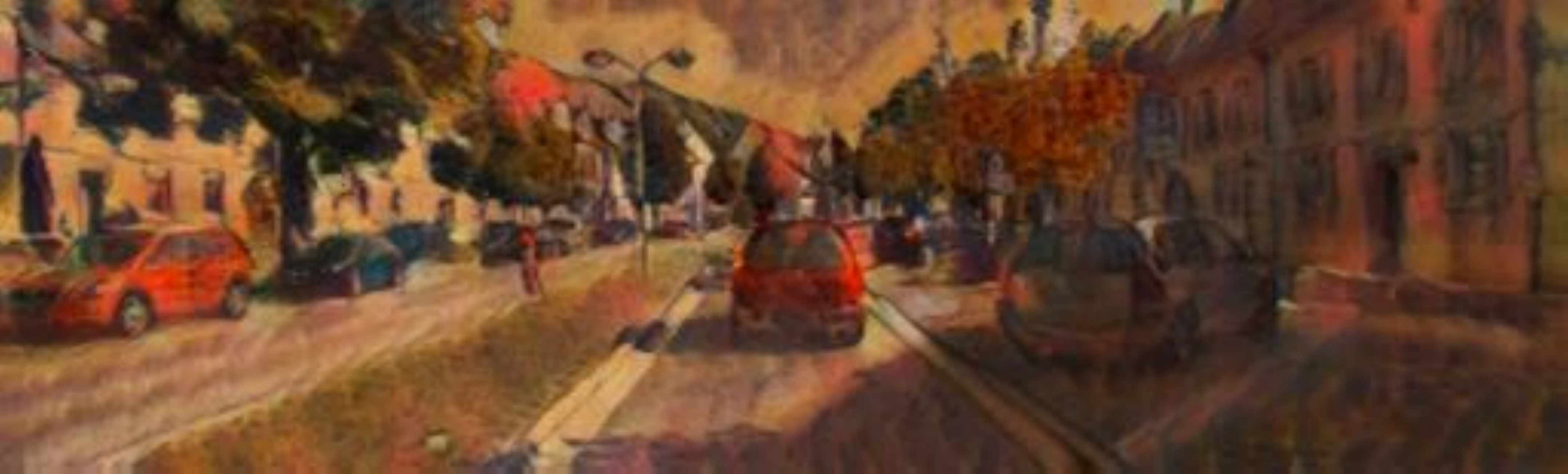}\\
\vspace{30mm}\\
\\\cmidrule{1-6}
\vspace{30mm}\\
\rotatebox[origin=c]{90}{\fontsize{\textw}{\texth}\selectfont Monodepth2\hspace{-310mm}}\hspace{10mm}
\includegraphics[width=\iw,height=\ih]{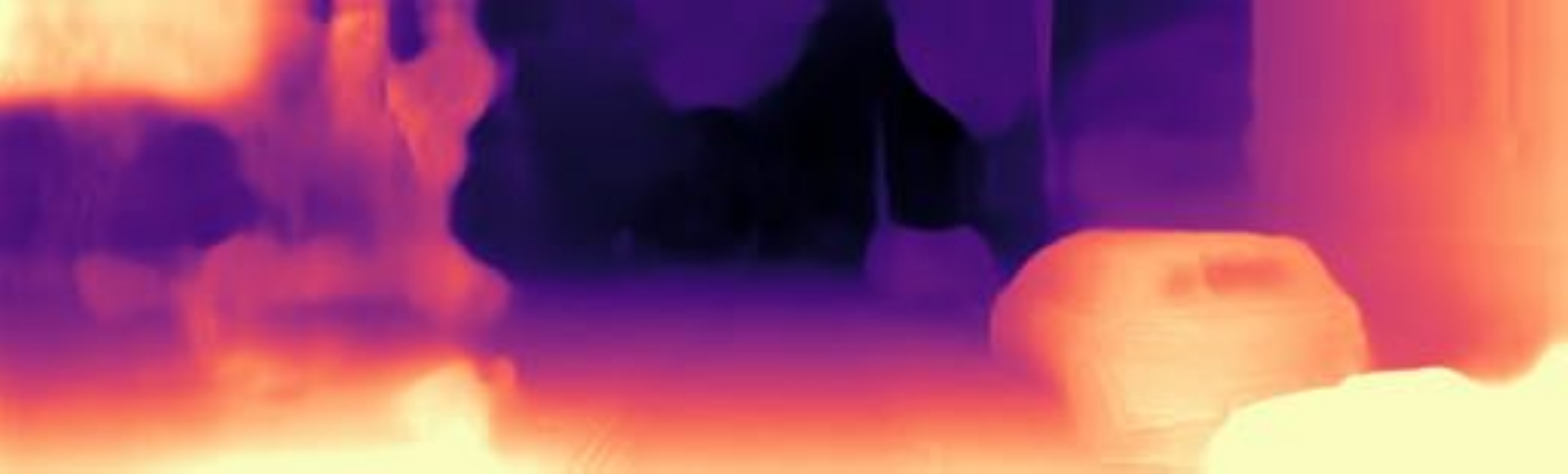} \qquad\qquad\quad &  
\includegraphics[width=\iw,height=\ih]{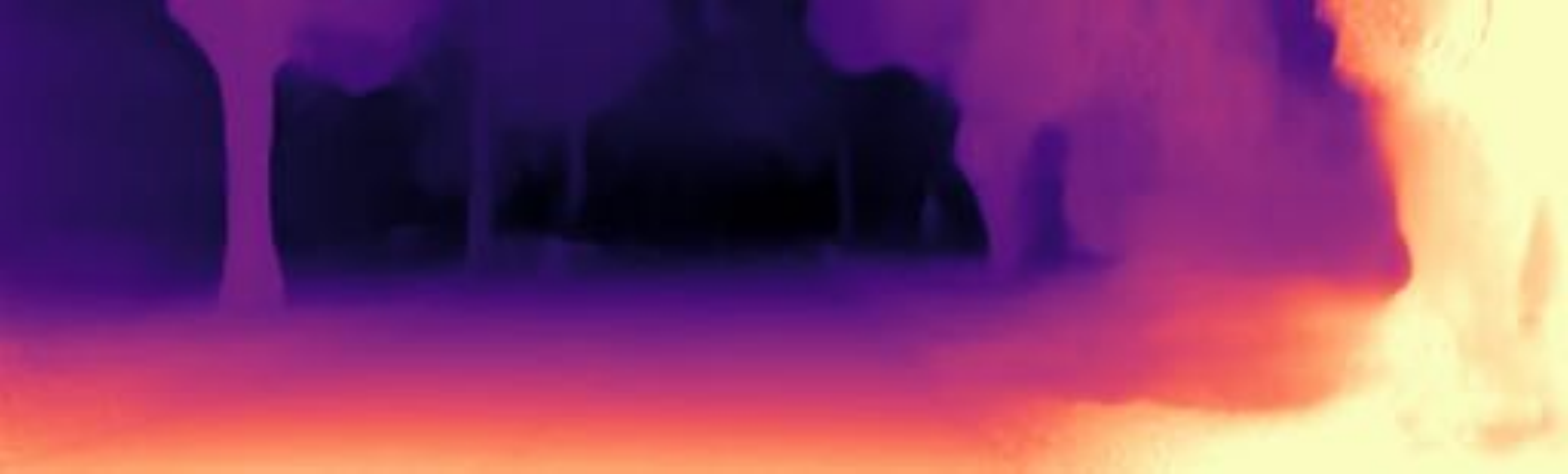} \qquad\qquad\quad & 
\includegraphics[width=\iw,height=\ih]{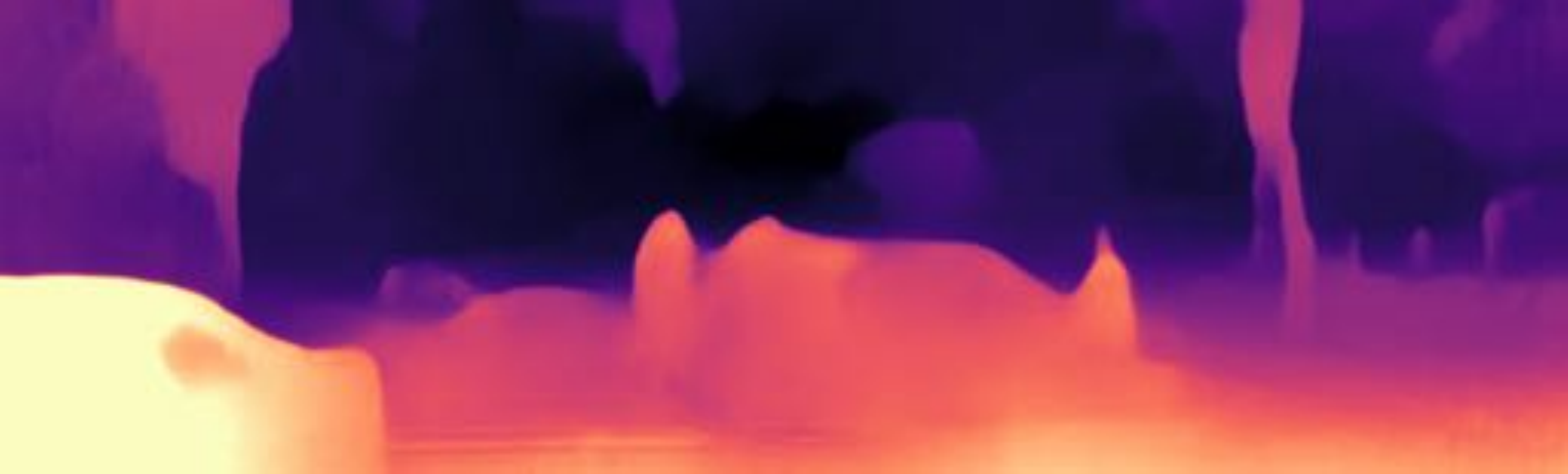} \qquad\qquad\quad & 
\includegraphics[width=\iw,height=\ih]{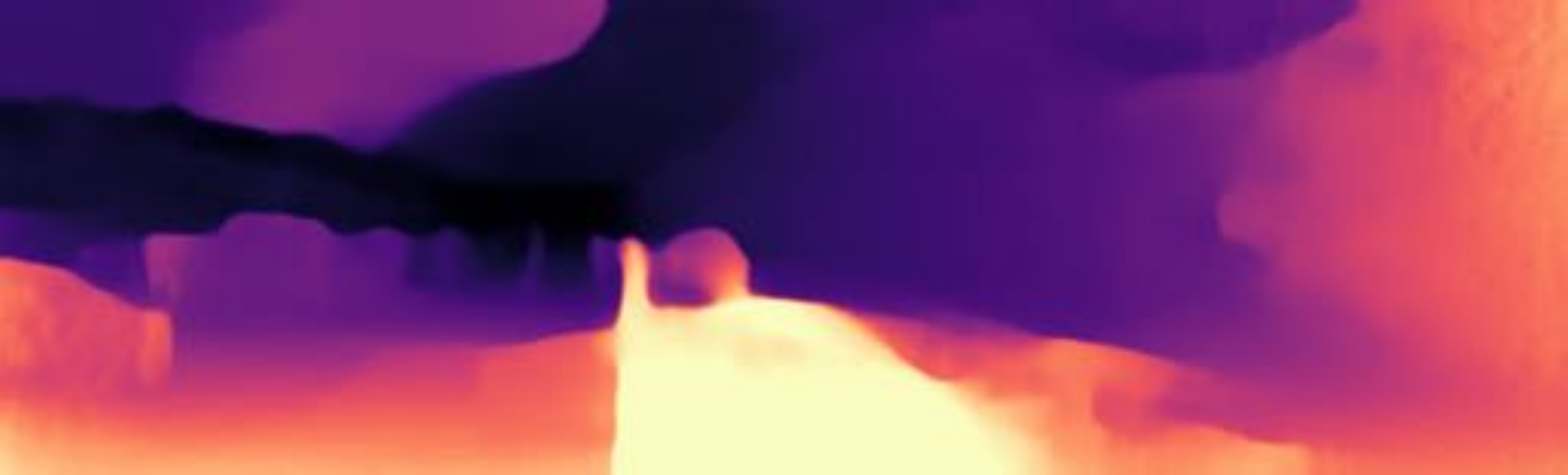} \qquad\qquad\quad & 
\includegraphics[width=\iw,height=\ih]{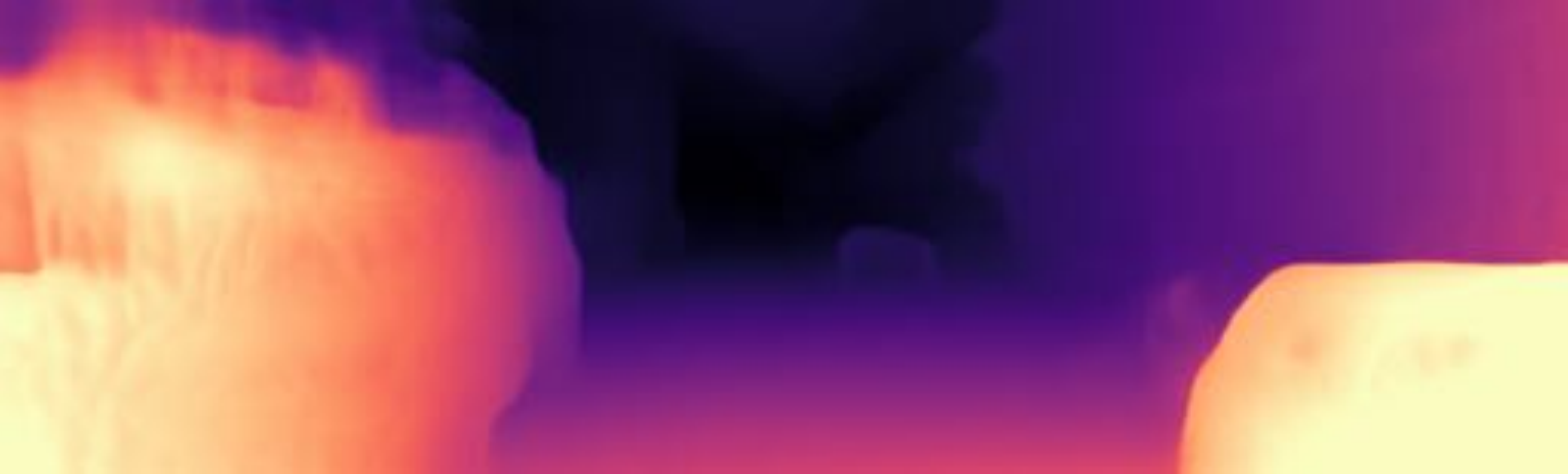} \qquad\qquad\quad & 
\includegraphics[width=\iw,height=\ih]{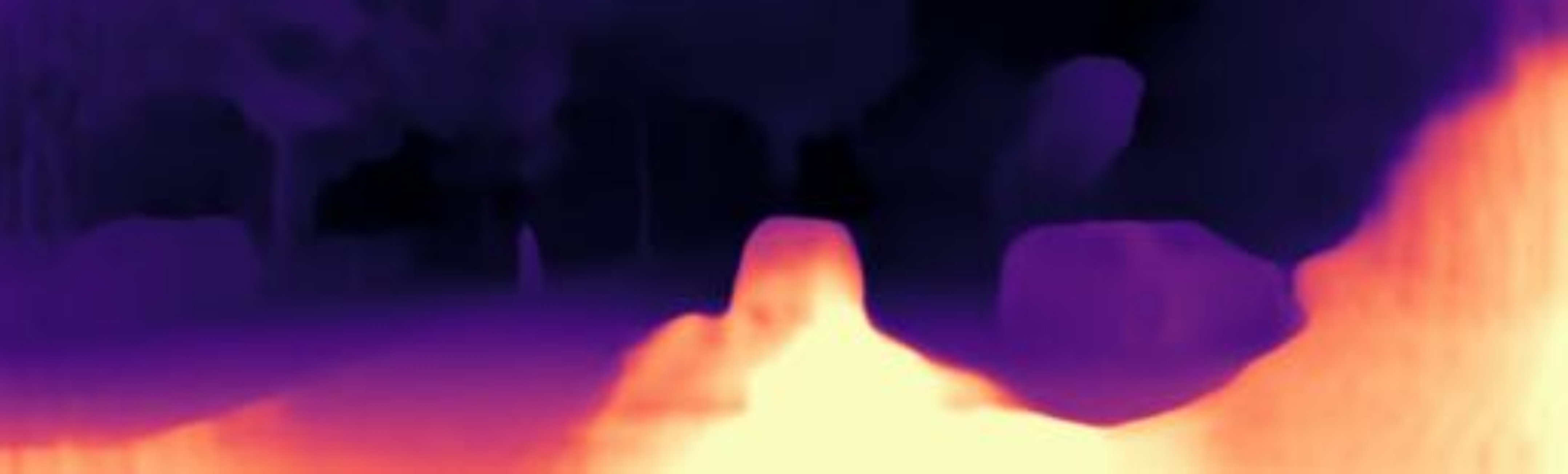}\\
\vspace{10mm}\\
\rotatebox[origin=c]{90}{\fontsize{\textw}{\texth}\selectfont PackNet-SfM\hspace{-300mm}}\hspace{20mm}
\includegraphics[width=\iw,height=\ih]{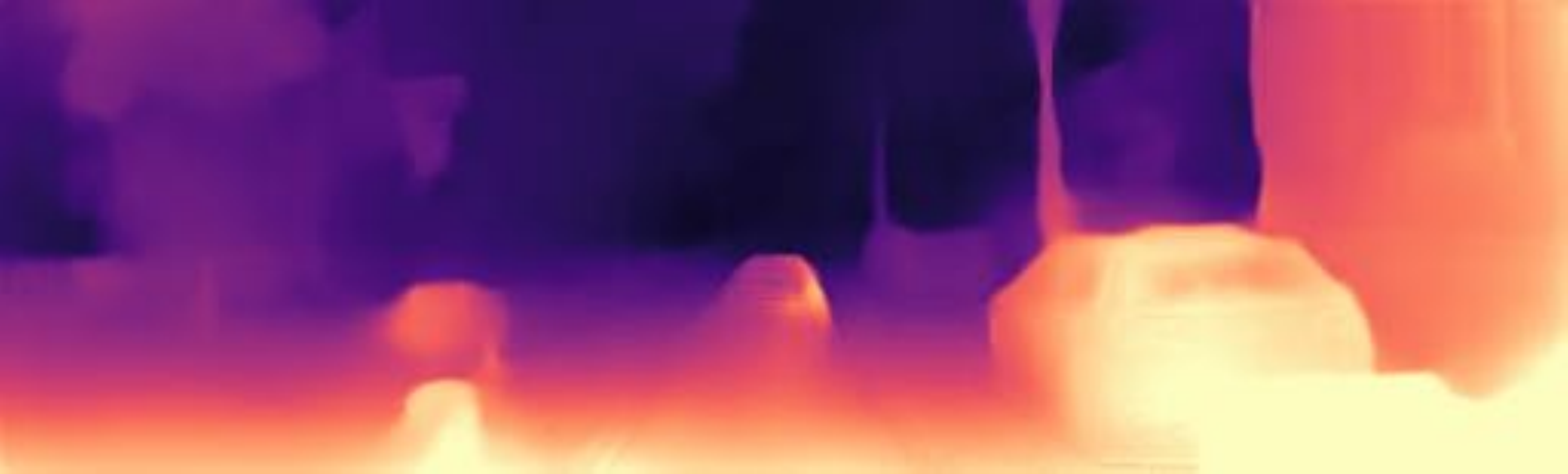}\qquad\qquad\quad &  
\includegraphics[width=\iw,height=\ih]{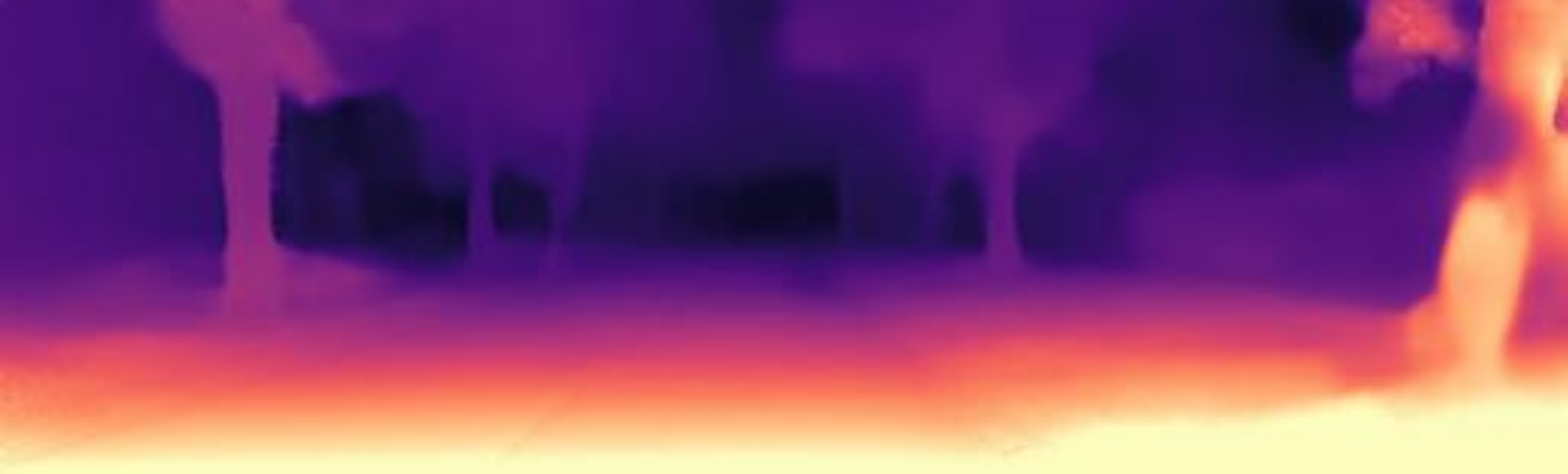}\qquad\qquad\quad & 
\includegraphics[width=\iw,height=\ih]{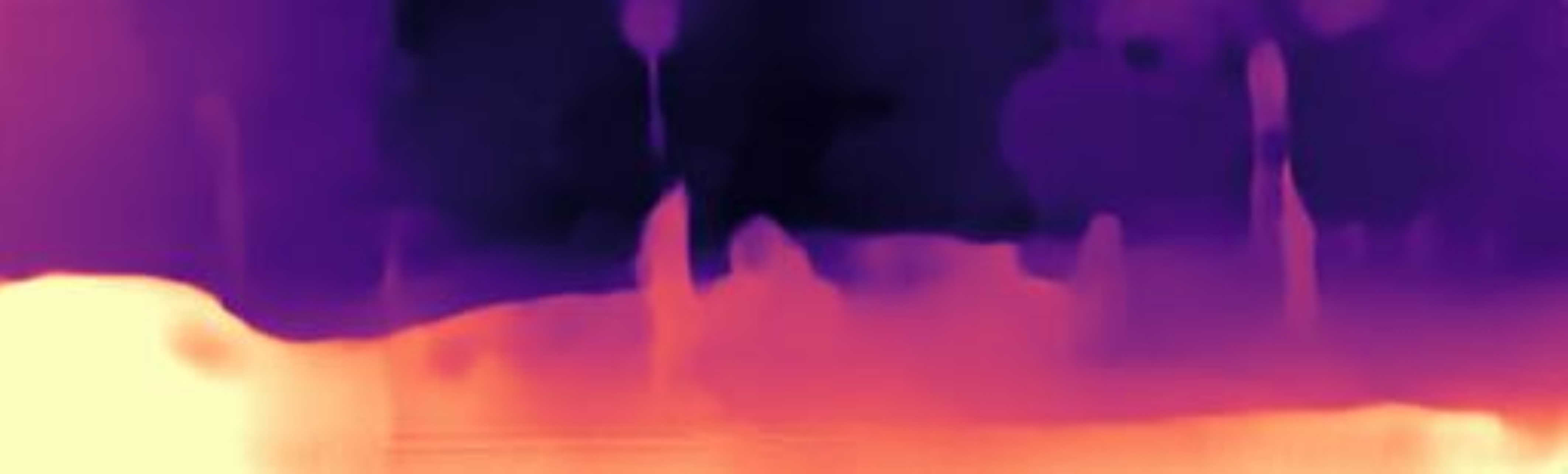}\qquad\qquad\quad & 
\includegraphics[width=\iw,height=\ih]{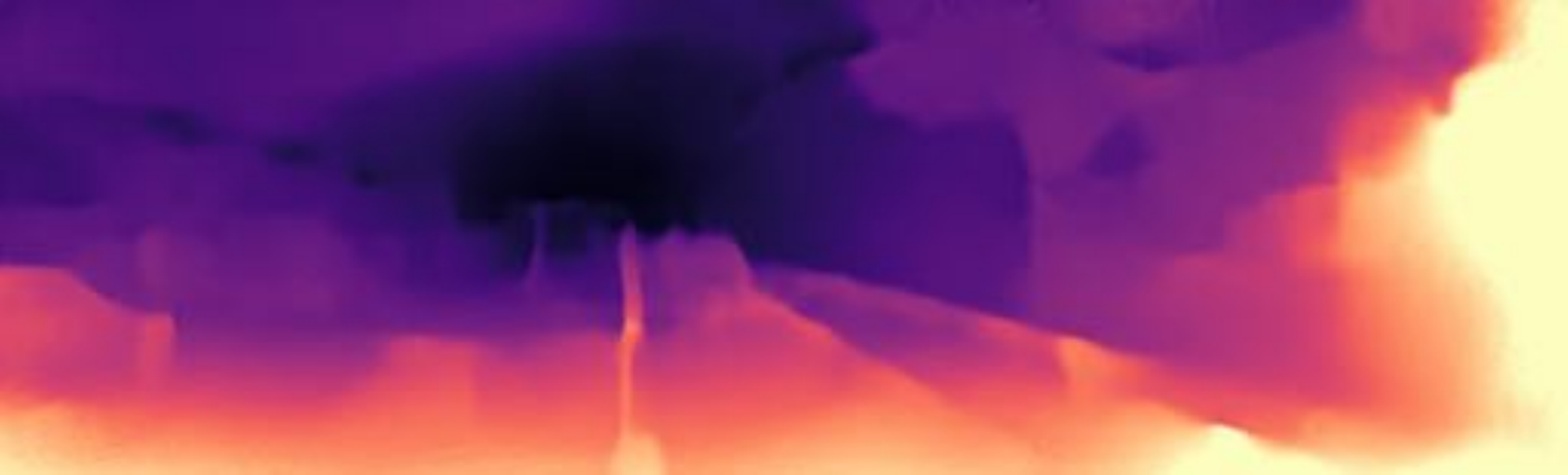}\qquad\qquad\quad & 
\includegraphics[width=\iw,height=\ih]{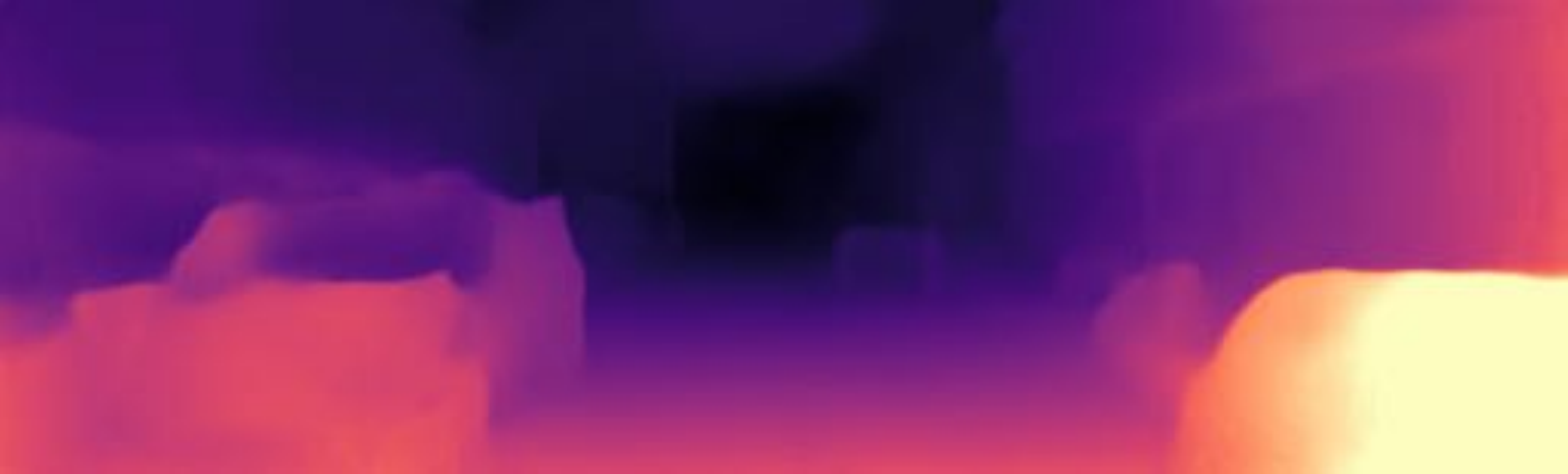}\qquad\qquad\quad & 
\includegraphics[width=\iw,height=\ih]{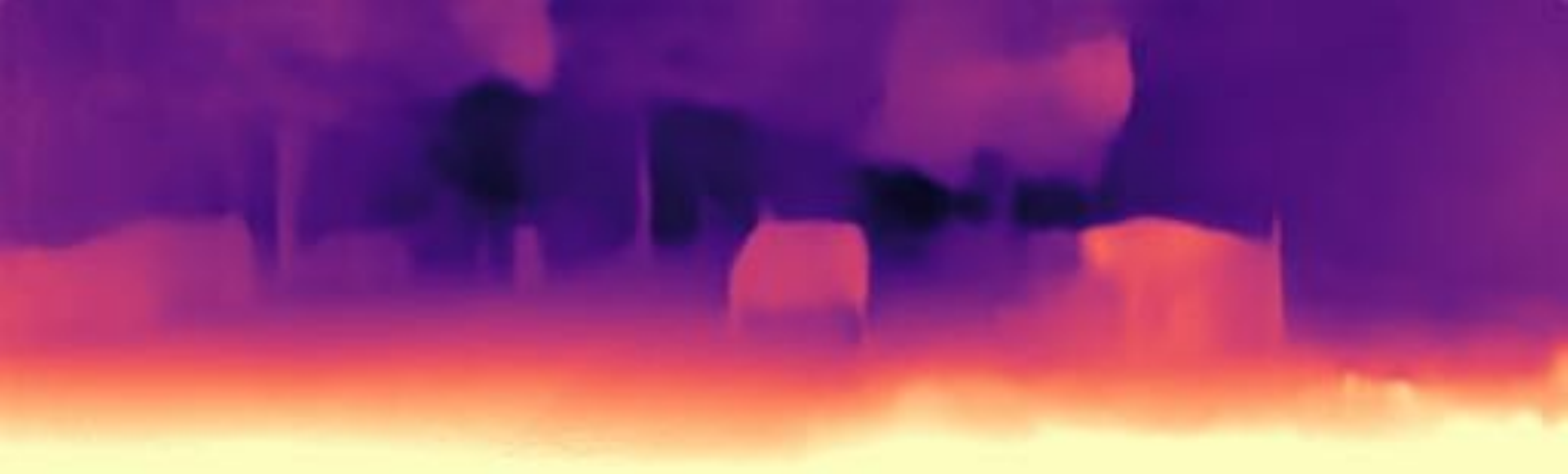}\\
\vspace{10mm}\\
\rotatebox[origin=c]{90}{\fontsize{\textw}{\texth}\selectfont R-MSFM6\hspace{-330mm}}\hspace{20mm}
\includegraphics[width=\iw,height=\ih]{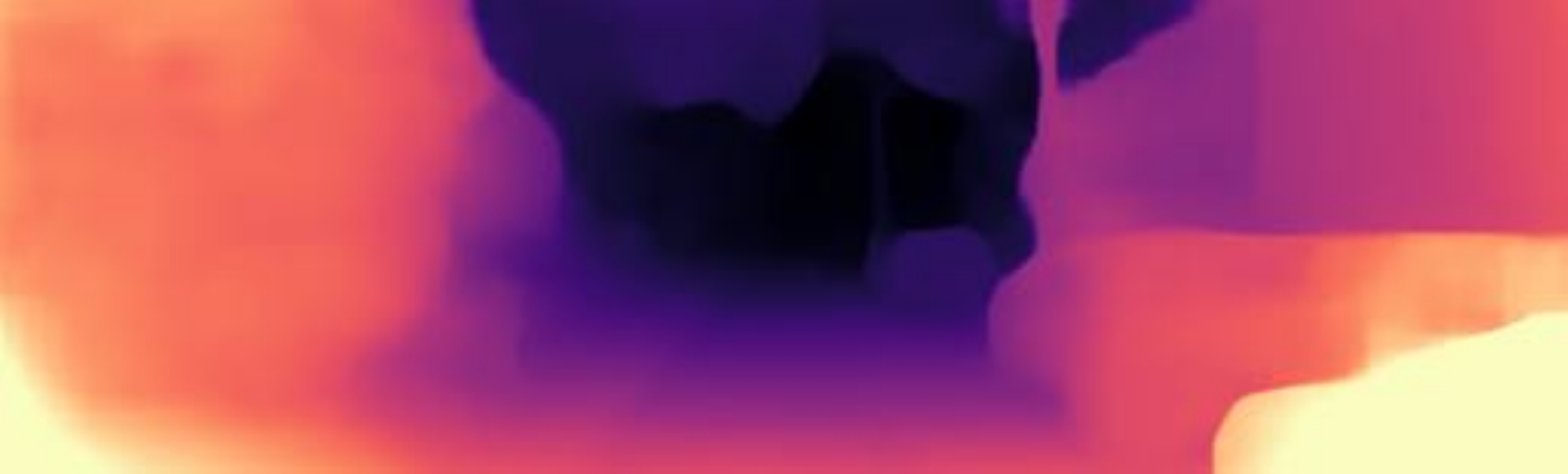}\qquad\qquad\quad &    
\includegraphics[width=\iw,height=\ih]{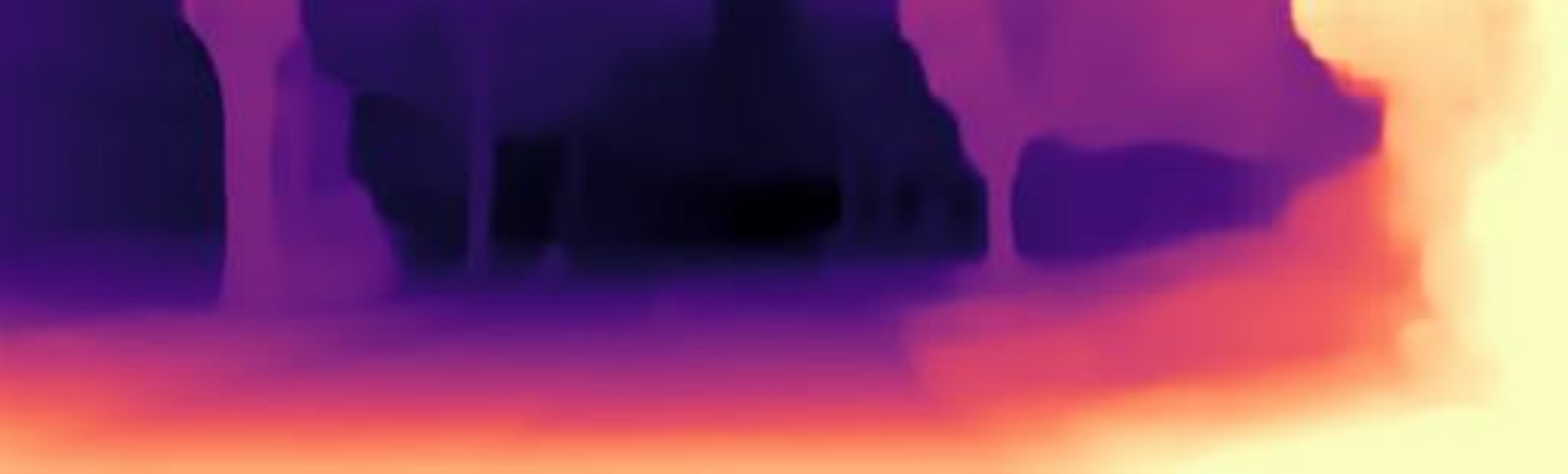}\qquad\qquad\quad &   
\includegraphics[width=\iw,height=\ih]{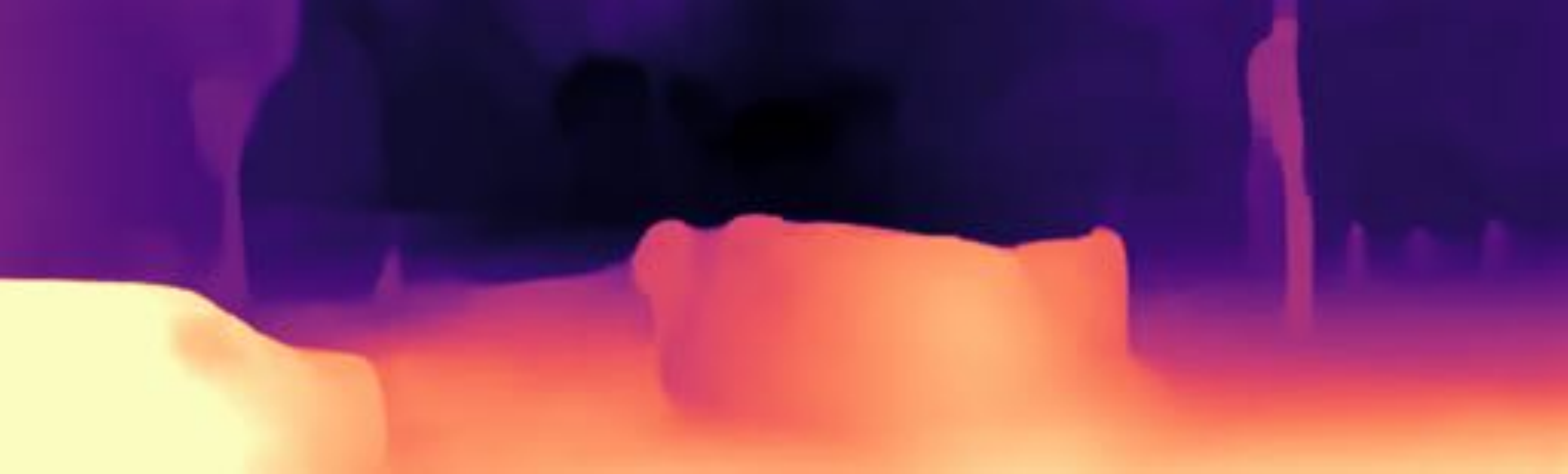}\qquad\qquad\quad &   
\includegraphics[width=\iw,height=\ih]{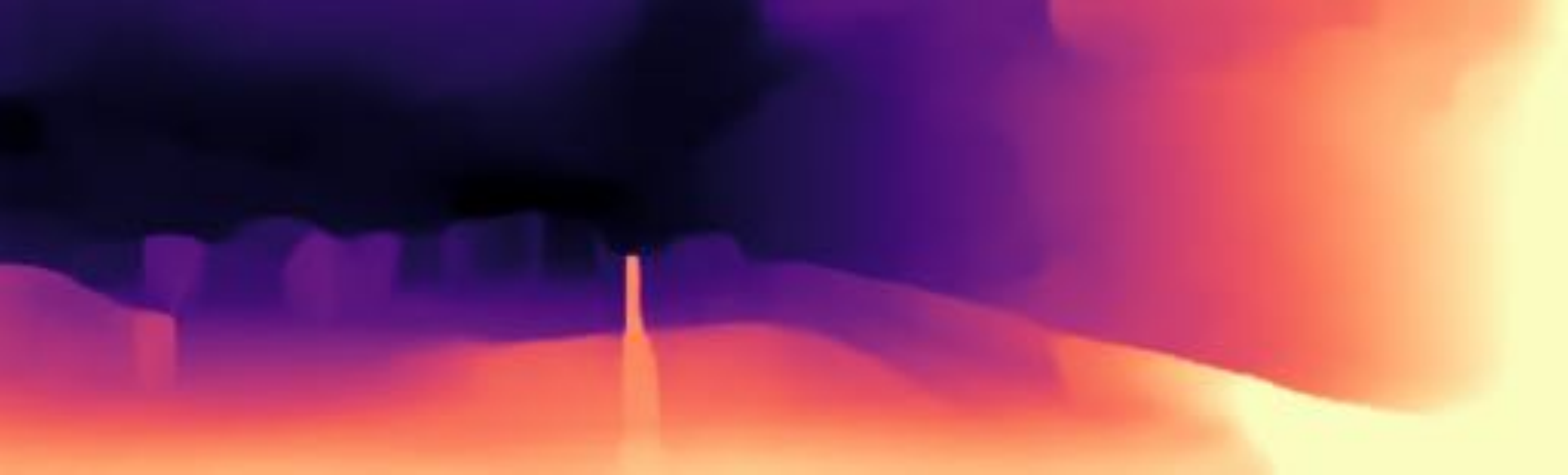}\qquad\qquad\quad &   
\includegraphics[width=\iw,height=\ih]{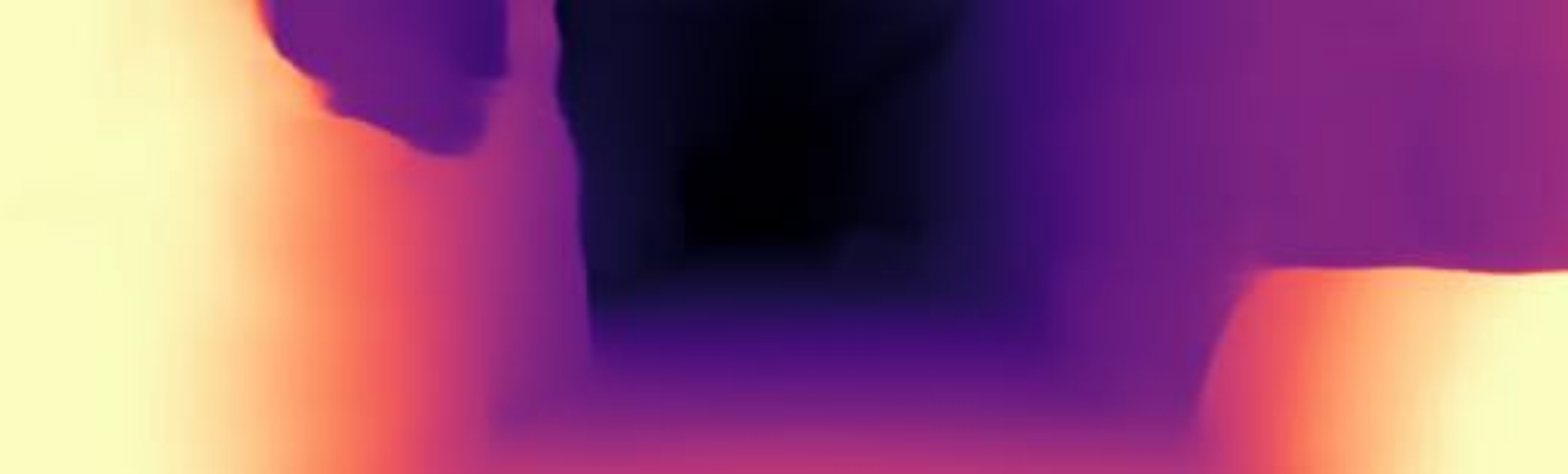}\qquad\qquad\quad &   
\includegraphics[width=\iw,height=\ih]{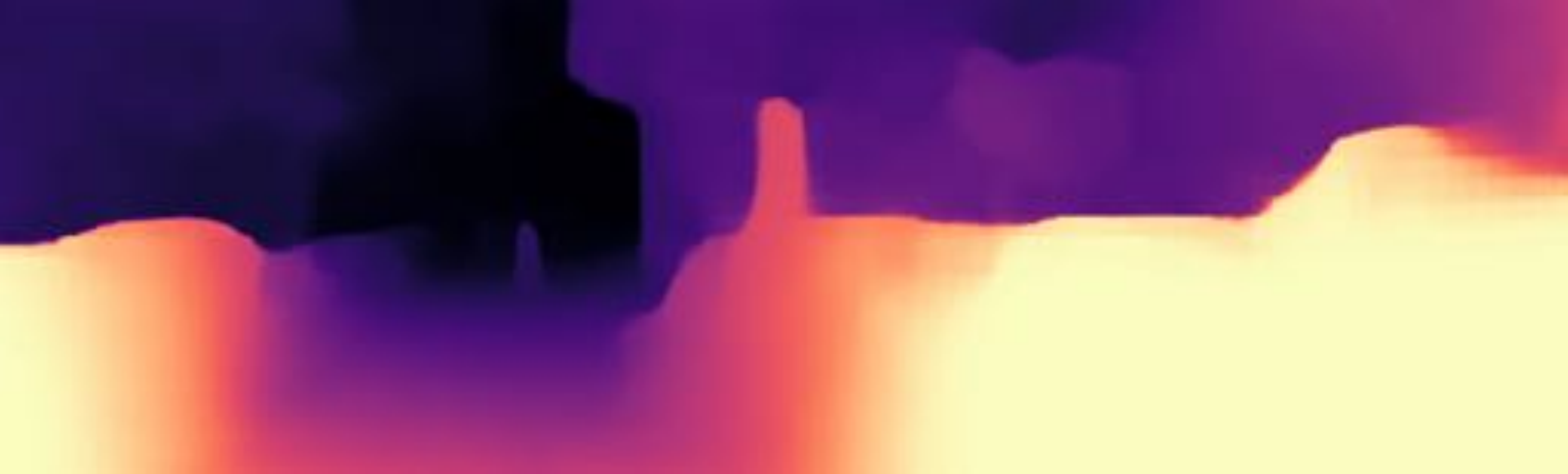}\\
\vspace{10mm}\\
\rotatebox[origin=c]{90}{\fontsize{\textw}{\texth}\selectfont MF-ConvNeXt\hspace{-320mm}}\hspace{20mm}
\includegraphics[width=\iw,height=\ih]{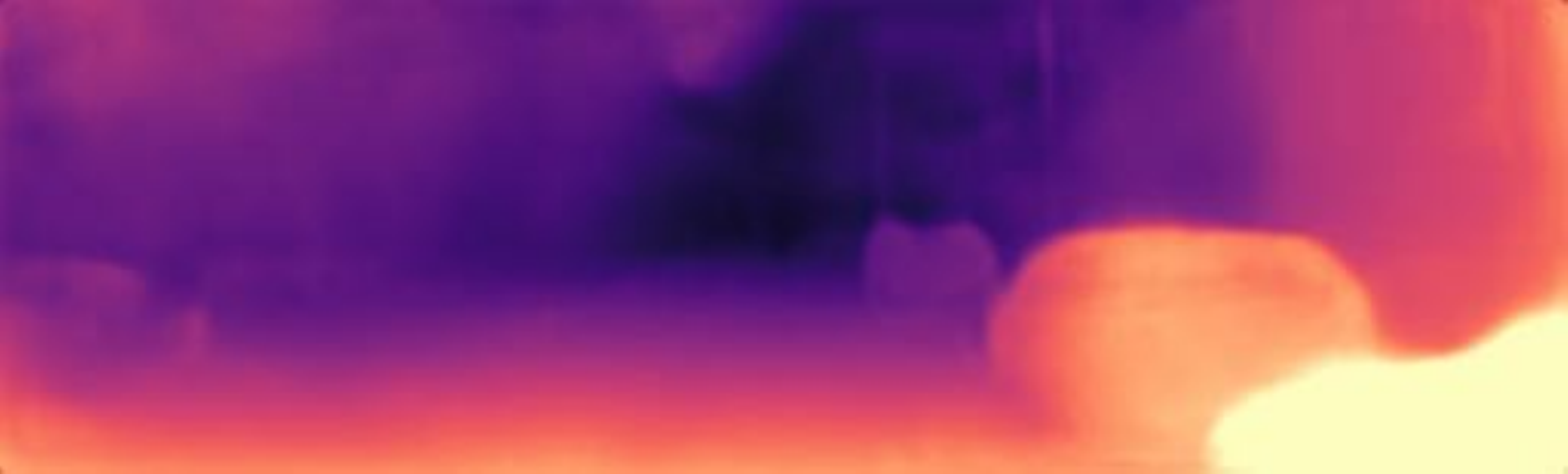}\qquad\qquad\quad &   
\includegraphics[width=\iw,height=\ih]{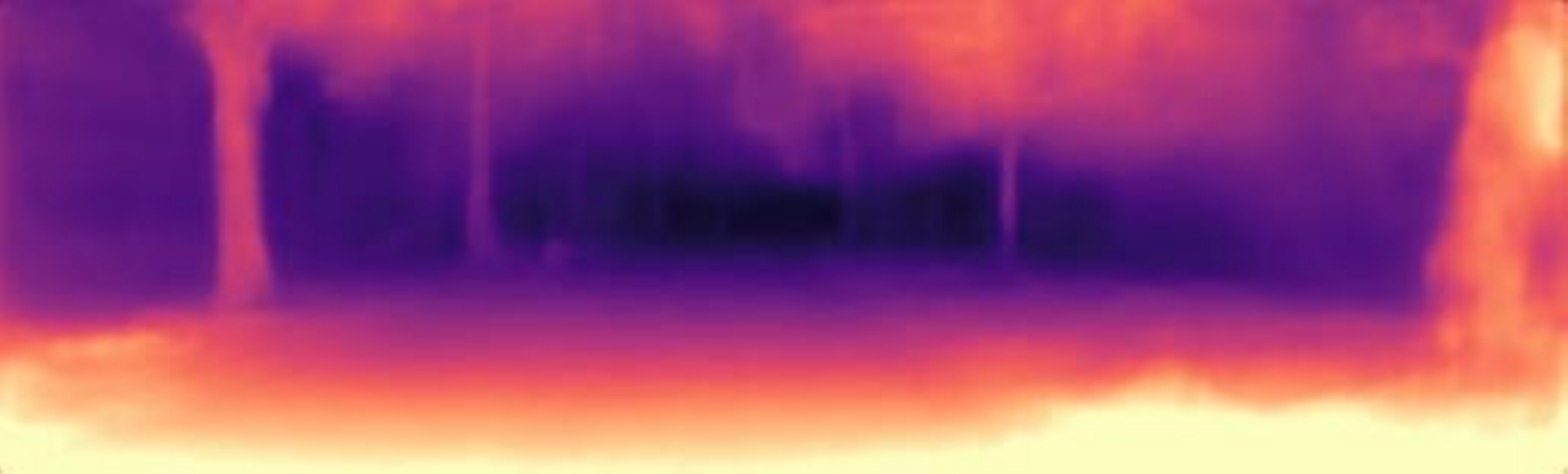}\qquad\qquad\quad &  
\includegraphics[width=\iw,height=\ih]{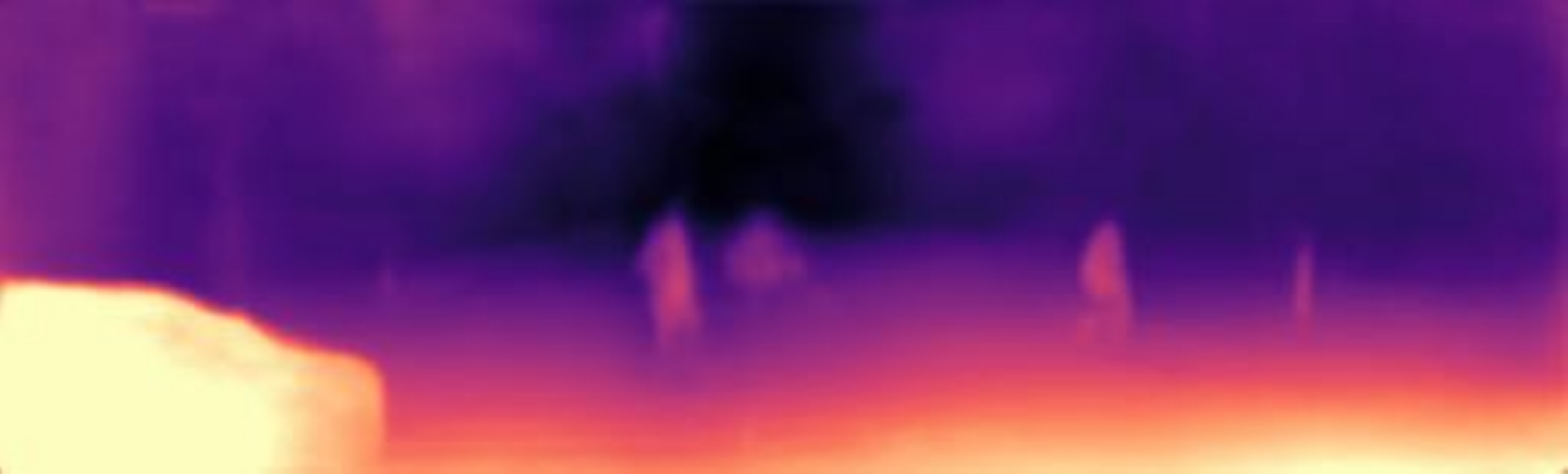}\qquad\qquad\quad &  
\includegraphics[width=\iw,height=\ih]{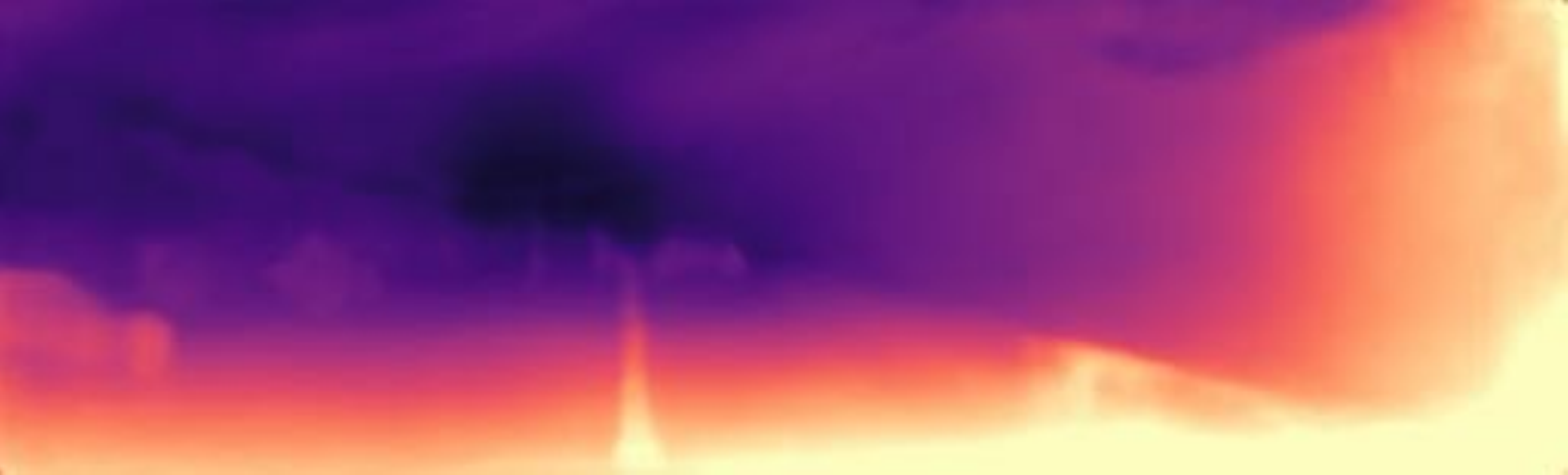}\qquad\qquad\quad &  
\includegraphics[width=\iw,height=\ih]{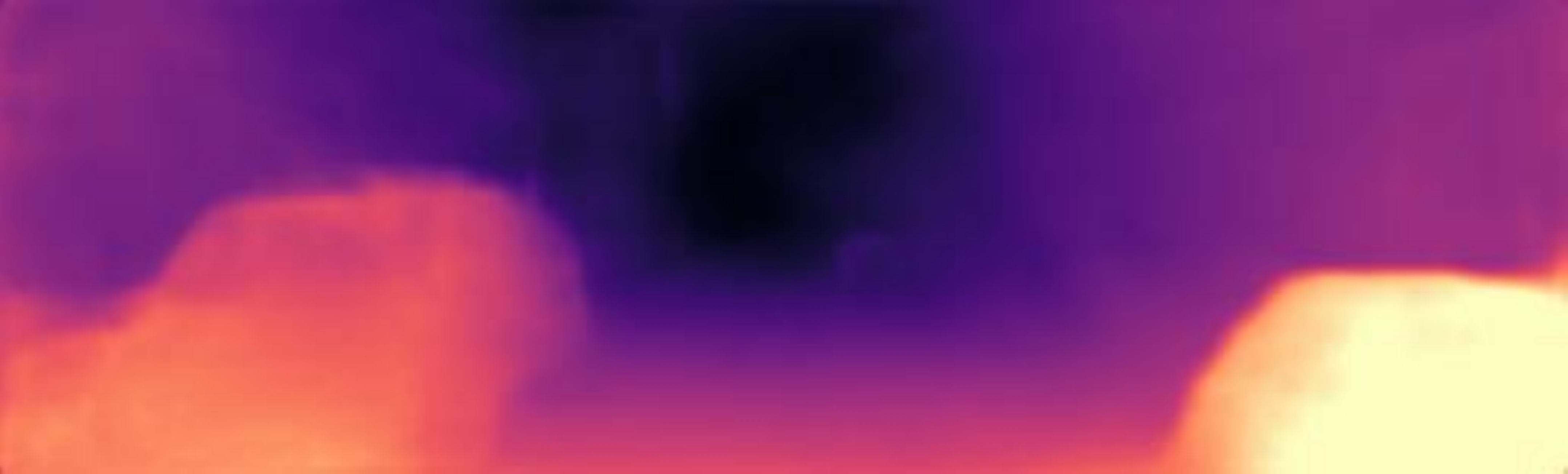}\qquad\qquad\quad &  
\includegraphics[width=\iw,height=\ih]{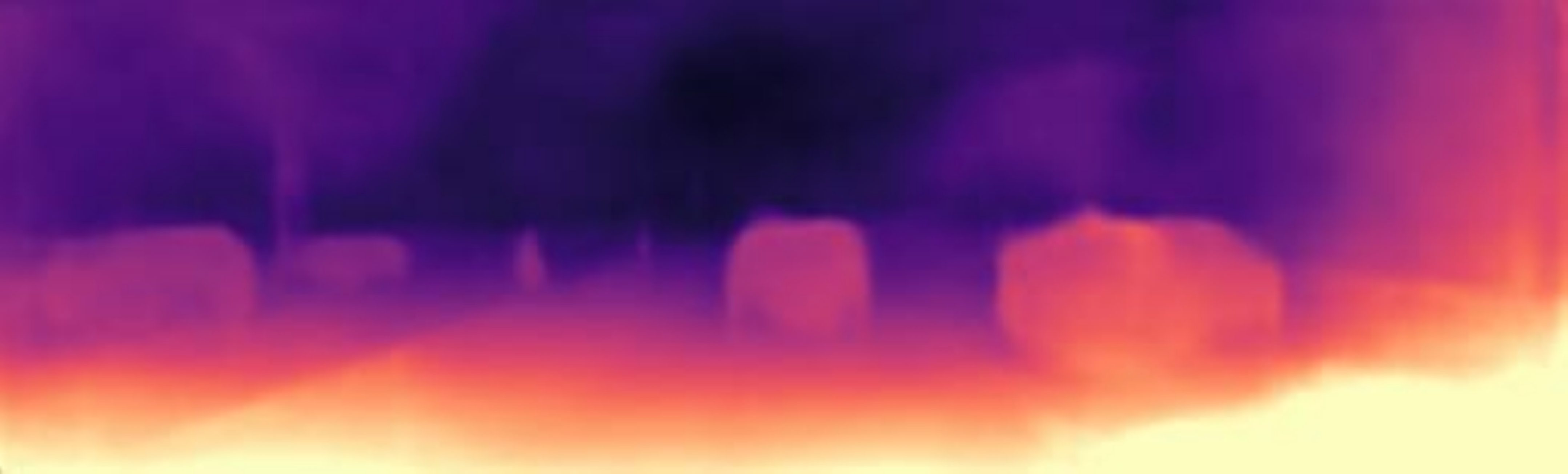}\\
\vspace{10mm}\\
\rotatebox[origin=c]{90}{\fontsize{\textw}{\texth}\selectfont MF-SLaK\hspace{-320mm}}\hspace{20mm}
\includegraphics[width=\iw,height=\ih]{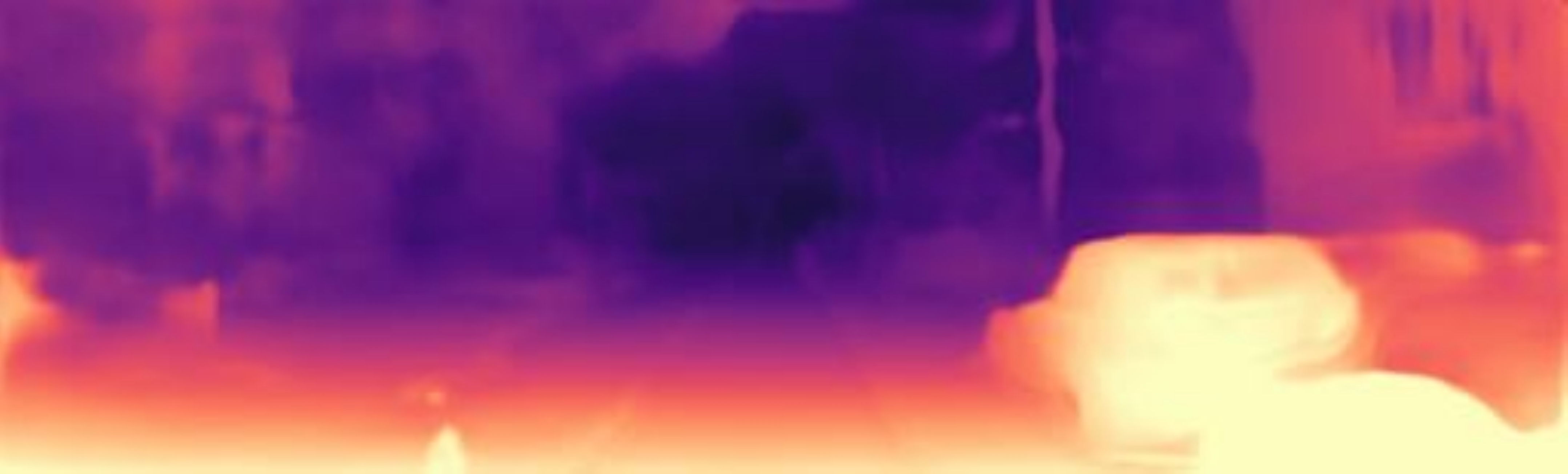}\qquad\qquad\quad &  
\includegraphics[width=\iw,height=\ih]{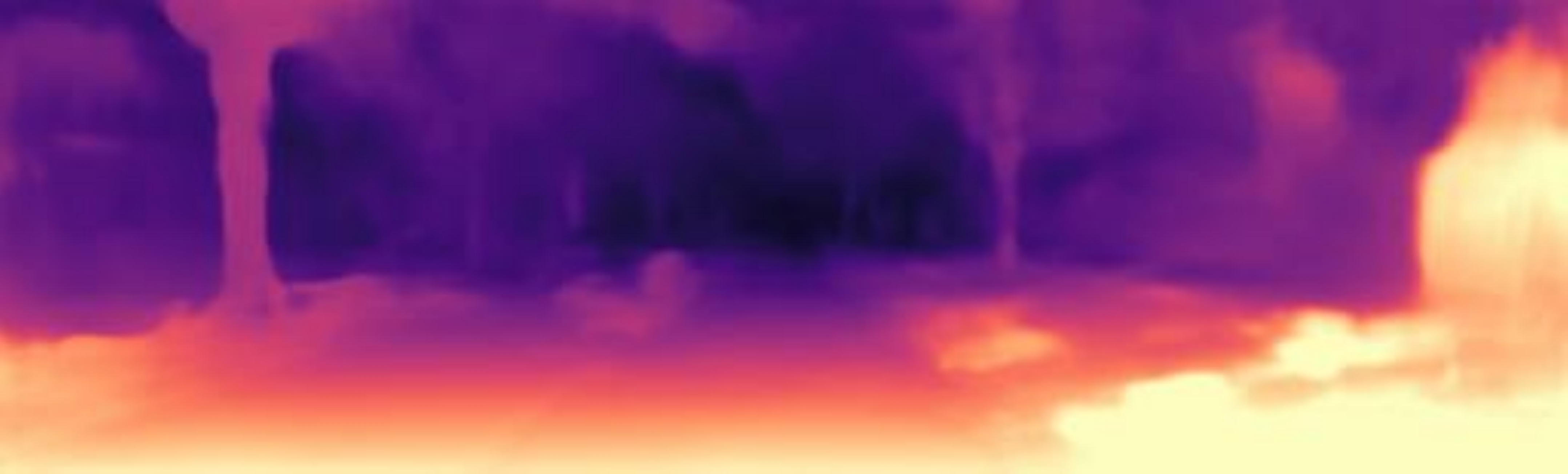}\qquad\qquad\quad & 
\includegraphics[width=\iw,height=\ih]{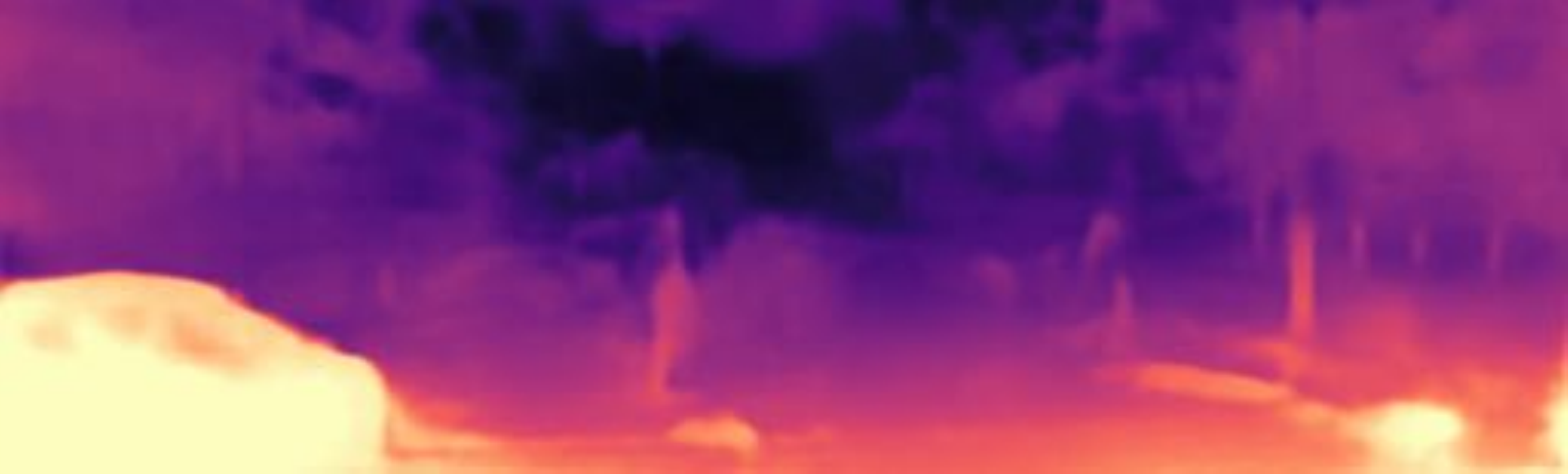}\qquad\qquad\quad & 
\includegraphics[width=\iw,height=\ih]{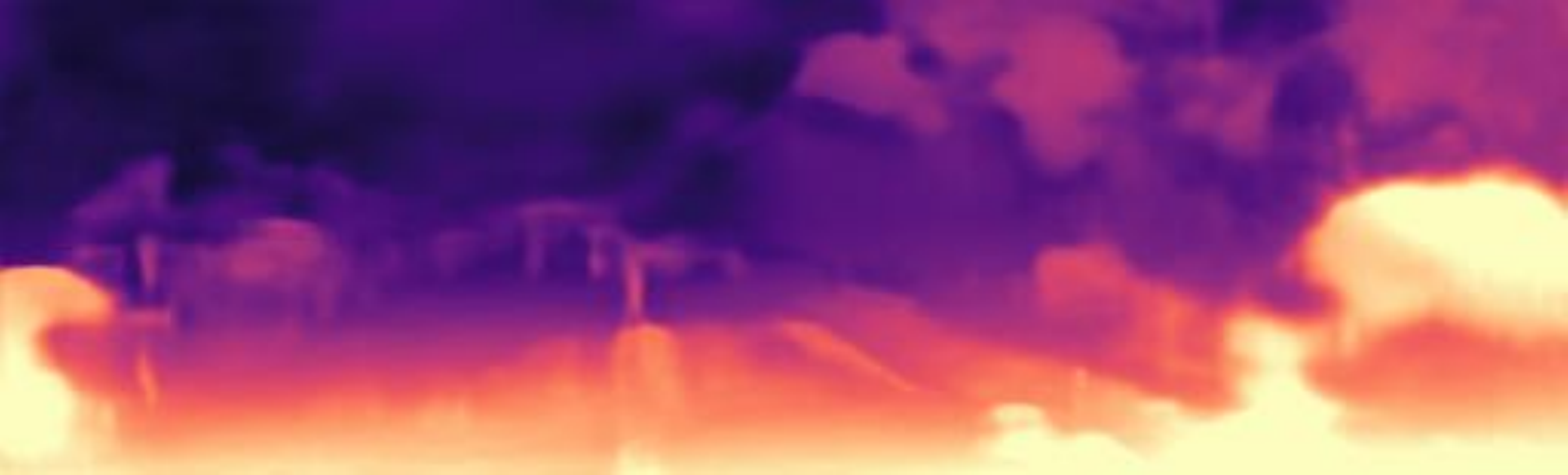}\qquad\qquad\quad & 
\includegraphics[width=\iw,height=\ih]{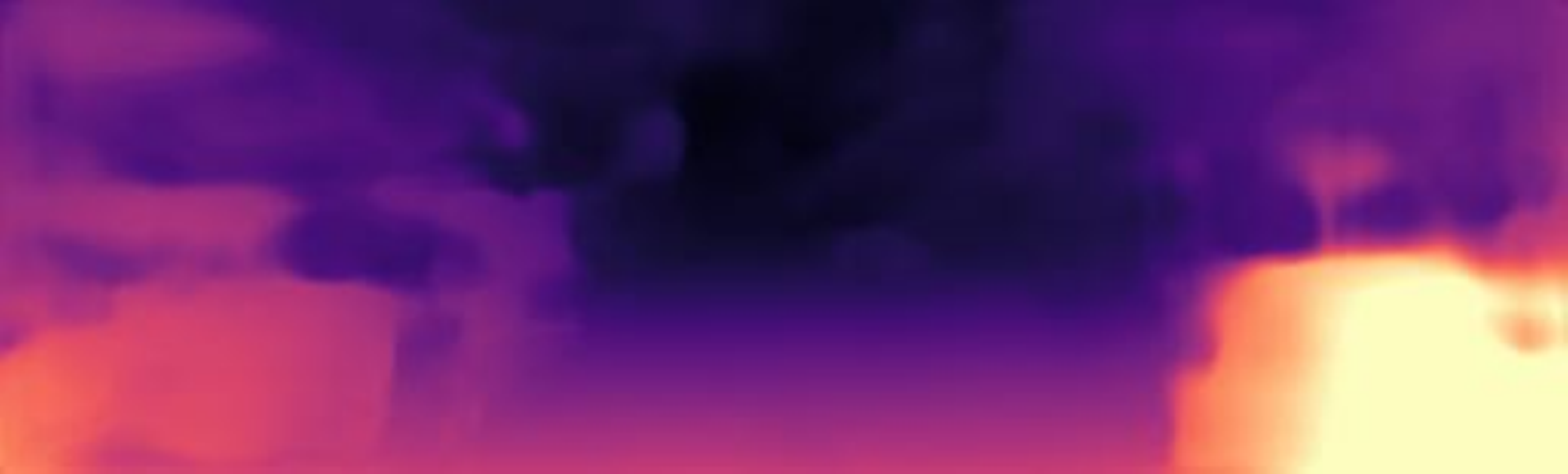}\qquad\qquad\quad & 
\includegraphics[width=\iw,height=\ih]{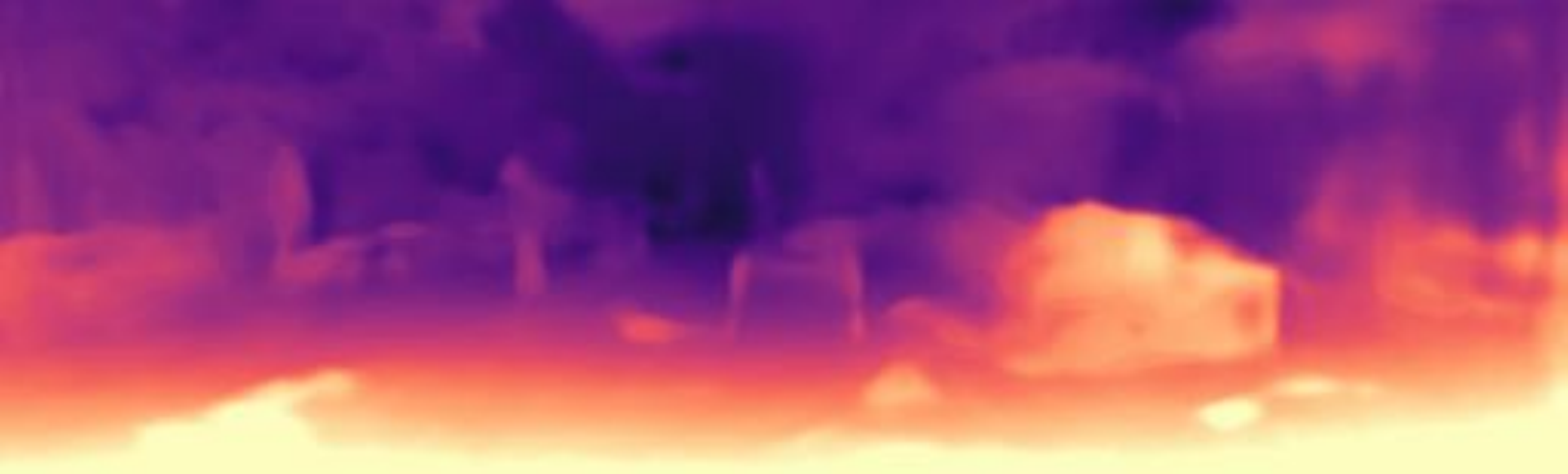}\\
\vspace{30mm}\\
\multicolumn{6}{c}{\fontsize{\w}{\h} \selectfont (a) Self-supervised CNN-based methods} \\
\vspace{30mm}\\
\rotatebox[origin=c]{90}{\fontsize{\textw}{\texth}\selectfont MF-ViT\hspace{-310mm}}\hspace{20mm}
\includegraphics[width=\iw,height=\ih]{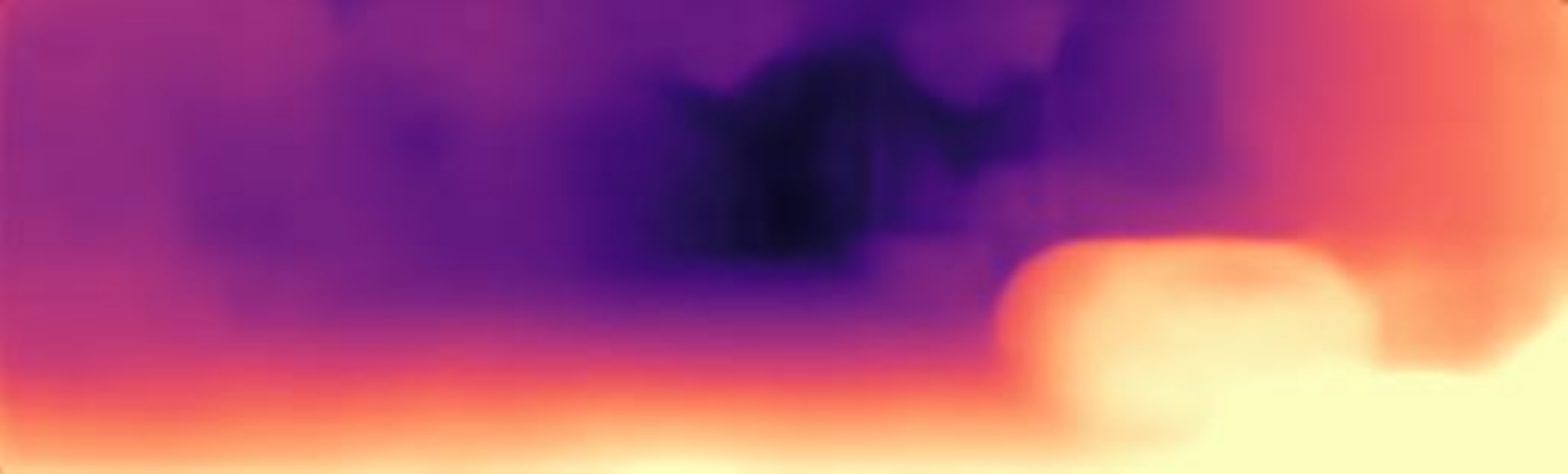}\qquad\qquad\quad & 
\includegraphics[width=\iw,height=\ih]{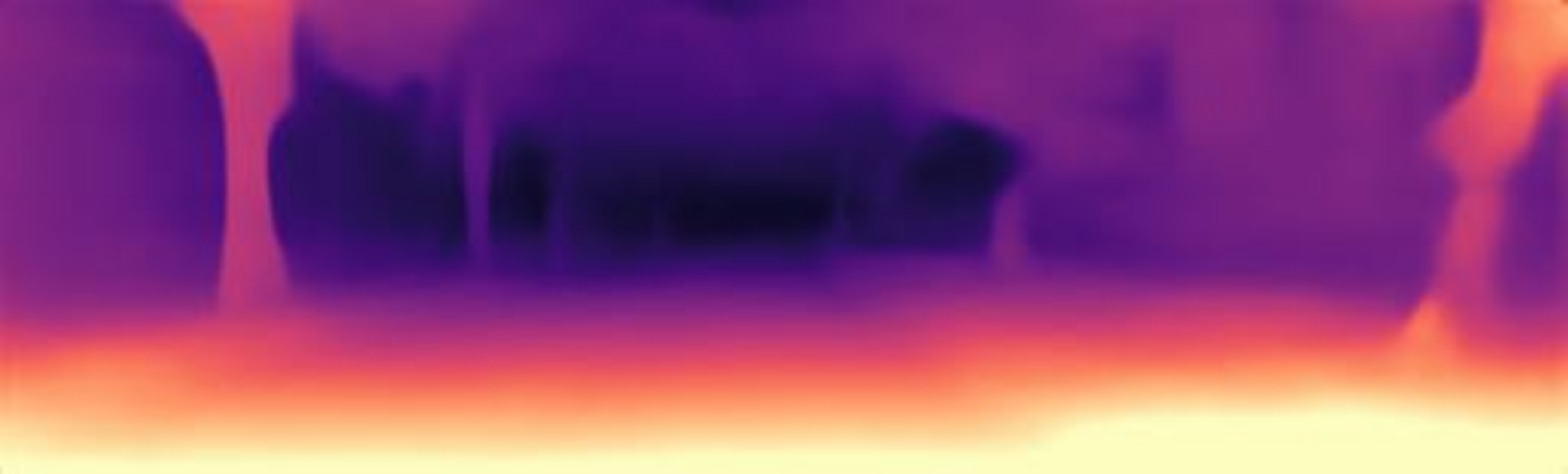}\qquad\qquad\quad & 
\includegraphics[width=\iw,height=\ih]{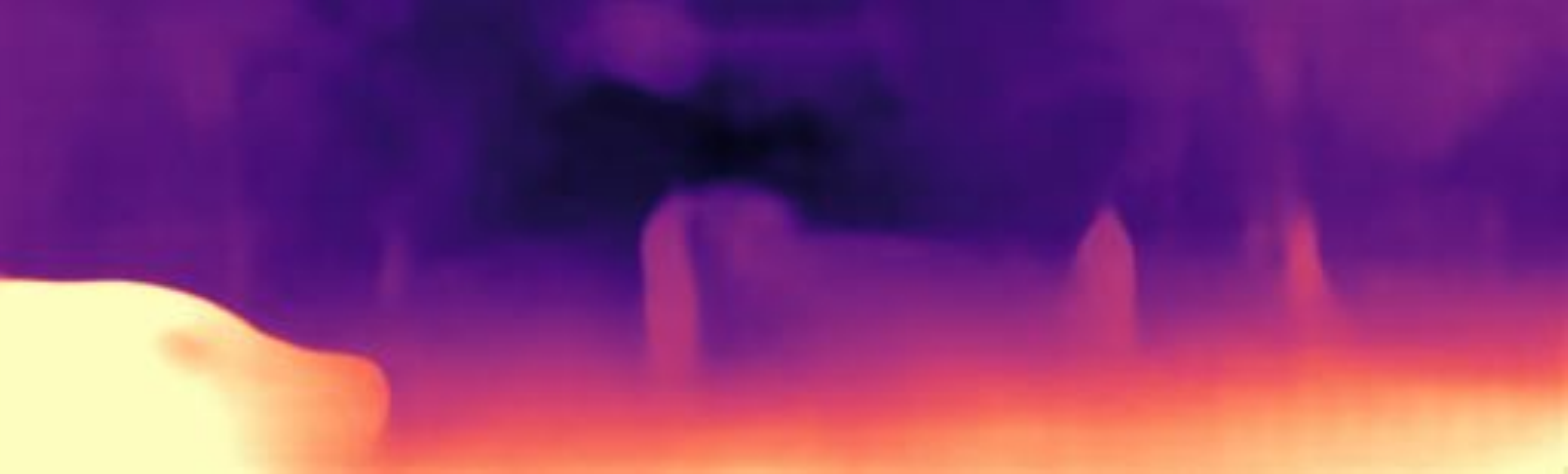}\qquad\qquad\quad & 
\includegraphics[width=\iw,height=\ih]{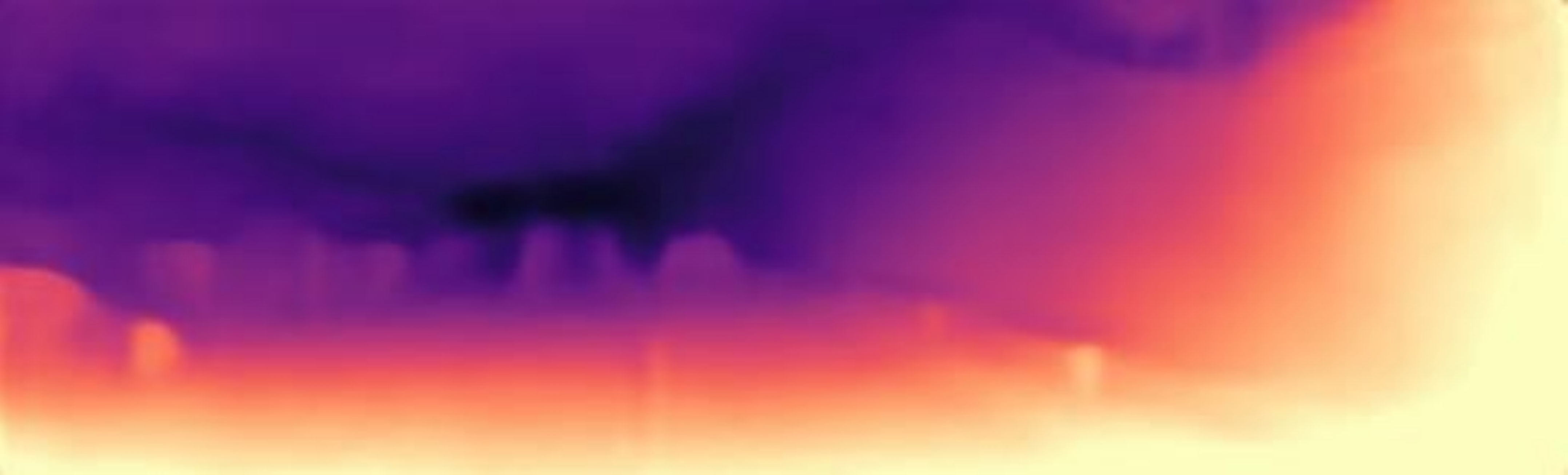}\qquad\qquad\quad & 
\includegraphics[width=\iw,height=\ih]{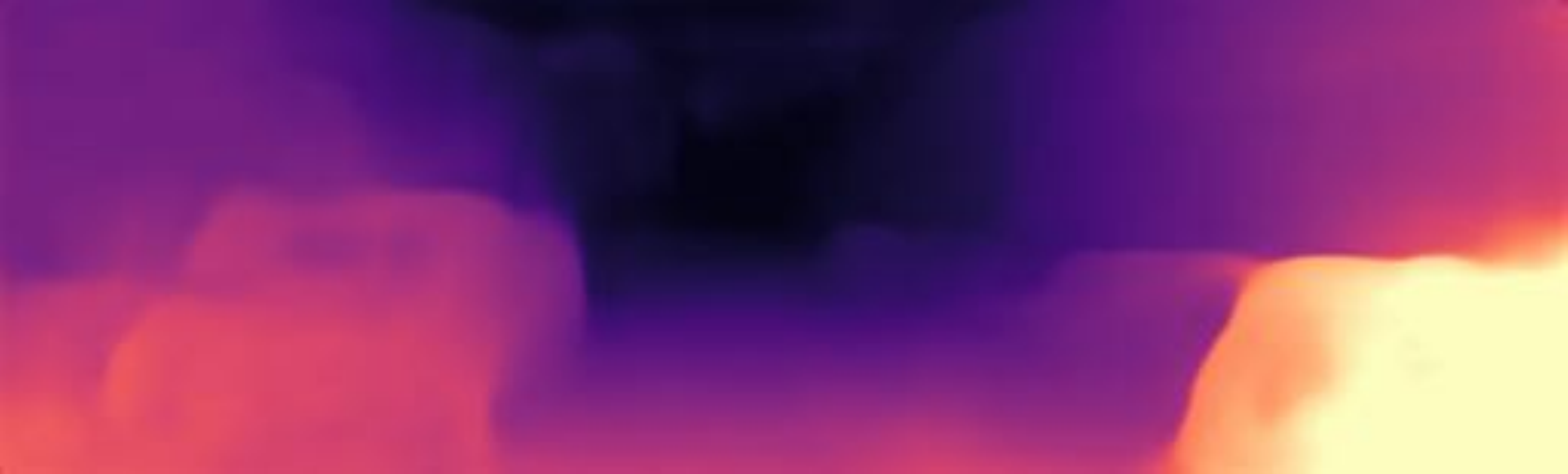}\qquad\qquad\quad & 
\includegraphics[width=\iw,height=\ih]{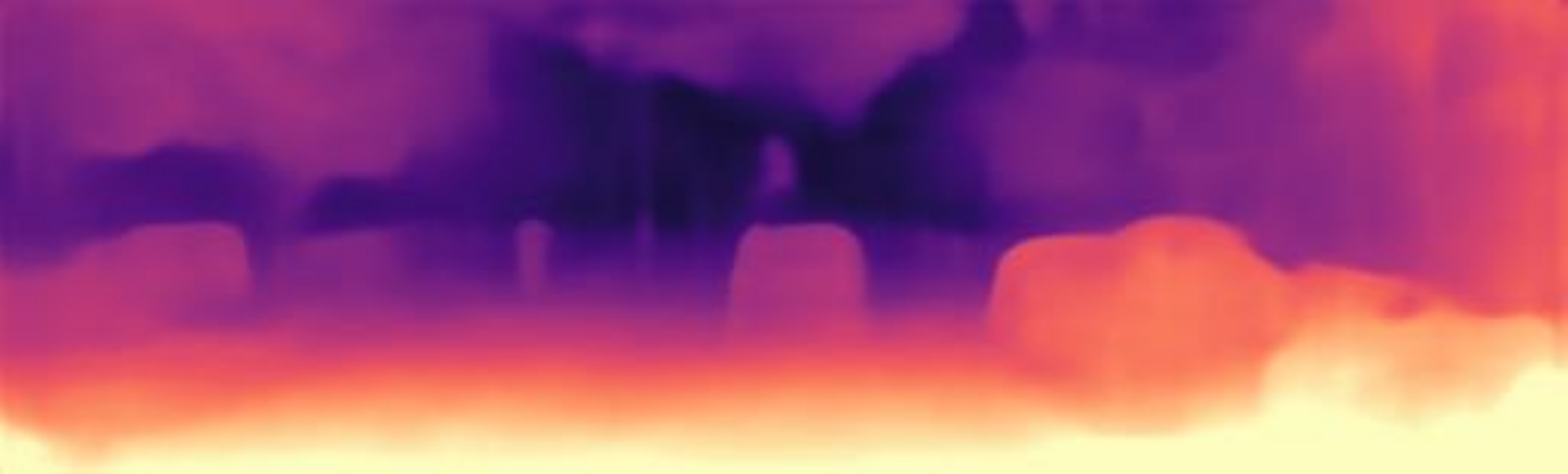}\\
\vspace{10mm}\\
\rotatebox[origin=c]{90}{\fontsize{\textw}{\texth}\selectfont MF-RegionViT\hspace{-320mm}}\hspace{10mm}
\includegraphics[width=\iw,height=\ih]{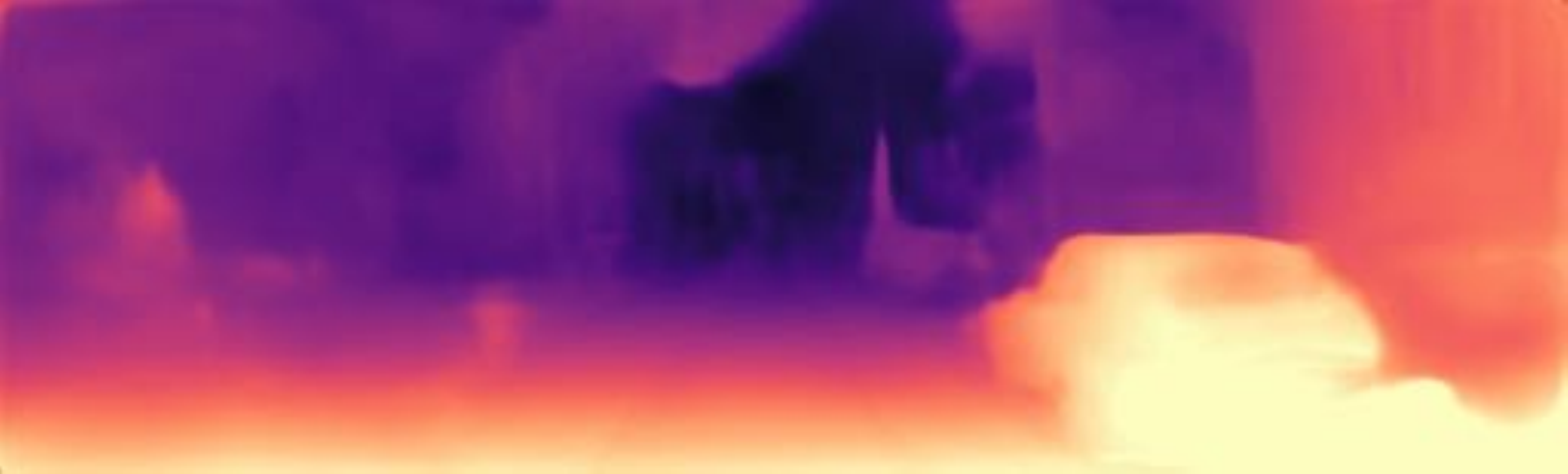}\qquad\qquad\quad & 
\includegraphics[width=\iw,height=\ih]{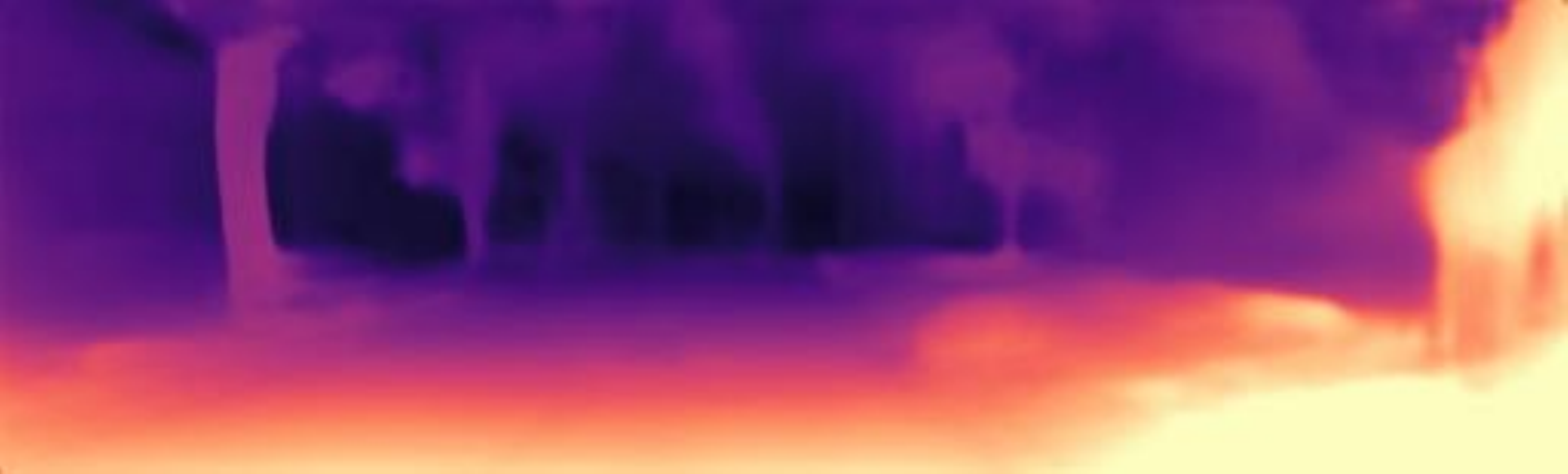}\qquad\qquad\quad &
\includegraphics[width=\iw,height=\ih]{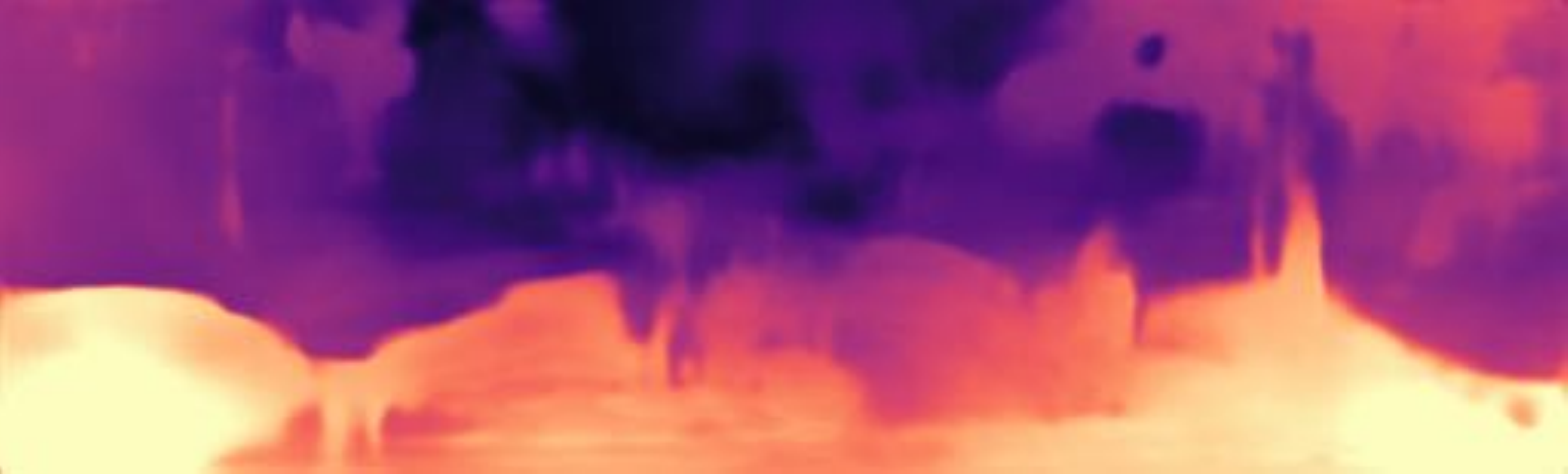}\qquad\qquad\quad &
\includegraphics[width=\iw,height=\ih]{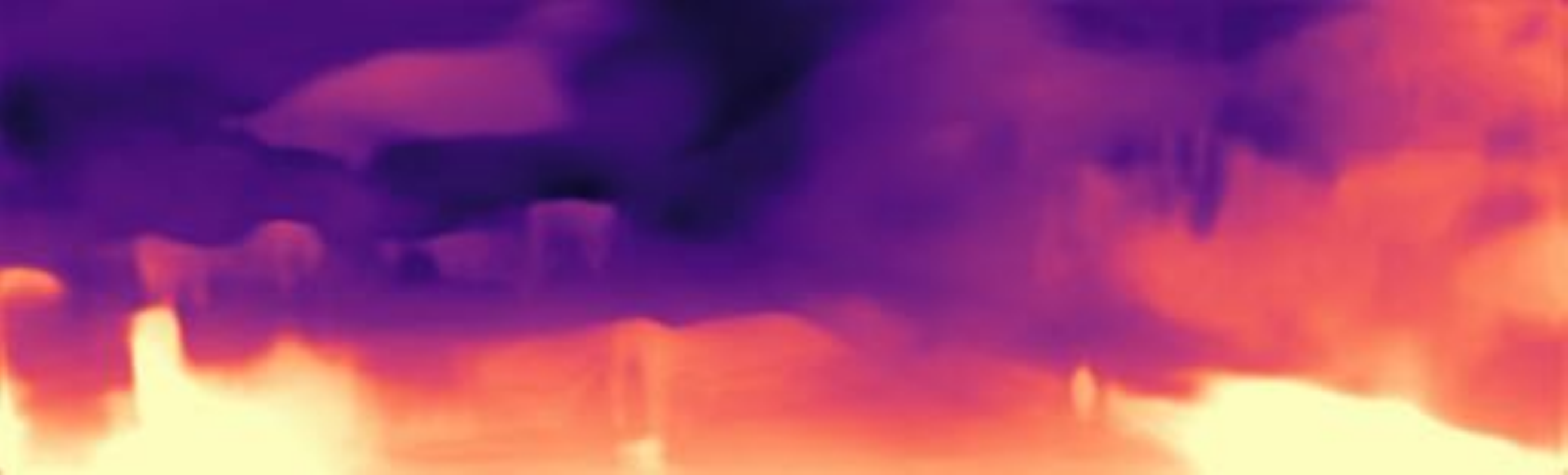}\qquad\qquad\quad &
\includegraphics[width=\iw,height=\ih]{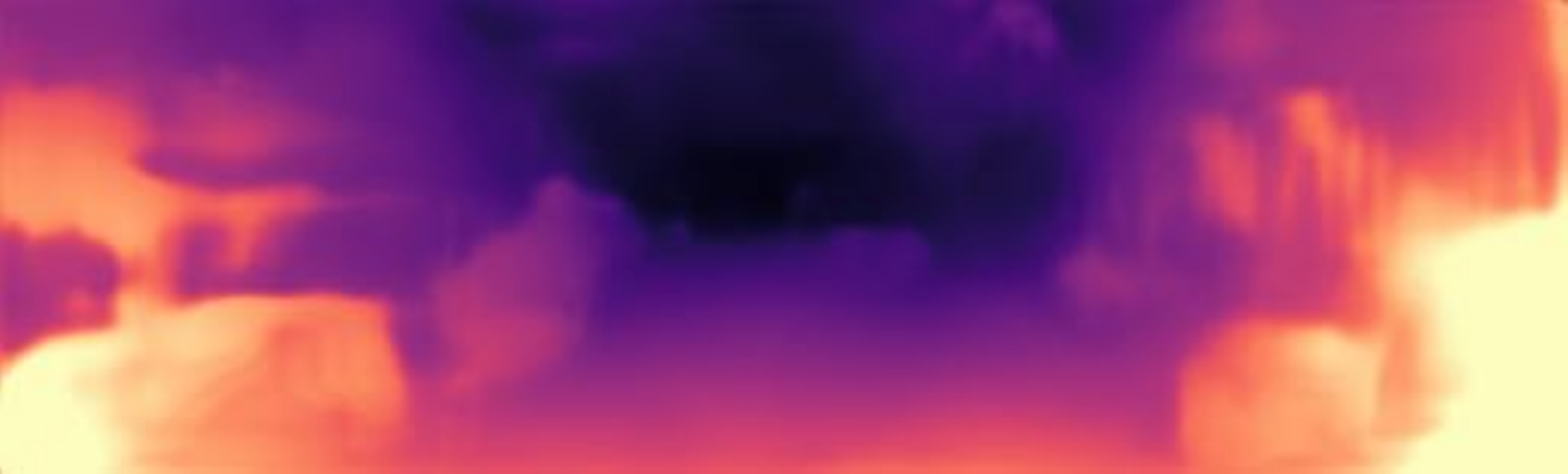}\qquad\qquad\quad &
\includegraphics[width=\iw,height=\ih]{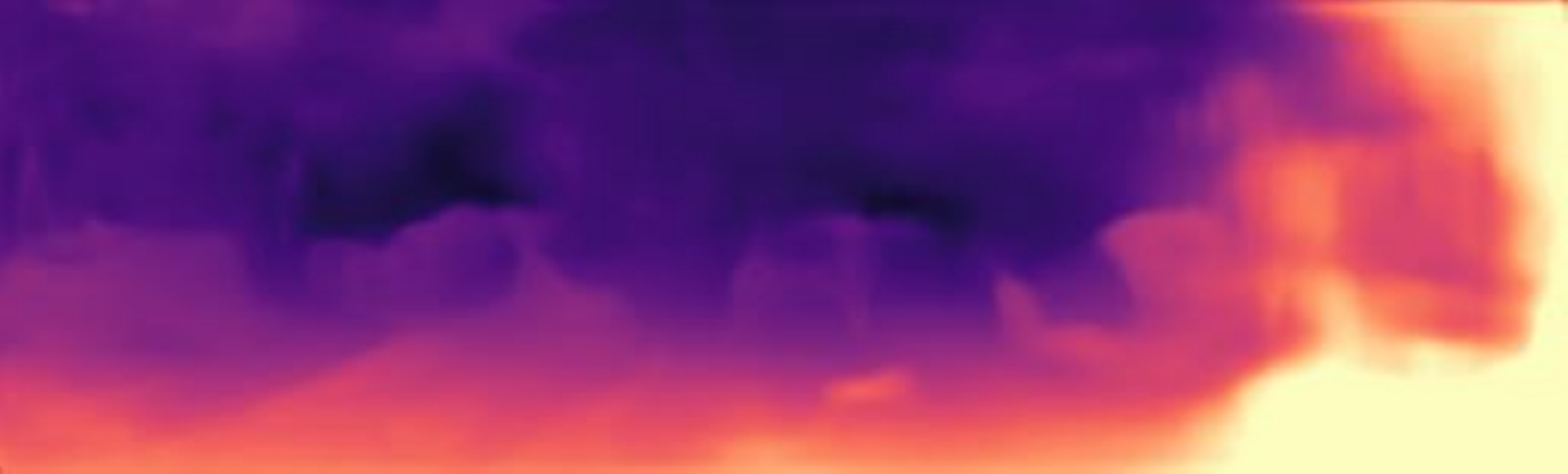}\\
\vspace{10mm}\\
\rotatebox[origin=c]{90}{\fontsize{\textw}{\texth}\selectfont MF-Twins\hspace{-310mm}}\hspace{20mm}
\includegraphics[width=\iw,height=\ih]{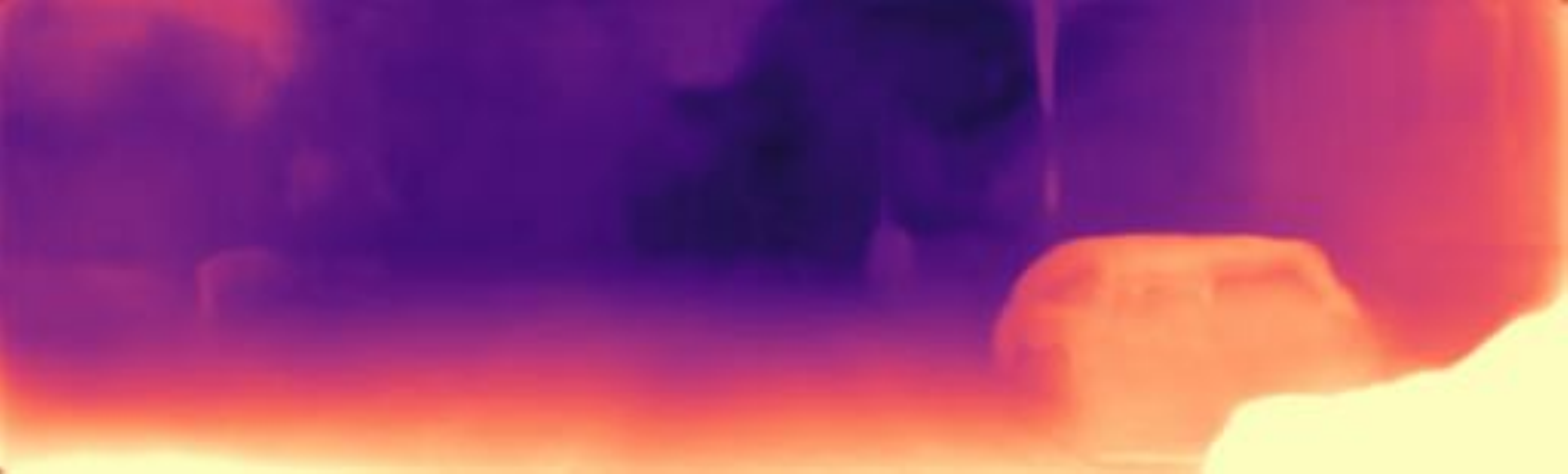}\qquad\qquad\quad &  
\includegraphics[width=\iw,height=\ih]{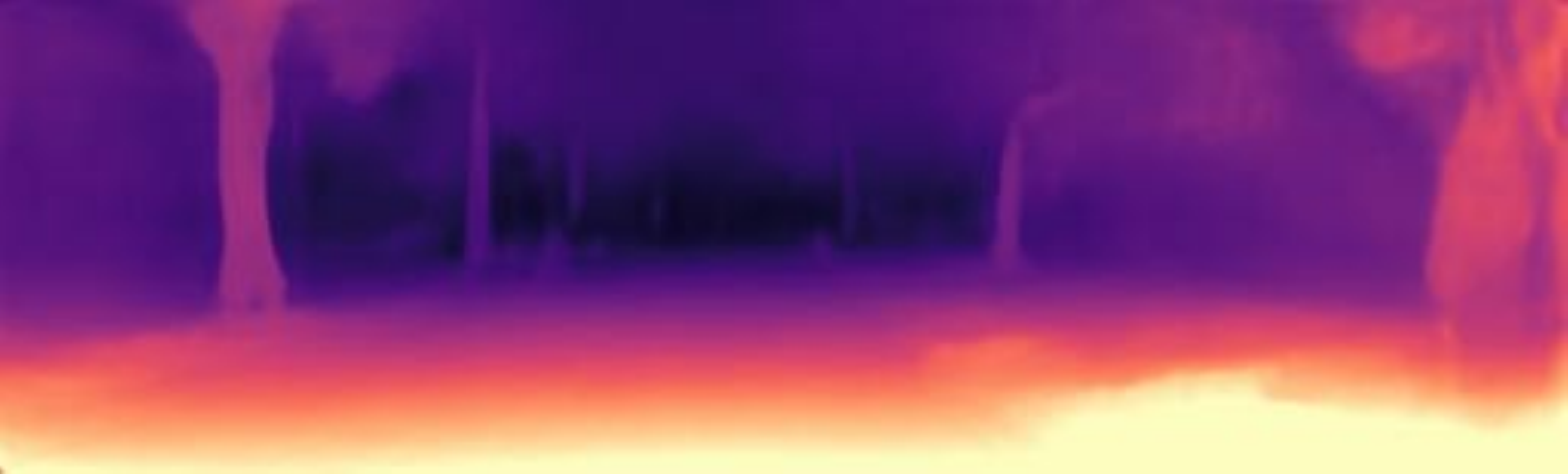}\qquad\qquad\quad & 
\includegraphics[width=\iw,height=\ih]{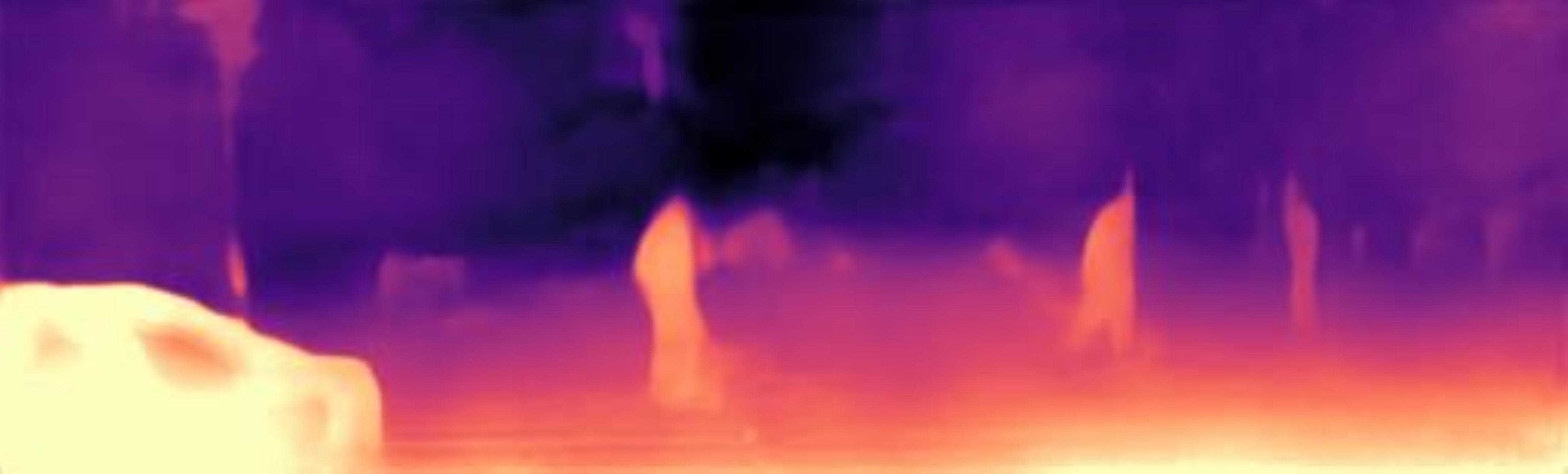}\qquad\qquad\quad & 
\includegraphics[width=\iw,height=\ih]{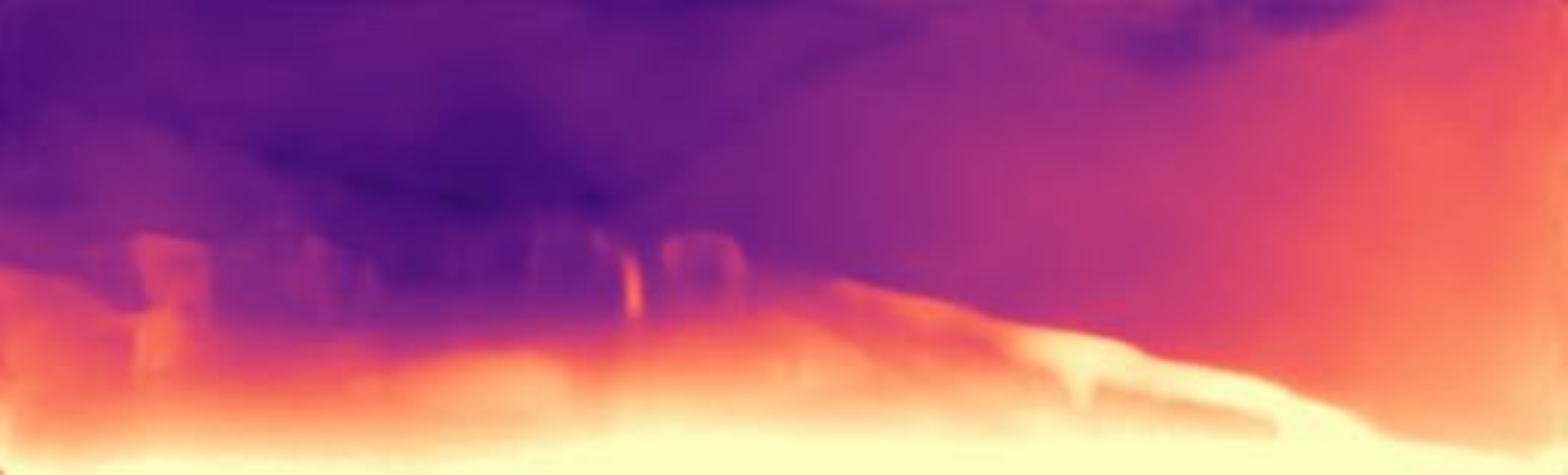}\qquad\qquad\quad & 
\includegraphics[width=\iw,height=\ih]{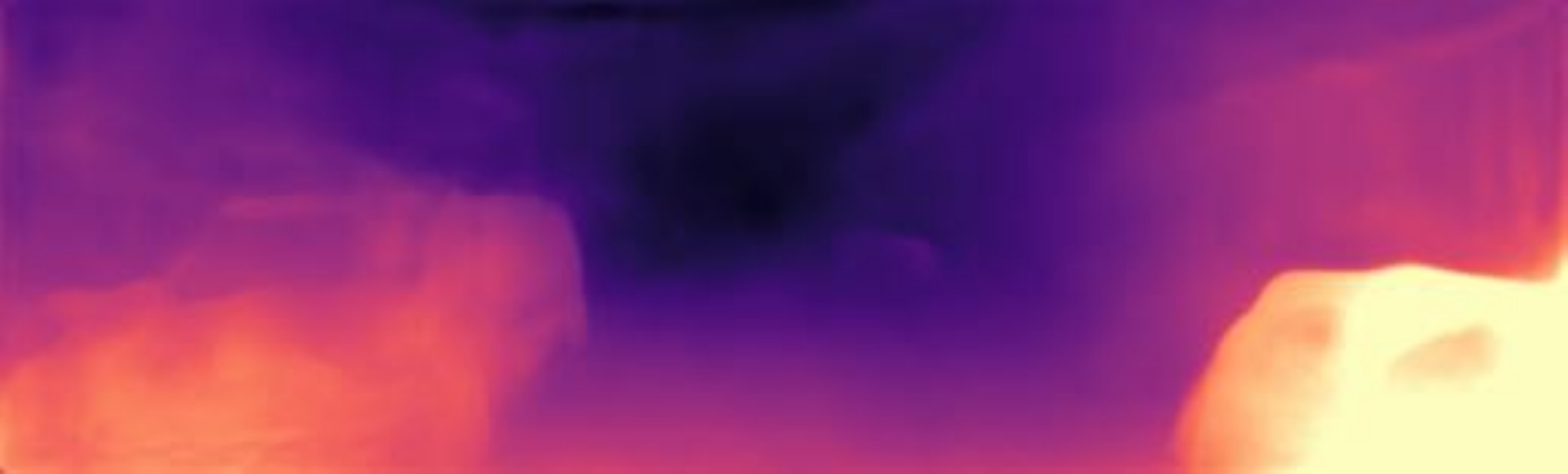}\qquad\qquad\quad &
\includegraphics[width=\iw,height=\ih]{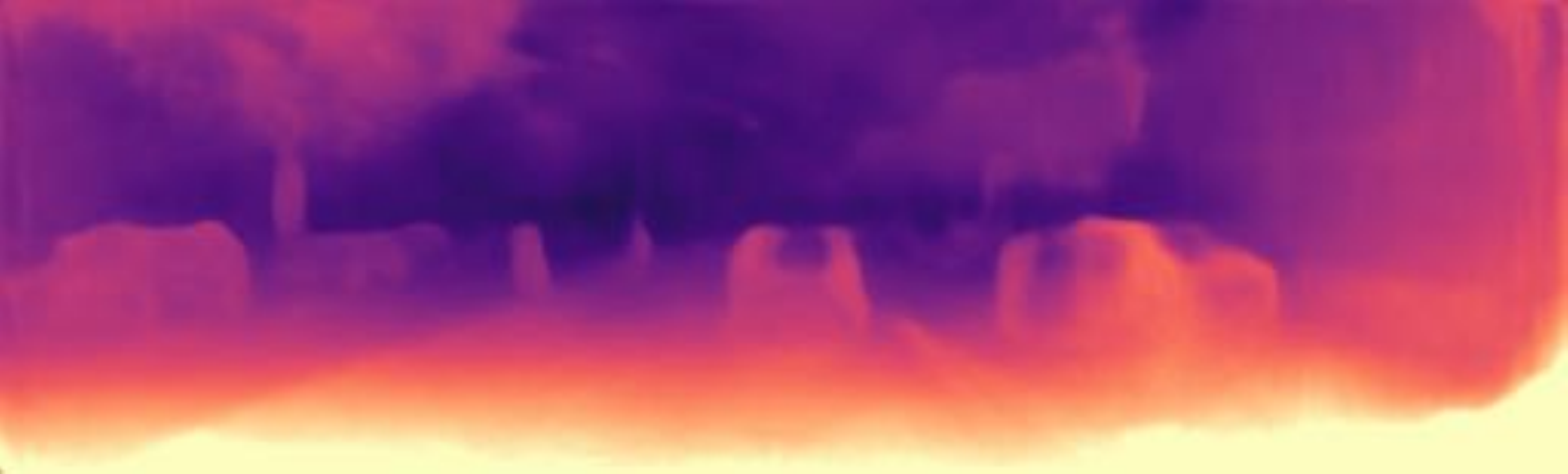}\\
\vspace{10mm}\\
\rotatebox[origin=c]{90}{\fontsize{\textw}{\texth}\selectfont MF-Ours\hspace{-300mm}}\hspace{20mm}
\includegraphics[width=\iw,height=\ih]{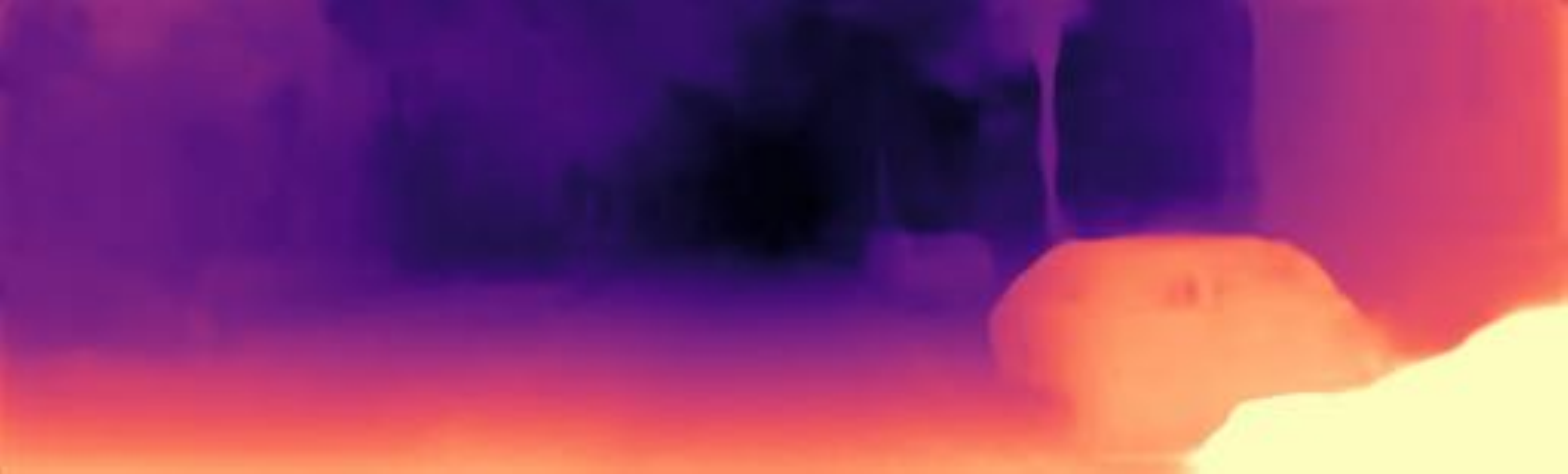}\qquad\qquad\quad &
\includegraphics[width=\iw,height=\ih]{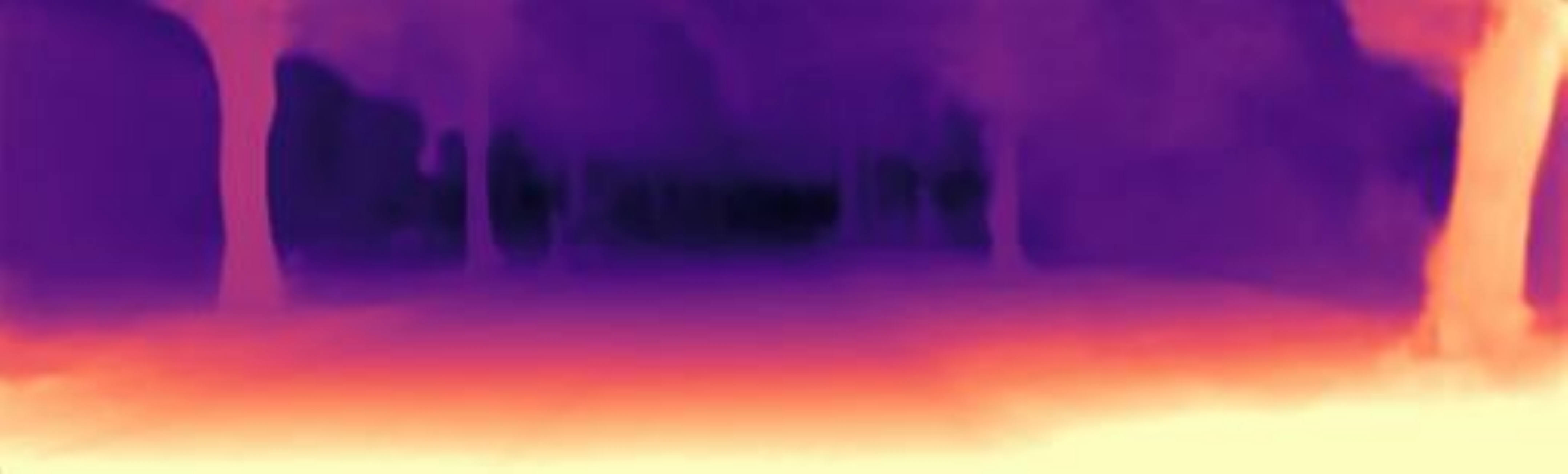}\qquad\qquad\quad &
\includegraphics[width=\iw,height=\ih]{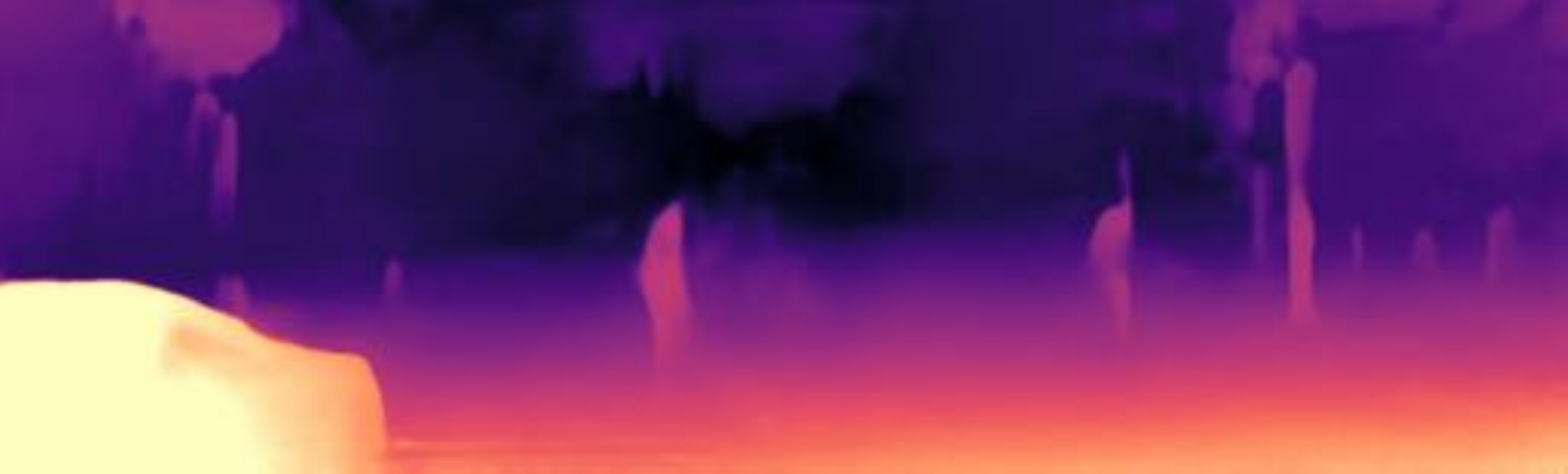}\qquad\qquad\quad &
\includegraphics[width=\iw,height=\ih]{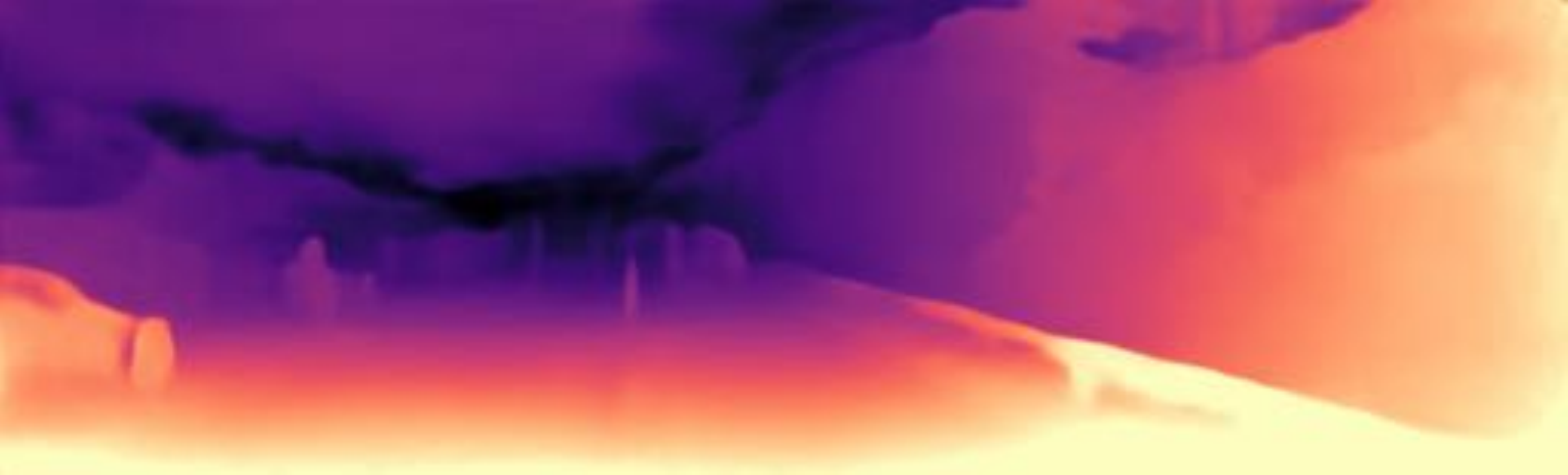}\qquad\qquad\quad &
\includegraphics[width=\iw,height=\ih]{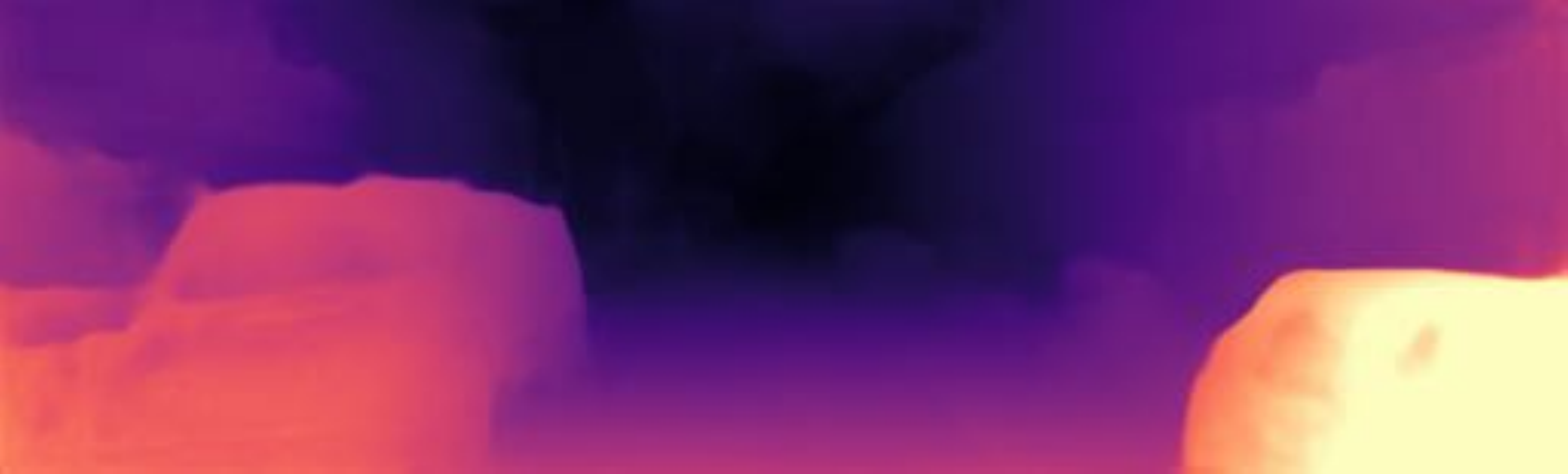}\qquad\qquad\quad &
\includegraphics[width=\iw,height=\ih]{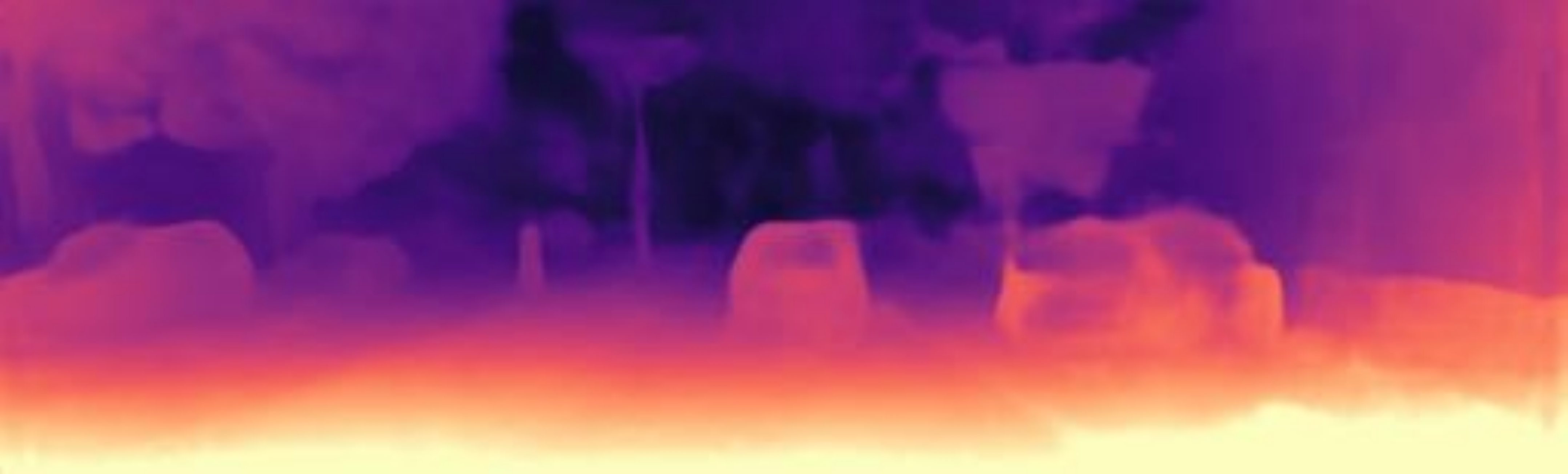}\\
\vspace{30mm}\\
\multicolumn{6}{c}{\fontsize{\w}{\h} \selectfont (b) Self-supervised Transformer-based methods} \\
\vspace{30mm}\\
\rotatebox[origin=c]{90}{\fontsize{\textw}{\texth}\selectfont BTS\hspace{-320mm}}\hspace{10mm} 
\includegraphics[width=\iw,height=\ih]{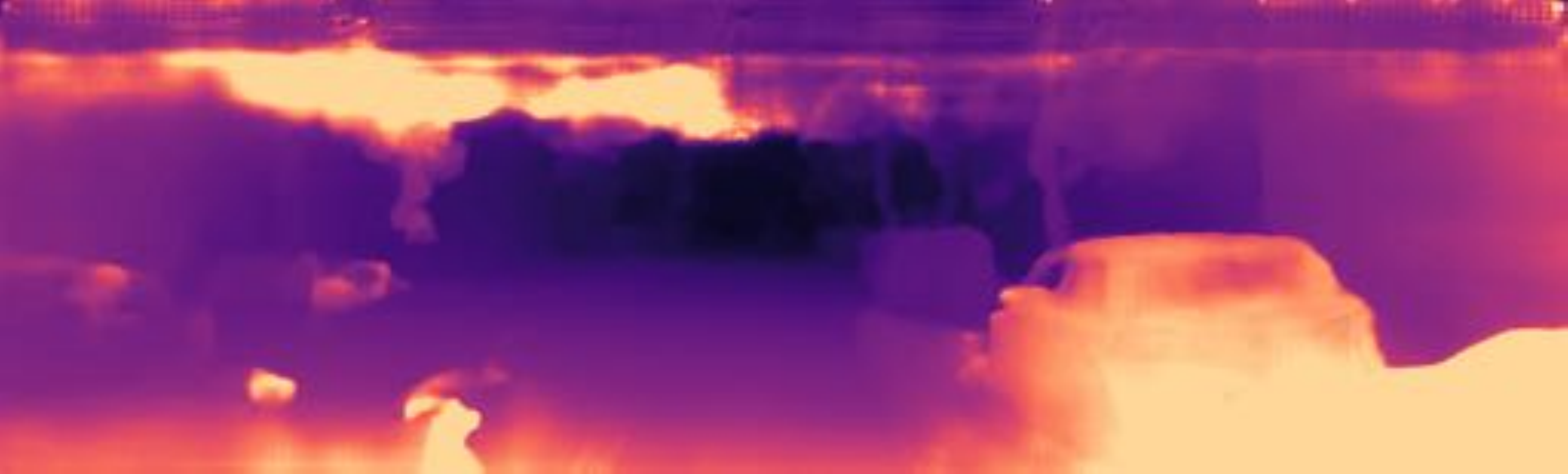}\qquad\qquad\quad &  
\includegraphics[width=\iw,height=\ih]{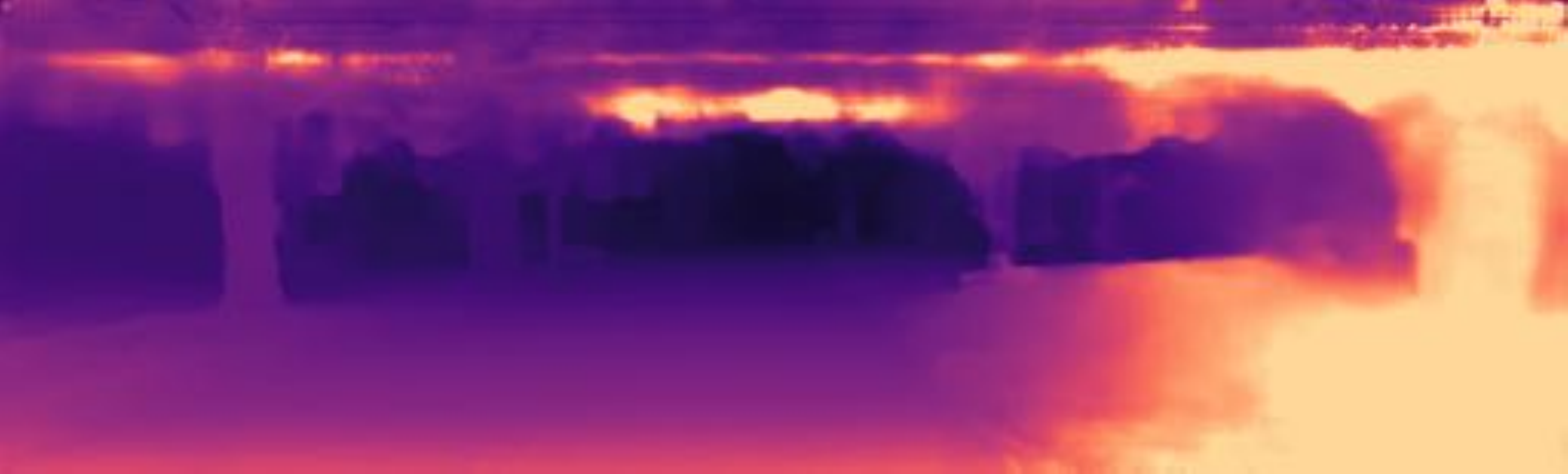}\qquad\qquad\quad & 
\includegraphics[width=\iw,height=\ih]{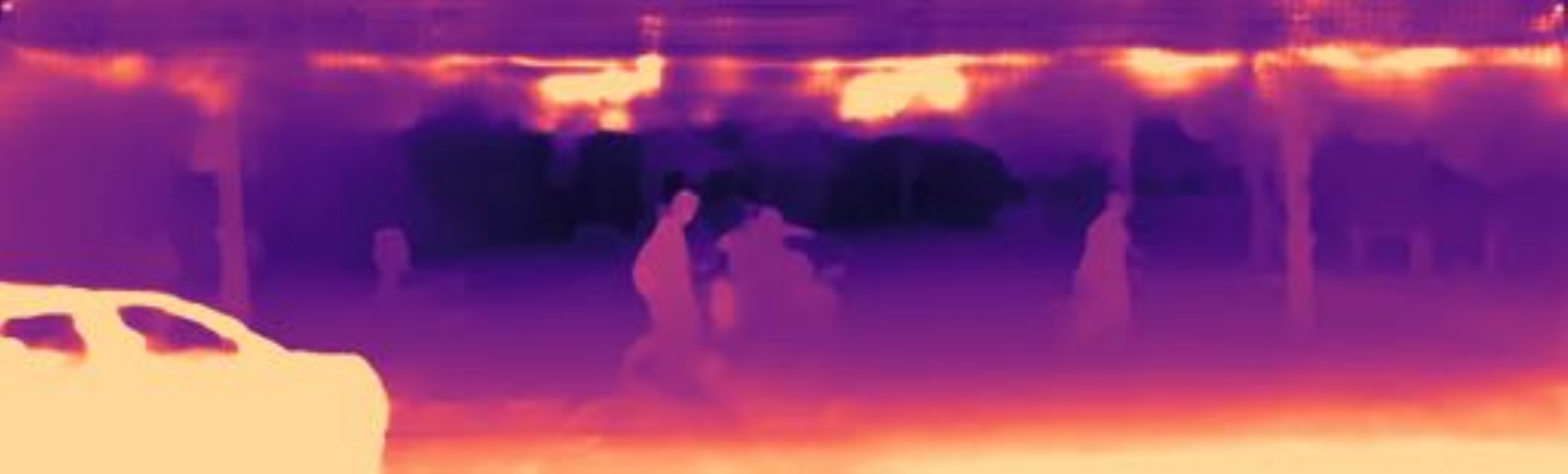}\qquad\qquad\quad & 
\includegraphics[width=\iw,height=\ih]{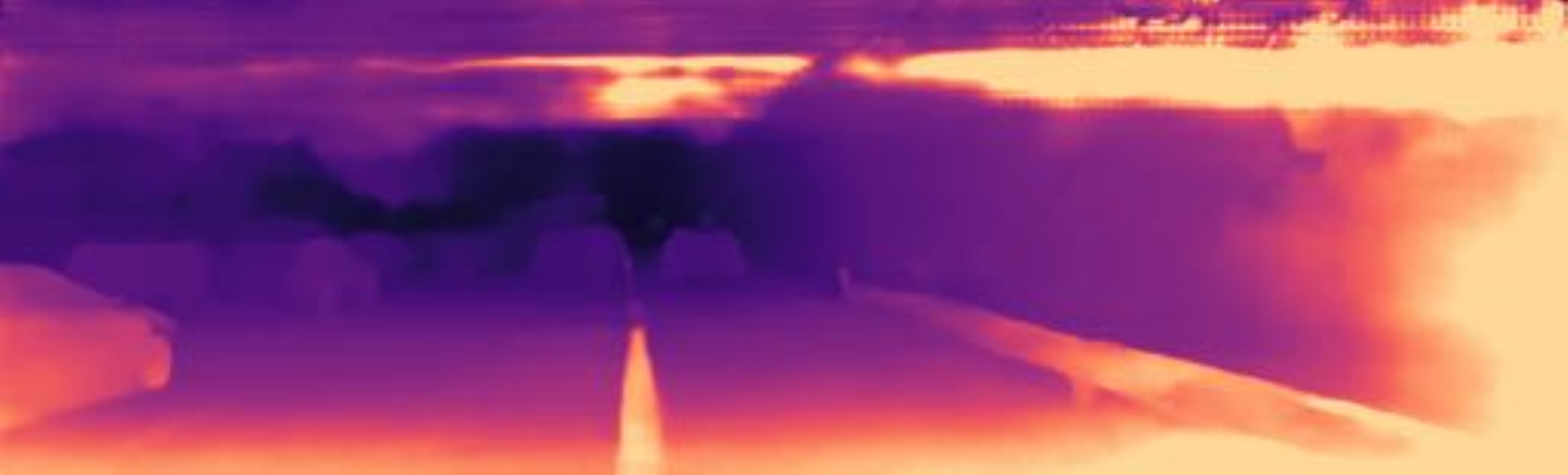}\qquad\qquad\quad & 
\includegraphics[width=\iw,height=\ih]{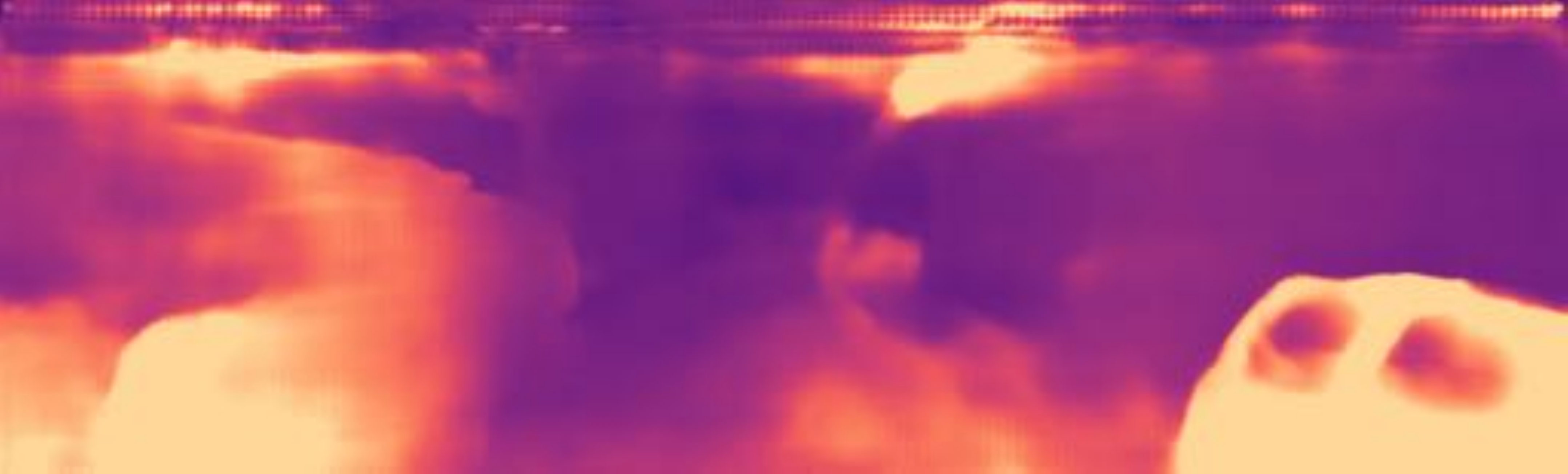}\qquad\qquad\quad & 
\includegraphics[width=\iw,height=\ih]{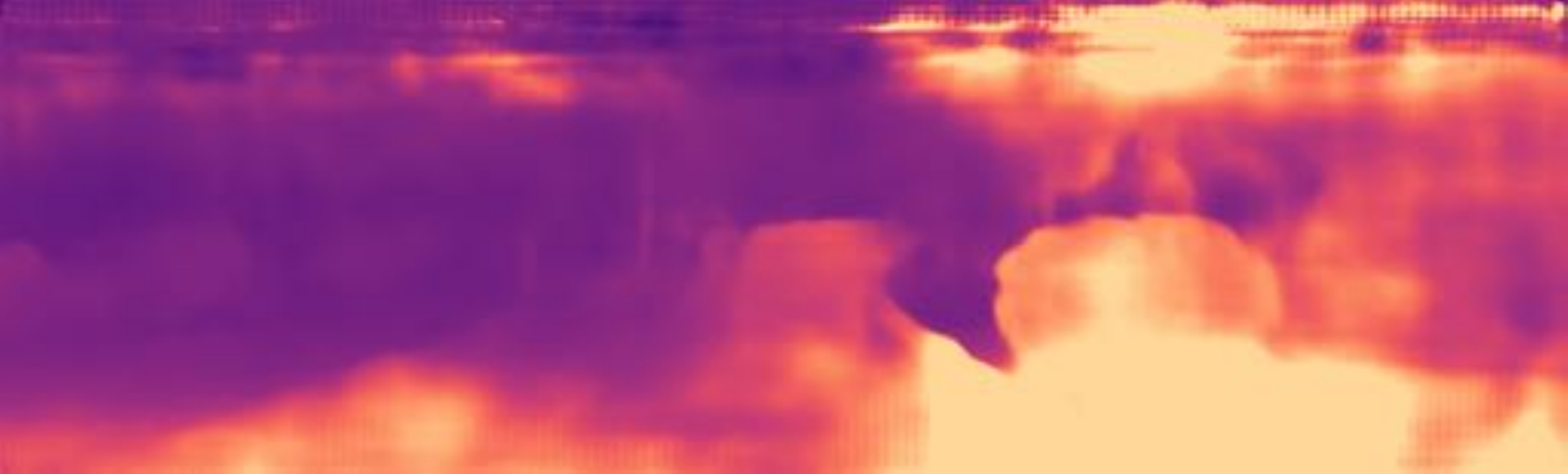}\\
\vspace{10mm}\\
\rotatebox[origin=c]{90}{\fontsize{\textw}{\texth}\selectfont Adabins\hspace{-300mm}}\hspace{10mm} 
\includegraphics[width=\iw,height=\ih]{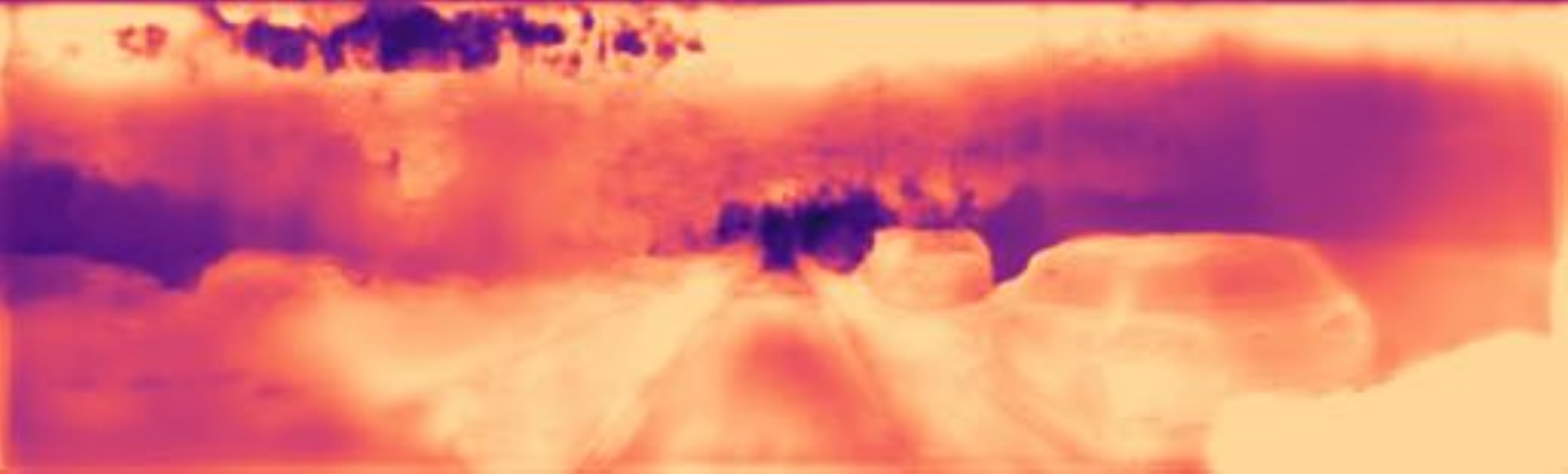}\qquad\qquad\quad & 
\includegraphics[width=\iw,height=\ih]{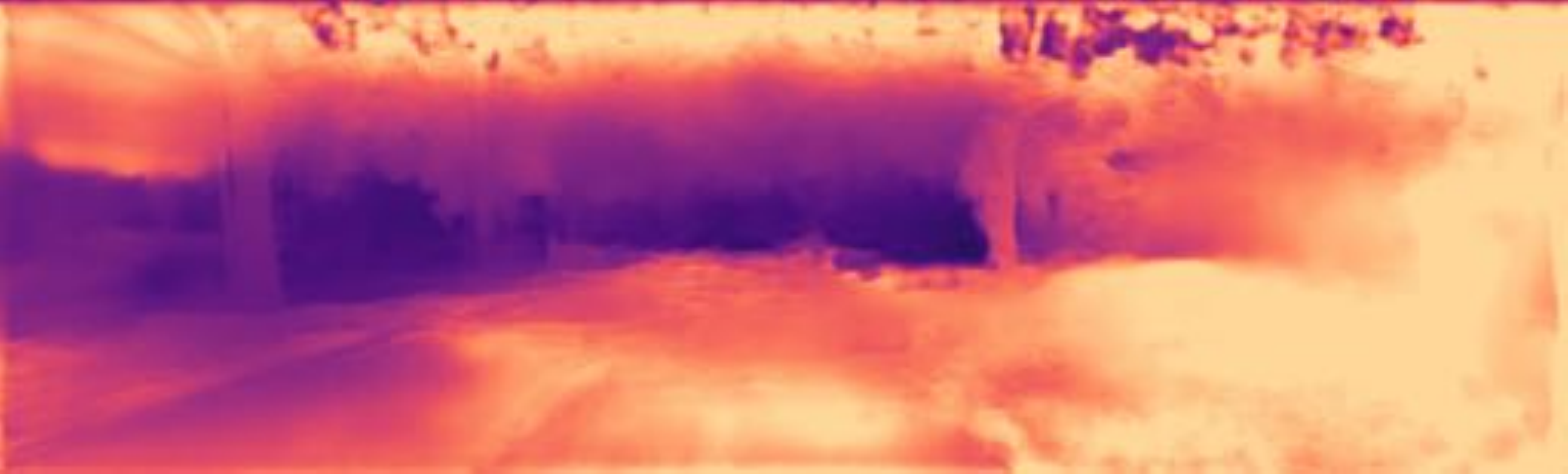}\qquad\qquad\quad & 
\includegraphics[width=\iw,height=\ih]{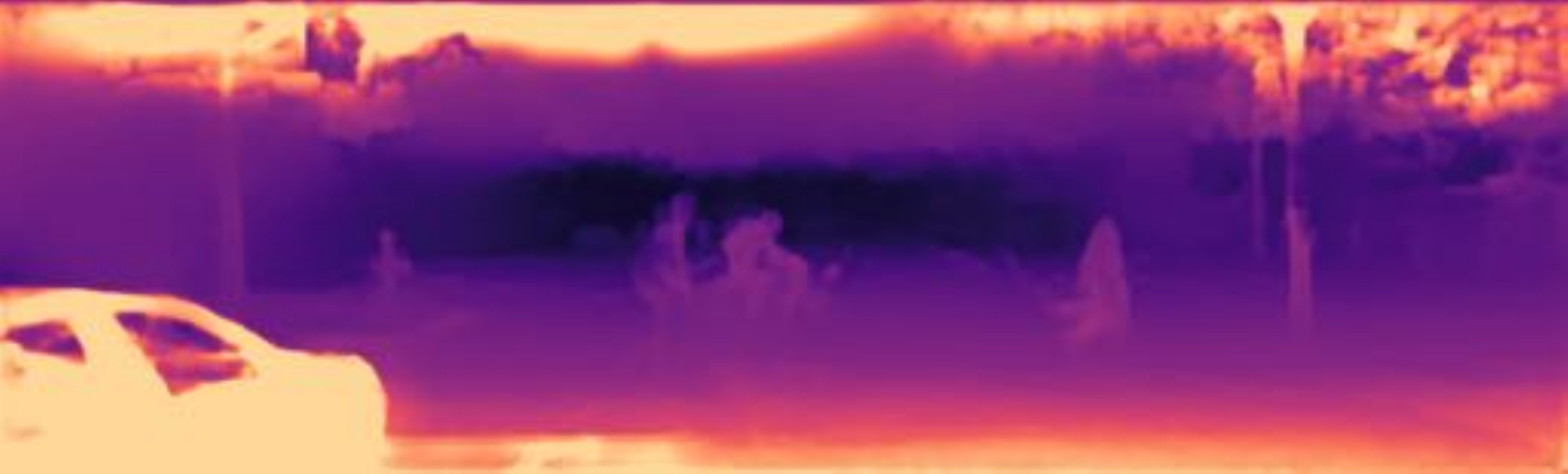}\qquad\qquad\quad & 
\includegraphics[width=\iw,height=\ih]{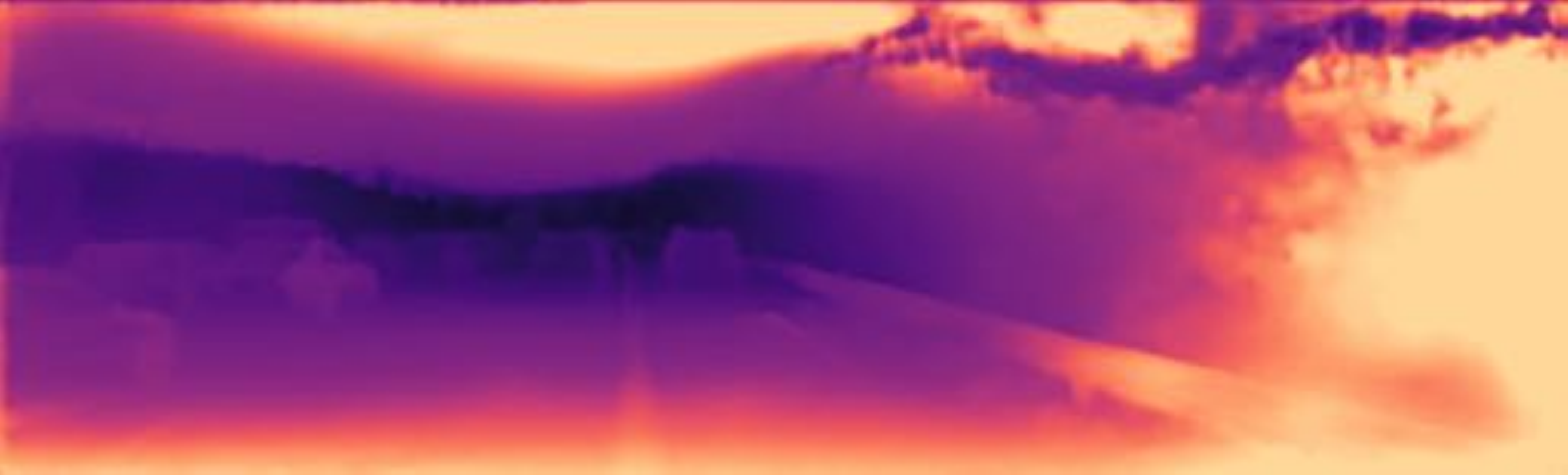}\qquad\qquad\quad & 
\includegraphics[width=\iw,height=\ih]{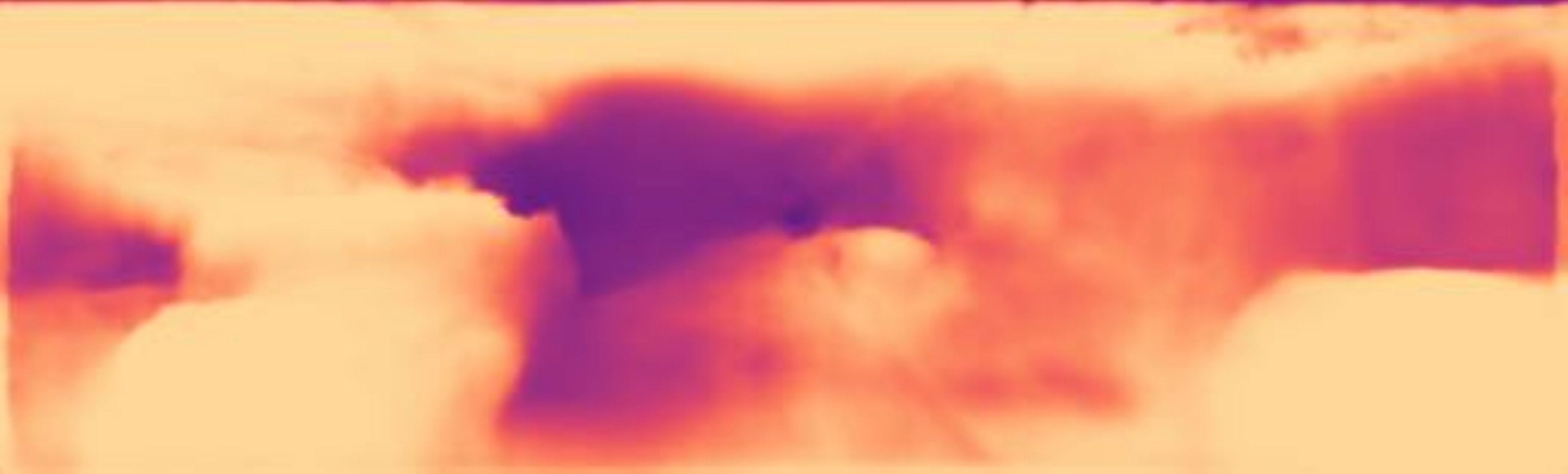}\qquad\qquad\quad & 
\includegraphics[width=\iw,height=\ih]{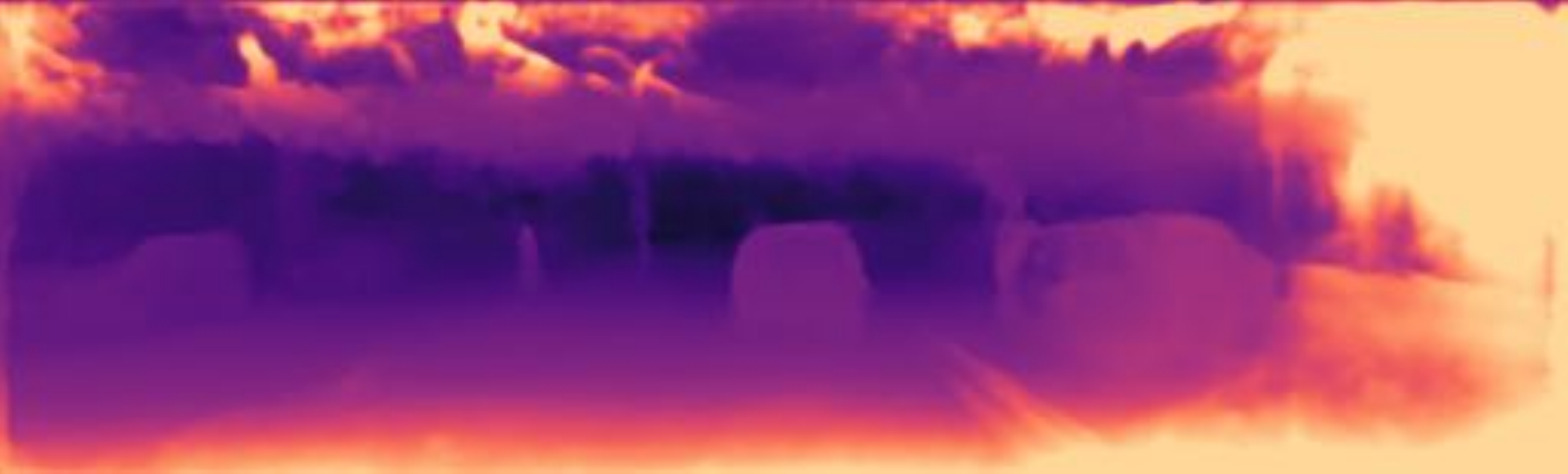}\\
\vspace{30mm}\\
\multicolumn{6}{c}{\fontsize{\w}{\h} \selectfont (c) Supervised CNN-based methods} \\
\vspace{30mm}\\
\rotatebox[origin=c]{90}{\fontsize{\textw}{\texth}\selectfont TransDepth\hspace{-310mm}}\hspace{10mm}
\includegraphics[width=\iw,height=\ih]{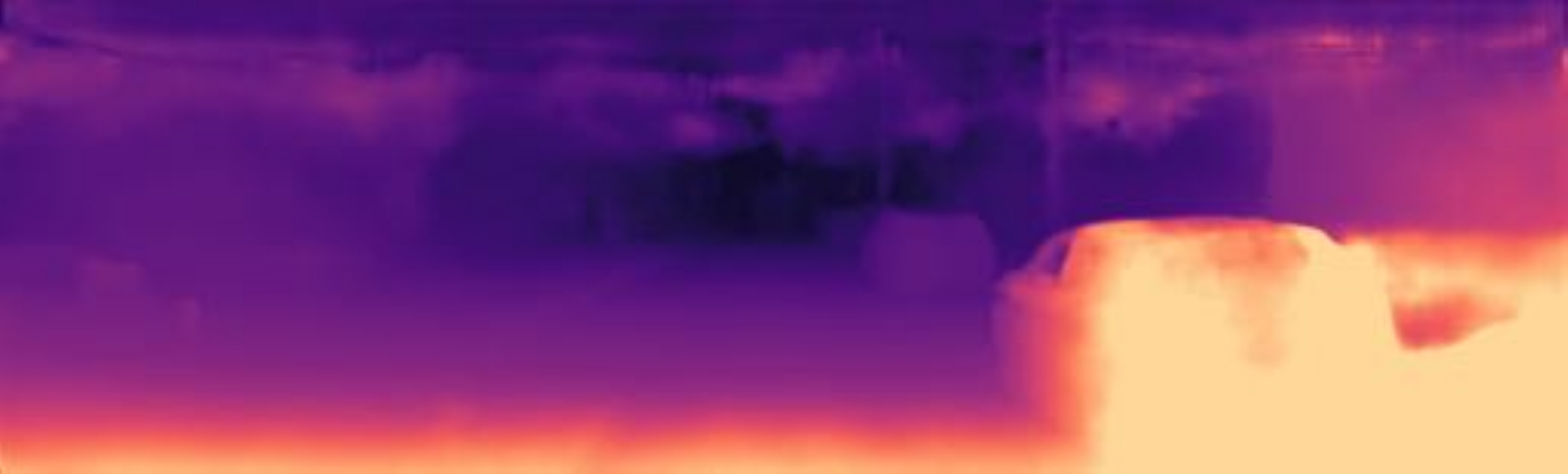}\qquad\qquad\quad &
\includegraphics[width=\iw,height=\ih]{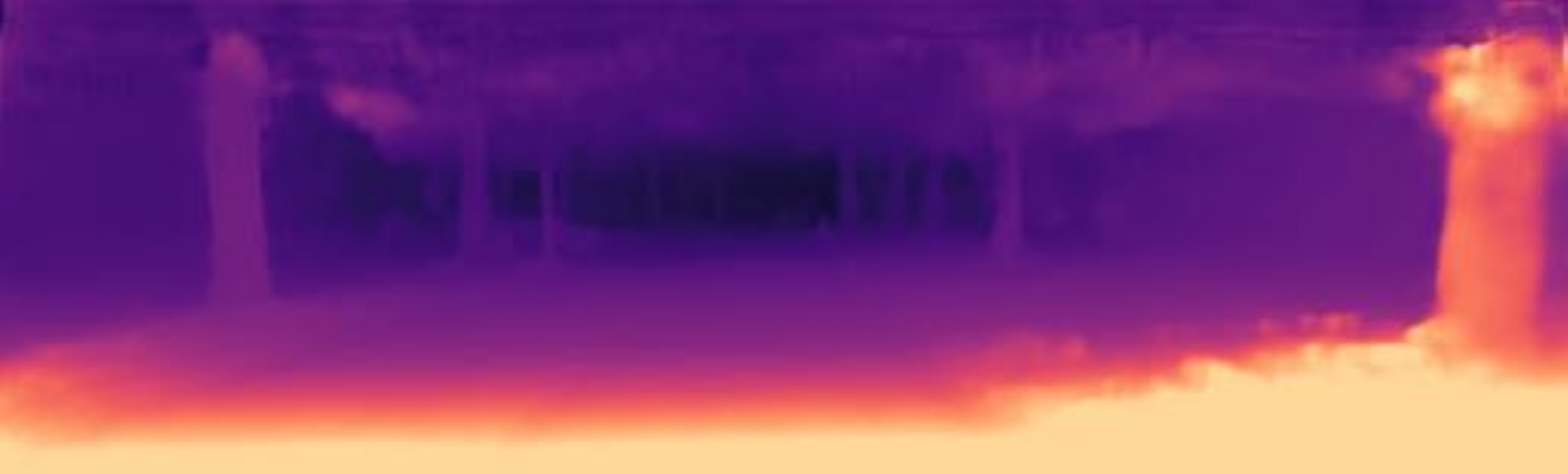}\qquad\qquad\quad &
\includegraphics[width=\iw,height=\ih]{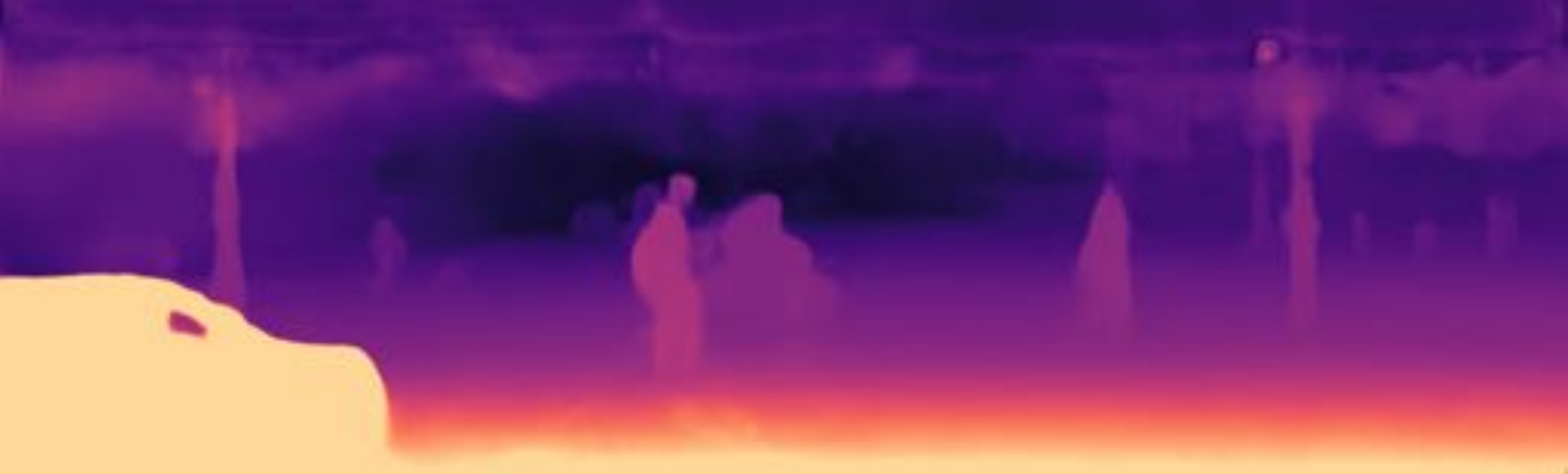}\qquad\qquad\quad &
\includegraphics[width=\iw,height=\ih]{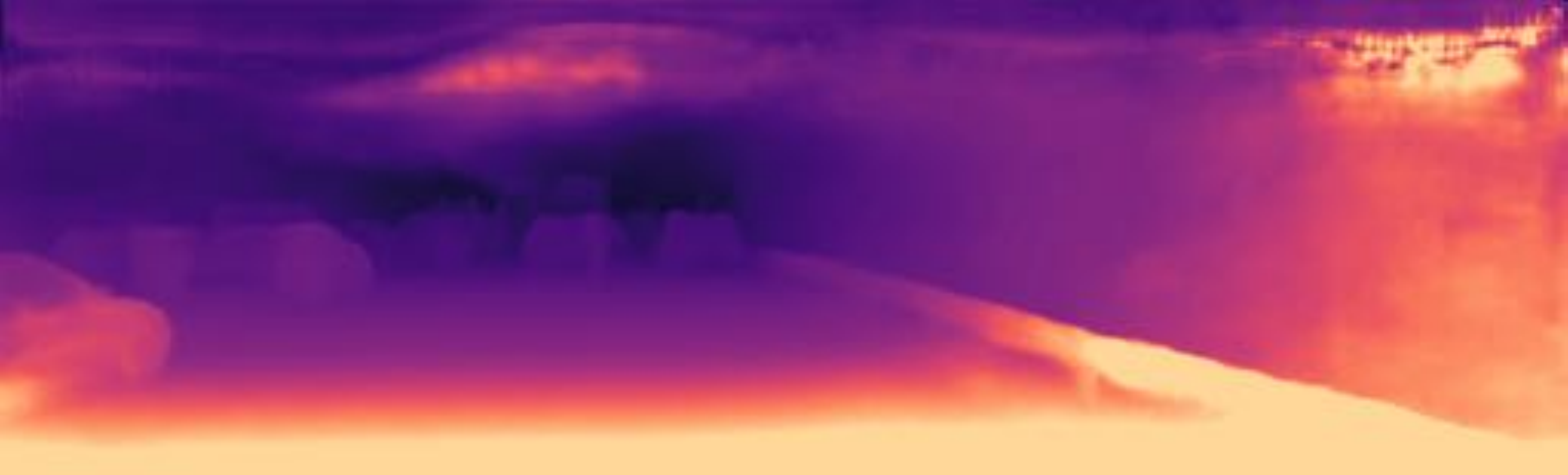}\qquad\qquad\quad &
\includegraphics[width=\iw,height=\ih]{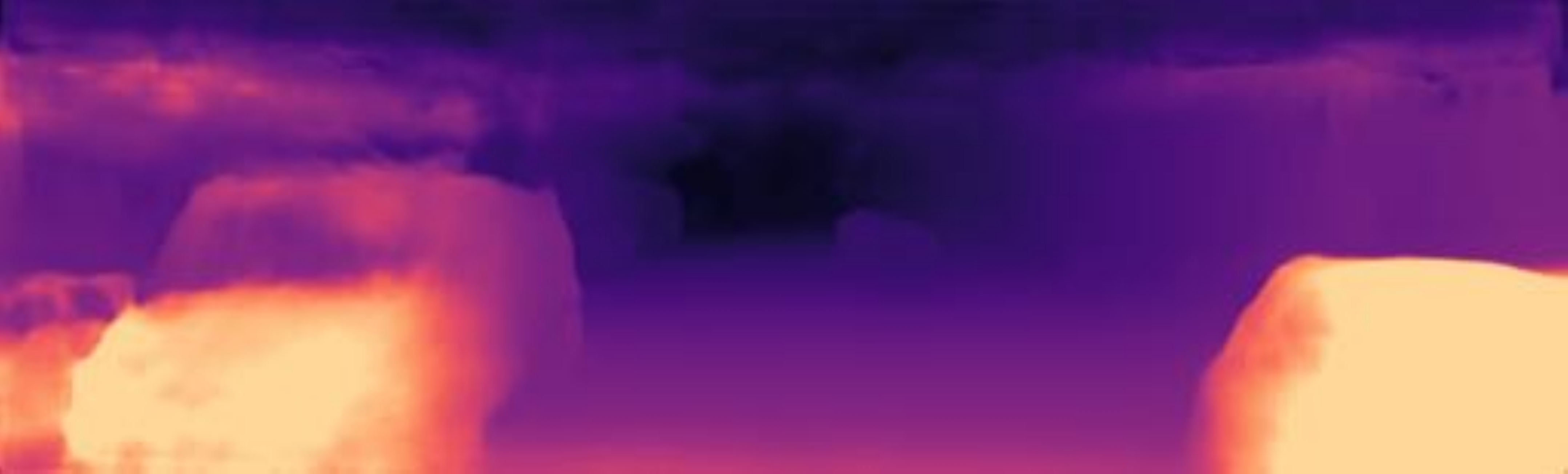}\qquad\qquad\quad &
\includegraphics[width=\iw,height=\ih]{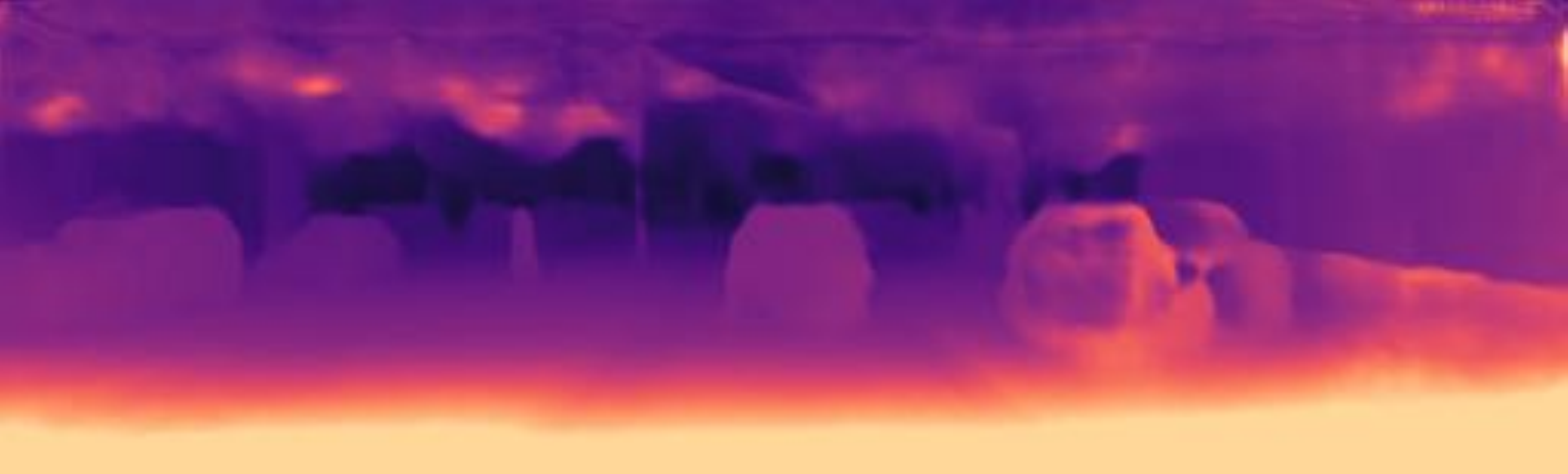}\\
\vspace{10mm}\\
\rotatebox[origin=c]{90}{\fontsize{\textw}{\texth}\selectfont DepthFormer\hspace{-310mm}}\hspace{10mm}
\includegraphics[width=\iw,height=\ih]{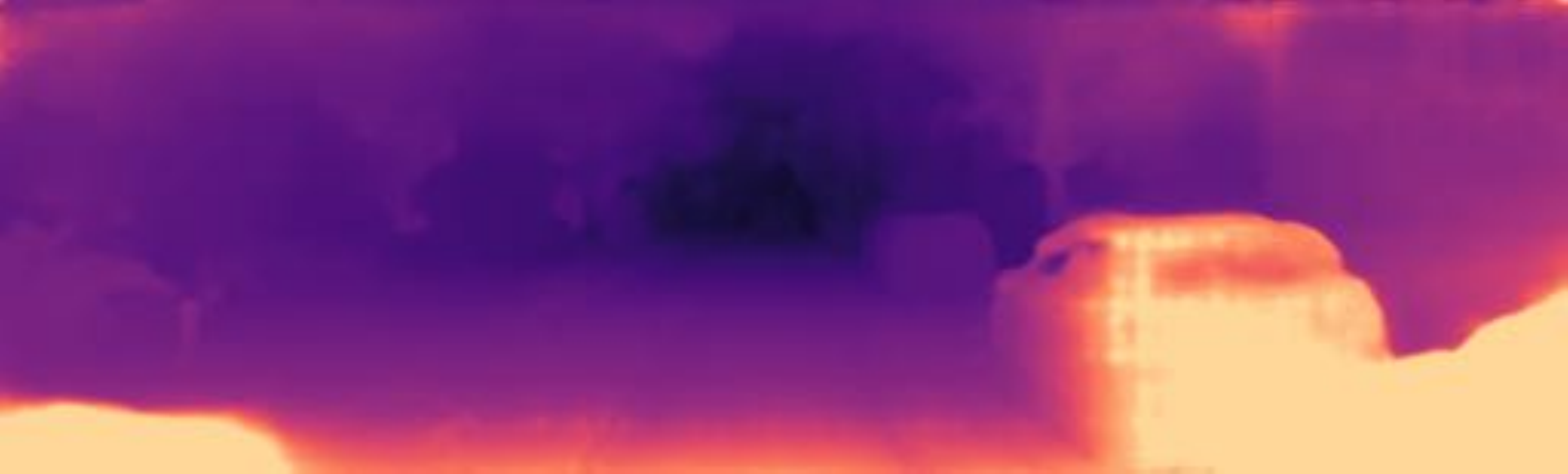}\qquad\qquad\quad &
\includegraphics[width=\iw,height=\ih]{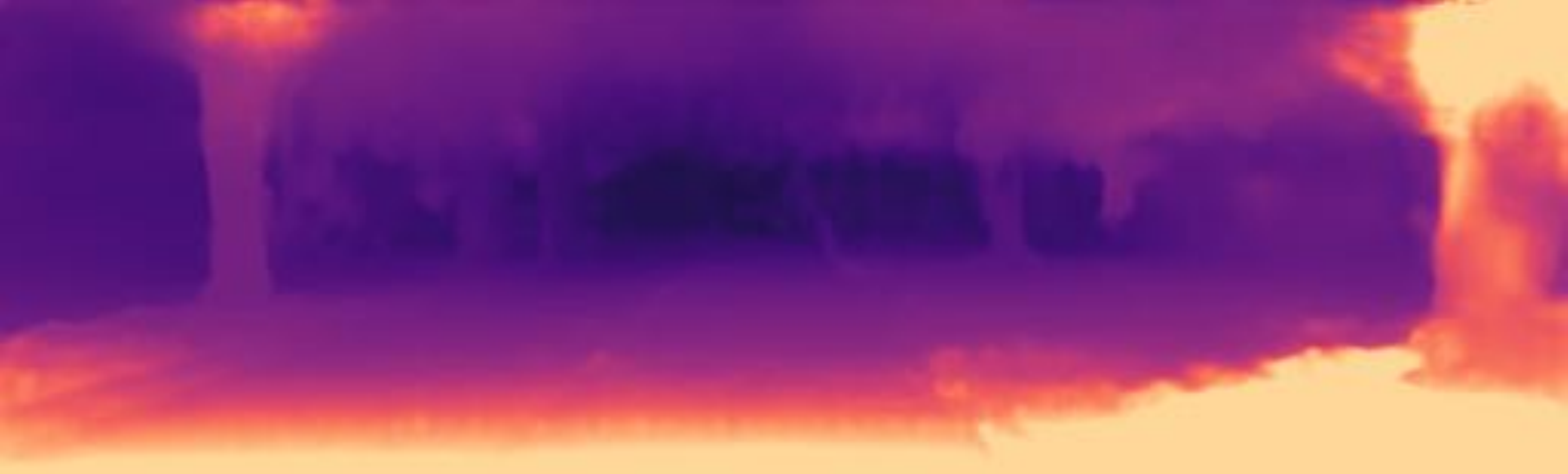}\qquad\qquad\quad &
\includegraphics[width=\iw,height=\ih]{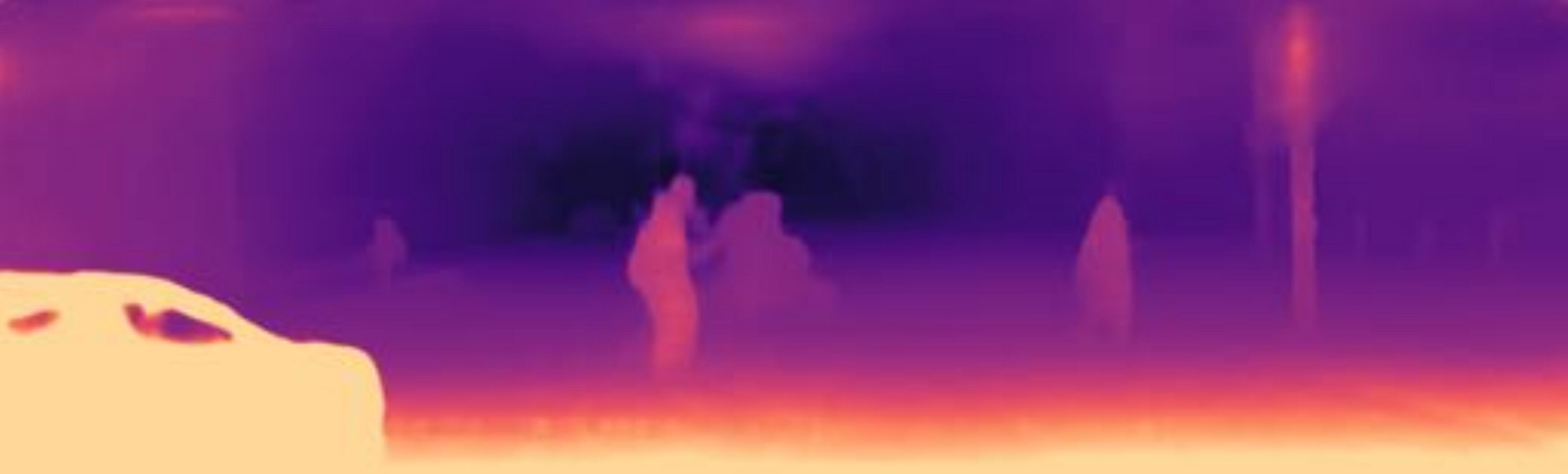}\qquad\qquad\quad &
\includegraphics[width=\iw,height=\ih]{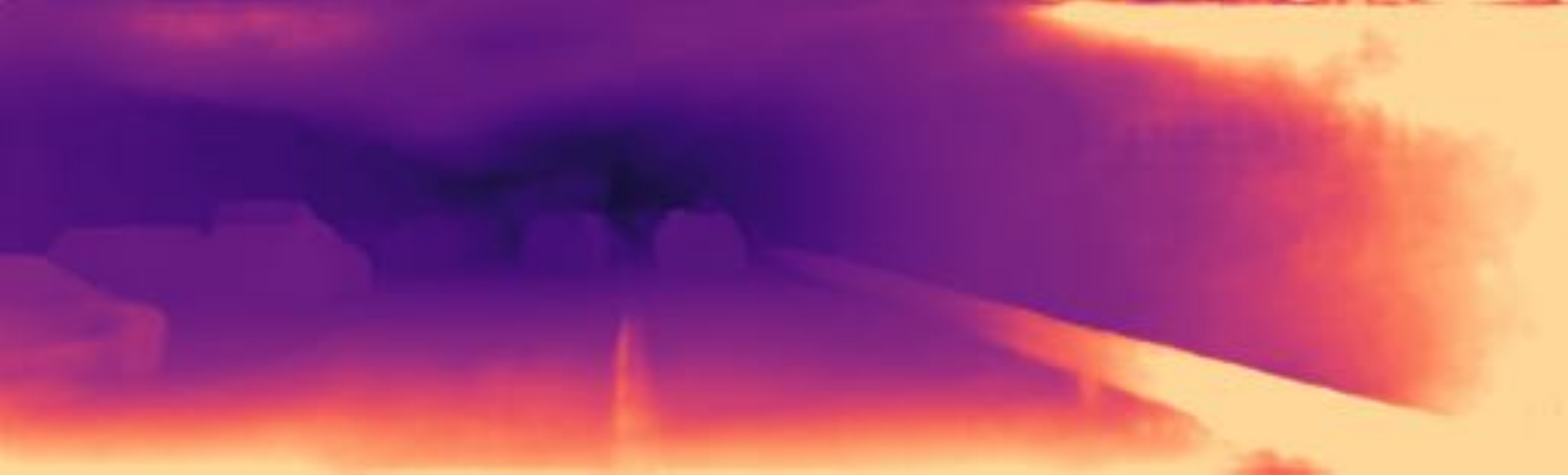}\qquad\qquad\quad &
\includegraphics[width=\iw,height=\ih]{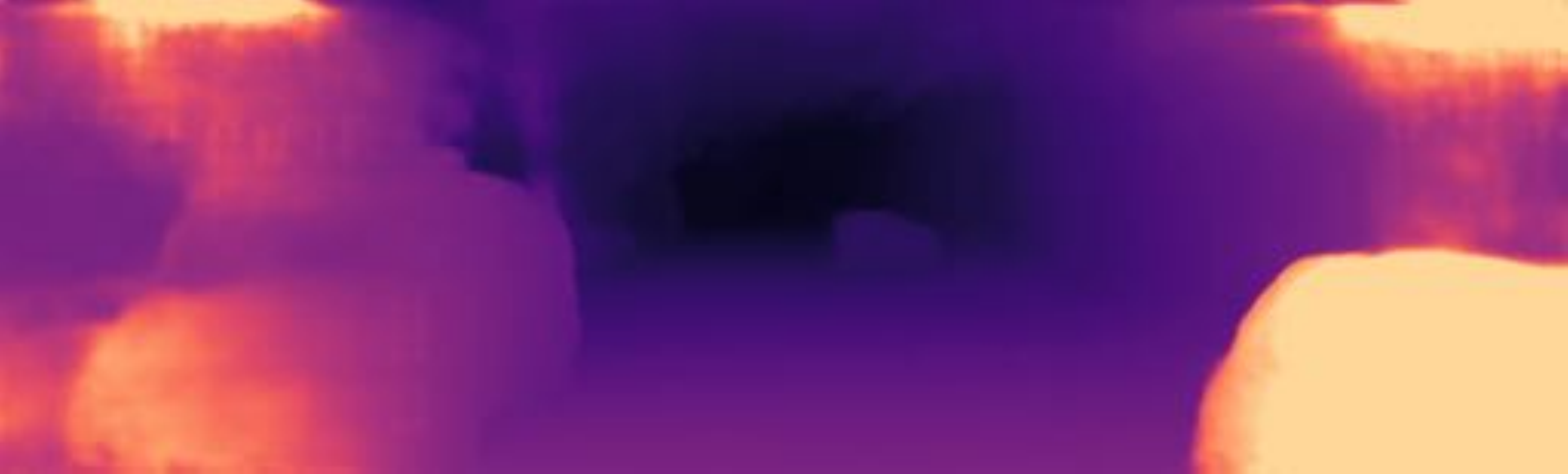}\qquad\qquad\quad &
\includegraphics[width=\iw,height=\ih]{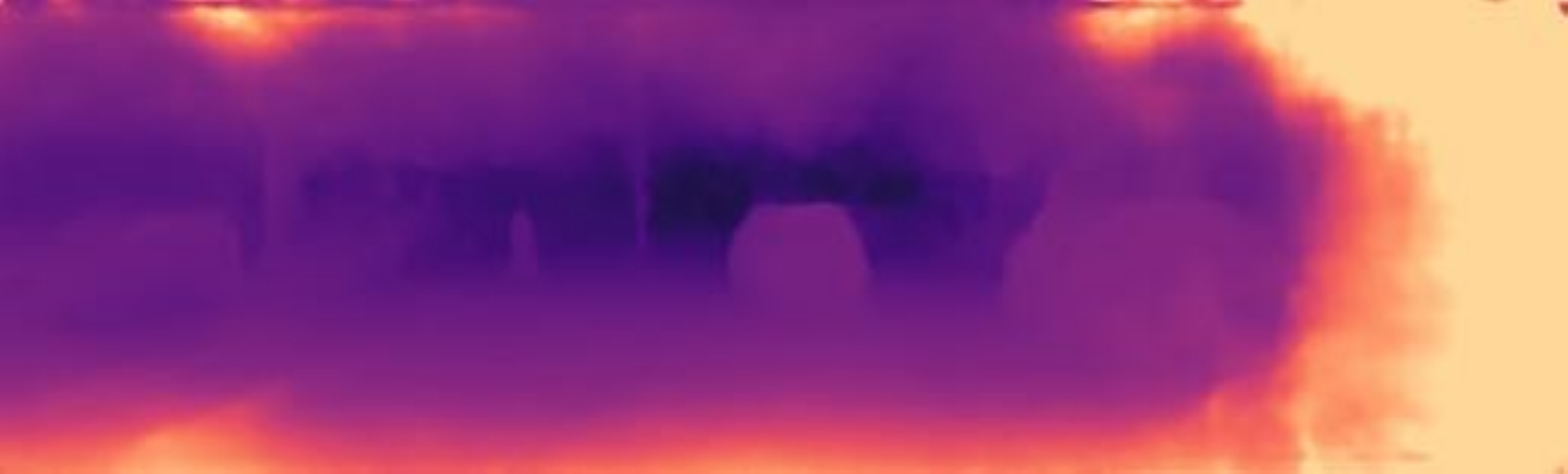}\\
\vspace{10mm}\\
\rotatebox[origin=c]{90}{\fontsize{\textw}{\texth}\selectfont GLPDepth\hspace{-310mm}}\hspace{10mm}
\includegraphics[width=\iw,height=\ih]{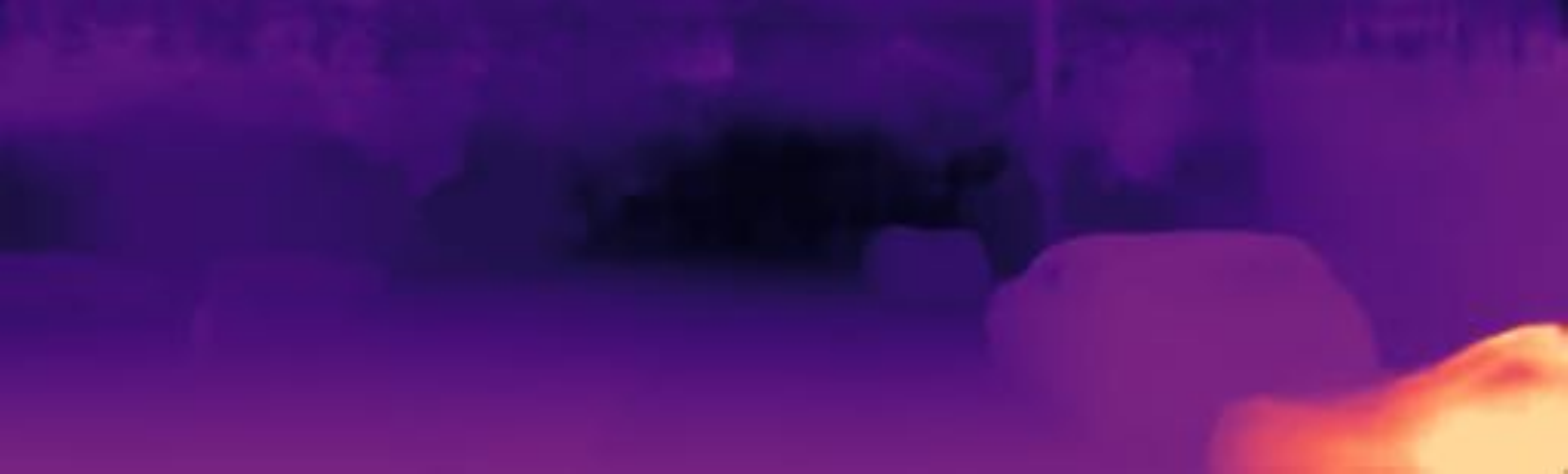}\qquad\qquad\quad &
\includegraphics[width=\iw,height=\ih]{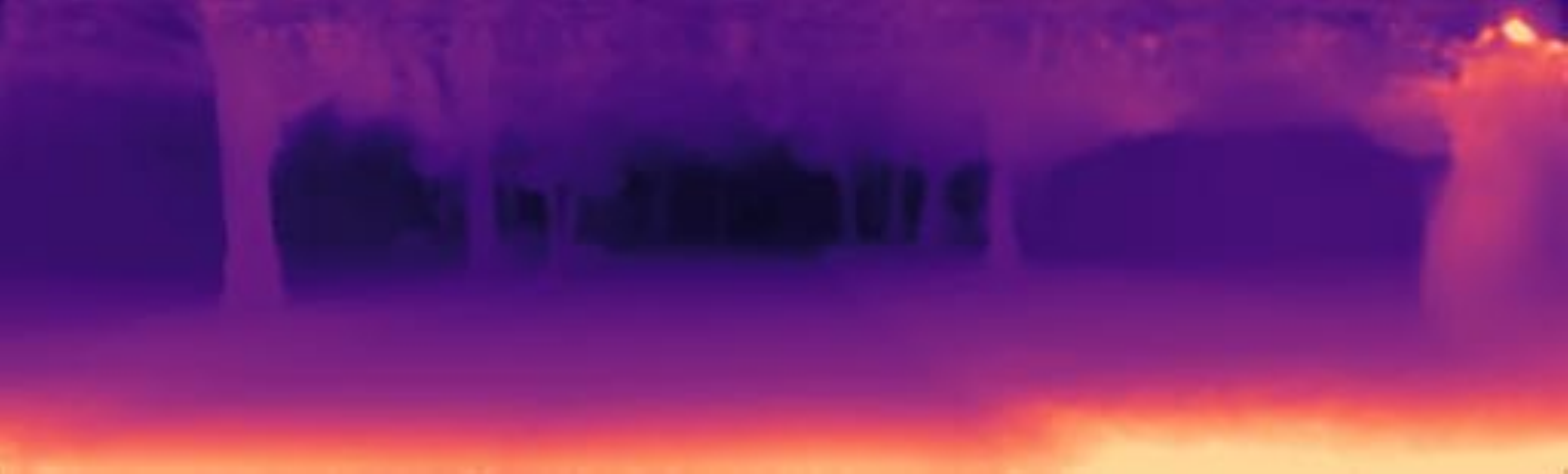}\qquad\qquad\quad &
\includegraphics[width=\iw,height=\ih]{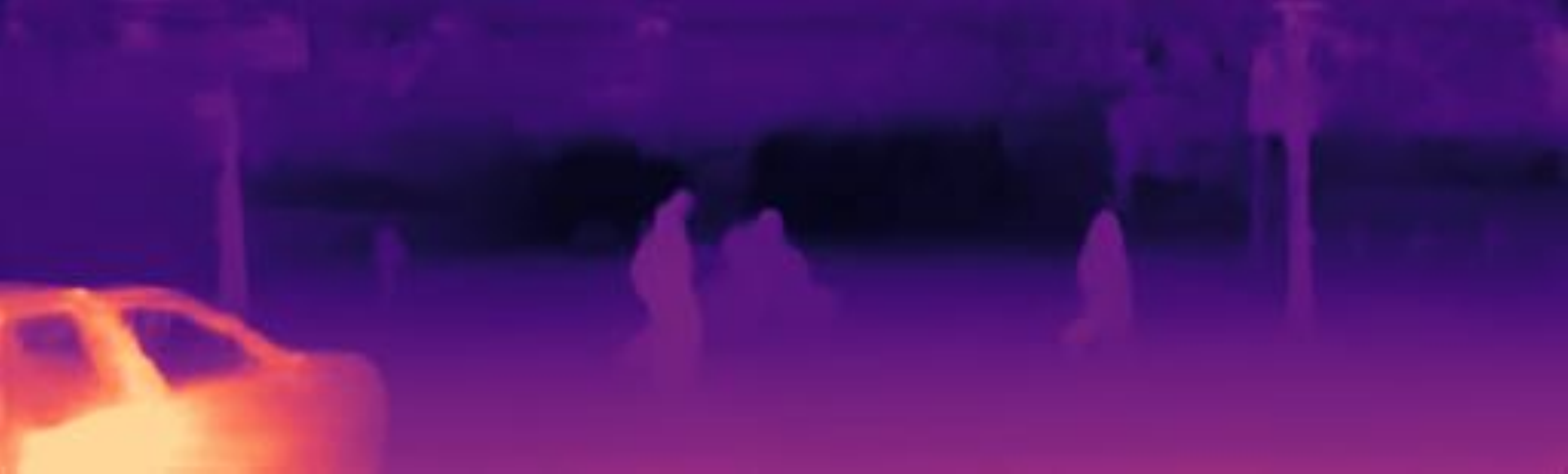}\qquad\qquad\quad &
\includegraphics[width=\iw,height=\ih]{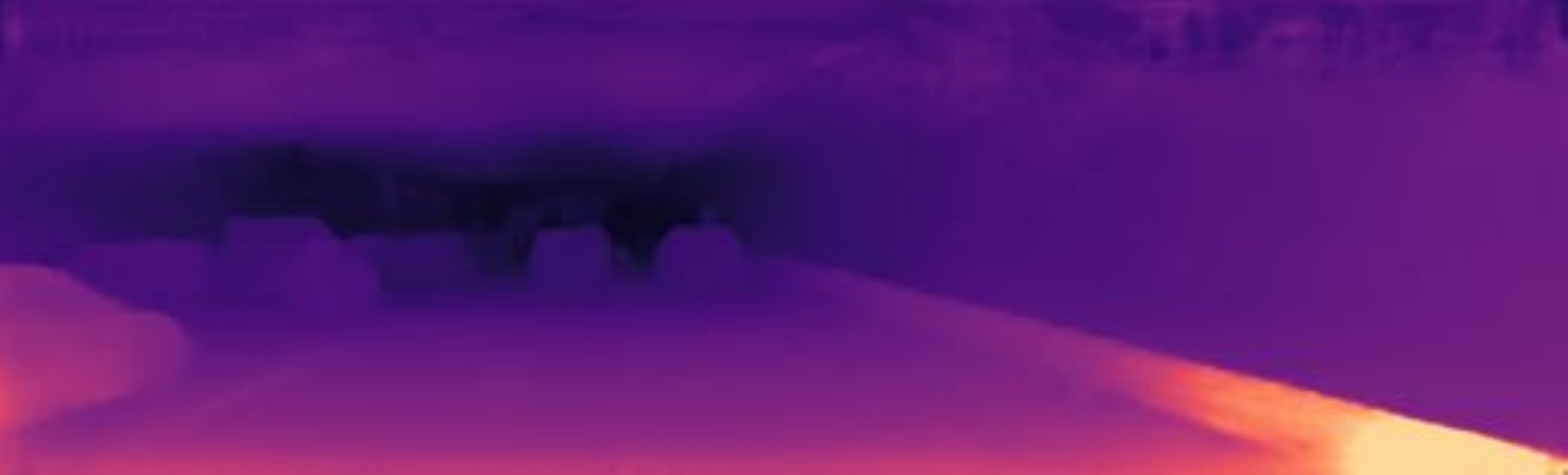}\qquad\qquad\quad &
\includegraphics[width=\iw,height=\ih]{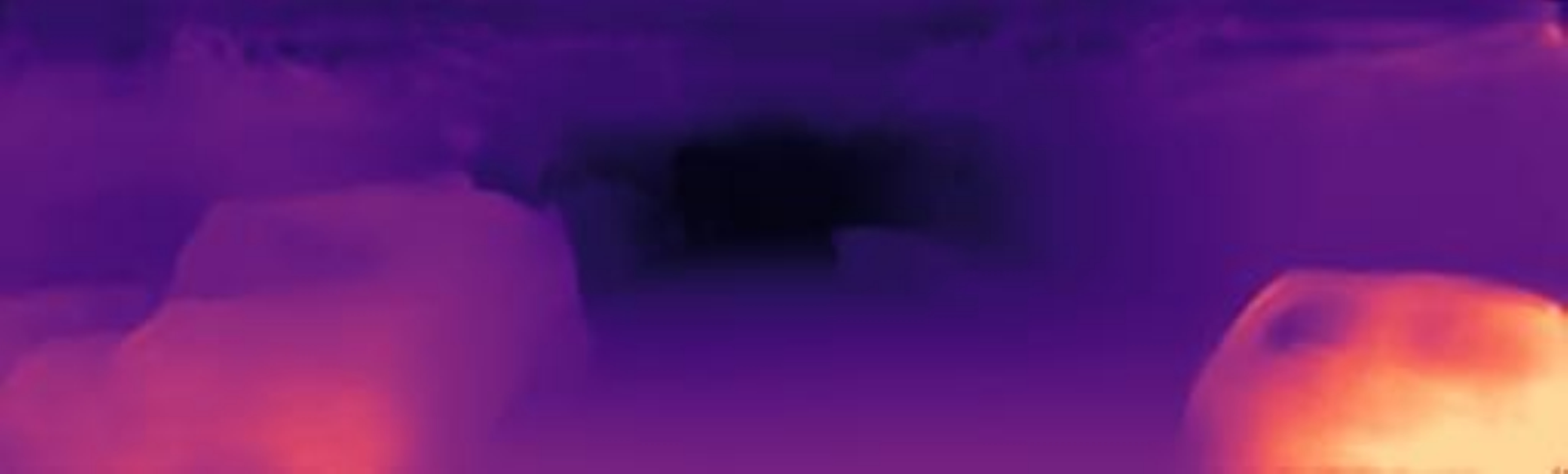}\qquad\qquad\quad &
\includegraphics[width=\iw,height=\ih]{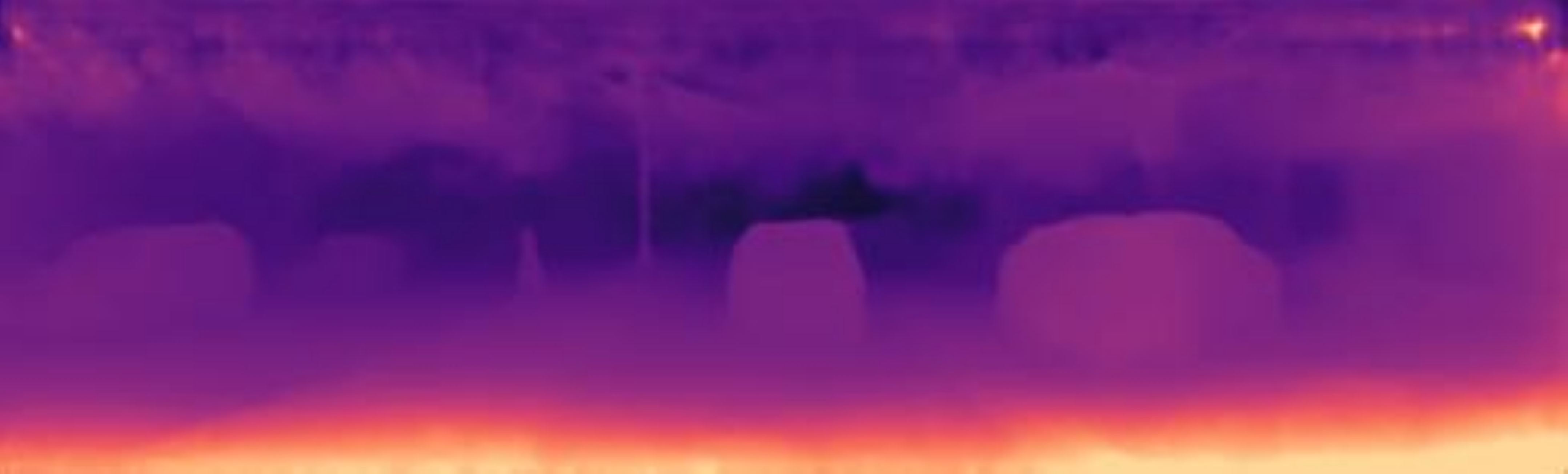}\\
\vspace{30mm}\\
\multicolumn{6}{c}{\fontsize{\w}{\h} \selectfont (d) Supervised Transformer-based methods} \\
\vspace{30mm}\\
\end{tabular}%
}
\caption{\textbf{Depth map results on synthetic texture-shifted (Watercolor, Pencil-sketch, Style-transfer) datasets.}}
\vspace{2mm}
\label{figure_result_texture_synthetic}
\end{figure*}

\begin{figure*}[t!]
\centering
\newcommand\textw{250}
\newcommand\texth{300}
\resizebox{\linewidth}{!}{
\begin{tabular}{cccc}
\includegraphics[scale=1.4]{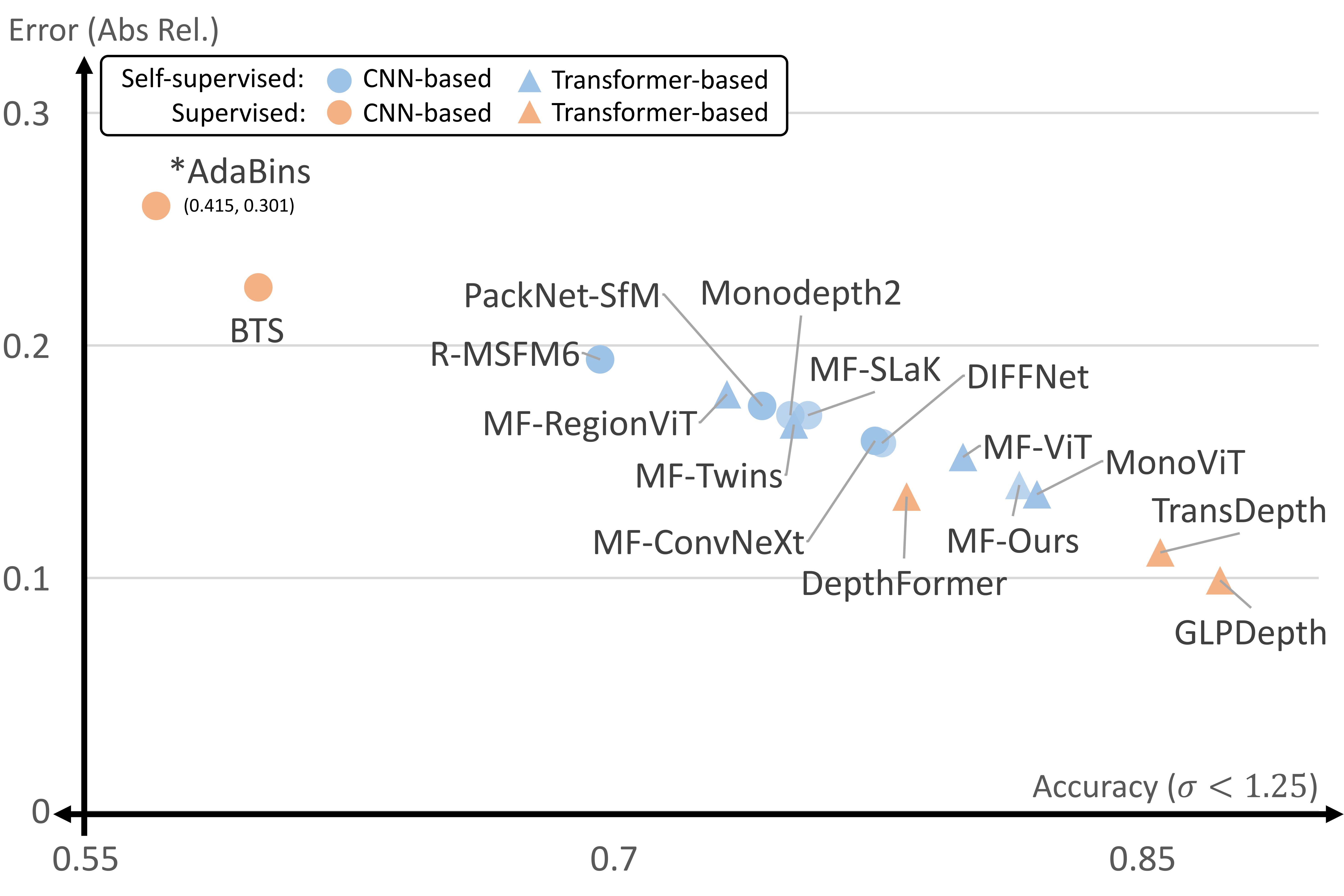} &
\includegraphics[scale=1.4]{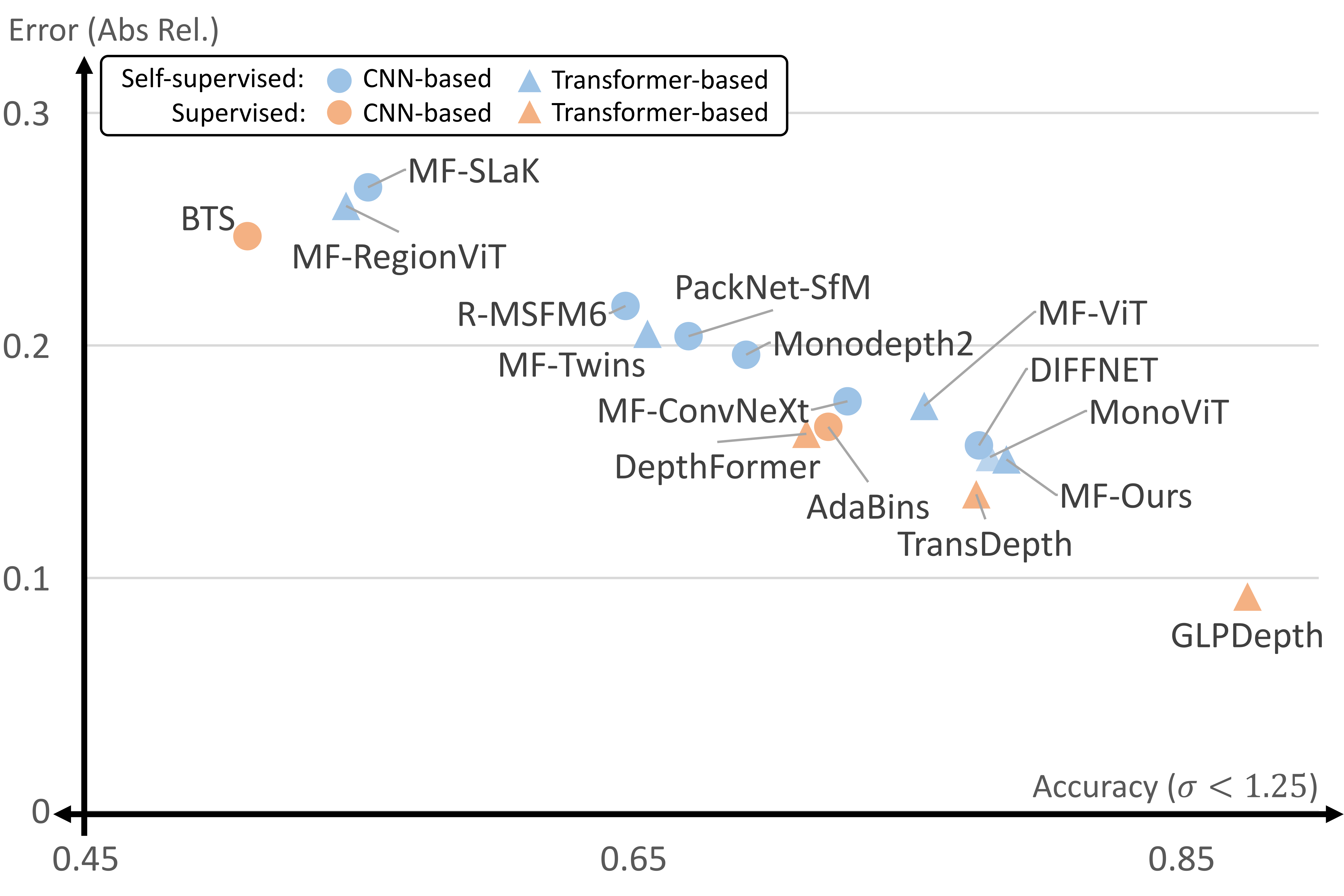} &
\includegraphics[scale=1.4]{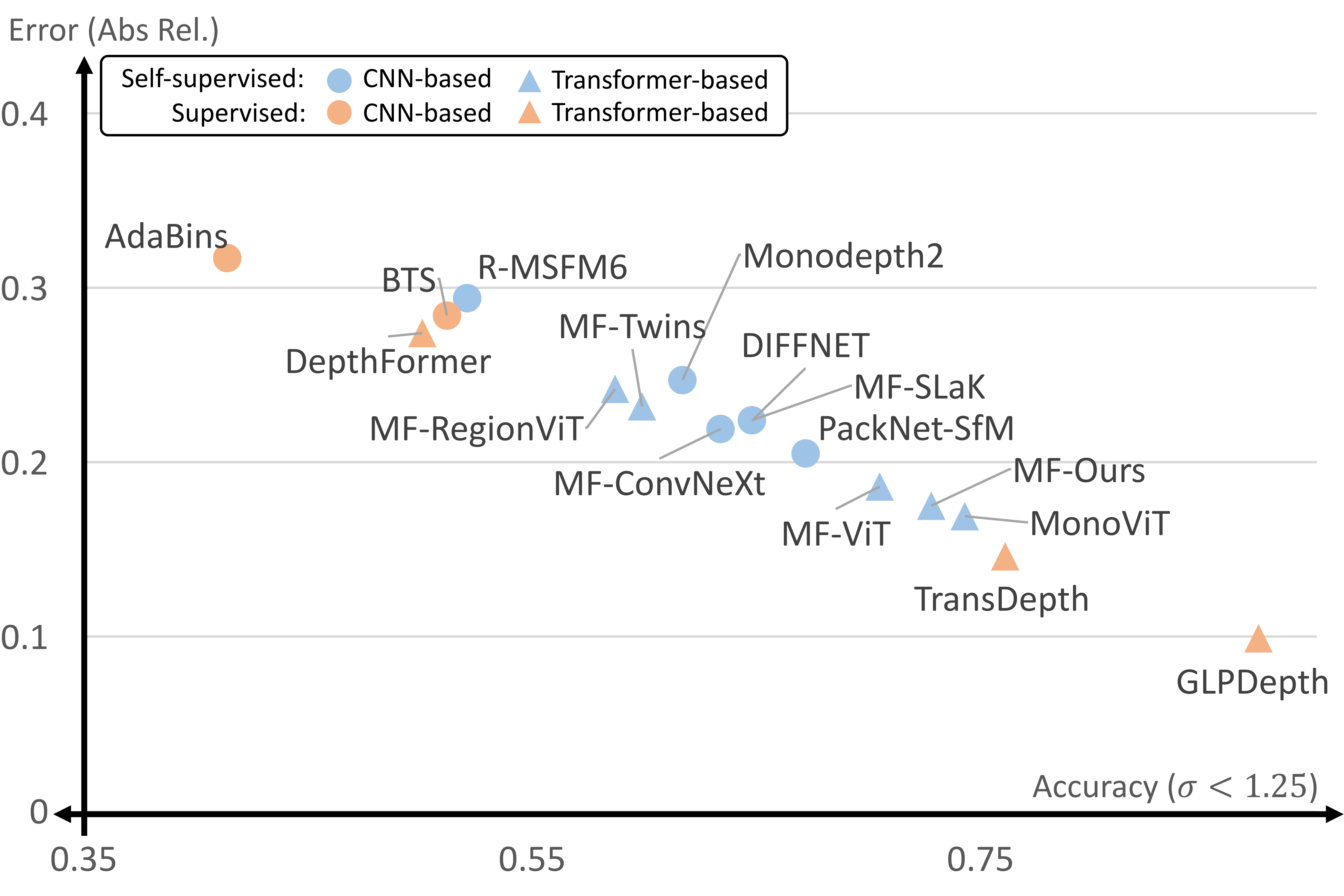} \vspace{30mm}\\
\fontsize{\textw}{\texth} \selectfont (a) Watercolor  & \fontsize{\textw}{\texth} \selectfont (b) Pencil-sketch & \fontsize{\textw}{\texth} \selectfont (c) Style-transfer  \\ 
\end{tabular}}
\caption{ \textbf{Depth Performance Comparison on synthetic texture-shifted (Watercolor, Pencil-sketch, Style-transfer) datasets.}}
\label{table_result_texture-shift}
\end{figure*}

\subsubsection{Texture-shifted datasets generation}
\label{data-generation}
In general, the texture is defined as an image's spatial color or pixel intensity pattern \cite{armi2019texture,lu2018deep,liu2019bow}. Inspired by \cite{geirhos2018imagenet}, we use three different texture shift strategies to investigate the impact of textures on the inference process in depth: texture smoothing (Watercolor), texture removal (Pencil-sketch), and texture transfer (Style-transfer).
The generated images and the corresponding results of each model are shown in \figref{figure_result_texture_synthetic}. The first two images are watercolored images, the middle two images and the last two images are pencil-sketch and style-transferred images, respectively. 
The following is a summary of the image generation:
\begin{table}[H]
\vspace{-1mm}
\resizebox{\columnwidth}{!}{ 
\normalsize
\begin{tabular}{lll}  
\textit{Watercolor} & \multicolumn{2}{p{0.85\linewidth}}{\raggedright We smooth the texture details from original images while preserving the color cues using \texttt{cv2.stylization}. The image looks like a watercolor pictures.} \\ 
\textit{Pencil-sketch} & \multicolumn{2}{p{0.85\linewidth}}{\raggedright We remove both textures and color from original images using \texttt{cv2.pencilSketch}. The image seems like a sketch drawn with pencils.} \\
\textit{Style-transfer} & \multicolumn{2}{p{0.85\linewidth}}{\raggedright We apply a new texture to the original image (context) by utilizing other images (style) using a style transfer \cite{gatys2016image}. 
The textures of the original images are changed.} \\ 
\end{tabular}}
\end{table}

\begin{figure*}[t!]
    \resizebox{\linewidth}{!}{%
    \includegraphics[]{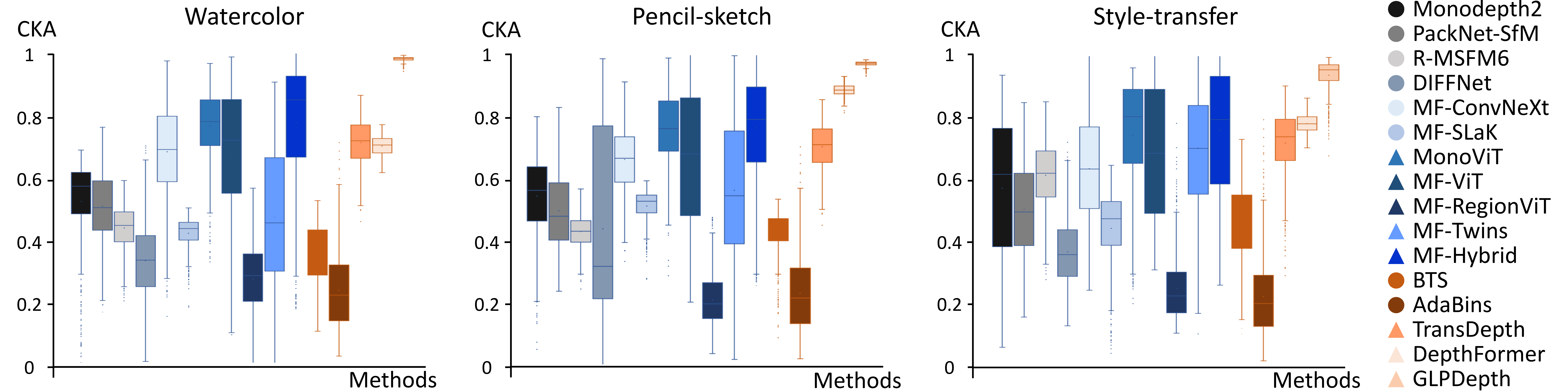}
    }
    \vspace{-2mm}
    \caption{\textbf{Box-and-Whisker plot of CKA similarity of all competitive methods on three different synthetic texture-shifted datasets.} The blue and red tones indicate self-supervised and supervised methods, respectively.
    The circle and triangle indicate CNN-based and Transformer-based models, respectively.}
    \vspace{-3mm}
\label{figure_result_feature}
\end{figure*}

\vspace{-4.5mm}
\begin{figure}[t!]
    \centering
    \resizebox{\columnwidth}{!}{
    \includegraphics{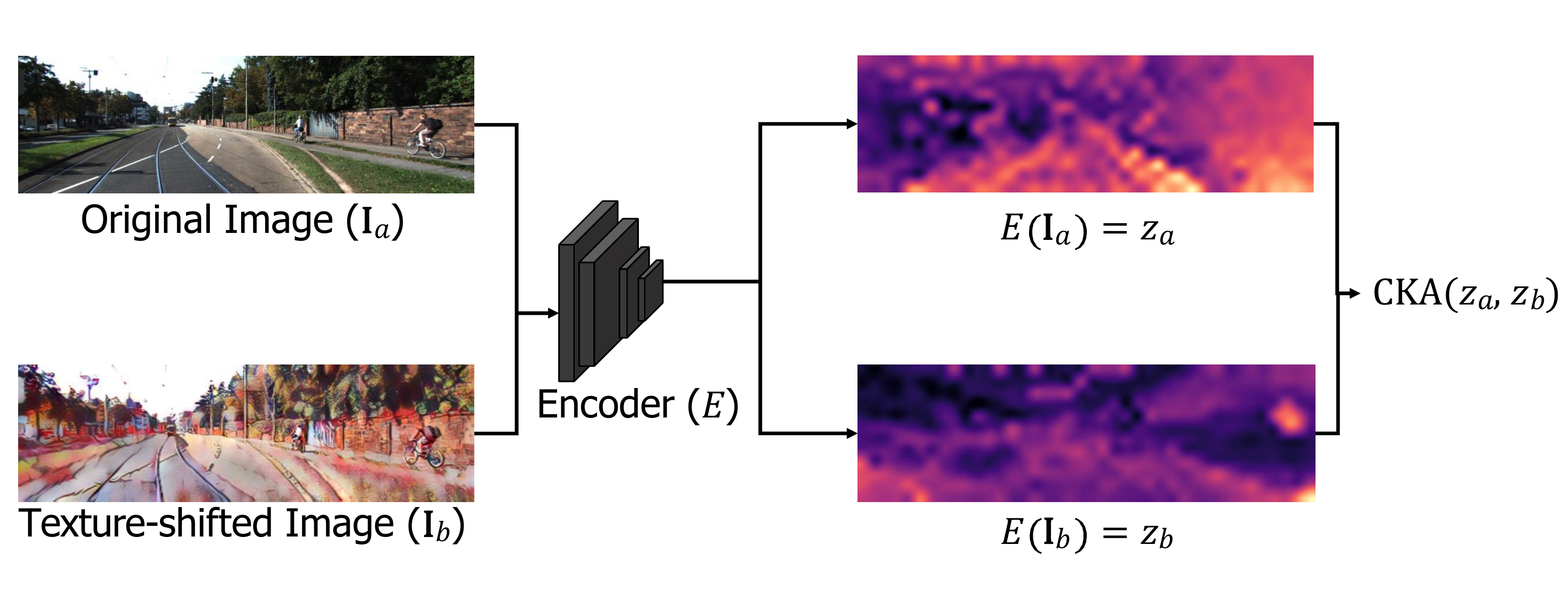}
    }
    \caption{Illustration of CKA \cite{kornblith2019similarity} similarity computation.}
    \label{figure_feature_process}
    \vspace{-4mm}
\end{figure}
\subsubsection{Evaluation on synthetic texture-shifted datasets}
\label{artifical-texture-exp}
We compare the performance of all the competitive methods and the modern backbones on the synthetic texture-shifted datasets, the same as the experiments in \secref{genealization-exp}. 
The qualitative and quantitative results are shown in \figref{figure_result_texture_synthetic}, \figref{table_result_texture-shift} and \tabref{table_result_texture}\footnote[2]{Note that this \tabref{table_result_texture} is in the appendix.}.
As seen in \secref{genealization-exp}, both the Transformer-based models produce better depth maps than pure CNN-based methods regardless of supervised or self-supervised methods.
In particular, the depth results from CNN-based models are unrecognizable with the strong texture-shifted datasets, especially the style-transferred data.
We also observe that MF-ConvNeXt shows a tolerable depth map on texture shift datasets and has lower errors than other CNN-based models, although it is purely CNN-based. These experiments support our two findings observed in \secref{genealization-exp}.
One is that networks whose encoder consists of Transformers are generally robust to texture changes.
However, MF-ConvNeXt and MF-RegionViT show different aspects from the CNN-based and Transformer-based models, which are the respective backbone models.
In the following section, we deeply analyze the intermediate feature representations of all backbones to verify the reason for these observations.

\begin{figure*}[p!]
\centering
\newcommand\w{240}
\newcommand\h{220}
\newcommand\iw{80cm}
\newcommand\ih{35cm}
\newcommand\textw{150}
\newcommand\texth{200}
\resizebox{\linewidth}{!}{%
\begin{tabular}{ccccccc}
\multicolumn{2}{c}{\fontsize{\w}{\h} \selectfont RobotCar } & 
\multicolumn{2}{c}{\fontsize{\w}{\h} \selectfont Foggy CityScapes } & 
\multicolumn{2}{c}{\fontsize{\w}{\h} \selectfont Rainy CityScapes }  \\
\vspace{30mm}\\
\rotatebox[origin=c]{90}{\fontsize{\textw}{\texth}\selectfont Input Images\hspace{-315mm}}\hspace{10mm}
\includegraphics[width=\iw,height=\ih]{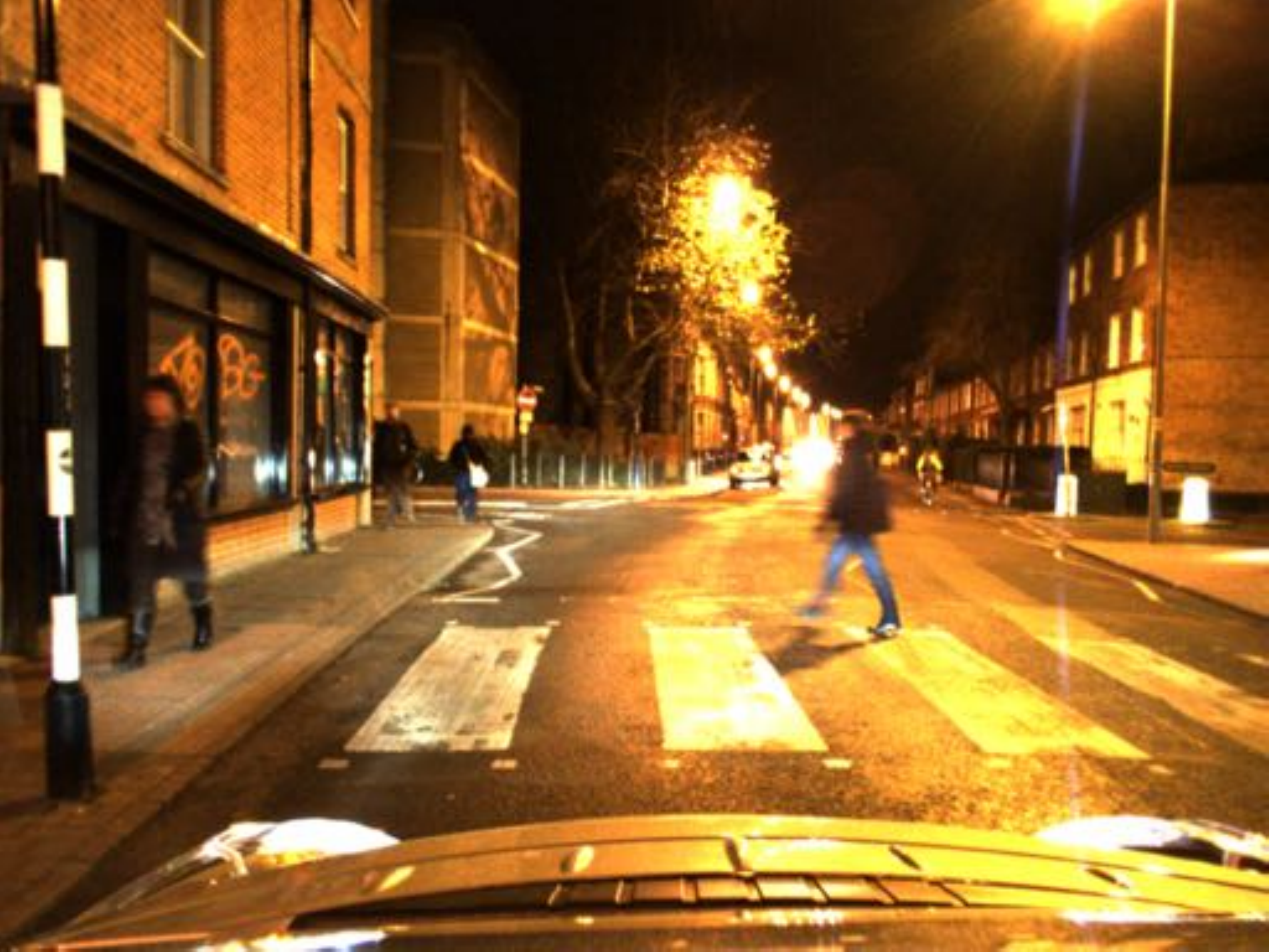}\qquad\qquad\quad &  
\includegraphics[width=\iw,height=\ih]{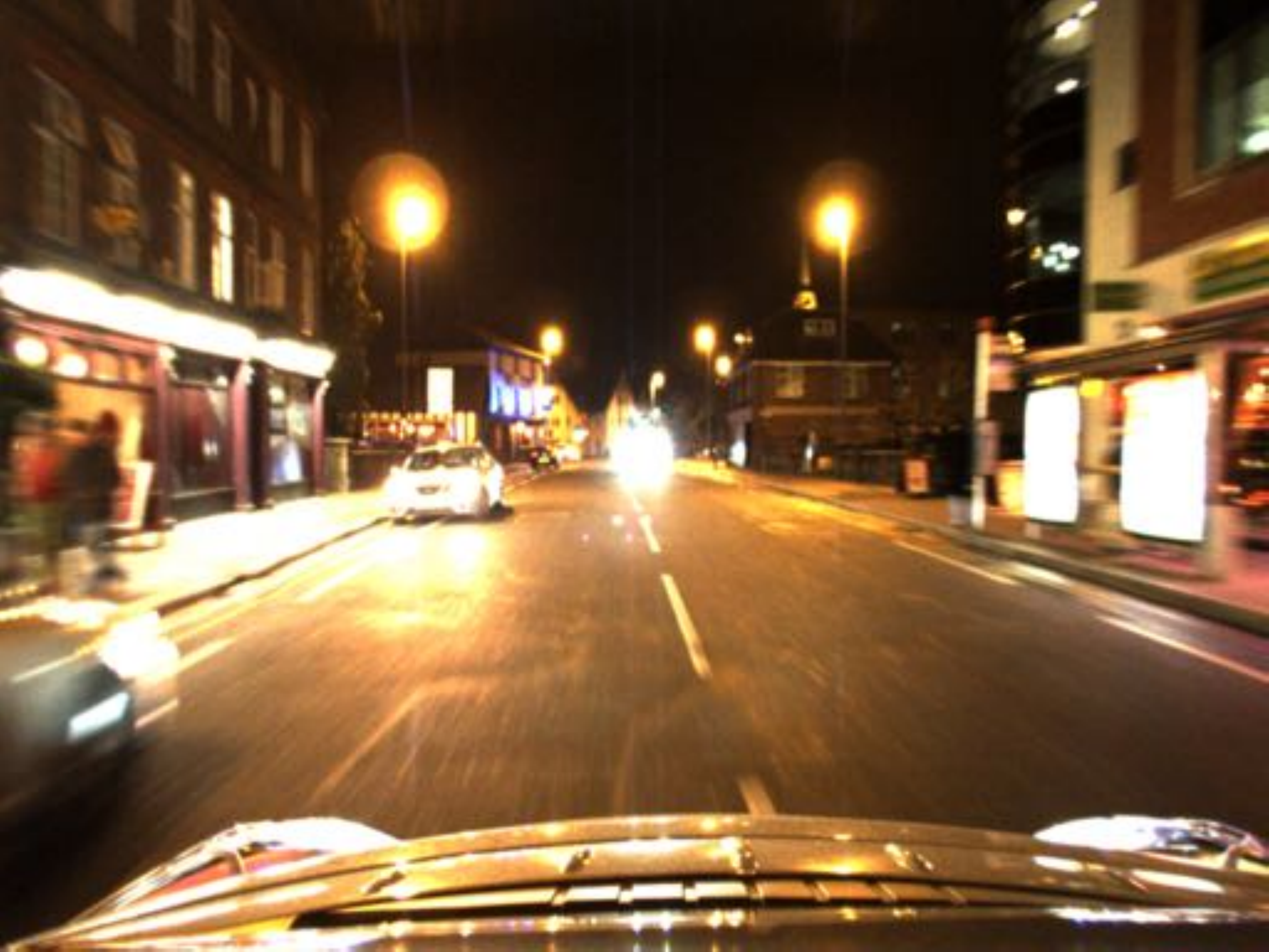}\qquad\qquad\quad &
\includegraphics[width=\iw,height=\ih]{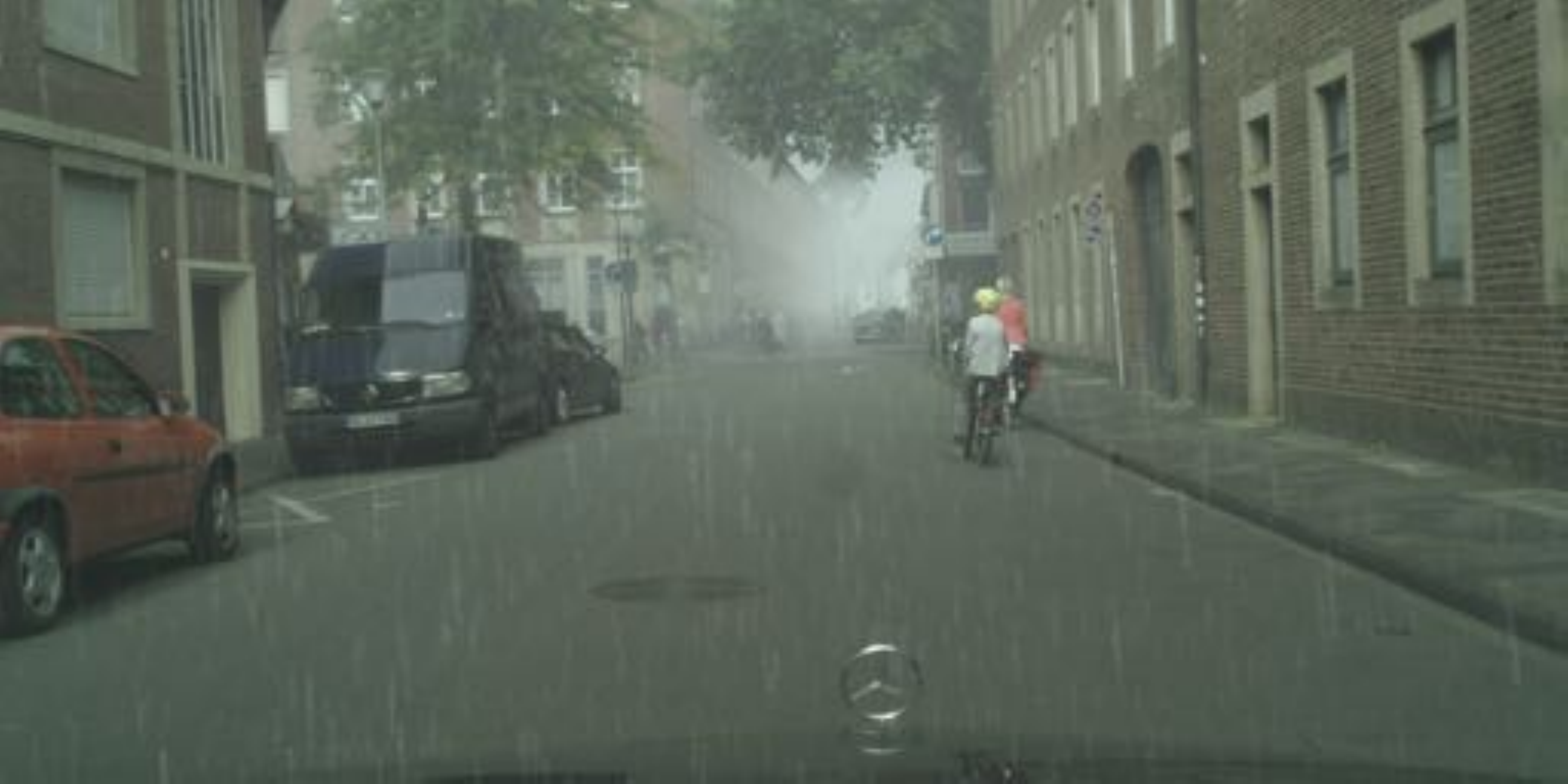}\qquad\qquad\quad &  
\includegraphics[width=\iw,height=\ih]{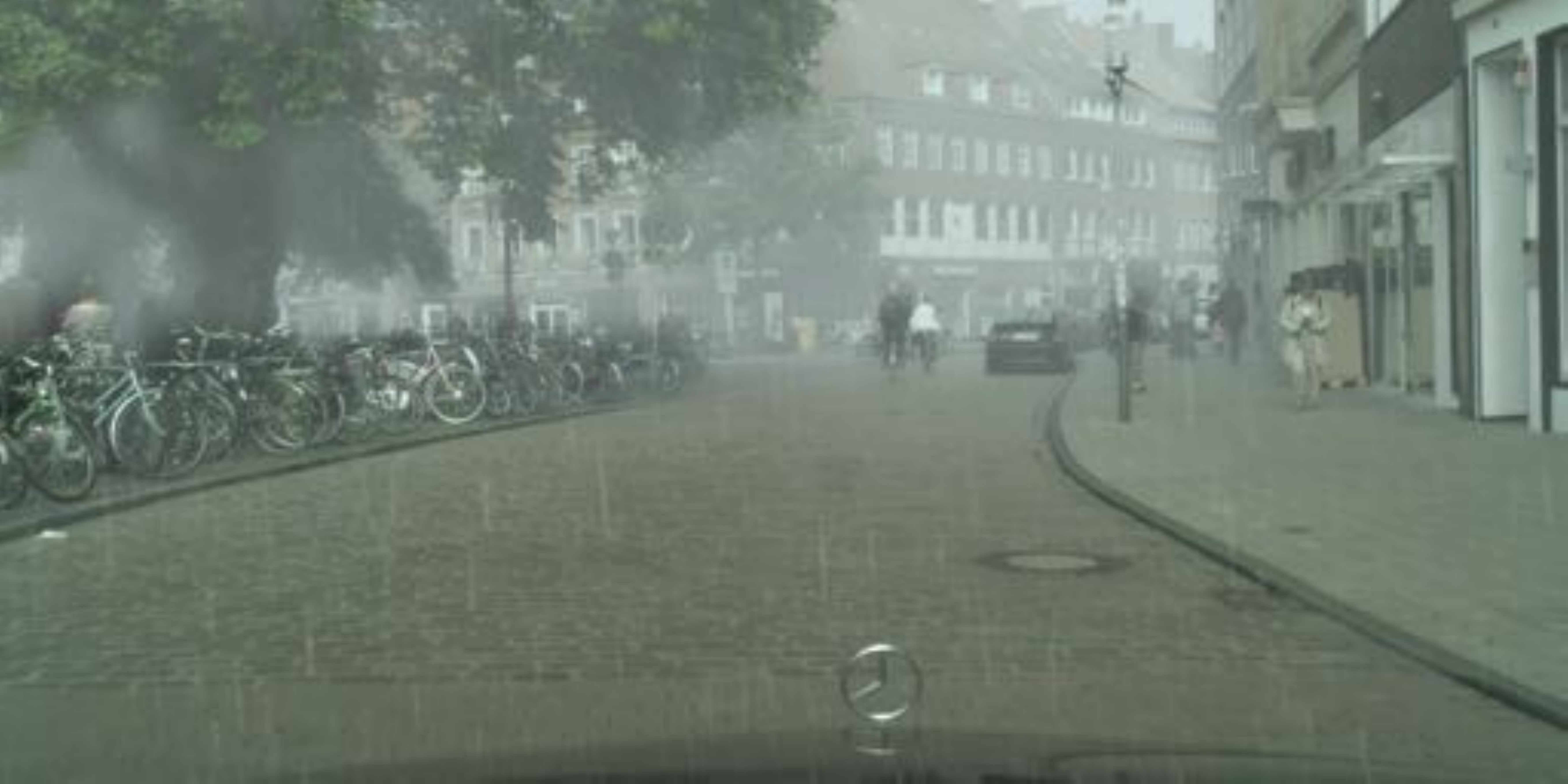}\qquad\qquad\quad &
\includegraphics[width=\iw,height=\ih]{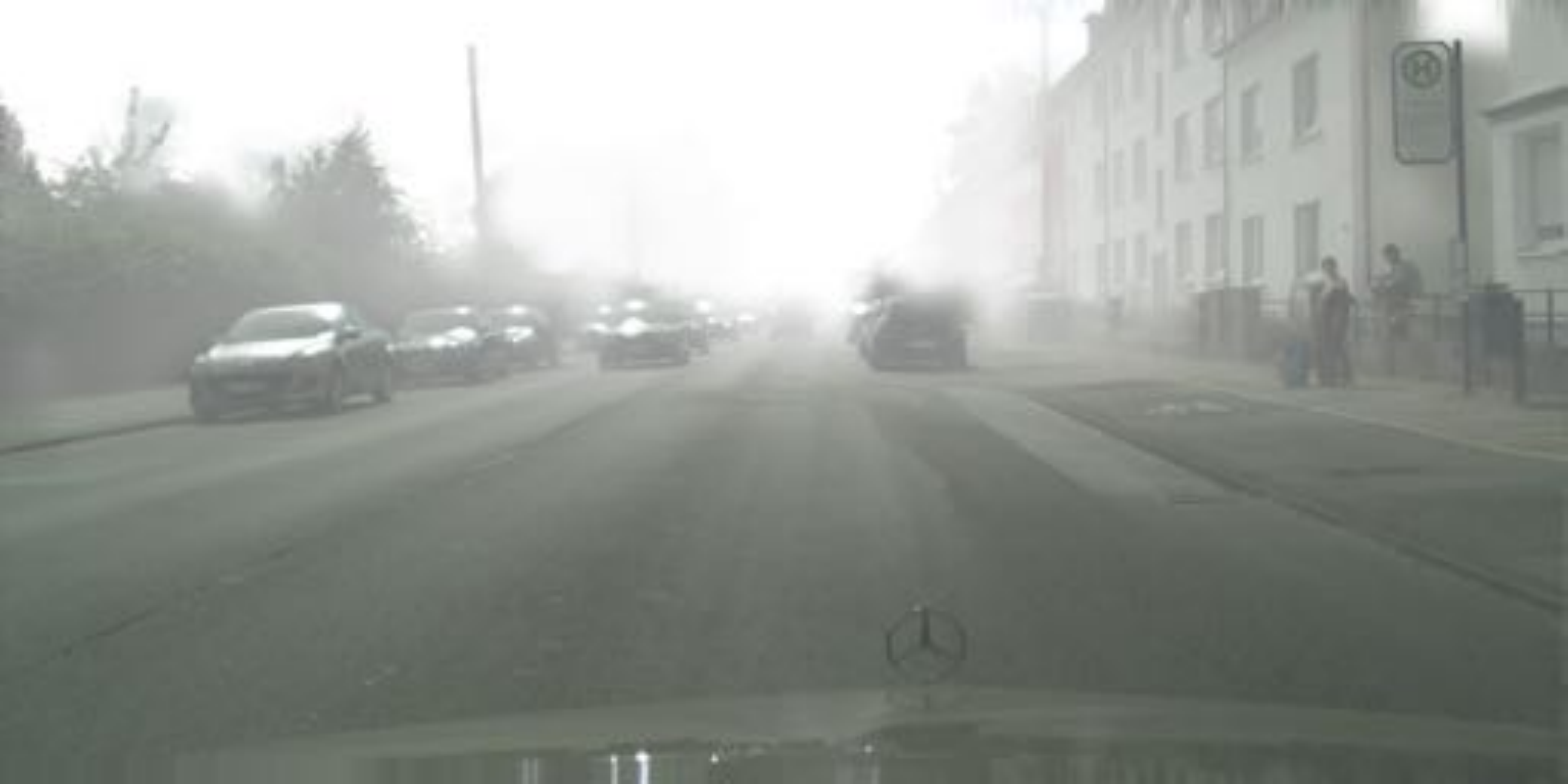}\qquad\qquad\quad &  
\includegraphics[width=\iw,height=\ih]{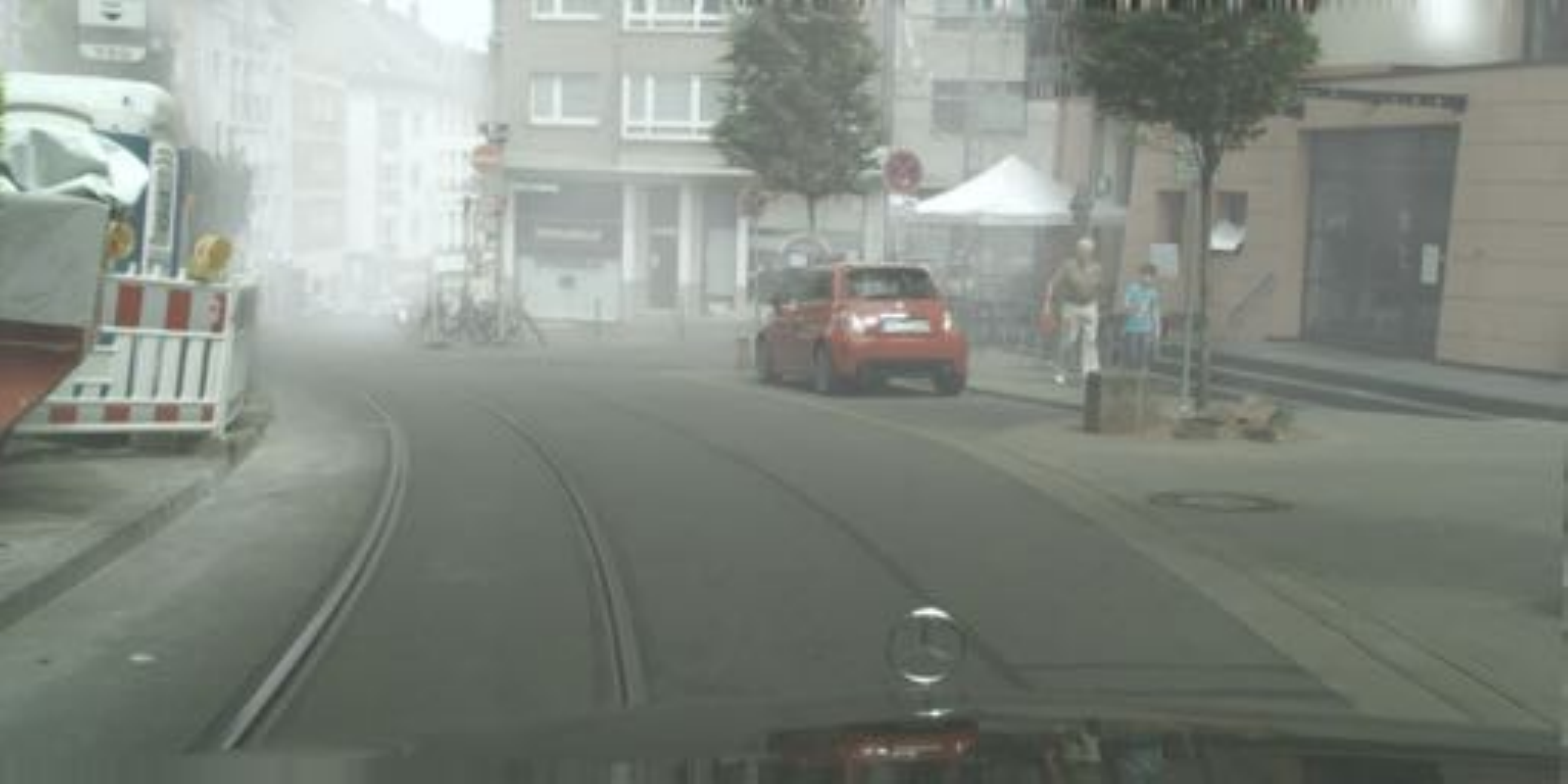}\\
\vspace{30mm}\\
\\\cmidrule{1-6}
\vspace{30mm}\\
\rotatebox[origin=c]{90}{\fontsize{\textw}{\texth}\selectfont Monodepth2\hspace{-320mm}}\hspace{8mm}
\includegraphics[width=\iw,height=\ih]{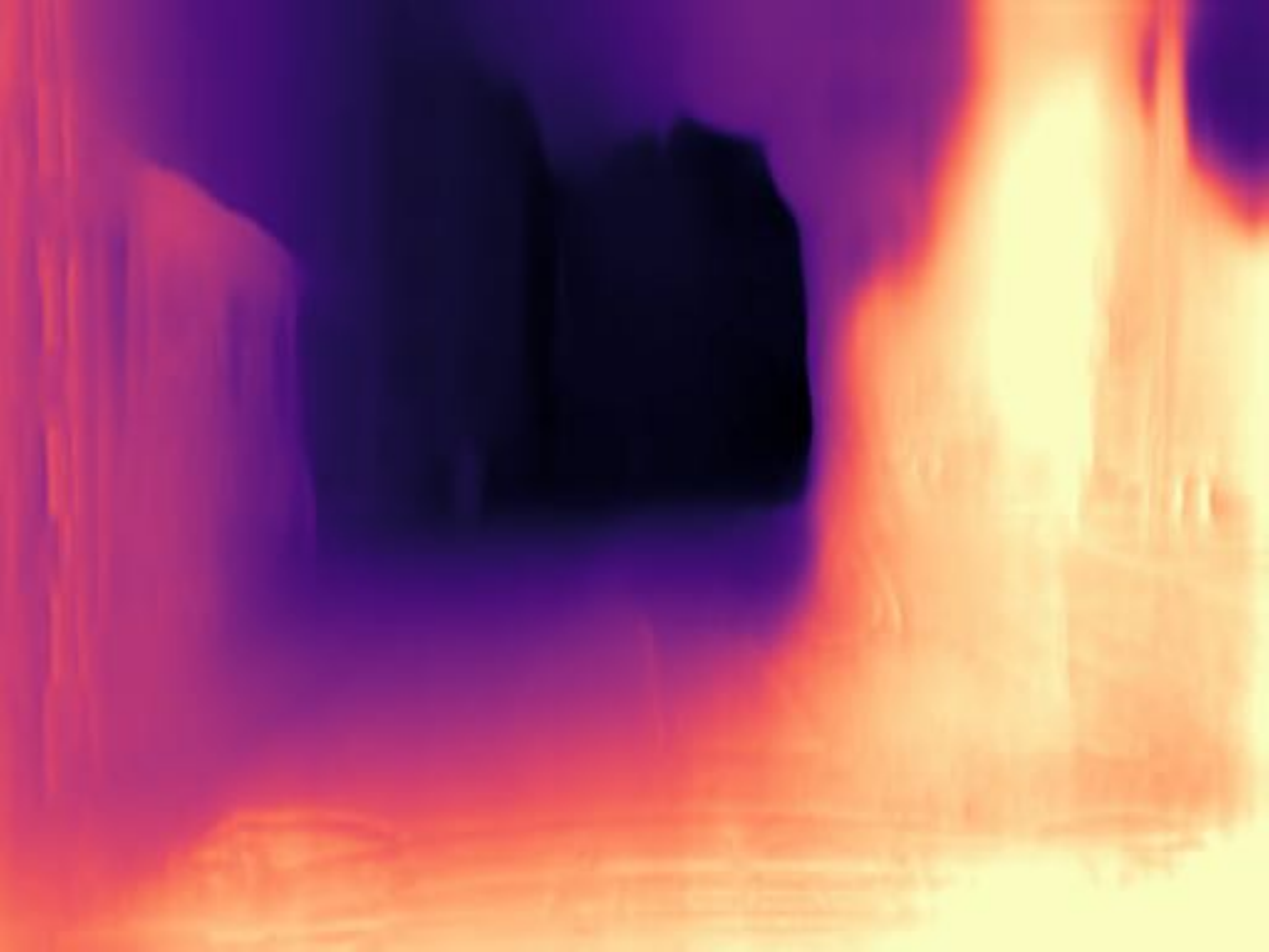}\qquad\qquad\quad &  
\includegraphics[width=\iw,height=\ih]{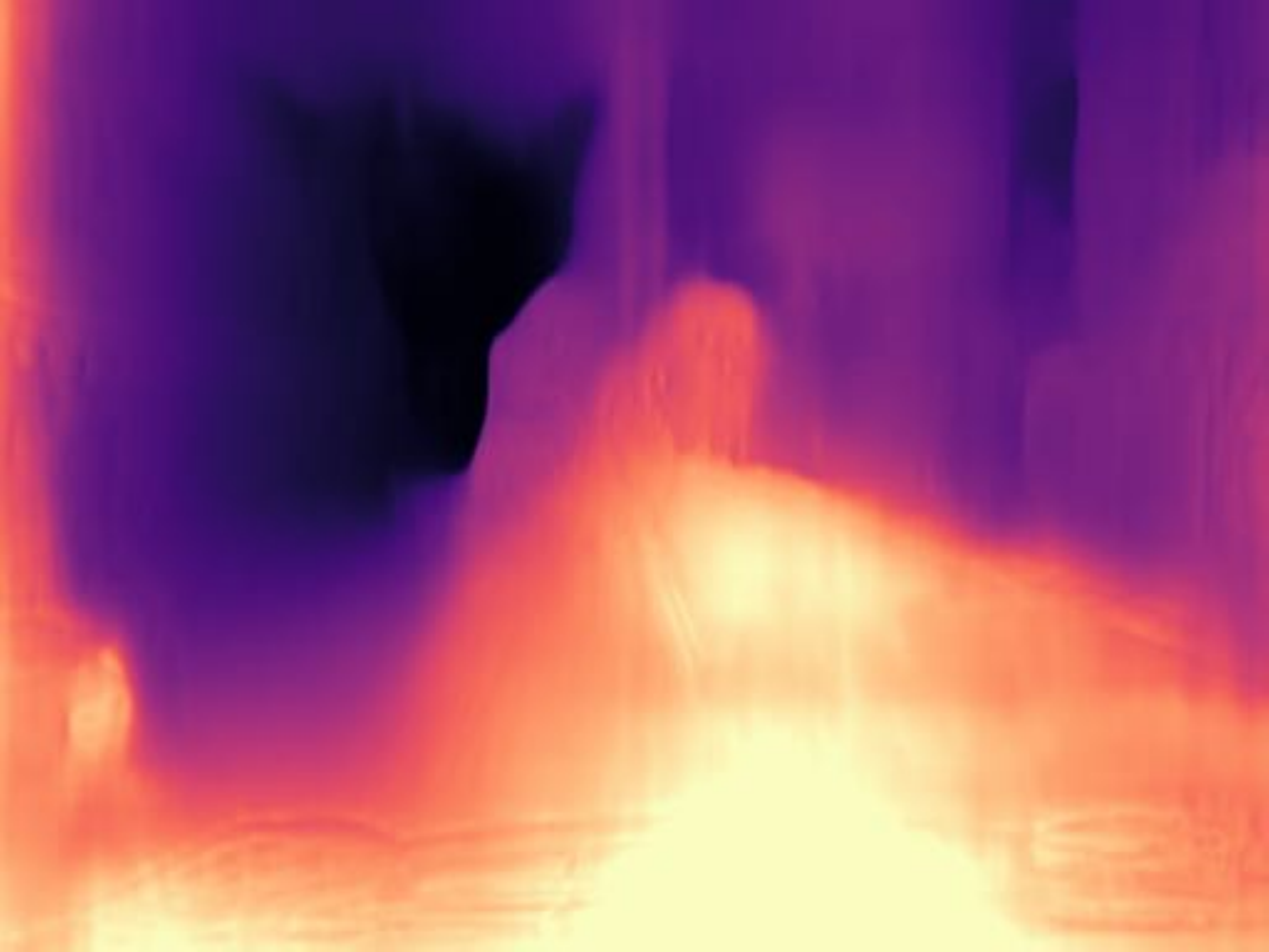}\qquad\qquad\quad &
\includegraphics[width=\iw,height=\ih]{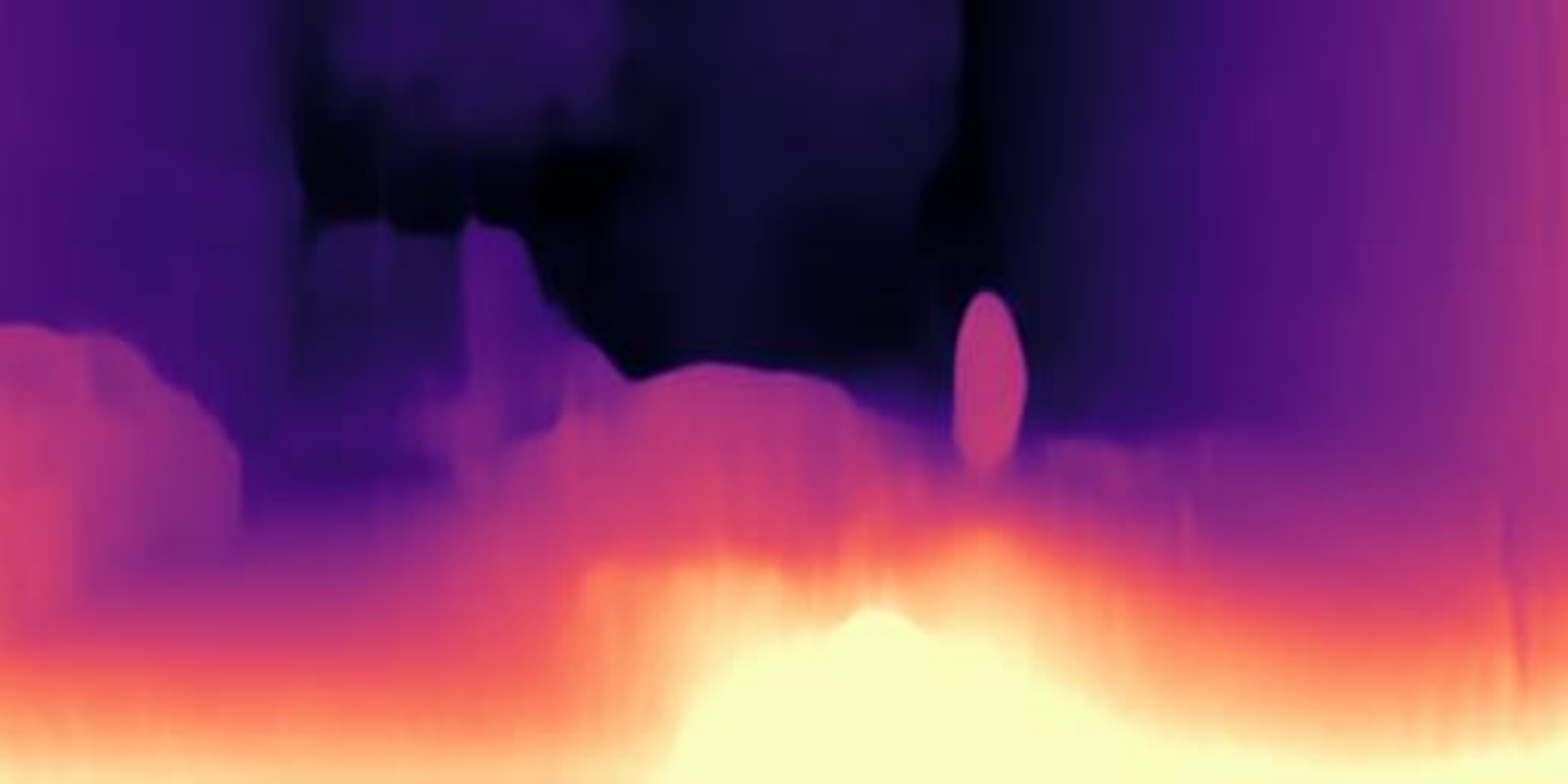}\qquad\qquad\quad &  
\includegraphics[width=\iw,height=\ih]{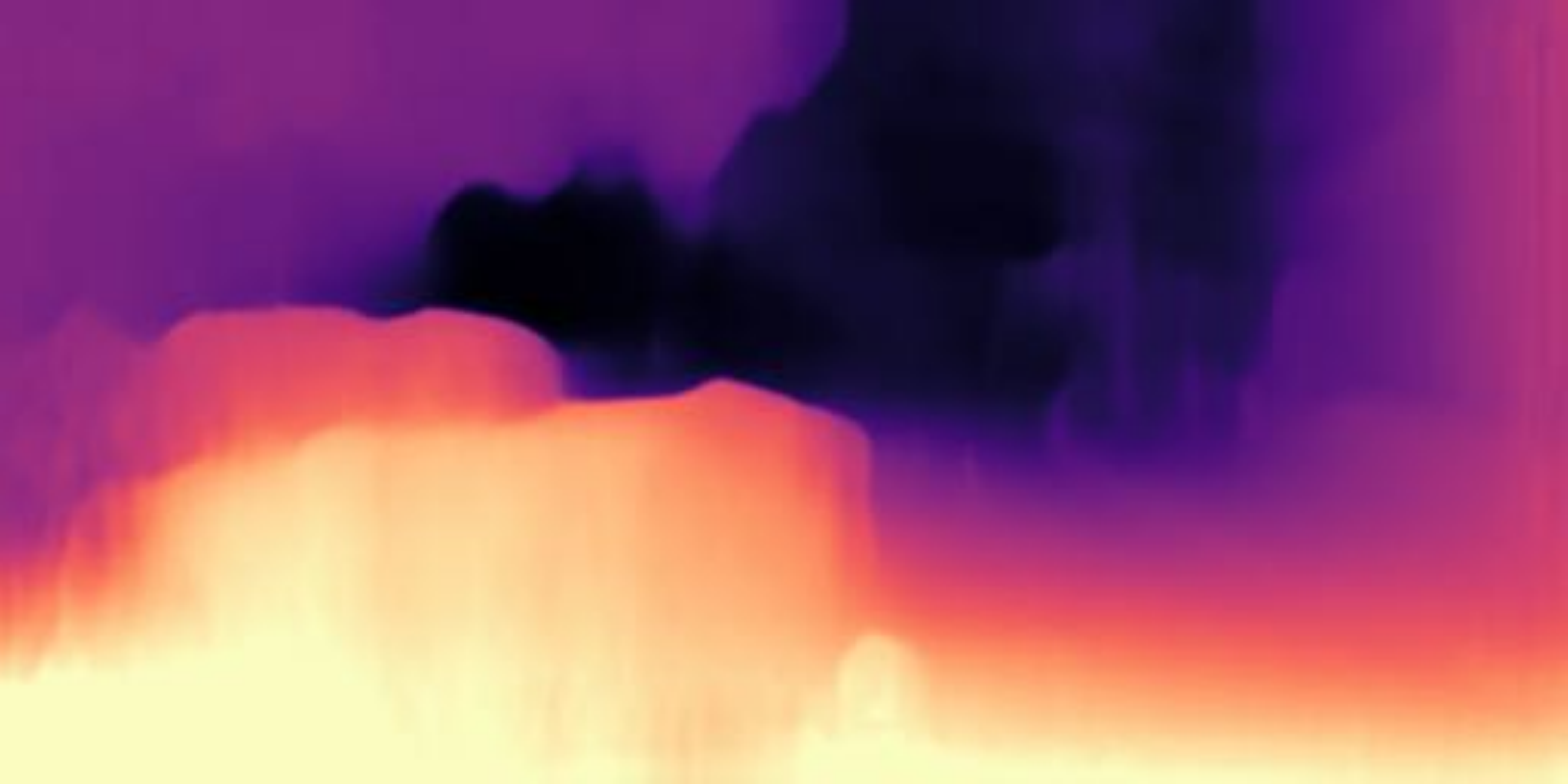}\qquad\qquad\quad & 
\includegraphics[width=\iw,height=\ih]{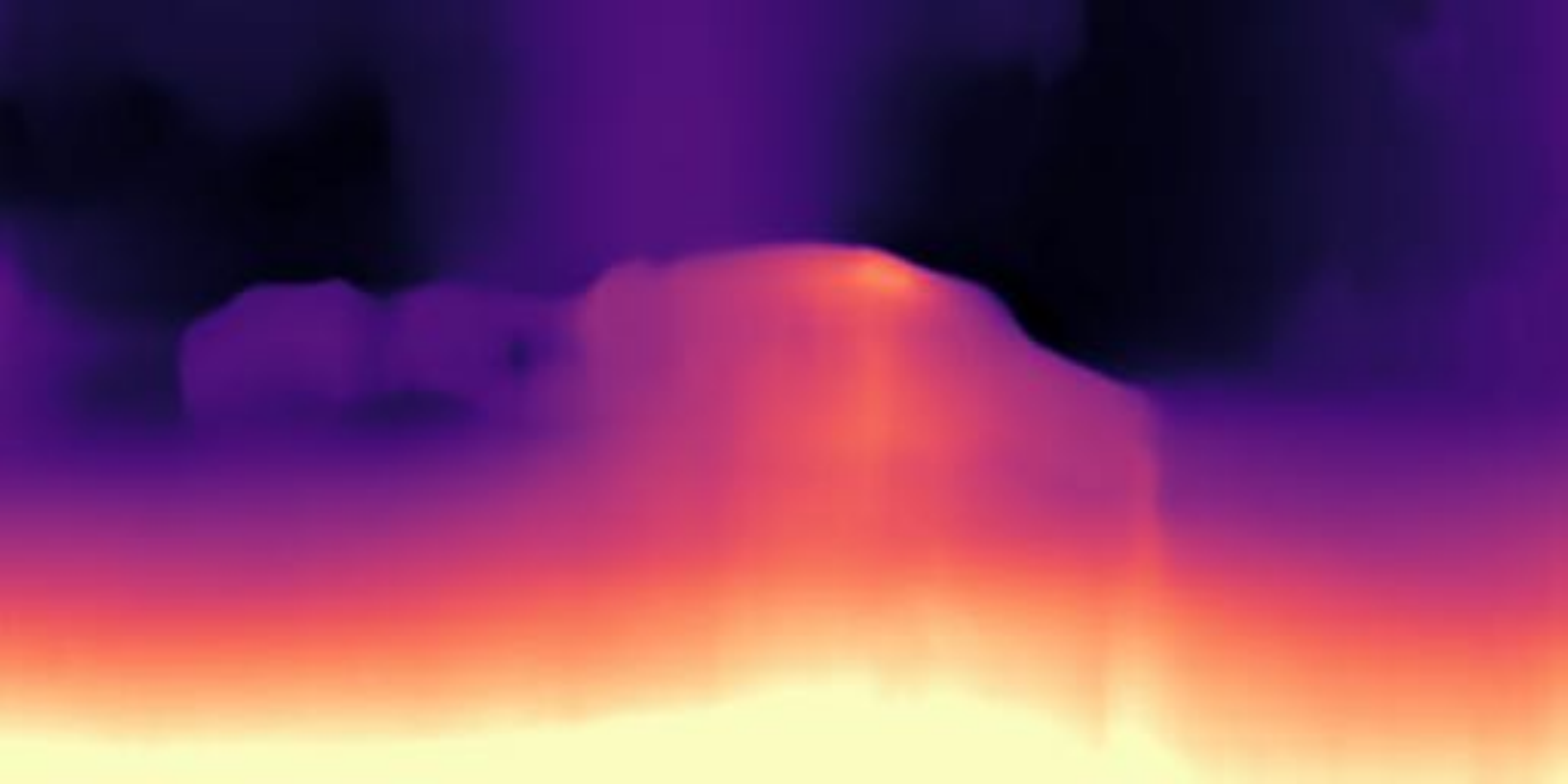}\qquad\qquad\quad &
\includegraphics[width=\iw,height=\ih]{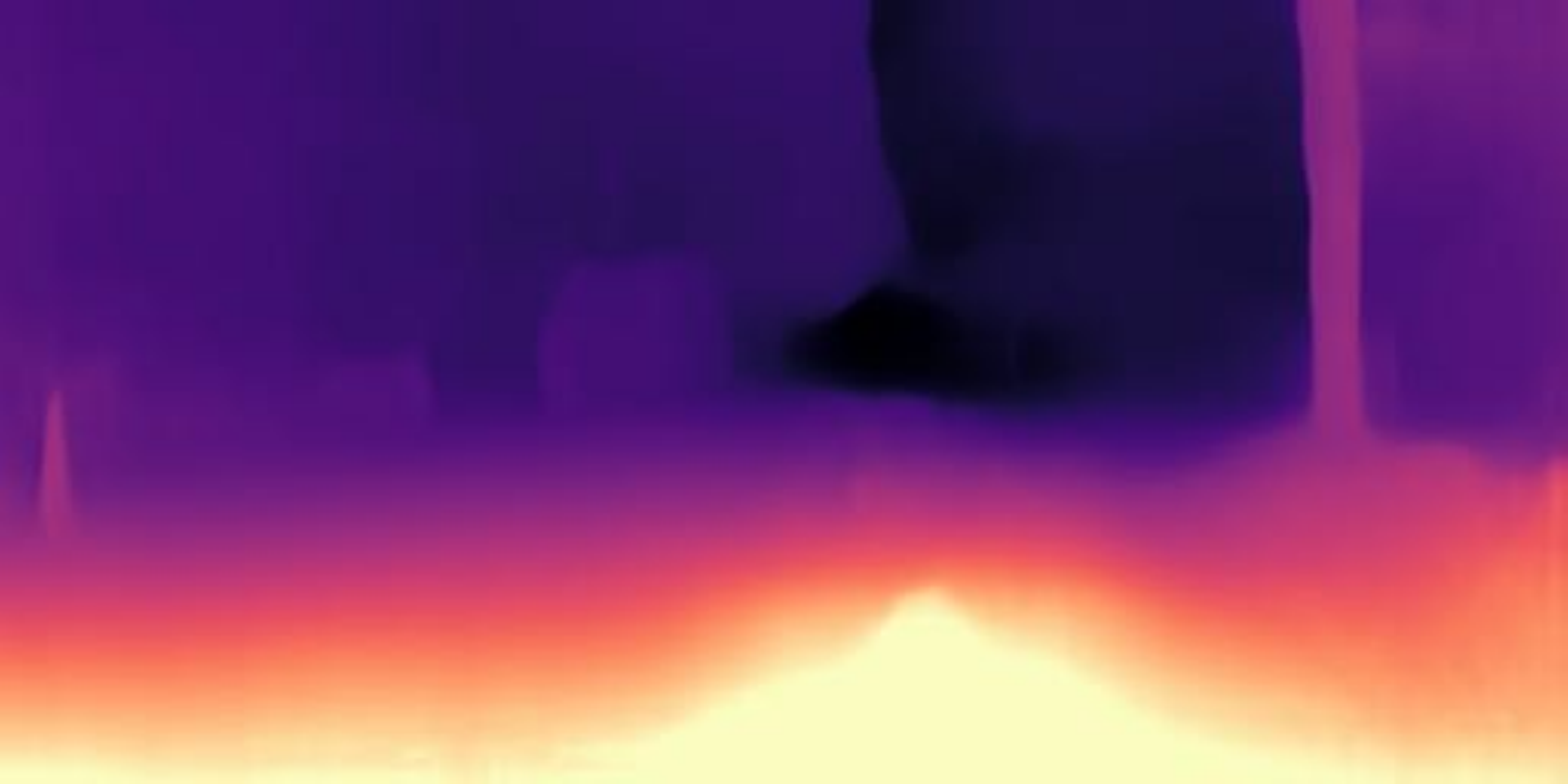}\\
\vspace{10mm} \\
\rotatebox[origin=c]{90}{\fontsize{\textw}{\texth}\selectfont PackNet-SfM\hspace{-310mm}}\hspace{15mm}
\includegraphics[width=\iw,height=\ih]{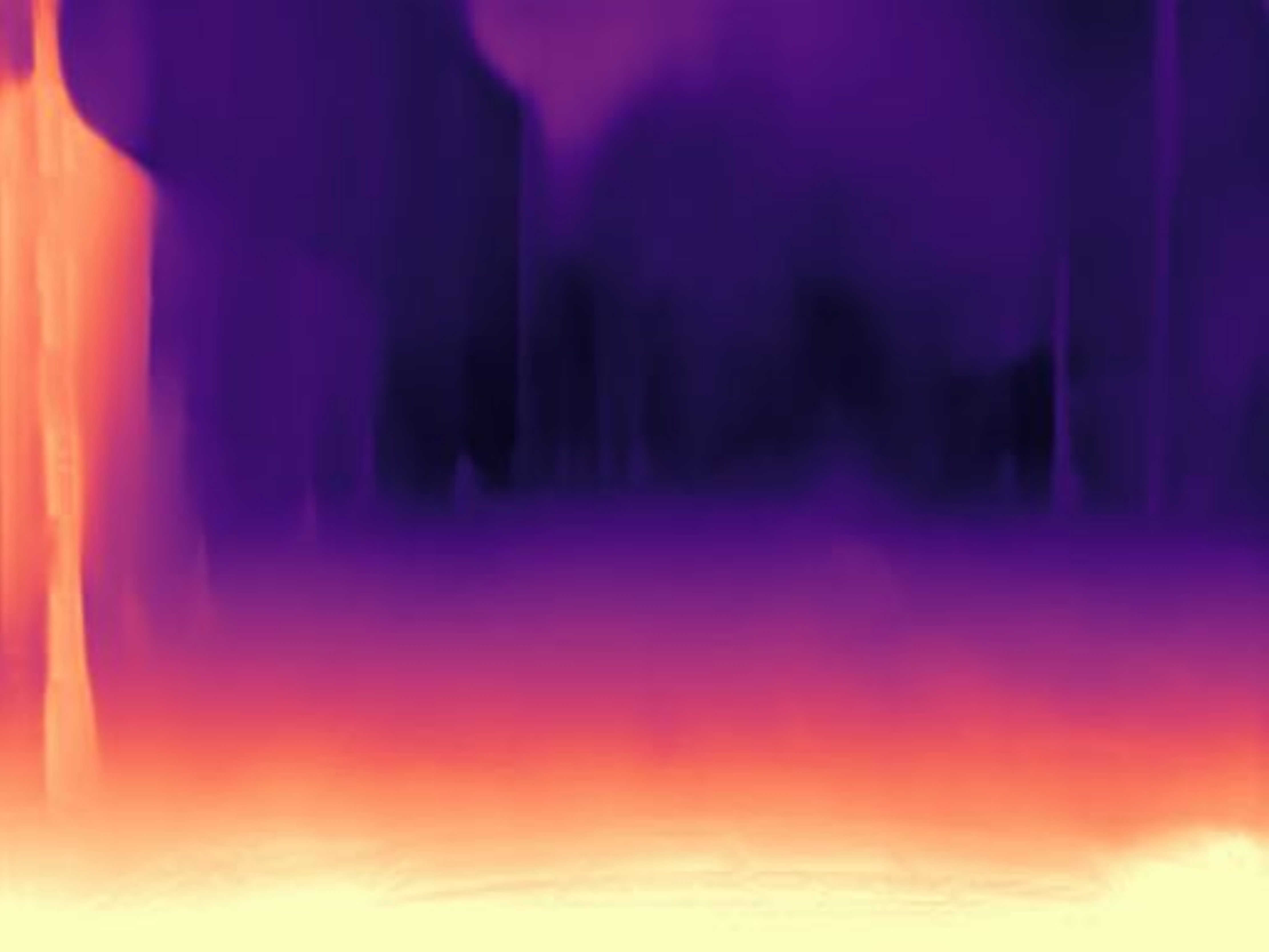}\qquad\qquad\quad &
\includegraphics[width=\iw,height=\ih]{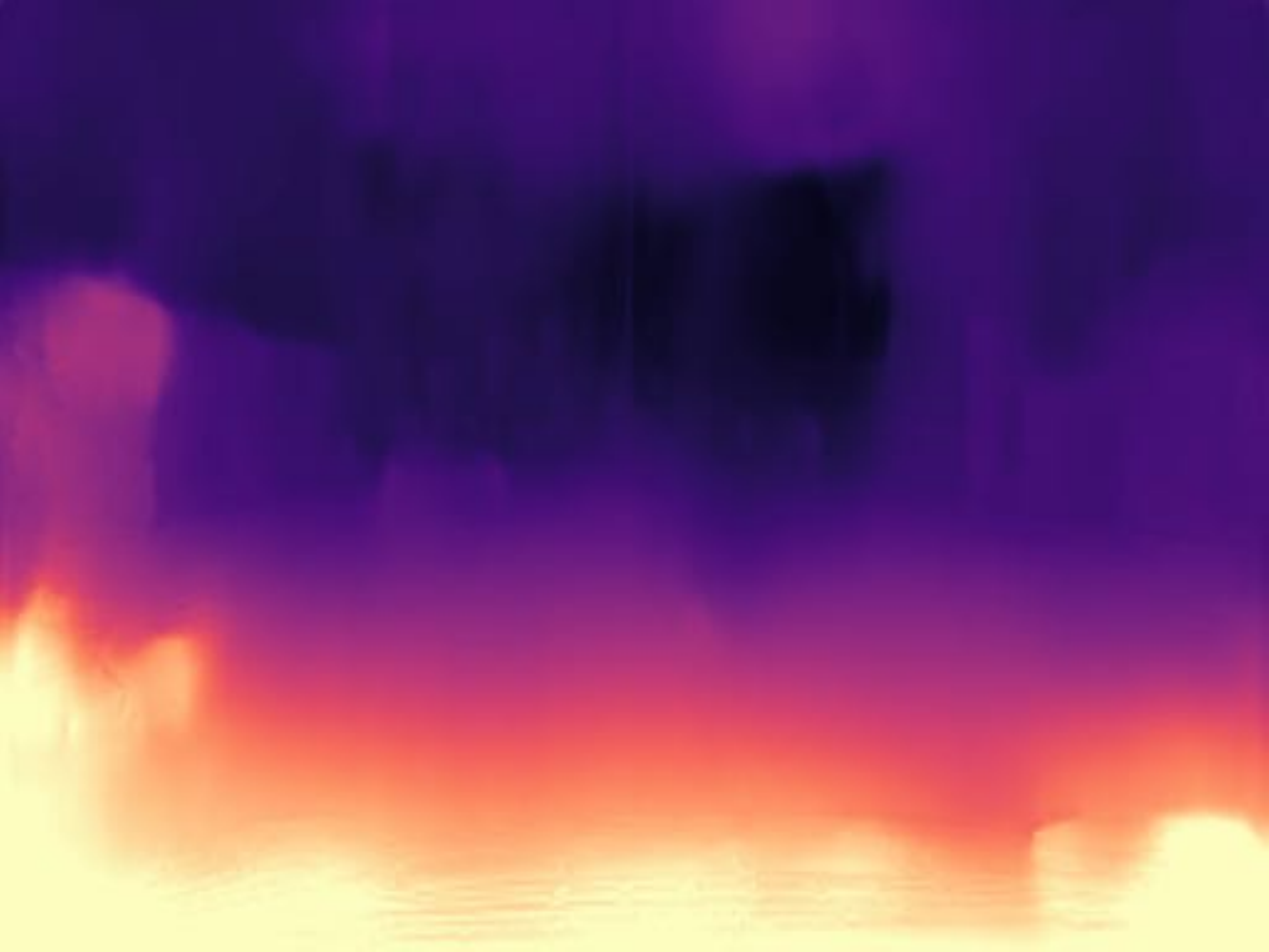}\qquad\qquad\quad &
\includegraphics[width=\iw,height=\ih]{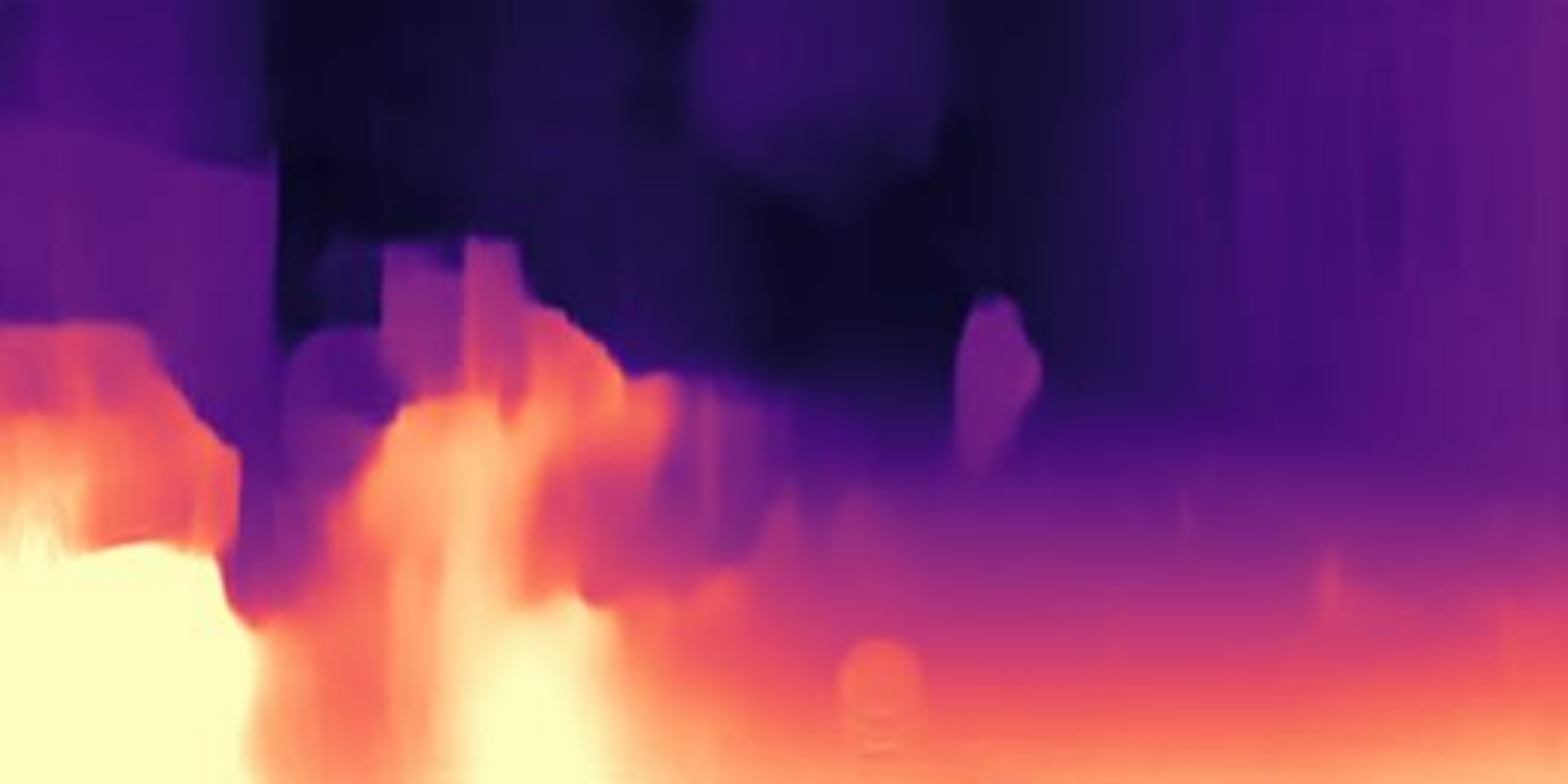}\qquad\qquad\quad &
\includegraphics[width=\iw,height=\ih]{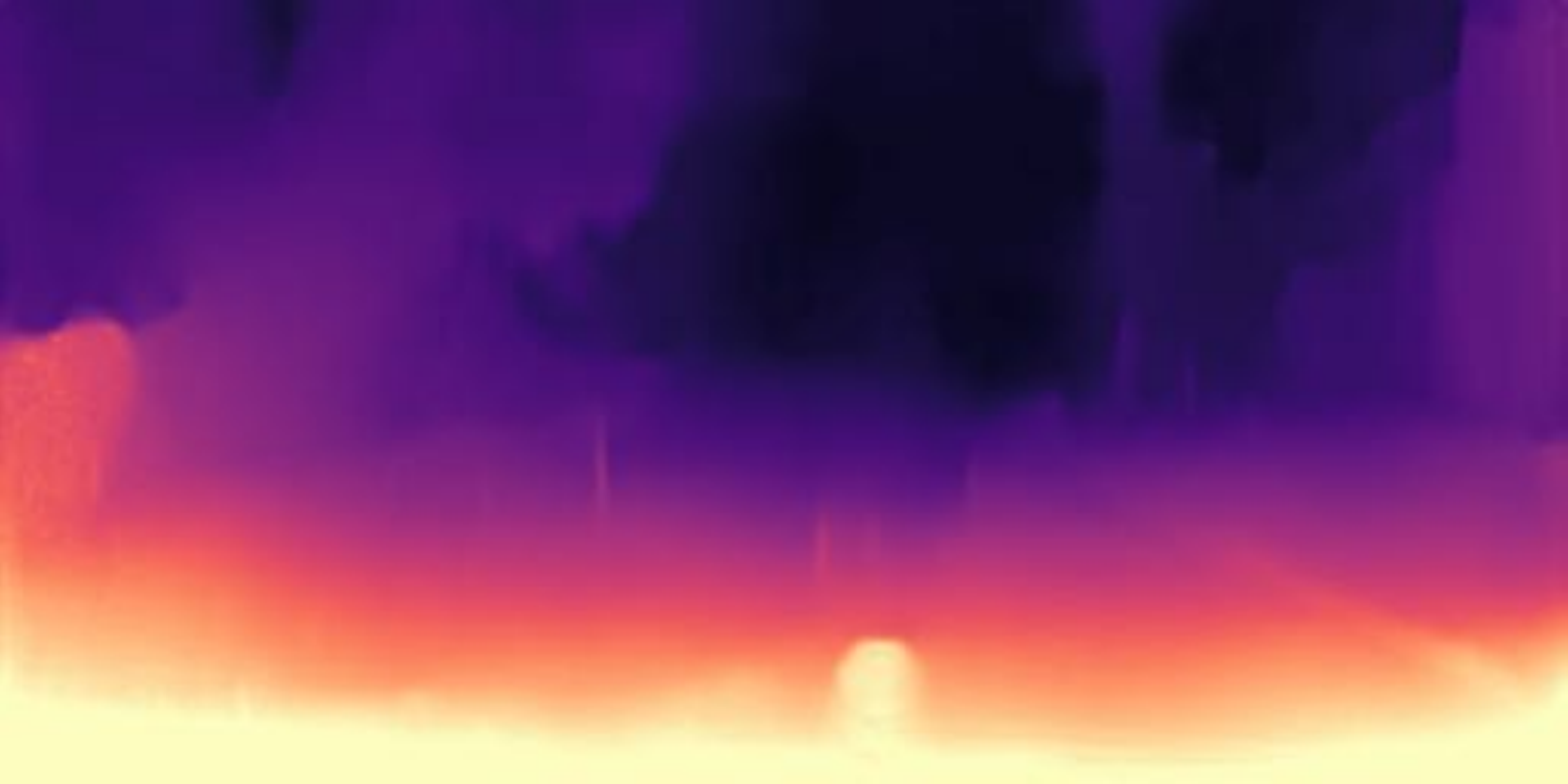}\qquad\qquad\quad &
\includegraphics[width=\iw,height=\ih]{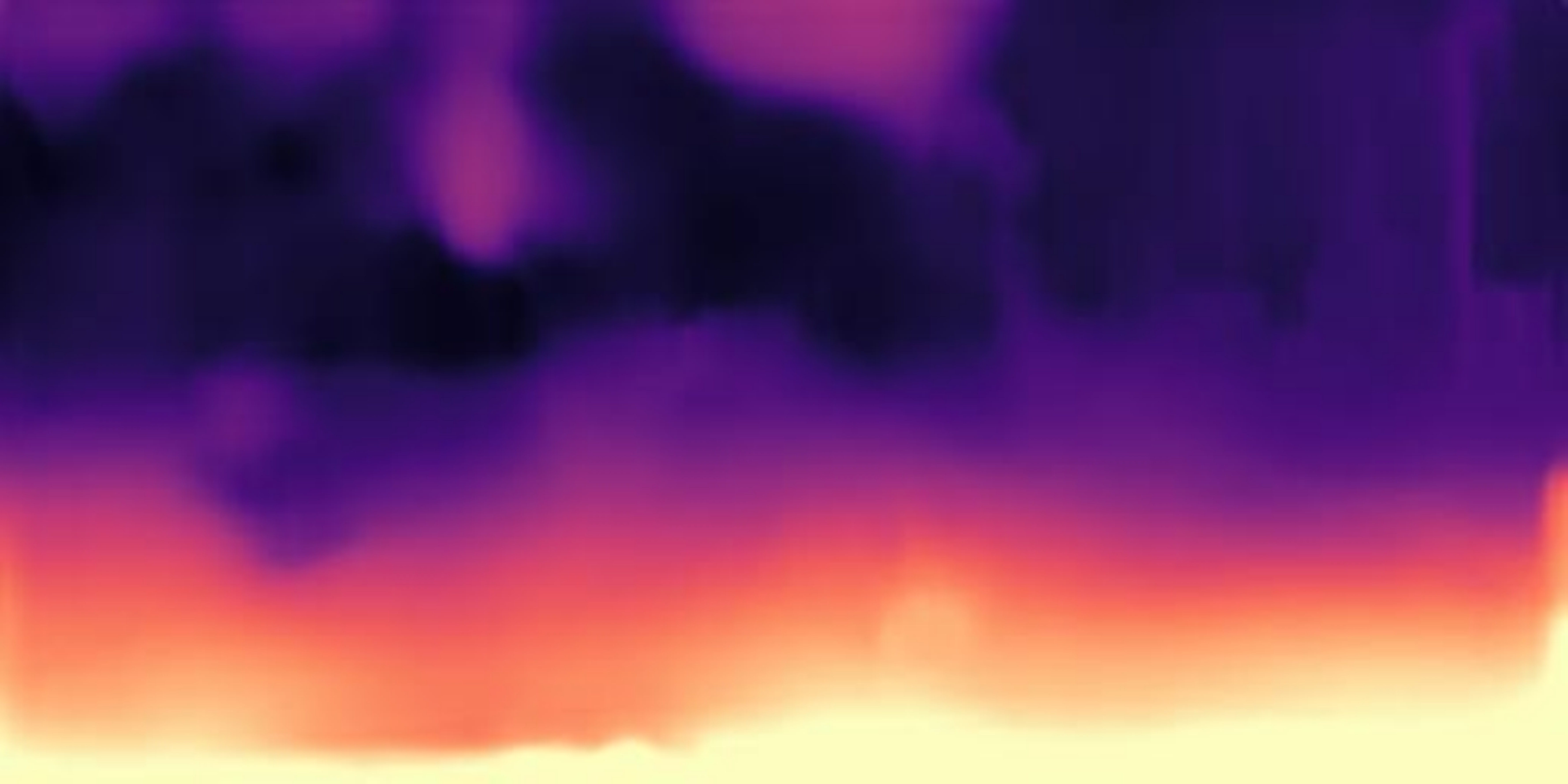}\qquad\qquad\quad &
\includegraphics[width=\iw,height=\ih]{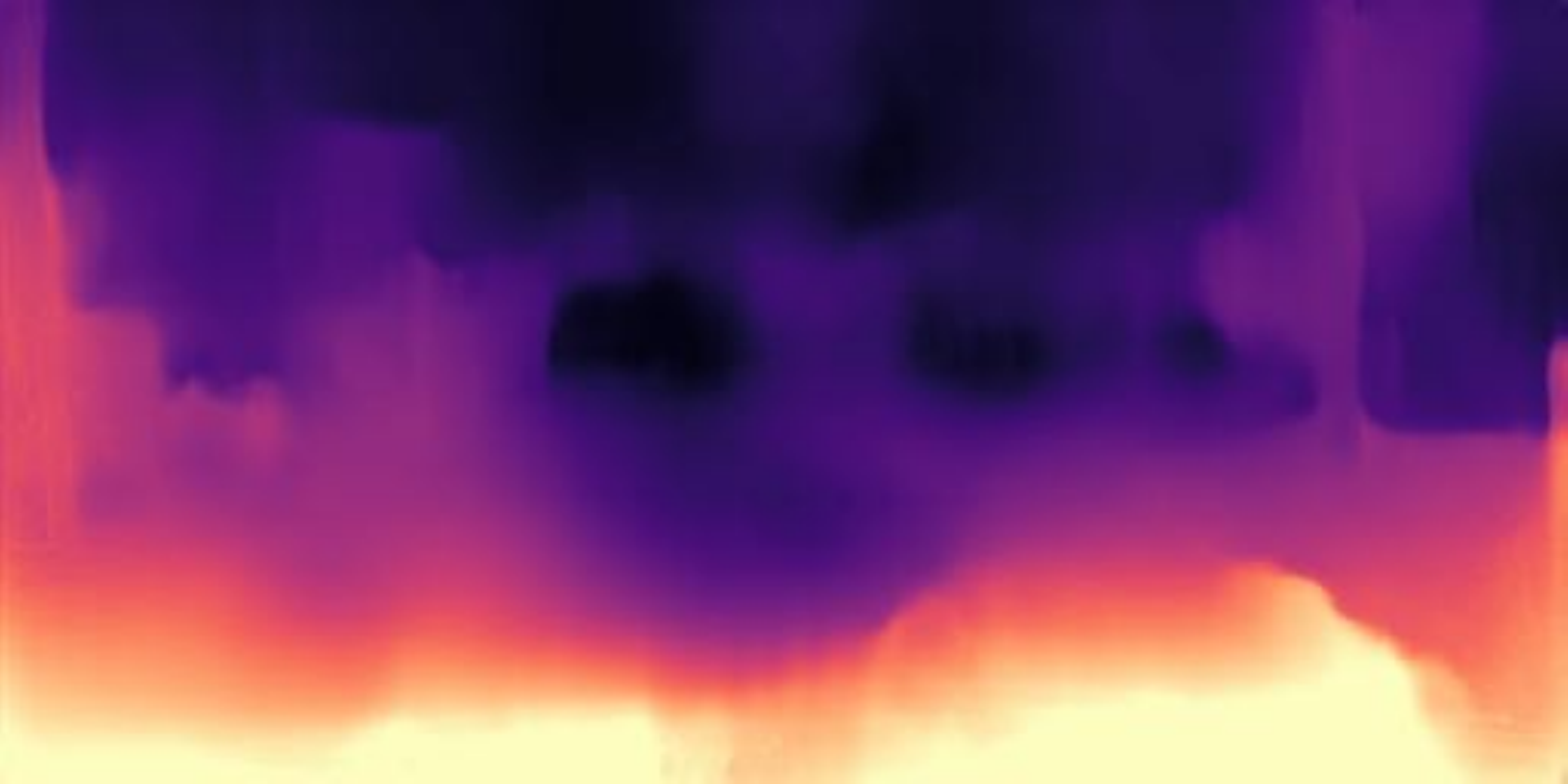}\\
\vspace{10mm} \\
\rotatebox[origin=c]{90}{\fontsize{\textw}{\texth}\selectfont R-MSFM6\hspace{-300mm}}\hspace{15mm}
\includegraphics[width=\iw,height=\ih]{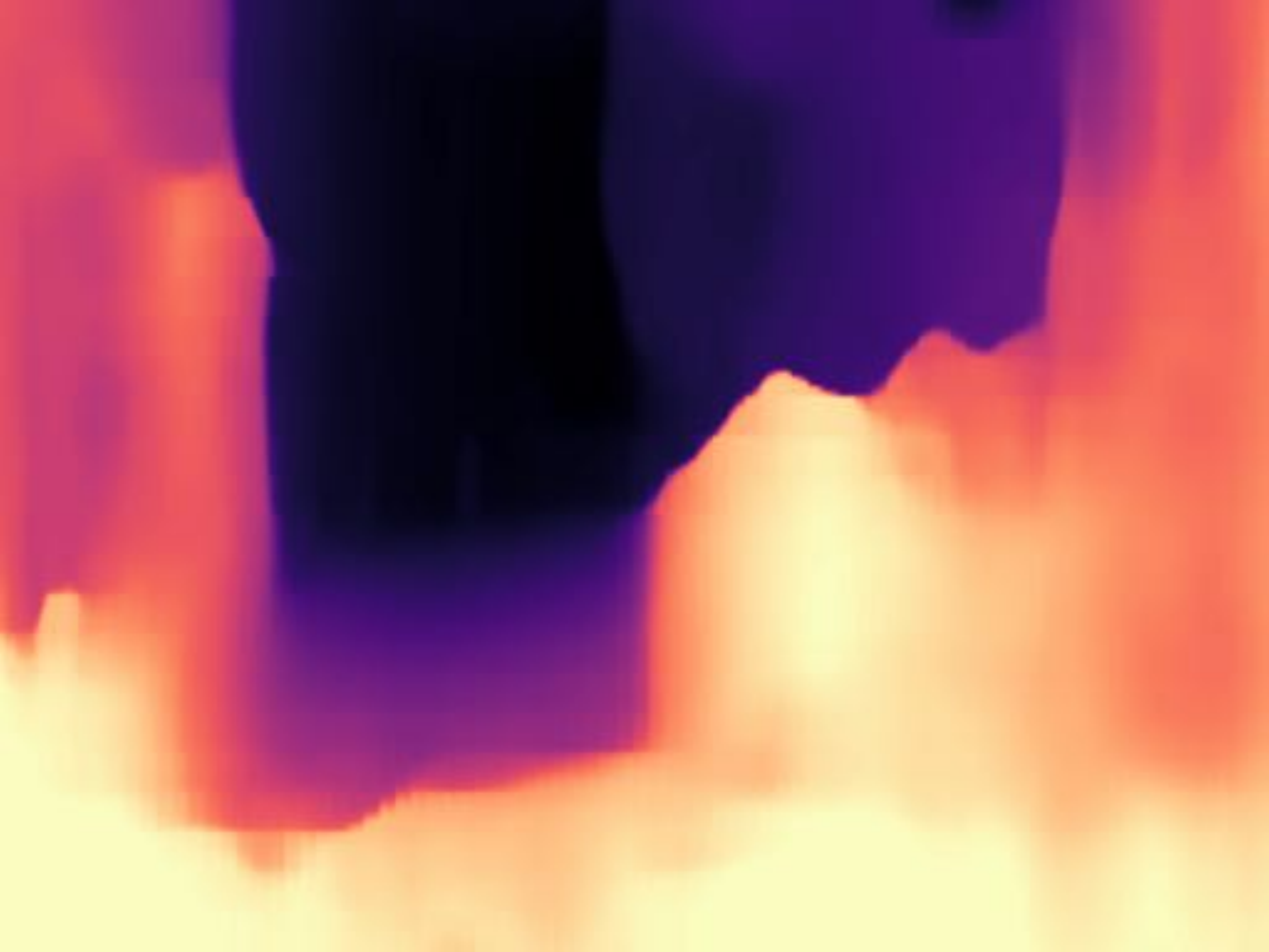}\qquad\qquad\quad &
\includegraphics[width=\iw,height=\ih]{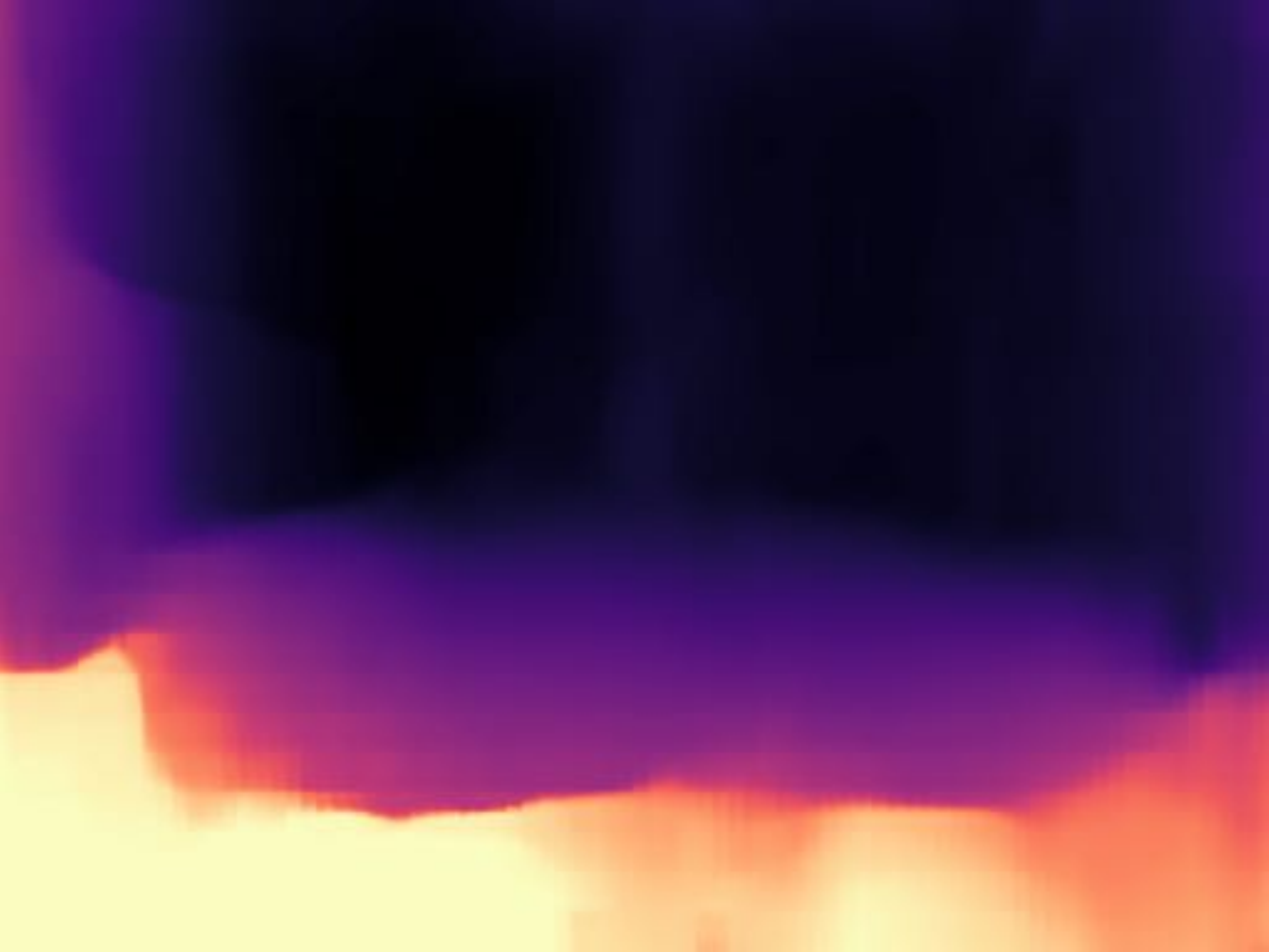}\qquad\qquad\quad &
\includegraphics[width=\iw,height=\ih]{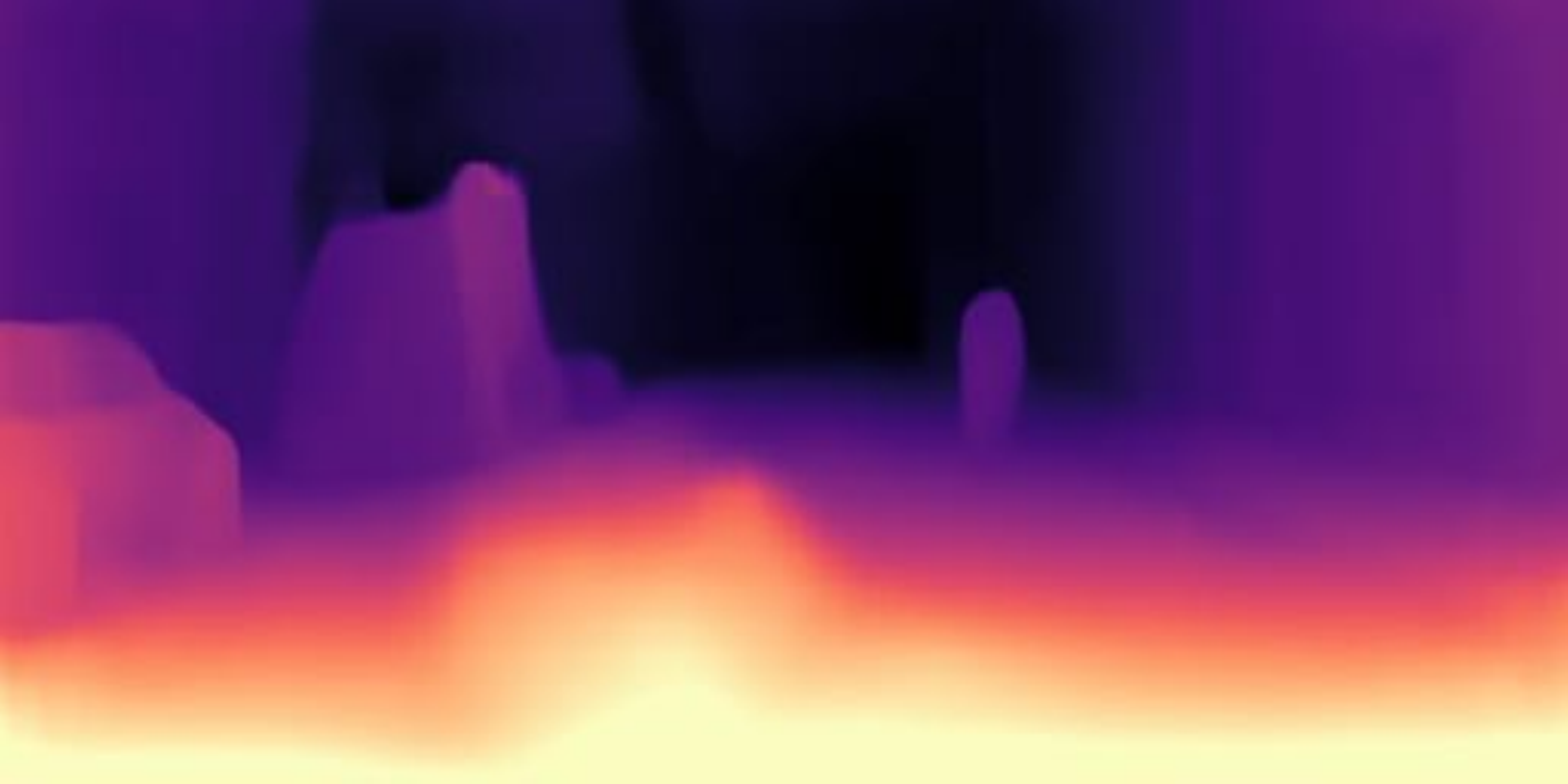}\qquad\qquad\quad &
\includegraphics[width=\iw,height=\ih]{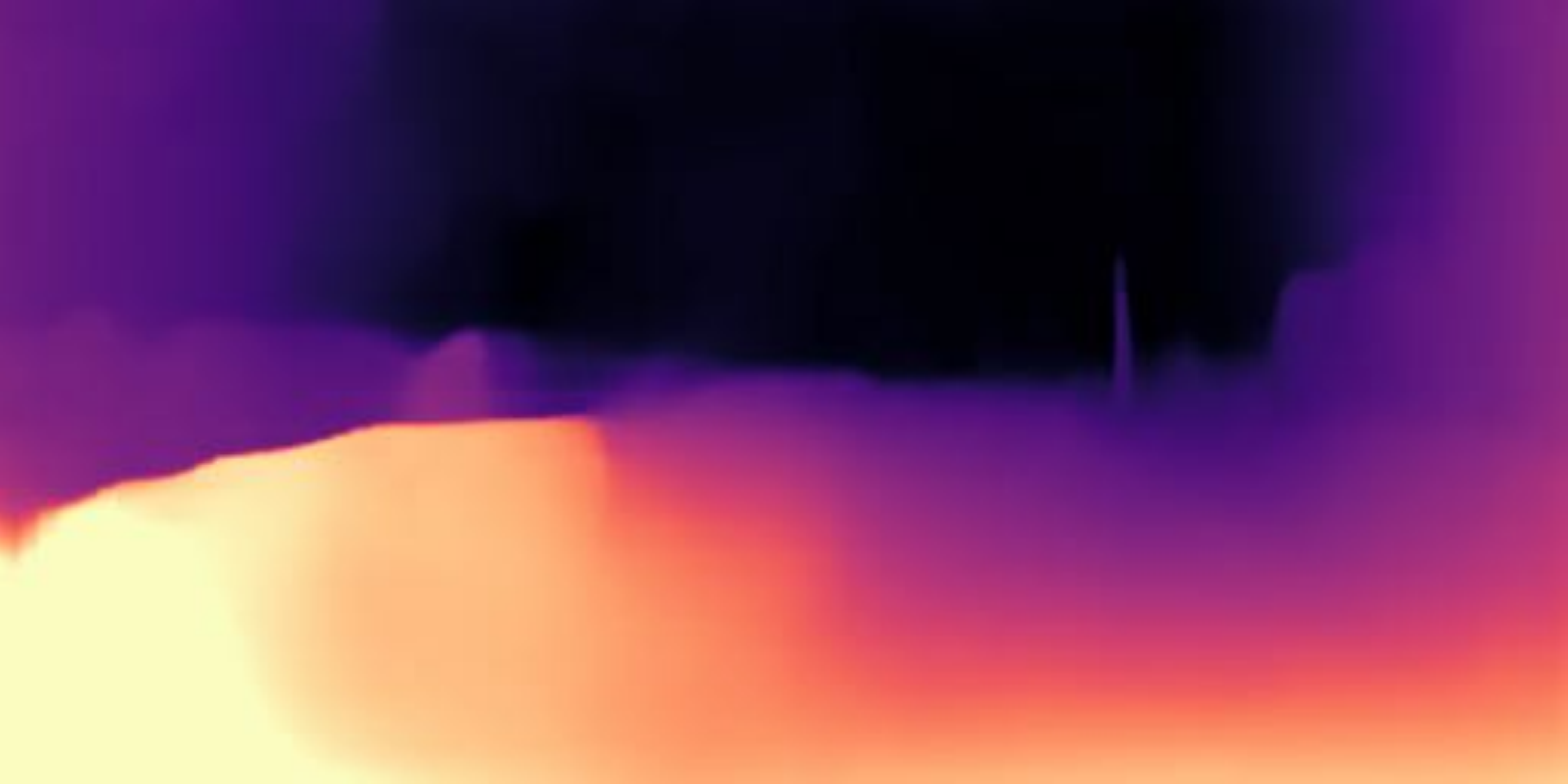}\qquad\qquad\quad &
\includegraphics[width=\iw,height=\ih]{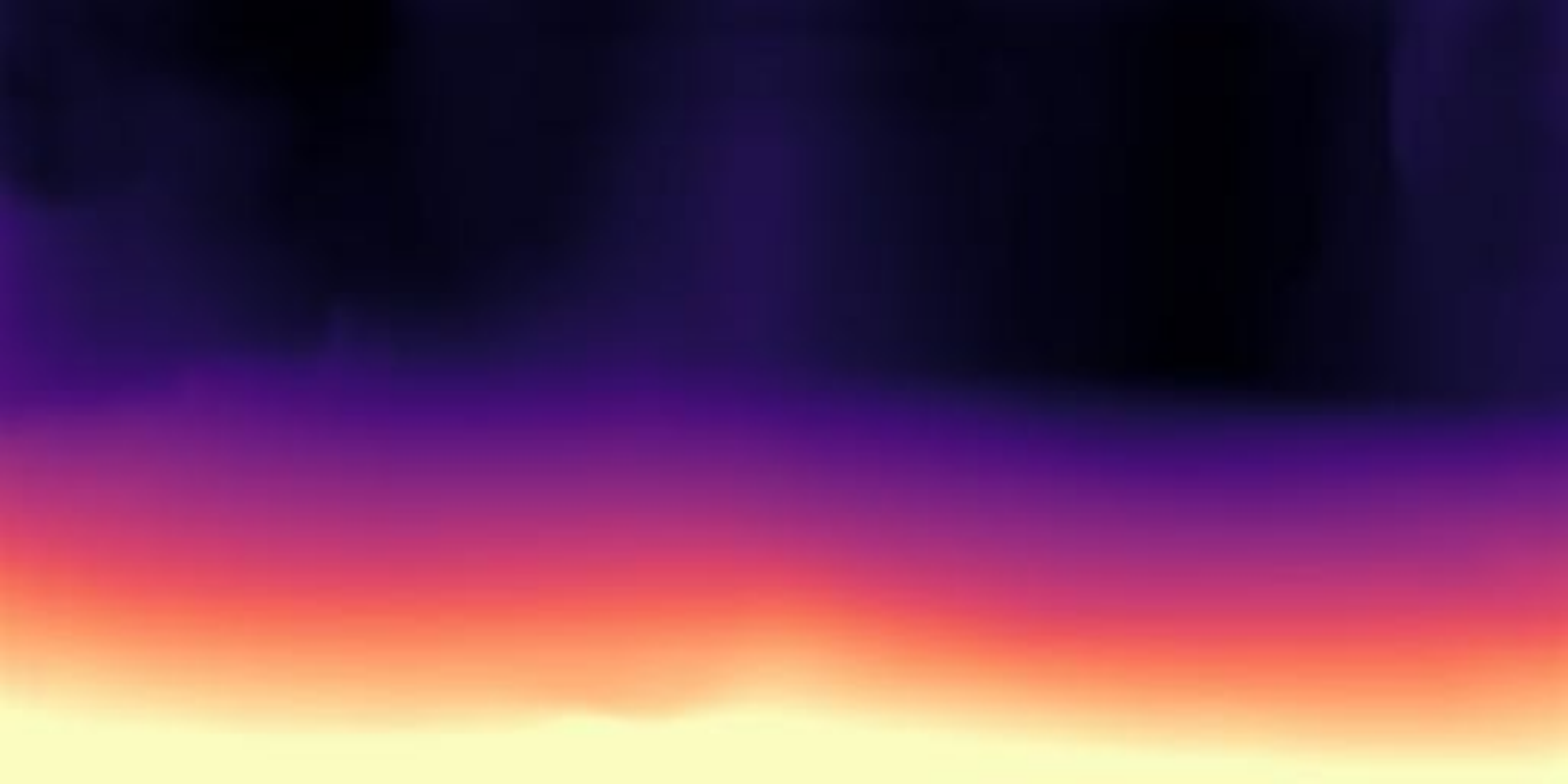}\qquad\qquad\quad &
\includegraphics[width=\iw,height=\ih]{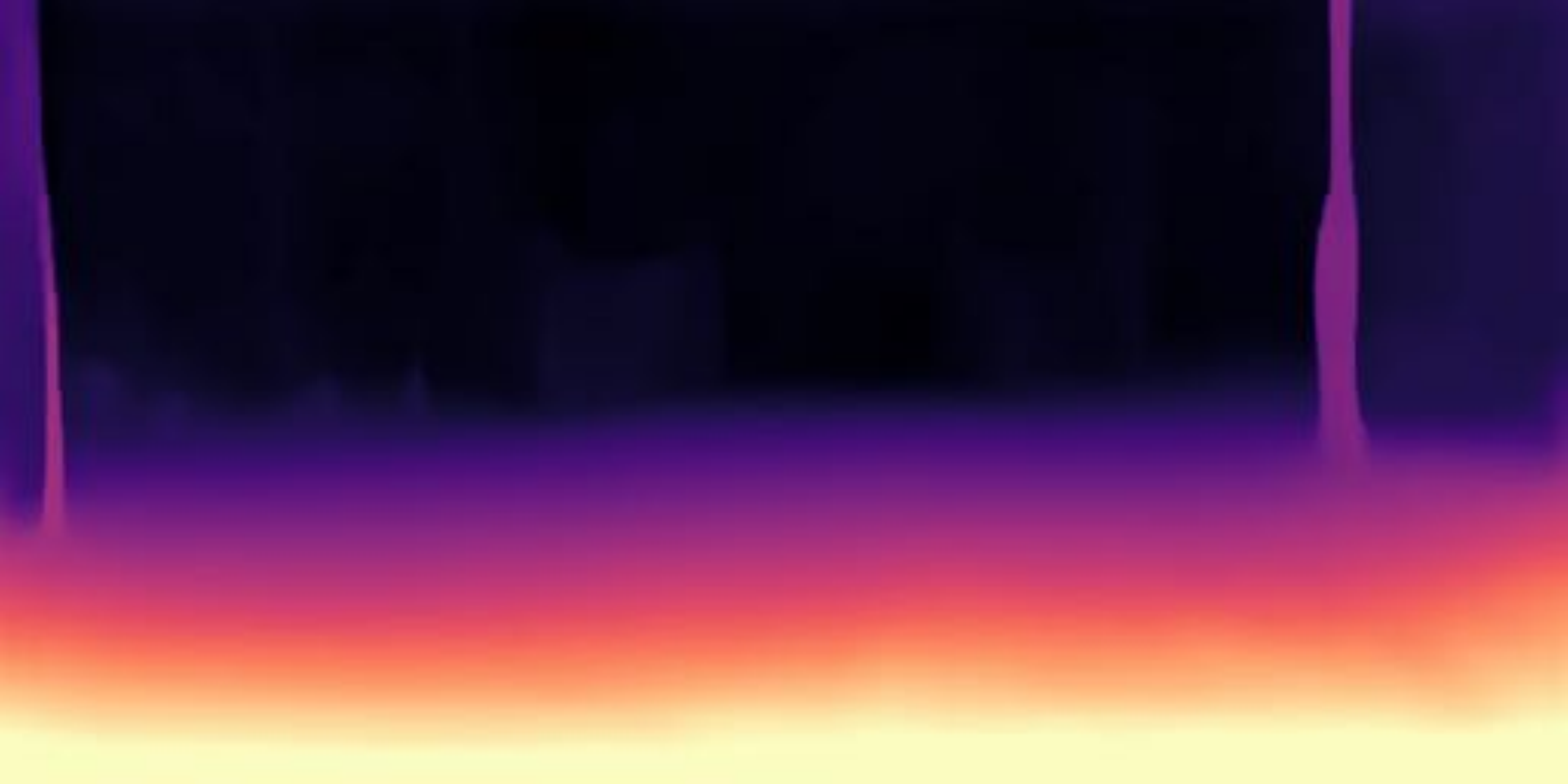}\\
\vspace{10mm} \\
\rotatebox[origin=c]{90}{\fontsize{\textw}{\texth}\selectfont MF-ConvNeXt\hspace{-290mm}}\hspace{15mm}
\includegraphics[width=\iw,height=\ih]{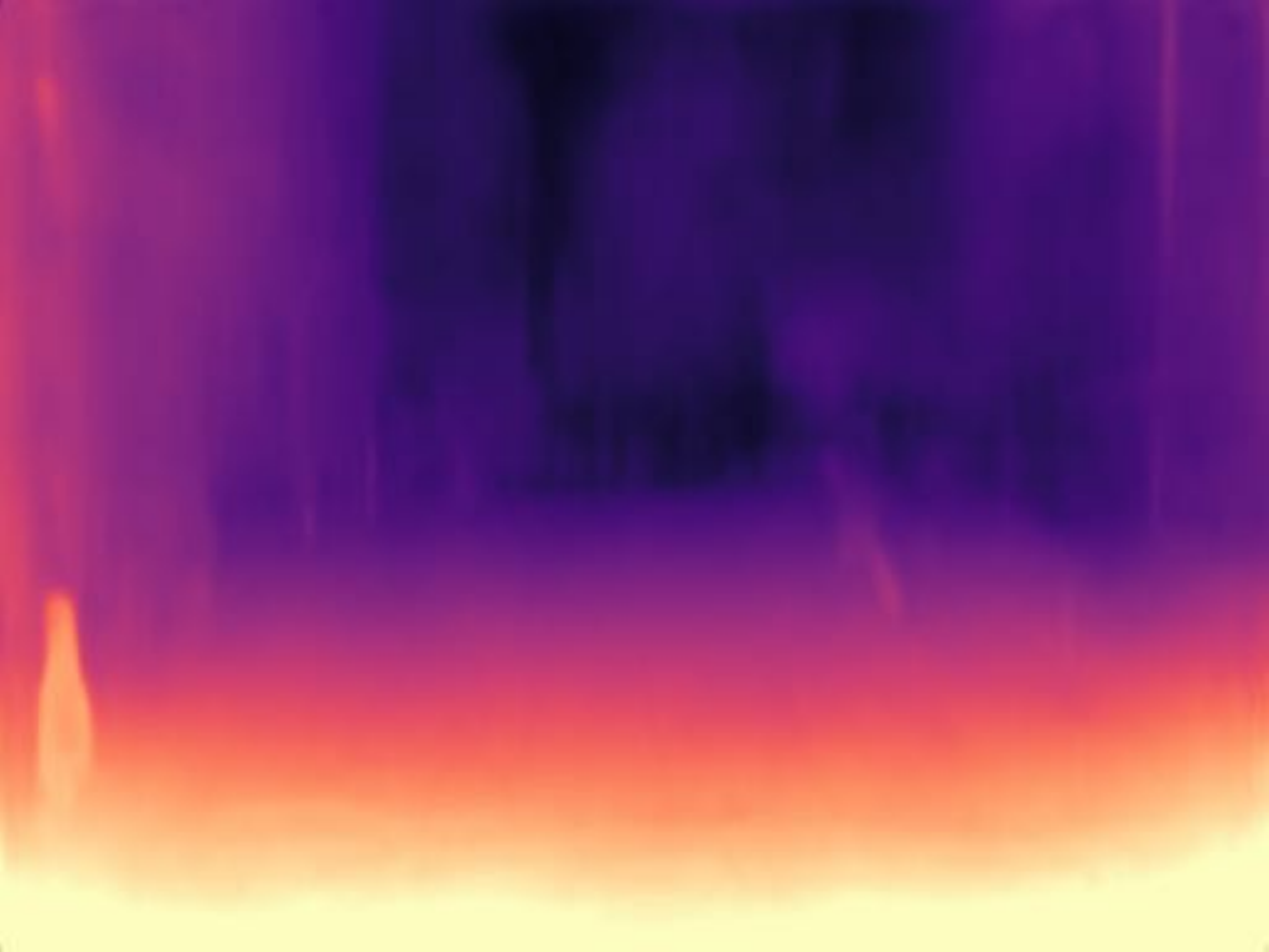}\qquad\qquad\quad &
\includegraphics[width=\iw,height=\ih]{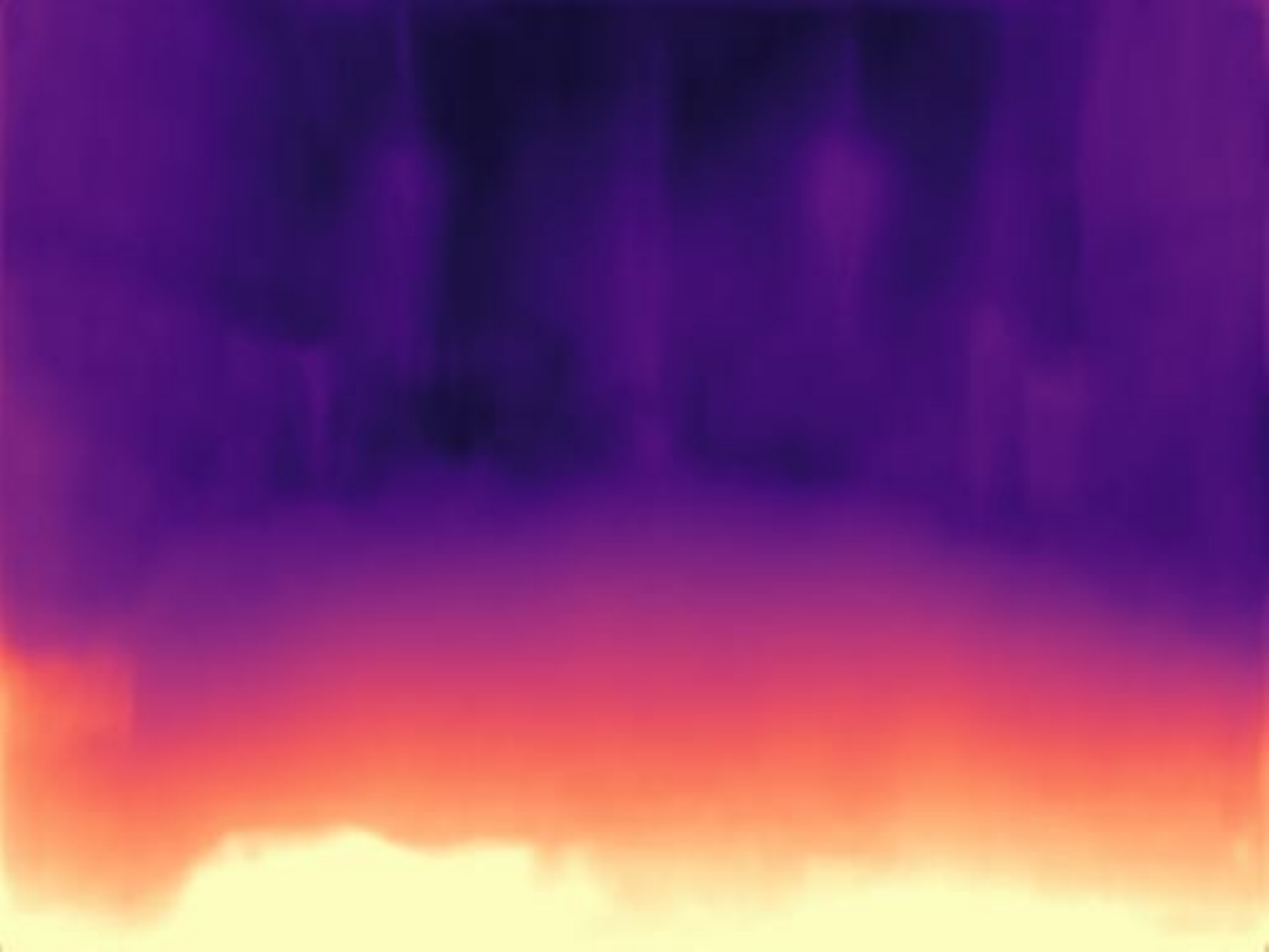}\qquad\qquad\quad &
\includegraphics[width=\iw,height=\ih]{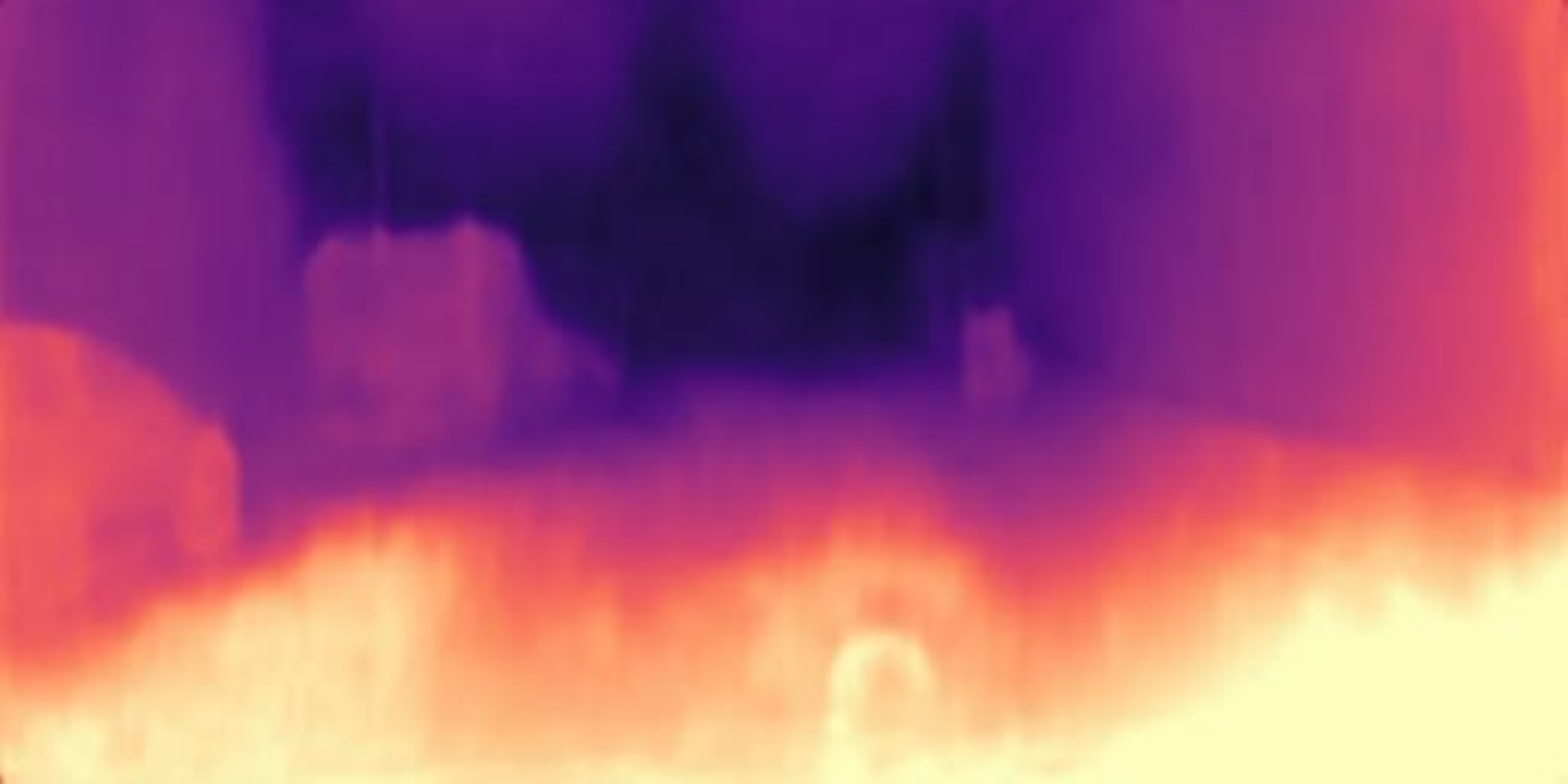}\qquad\qquad\quad &
\includegraphics[width=\iw,height=\ih]{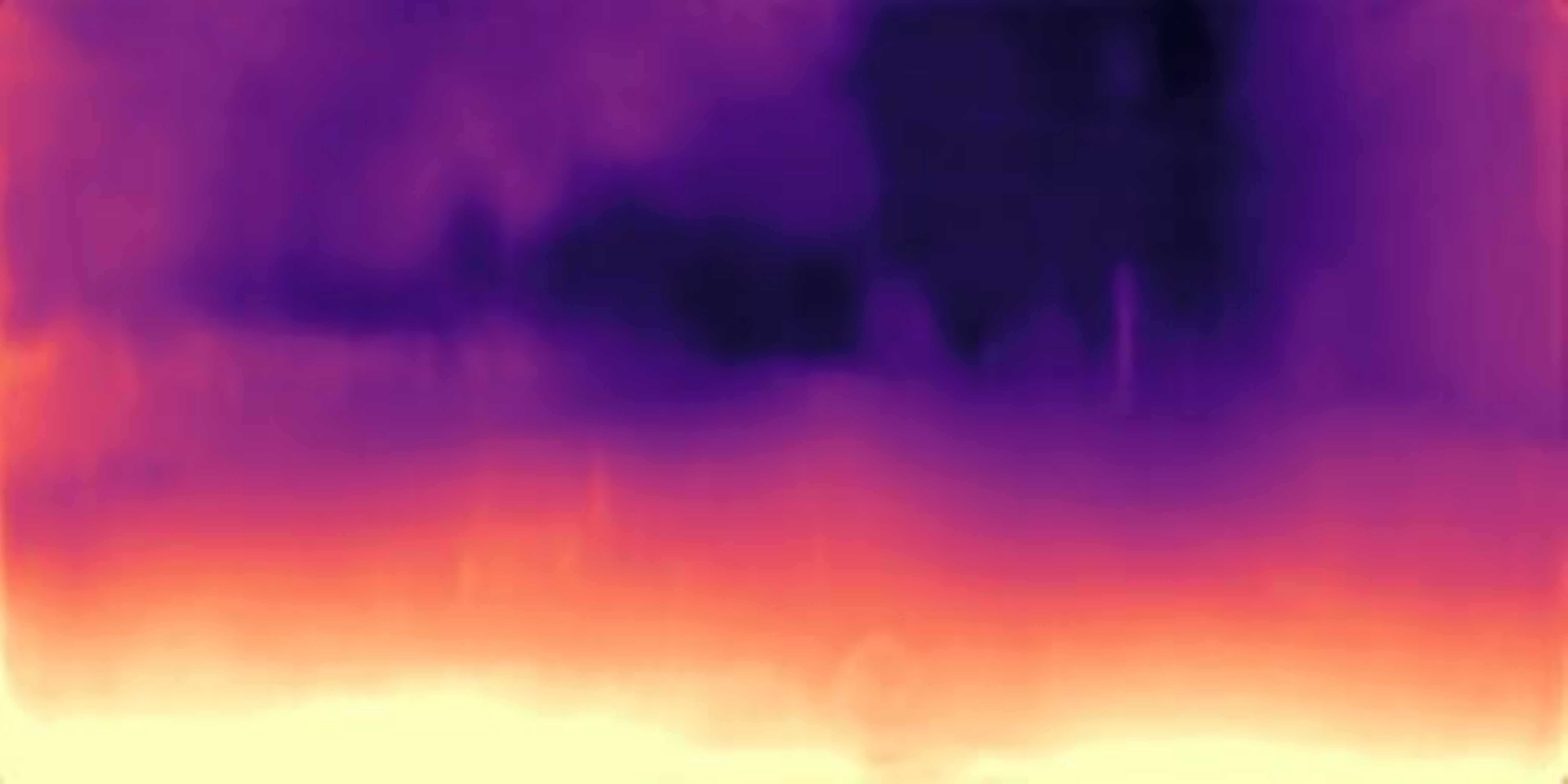}\qquad\qquad\quad &
\includegraphics[width=\iw,height=\ih]{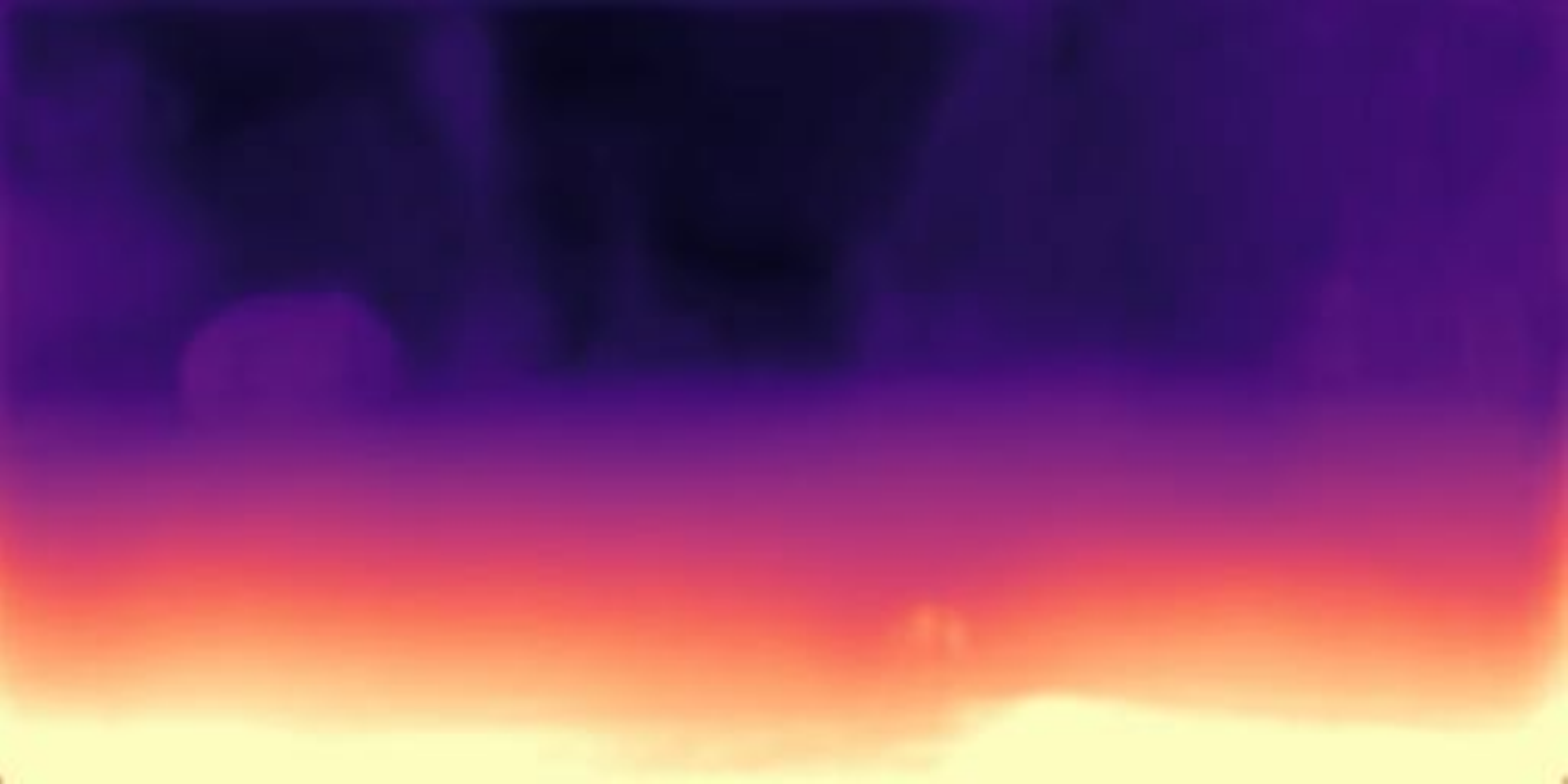}\qquad\qquad\quad &
\includegraphics[width=\iw,height=\ih]{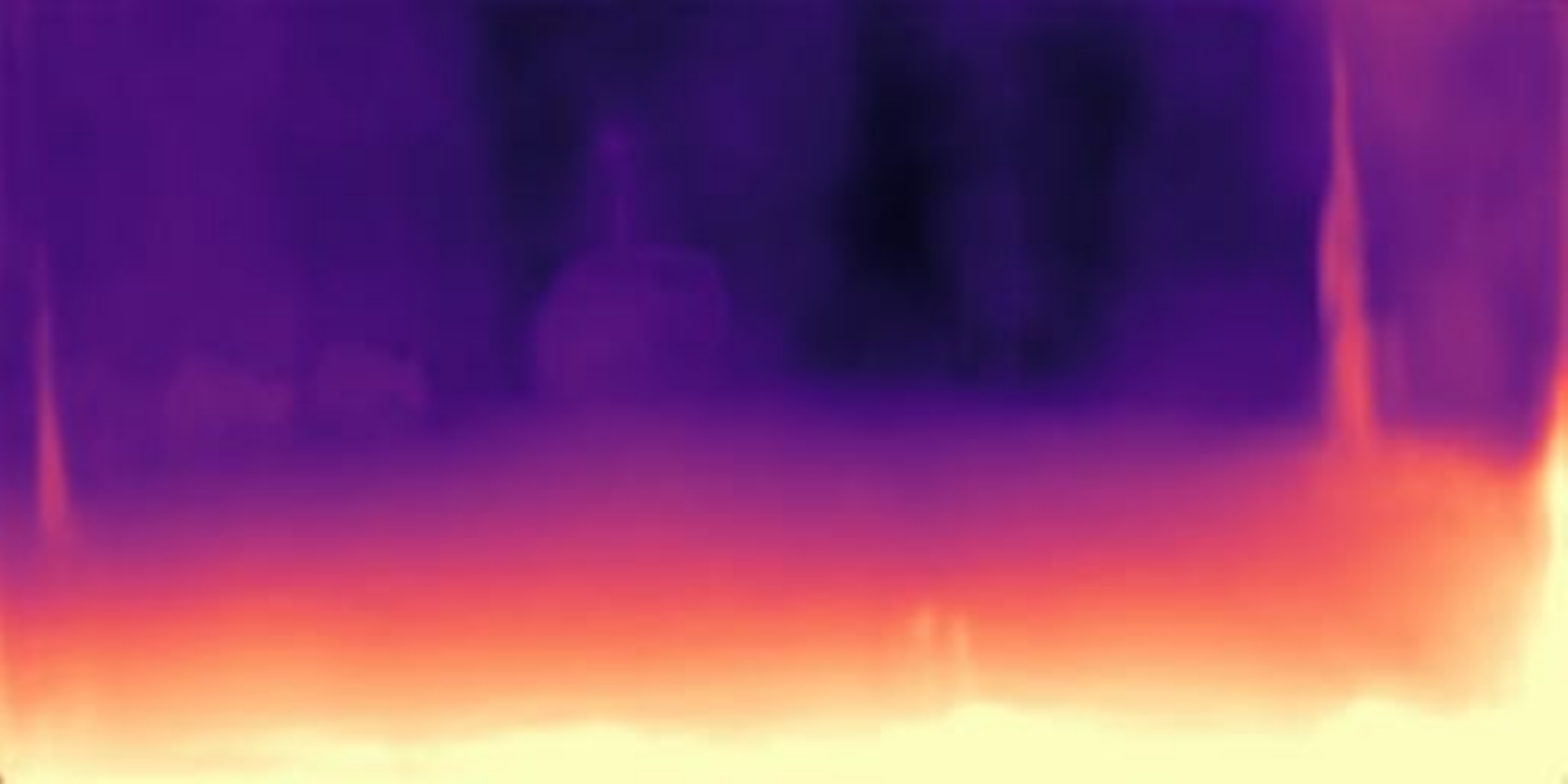}\\
\vspace{10mm} \\
\rotatebox[origin=c]{90}{\fontsize{\textw}{\texth}\selectfont MF-SLaK\hspace{-320mm}}\hspace{15mm}
\includegraphics[width=\iw,height=\ih]{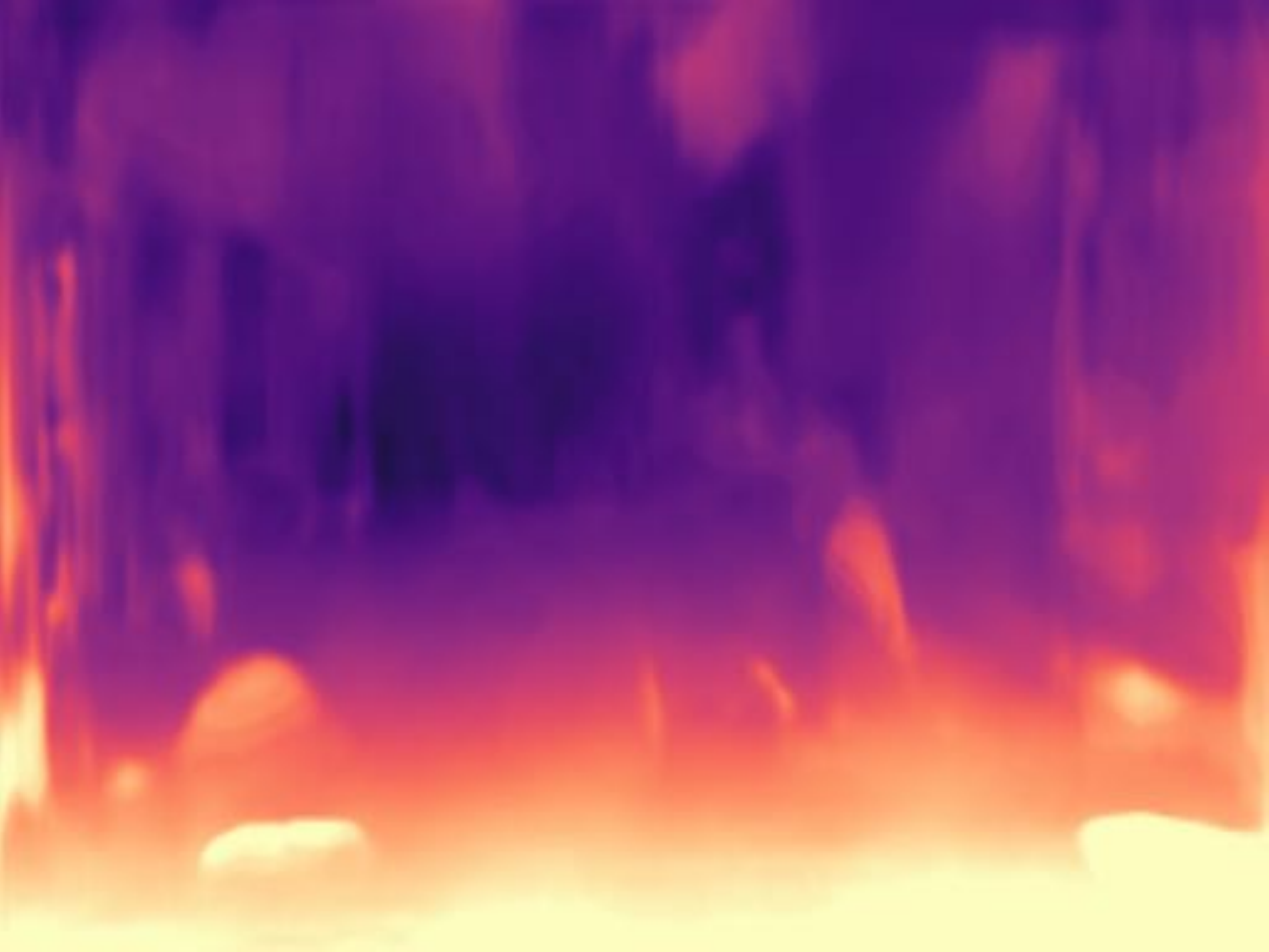}\qquad\qquad\quad &  
\includegraphics[width=\iw,height=\ih]{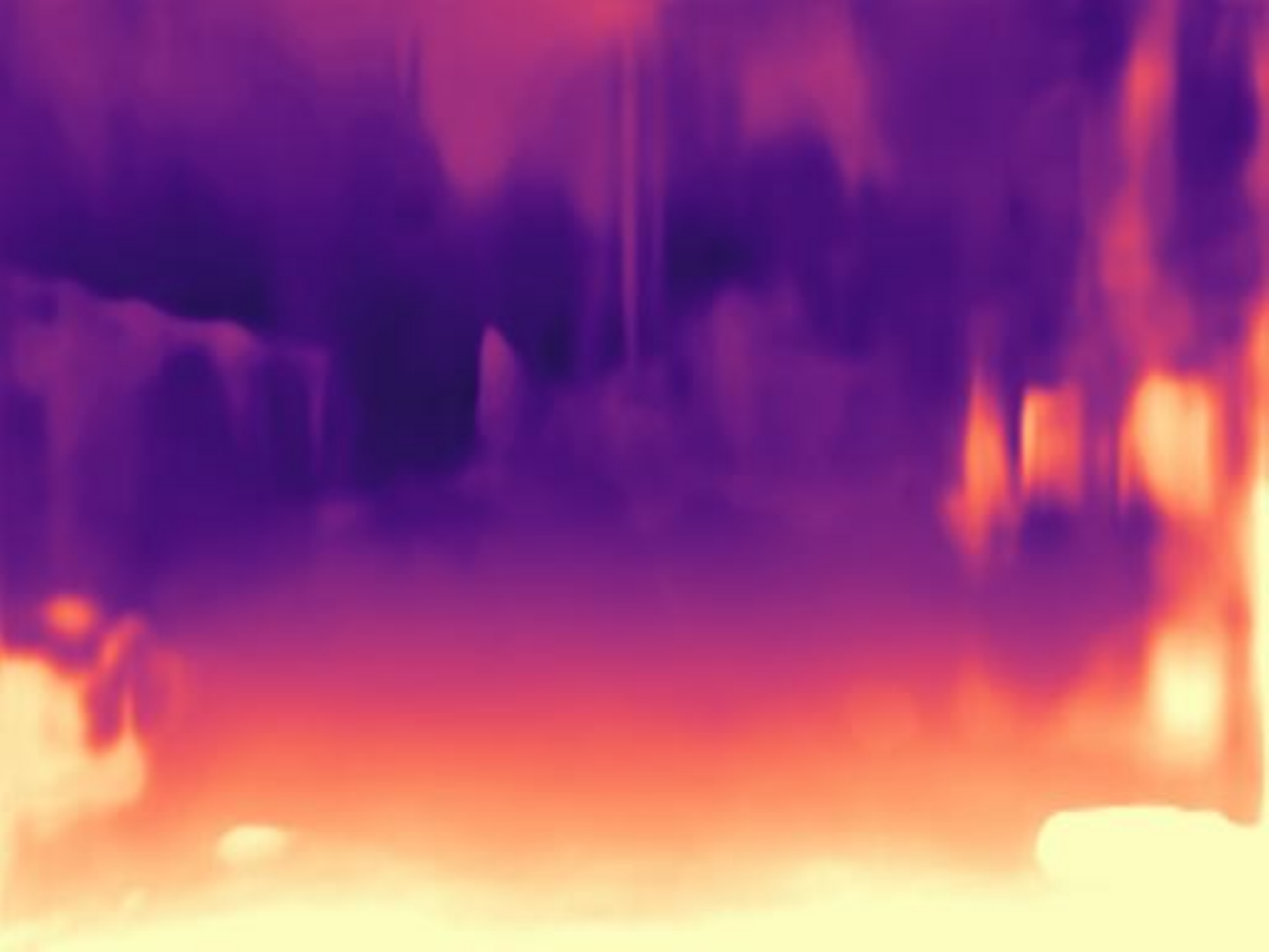}\qquad\qquad\quad &
\includegraphics[width=\iw,height=\ih]{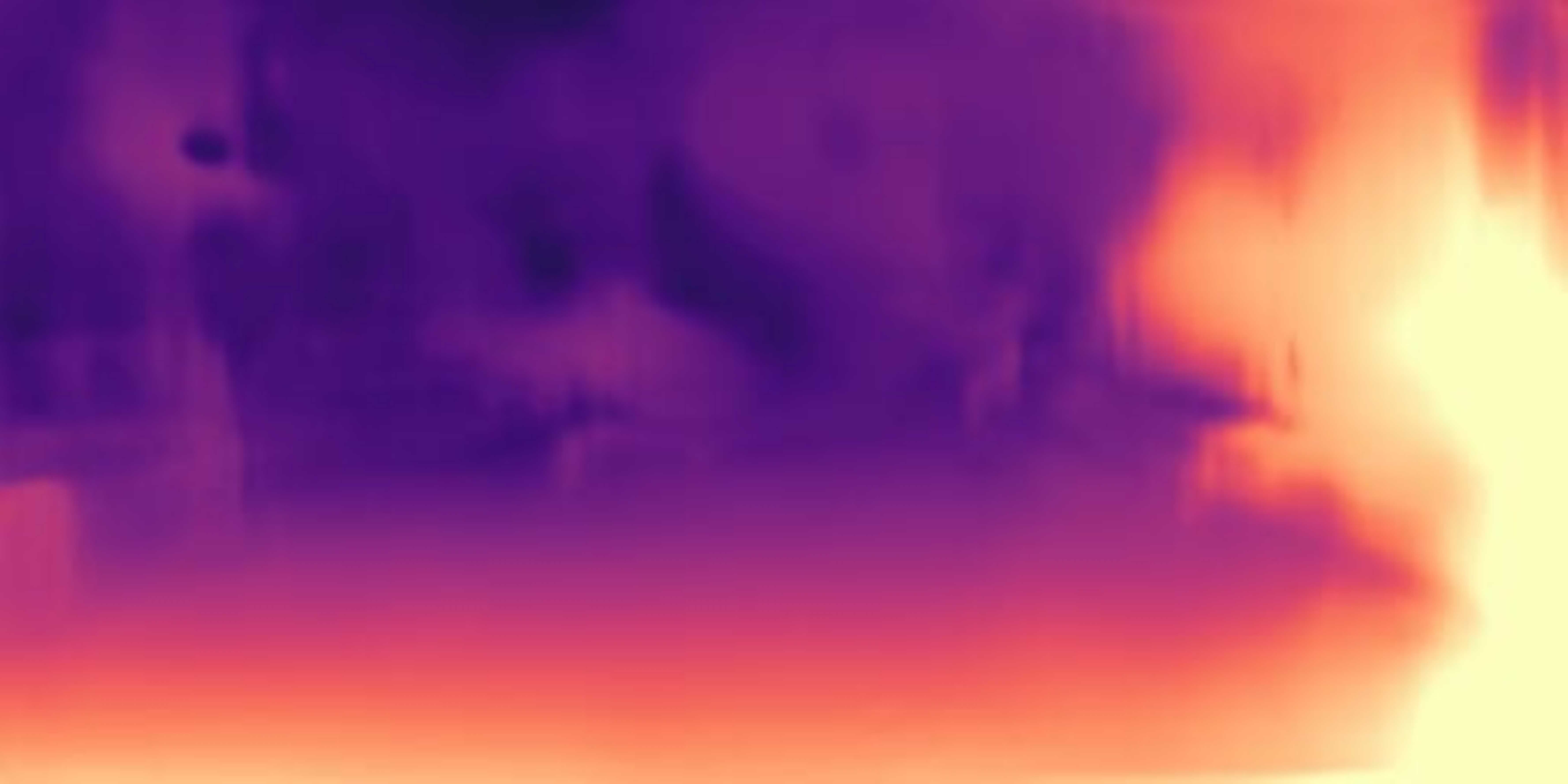}\qquad\qquad\quad &  
\includegraphics[width=\iw,height=\ih]{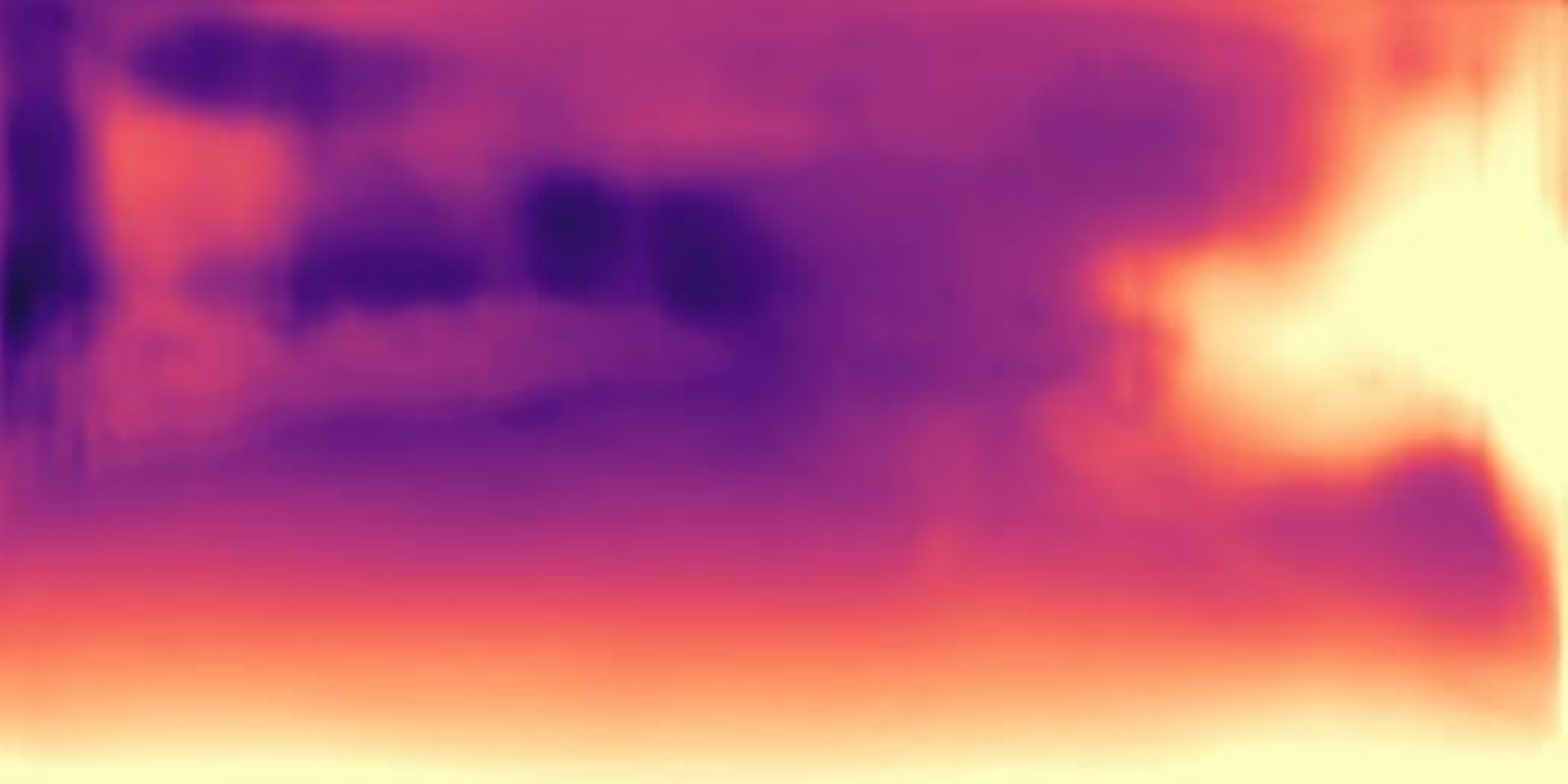}\qquad\qquad\quad & 
\includegraphics[width=\iw,height=\ih]{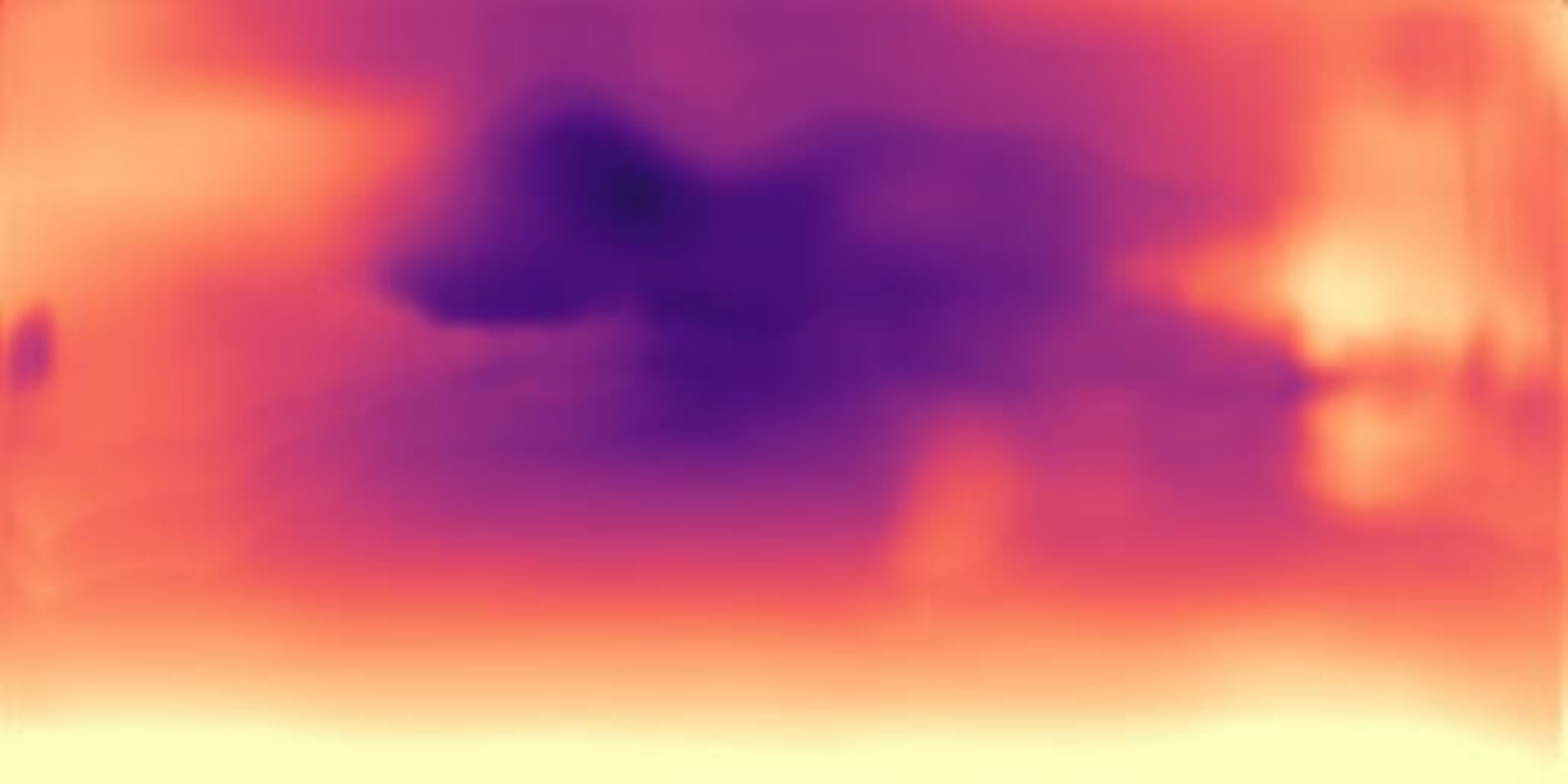}\qquad\qquad\quad &
\includegraphics[width=\iw,height=\ih]{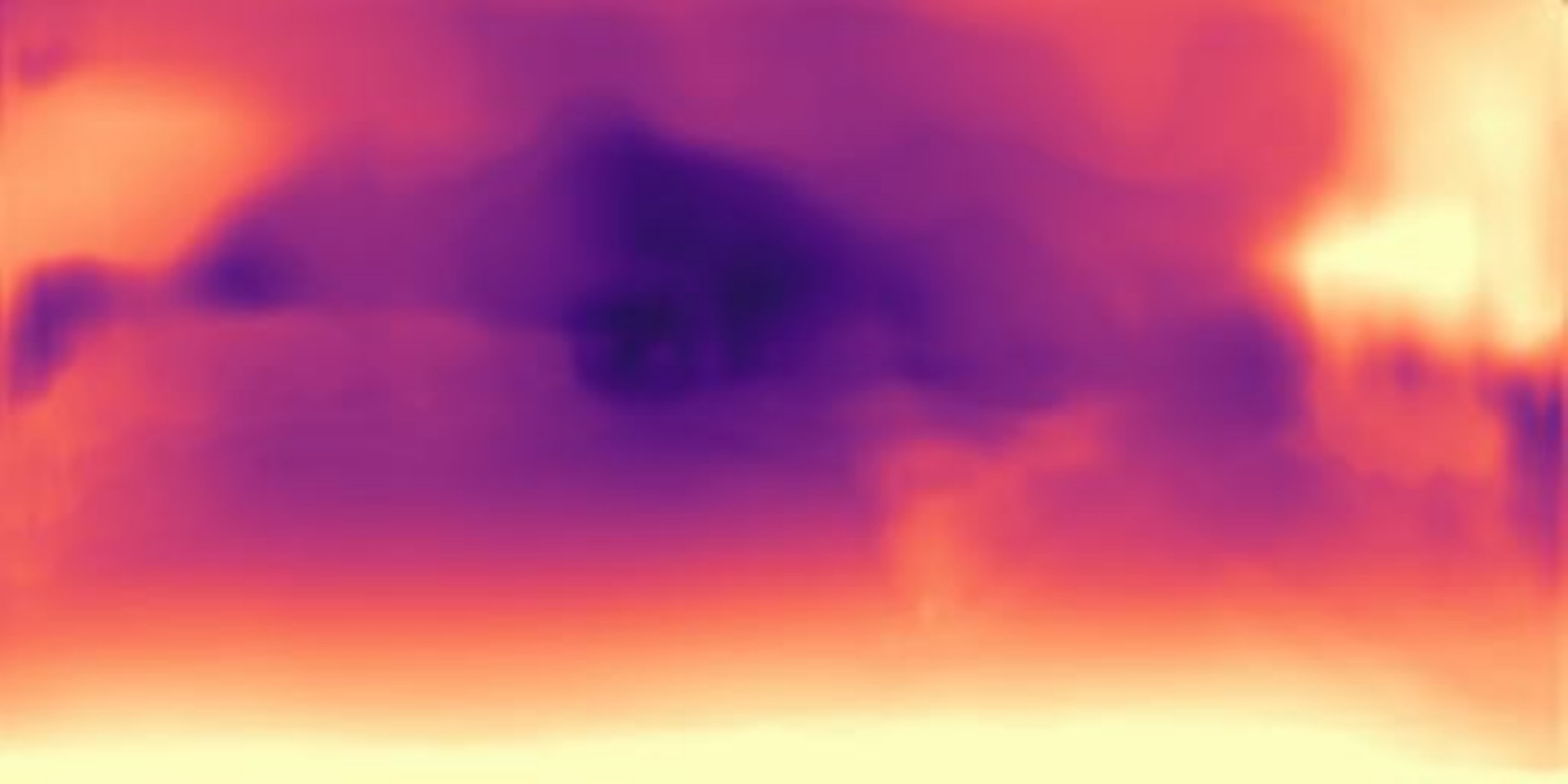}\\
\vspace{30mm} \\
\multicolumn{6}{c}{\fontsize{\w}{\h} \selectfont (a) Self-supervised CNN-based methods} \\
\vspace{30mm} \\
\rotatebox[origin=c]{90}{\fontsize{\textw}{\texth}\selectfont MF-ViT\hspace{-320mm}}\hspace{15mm}
\includegraphics[width=\iw,height=\ih]{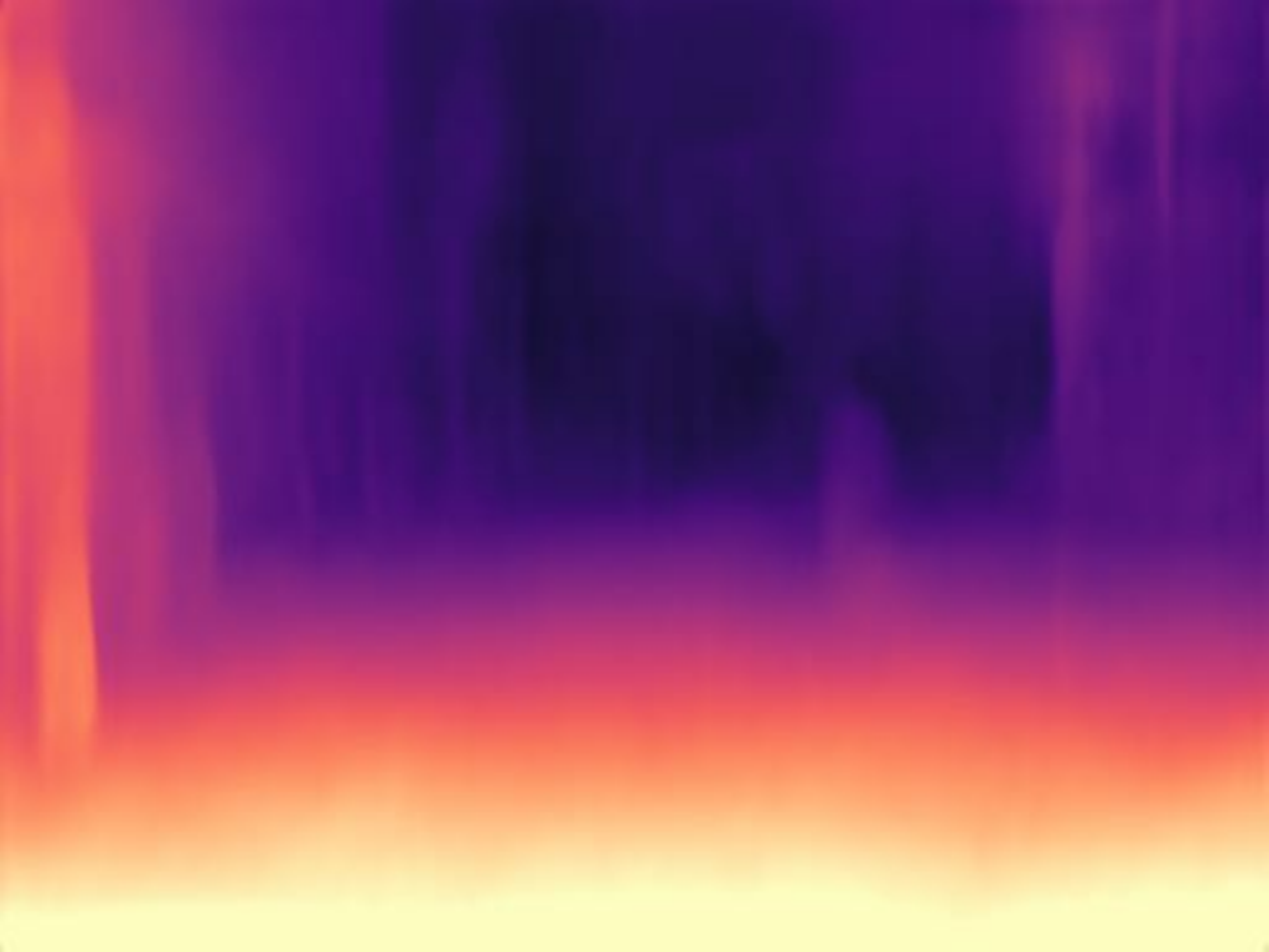}\qquad\qquad\quad &  
\includegraphics[width=\iw,height=\ih]{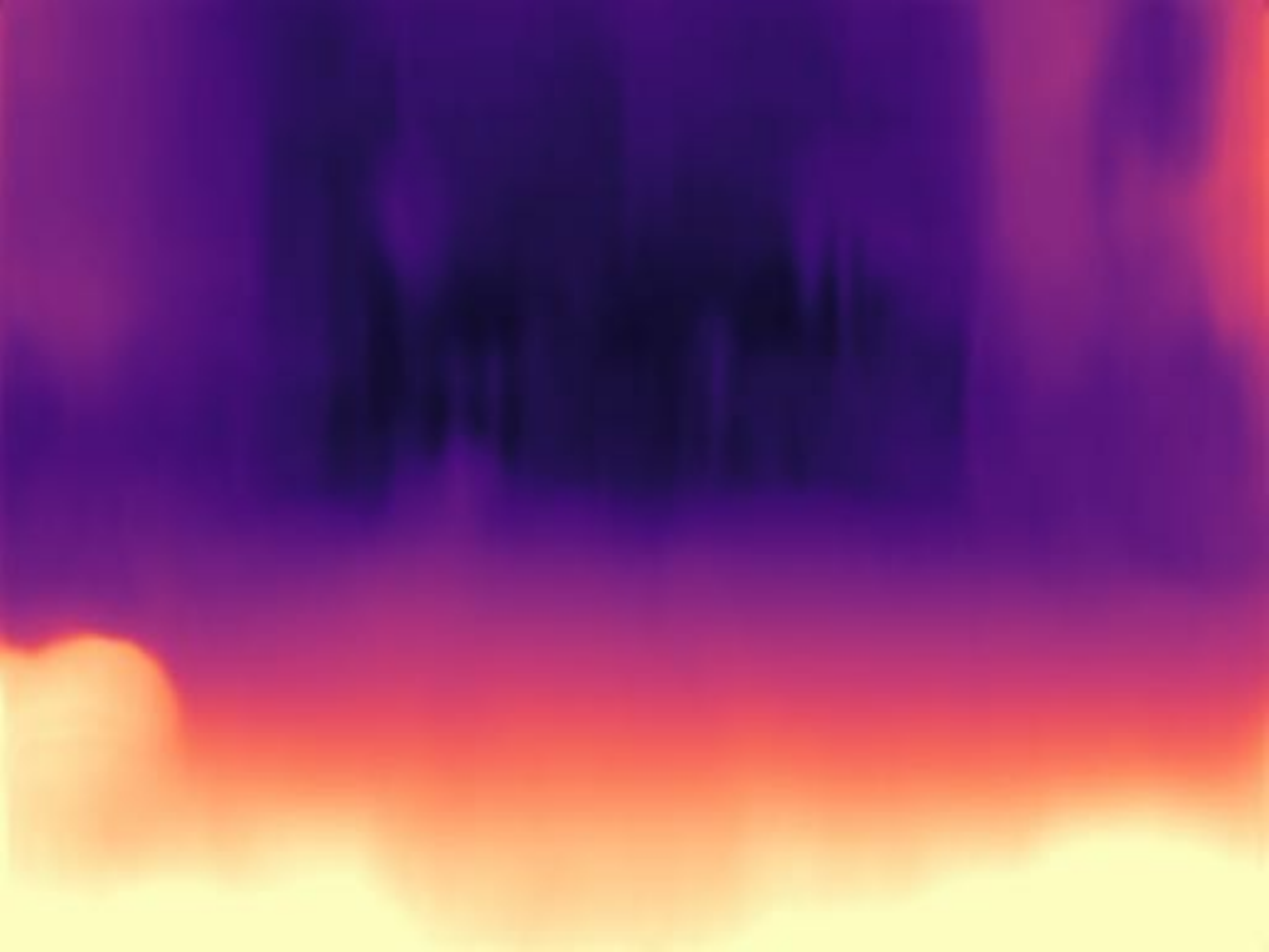}\qquad\qquad\quad &
\includegraphics[width=\iw,height=\ih]{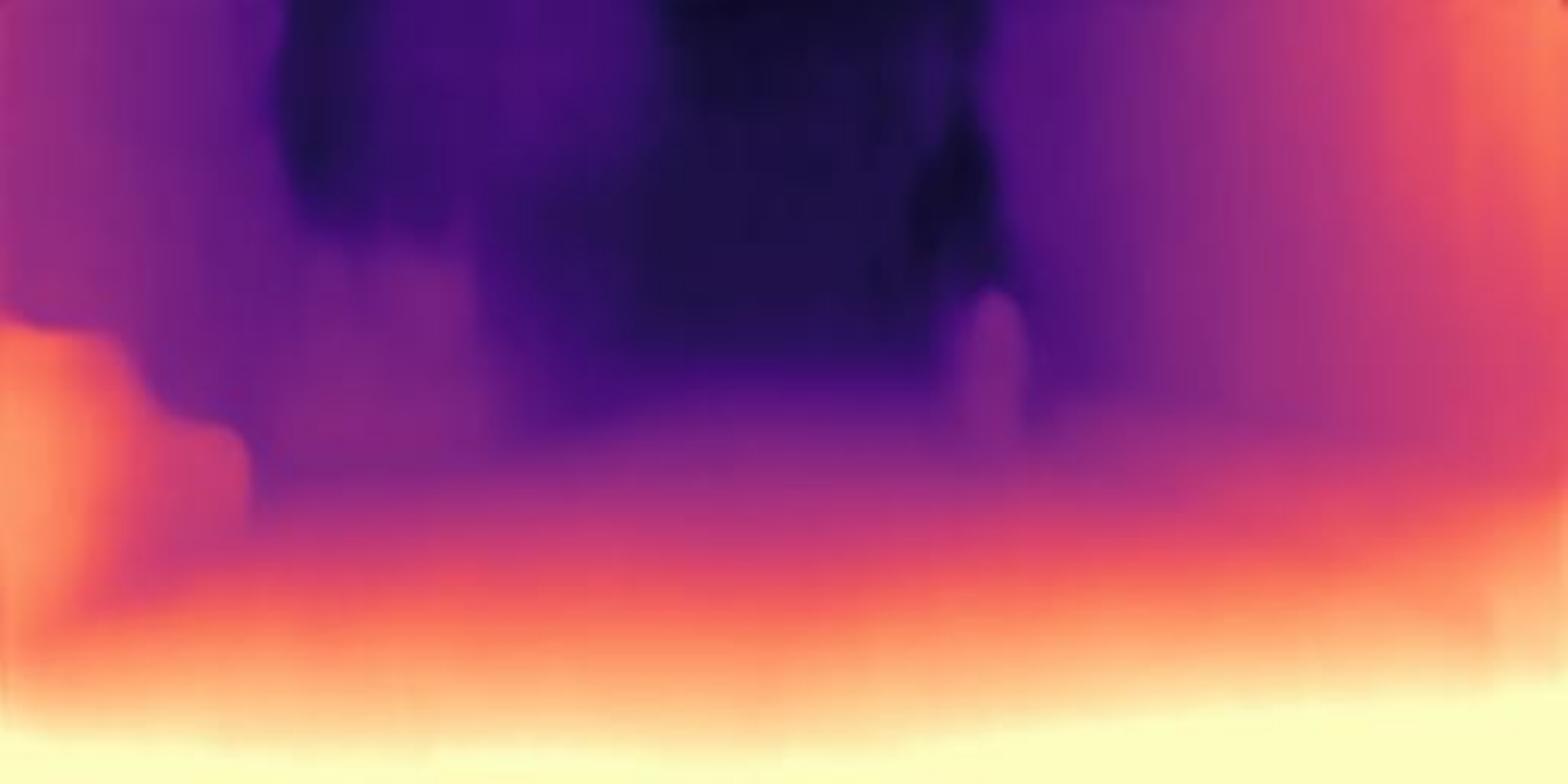}\qquad\qquad\quad &
\includegraphics[width=\iw,height=\ih]{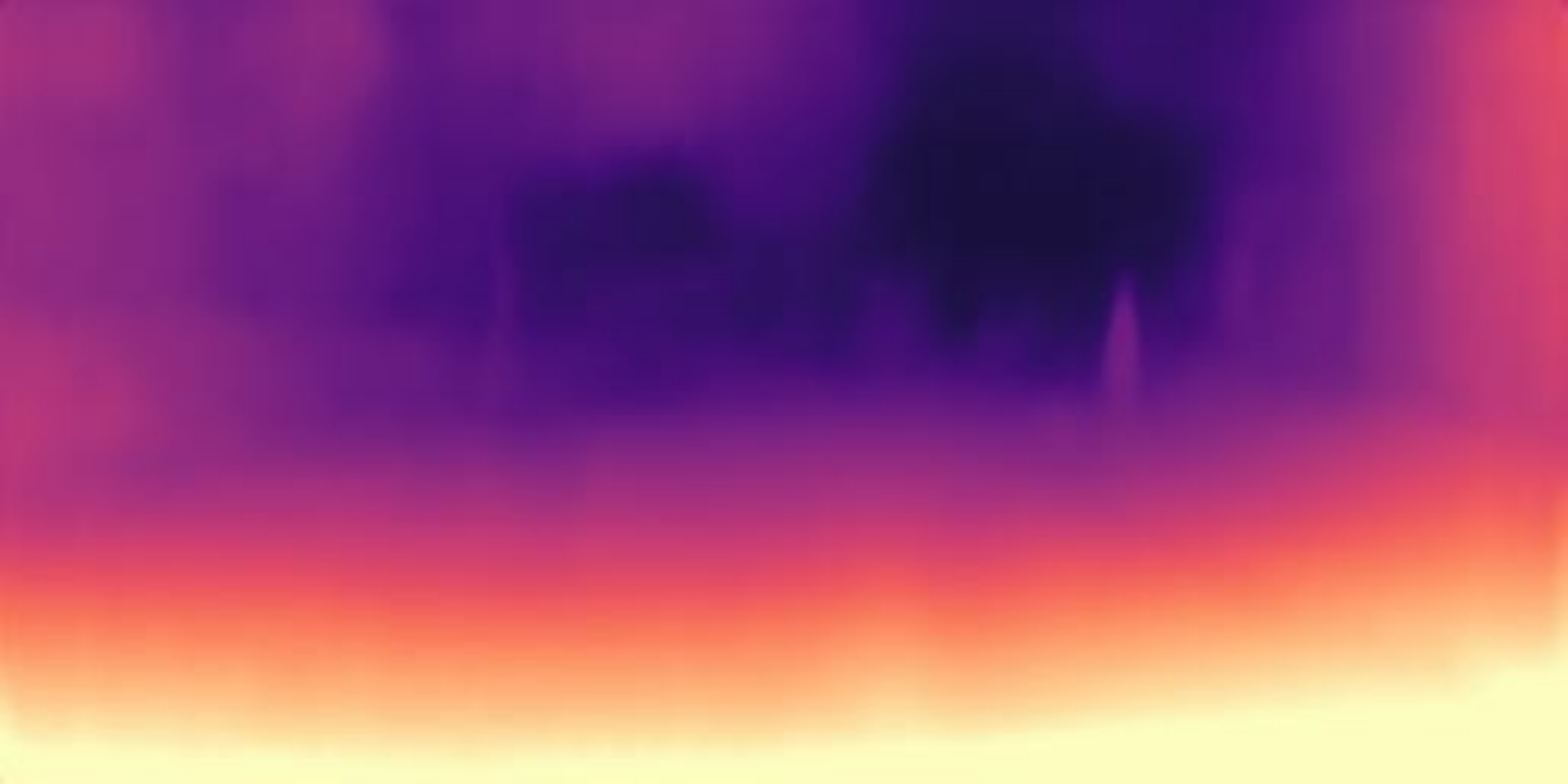}\qquad\qquad\quad &
\includegraphics[width=\iw,height=\ih]{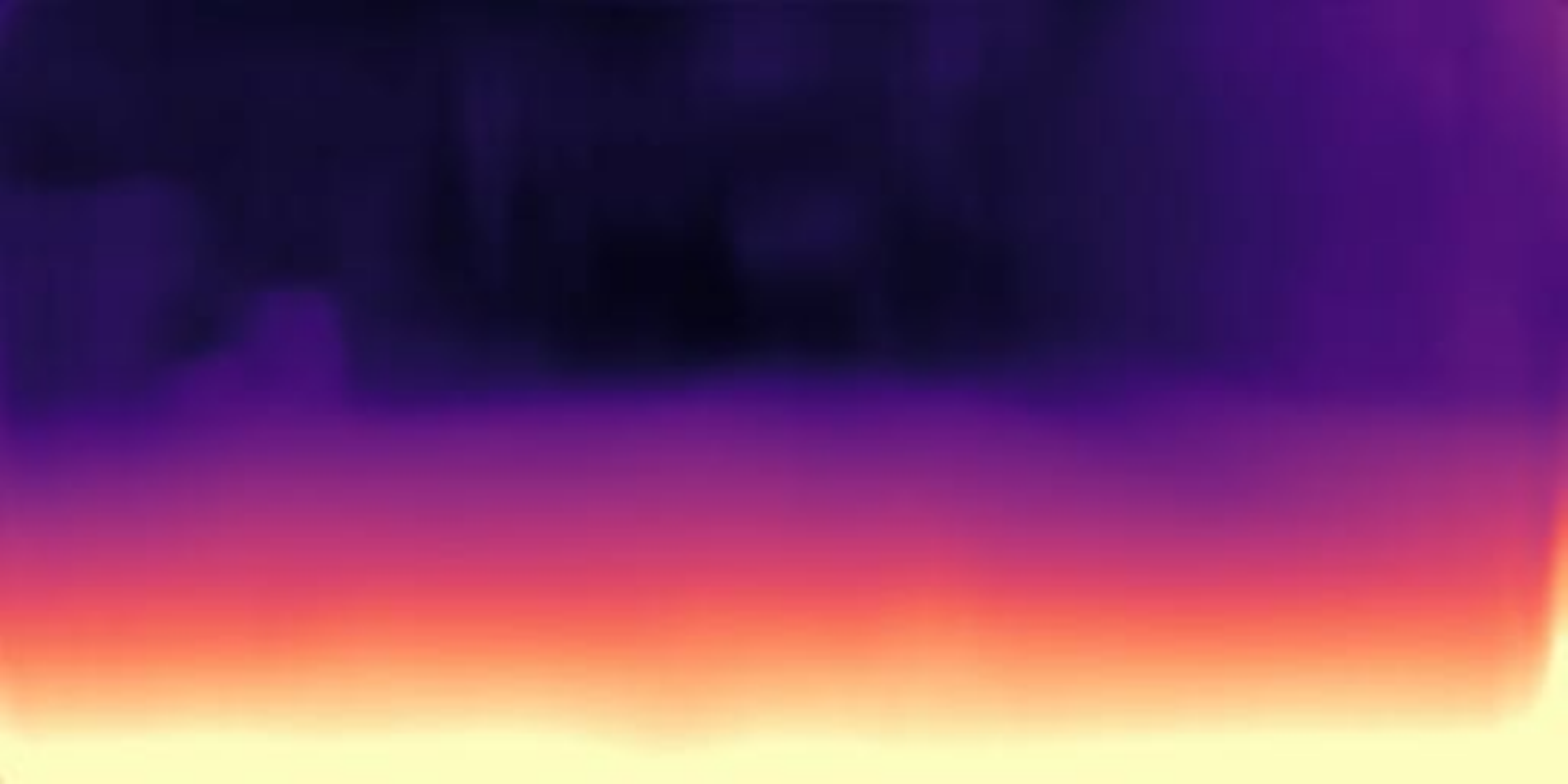}\qquad\qquad\quad &
\includegraphics[width=\iw,height=\ih]{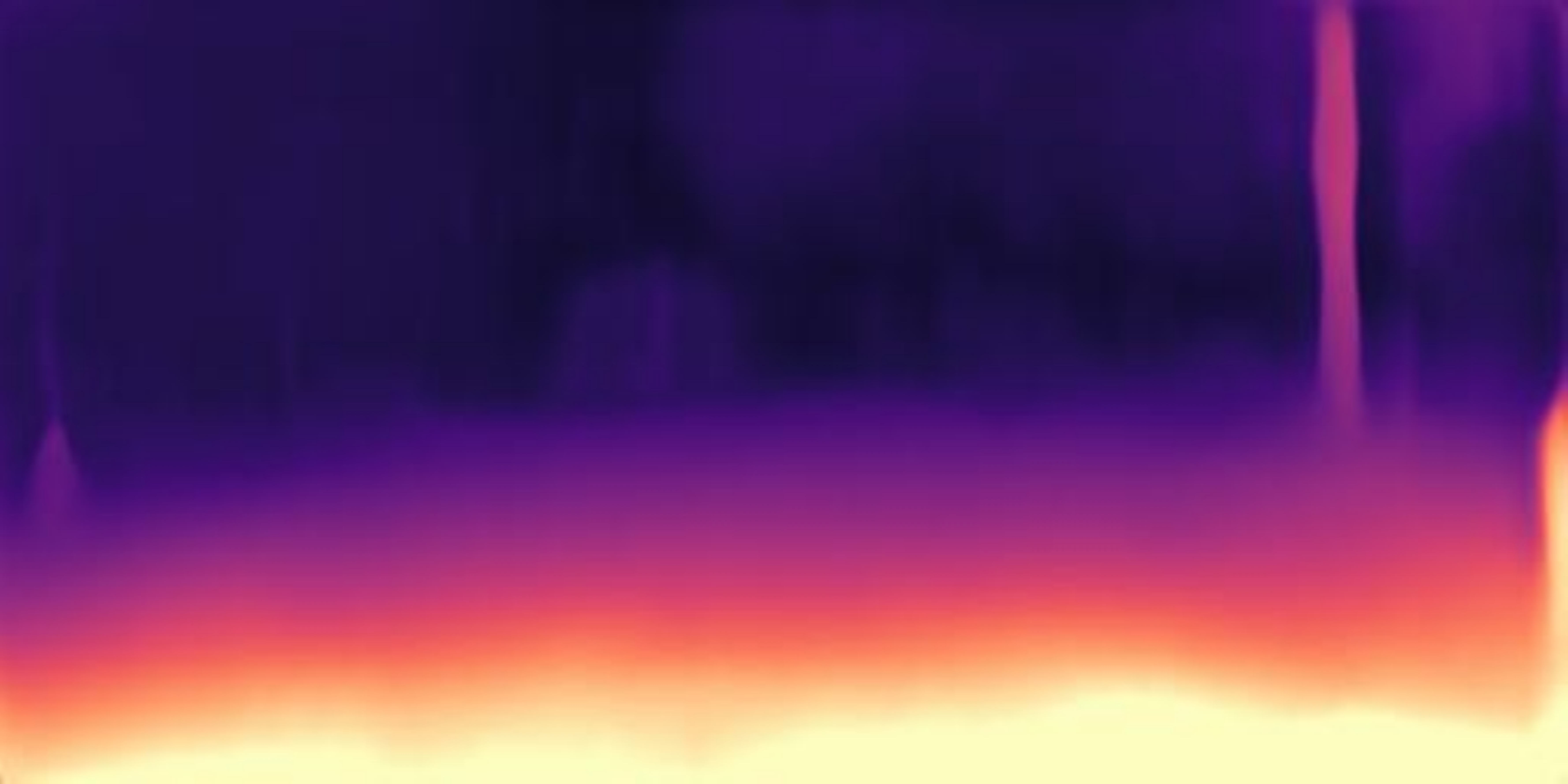}\\
\vspace{10mm} \\
\rotatebox[origin=c]{90}{\fontsize{\textw}{\texth}\selectfont MF-RegionViT\hspace{-300mm}}\hspace{15mm}
\includegraphics[width=\iw,height=\ih]{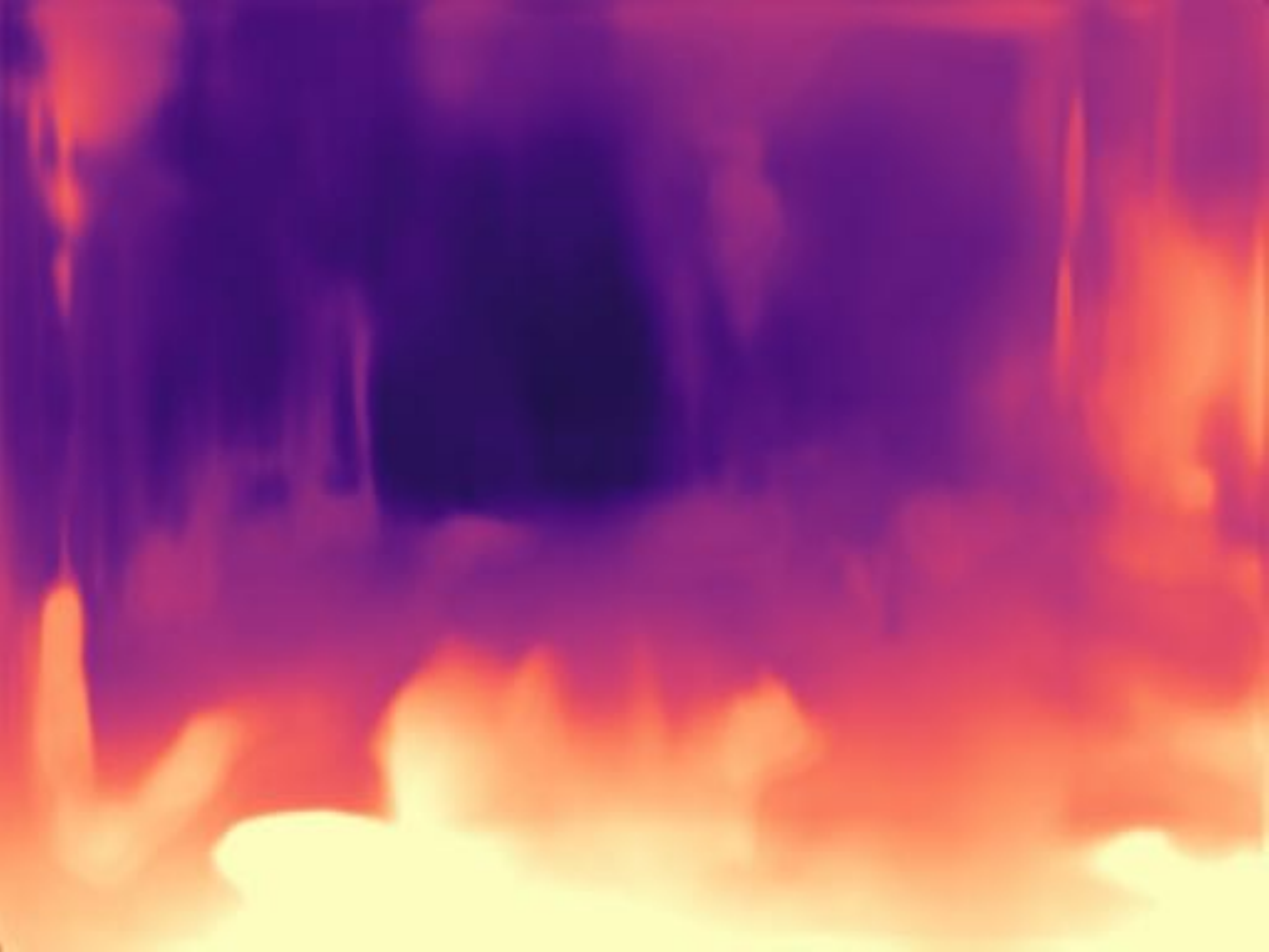}\qquad\qquad\quad &  
\includegraphics[width=\iw,height=\ih]{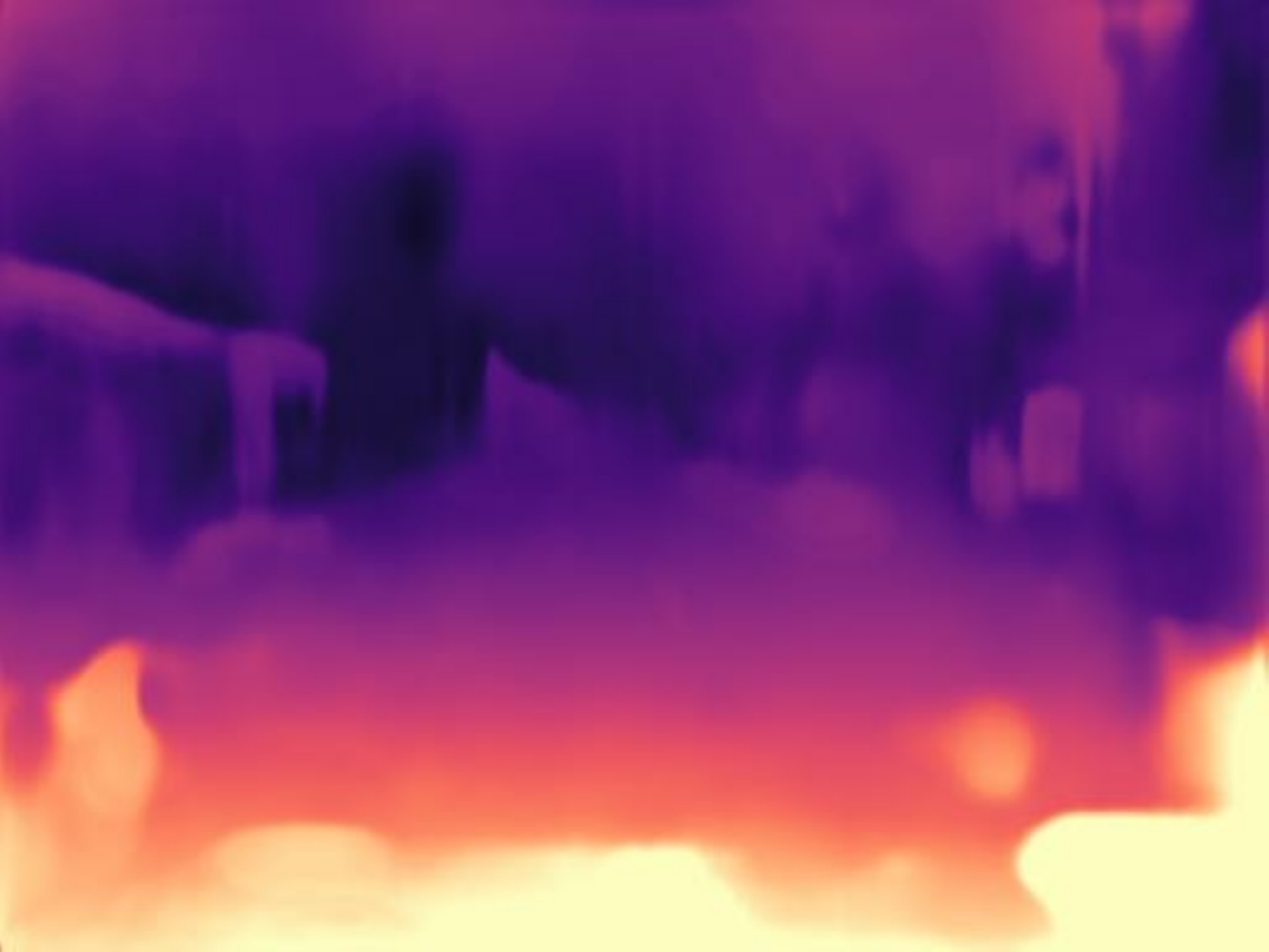}\qquad\qquad\quad &
\includegraphics[width=\iw,height=\ih]{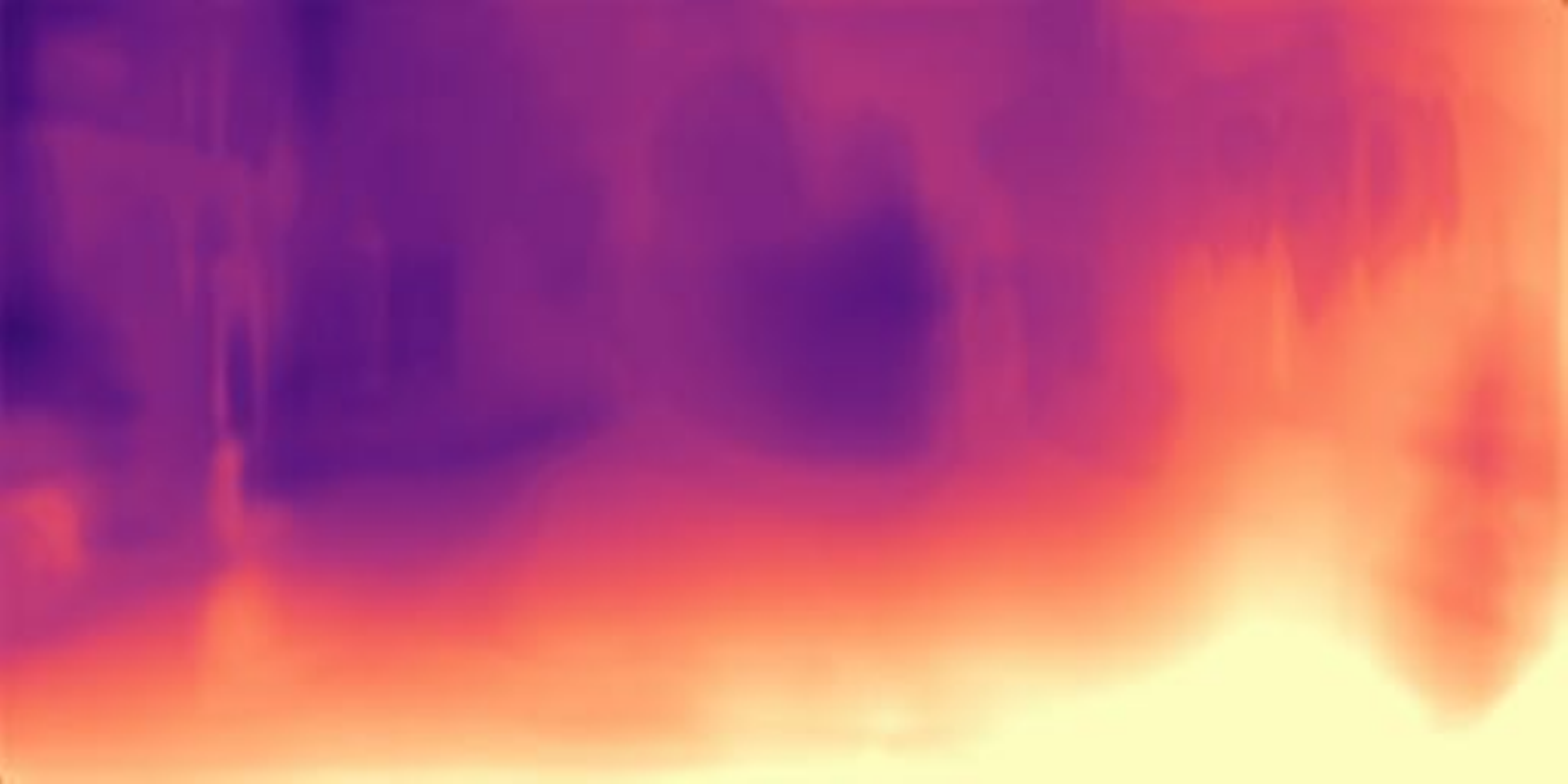}\qquad\qquad\quad &
\includegraphics[width=\iw,height=\ih]{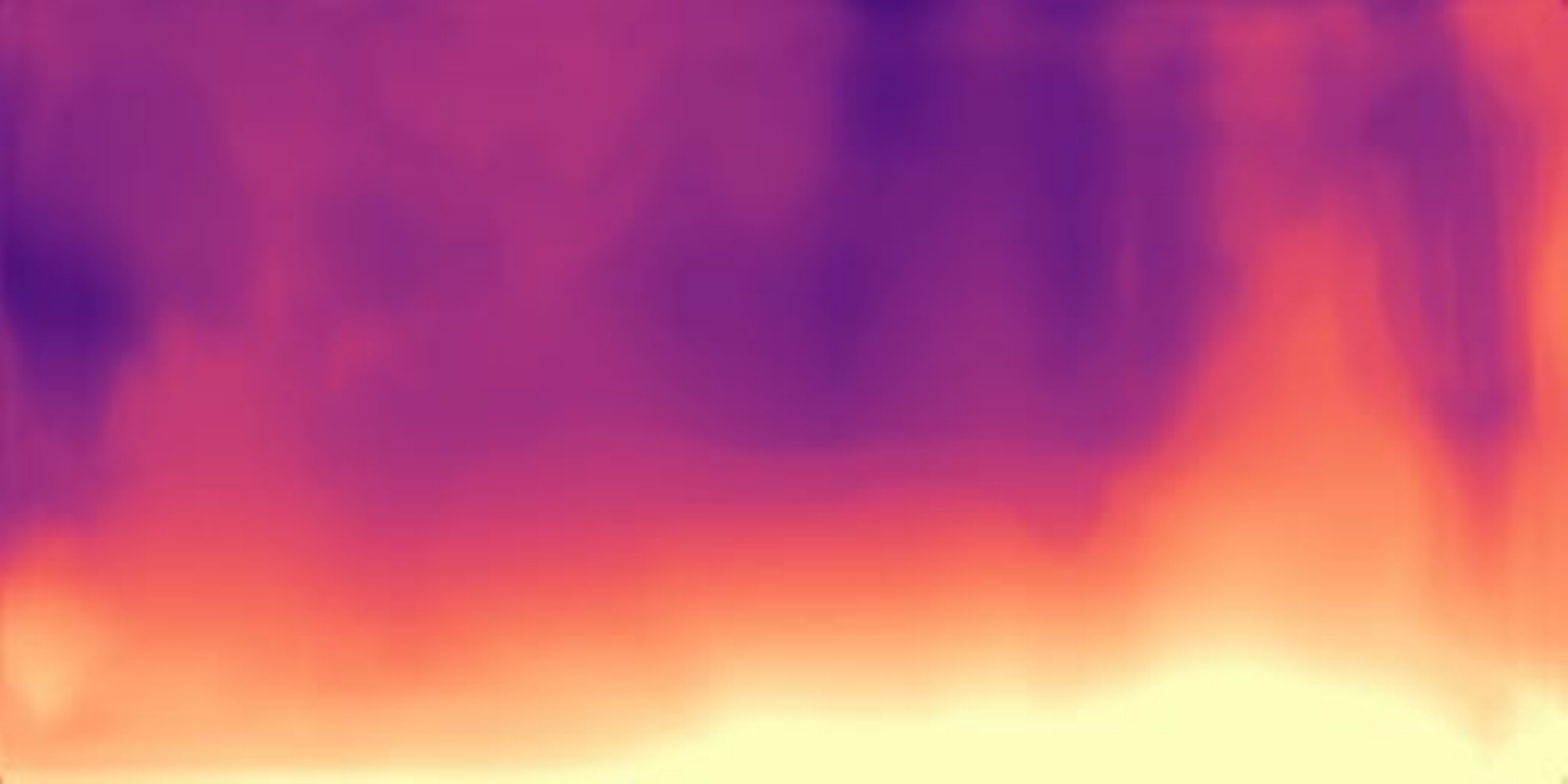}\qquad\qquad\quad &
\includegraphics[width=\iw,height=\ih]{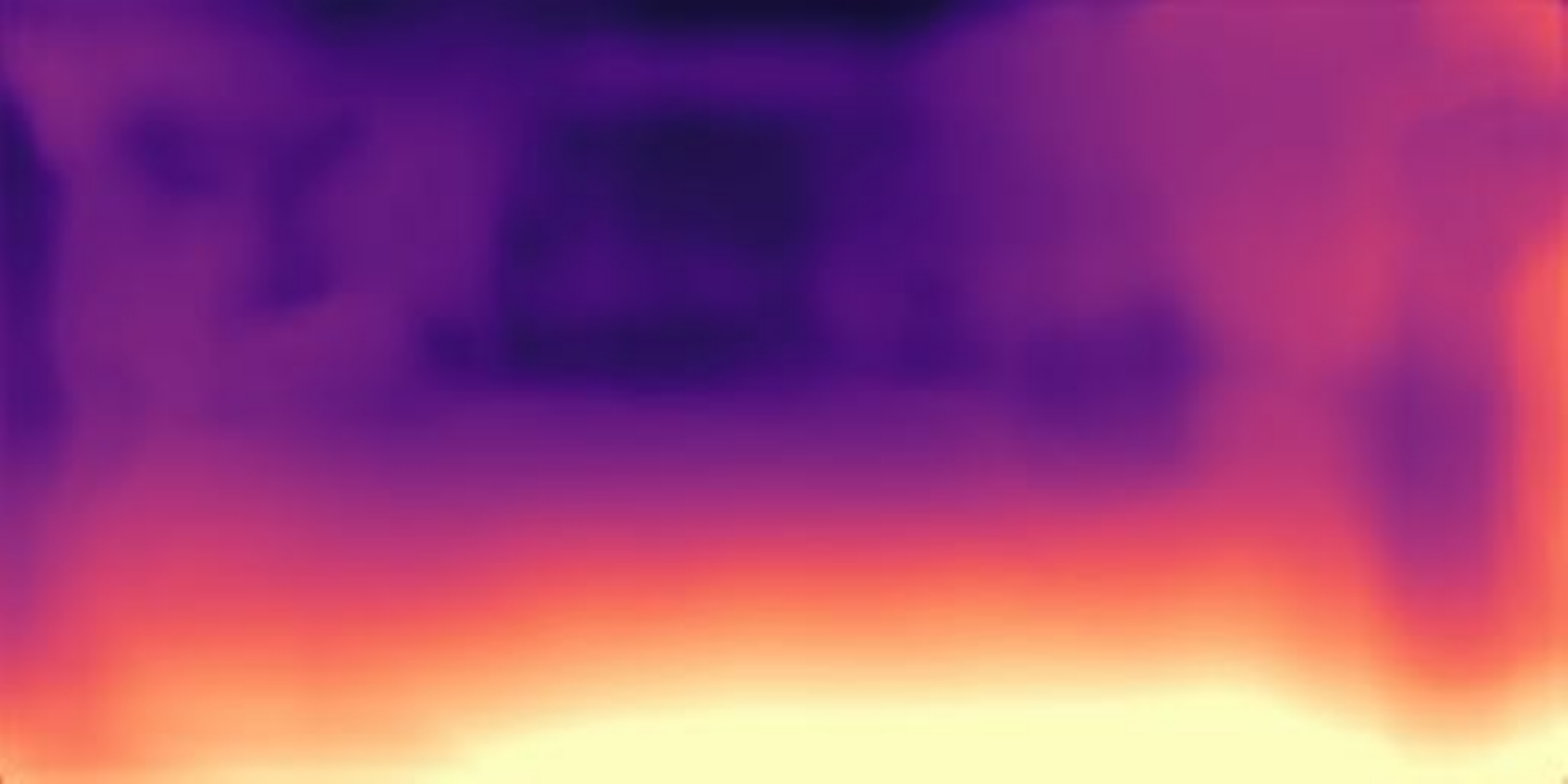}\qquad\qquad\quad &
\includegraphics[width=\iw,height=\ih]{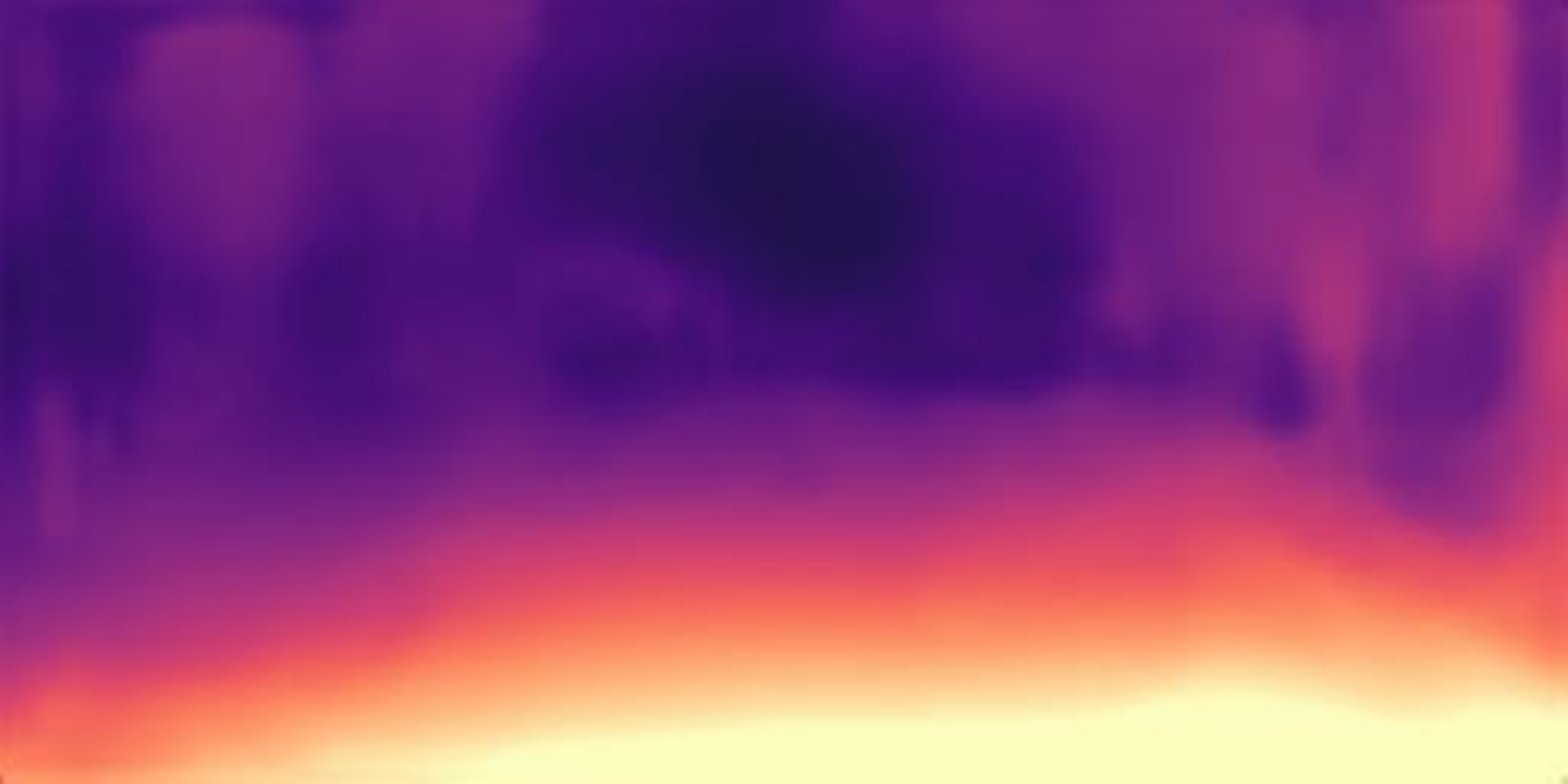}\\
\vspace{10mm} \\
\rotatebox[origin=c]{90}{\fontsize{\textw}{\texth}\selectfont MF-Twins\hspace{-320mm}}\hspace{15mm}
\includegraphics[width=\iw,height=\ih]{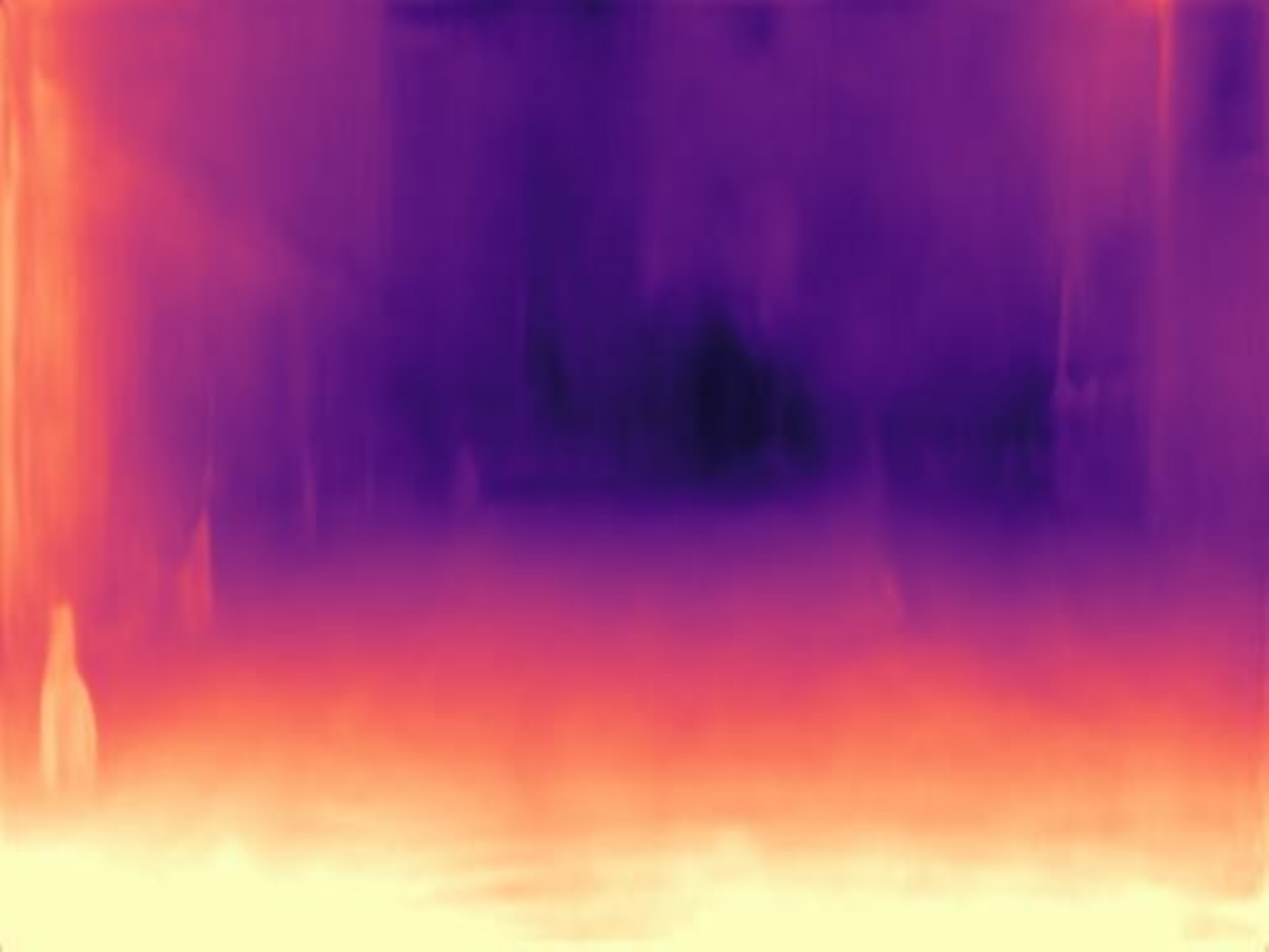}\qquad\qquad\quad &
\includegraphics[width=\iw,height=\ih]{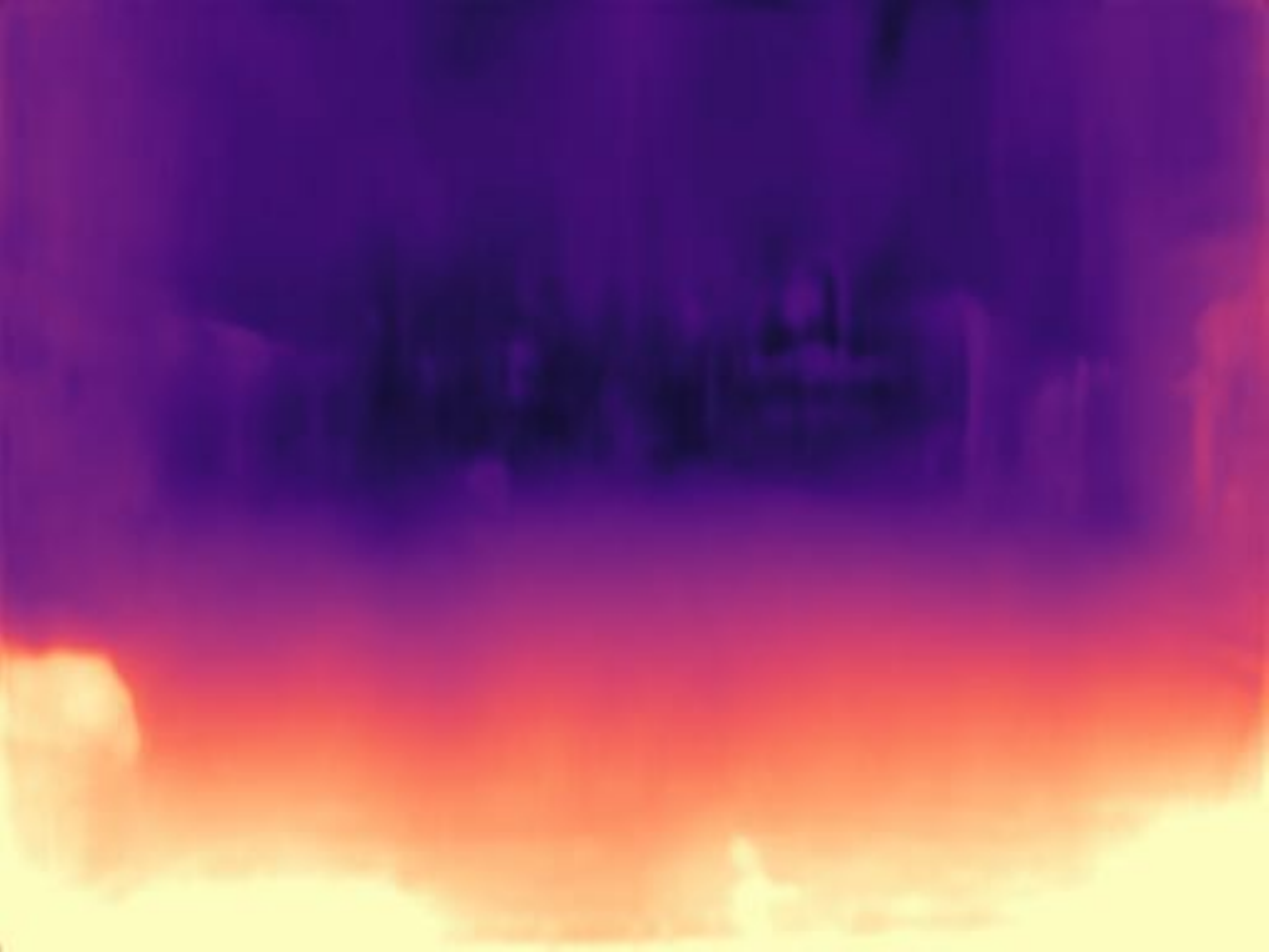}\qquad\qquad\quad &
\includegraphics[width=\iw,height=\ih]{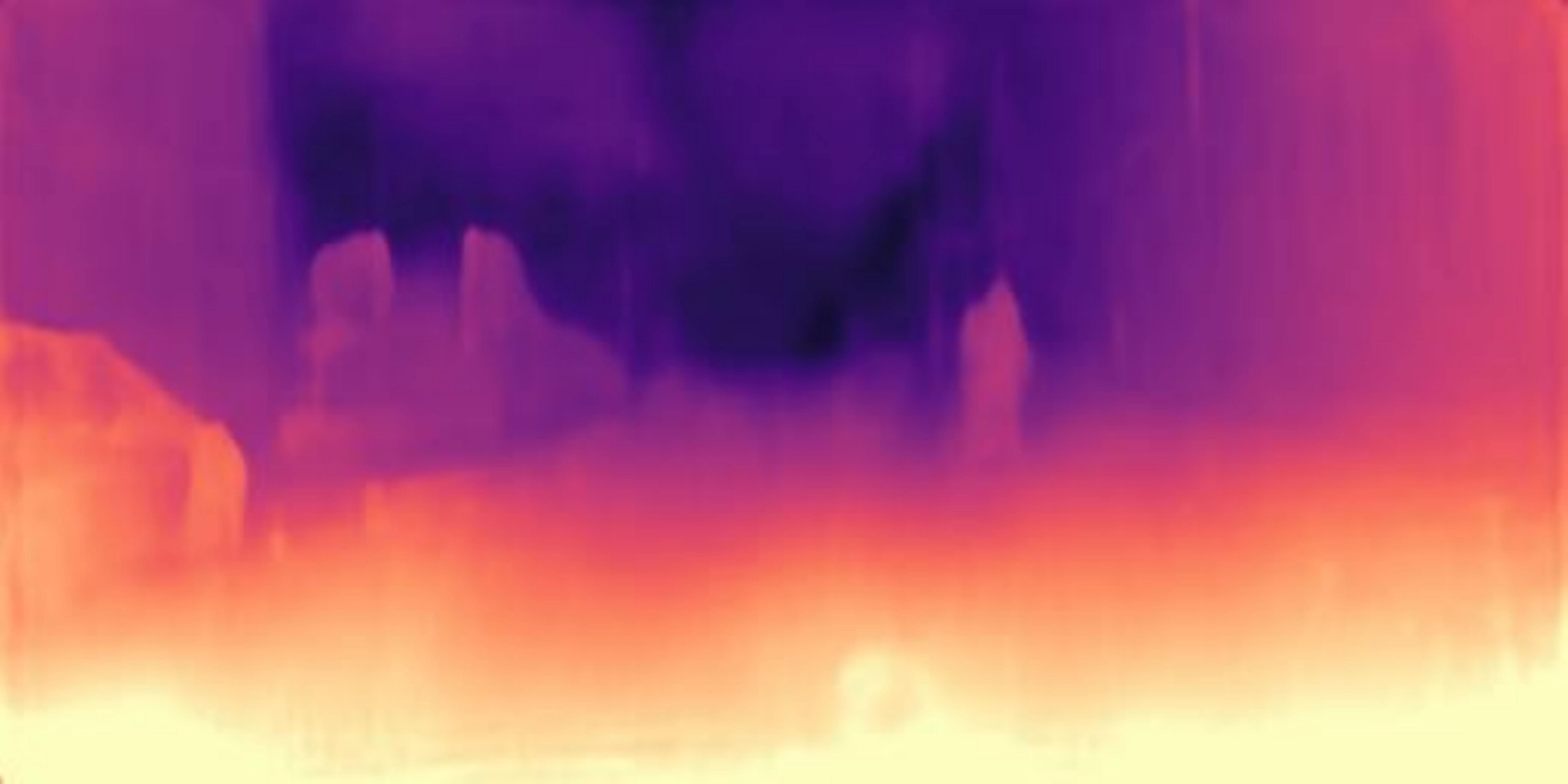}\qquad\qquad\quad &
\includegraphics[width=\iw,height=\ih]{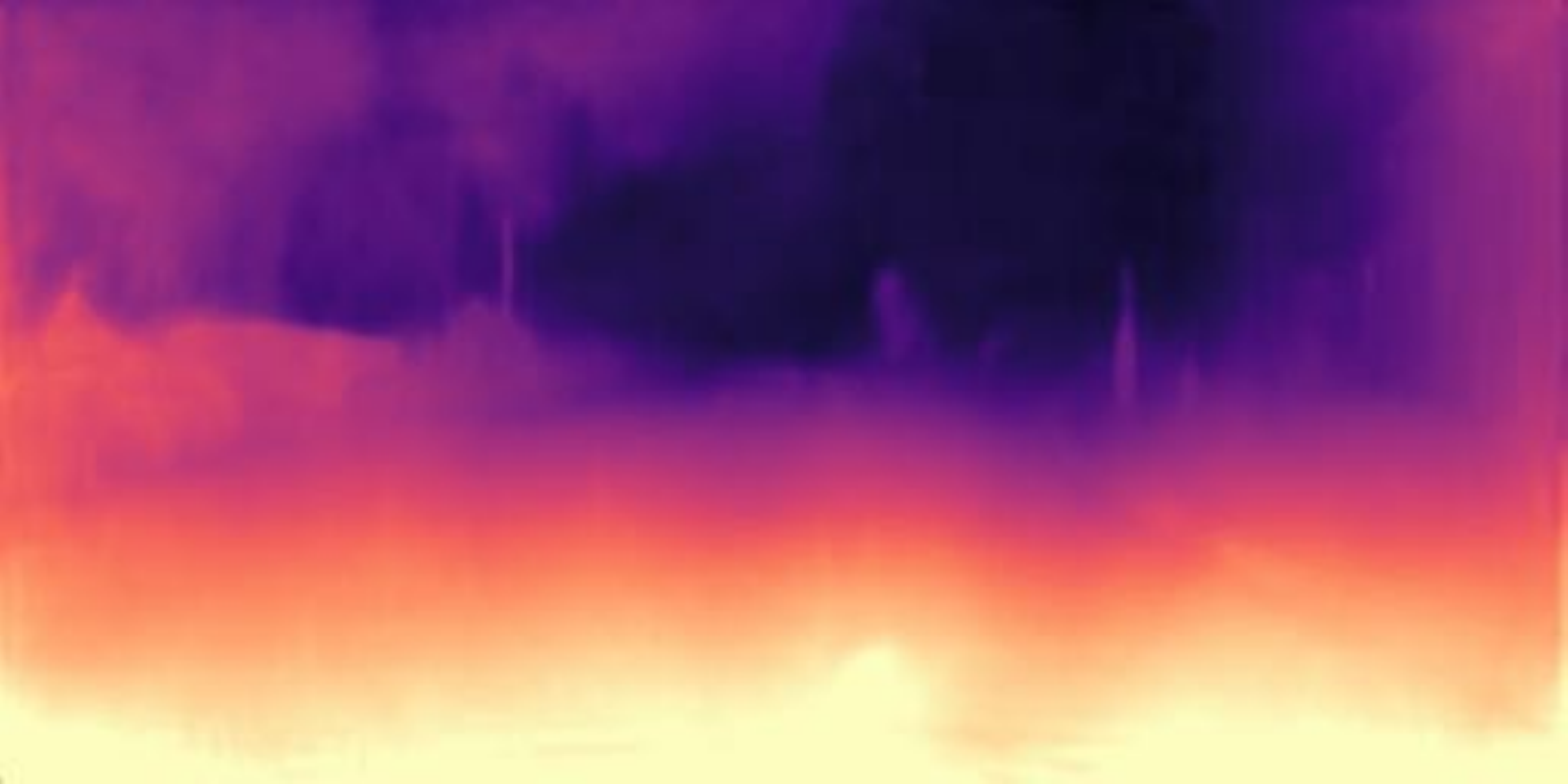}\qquad\qquad\quad &
\includegraphics[width=\iw,height=\ih]{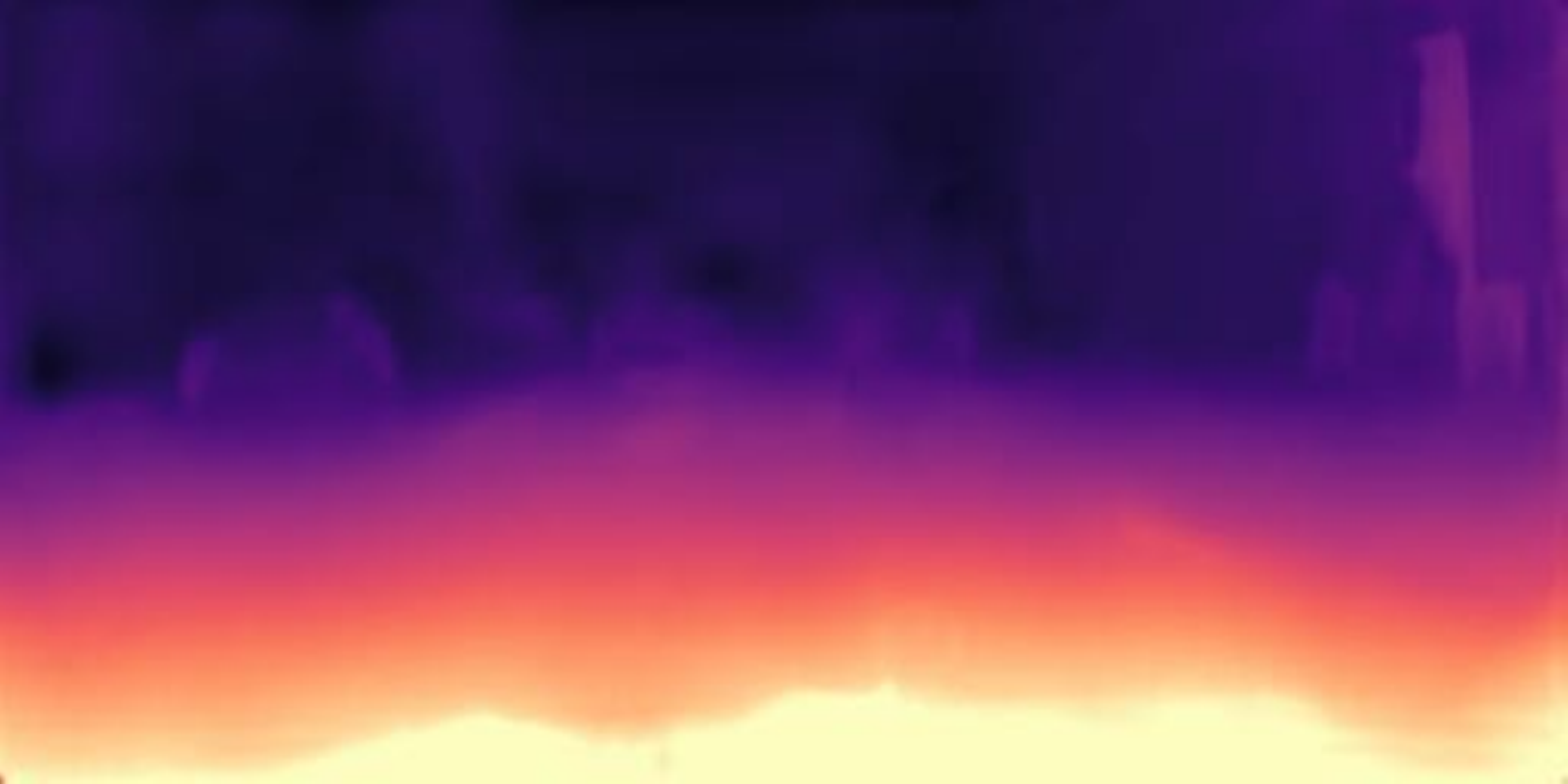}\qquad\qquad\quad &
\includegraphics[width=\iw,height=\ih]{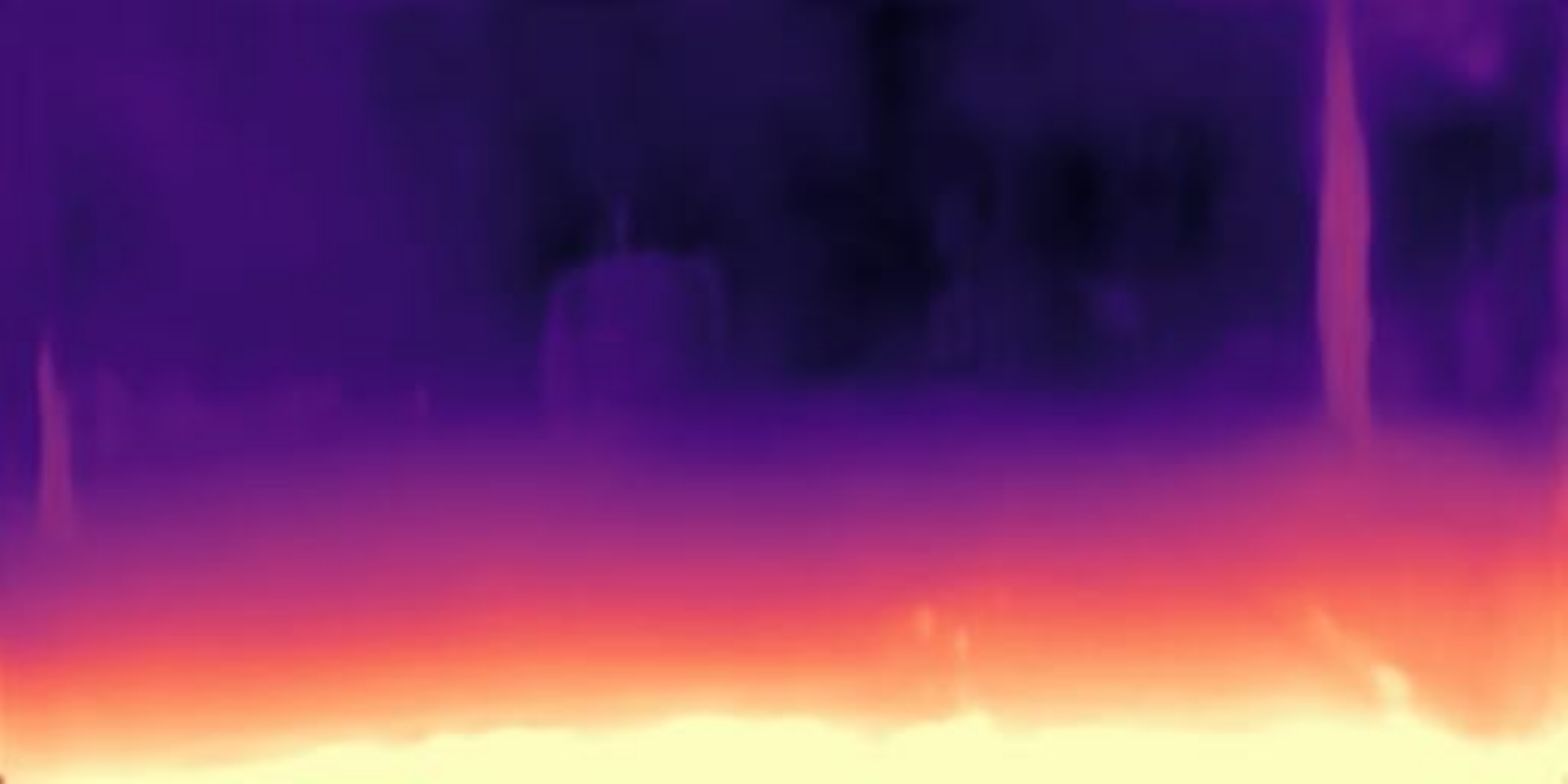}\\
\vspace{10mm} \\
\rotatebox[origin=c]{90}{\fontsize{\textw}{\texth}\selectfont MF-Ours\hspace{-310mm}}\hspace{15mm}
\includegraphics[width=\iw,height=\ih]{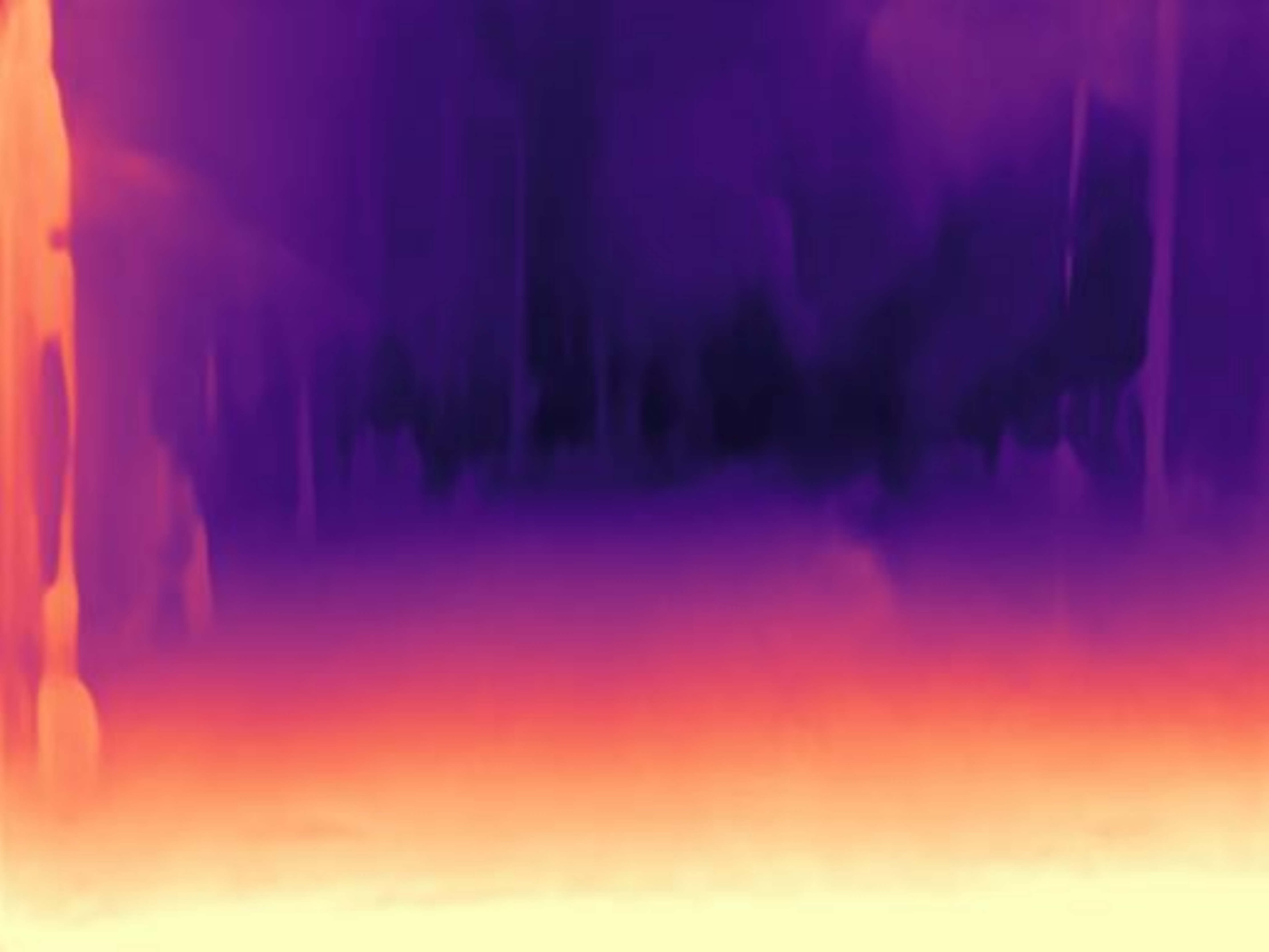}\qquad\qquad\quad &  
\includegraphics[width=\iw,height=\ih]{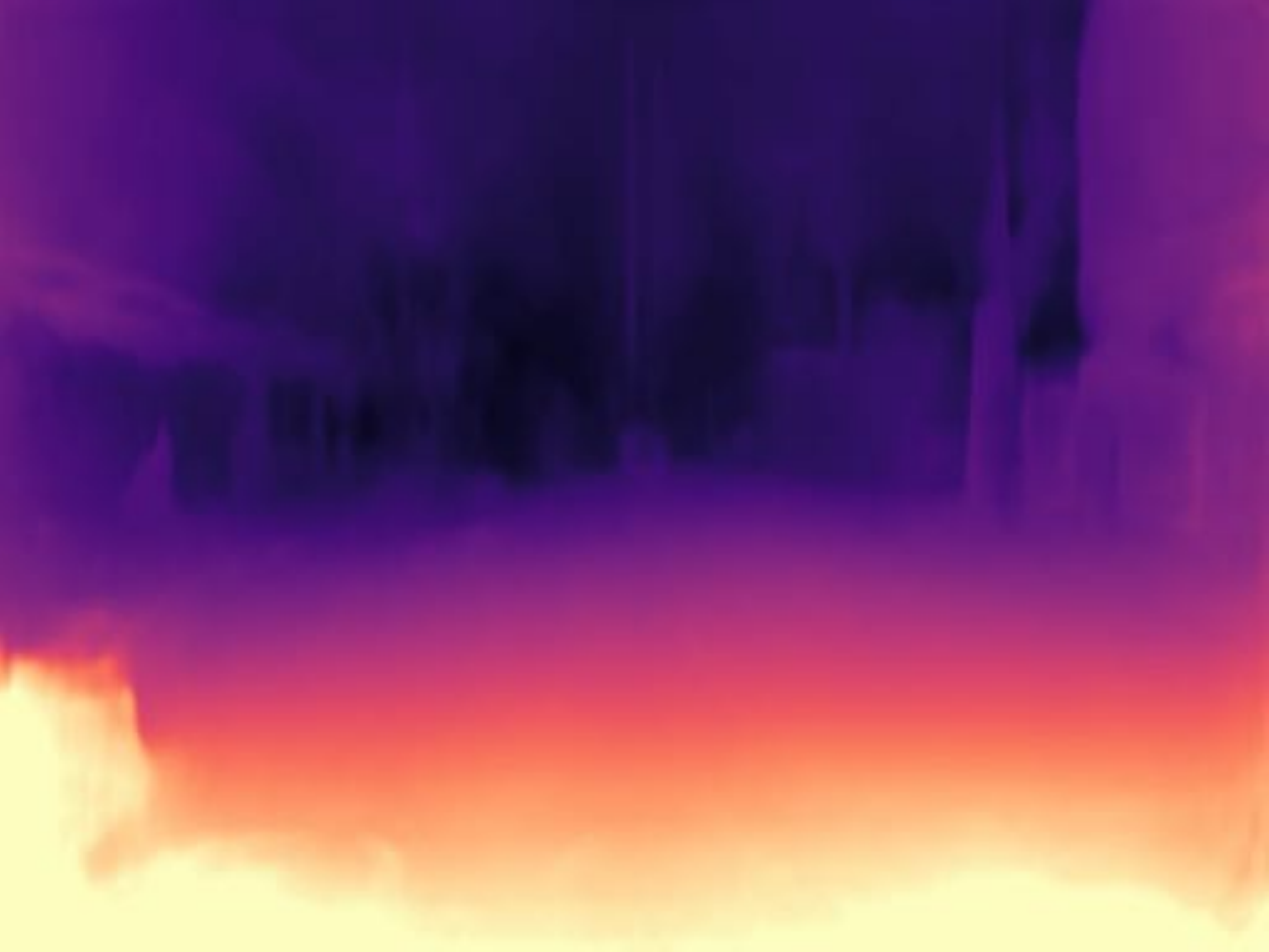}\qquad\qquad\quad &
\includegraphics[width=\iw,height=\ih]{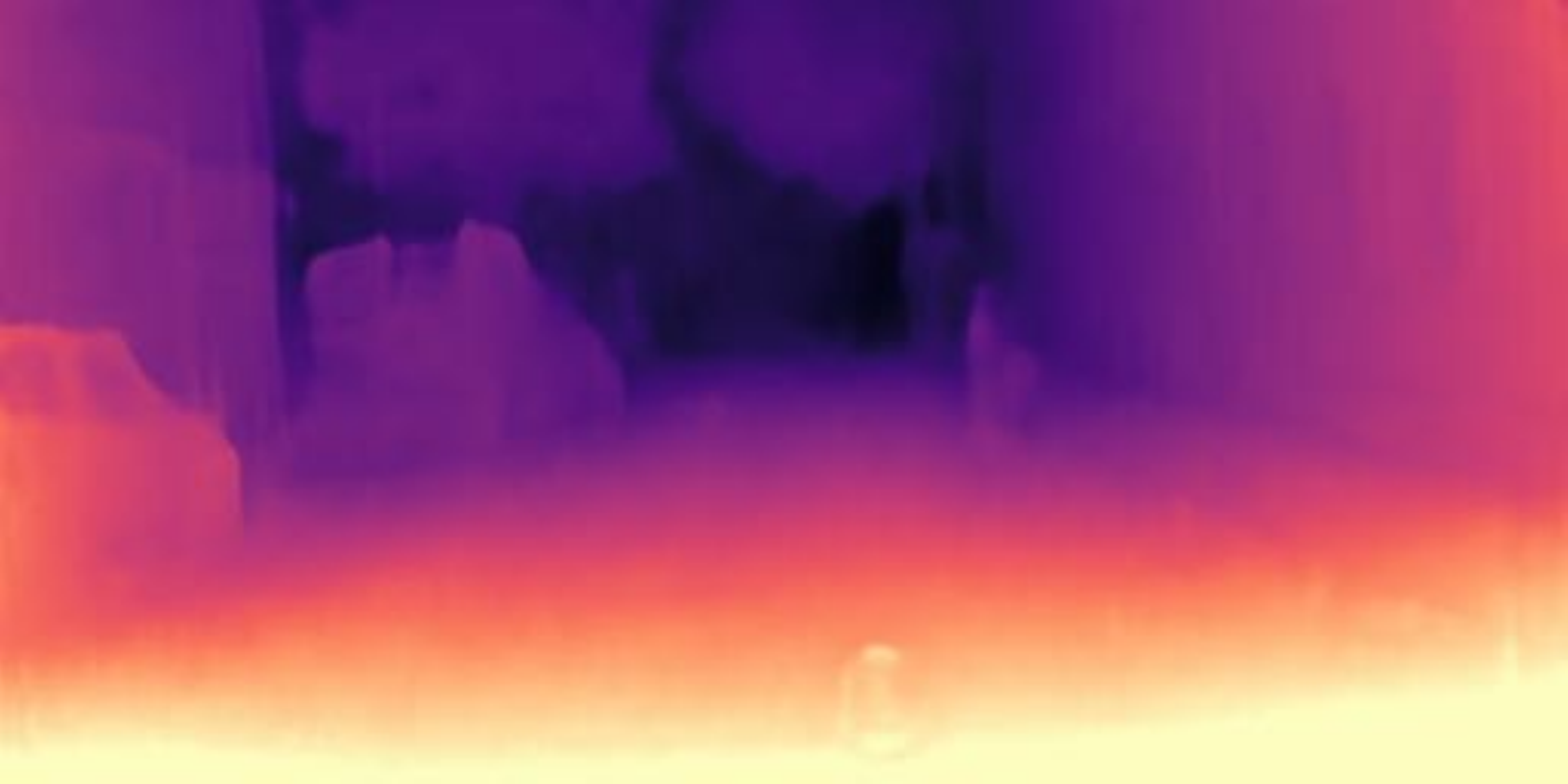}\qquad\qquad\quad &
\includegraphics[width=\iw,height=\ih]{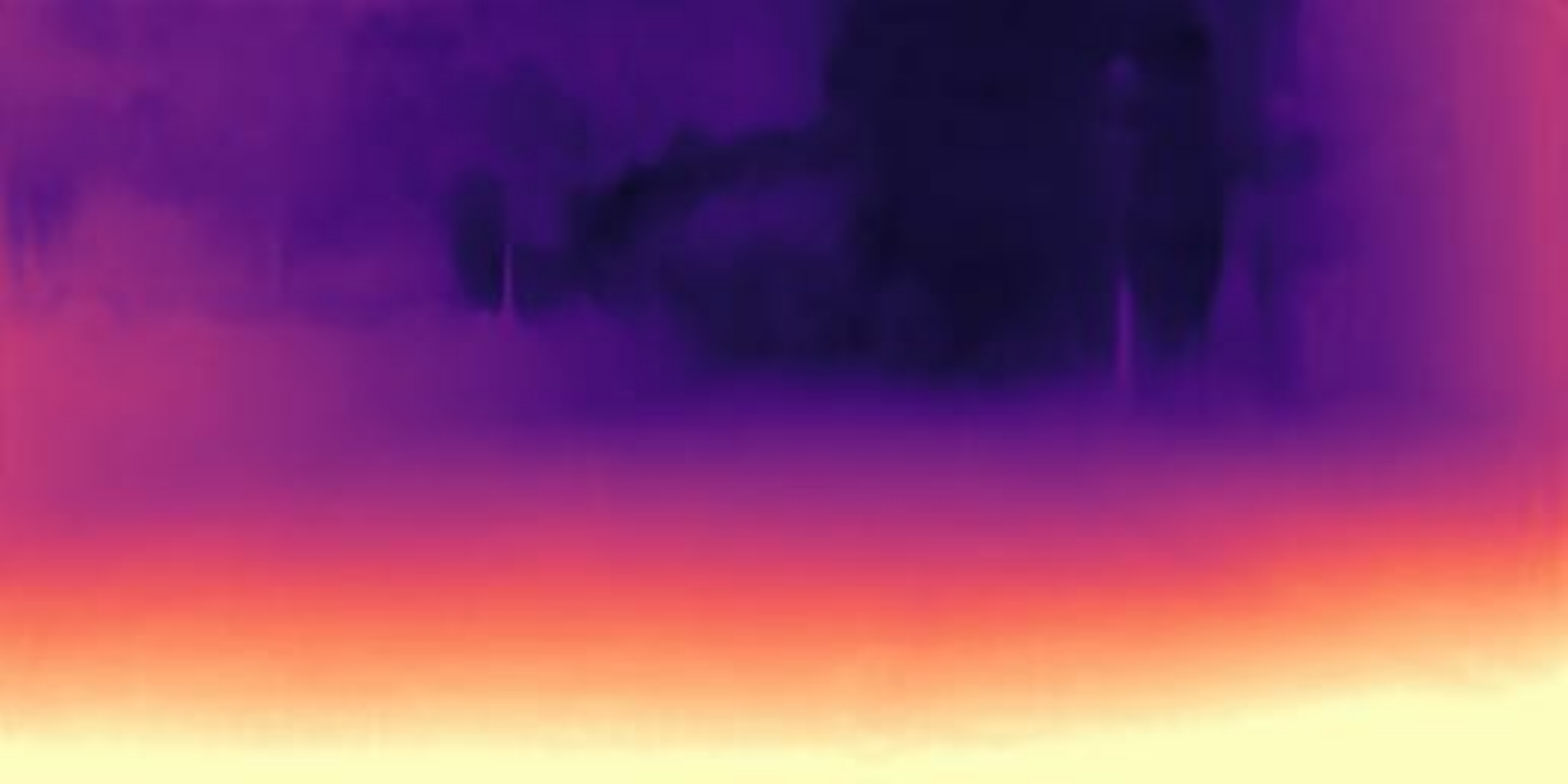}\qquad\qquad\quad &
\includegraphics[width=\iw,height=\ih]{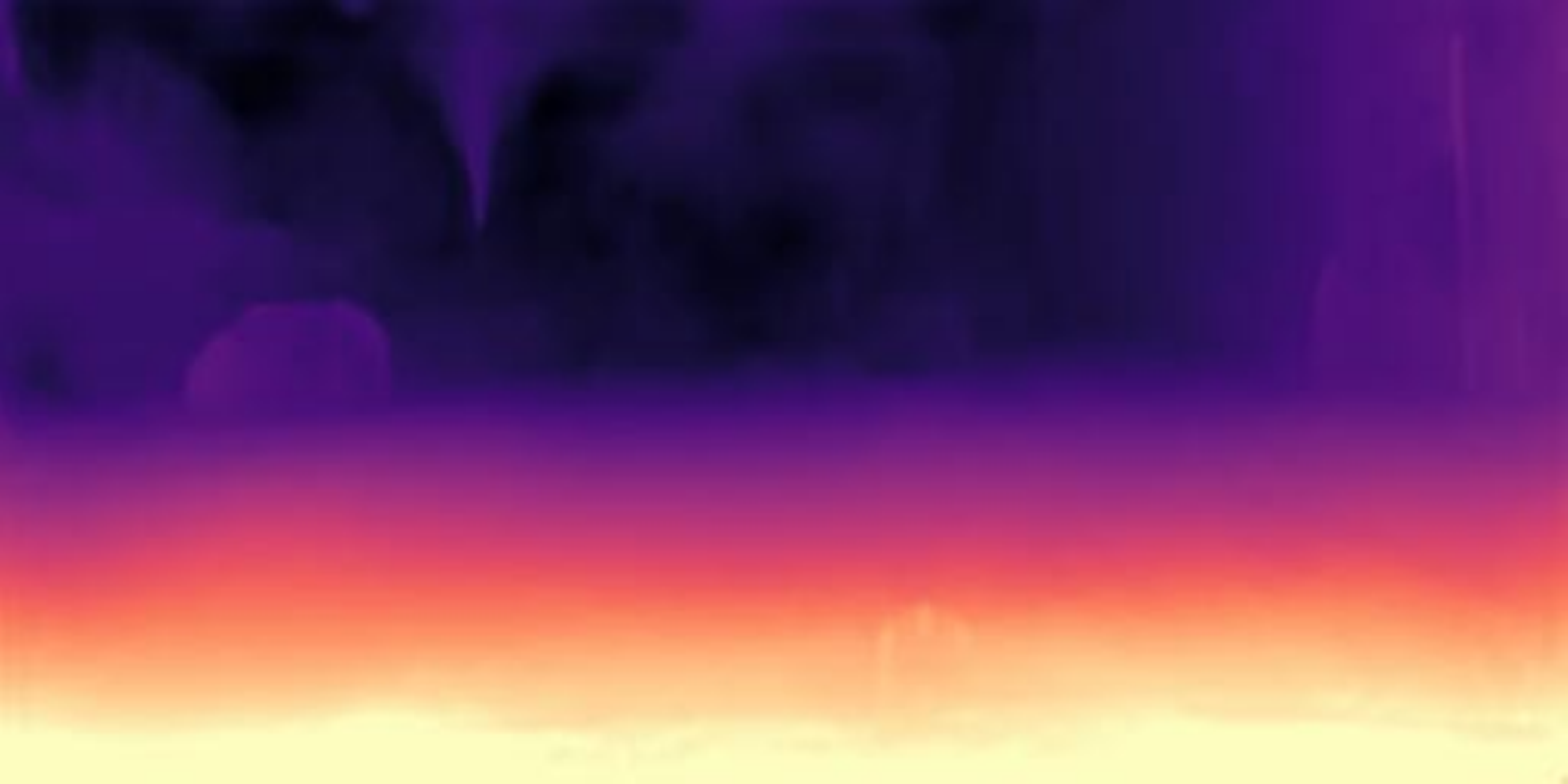}\qquad\qquad\quad &
\includegraphics[width=\iw,height=\ih]{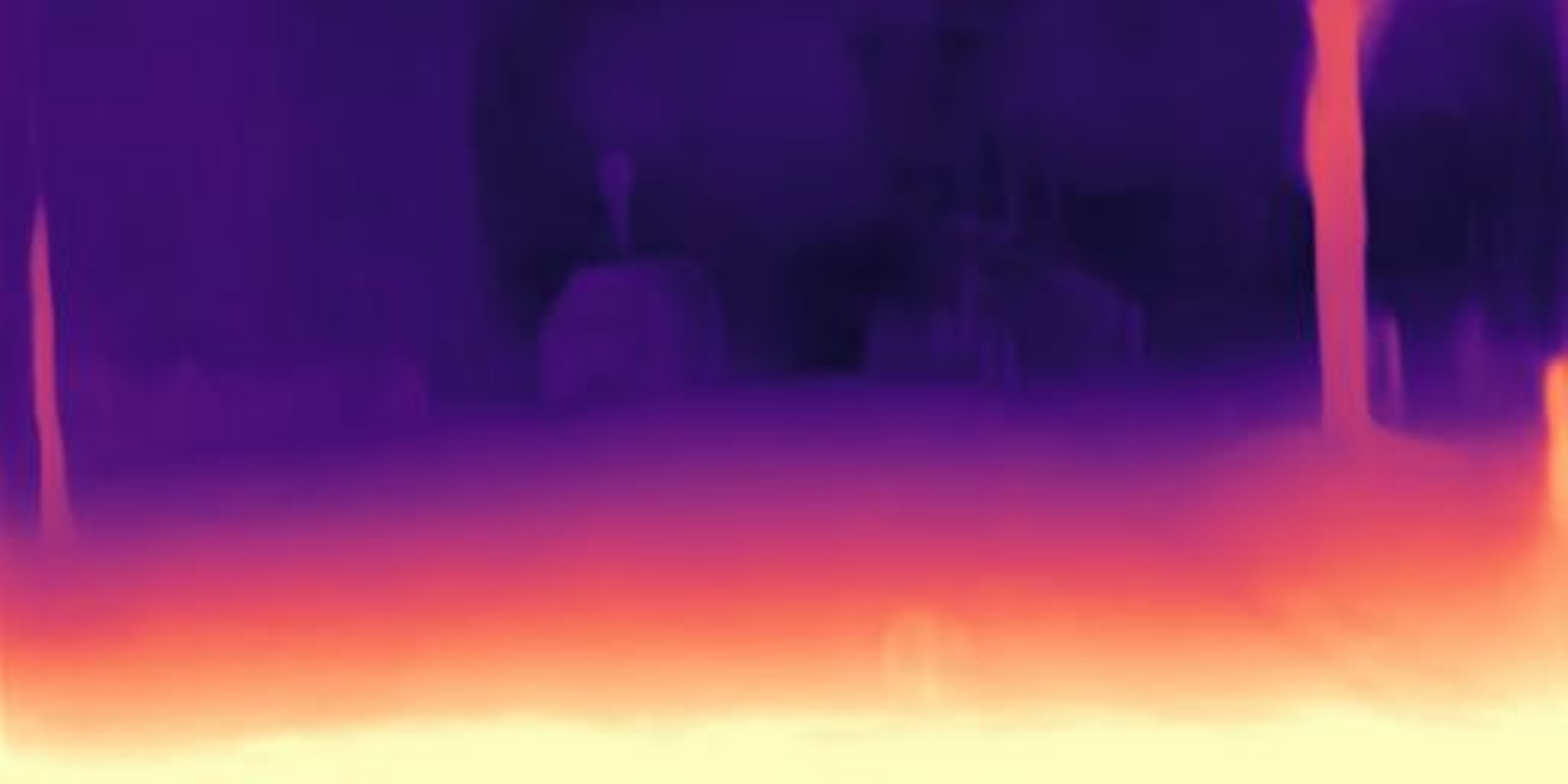}\\
\vspace{30mm} \\
\multicolumn{6}{c}{\fontsize{\w}{\h} \selectfont (b) Self-supervised Transformer-based methods} \\
\vspace{30mm} \\
\rotatebox[origin=c]{90}{\fontsize{\textw}{\texth}\selectfont BTS\hspace{-320mm}}\hspace{15mm}
\includegraphics[width=\iw,height=\ih]{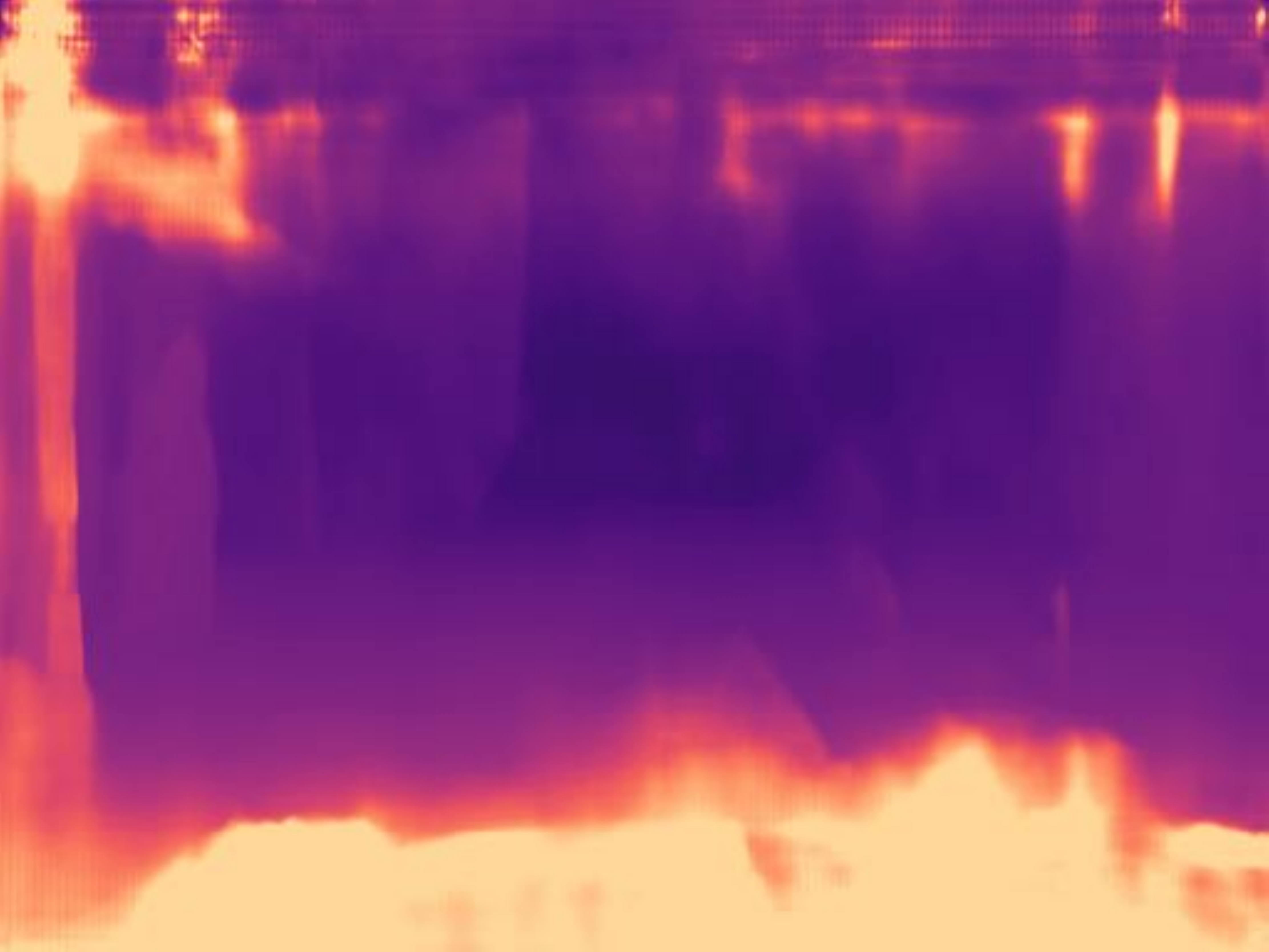}\qquad\qquad\quad &
\includegraphics[width=\iw,height=\ih]{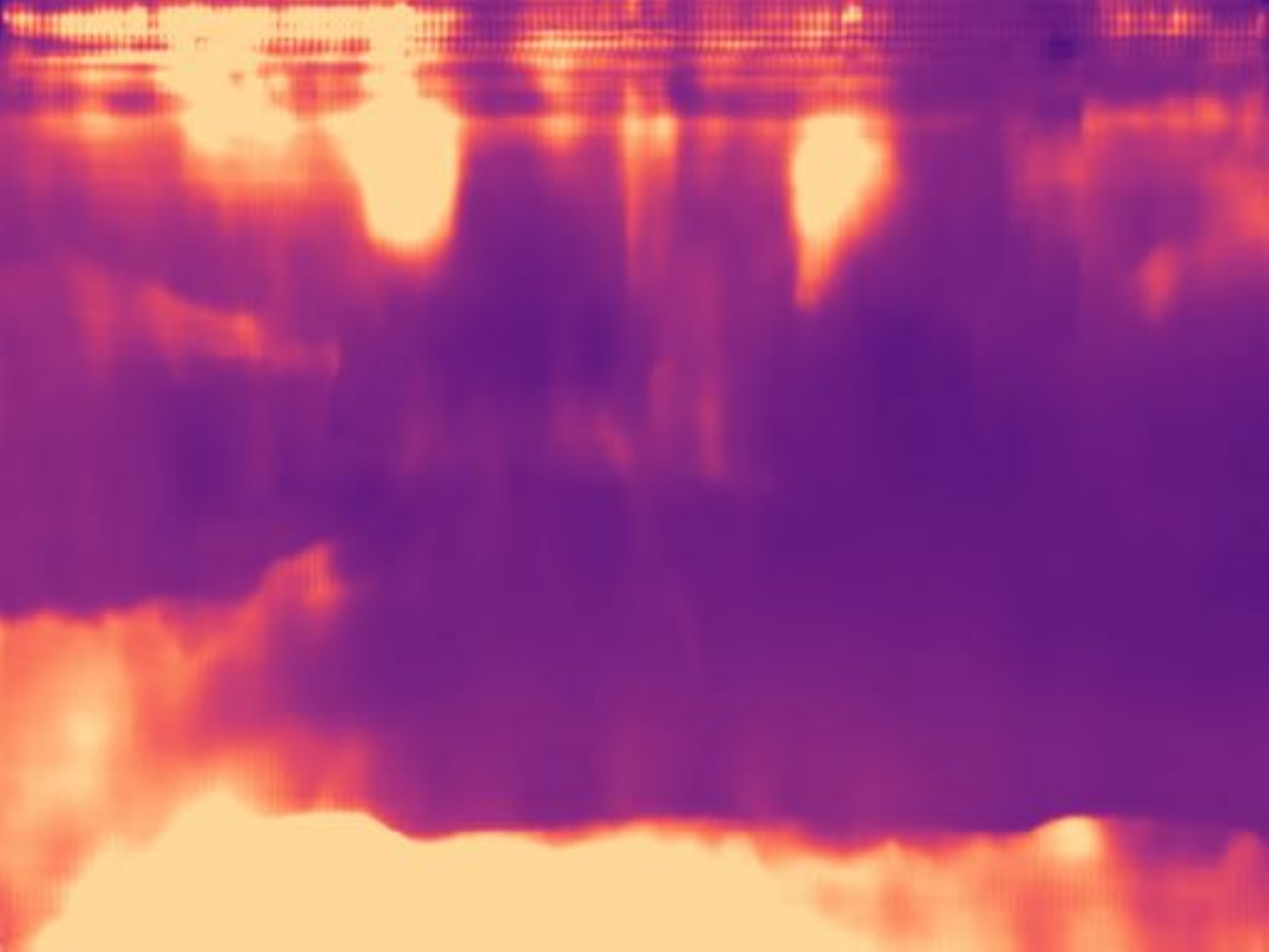}\qquad\qquad\quad &
\includegraphics[width=\iw,height=\ih]{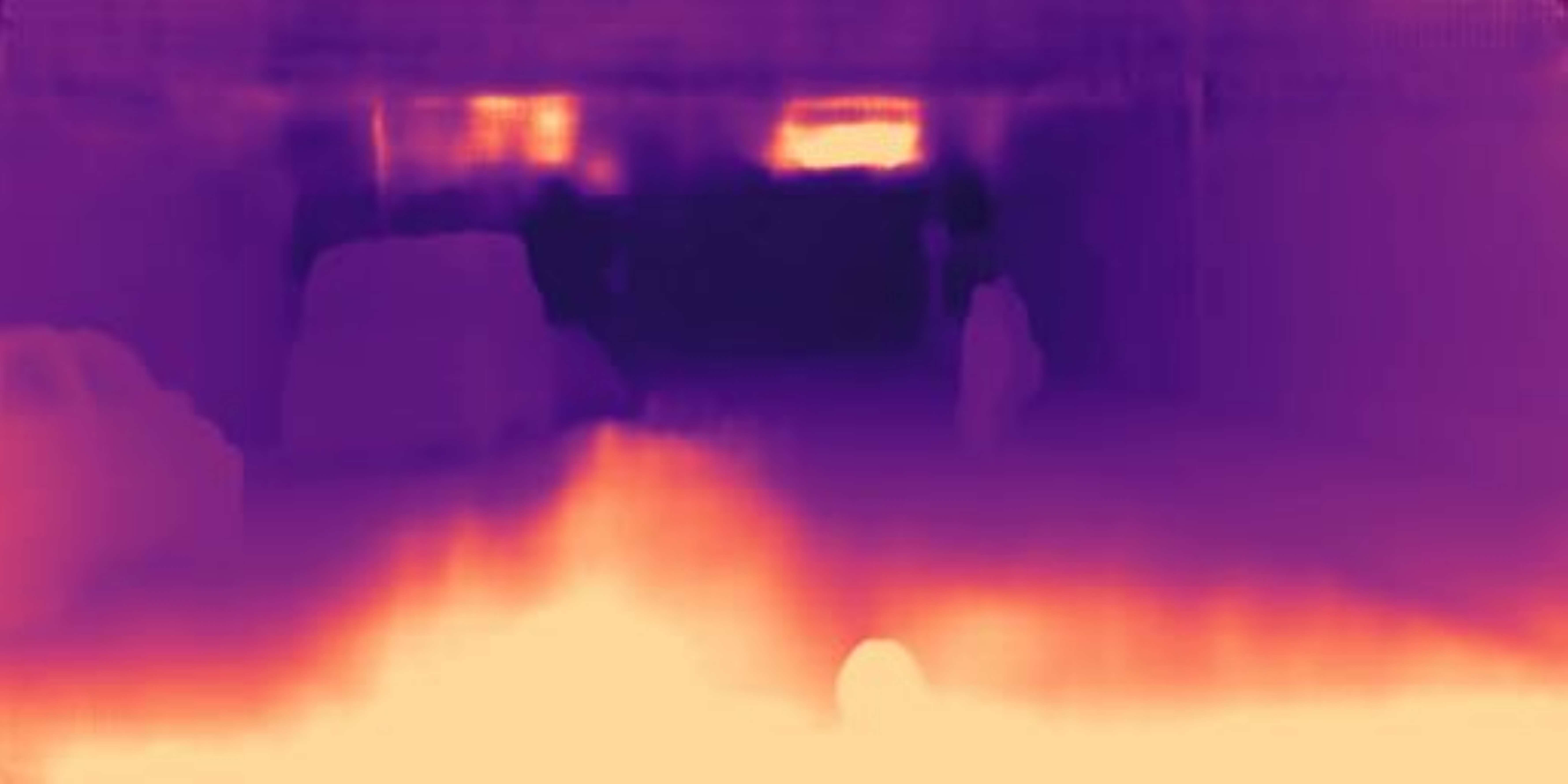}\qquad\qquad\quad &
\includegraphics[width=\iw,height=\ih]{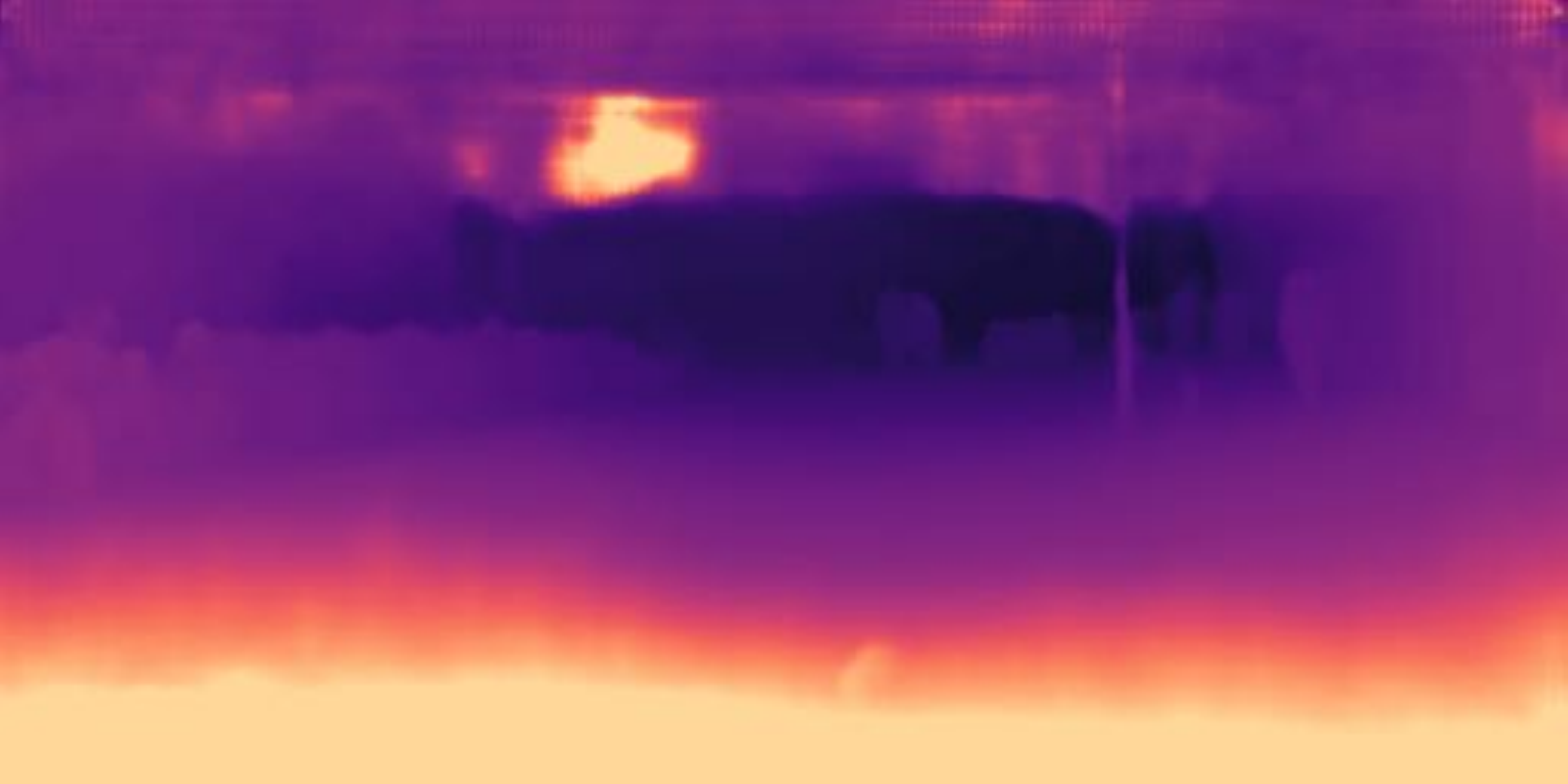}\qquad\qquad\quad &
\includegraphics[width=\iw,height=\ih]{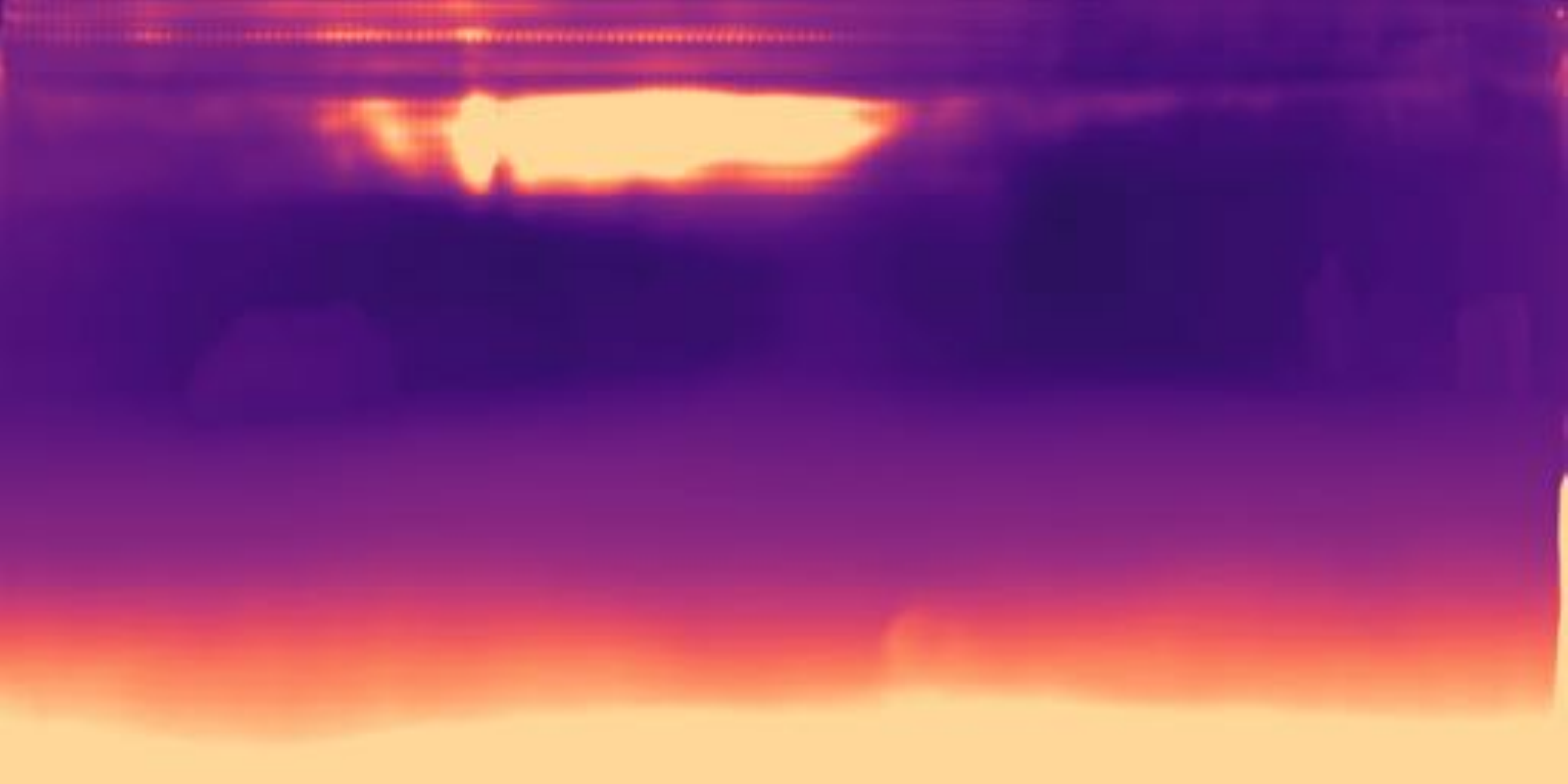}\qquad\qquad\quad &
\includegraphics[width=\iw,height=\ih]{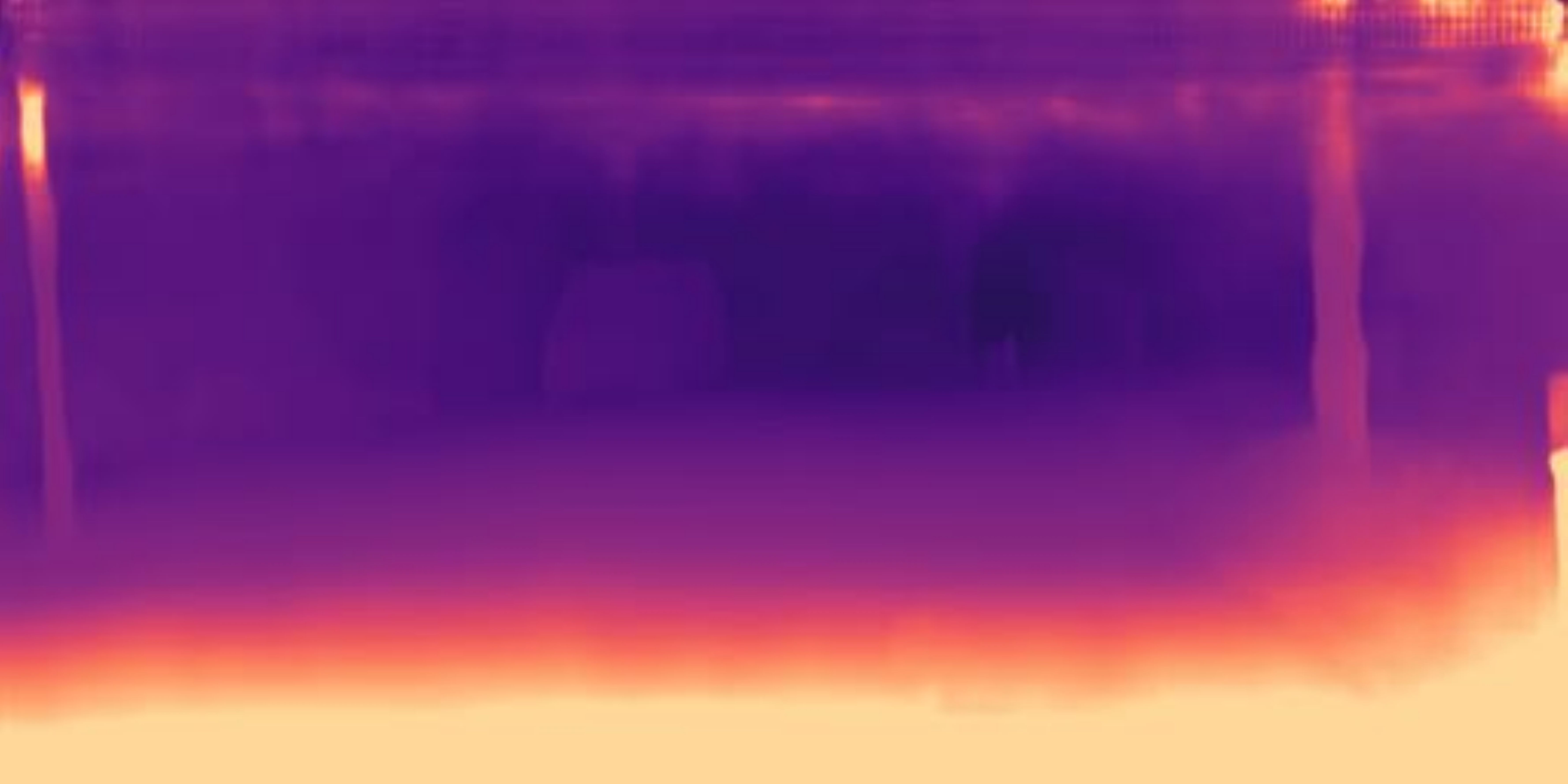}\\
\vspace{10mm} \\
\rotatebox[origin=c]{90}{\fontsize{\textw}{\texth}\selectfont Adabins\hspace{-320mm}}\hspace{15mm}
\includegraphics[width=\iw,height=\ih]{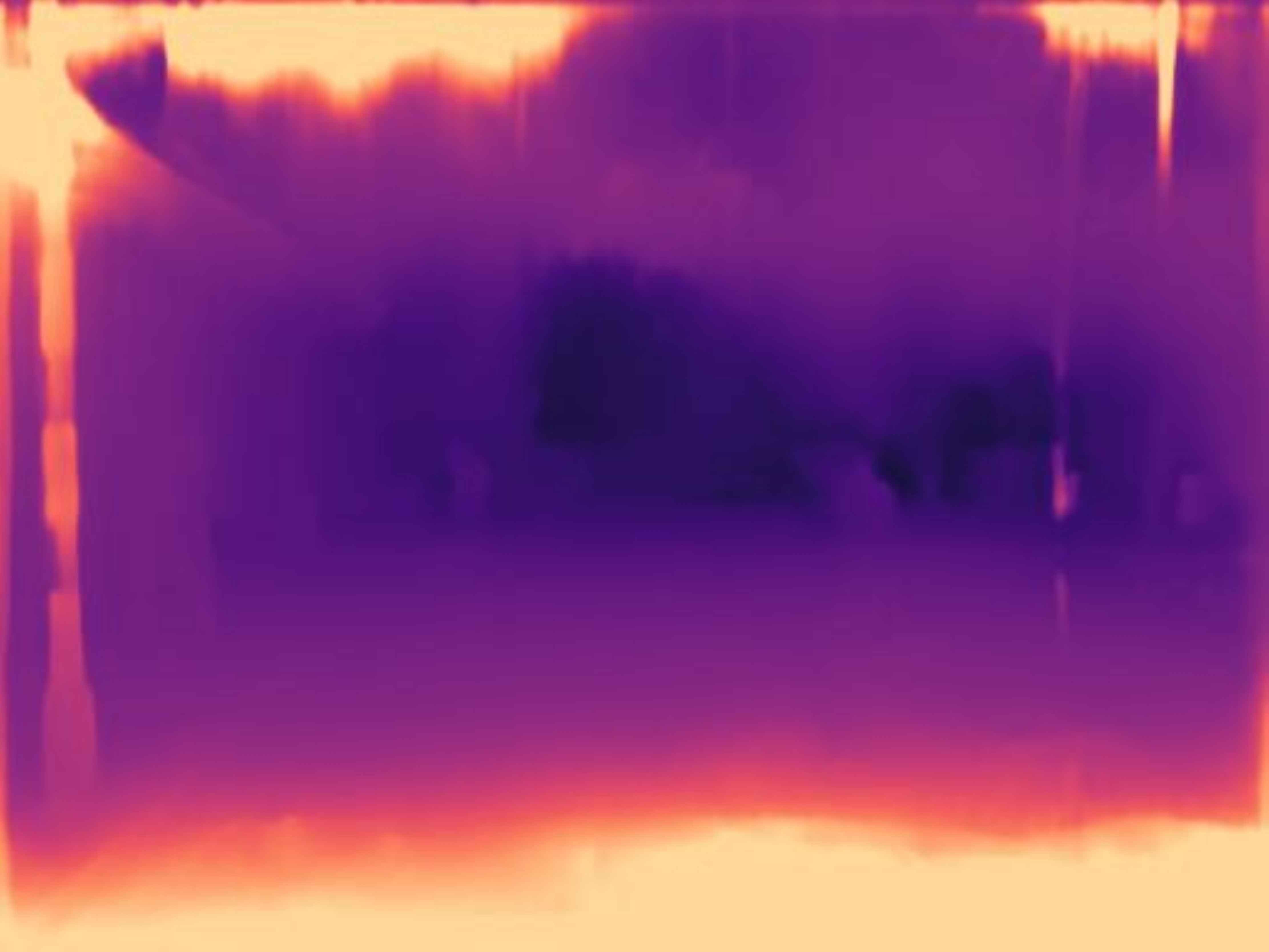}\qquad\qquad\quad &
\includegraphics[width=\iw,height=\ih]{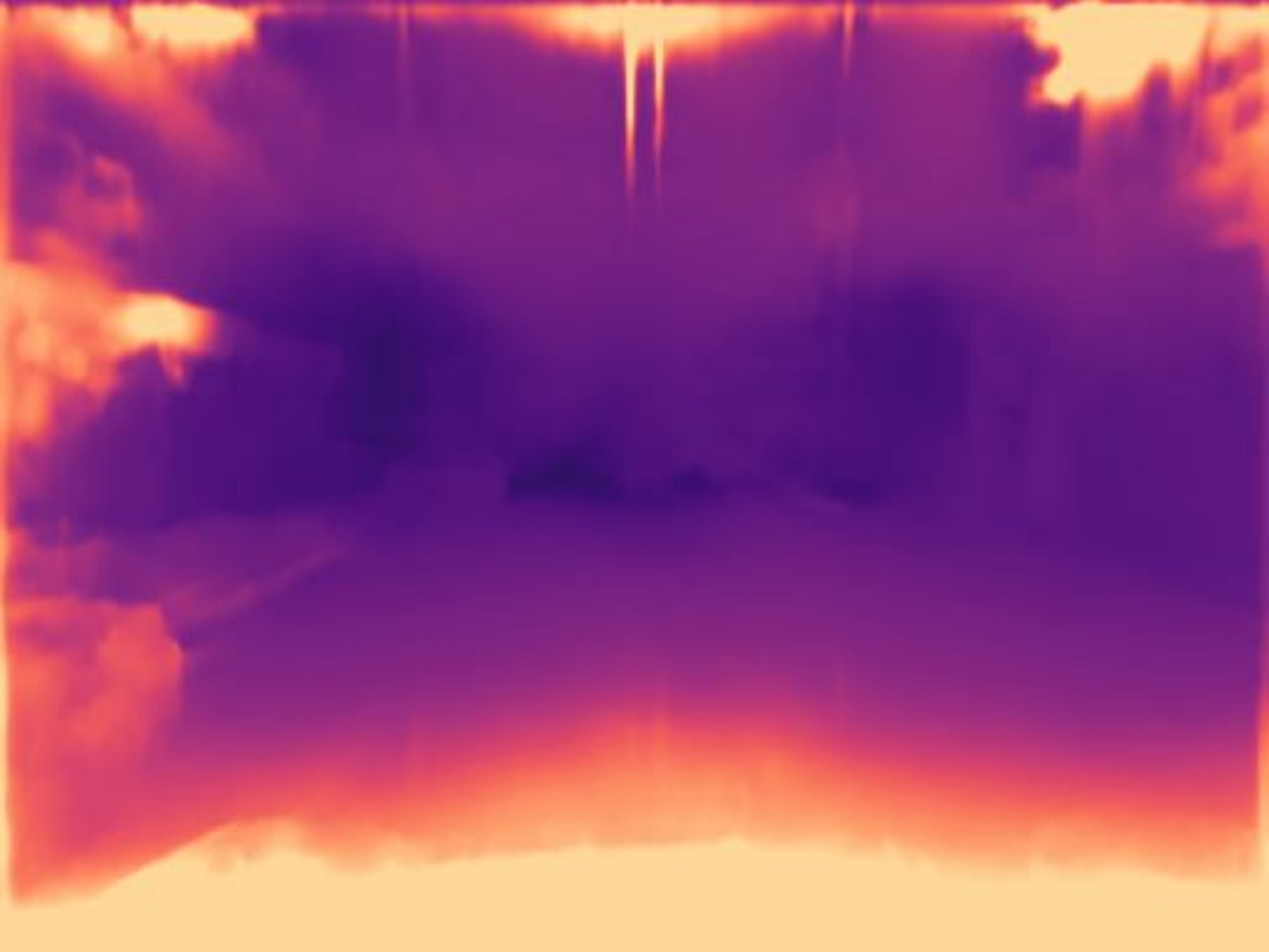}\qquad\qquad\quad &
\includegraphics[width=\iw,height=\ih]{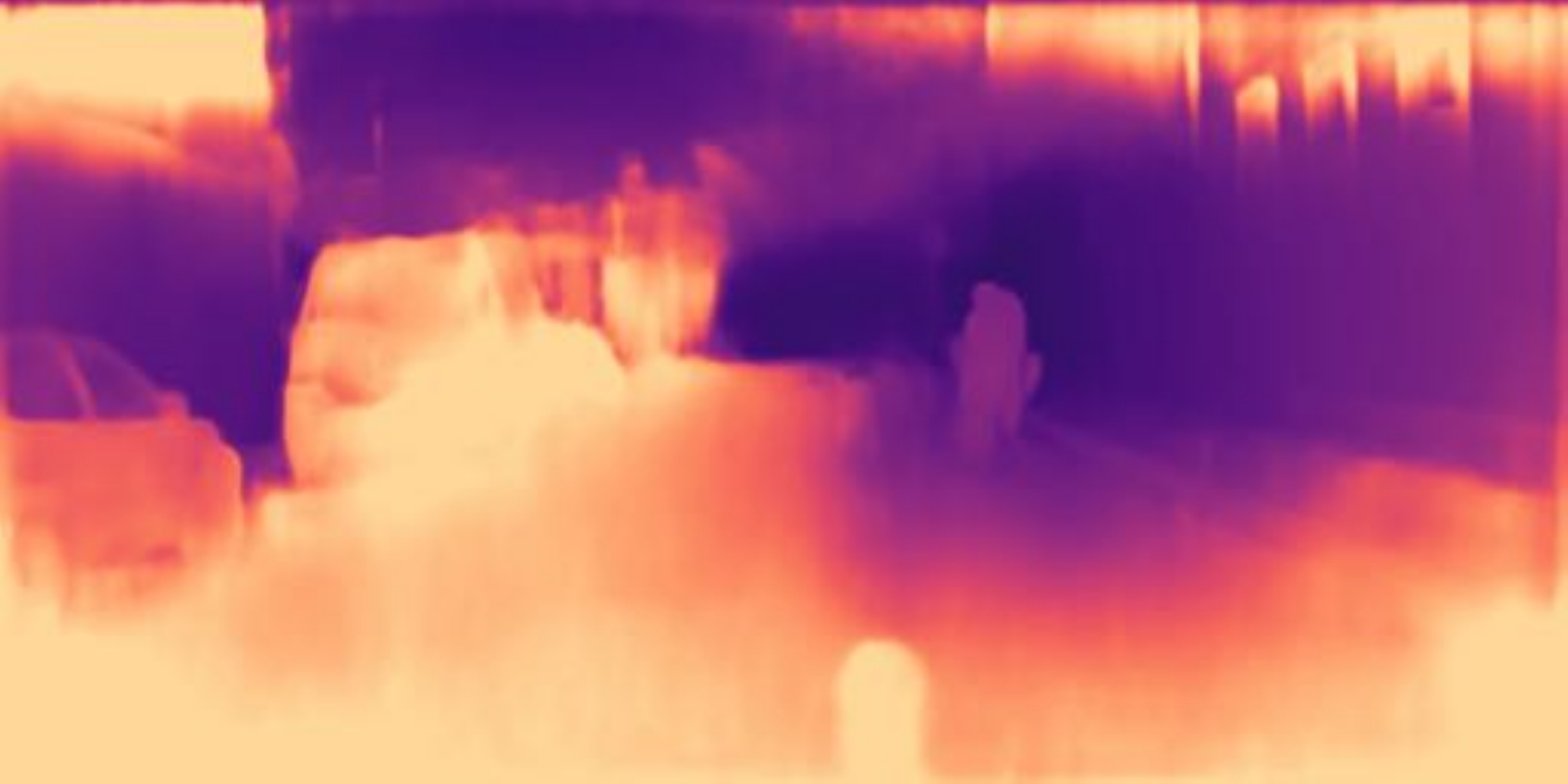}\qquad\qquad\quad &
\includegraphics[width=\iw,height=\ih]{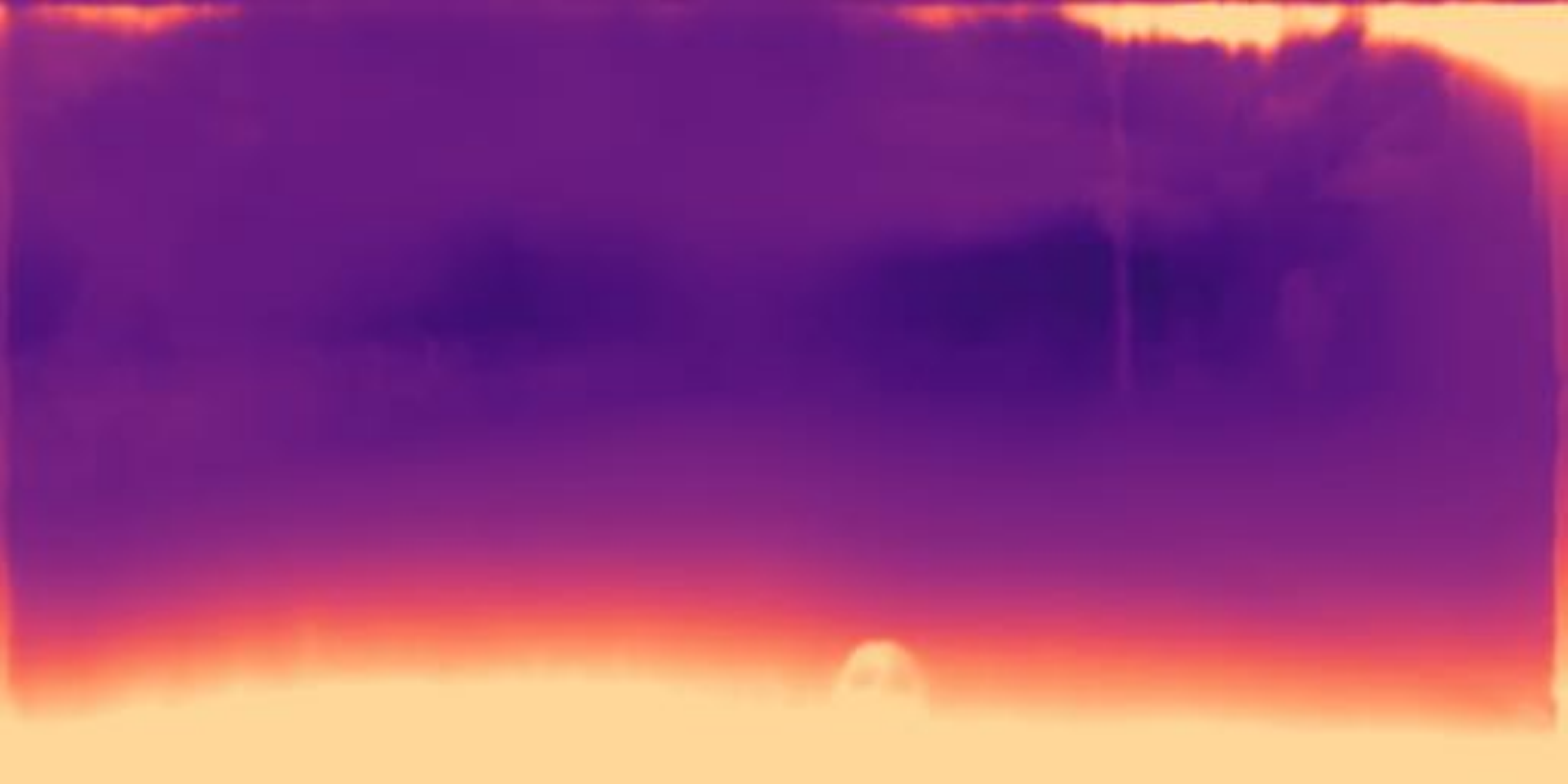}\qquad\qquad\quad &
\includegraphics[width=\iw,height=\ih]{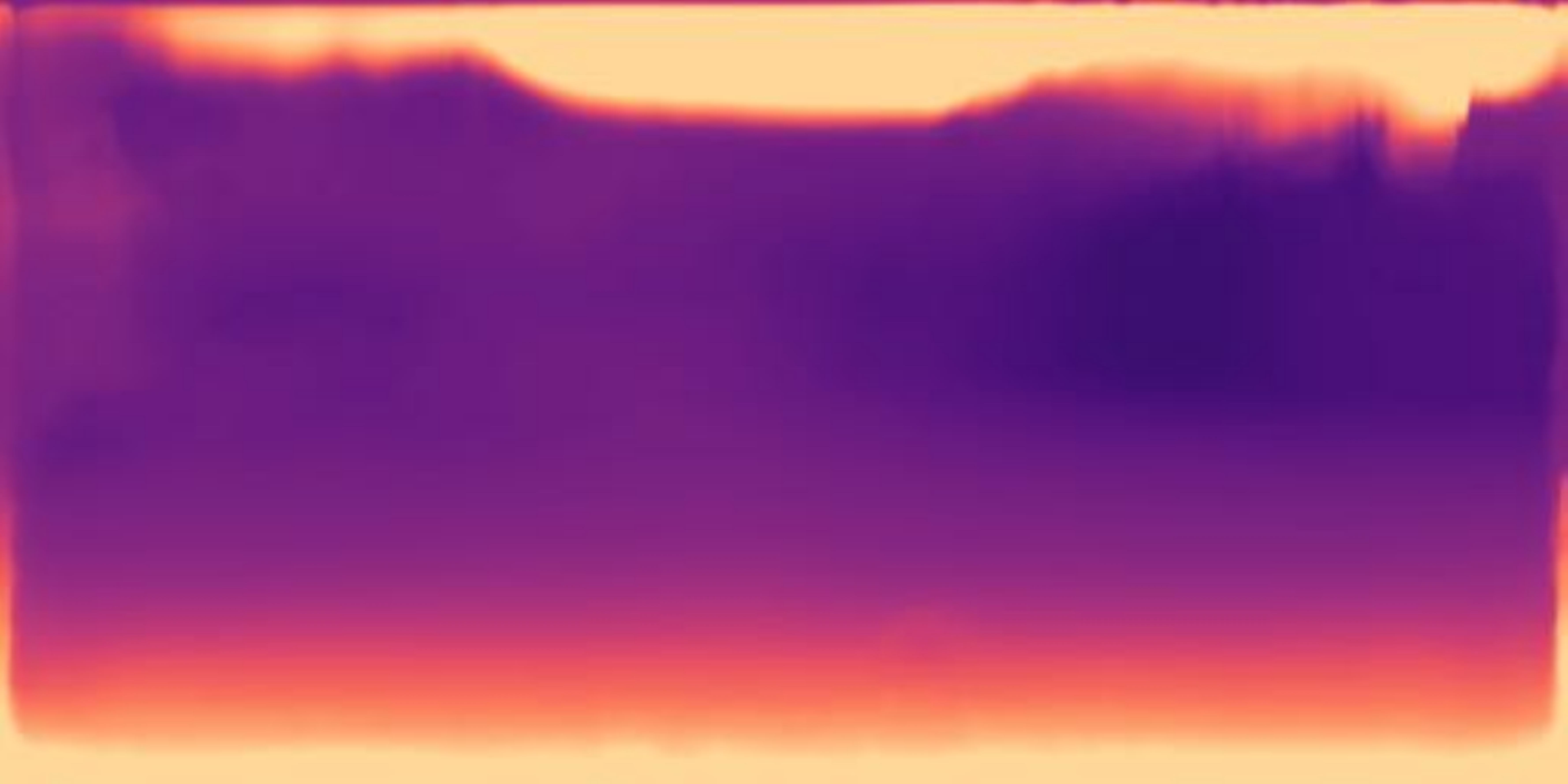}\qquad\qquad\quad &
\includegraphics[width=\iw,height=\ih]{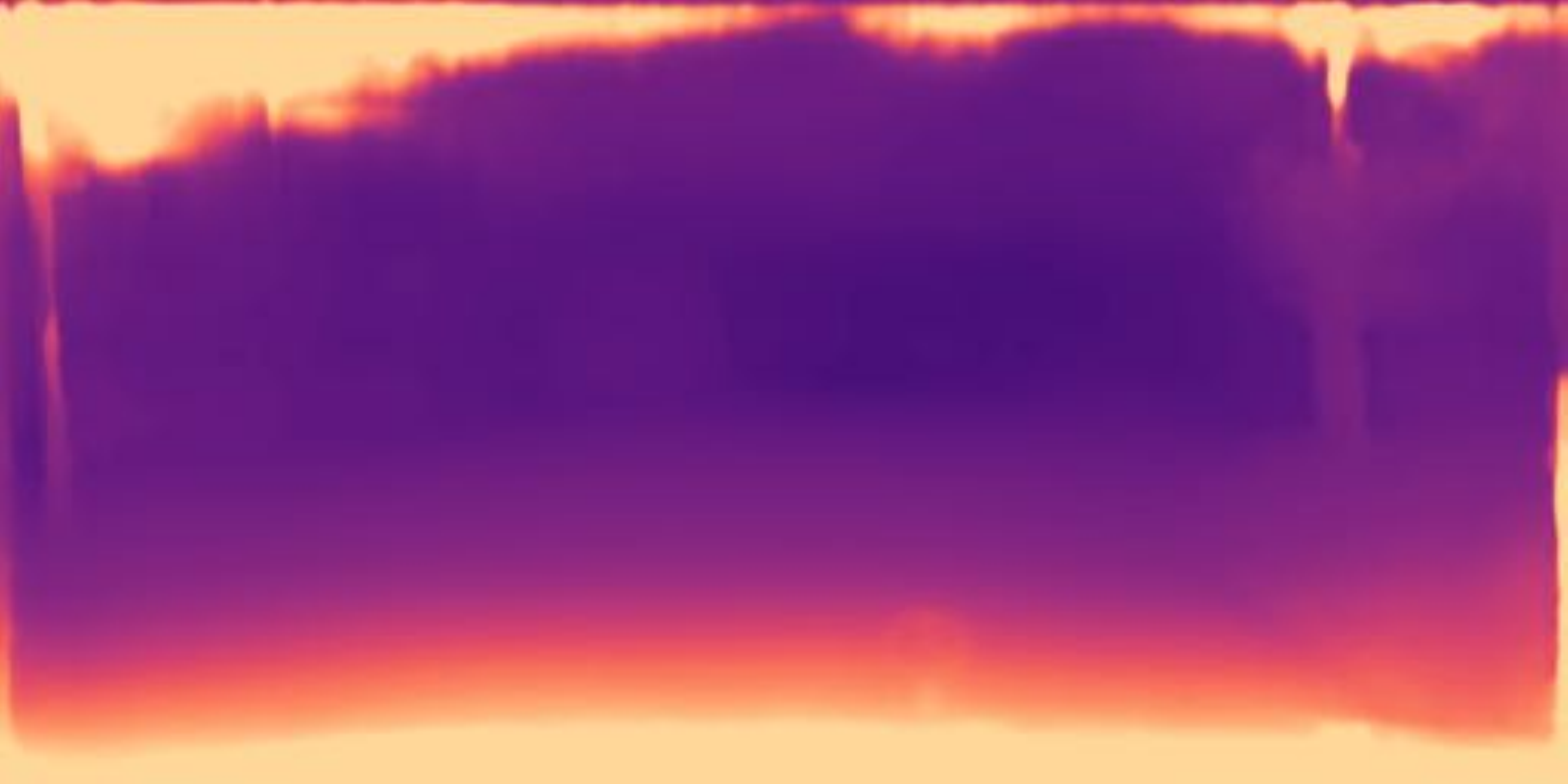}\\
\vspace{30mm} \\
\multicolumn{6}{c}{\fontsize{\w}{\h} \selectfont (c) Supervised CNN-based methods} \\
\vspace{30mm} \\
\rotatebox[origin=c]{90}{\fontsize{\textw}{\texth}\selectfont TransDepth\hspace{-300mm}}\hspace{15mm}
\includegraphics[width=\iw,height=\ih]{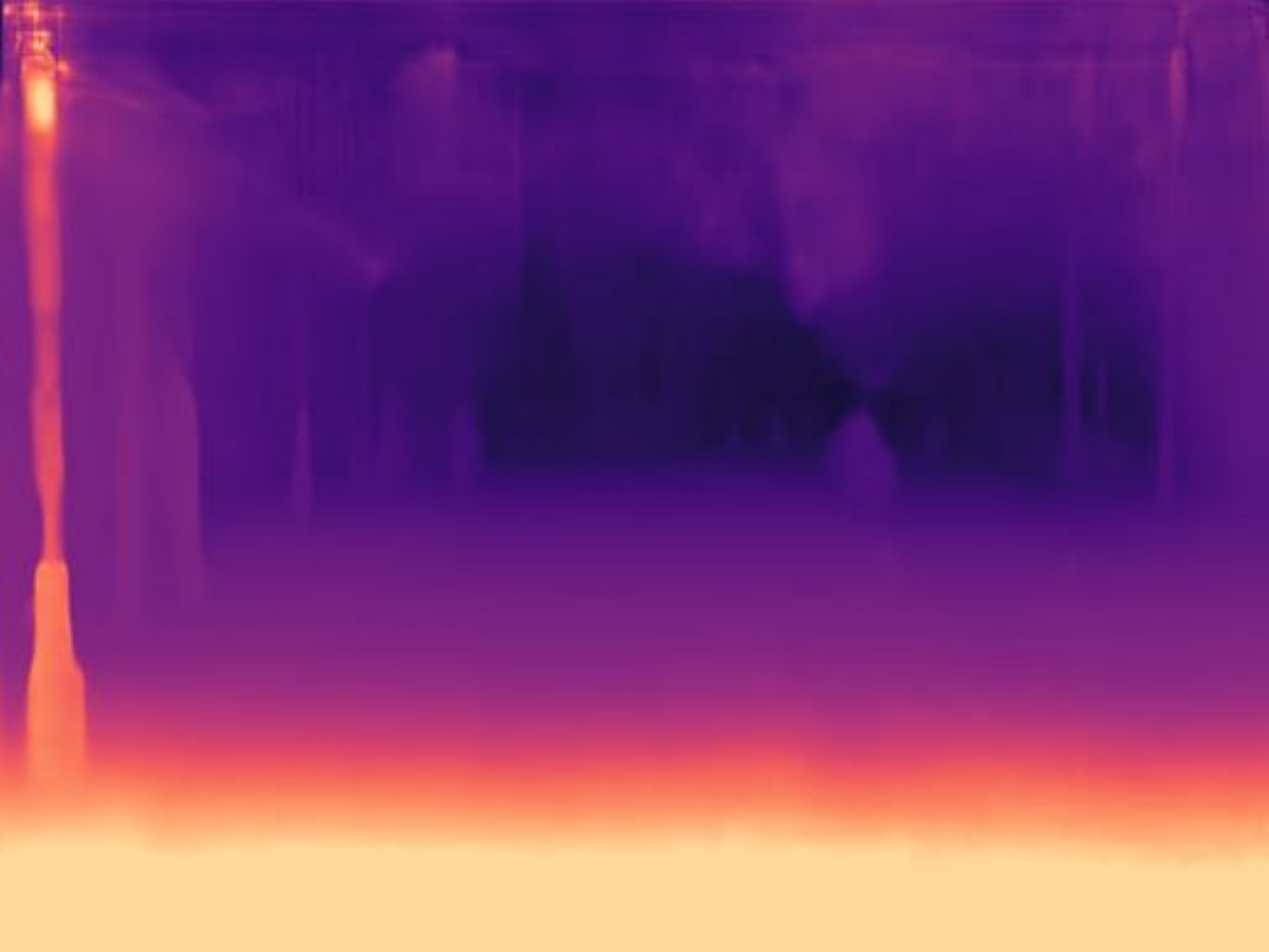}\qquad\qquad\quad &
\includegraphics[width=\iw,height=\ih]{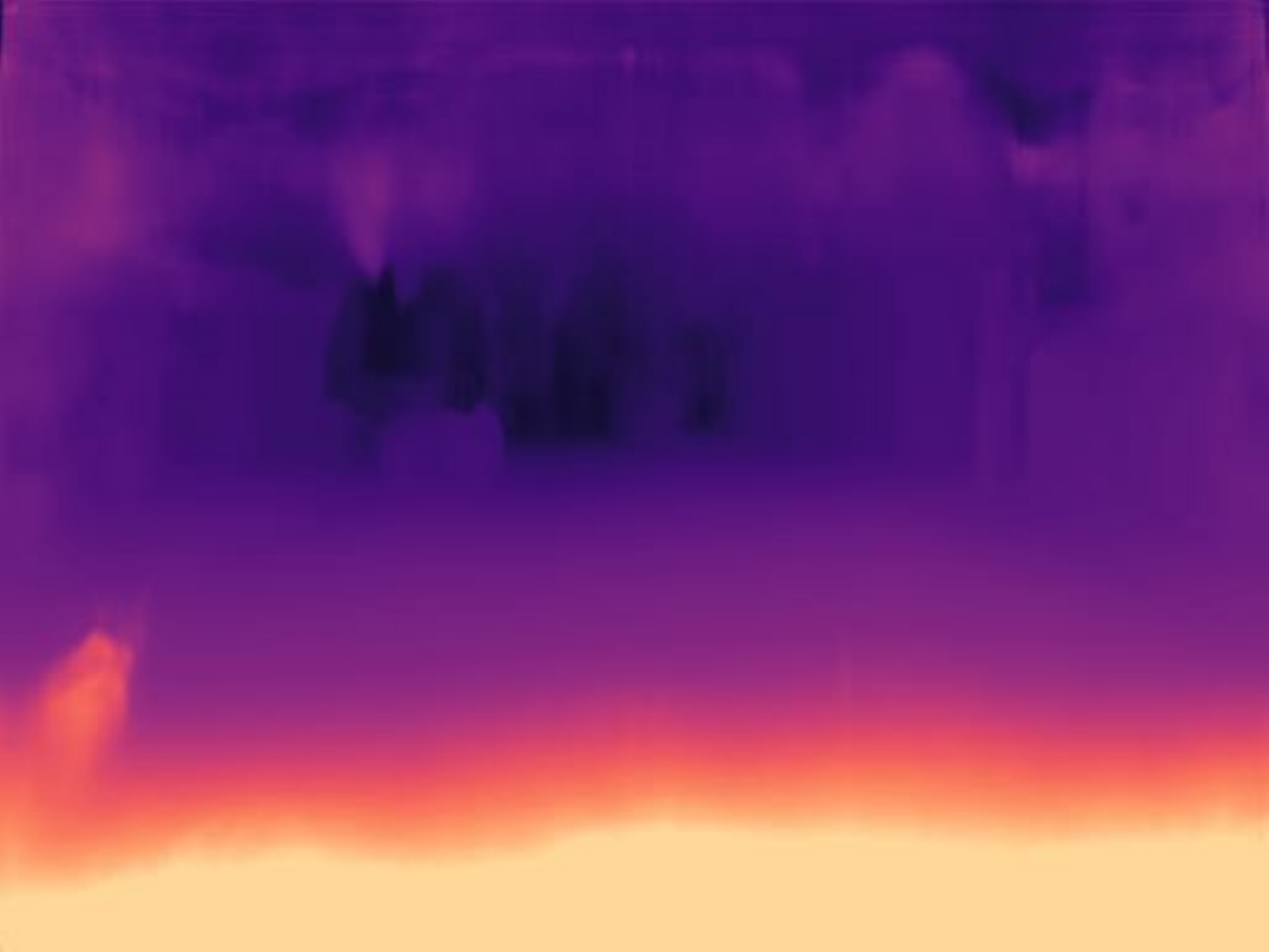}\qquad\qquad\quad &
\includegraphics[width=\iw,height=\ih]{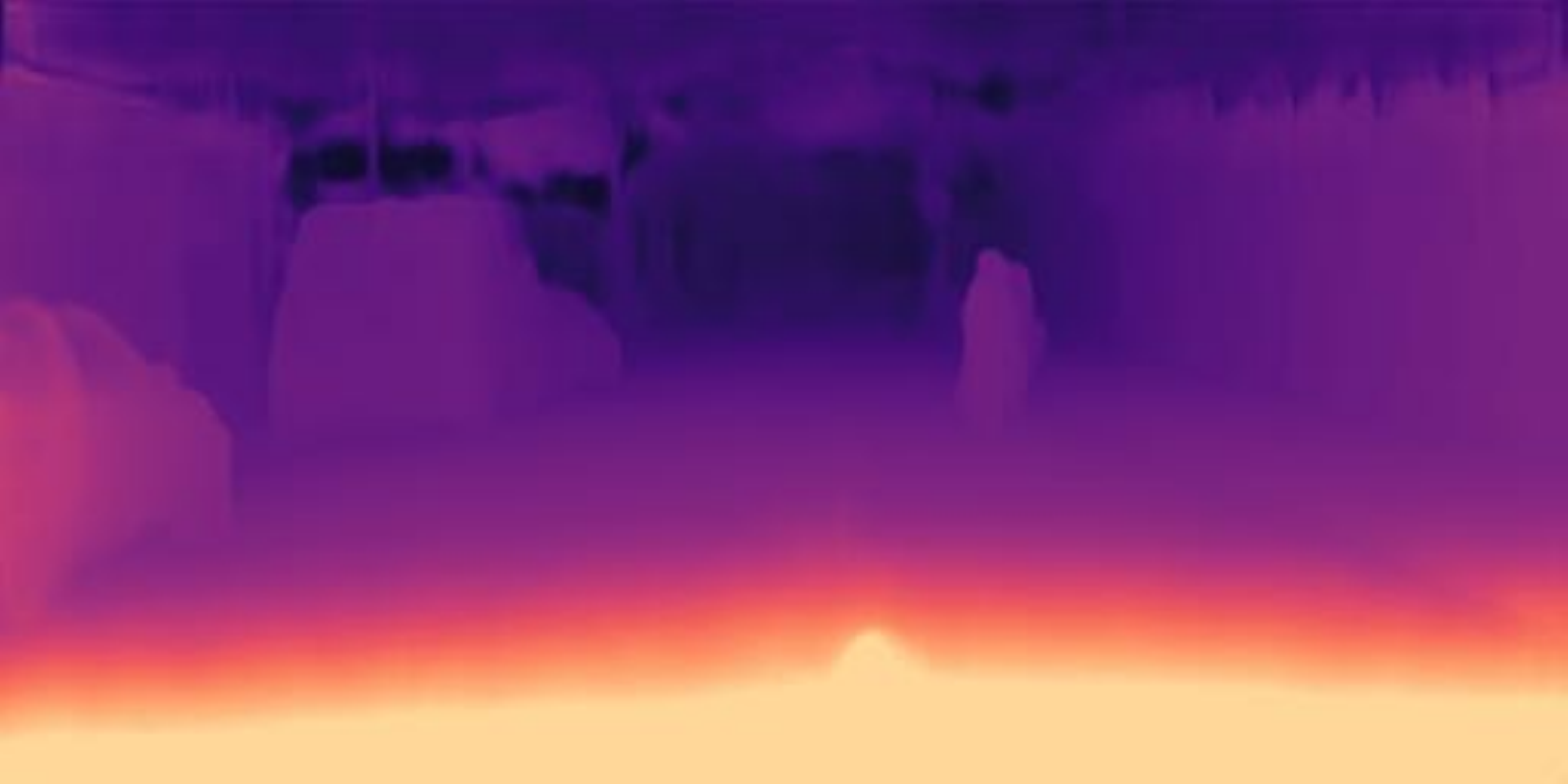}\qquad\qquad\quad &
\includegraphics[width=\iw,height=\ih]{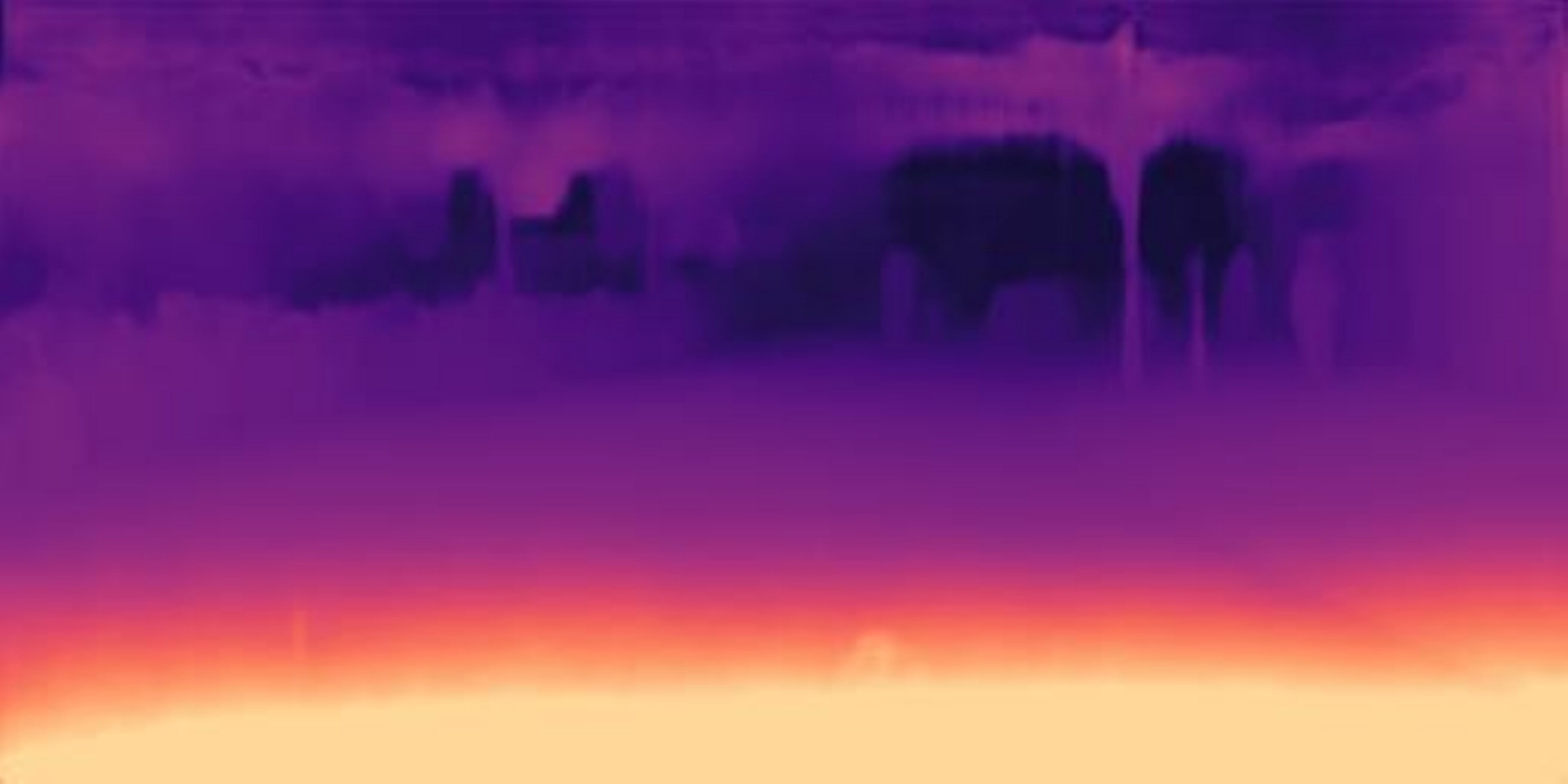}\qquad\qquad\quad &
\includegraphics[width=\iw,height=\ih]{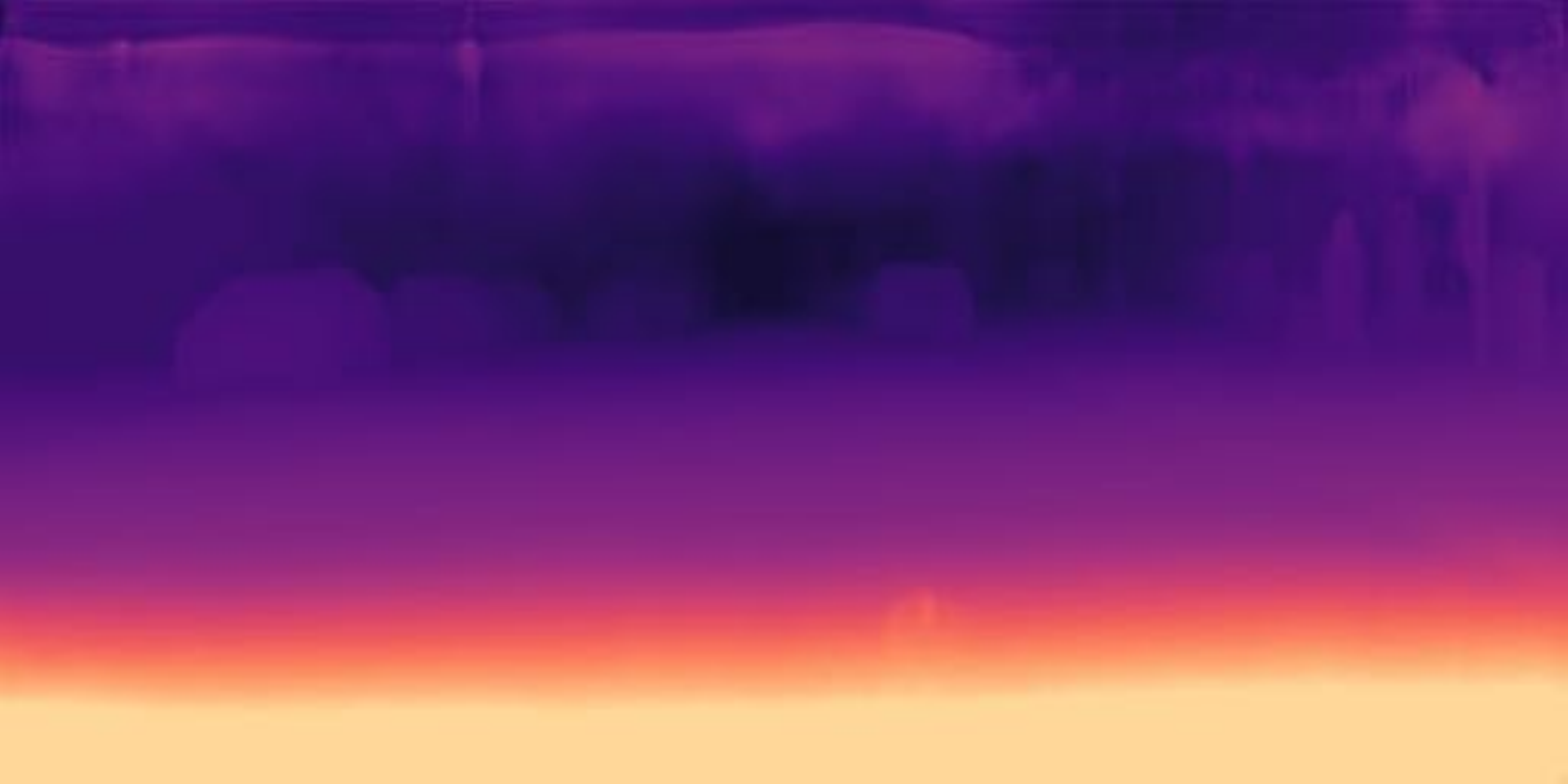}\qquad\qquad\quad &
\includegraphics[width=\iw,height=\ih]{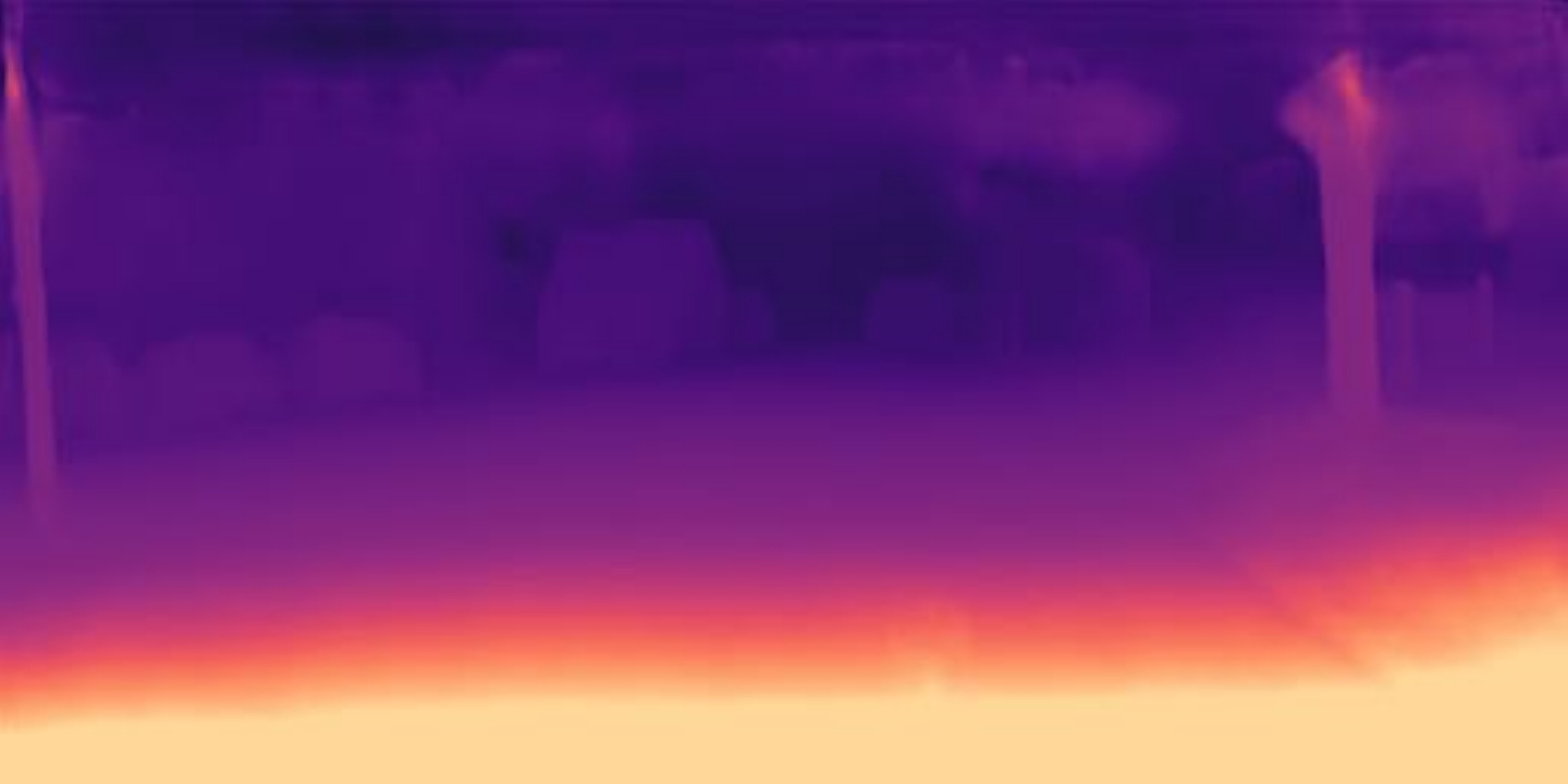}\\
\vspace{10mm} \\
\rotatebox[origin=c]{90}{\fontsize{\textw}{\texth}\selectfont DepthFormer\hspace{-300mm}}\hspace{15mm}
\includegraphics[width=\iw,height=\ih]{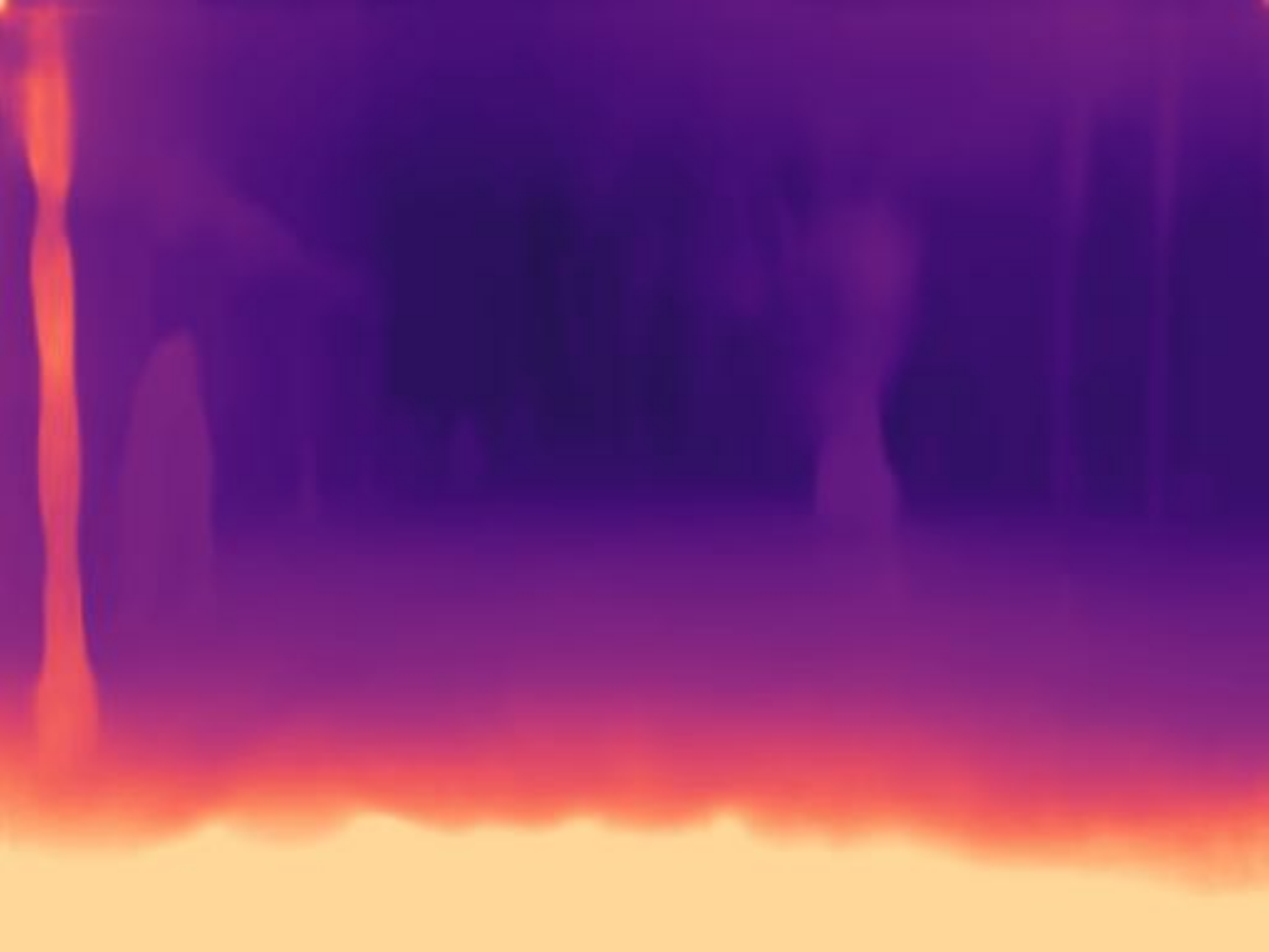}\qquad\qquad\quad &
\includegraphics[width=\iw,height=\ih]{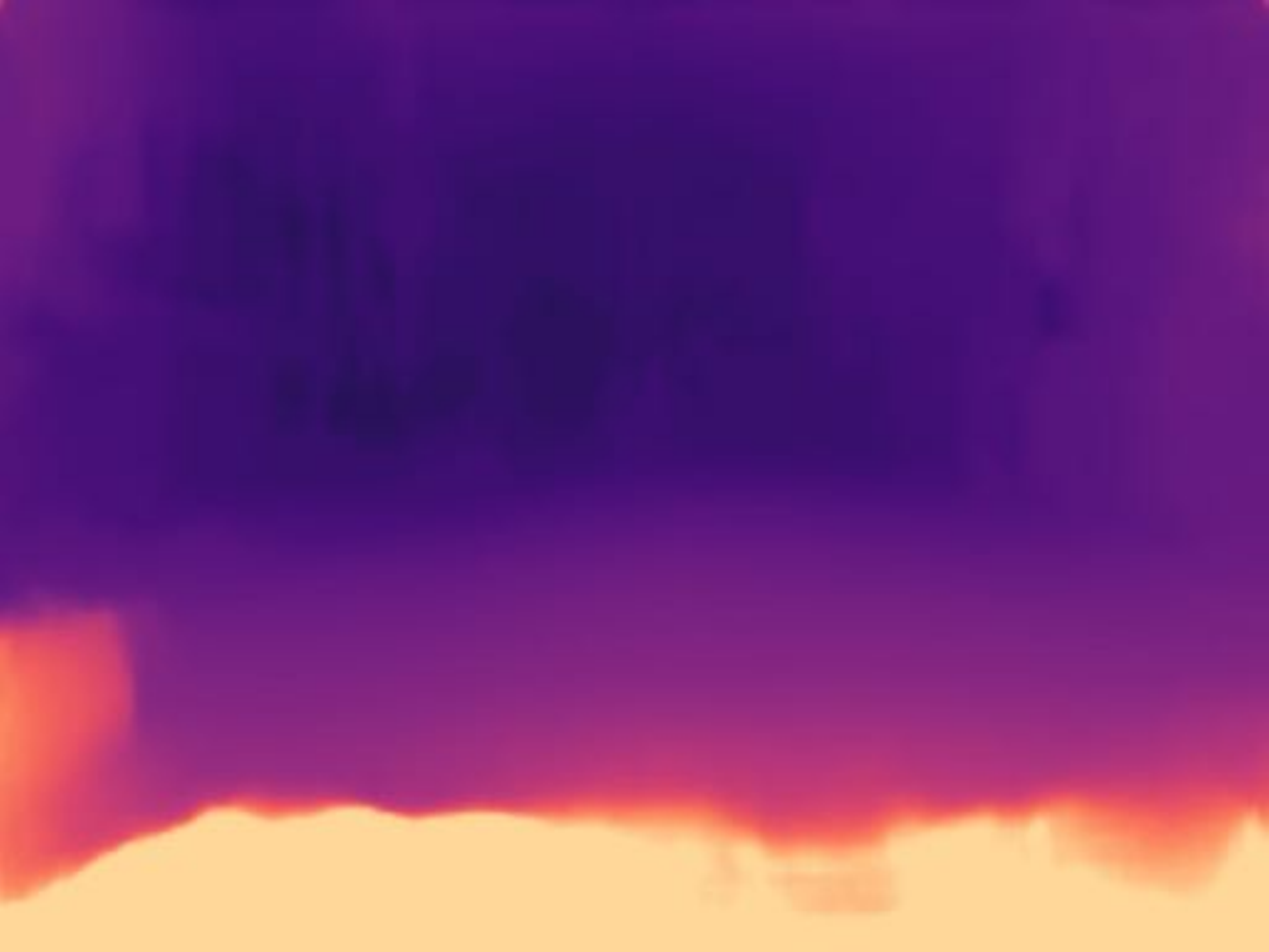}\qquad\qquad\quad &
\includegraphics[width=\iw,height=\ih]{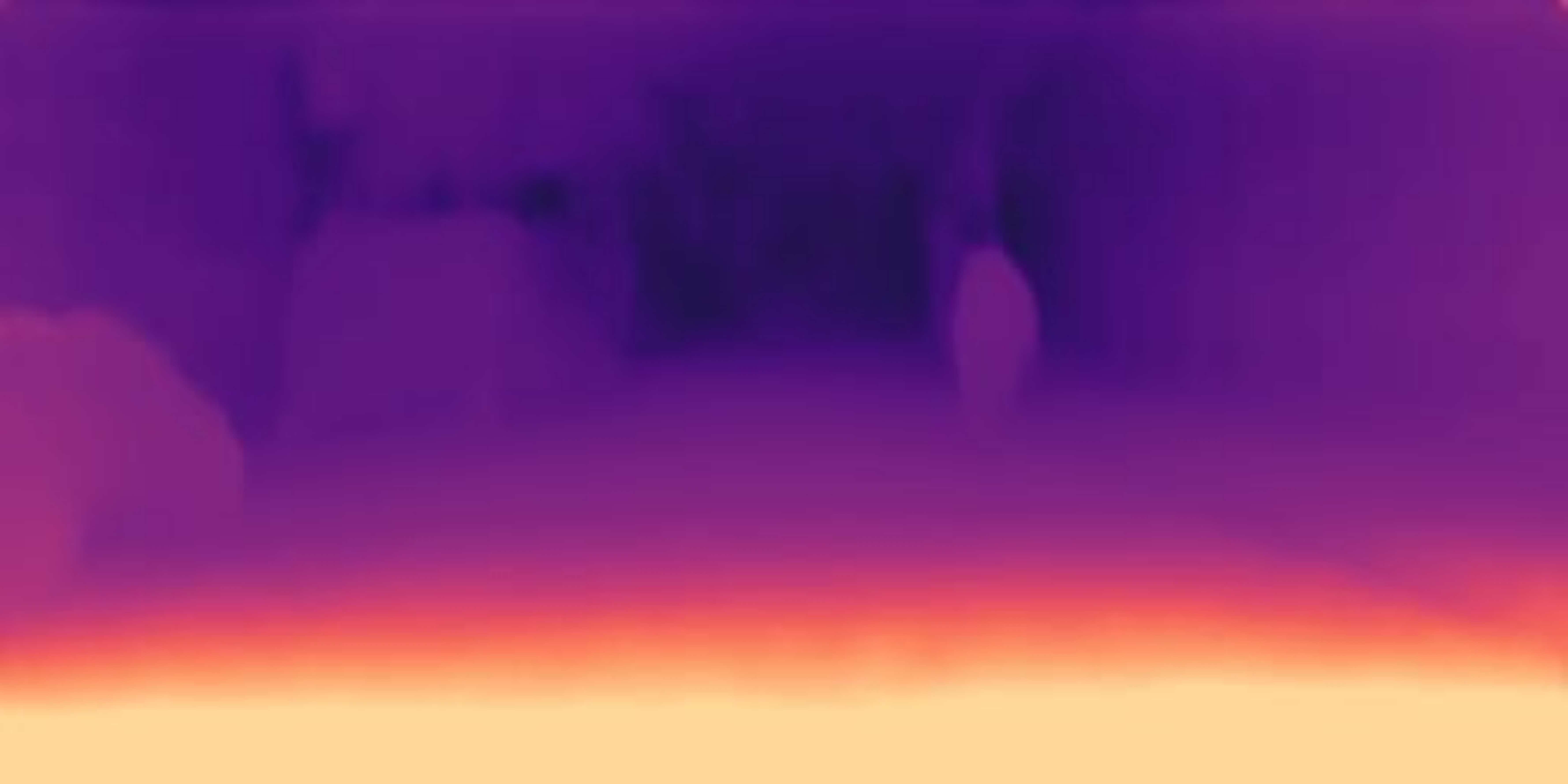}\qquad\qquad\quad &
\includegraphics[width=\iw,height=\ih]{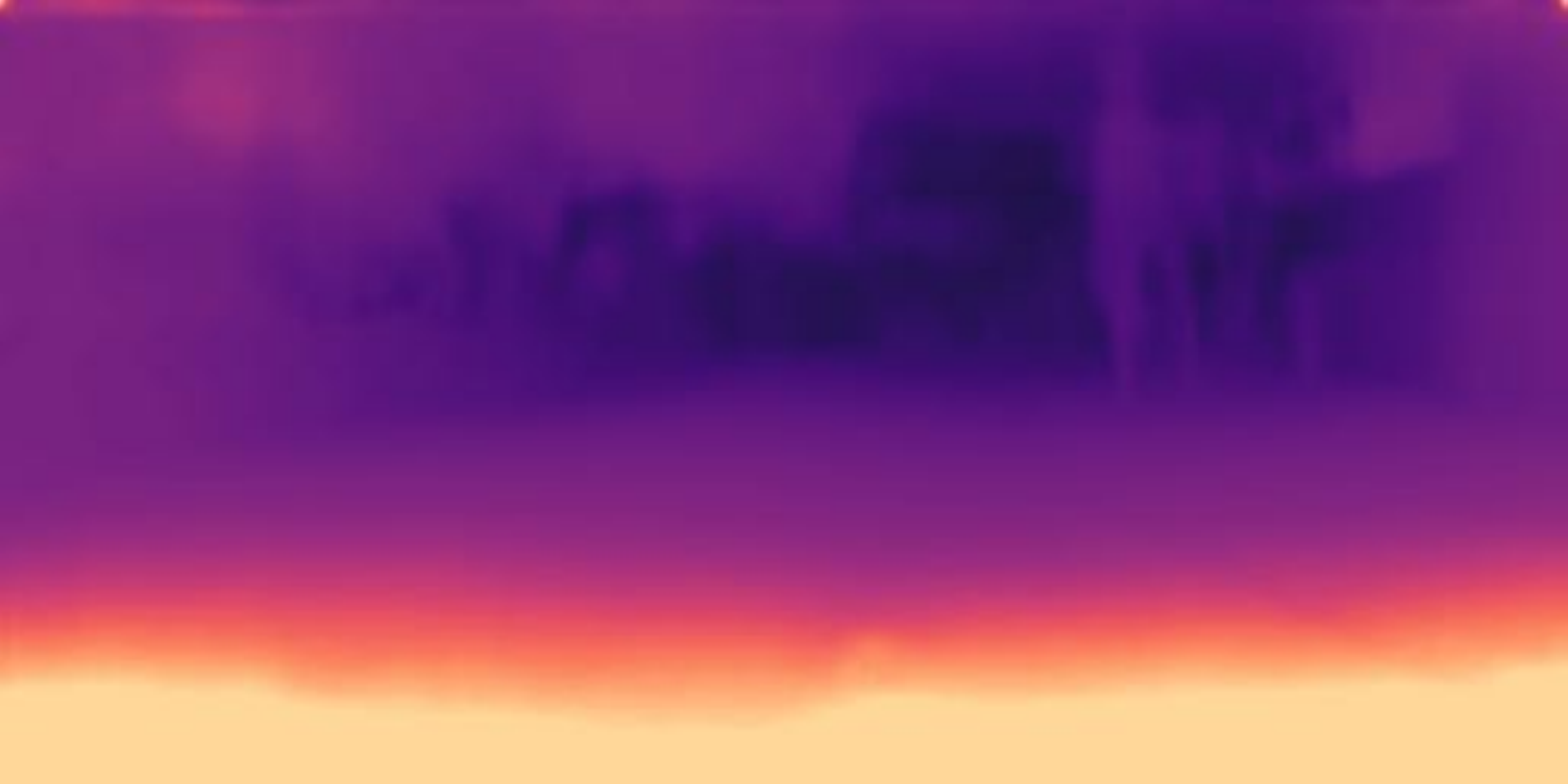}\qquad\qquad\quad &
\includegraphics[width=\iw,height=\ih]{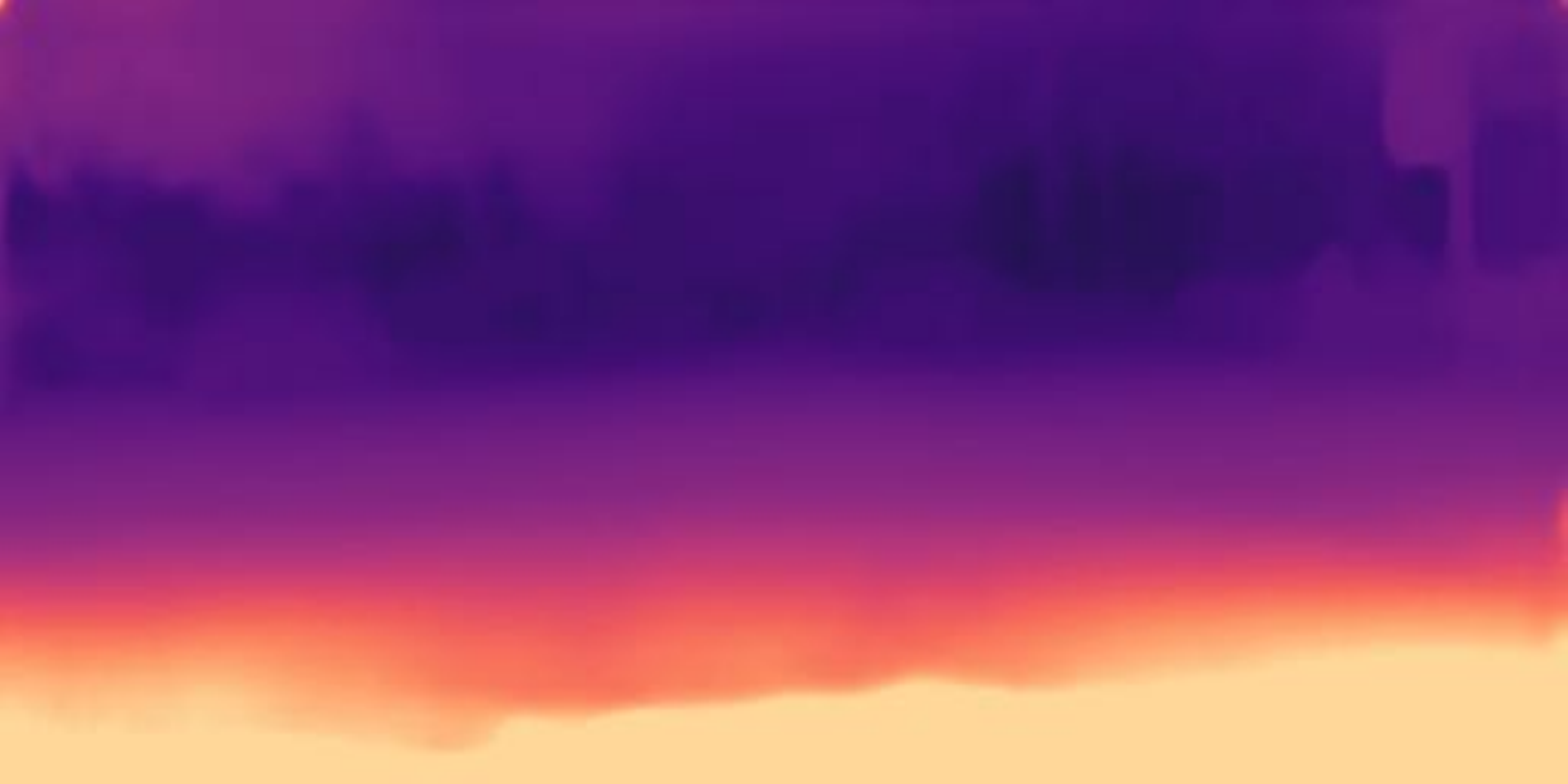}\qquad\qquad\quad &
\includegraphics[width=\iw,height=\ih]{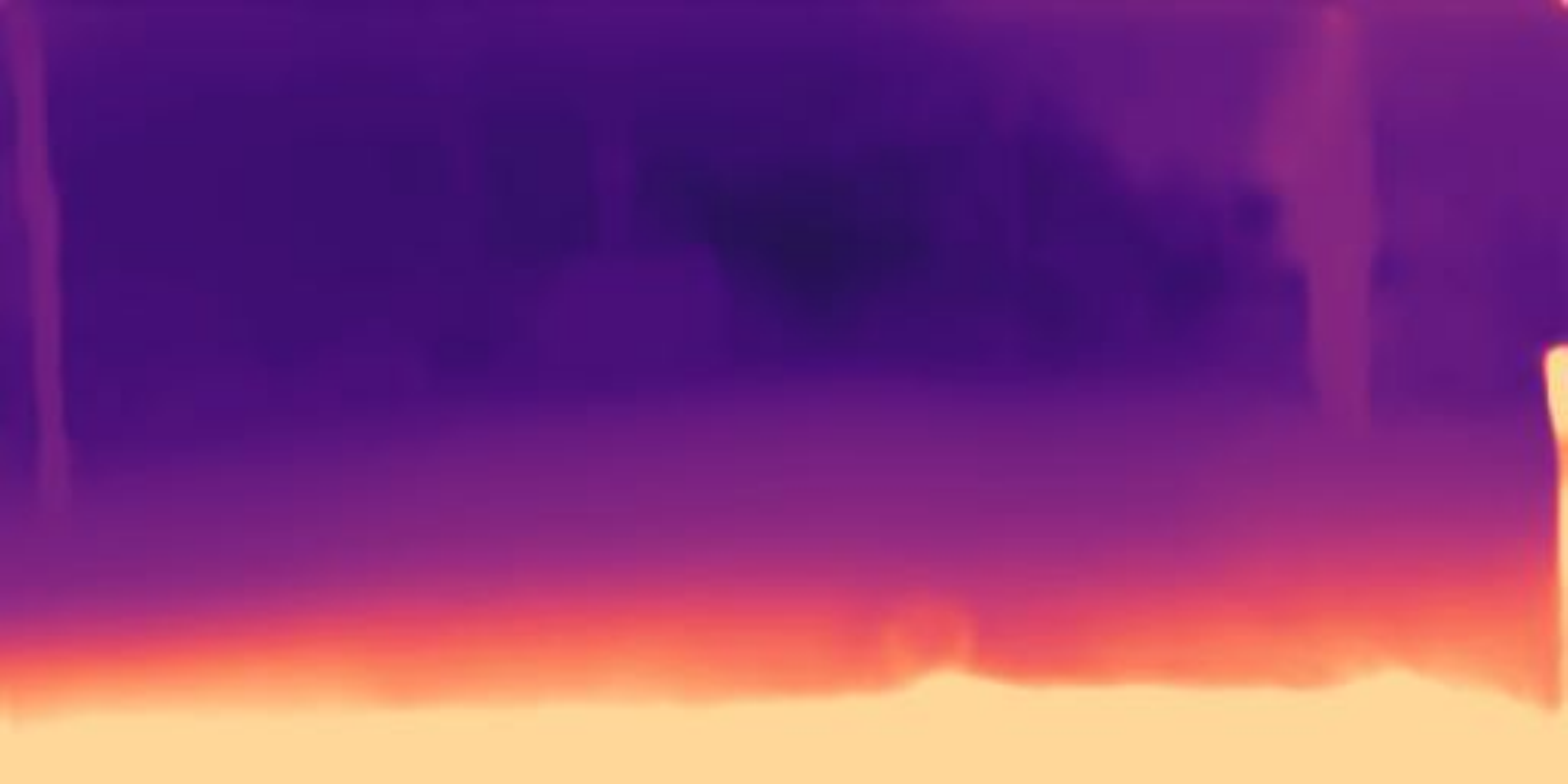}\\
\vspace{10mm} \\
\rotatebox[origin=c]{90}{\fontsize{\textw}{\texth}\selectfont GLPDepth\hspace{-310mm}}\hspace{15mm}
\includegraphics[width=\iw,height=\ih]{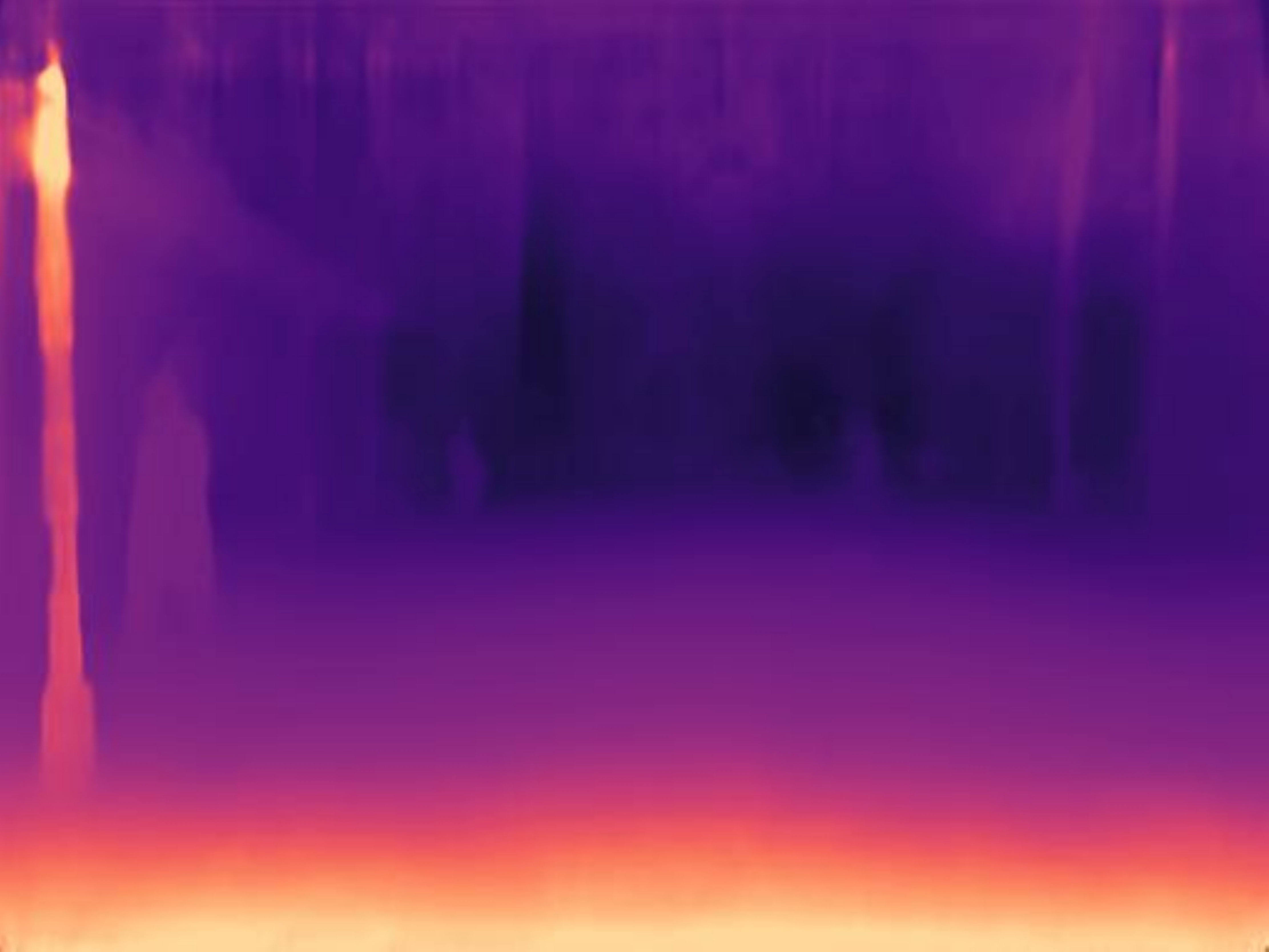}\qquad\qquad\quad &
\includegraphics[width=\iw,height=\ih]{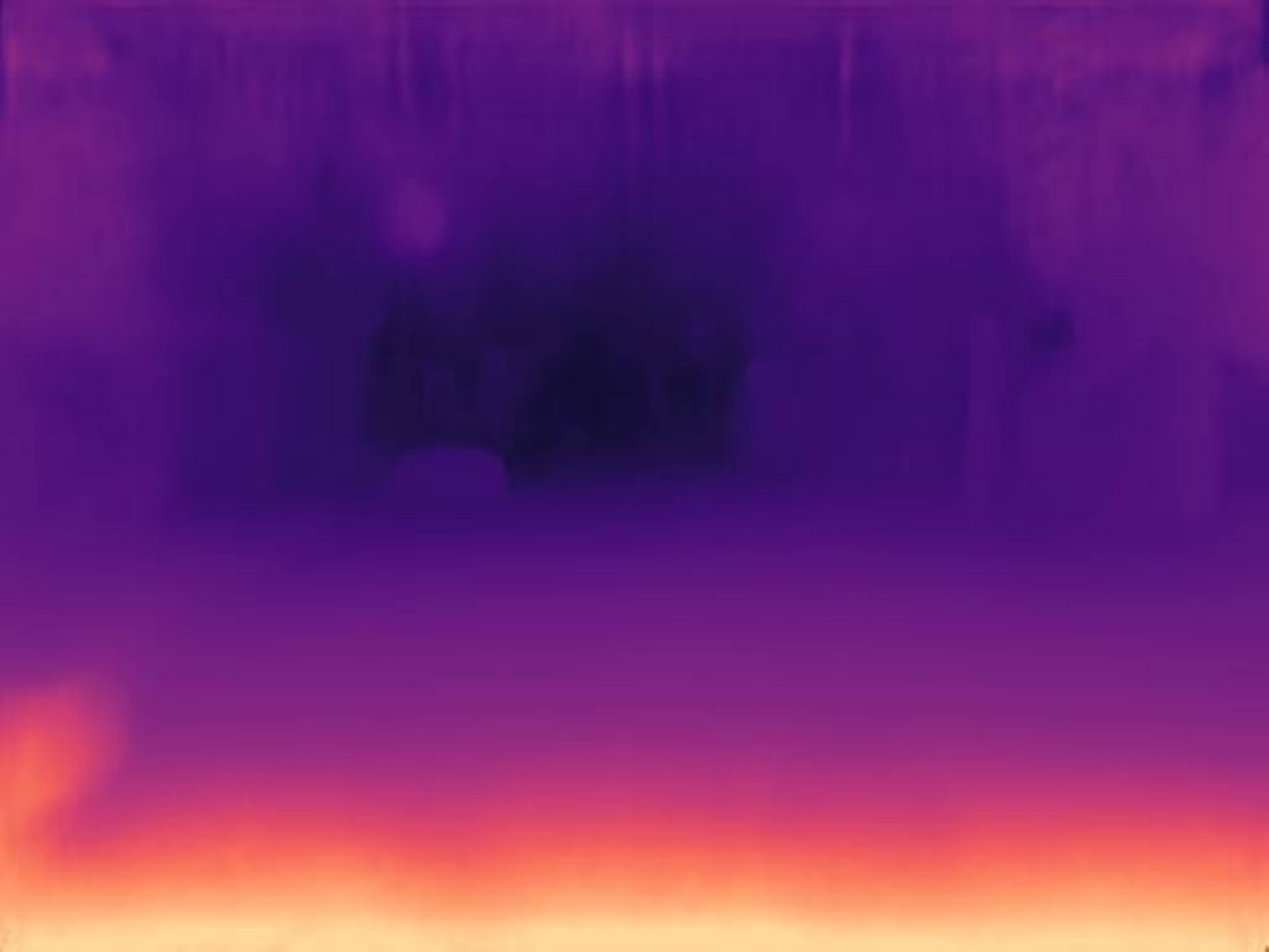}\qquad\qquad\quad &
\includegraphics[width=\iw,height=\ih]{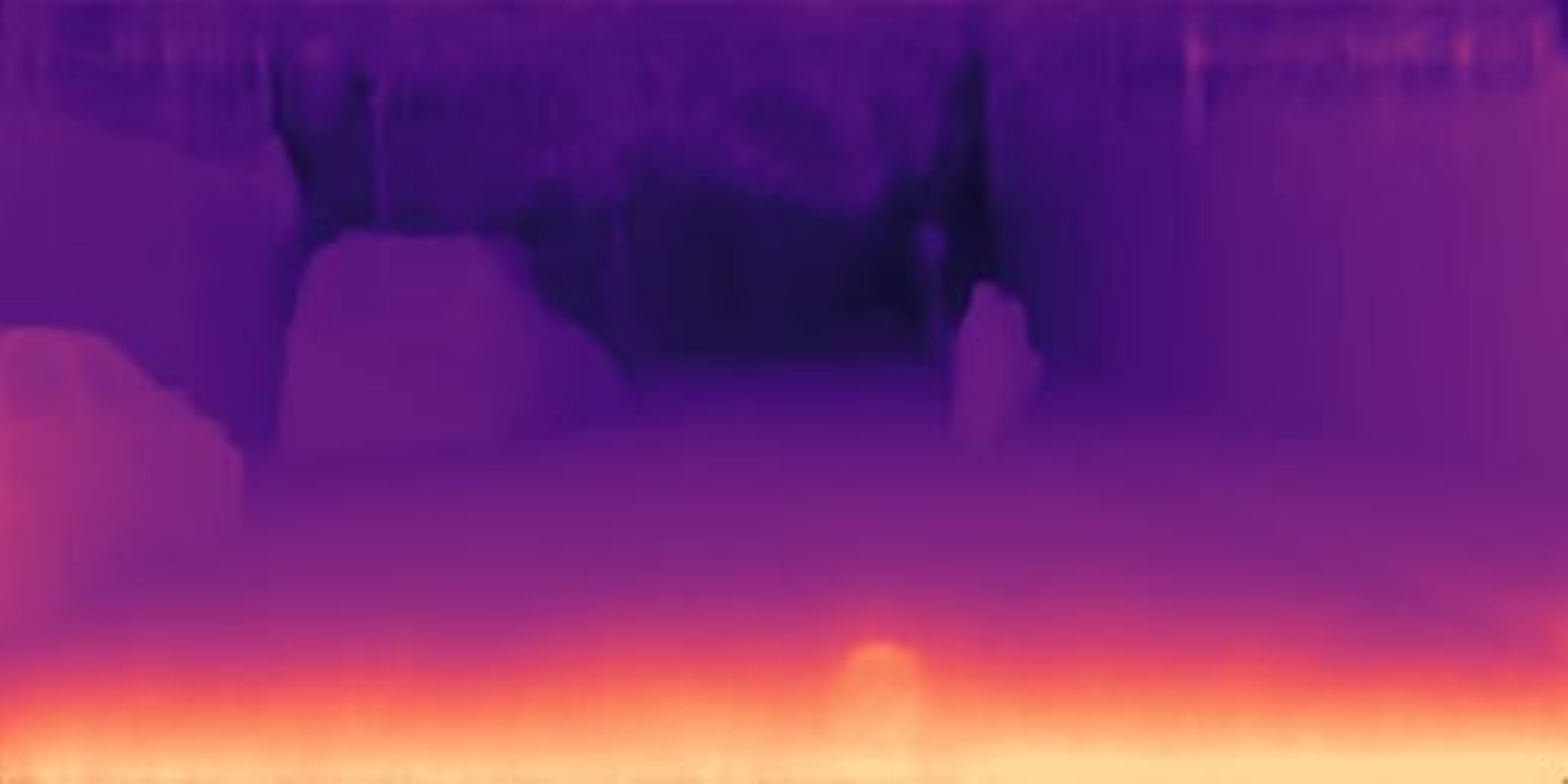}\qquad\qquad\quad &
\includegraphics[width=\iw,height=\ih]{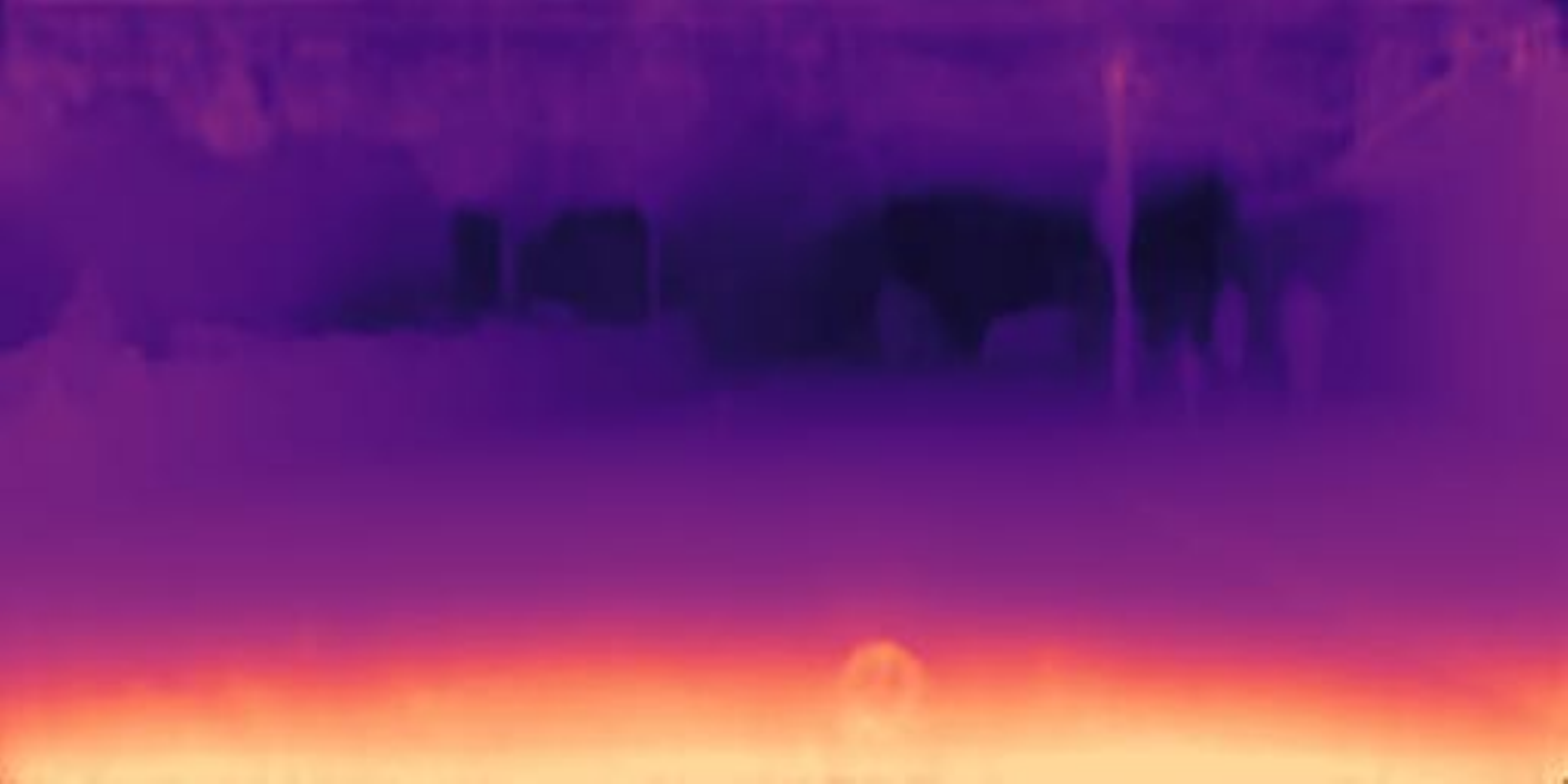}\qquad\qquad\quad &
\includegraphics[width=\iw,height=\ih]{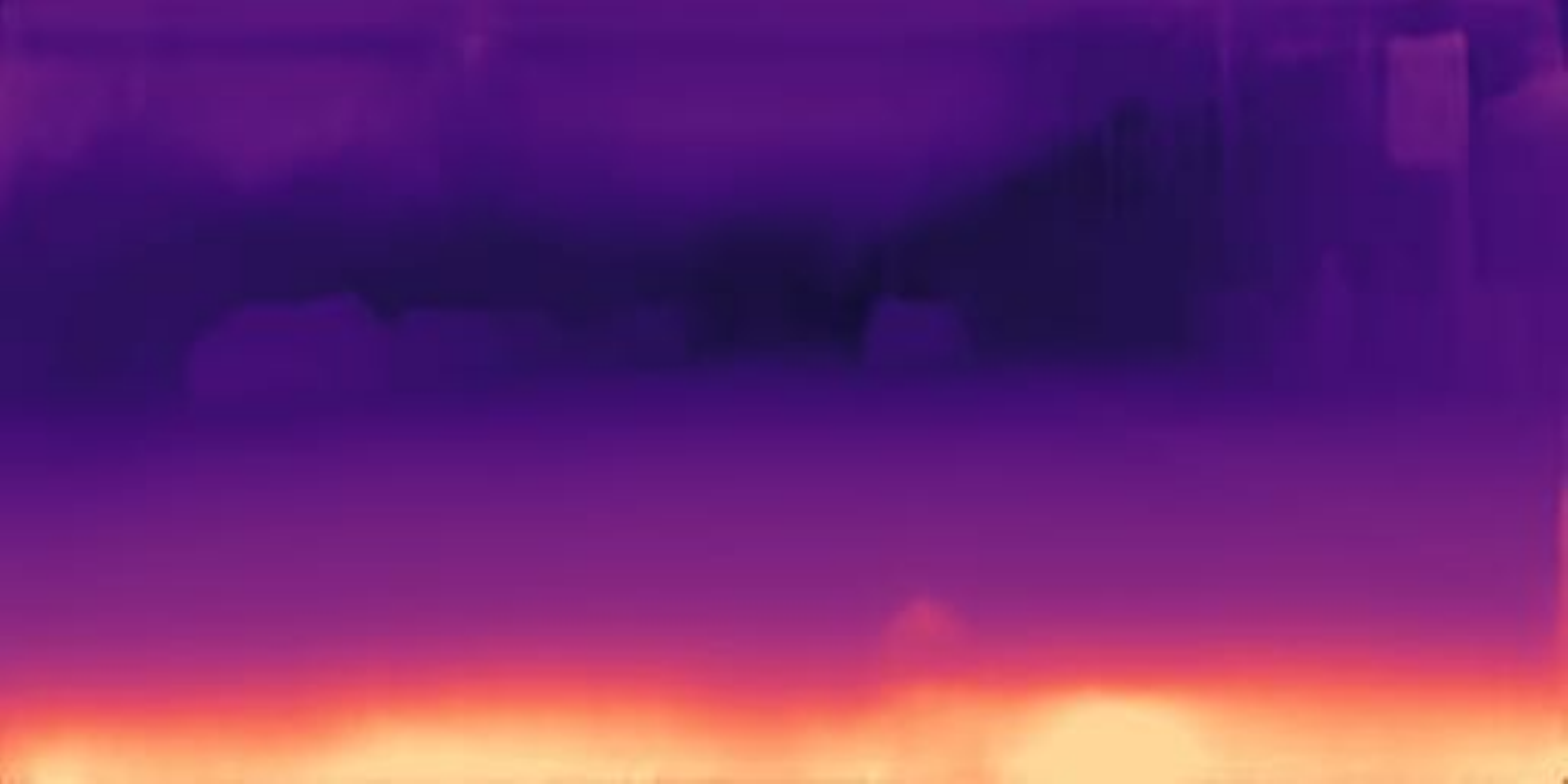}\qquad\qquad\quad &
\includegraphics[width=\iw,height=\ih]{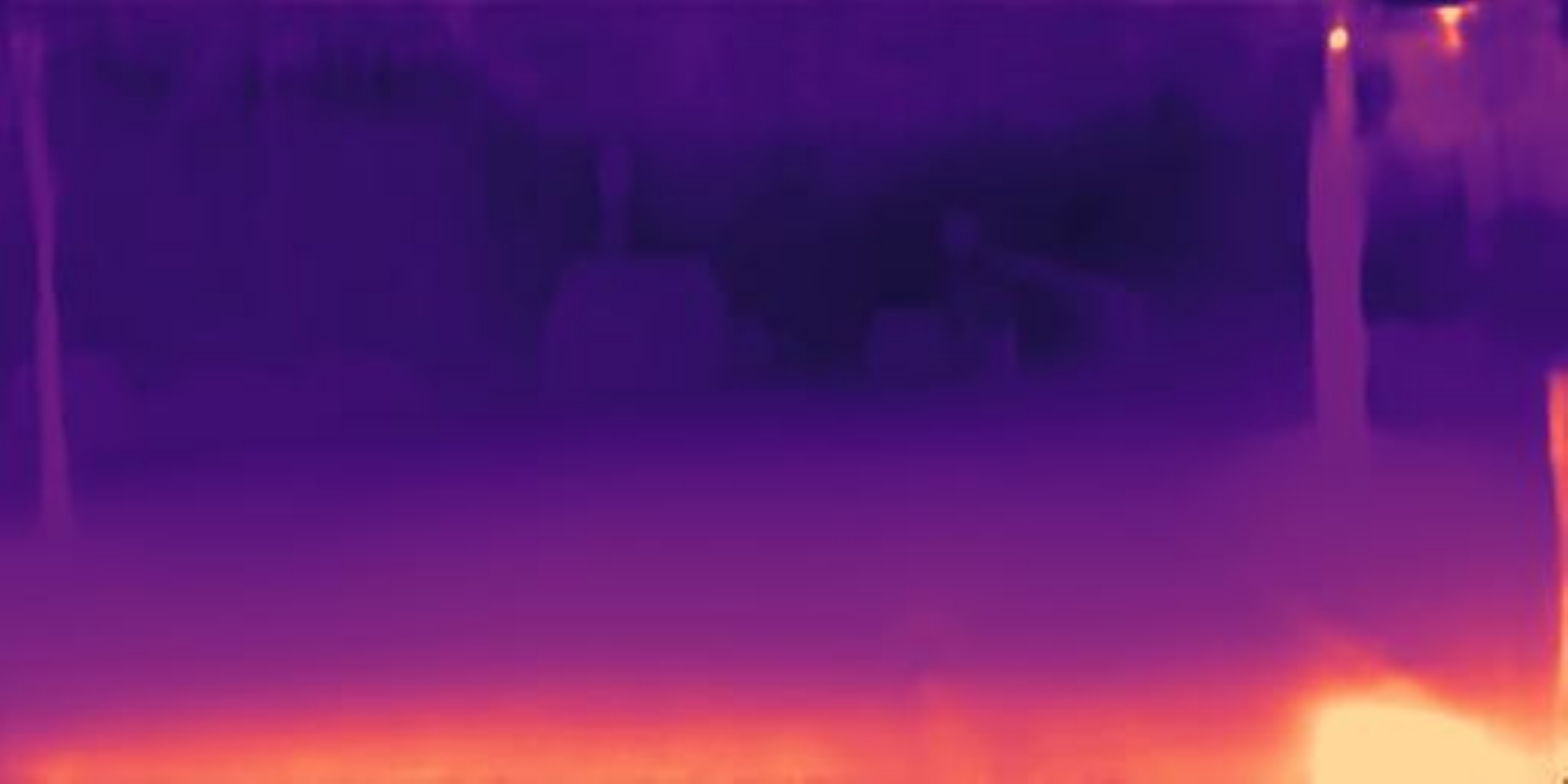}\\
\vspace{30mm} \\
\multicolumn{6}{c}{\fontsize{\w}{\h} \selectfont (d) Supervised Transformer-based methods} \\
\vspace{30mm} \\
\end{tabular}%
}
\caption{\textbf{Depth map results on real-world texture-shifted (RobotCar, Foggy and Rainy CityScapes) datasets.}  }
\label{figure_result_practical}
\end{figure*}

\subsubsection{Analysis on feature representation from backbones}
\label{feature-representation}
To analyze the internal properties of CNNs and Transformers, we employ the centered kernel alignment (CKA), which is the similarity measure of internal representations of neural networks by following previous works \cite{raghu2021vision,cortes2012algorithms,kornblith2019similarity}.
The CKA is widely used to analyze feature representations of neural networks \cite{song2012feature,kornblith2019similarity}, because of its invariant properties to the orthogonal transformation of representations and isotropic scaling \cite{raghu2021vision}.
We freeze the encoder $E$ of the depth networks trained on the original KITTI dataset. 
We extract features $z_a, z_b$ from the last layer of the encoder by passing the original image $\mathbf{I}_a$ and texture-shifted image $\mathbf{I}_b$, respectively.
Then, the CKA is computed given $K=z_a z_a^T$ and $L=z_b z_b^T$ with $m$ number of samples as follows:
\begin{align}
    CKA(K,L) = \frac{HSIC(K,L)}{\sqrt{HSIC(K,K)HSIC(L,L)}}, \\
    HSIC(K,L) = (KHLH)/(m-1)^2, 
\end{align}
where $H$ is the centering matrix $H_n= I_n - \frac{1}{n}\textbf{11}^T$.
The overall process is illustrated in \figref{figure_feature_process}.
We use 697 image pairs from the KITTI Eigen dataset and their corresponding texture-shifted datasets.
We compute the CKA with three types of texture-shifted datasets, including watercolor, pencil-sketch, and style-transfer, as shown in \figref{figure_result_feature}.  

\begin{figure*}[t!]
\centering
\newcommand\textw{250}
\newcommand\texth{300}
\resizebox{\linewidth}{!}{
\begin{tabular}{cccc}
\includegraphics[scale=1.4]{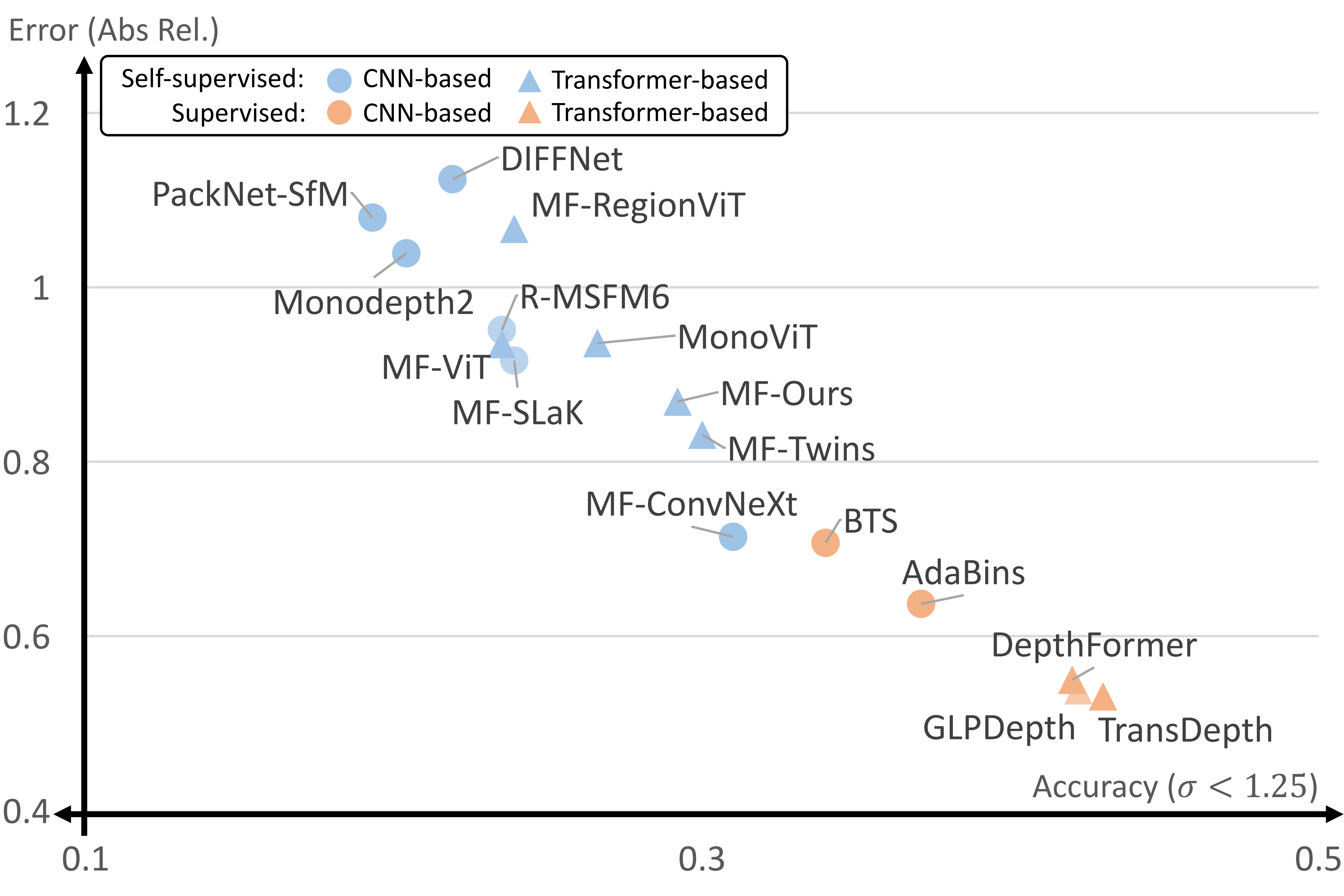} &
\includegraphics[scale=1.4]{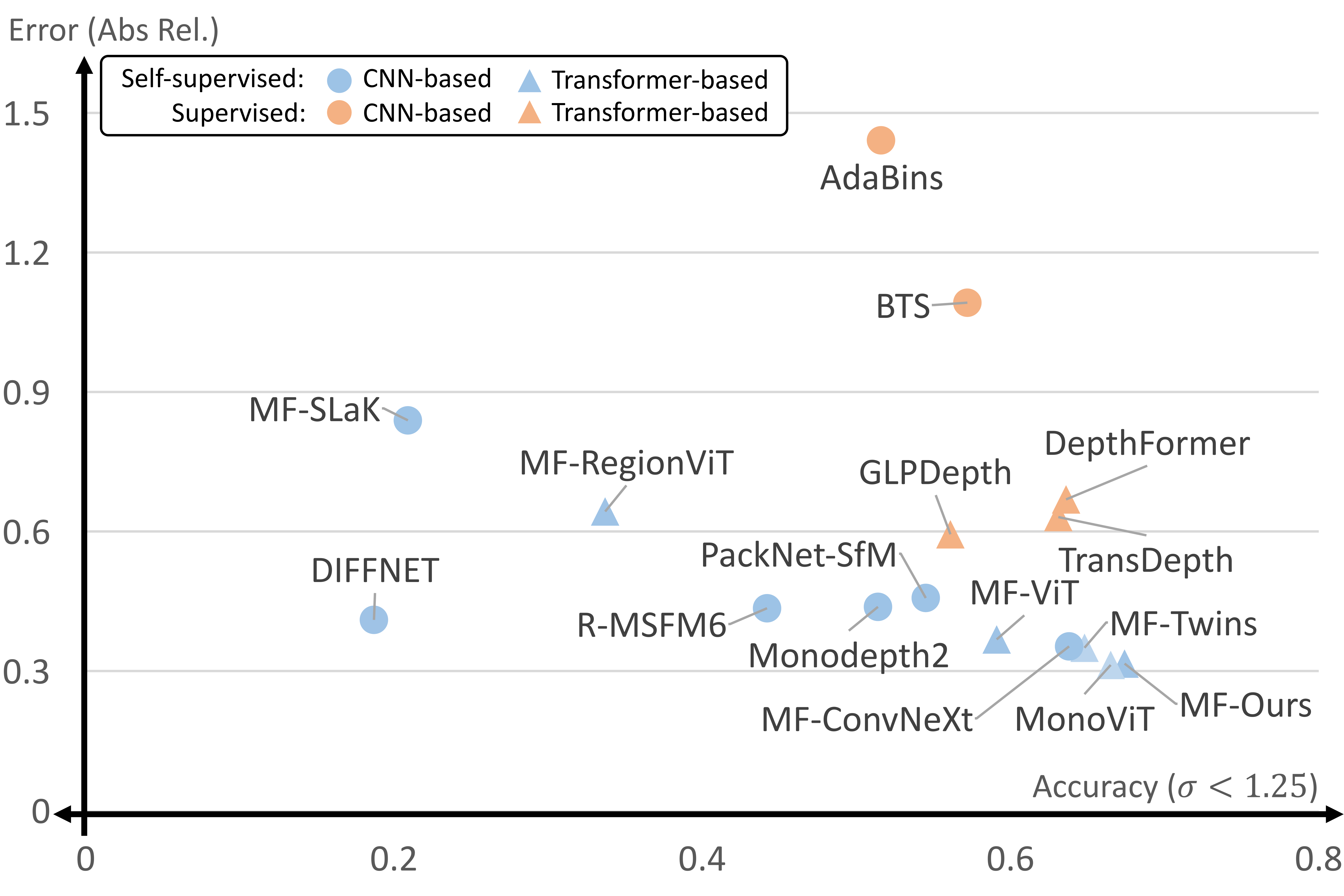} &
\includegraphics[scale=1.4]{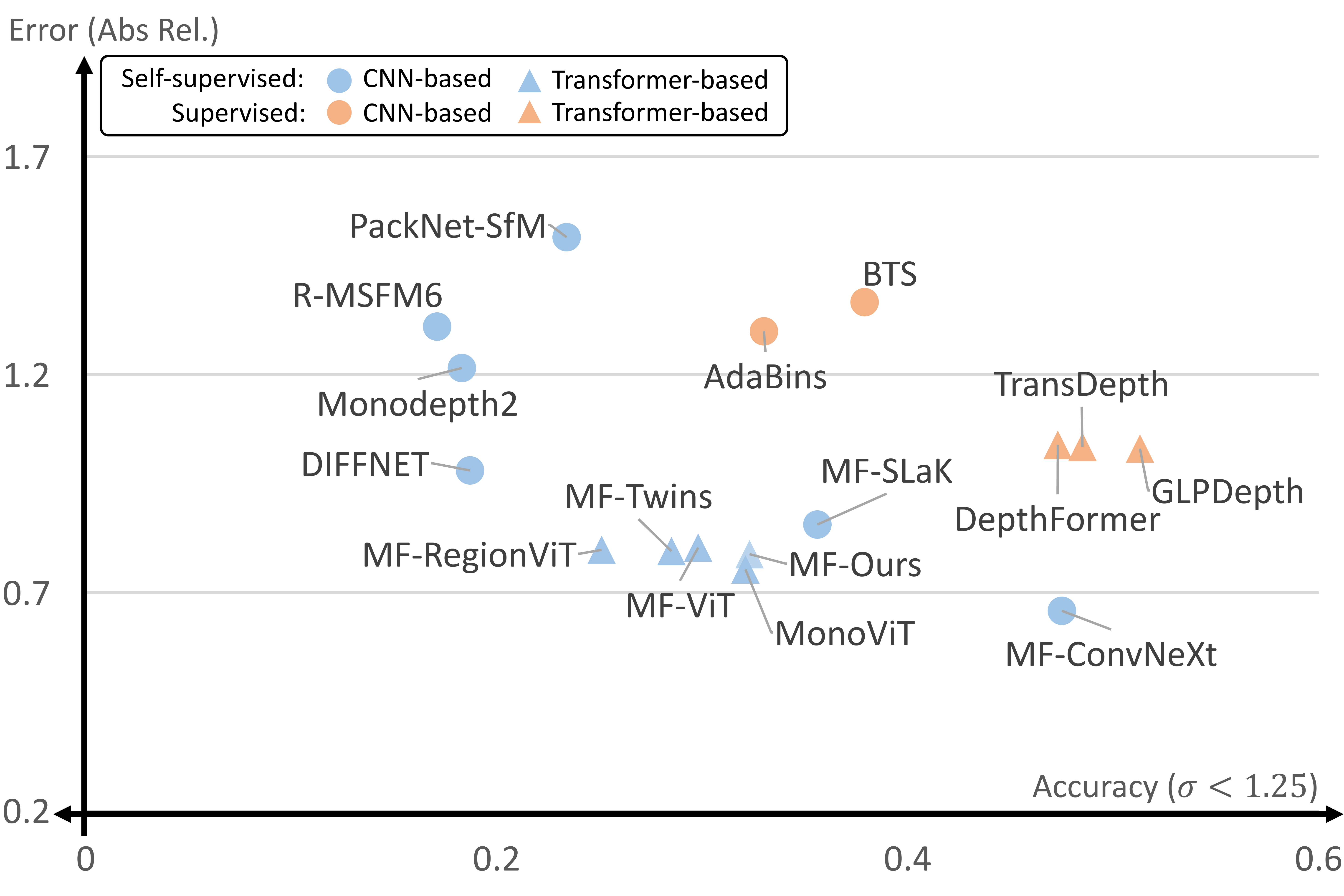} \vspace{30mm} \\
\fontsize{\textw}{\texth} \selectfont (a)  RobotCar  & \fontsize{\textw}{\texth} \selectfont (b)  Foggy CityScapes & \fontsize{\textw}{\texth} \selectfont (c)  Rainy CityScapes  \\ 
\end{tabular}}
\caption{ \textbf{Depth Performance Comparison on real-world texture-shifted (RobotCar, Foggy and Rainy CityScapes) datasets.}}
\label{table_result_robotcar}
\end{figure*}

The results of feature similarity show a similar aspect of the quantitative results on synthetic texture-shifted datasets in \tabref{table_result_texture}\footnote[3]{Note that these \tabref{table_result_texture} and \tabref{table_result_practical} are in the appendix.\label{fn:reffoot}}. 
MF-Ours, MonoViT and GLPDepth extract the most similar features from the original and texture-shifted datasets in the self-supervised and supervised methods.
Generally, Transformer-based models (MF-ViT, MF-Twins, GLPDepth, MF-Ours, TransDepth, DepthFormer) extract features of texture-shifted images similar to the original images. 
On the other hand, most CNN-based models (Monodepth2, PackNet-SfM, R-MSFM6, BTS, AdaBins) extract inconsistent features from original and texture-shifted images. 
The results are consistently observed regardless of the types of texture changes.
It demonstrates that the robustness to environmental changes comes from consistent feature extraction from encoders.
These results also support our observations in \secref{artifical-texture-exp} that the Transformers have a strong shape-bias whereas CNNs have a strong texture bias. Moreover, models with shape-biased representation show better generalization performance than models with texture bias for monocular depth estimation. 
We also observe that the pure Transformer-based methods (MF-ViT, MF-Twins, GLPDepth) and CNN-Transformer hybrid methods (MF-Ours, TransDepth, DepthFormer) show similar properties. 
The experiments provide us with some interesting observations on modern network structures.

First, it is noteworthy that MF-ConvNeXt, one of the CNN-based models, shows a relatively higher CKA similarity than the other CNN-based models.
Even the similarity of the models is on par with Transformer-based models.
ConvNeXt consists of a patchify stem, and a depth-wise convolution \cite{guo2019depthwise} with a large receptive field that mimics the self-attention mechanism using Transformer's macro design \cite{liu2022convnet}. 
The design intuition of ConvNeXt makes that ConvNeXt can focus on global information, similar to self-attention. Because of the modern macro design, MF-ConvNeXt consisting of only convolutional layers is robust to texture-shifted datasets and has a strong shape-bias.

Second, unlike other Transformer-based models, MF-RegionViT is sensitive to texture changes, similar to CNN-based models.
It is highly biased toward texture information rather than shape information, even though it is a transformer-based model. RegionViT \cite{chen2021regionvit} employs region-to-local attention, which is getting interaction between local regions. It improves the locality of the networks through local attention by capturing fine-grained spatial information. We believe that the strong locality makes MF-RegionViT rely heavily on spatial information.



Based on these experiments, we find that a self-attention module capturing global information, strengthen the shape-bias of the network. We also observe that the locality from local attention or the intrinsic property of CNNs improves the texture-bias of the network.
In terms of network design, a self-attention mechanism is ultimately key to generalization. 
Additionally, the locality of the network weakens generality, even if it helps to optimize in the training process.

\subsection{Evaluation on real-world texture-shifted datasets}
\label{practical-texture-exp}
In addition, we conduct experiments on practical texture changes that can occur in real driving scenarios to demonstrate their applicability to the real-world. 
We evaluate all models using practical texture-shifted datasets consisting of the night (Oxford RobotCar \cite{maddern20171}), foggy (Foggy CityScapes \cite{sakaridis2018semantic}), and rainy driving scenes (Rainy CityScapes \cite{hu2019depth}).
Overall, the experiment demonstrates that the properties of CNNs and Transformers identified in \secref{texture-shifted-exp} are similarly observed in real-world texture-shifted scenarios. 
Quantitative and qualitative results and detailed analyzes for each network are described in the following paragraphs.

The depth map results of each model are shown in \figref{figure_result_practical}. Similar to the experiments in \secref{artifical-texture-exp}, Transformer-based models show plausible depth maps, except MF-RegionViT.
On the other hand, CNN-based models fail to estimate depth even for objects such as a car in scenes similar to the training dataset, with the exception of MF-ConvNeXt.
The quantitative results in \figref{table_result_robotcar} and \tabref{table_result_practical}\footref{fn:reffoot} also show that CNN-based models have higher errors than Transformer-based models.
MF-SLaK is no different from other CNN models, but MF-ConvNeXt shows low errors and feasible depth maps by mimicking Transformers, as shown in \figref{figure_result_practical}.
MF-Twins show similar results to other Transformer-based models because only a weak locality is added compared to MF-RegionViT. However, MF-RegionViT indicates texture-biased results like CNN-based models due to its strong locality.

The results on rainy and foggy datasets (Rainy and Foggy CityScapes) also indicate similar aspects to the night scenes. Transformer-based models estimate the plausible depth and distinguish even tiny objects such as bollards. On the contrary, CNN-based models predict incorrect depth maps. In particular, CNN-based models exhibit significant errors in areas such as roads despite similar texture to training datasets. MF-ConvNeXt and MF-Twins also infer plausible depth maps regardless of changes in the weather environment, but MF-RegionViT and MF-SLaK estimate bizarre depth maps. 
In these experiments, we observe that textural shifts like weather and illumination changes confuse CNN-based models. We also find that CNN-based models are affected by changes in texture as well as loss of texture information, which can be a fatal problem in real-world driving applications.

\section{Conclusion}
In this paper, we have proposed a self-supervised monocular depth estimation method called MonoFormer (MF-Ours). 
More importantly, we have presented an in-depth analysis of the generalization performance of various modern backbone structures as well as state-of-the-art self-supervised and supervised methods for monocular depth estimation using various datasets. 
The in-distribution datasets are used to compare common performance on the KITTI benchmark, while the out-of-distribution and texture-shifted datasets are used to compare the generalization performance.
We deeply analyze the properties of the features extracted from each model using the synthetic texture-shifted datasets.
Finally, we demonstrate the applicability of the model in real-world scenarios of changing environments using day-night, foggy and rainy datasets.

Through extensive experiments, we observe that MF-Ours and GLPDepth have the best generalization performance among all self-supervised and supervised methods, respectively.
The supervised method, GLPDepth, achieves the best generalization performance among all monocular depth estimation methods.
More interestingly, we provide three important observations about the generality of monocular depth estimation.
First, CNN-based models heavily rely on texture information to recognize scenes and objects, while Transformer-based models use shape information to a greater degree for the monocular depth estimation task. 
Second, texture-biased representations result in poor generalization performance for environmental changes such as differences in illumination and weather. 
In contrast, shape-biased representations are more robust to such texture-shifted environments.
Lastly, ConvNeXt and RegionViT are CNN-based and Transformer-based, respectively, but have properties different from those of the backbone structure.
It shows that the texture-bias and shape-bias are not the intrinsic properties of CNNs and Transformers, respectively. 
Instead, the intrinsic locality of CNNs induces texture-biased characteristics, while the self-attention mechanism, the base layer of Transformers, induces shape-biased properties. 

We believe our observations provide valuable insights into best practices in network design not only for monocular depth estimation but also for a variety of dense prediction tasks such as semantic segmentation, depth completion, normal estimation, etc.
Based on these observations, the best way to design generalized networks for dense prediction is to utilize multi-self-attention (MSA) from Transformers and supplement the local information loss of global self-attention by using weak local attention as an auxiliary.


\section*{Acknowledgment}
This work was partly supported by Institute of Information \& communications Technology Planning \& Evaluation (IITP) grant funded by the Korea government(MSIT) (2020-0-00231, Development of Low Latency VR·AR Streaming Technology based on 5G edge cloud, 50\%) and the National Research Foundation of Korea (NRF) grant funded by the Korea government (MSIT) (No. RS-2023-00210908, 50\%).

\ifCLASSOPTIONcaptionsoff
  \newpage
\fi



\bibliographystyle{IEEEtran}
\bibliography{IEEEtran}

%



%

\vspace{-2cm}
\begin{IEEEbiography}[{\includegraphics[width=1in,height=1.25in,clip,keepaspectratio]{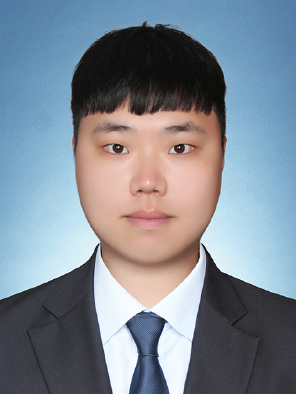}}]{Jinwoo Bae} received the B.S. degree in the Department of Electrical and Information Engineering from the Seoul National University of Science and Technology in 2020, and the M.S. degree in the Department of Electrical Engineering and Computer Science from DGIST in 2023.
He is currnetly a computer vision engineer in Hyundai Motor Group. His research interests include 3D vision such as depth estimation, robot vision, and multi-camera framework. He was a research intern at KIST (Seoul, Korea) in 2020 and ETRI (Pangyo, Korea) in 2019.
\end{IEEEbiography}

\vspace{-1.5cm}
\begin{IEEEbiography}[{\includegraphics[width=1in,height=1.25in,clip,keepaspectratio]{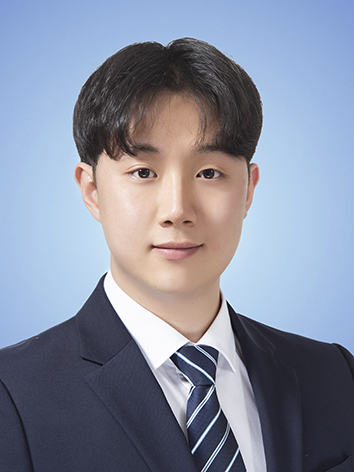}}]{Kyumin Hwang} received the B.S. degree in the Department of Computer Science and Engineering from the Kyungpook National University in 2020, and the M.S. degree in the Department of Information and Communication Engineering from DGIST in 2022. He is currently pursuing the Ph.D. degree in the Department of Electrical Engineering and Computer Science at DGIST. His research interests include 3D computer vision, including multi-view stereo for 3D reconstruction, and deep learning with geometry for autonomous driving. He was a research intern at ETRI (Daegu, Korea) in 2019.
\end{IEEEbiography}

\begin{IEEEbiography}[{\includegraphics[width=1in,height =1.25in,clip,keepaspectratio]{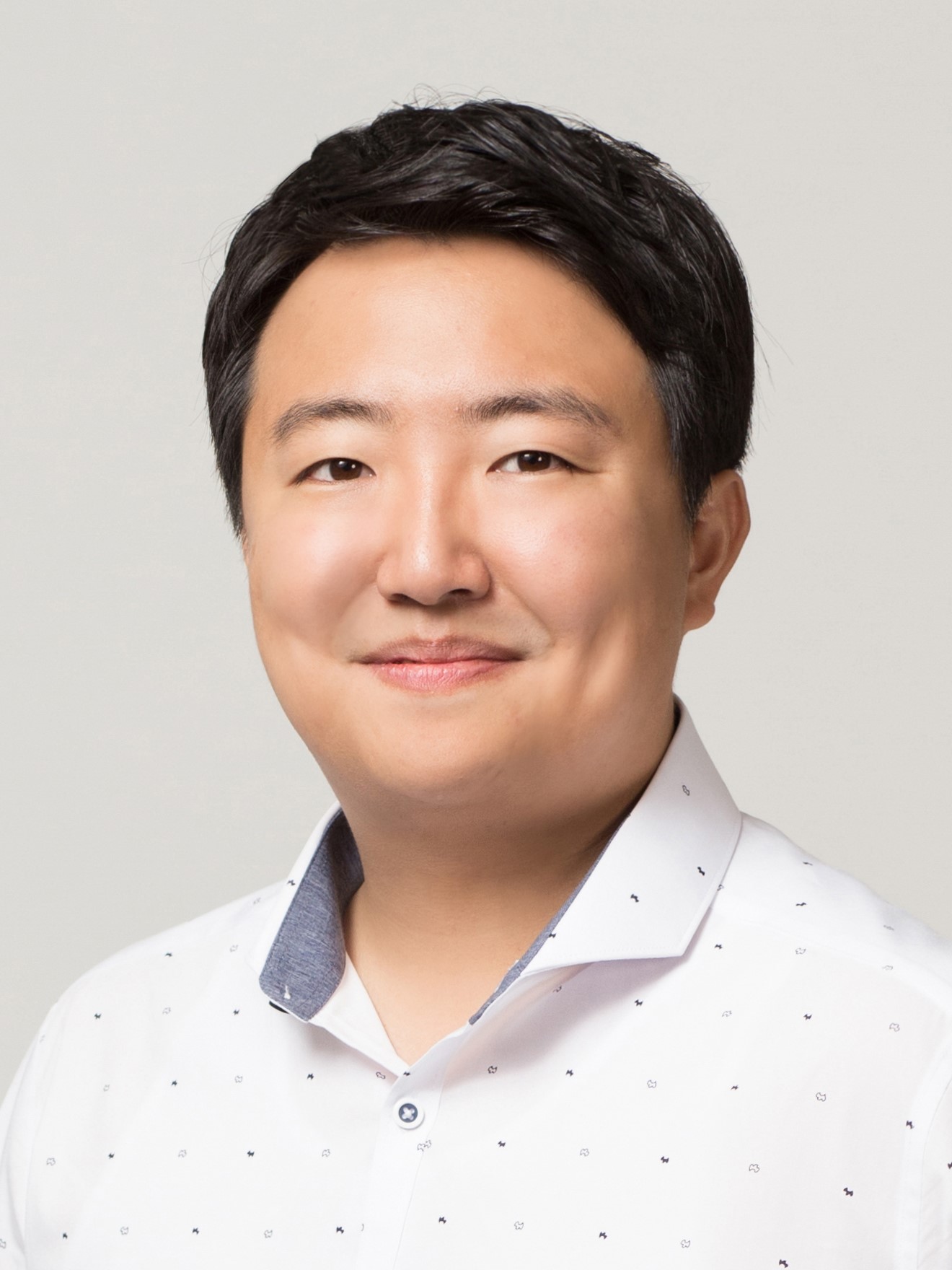}}]{Sunghoon Im} 
    received the B.S. degree in the Department of Electronic Engineering from Sogang University in 2014, and the M.S. and Ph.D. degree in the School of Electrical Engineering from KAIST in 2016 and 2019. He joined the Department of Electrical Engineering and Computer Science at DGIST, Daegu, Korea, in 2019, where he is currently working as an associate professor. He was a recipient of Microsoft Research Asia fellowship and Global Ph.D. fellowship from NRF of Korea. His research interests include computer vision, robot vision, and machine learning.
\end{IEEEbiography}

\clearpage
\onecolumn
\appendix
\setcounter{table}{0}
\setcounter{figure}{0}

\begin{table*}[h!]
\centering
\Large
\resizebox{\linewidth}{!}{%
\begin{tabular}{clcccccclcccccclcccccc}
\hline \hline 
\multirow{2}{*}{Datasets} &  & \multicolumn{6}{c}{RGBD}   &  & \multicolumn{6}{c}{SUN3D}  &  & \multicolumn{6}{c}{Scenes11}                                                                  \\ \cline{3-8} \cline{10-15} \cline{17-22} 
                    &  & Abs Rel  & Sq Rel  & RMSE  & $\delta < 1.25$  & $\delta < 1.25^2 $ & $\delta < 1.25^3 $&  & Abs Rel  & Sq Rel  & RMSE  & $\delta < 1.25$  & $\delta < 1.25^2 $ & $\delta < 1.25^3 $&  & Abs Rel  & Sq Rel  & RMSE  & $\delta < 1.25$  & $\delta < 1.25^2 $ & $\delta < 1.25^3 $    \\ \hline
Monodepth2          &  &0.610 & 0.508 & 0.488 & 0.292 & 0.520 & 0.681 & & 0.554 & 0.535 & 0.576 & 0.324 & 0.556 & 0.718&  &1.647 & 0.763 & 0.356 & 0.312 & 0.529 & 0.671 \\
PackNet-SfM         &  &0.593 & 0.416 & 0.460 & 0.318 & 0.562 & 0.731 & & 0.466 & 0.336 & 0.471 & 0.350 & 0.612 & 0.792&  & 2.065 & 0.837 & 0.330 & 0.310 & 0.530 & 0.674 \\
R-MSFM6             &  &0.695 & 0.553 & 0.490 & 0.261 & 0.471 & 0.627 & & 0.523 & 0.406 & 0.506 & 0.310 & 0.544 & 0.721&  &1.727 & 0.726 & 0.361 & 0.280 & 0.494 & 0.636 \\
DIFFNet             &  &0.573 & 0.333 & 0.359 & 0.278 & 0.509 & 0.683 & & 0.517 & 0.387 & 0.370 & 0.347 & 0.598 & 0.747 & & 1.444 & 0.387 & 0.240 & 0.342 & 0.571 & 0.708 \\
MF-ConvNeXt         &  & 0.346 & 0.094 & 0.208 & 0.438 & 0.706 & 0.860 & & 0.254 & 0.061 & 0.182 & 0.538 & 0.833 & 0.942& & 1.827 & 0.620 & 0.251 & 0.350 & 0.576 & 0.718\\
MF-SLaK             &  &0.481 & 0.198 & 0.272 & 0.349 & 0.607 & 0.770 & & 0.379 & 0.134 & 0.258 & 0.396 & 0.673 & 0.838&  &1.913 & 0.533 & 0.263 & 0.295 & 0.525 & 0.680 \\ \cline{1-1} \cline{3-8} \cline{10-15} \cline{17-22} 
MonoViT              &  & 0.425 & 0.150 & 0.258 & 0.361 & 0.612 & 0.759 & & 0.372 & 0.145 & 0.261 & 0.431 & 0.686 & 0.815 & & \textbf{1.163} & 0.270 & 0.229 & 0.377 & 0.609 & 0.736 \\
MF-ViT              &  & 0.337 & 0.098 & 0.207 & 0.475 &0.747  & 0.874 & & 0.244 & 0.059  & 0.174 &0.570  & 0.844 & 0.946& & 1.476 & 0.314 & 0.219 & 0.419  & 0.657 & 0.776 \\
MF-RegionViT        &  & 0.484 & 0.225 & 0.276 & 0.369 & 0.626 & 0.791 & & 0.358 & 0.121 & 0.247 & 0.413 & 0.696 & 0.855&  &1.866 & 0.486 & 0.257 & 0.316 & 0.546 & 0.695 \\ 
MF-Twins            &  & 0.323 &\textbf{0.085} & 0.194 & 0.522 & 0.749 & 0.876 & & 0.256 & 0.058 & 0.178 & 0.537 & 0.832 & 0.941&  &1.503 & 0.316 & 0.233 & 0.387 & 0.615 & 0.749 \\
MF-Ours           &  &  
 \textbf{0.302} &
 \textbf{0.085} & 
 \textbf{0.193} & 
 \textbf{0.532} & 
 \textbf{0.766} & 
 \textbf{0.888} &  & 
 \textbf{0.232} & 
 \textbf{0.051} & 
 \textbf{0.167} & 
 \textbf{0.586} & 
 \textbf{0.860} & 
 \textbf{0.953}&  &1.220 & \textbf{0.213} & \textbf{0.214} & \textbf{0.433} & \textbf{0.667} & \textbf{0.780}  \\ \hline \hline
BTS                       &  & 0.551 & 0.197 & 0.293 & 0.278 & 0.504 & 0.695 & & 0.403 & 0.145 & 0.268 & 0.380 & 0.664 & 0.842 &  & 2.446 & 0.844 & 0.269 & 0.269 & 0.496 & 0.663 \\
AdaBins                   &  & 0.448 & 0.126 & 0.261 & 0.269 & 0.539 & 0.760 & & 0.353 & 0.112 & 0.240 & 0.400 & 0.683 & 0.854&  & 2.583 & 0.992 & 0.265 & 0.277 & 0.512 & 0.675 \\ \cline{1-1} \cline{3-8} \cline{10-15} \cline{17-22} 
TransDepth                &  & 0.416 & 0.138 & 0.237 & 0.409 & 0.663 & 0.826 & & 0.297 & 0.087 & 0.218 & 0.511 & 0.793 & 0.908&  & 1.830 & 0.464 & 0.233 & 0.313 & 0.565 & 0.723 \\
DepthFormer               &  & 0.465 & 0.164 & 0.260 & 0.358 & 0.614 & 0.789 & & 0.277 & 0.083 & \textbf{0.204} & \textbf{0.563} & \textbf{0.836} & \textbf{0.935}&  & 2.470 & 0.930 & 0.251 & 0.324 & 0.560 & 0.708  \\ 
GLPDepth                  &  & \textbf{0.381} & \textbf{0.111} & \textbf{0.225} & \textbf{0.442} & \textbf{0.680} & \textbf{0.837} & & \textbf{0.273} & \textbf{0.074} & 0.207 & 0.533 & 0.819 & 0.928&  &\textbf{1.811} & \textbf{0.439} & \textbf{0.226} & \textbf{0.391} & \textbf{0.625} & \textbf{0.757}  \\ \hline \hline 
\end{tabular}%
}
\caption{\textbf{Quantitative results on the out-of-distribution (Common indoor environments - RGBD, SUN3D and Synthetic from graphics tools - Scenes11) datasets.} }
\label{table_result_indoor}
\vspace{-0.3cm}
\end{table*}

\begin{table*}[h!]
\centering
\resizebox{\columnwidth}{!}{%
\begin{tabular}{clcccccclcccccc}
\hline \hline
\multirow{2}{*}{Datasets} & & \multicolumn{6}{c}{MVS} & & \multicolumn{6}{c}{ETH3D}  \\ \cline{3-8} \cline{10-15}
 & & Abs Rel  & Sq Rel  & RMSE  & $\delta < 1.25$  & $\delta < 1.25^2 $ & $\delta < 1.25^3 $ & & Abs Rel  & Sq Rel  & RMSE  & $\delta < 1.25$  & $\delta < 1.25^2 $ & $\delta < 1.25^3 $ \\ \hline
Monodepth2  & & 0.471 & 0.407 & 0.503 & 0.408 & 0.661 & 0.806 & & 0.449&0.219&0.267&0.397&0.629&0.753 \\
PackNet-SfM  & & 0.449 & 0.295 & 0.429 & 0.397 & 0.670 & 0.837 &  & 0.359&0.119&0.218&0.436&0.700&0.839  \\
R-MSFM6  & & 0.550 & 0.603 & 0.583 & 0.352 & 0.591 & 0.756 & & 0.441&0.168&0.249&0.379&0.624&0.757\\
DIFFNet  & & 0.341 & 0.140 & 0.252 & 0.477 & 0.756 & 0.882 & & 0.425 & 0.190 & 0.246 & 0.420 & 0.645 & 0.775\\
MF-ConvNeXt    & & 0.259 & 0.060 & 0.171 & 0.582 & 0.863 & 0.949 & & 0.271&0.063&0.178&0.527&0.802&0.920 \\
MF-SLaK   & & 0.424 & 0.149 & 0.264 & 0.375 & 0.651 & 0.820 &  &0.386&0.115&0.226&0.375&0.652&0.821  \\ \cline{1-1} \cline{3-8} \cline{10-15}  
MonoViT  & & 0.284 & 0.083 & 0.203 & 0.538 & 0.814 & 0.916 & & 0.334 & 0.103 & 0.203 & 0.472 & 0.723 & 0.842 \\
MF-ViT  & & 0.254 & 0.061 & 0.179 & 0.589 &0.865  & 0.952 & & 0.249&0.056&0.167&0.559&0.826&0.933 \\
MF-RegionViT & & 0.376 & 0.124 & 0.238 & 0.436 & 0.714 & 0.863 & &0.364&0.098&0.136&0.410&0.695&0.853 \\
MF-Twins  & & 0.247 & 0.056 & 0.171 & 0.590 & 0.861 & 0.952 & & 0.270&0.062&0.182&0.531&0.801&0.921 \\ 
MF-Ours  & & \textbf{0.231} & \textbf{0.052} & \textbf{0.167} & \textbf{0.617} & \textbf{0.883} & \textbf{0.960} & & \textbf{0.232}&\textbf{0.050}&\textbf{0.160}&\textbf{0.587}&\textbf{0.851}&\textbf{0.946} \\ \hline \hline
BTS   & & 0.530 & 0.280 & 0.333 & 0.343 & 0.635 & 0.825 & & 0.508&0.199&0.287&0.324&0.582&0.758 \\ 
AdaBins  & & 0.409 & 0.157 & 0.246 & 0.392 & 0.689 & 0.870 & & 0.478&0.152&0.256&0.300&0.571&0.758 \\ \cline{1-1} \cline{3-8} \cline{10-15}    
TransDepth   & & 0.416 & 0.191 & 0.280 & 0.440 & 0.714 & 0.860 & &0.370&0.108&0.220&0.427&0.696&0.846  \\
DepthFormer   & & 0.369 & 0.158 & 0.259 & 0.516 & 0.793 & 0.906 & & \textbf{0.285}&0.079&\textbf{0.182}&\textbf{0.573}&\textbf{0.816}&\textbf{0.921}  \\
GLPDepth   & & \textbf{0.279} & \textbf{0.075} & \textbf{0.195} & \textbf{0.551} & \textbf{0.839} & \textbf{0.948} &  & 0.290&\textbf{0.071}&0.191&0.519&0.786&0.905  \\ \hline \hline
\end{tabular}}
\caption{\textbf{Quantitative results on the out-of-distribution (Man-made in/outdoor environments - MVS, ETH3D) datasets.} }
\label{table_result_inoutdoor}
\end{table*}

\begin{table*}[h!]
\centering
\Large
\resizebox{\linewidth}{!}{%
\begin{tabular}{clcccccclcccccclcccccc}
\hline \hline 
\multirow{2}{*}{Datasets} &  & \multicolumn{6}{c}{Watercolor}                                                                     &  & \multicolumn{6}{c}{Pencil-sketch}                                                                   &  & \multicolumn{6}{c}{Style-transfer}                                                                  \\ \cline{3-8} \cline{10-15} \cline{17-22} 
                    &  & Abs Rel  & Sq Rel  & RMSE  & $\delta < 1.25$  & $\delta < 1.25^2 $ & $\delta < 1.25^3 $&  & Abs Rel  & Sq Rel  & RMSE  & $\delta < 1.25$  & $\delta < 1.25^2 $ & $\delta < 1.25^3 $&  & Abs Rel  & Sq Rel  & RMSE  & $\delta < 1.25$  & $\delta < 1.25^2 $ & $\delta < 1.25^3 $    \\ \hline
Monodepth2          &  & 0.170          & 1.345          & 6.175          & 0.750          & 0.909          & 0.960          &  & 0.196          & 1.522          & 6.232          & 0.691          & 0.898          & 0.962          &  & 0.247          & 2.440          & 7.525          & 0.617          & 0.828          & 0.920          \\
PackNet-SfM         &  & 0.174          & 1.364          & 6.334          & 0.742          & 0.906          & 0.961          &  & 0.204          & 1.569          & 6.568          & 0.670          & 0.888          & 0.957          &  & 0.205          & 1.606          & 6.778          & 0.672          & 0.876          & 0.948          \\
R-MSFM6             &  & 0.194          & 1.613          & 7.173          & 0.696          & 0.876          & 0.943          &  & 0.217          & 1.698          & 6.719          & 0.647          & 0.872          & 0.951          &  & 0.294          & 2.788          & 8.579          & 0.521          & 0.770          & 0.894          \\
DIFFNet             &  & 0.158 & 1.209 & 5.945 & 0.776 & 0.921 & 0.965 &  & 0.157 & 1.097 & 5.654 & 0.776 & \textbf{0.935} & 0.973 &  & 0.225 & 1.988 &  7.328 & 0.649 & 0.845 & 0.926\\
MF-ConvNeXt         &  & 0.159          & 1.175          & 5.971          & 0.774          & 0.921          & 0.969          &  & 0.176          & {1.280} & 6.582          & 0.728          & 0.907          & 0.965          &  & 0.219          & 1.722          & 7.214          & 0.634          & 0.861          & 0.942          \\
MF-SLaK             &  & 0.170          & 1.272          & 6.316          & 0.755          & 0.907          & 0.961          &  & 0.268          & 2.268          & 8.042          & 0.553          & 0.811          & 0.915          &  & 0.224          & 1.859          & 7.423          & 0.648          & 0.855          & 0.934          \\ \cline{1-1} \cline{3-8} \cline{10-15} \cline{17-22} 
MonoViT              &  & \textbf{0.136} & \textbf{0.998} & \textbf{5.503} & \textbf{0.820} & \textbf{0.940} & \textbf{0.975} &  & 0.152 & 1.085 & 5.933 & 0.780 & 0.932 & 0.974 &  & \textbf{0.169} & \textbf{1.300} & \textbf{6.410} & \textbf{0.743} & \textbf{0.909} & \textbf{0.962 }\\
MF-ViT              &  & {0.152} & {1.196} & {5.668} & {0.799} & {0.932} & {0.973} &  & {0.174} & 1.311          & {5.770} & {0.756} & {0.920} & {0.967} &  & {0.186} & {1.379} & {6.652} & {0.705} & {0.898} & {0.959} \\
MF-RegionViT        &  & 0.179          & 1.341          & 6.309          & 0.732          & 0.902          & 0.958          &  & 0.260          & 2.012          & 7.491          & 0.545          & 0.832          & 0.939          &  & 0.242          & 2.030          & 7.944          & 0.587          & 0.835          & 0.927          \\ 
MF-Twins            &  & 0.166          & 1.295          & 6.558          & 0.751          & 0.911          & 0.963          &  & 0.205          & 1.630          & 7.512          & 0.655          & 0.871          & 0.949          &  & 0.232          & 1.974          & 8.045          & 0.599          & 0.839          & 0.931          \\
MF-Ours           &  & 0.140 & 1.053 & 5.665 & 0.815 & 0.936 & \textbf{0.975} &  & \textbf{0.151} & \textbf{1.084} & \textbf{5.615} & \textbf{0.786} & 0.934 & \textbf{0.976} &  & 0.175 & 1.307 & 6.435 & 0.728 & 0.906 & \textbf{0.962} \\ \hline \hline
BTS                       &  & 0.225          & 2.469          & 8.503          & 0.599          & 0.753          & 0.844          &  & 0.247          & 2.180          & 7.956          & 0.509          & 0.745          & 0.871          &  & 0.284          & 3.233          & 9.300          & 0.512          & 0.674          & 0.770          \\
AdaBins                   &  & 0.301          & 2.938          & 8.761          & 0.415          & 0.642          & 0.782          &  & 0.165          & 1.133          & 5.812          & 0.721          & 0.889          & 0.957          &  & 0.317          & 3.500          & 9.594          & 0.414          & 0.626          & 0.751          \\ \cline{1-1} \cline{3-8} \cline{10-15} \cline{17-22} 
TransDepth                &  & {0.111} & {0.728} & {4.886} & {0.855} & {0.955} & {0.985} &  & {0.136} & {0.987} & {5.774} & {0.775} & {0.926} & {0.975} &  & {0.146} & {1.127} & {5.910} & {0.761} & {0.911} & {0.964} \\
DepthFormer               &  & 0.135          & 1.102          & 5.983          & 0.783          & 0.905          & 0.955          &  & 0.162          & 1.290          & 6.498          & 0.713          & 0.885          & 0.953          &  & 0.274          & 2.766          & 8.366          & 0.501          & 0.691          & 0.801          \\ 
GLPDepth                  &  & \textbf{0.099} & \textbf{0.573} & \textbf{4.121} & \textbf{0.872} & \textbf{0.960} & \textbf{0.986} &  & \textbf{0.092} & \textbf{0.513} & \textbf{3.950} & \textbf{0.874} & \textbf{0.970} & \textbf{0.992} &  & \textbf{0.099} & \textbf{0.513} & \textbf{3.950} & \textbf{0.874} & \textbf{0.970} & \textbf{0.992} \\ \hline \hline 
\end{tabular}%
}
\caption{\textbf{Quantitative results on synthetic texture-shifted (Watercolor, Pencil-sketch, Style-transfer) datasets.} }
\label{table_result_texture}
\vspace{-0.3cm}
\end{table*}

\begin{table*}[h!]
\centering
\large
\resizebox{\linewidth}{!}{%
\begin{tabular}{clcccccclcccccclcccccc}
\hline \hline
\multirow{2}{*}{Datasets} &  & \multicolumn{6}{c}{Oxford RobotCar}                                                                     &  & \multicolumn{6}{c}{Foggy CityScapes}                                                                   &  & \multicolumn{6}{c}{Rainy CityScapes}                                                                  \\ \cline{3-8} \cline{10-15} \cline{17-22} 
                    &  & Abs Rel  & Sq Rel  & RMSE  & $\delta < 1.25$  & $\delta < 1.25^2 $ & $\delta < 1.25^3 $    &  & Abs Rel  & Sq Rel  & RMSE  & $\delta < 1.25$  & $\delta < 1.25^2 $ & $\delta < 1.25^3 $ &  & Abs Rel  & Sq Rel  & RMSE  & $\delta < 1.25$  & $\delta < 1.25^2 $ & $\delta < 1.25^3 $ \\ \hline
Monodepth2          &  & 1.039          & 0.044          & \textbf{0.159} & 0.204          & 0.406          & 0.564          &           & 0.438          & 0.060          & 0.109          & 0.514          & 0.757          & 0.853          &           & 1.215          & 2.194          & 1.289          & 0.183          & 0.348          & 0.495          \\
PackNet-SfM         &  & 1.080          & 0.045          & \textbf{0.159} & 0.193          & 0.378          & 0.536          &           & 0.457          & 0.055          & 0.102          & 0.545          & 0.782          & 0.873          &           & 1.515          & 1.771          & 1.169          & 0.234          & 0.444          & 0.609          \\
R-MSFM6             &  & 0.951          & 0.043          & {0.160} & 0.235          & 0.362          & 0.469          &           & 0.435          & 0.062          & 0.124          & 0.442          & 0.697          & 0.809          &           & 1.310          & 3.028          & 1.509          & 0.171          & 0.315          & 0.441          \\
DIFFNet             &  & 1.124 & 0.049 & 0.160 & 0.219 & 0.336 & 0.556 & & 0.410 & 0.214 & 0.099 & 0.187 & 0.352 & 0.515 & & 0.980 & 1.214 & 0.990 & 0.187 & 0.352 & 0.515 \\
MF-ConvNeXt         &  & \textbf{0.714} & \textbf{0.037} & \textbf{0.159} & \textbf{0.310} & {0.546} & \textbf{0.705} &           & {0.353} & 0.031          & 0.090          & 0.638          & {0.864} & {0.922} &           & \textbf{0.658} & \textbf{0.332} & \textbf{0.396} & \textbf{0.475} & \textbf{0.739} & \textbf{0.900} \\
MF-SLaK             &  & 0.916          & 0.041          & {0.160} & 0.239          & 0.456          & 0.623          &           & 0.839          & 0.151          & 0.205          & 0.209          & 0.440          & 0.652          &           & 0.856          & {0.510} & {0.460} & {0.356} & {0.631} & {0.822} \\ \cline{1-1} \cline{3-8} \cline{10-15} \cline{17-22} 
MonoViT              &  & 0.936 & 0.041 & 0.163 & 0.266 & 0.473 & 0.589 & {} & \textbf{0.313} & 0.044 & 0.081 & 0.665 & 0.862 & 0.921 & {} & 0.753 & 0.717 & 0.891 & 0.321 & 0.466 & 0.775\\
MF-ViT              &  & 0.935          & 0.042          & \textbf{0.159} & 0.235          & 0.444          & 0.602          & {} & 0.368          & 0.033          & 0.097          & 0.591          & 0.845          & 0.917          & {} & 0.803          & 0.891          & 0.868          & 0.298          & 0.545          & 0.709          \\
MF-RegionViT        &  & 1.067          & 0.041          & {0.160} & 0.239          & 0.456          & 0.623          &           & 0.643          & 0.088          & 0.167          & 0.337          & 0.605          & 0.782          &           & 0.798          & 0.773          & 0.805          & 0.251          & 0.493          & 0.676          \\ 
MF-Twins            &  & {0.831} & \textbf{0.037} & \textbf{0.159} & 0.300          & 0.516          & 0.686          &           & 0.350          & {0.029} & {0.087} & {0.648} & 0.862          & 0.920          &           & 0.795          & 0.825          & 0.829          & 0.285          & 0.526          & 0.688          \\ 
MF-Ours           &  & 0.869          & {0.040} & \textbf{0.159} & {0.292} & \textbf{0.555} & {0.691} & \textbf{} & 0.316 & \textbf{0.025} & \textbf{0.080} & \textbf{0.674} & \textbf{0.879} & \textbf{0.929} & \textbf{} & {0.788} & 0.707          & 0.752          & 0.323          & 0.574          & 0.738          \\ \hline \hline
BTS                       &  & 0.707          & 0.039          & {0.160} & 0.340          & 0.604          & 0.723          &           & 1.092          & 0.294          & 0.117          & 0.572          & 0.716          & 0.794          &           & 1.366          & 1.632          & 0.710          & 0.379          & 0.623          & 0.756          \\
AdaBins                   &  & 0.637          & {0.038} & {0.160} & 0.371          & 0.640          & 0.771          &           & 1.441          & 0.524          & 0.185          & 0.516          & 0.667          & 0.744          &           & 1.299          & 1.452          & {0.668} & 0.330          & 0.585          & 0.771          \\ \cline{1-1} \cline{3-8} \cline{10-15} \cline{17-22} 
TransDepth                &  & \textbf{0.531} & \textbf{0.037} & \textbf{0.159} & \textbf{0.430} & {0.645} & {0.768} & {} & {0.631} & {0.075} & \textbf{0.069} & {0.631} & {0.797} & {0.860} & {} & {1.034} & {1.003} & 0.793          & {0.485} & {0.698} & {0.771} \\
DepthFormer               &  & 0.550          & \textbf{0.037} & \textbf{0.159} & 0.420          & 0.629          & 0.756          &           & 0.669          & 0.083          & {0.070} & \textbf{0.636} & 0.794          & 0.857          &           & 1.039          & 1.016          & 0.816          & 0.473          & 0.690          & 0.764          \\ 
GLPDepth                  &  & {0.538} & \textbf{0.037} & \textbf{0.159} & {0.422} & \textbf{0.673} & \textbf{0.780} &           & \textbf{0.594} & \textbf{0.070} & 0.103          & 0.561          & \textbf{0.799} & \textbf{0.872} &           & \textbf{1.030} & \textbf{0.928} & \textbf{0.560} & \textbf{0.513} & \textbf{0.742} & \textbf{0.822} \\ \hline \hline
\end{tabular}%
}
\caption{\textbf{Quantitative results on real-world texture-shifted (RobotCar, Foggy and Rainy CityScapes) datasets.}}
\vspace{-0.2cm}
\label{table_result_practical}
\end{table*}

\newpage
\begin{figure*}[p]
\centering
\newcommand\iw{80cm}
\newcommand\ih{30cm}
\newcommand\w{200}
\newcommand\h{180}
\newcommand\textw{120}
\newcommand\texth{200}
\resizebox{\linewidth}{!}{%
\begin{tabular}{ccccccccc}
\multicolumn{1}{c}{\fontsize{\w}{\h} \selectfont RGBD } & 
\multicolumn{1}{c}{\fontsize{\w}{\h} \selectfont SUN3D } & 
\multicolumn{1}{c}{\fontsize{\w}{\h} \selectfont MVS } & 
\multicolumn{1}{c}{\fontsize{\w}{\h} \selectfont ETH3D } & 
\multicolumn{1}{c}{\fontsize{\w}{\h} \selectfont Scenes11 } \\
\vspace{30mm}\\
\rotatebox[origin=c]{90}{\fontsize{\textw}{\texth}\selectfont Input Images\hspace{-310mm}}\hspace{10mm} 
\includegraphics[width=\iw,height=\ih]{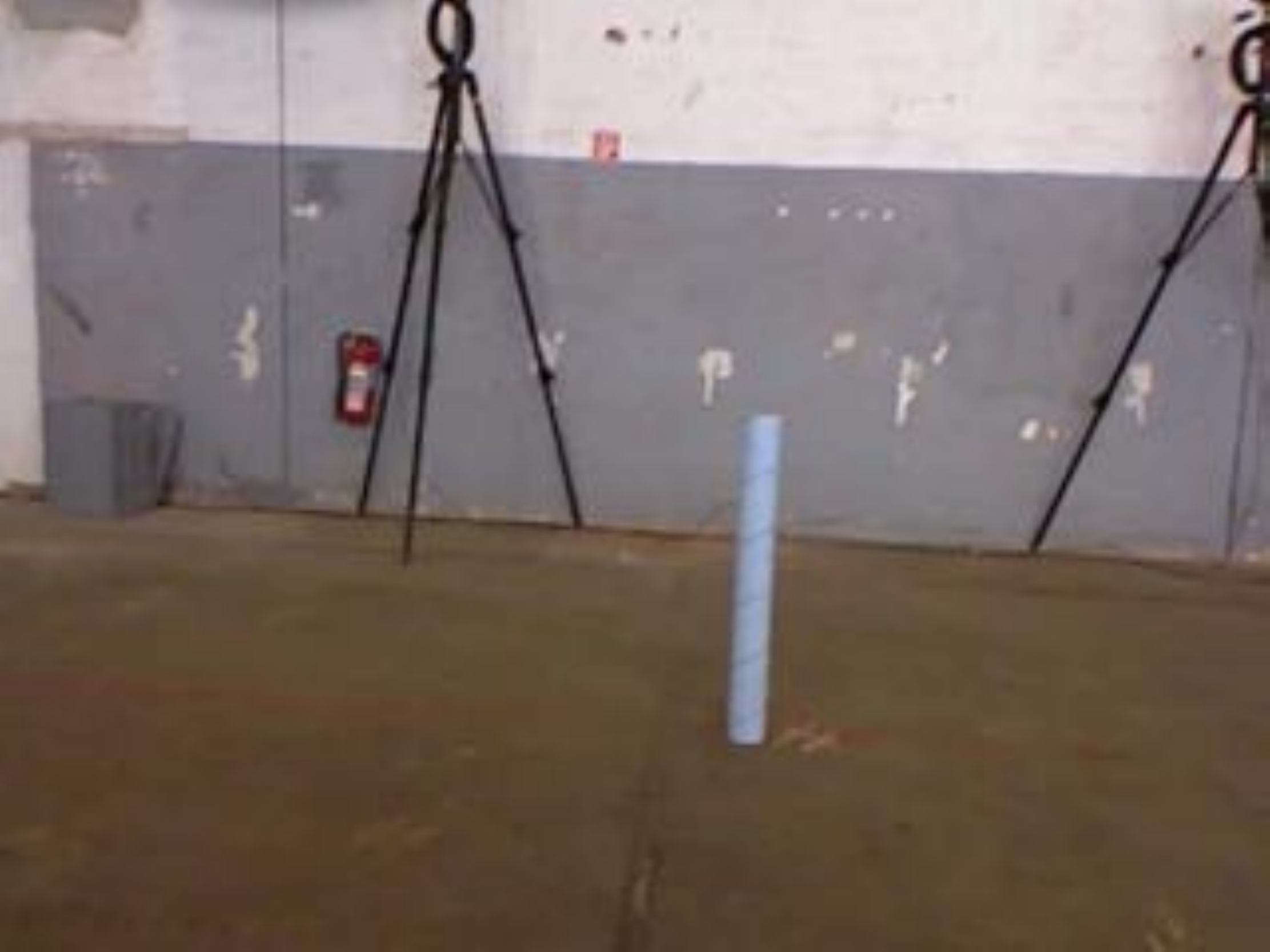}  \qquad\qquad\quad &  
\includegraphics[width=\iw,height=\ih]{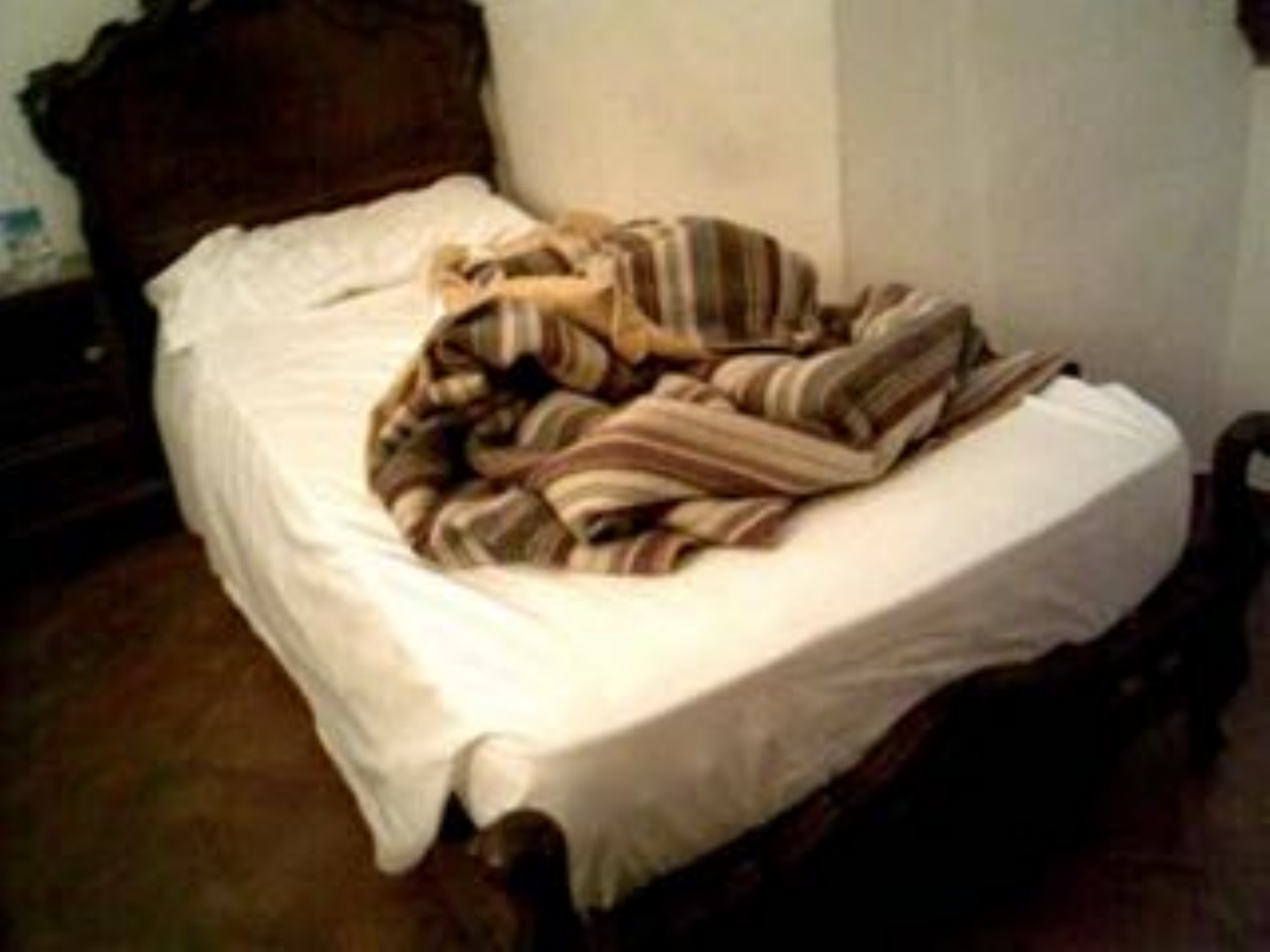}  \qquad\qquad\quad &  
\includegraphics[width=\iw,height=\ih]{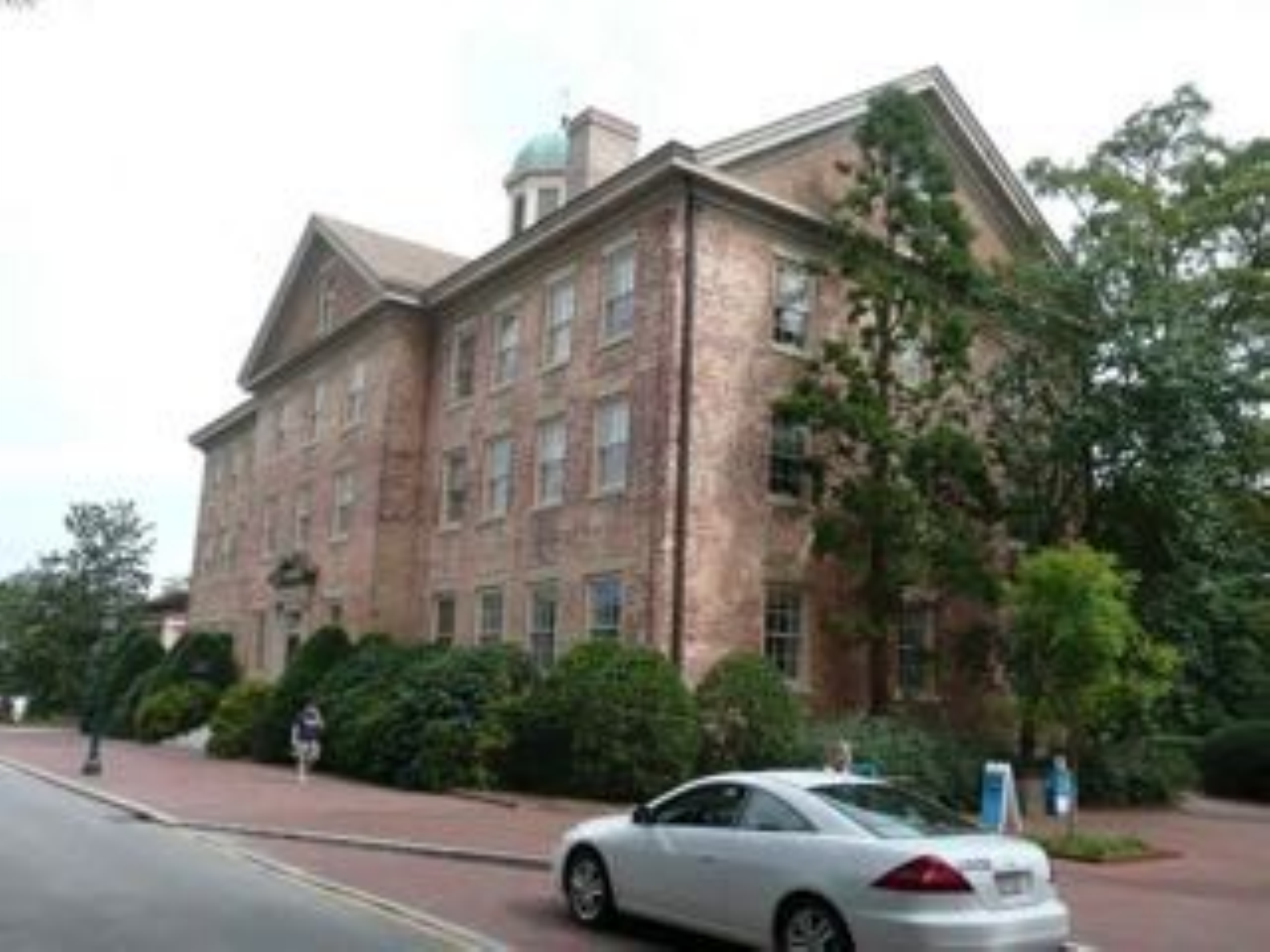} \qquad\qquad\quad &  
\includegraphics[width=\iw,height=\ih]{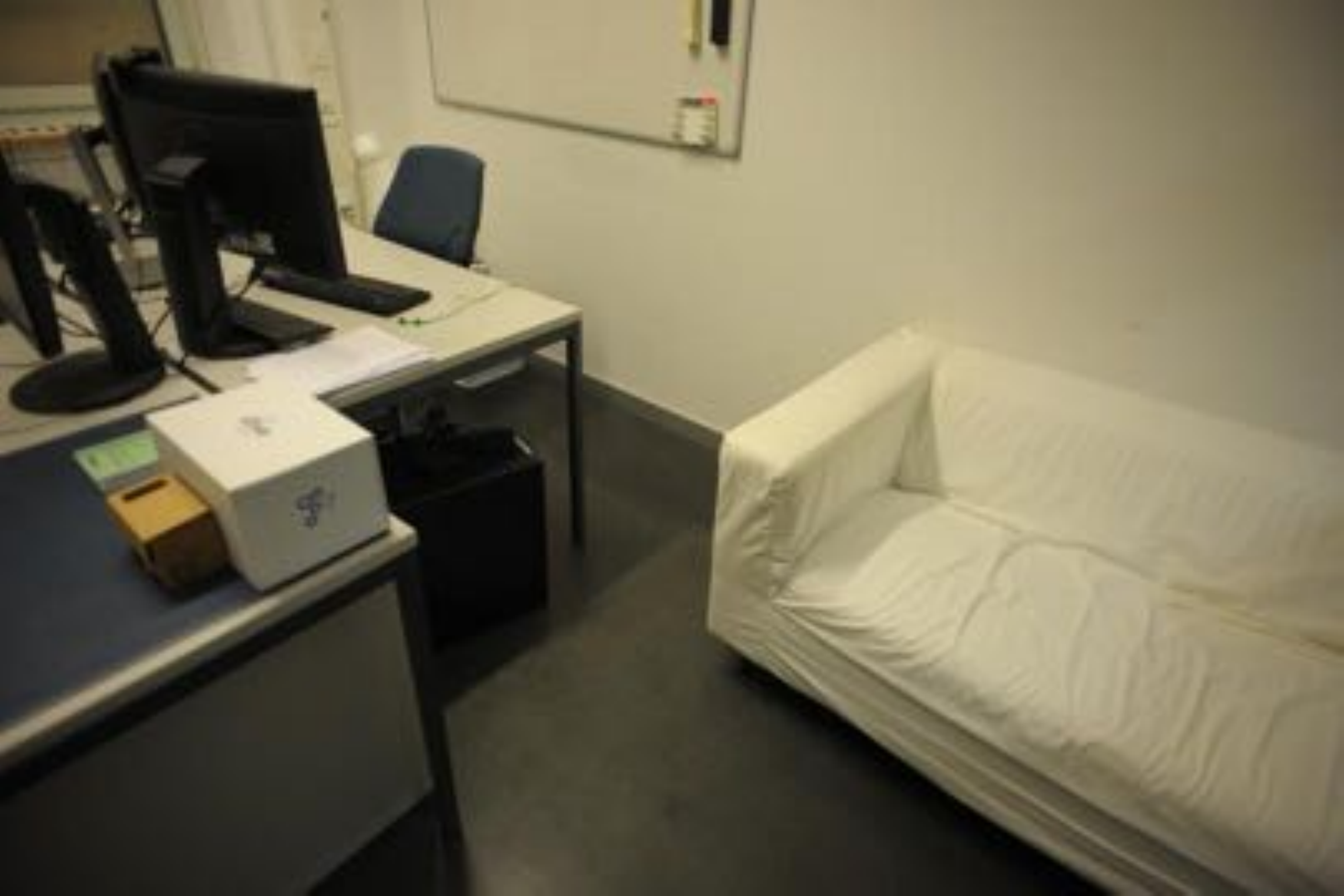} \qquad\qquad\quad &  
\includegraphics[width=\iw,height=\ih]{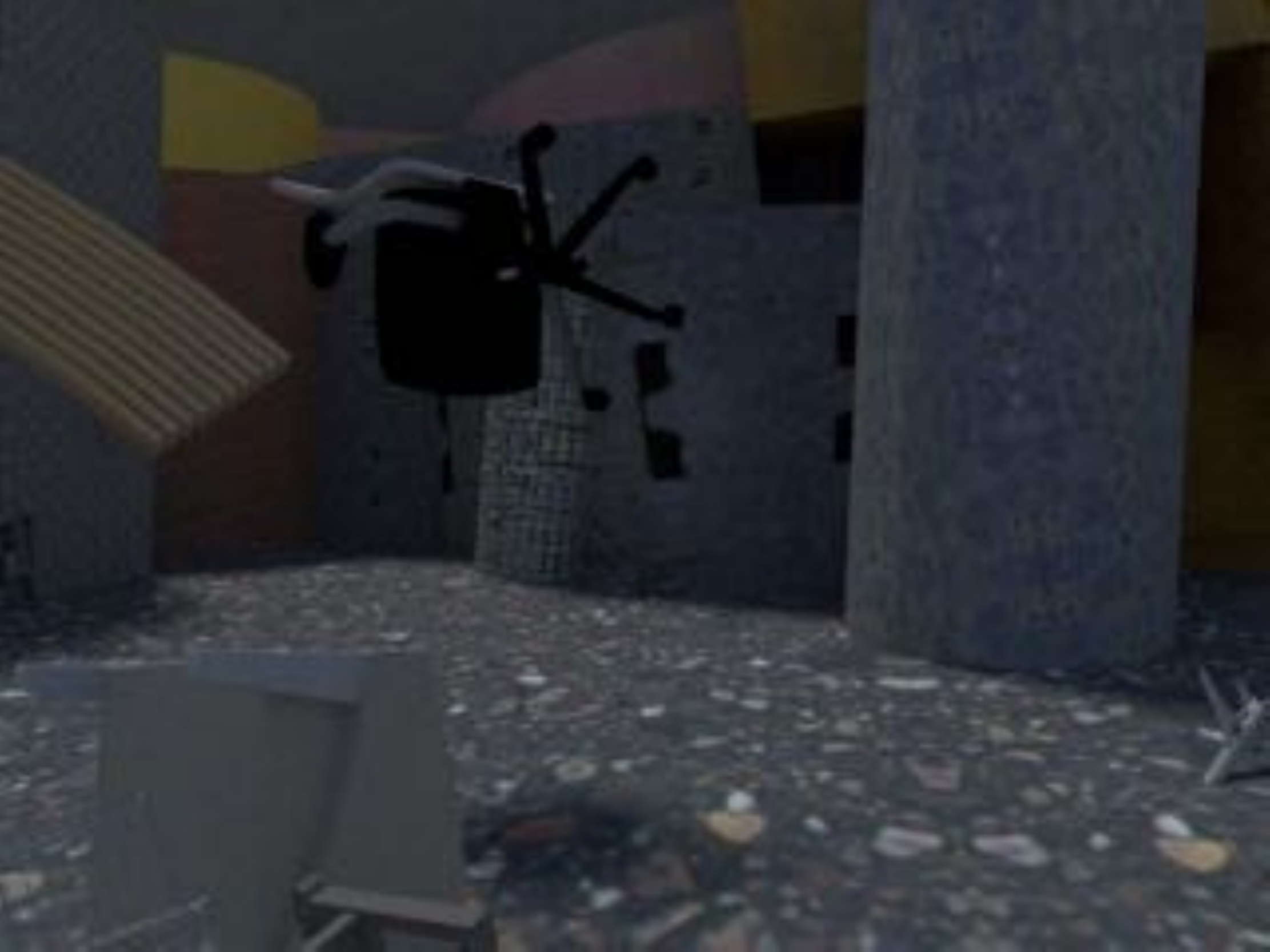}\\
\vspace{10mm}\\
\rotatebox[origin=c]{90}{\fontsize{\textw}{\texth} \selectfont Groundtruth\hspace{-250mm}}\hspace{25mm}
\includegraphics[width=\iw,height=\ih]{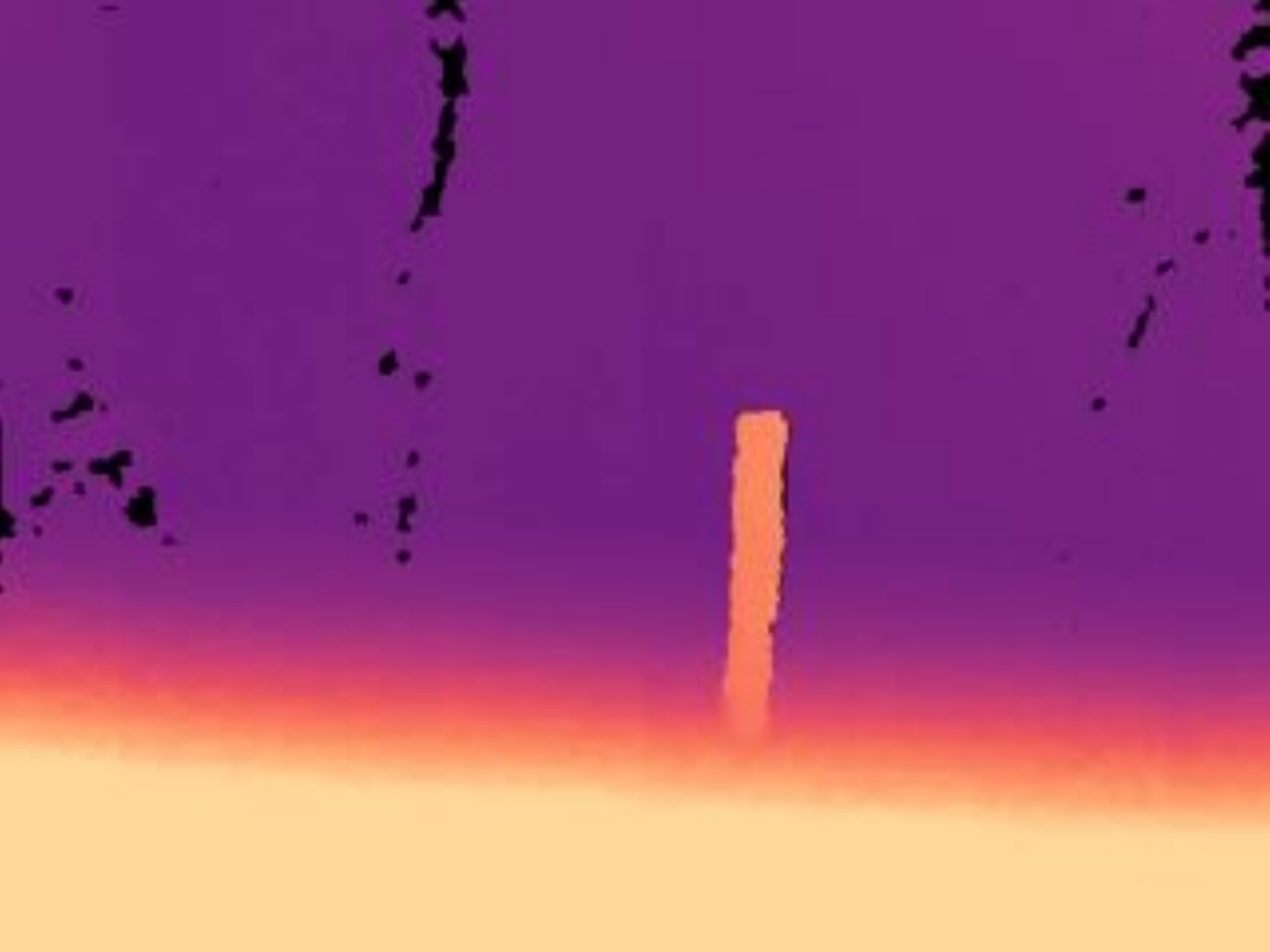} \qquad\qquad\quad &  
\includegraphics[width=\iw,height=\ih]{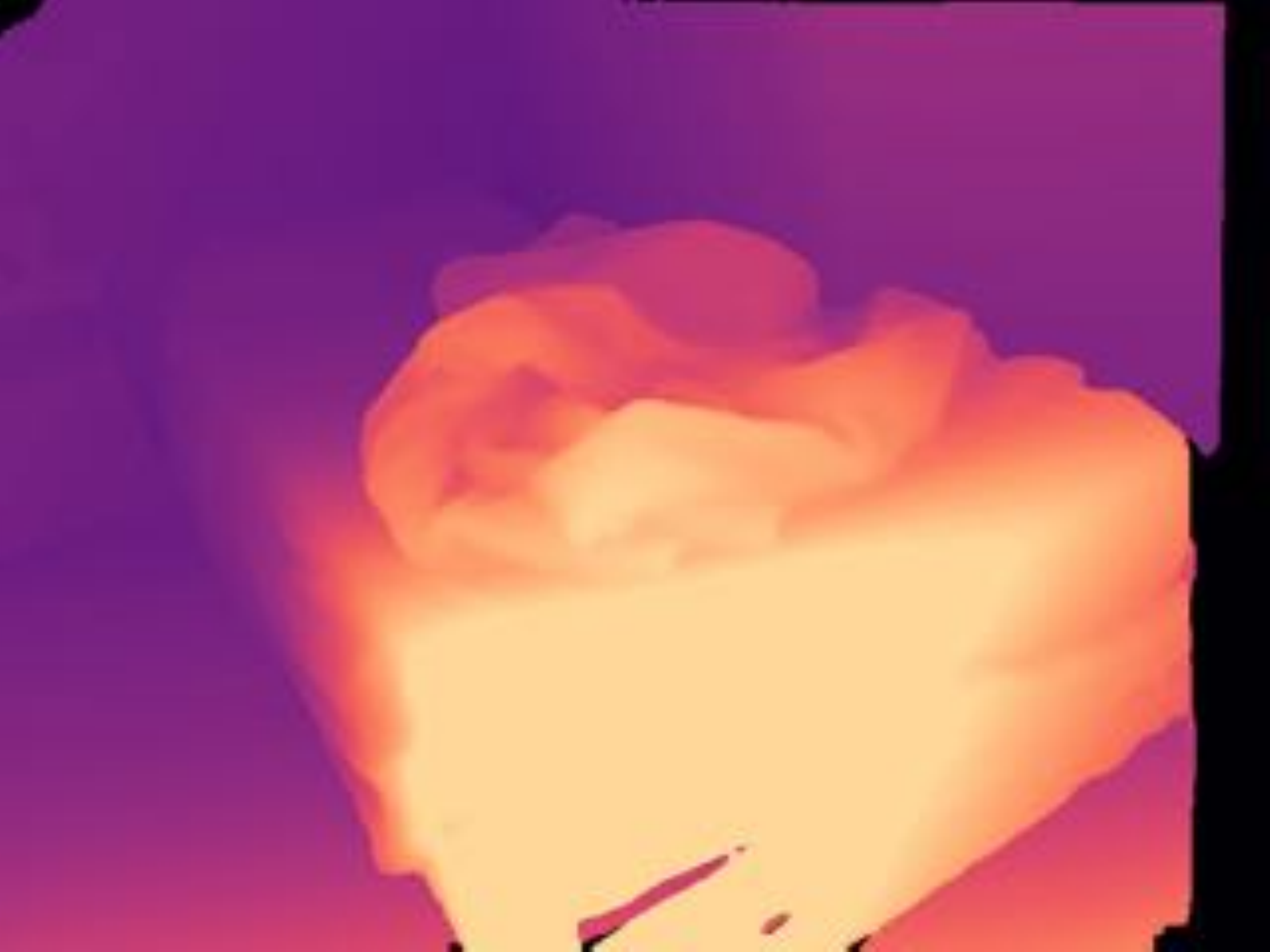} \qquad\qquad\quad &  
\includegraphics[width=\iw,height=\ih]{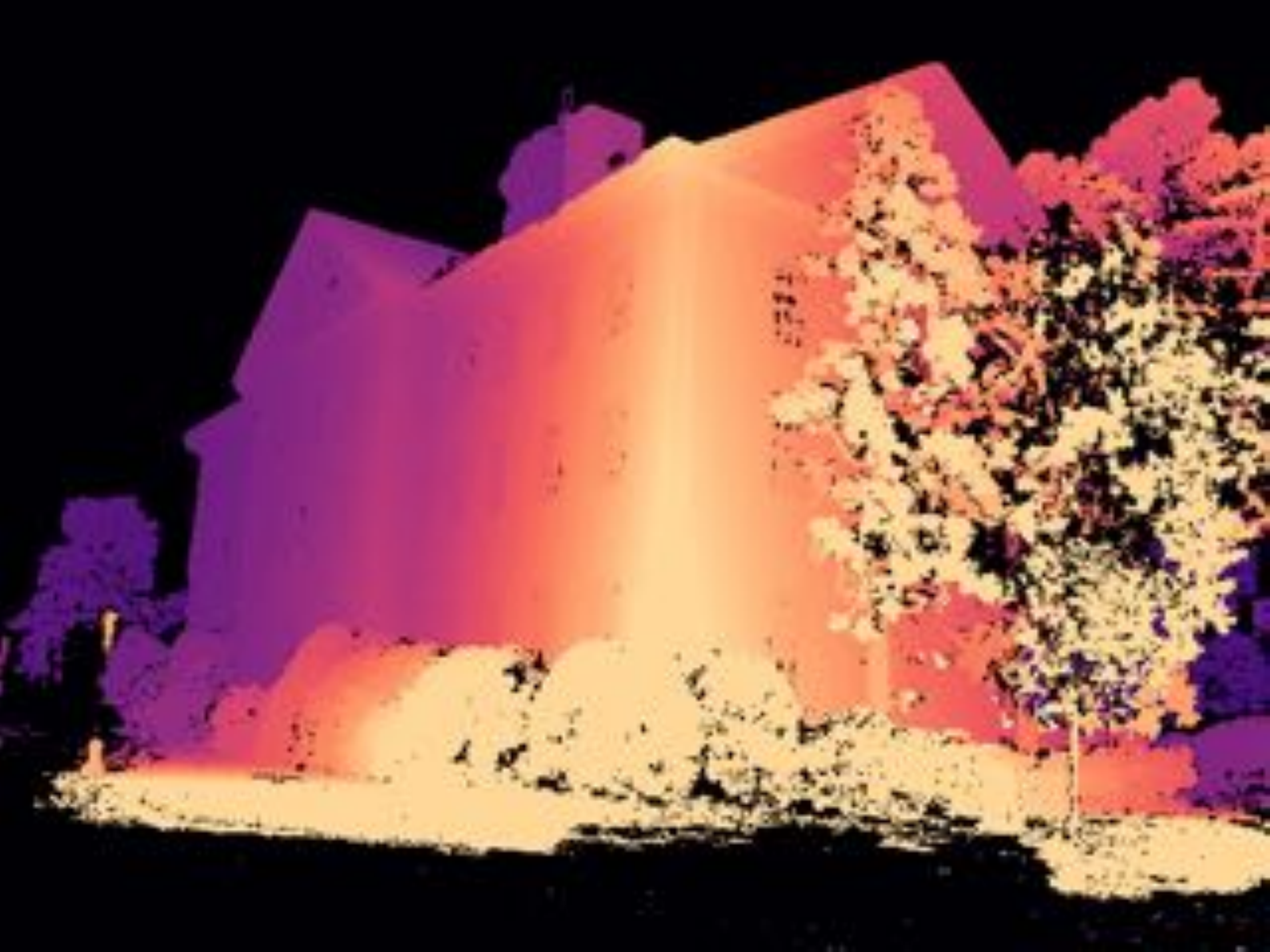} \qquad\qquad\quad &  
\includegraphics[width=\iw,height=\ih]{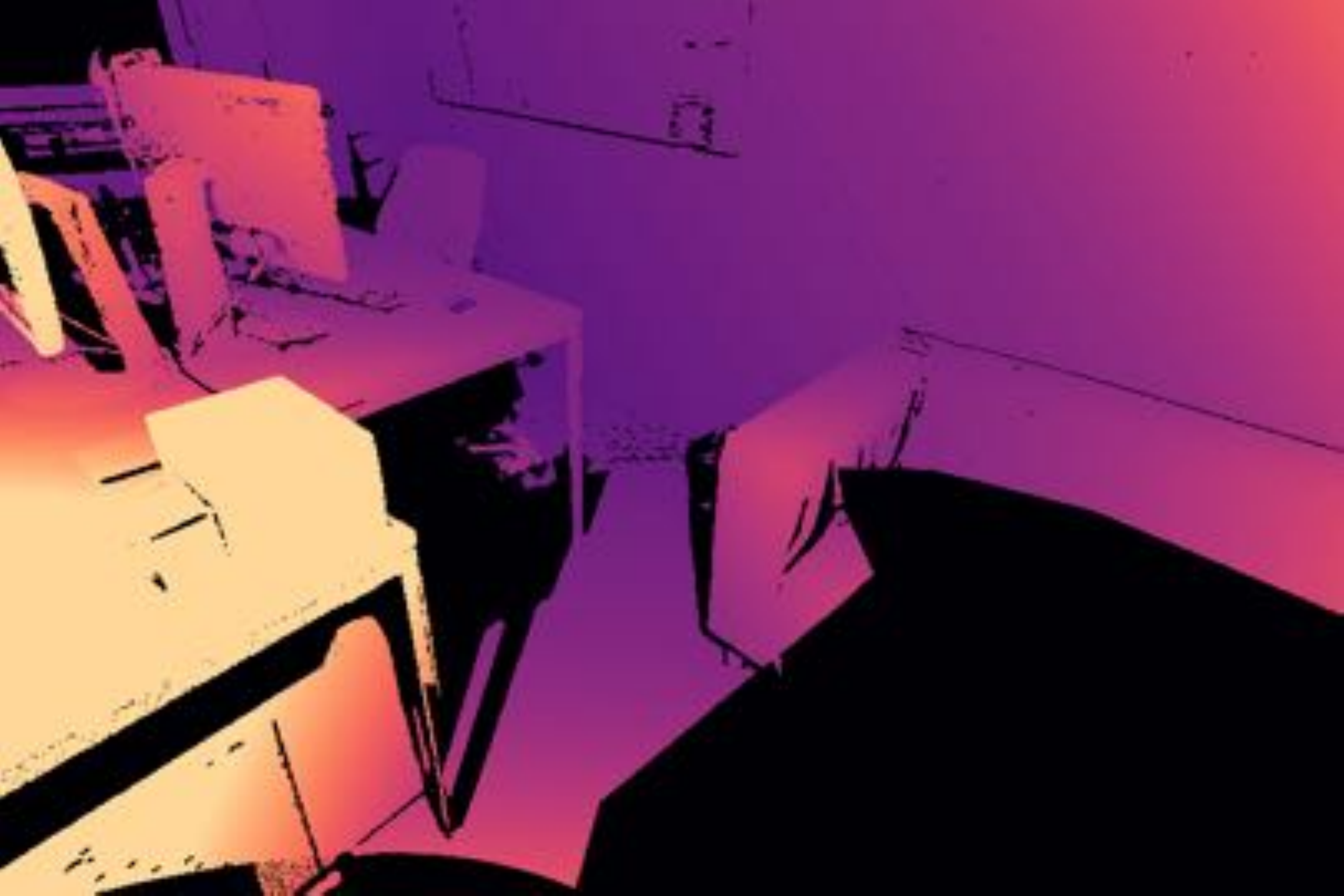} \qquad\qquad\quad &  
\includegraphics[width=\iw,height=\ih]{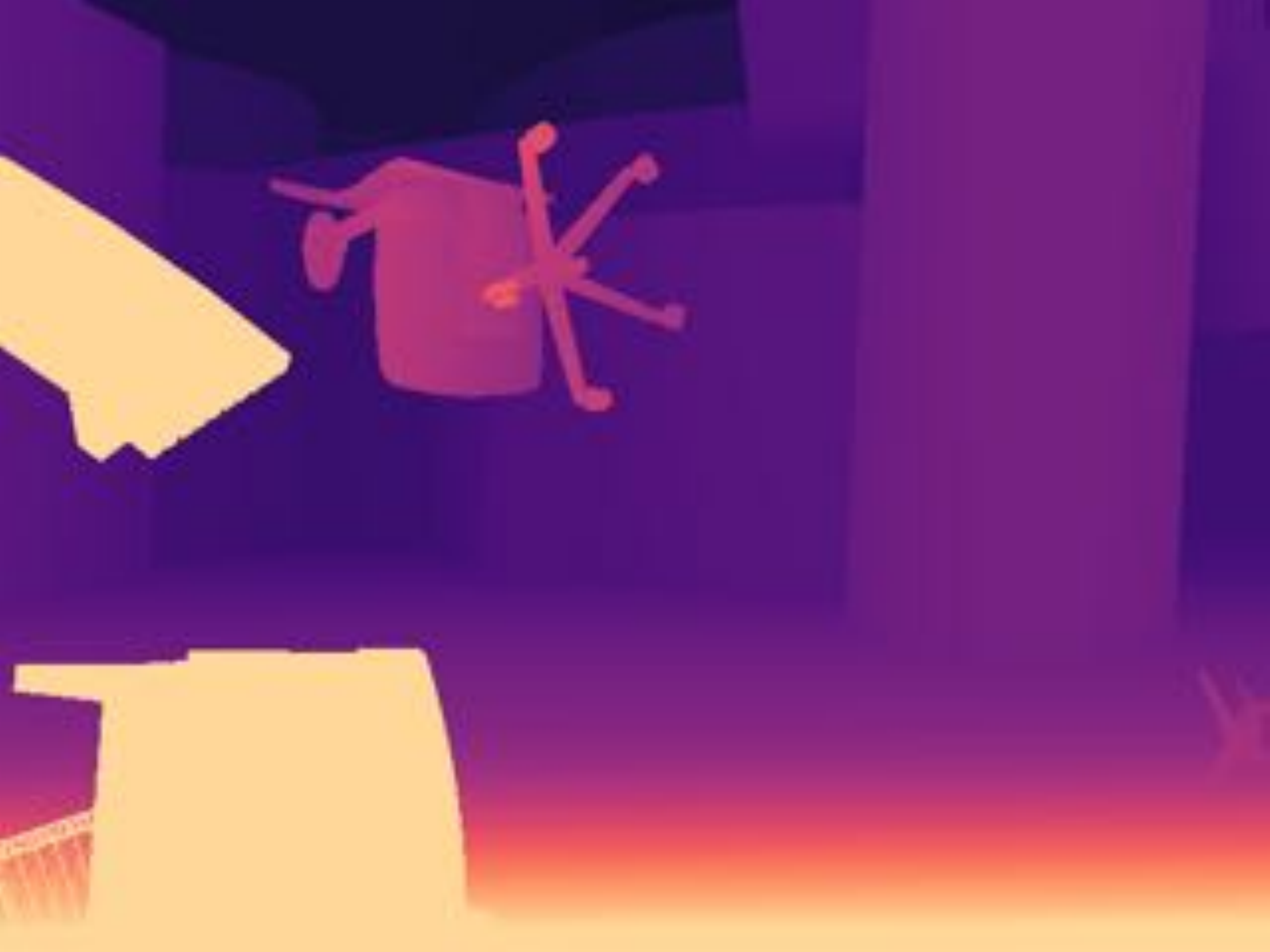}\\

\vspace{30mm}\\
\\\cmidrule{1-5}
\vspace{30mm}\\
\rotatebox[origin=c]{90}{\fontsize{\textw}{\texth} \selectfont Monodepth2\hspace{-270mm}}\hspace{10mm}
\includegraphics[width=\iw,height=\ih]{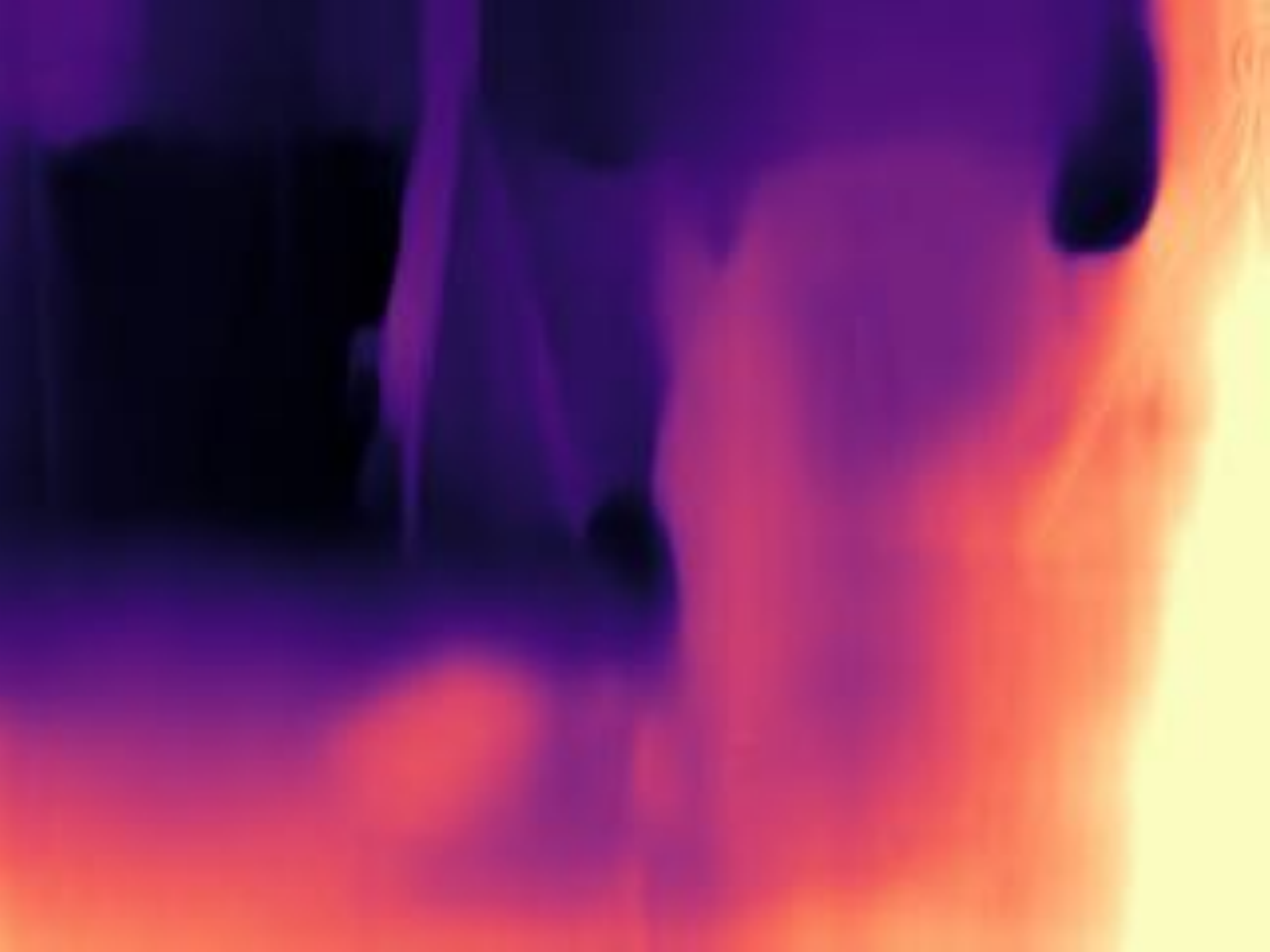} \qquad\qquad\quad &  
\includegraphics[width=\iw,height=\ih]{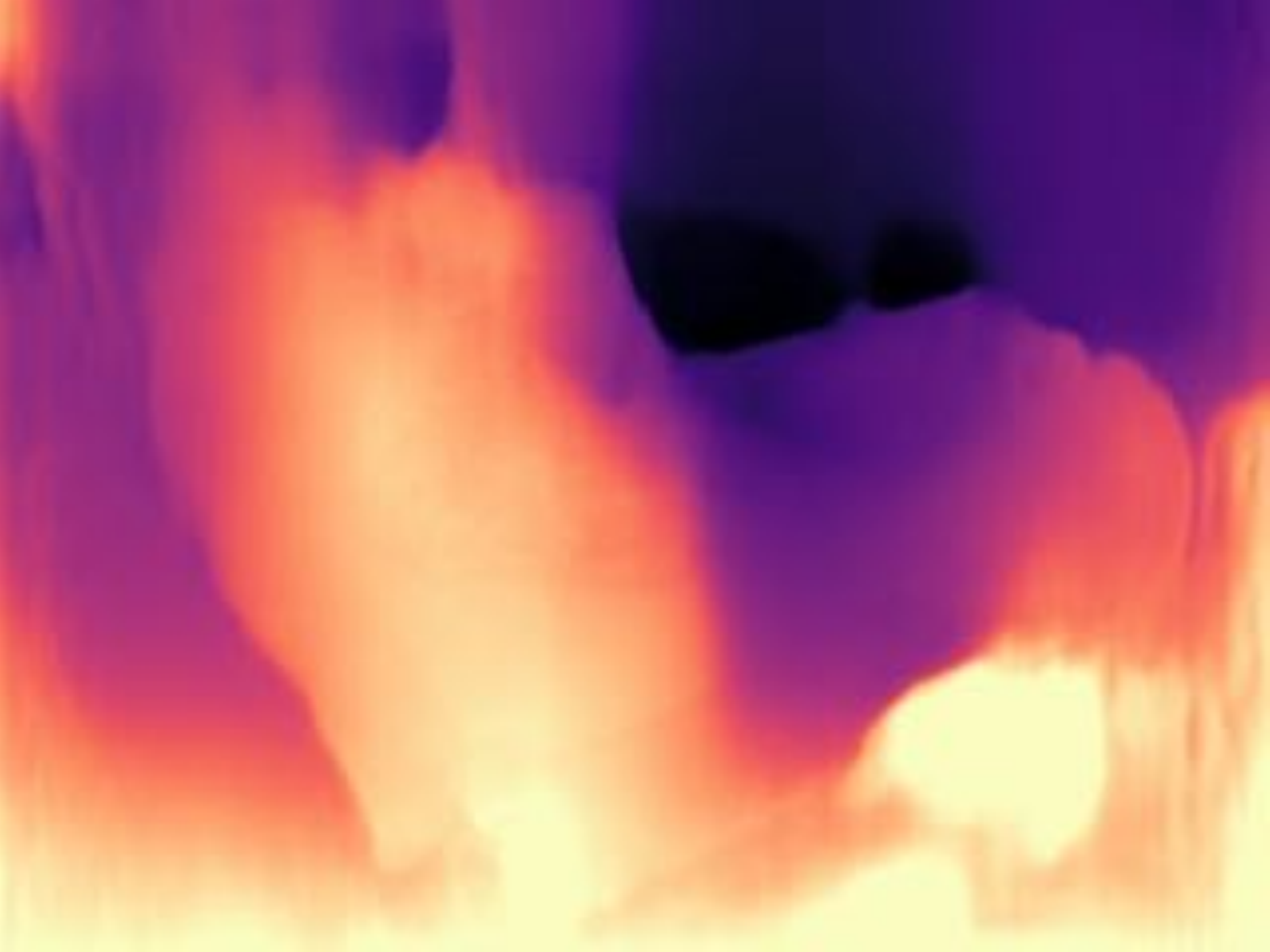} \qquad\qquad\quad &  
\includegraphics[width=\iw,height=\ih]{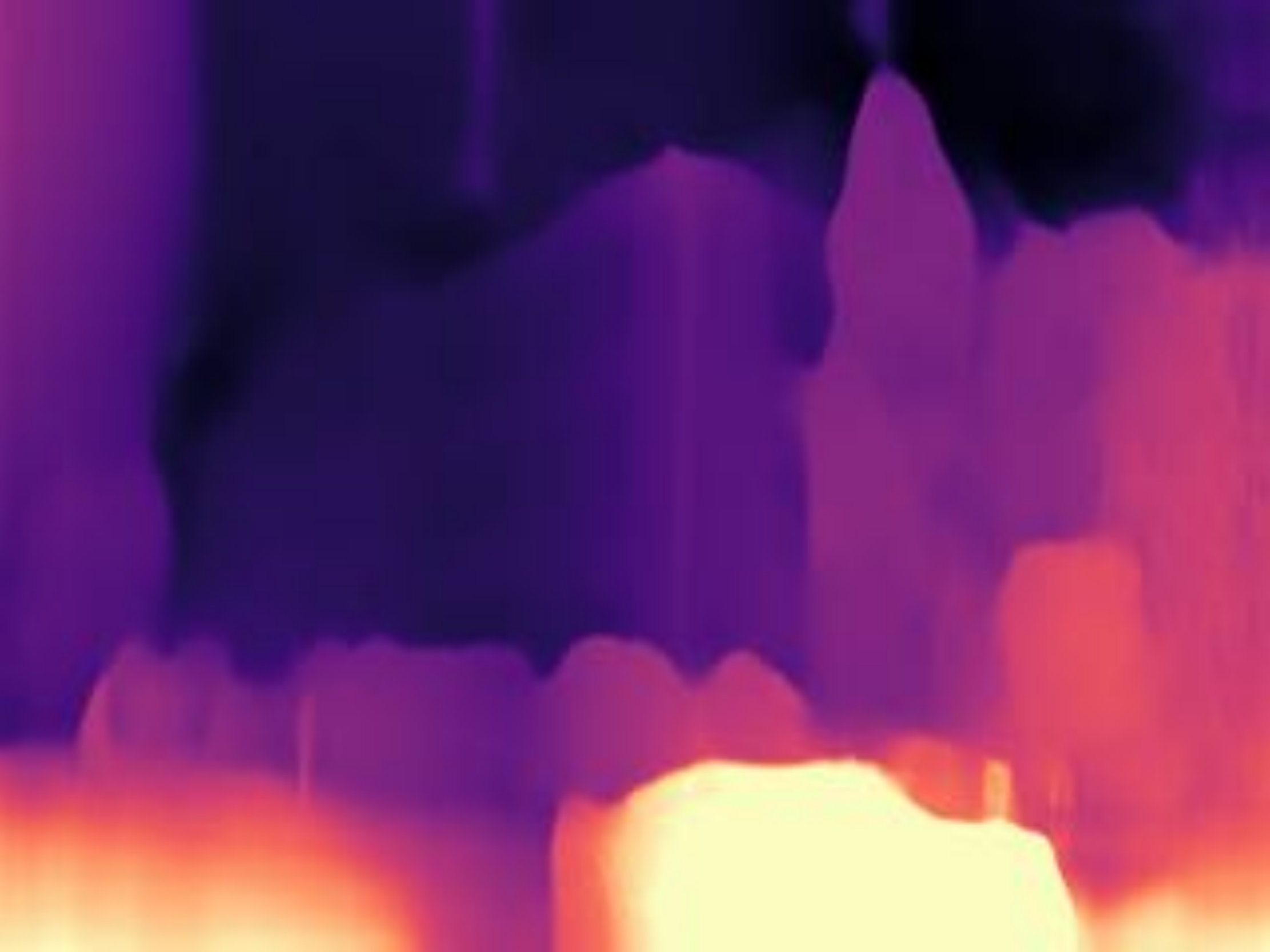} \qquad\qquad\quad &  
\includegraphics[width=\iw,height=\ih]{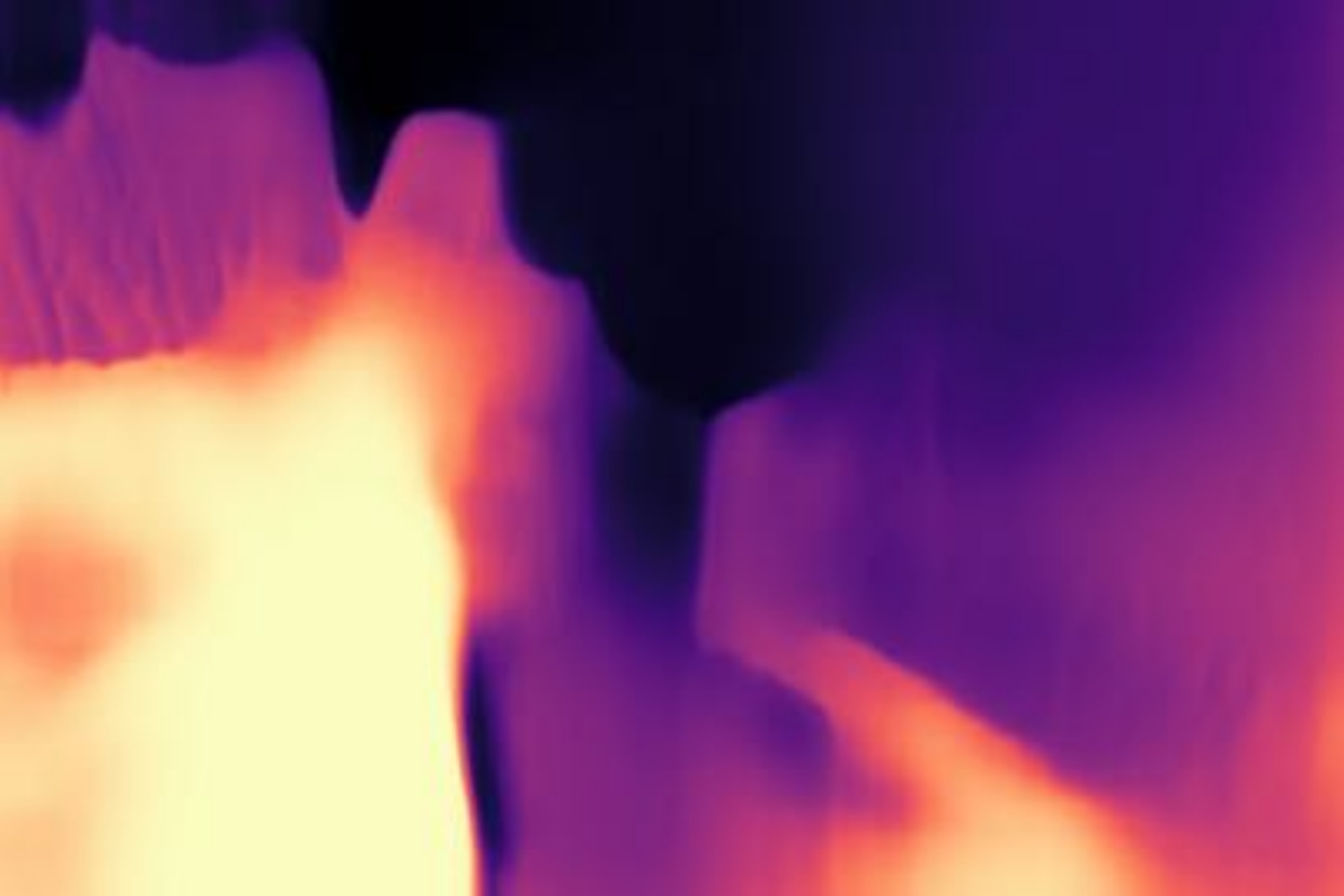} \qquad\qquad\quad &  
\includegraphics[width=\iw,height=\ih]{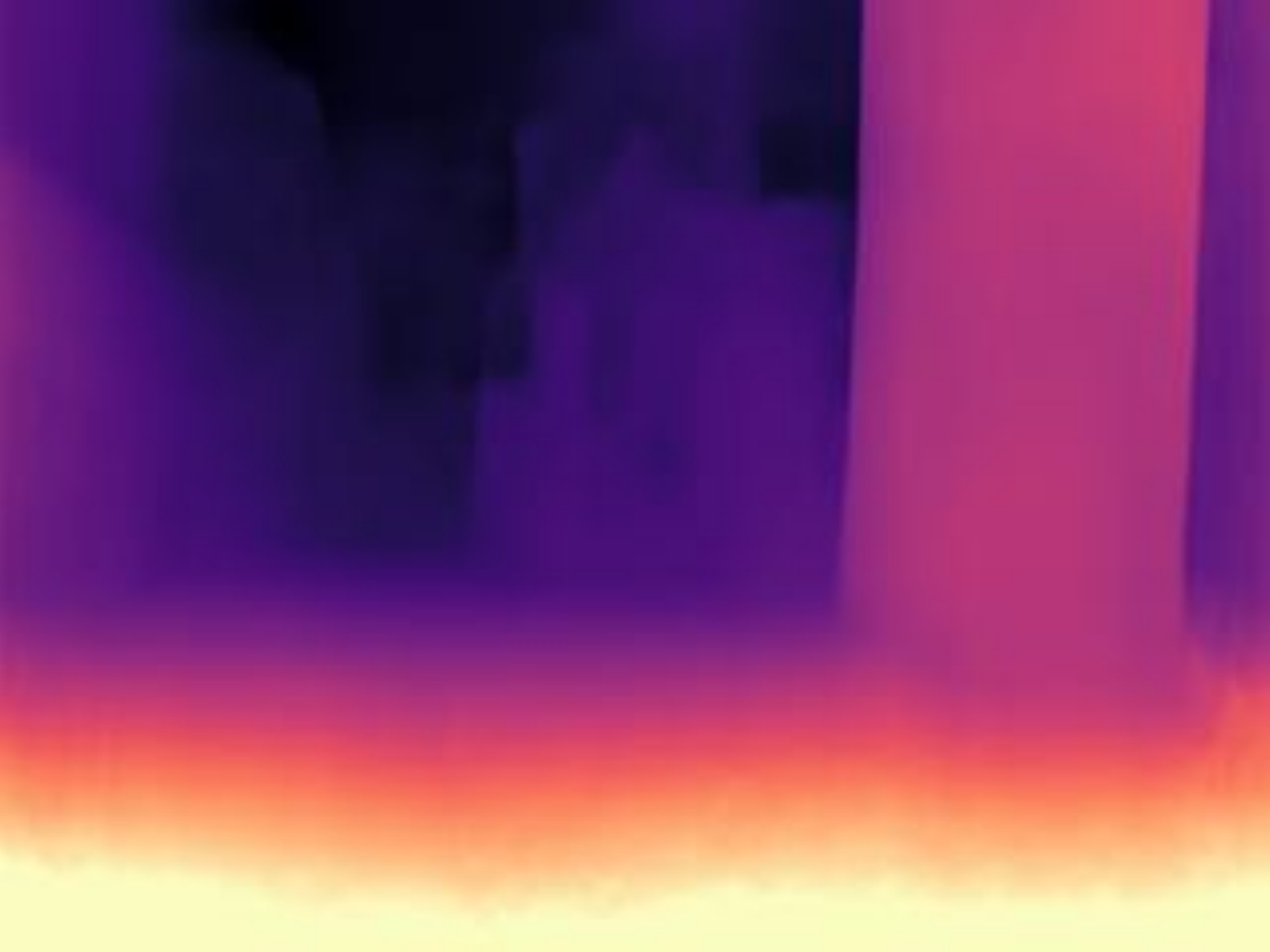}\\

\vspace{10mm}\\
\rotatebox[origin=c]{90}{\fontsize{\textw}{\texth} \selectfont PackNet-SfM\hspace{-270mm}}\hspace{25mm}
\includegraphics[width=\iw,height=\ih]{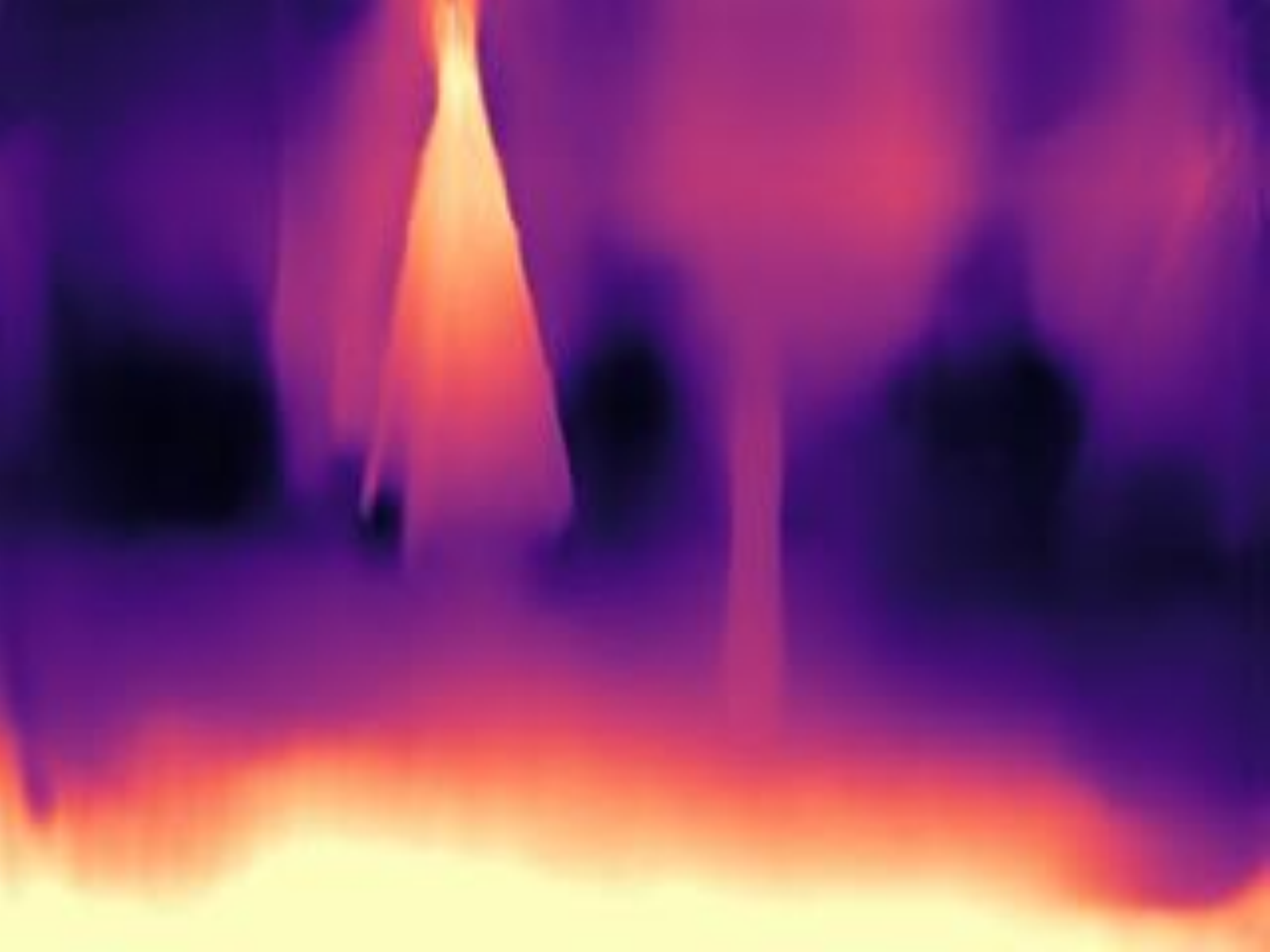} \qquad\qquad\quad &  
\includegraphics[width=\iw,height=\ih]{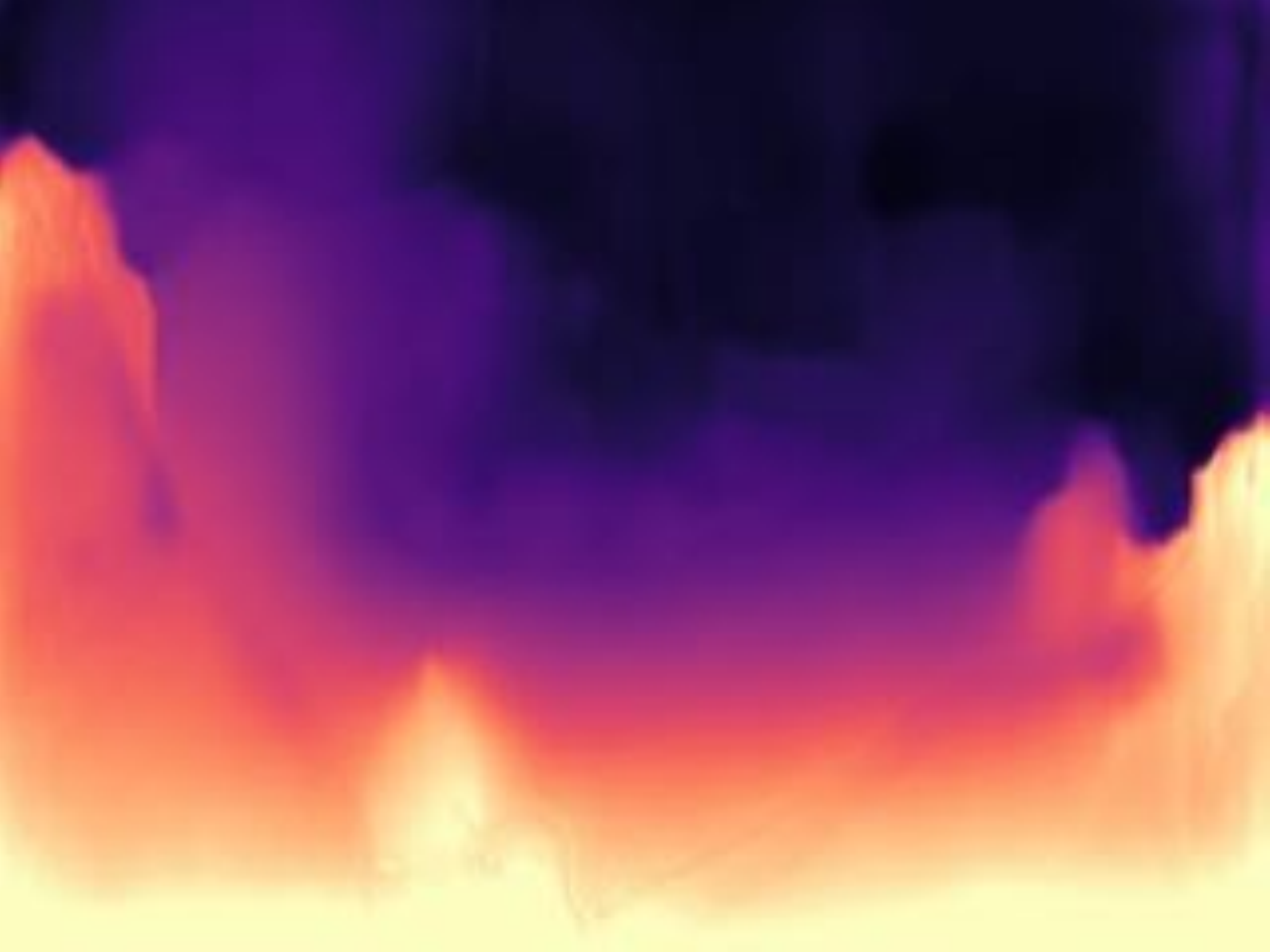} \qquad\qquad\quad &  
\includegraphics[width=\iw,height=\ih]{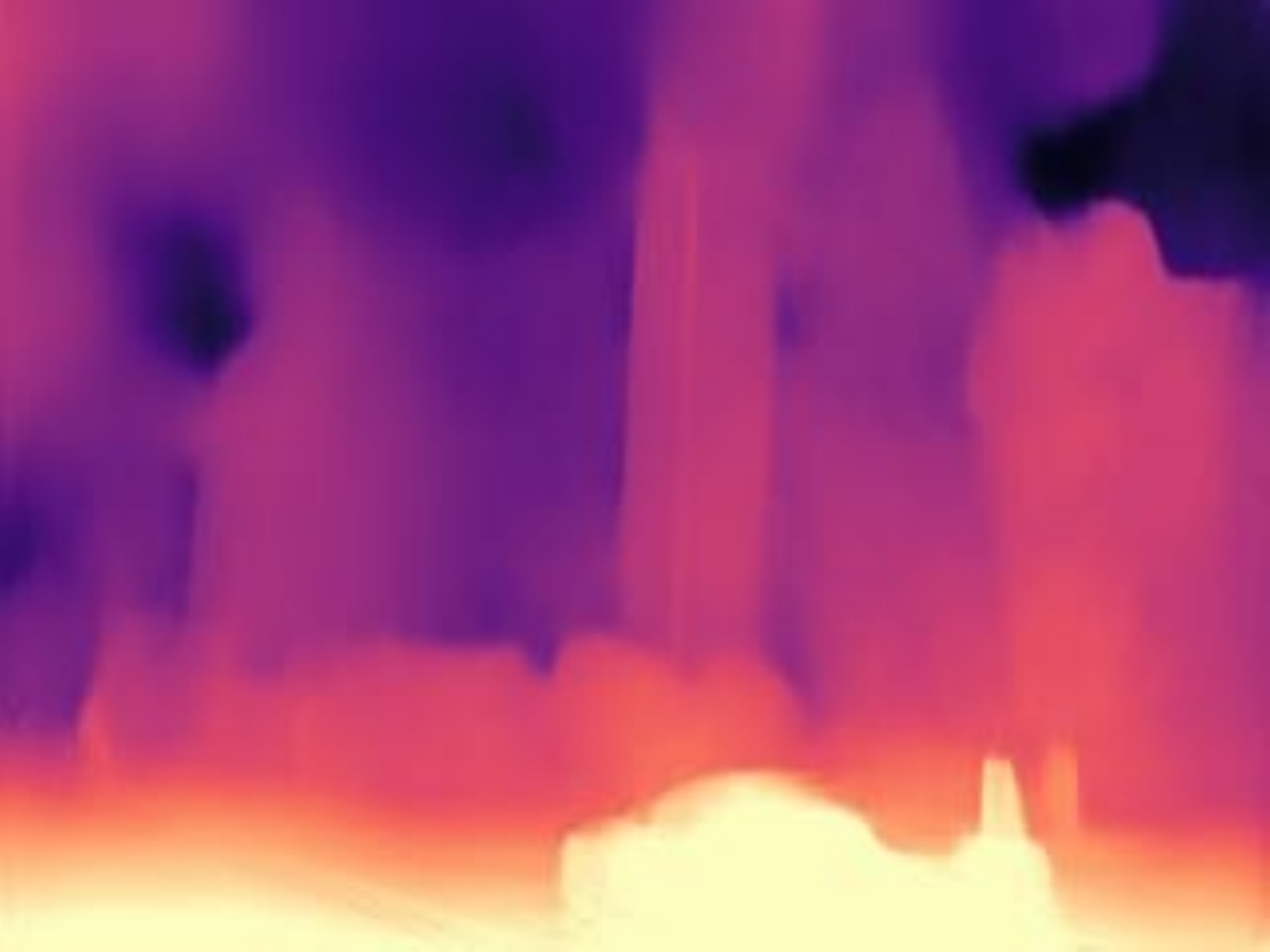} \qquad\qquad\quad &  
\includegraphics[width=\iw,height=\ih]{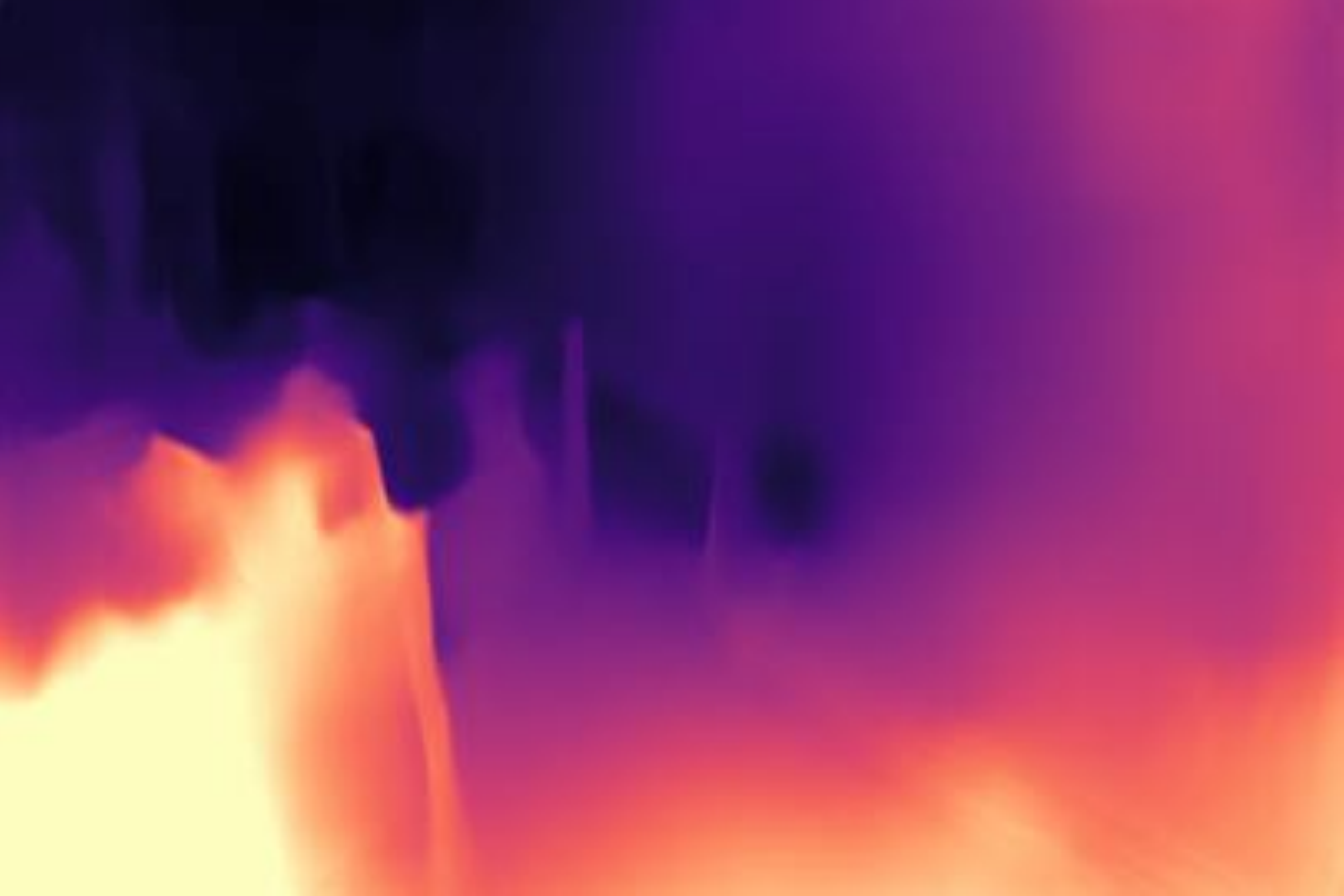} \qquad\qquad\quad &  
\includegraphics[width=\iw,height=\ih]{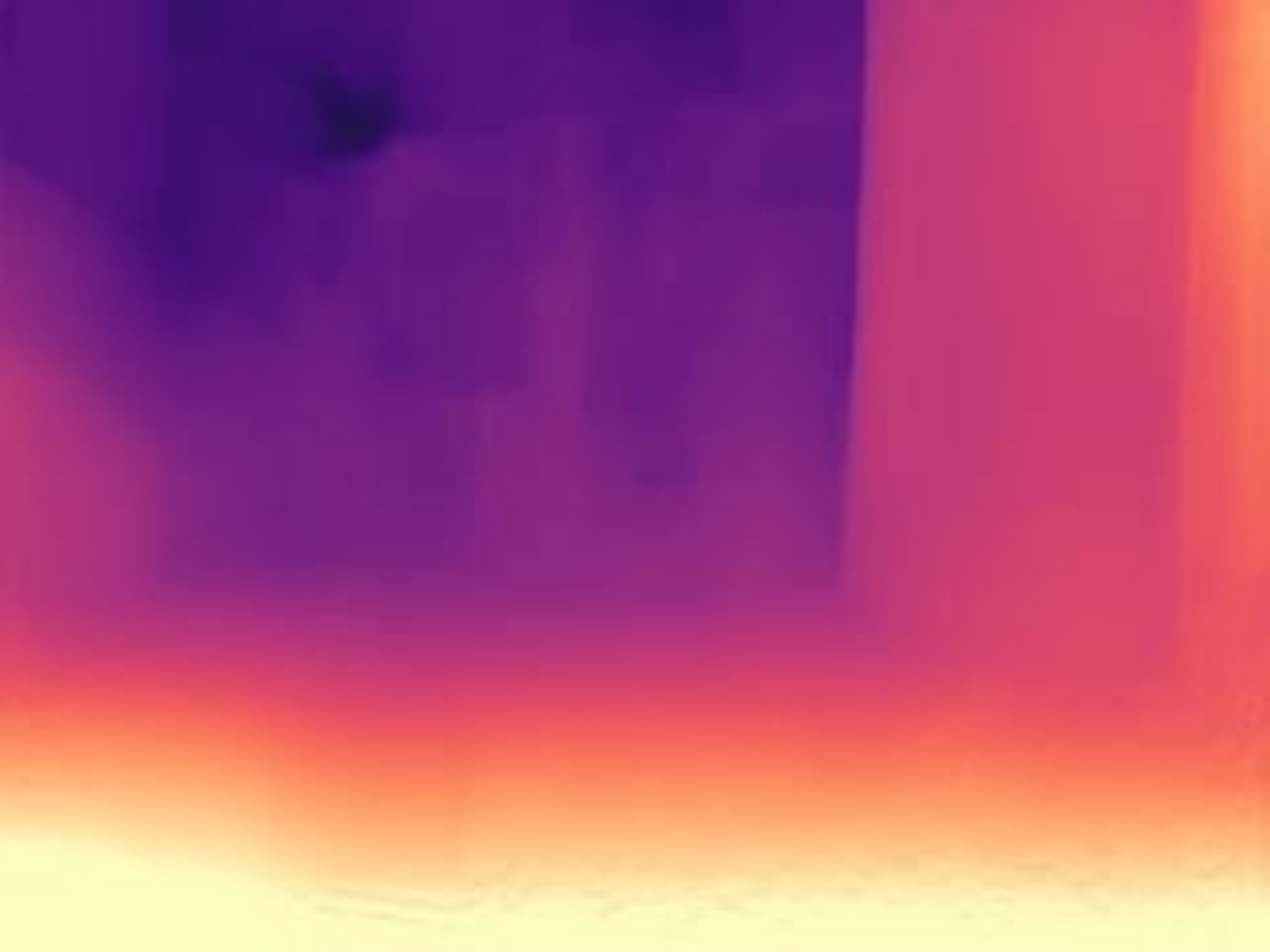}\\
\vspace{10mm}\\
\rotatebox[origin=c]{90}{\fontsize{\textw}{\texth} \selectfont R-MSFM6\hspace{-270mm}}\hspace{24mm}
\includegraphics[width=\iw,height=\ih]{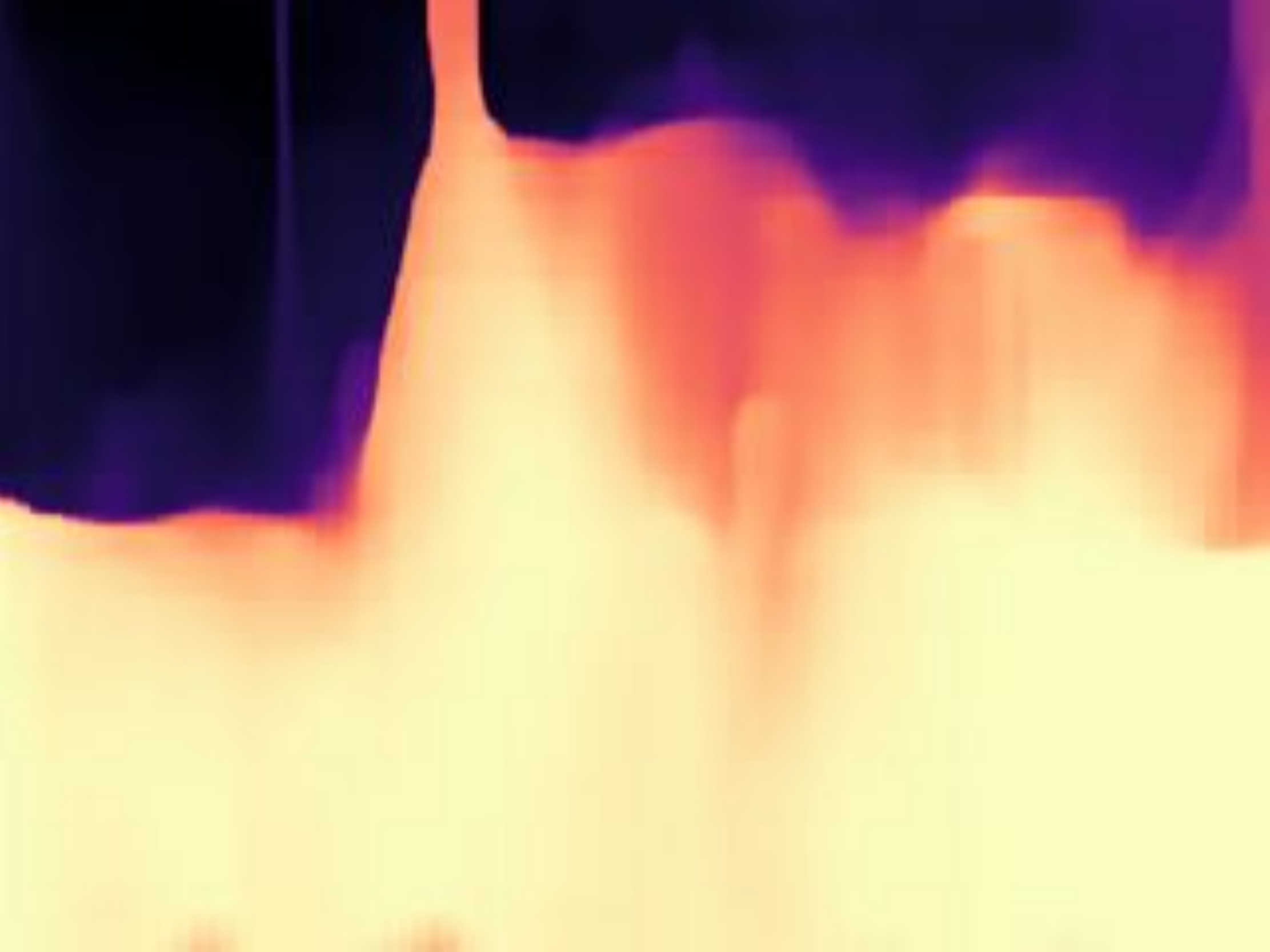} \qquad\qquad\quad &  
\includegraphics[width=\iw,height=\ih]{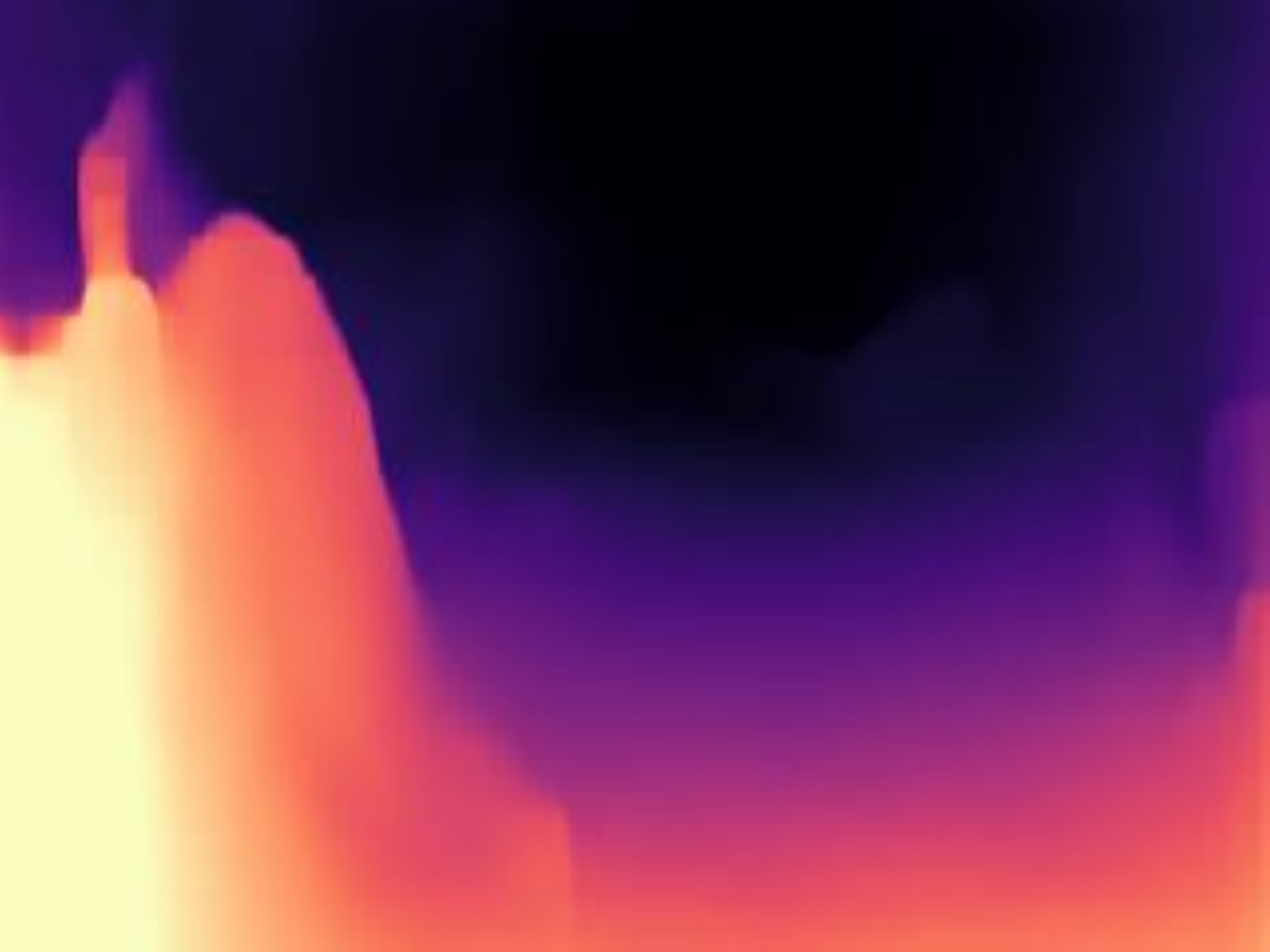} \qquad\qquad\quad &  
\includegraphics[width=\iw,height=\ih]{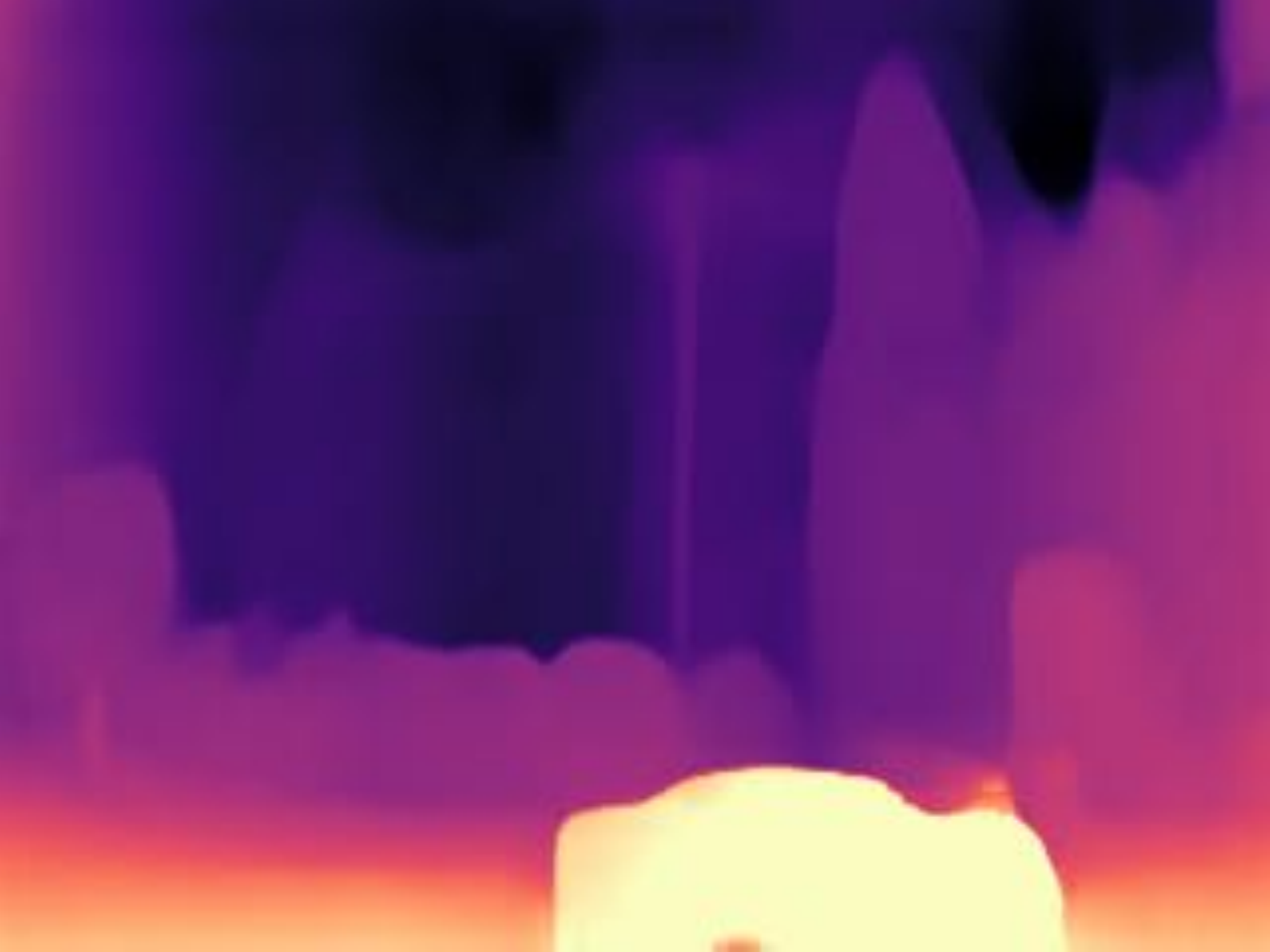} \qquad\qquad\quad &  
\includegraphics[width=\iw,height=\ih]{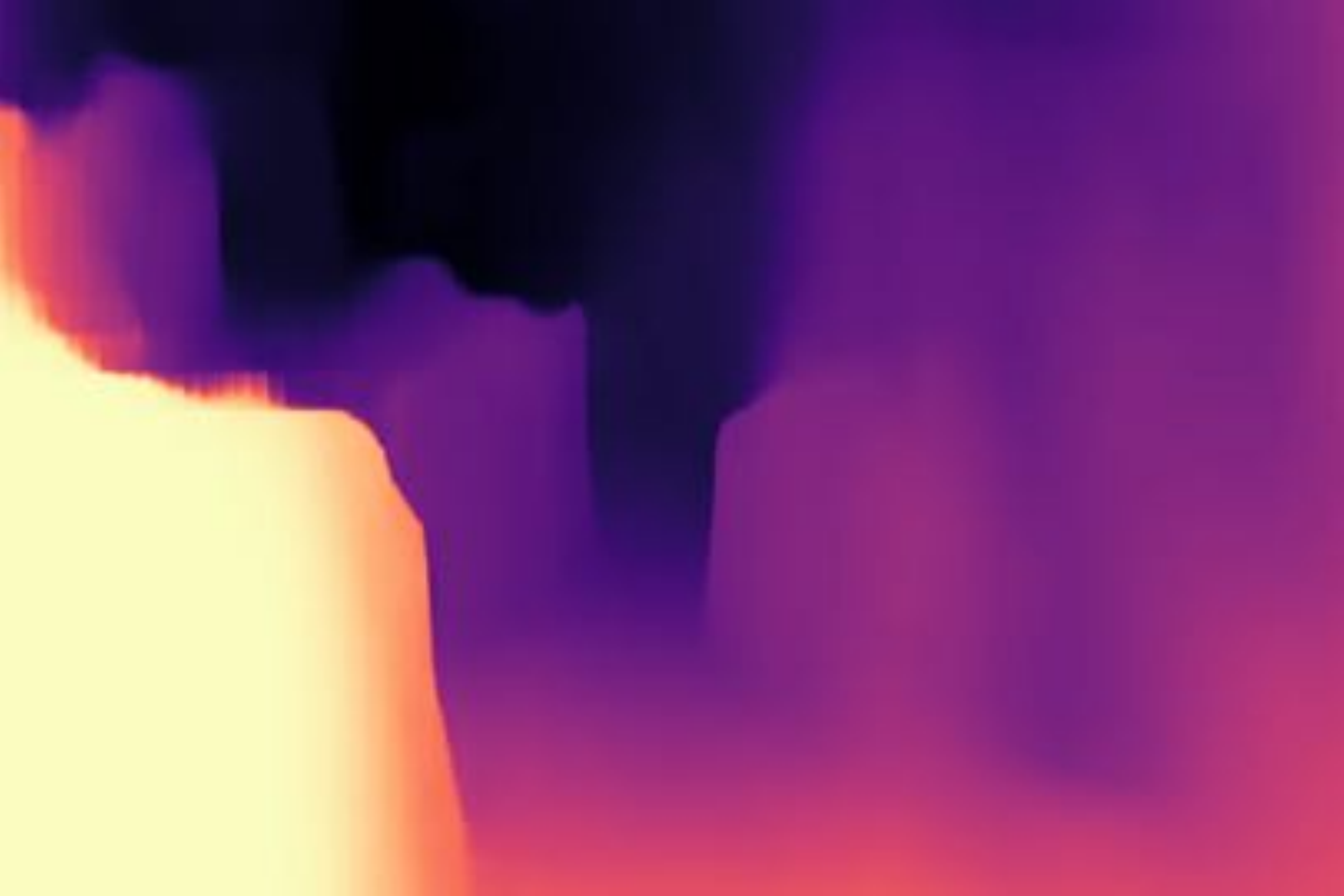} \qquad\qquad\quad &  
\includegraphics[width=\iw,height=\ih]{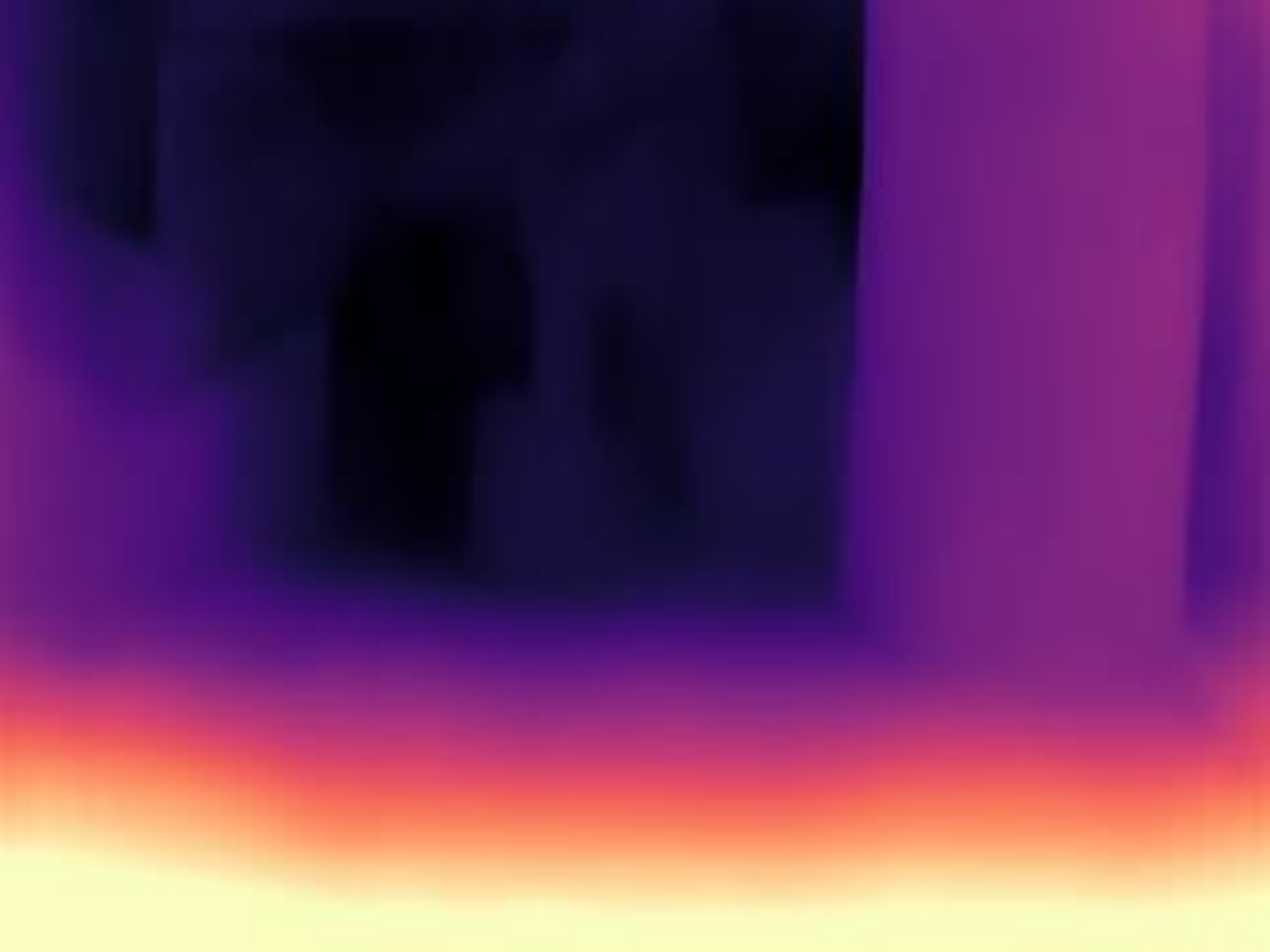}\\
\vspace{10mm}\\
\rotatebox[origin=c]{90}{\fontsize{\textw}{\texth} \selectfont MF-ConvNeXt\hspace{-260mm}}\hspace{24mm}
\includegraphics[width=\iw,height=\ih]{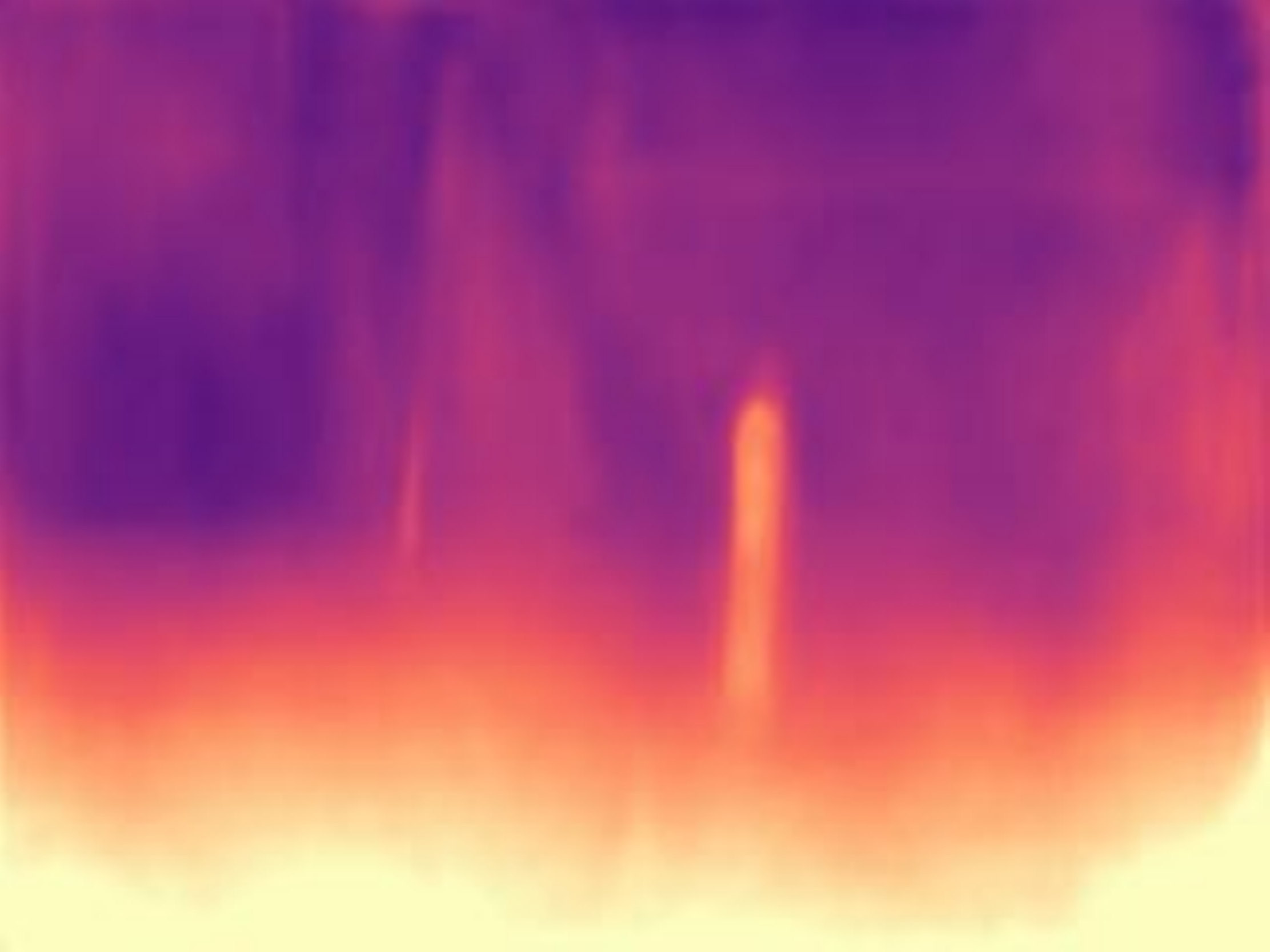} \qquad\qquad\quad & 
\includegraphics[width=\iw,height=\ih]{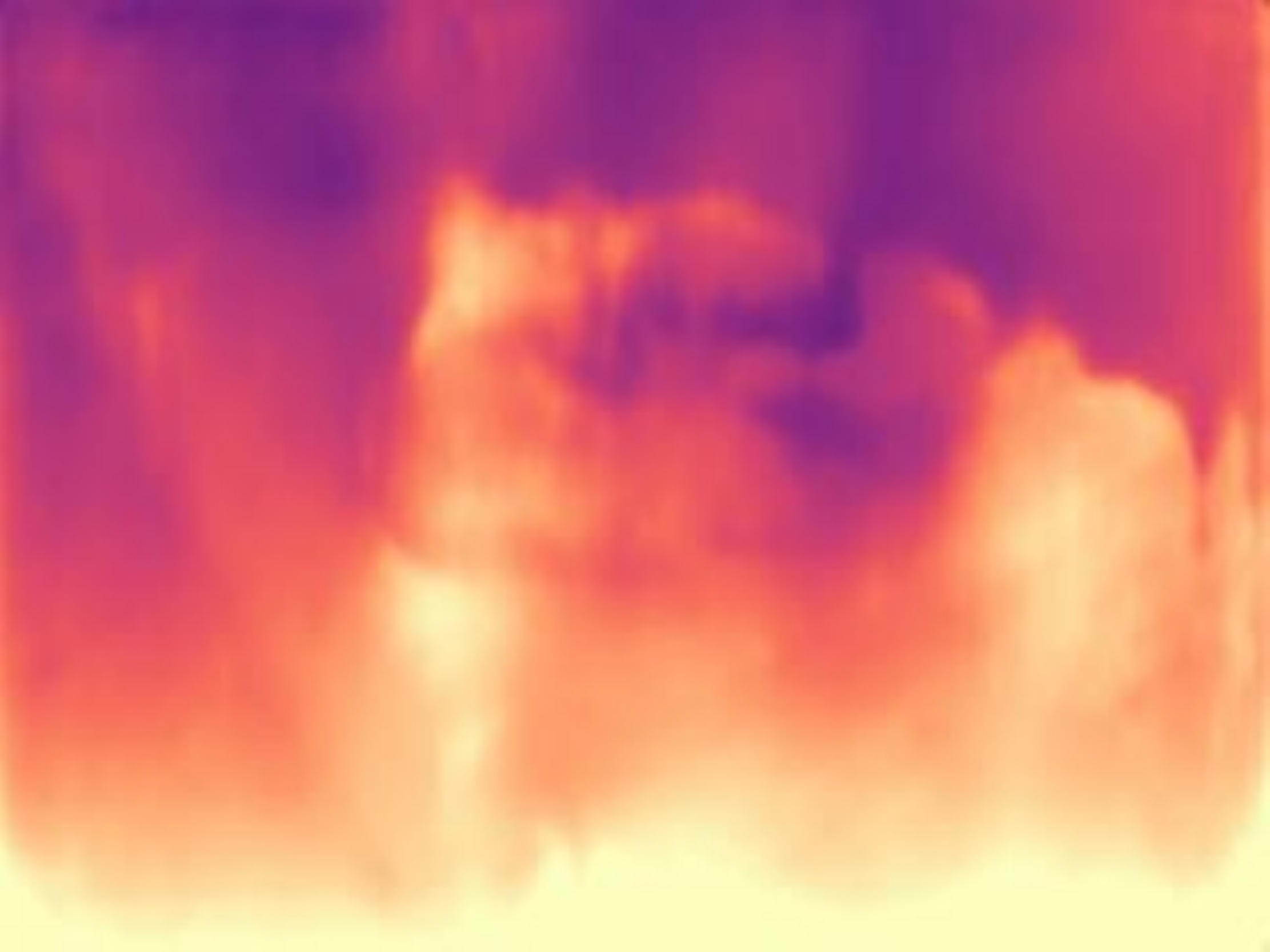} \qquad\qquad\quad & 
\includegraphics[width=\iw,height=\ih]{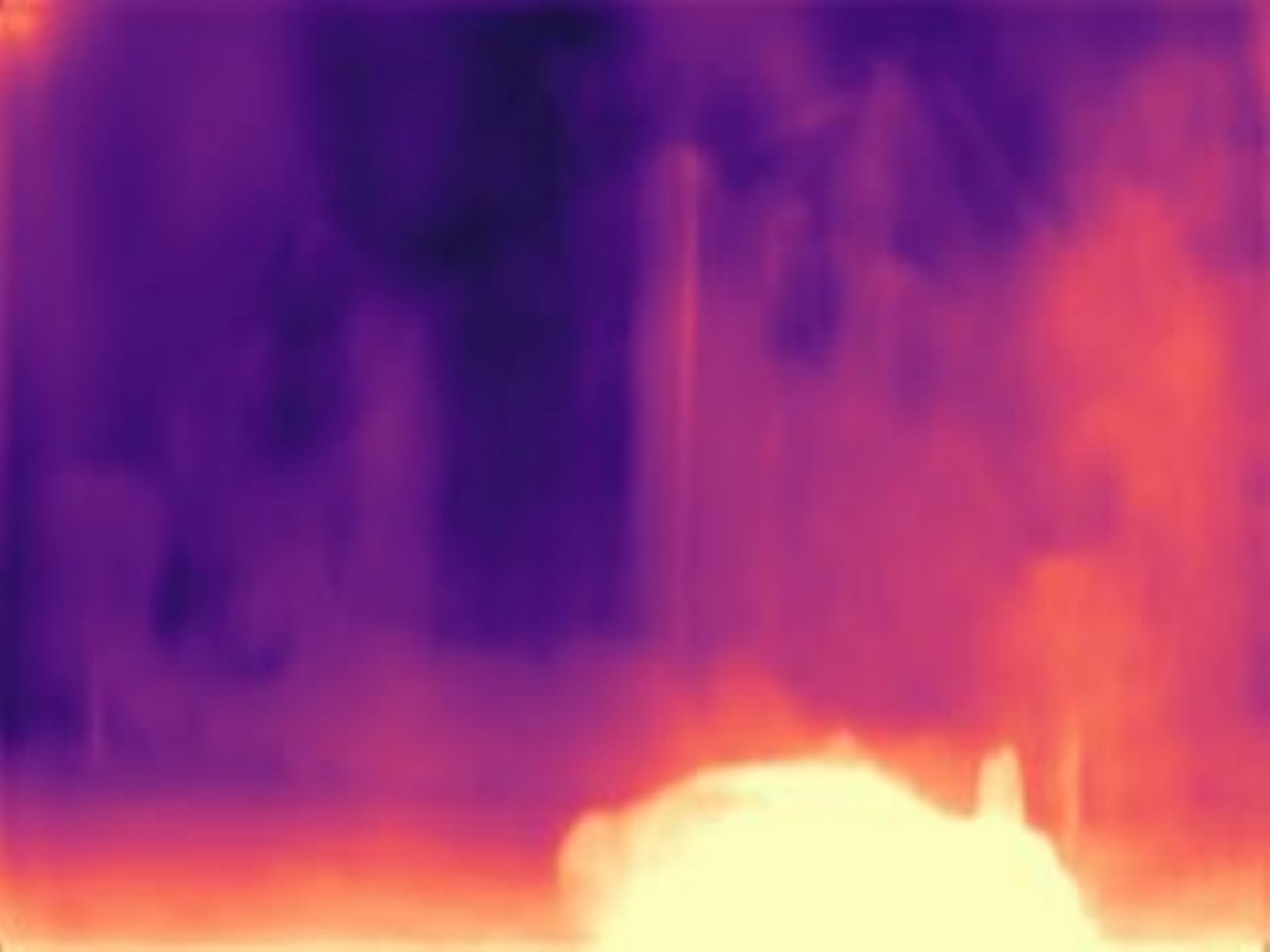} \qquad\qquad\quad & 
\includegraphics[width=\iw,height=\ih]{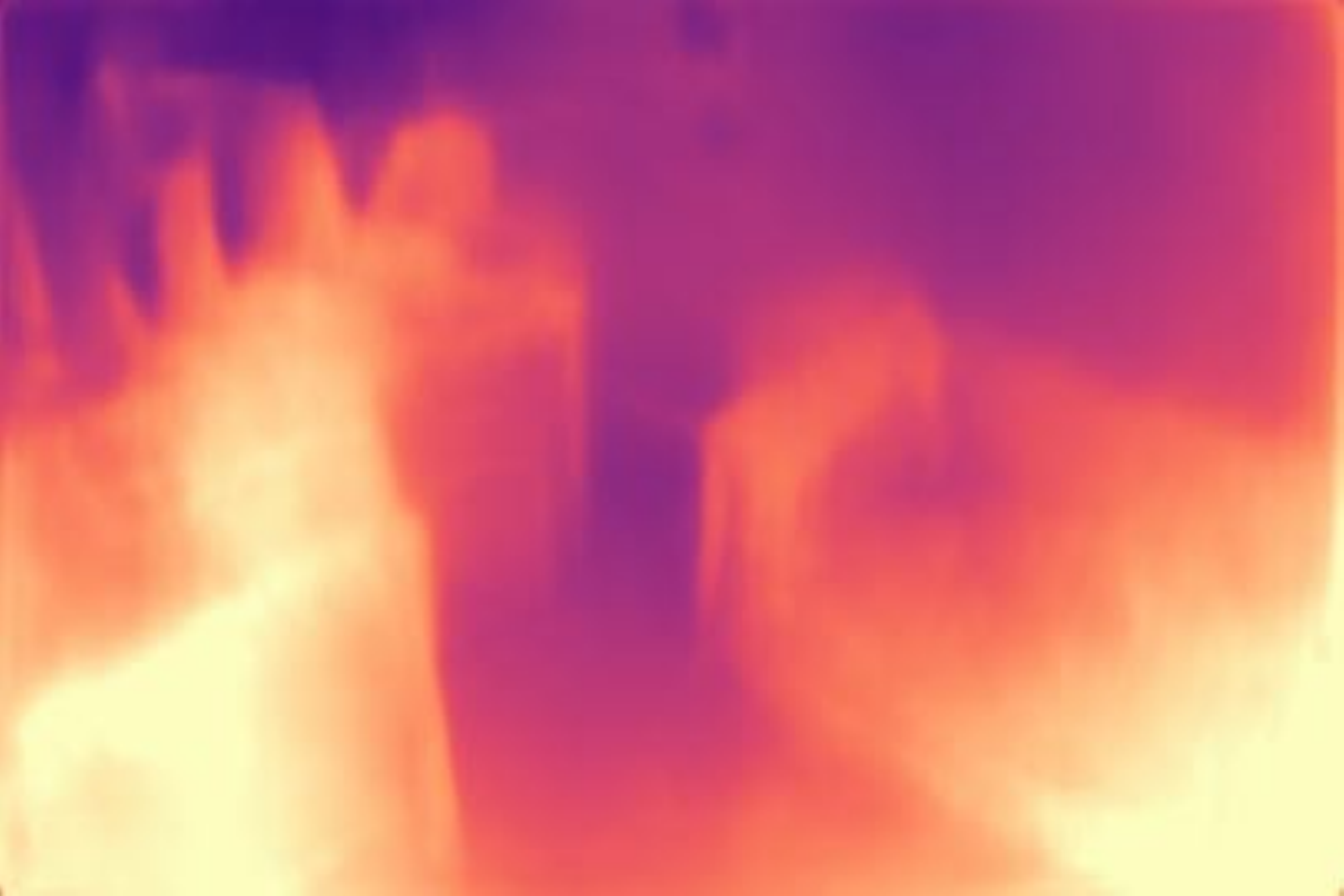} \qquad\qquad\quad & 
\includegraphics[width=\iw,height=\ih]{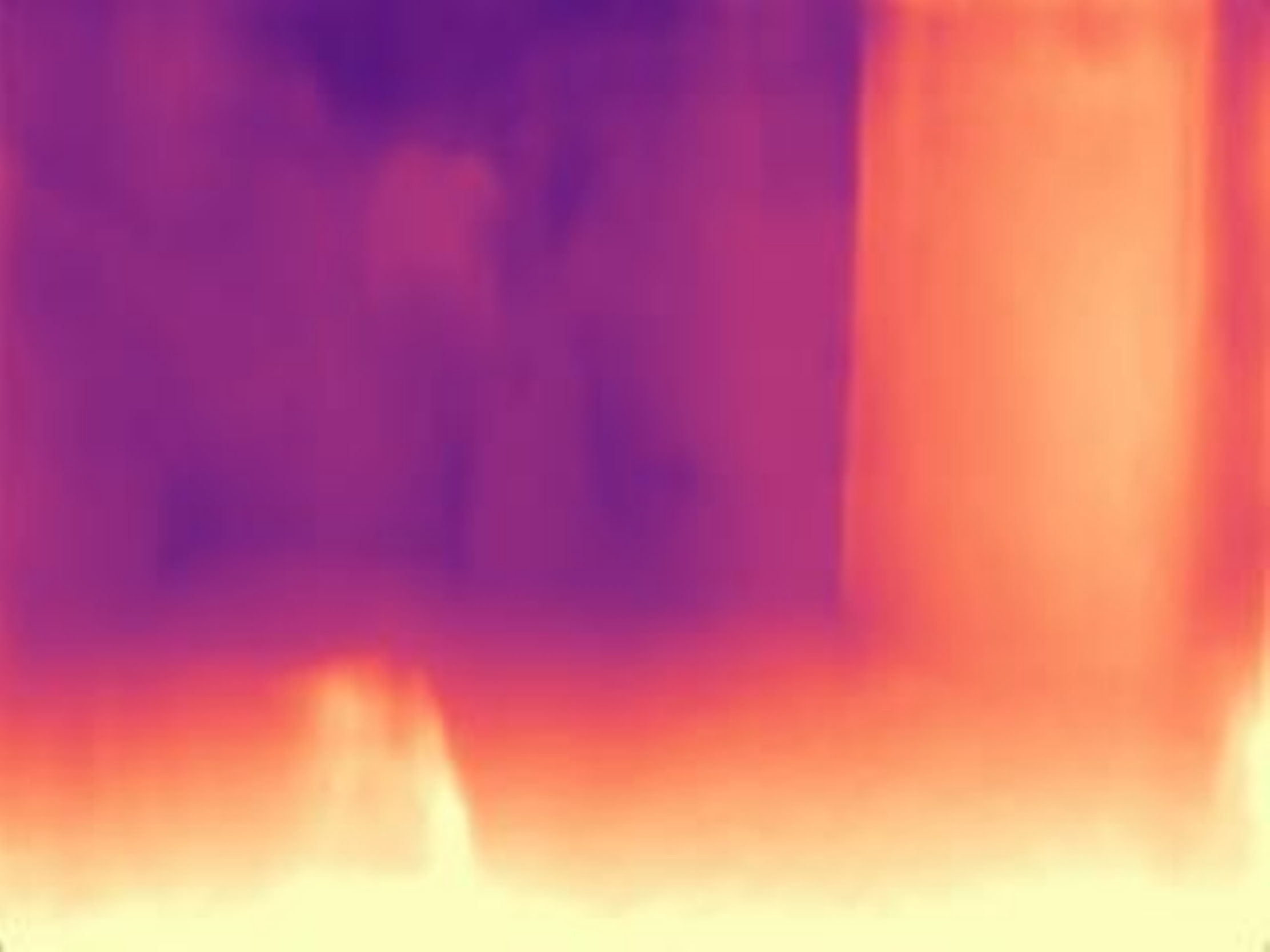}\\
\vspace{10mm}\\
\rotatebox[origin=c]{90}{\fontsize{\textw}{\texth} \selectfont MF-SLaK\hspace{-270mm}}\hspace{24mm}
\includegraphics[width=\iw,height=\ih]{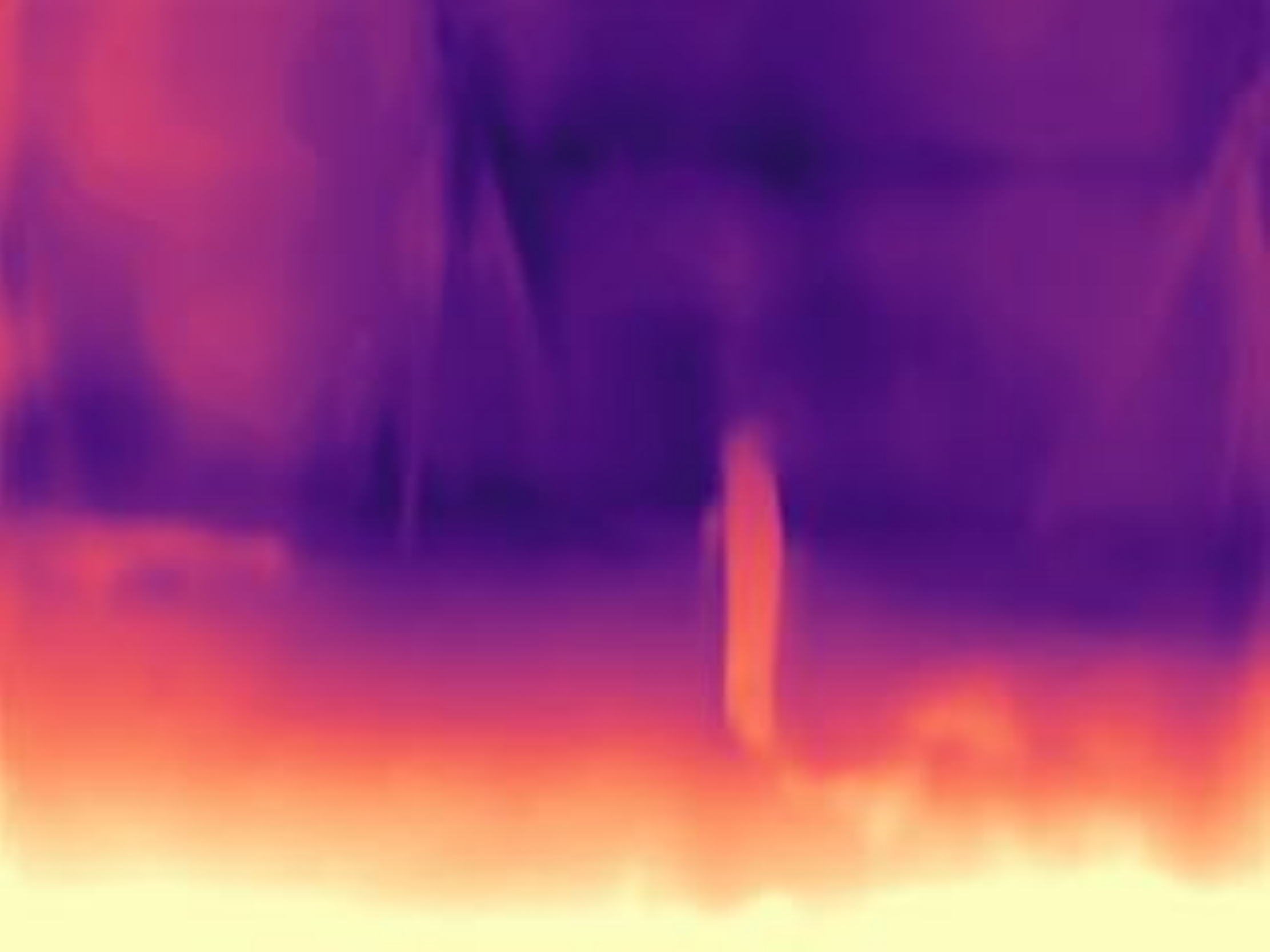} \qquad\qquad\quad & 
\includegraphics[width=\iw,height=\ih]{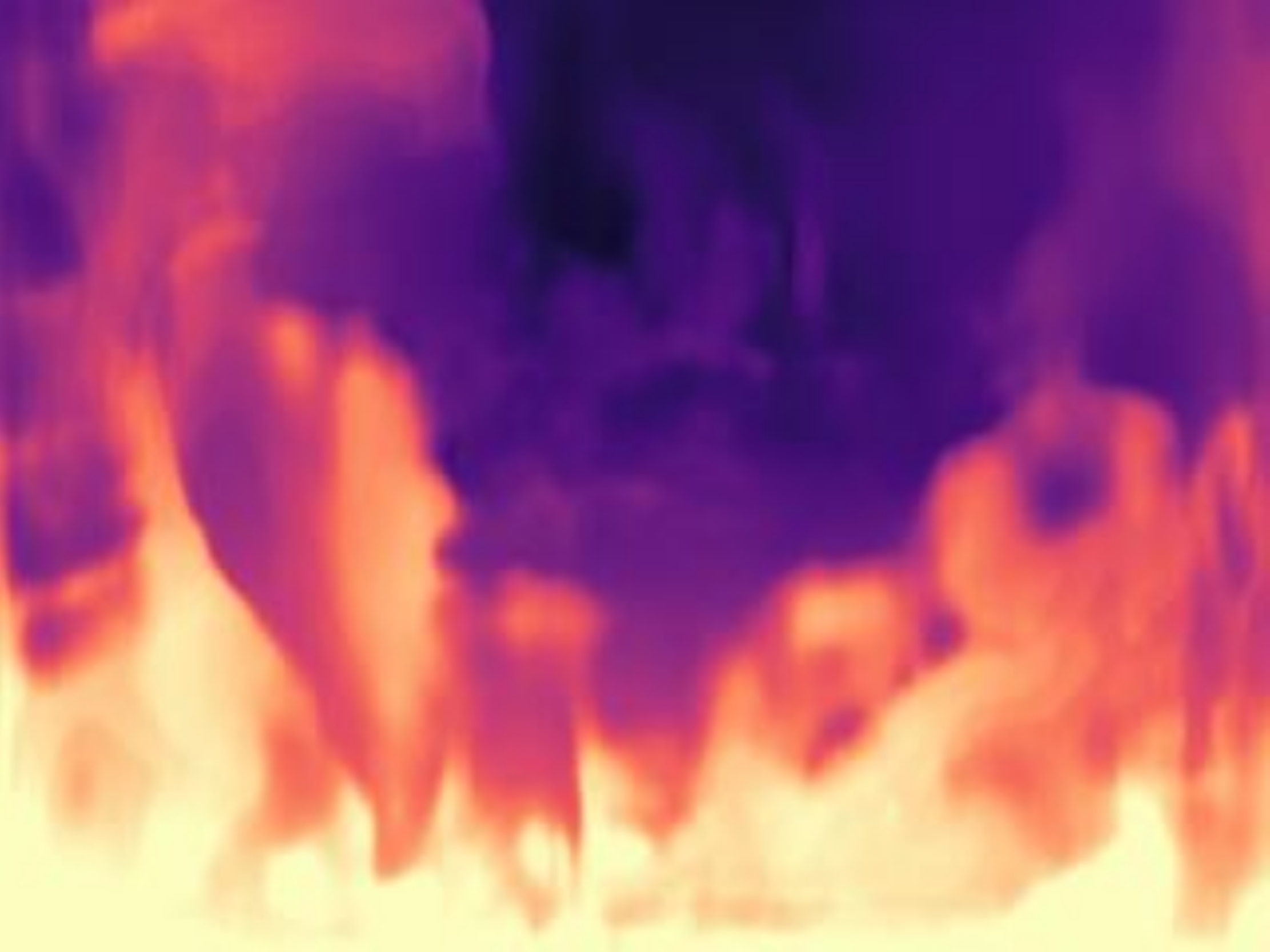} \qquad\qquad\quad & 
\includegraphics[width=\iw,height=\ih]{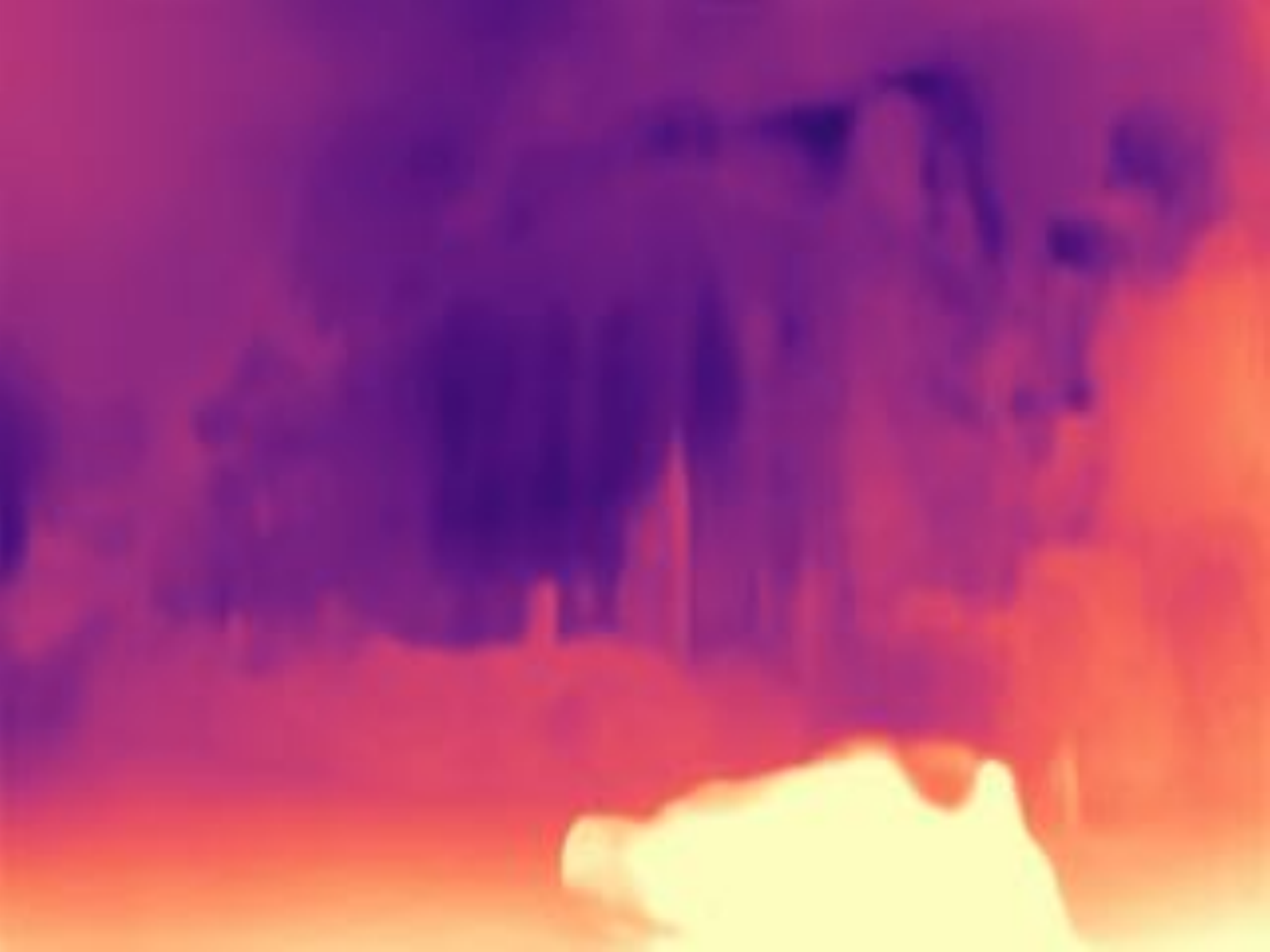} \qquad\qquad\quad & 
\includegraphics[width=\iw,height=\ih]{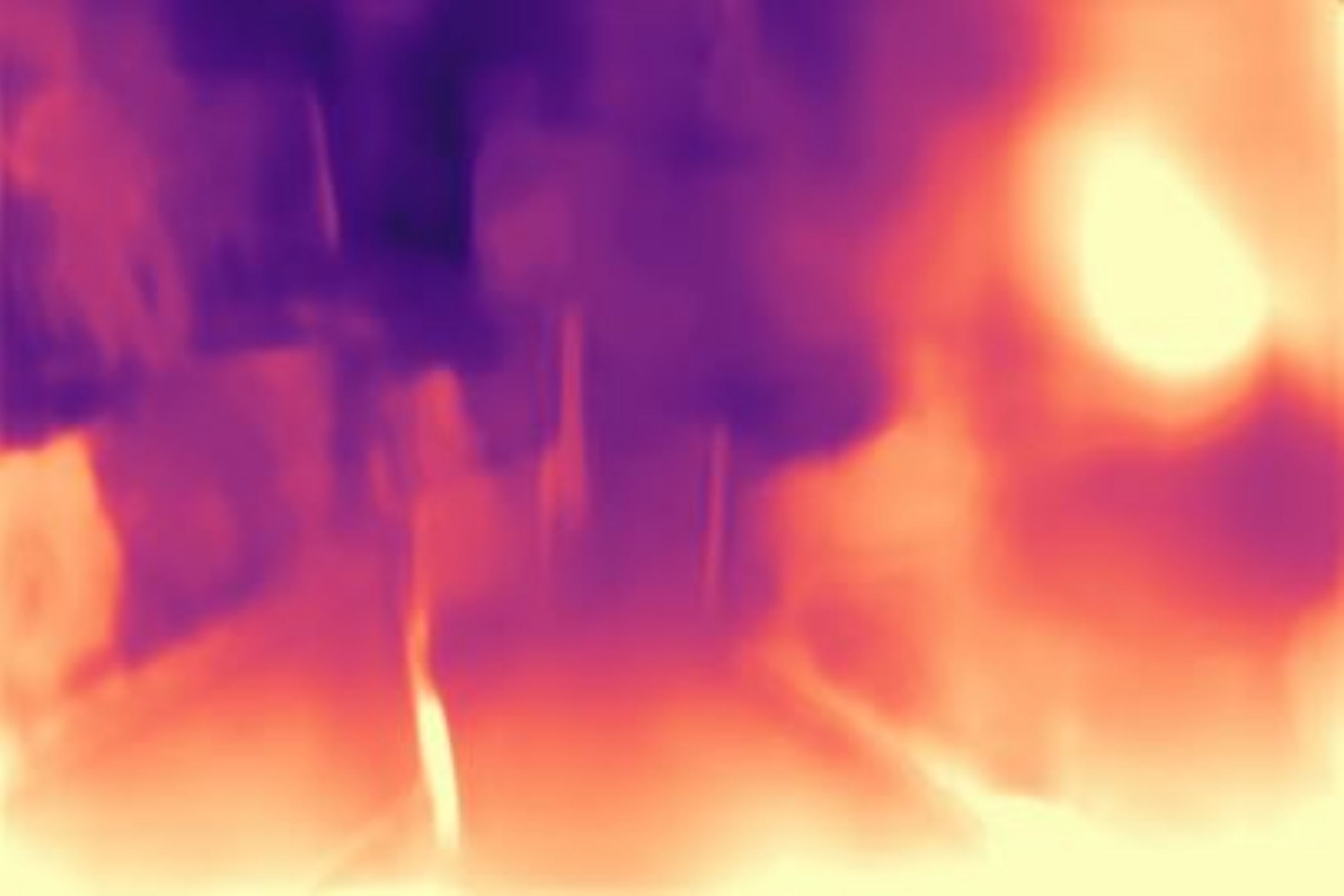} \qquad\qquad\quad & 
\includegraphics[width=\iw,height=\ih]{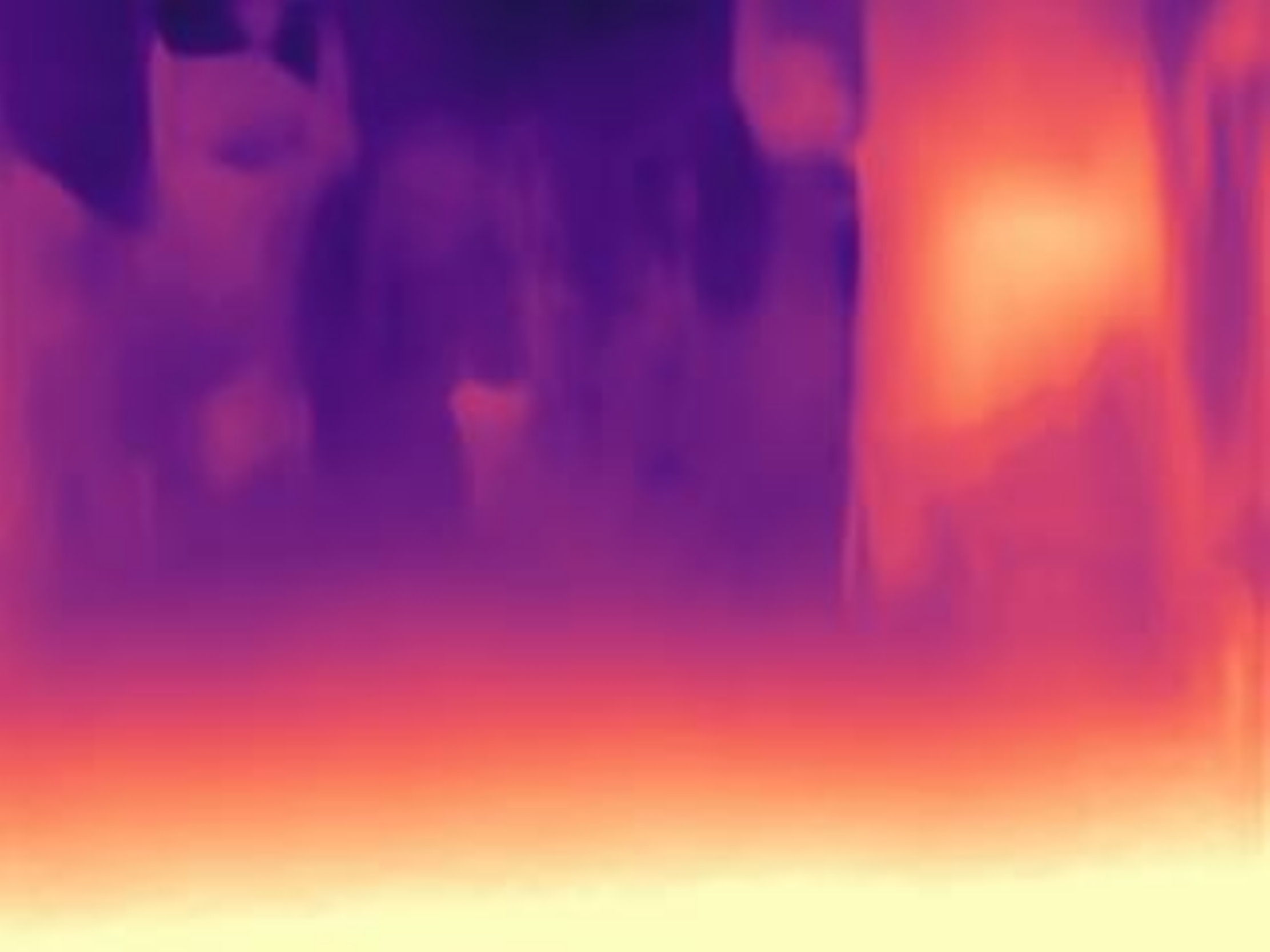}\\
\vspace{30mm}\\
\multicolumn{5}{c}{\fontsize{\w}{\h} \selectfont (a) Self-supervised CNN-based methods } & 
\vspace{30mm}\\
\rotatebox[origin=c]{90}{\fontsize{\textw}{\texth} \selectfont MF-ViT\hspace{-270mm}}\hspace{24mm}
\includegraphics[width=\iw,height=\ih]{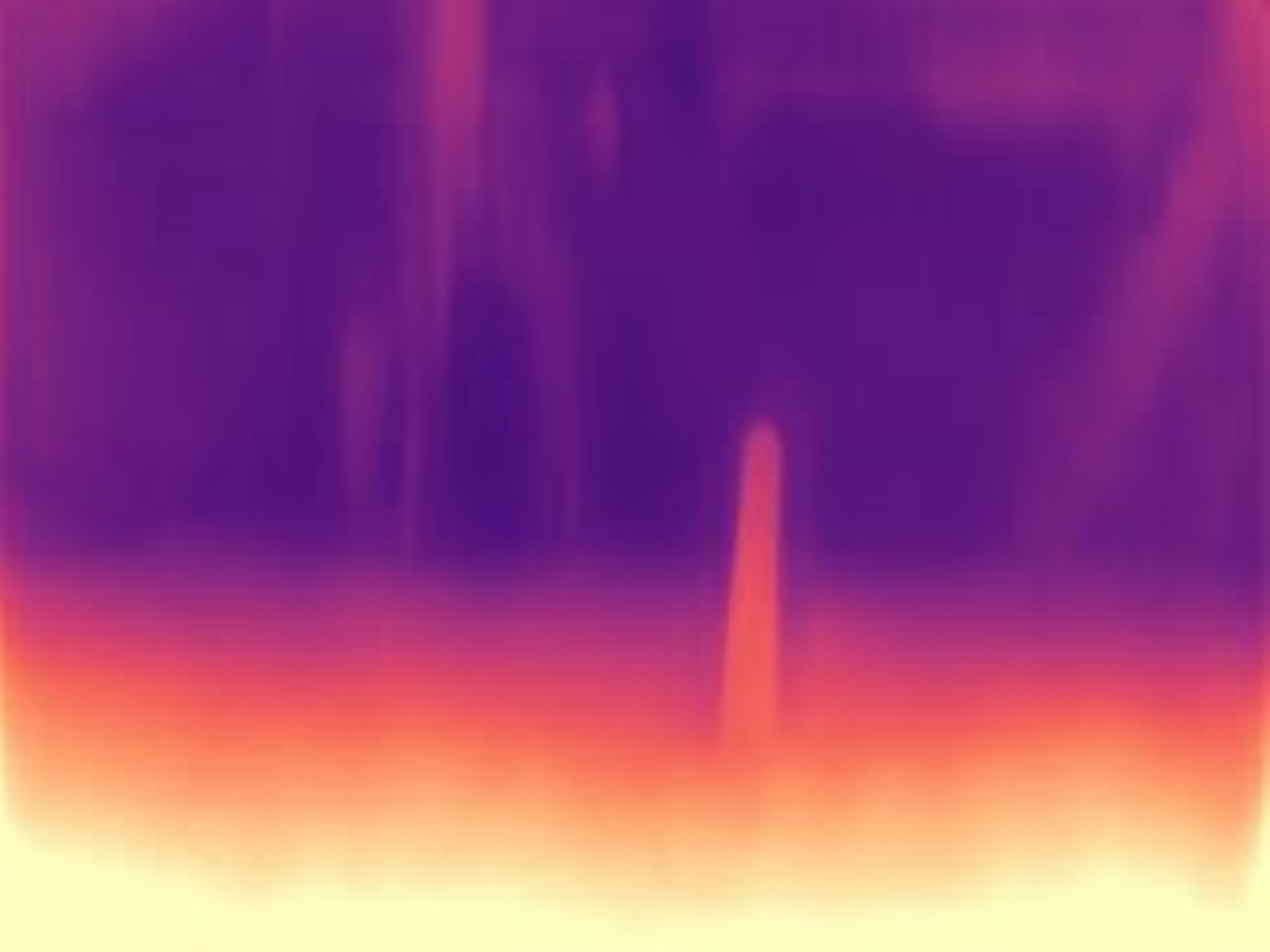} \qquad\qquad\quad & 
\includegraphics[width=\iw,height=\ih]{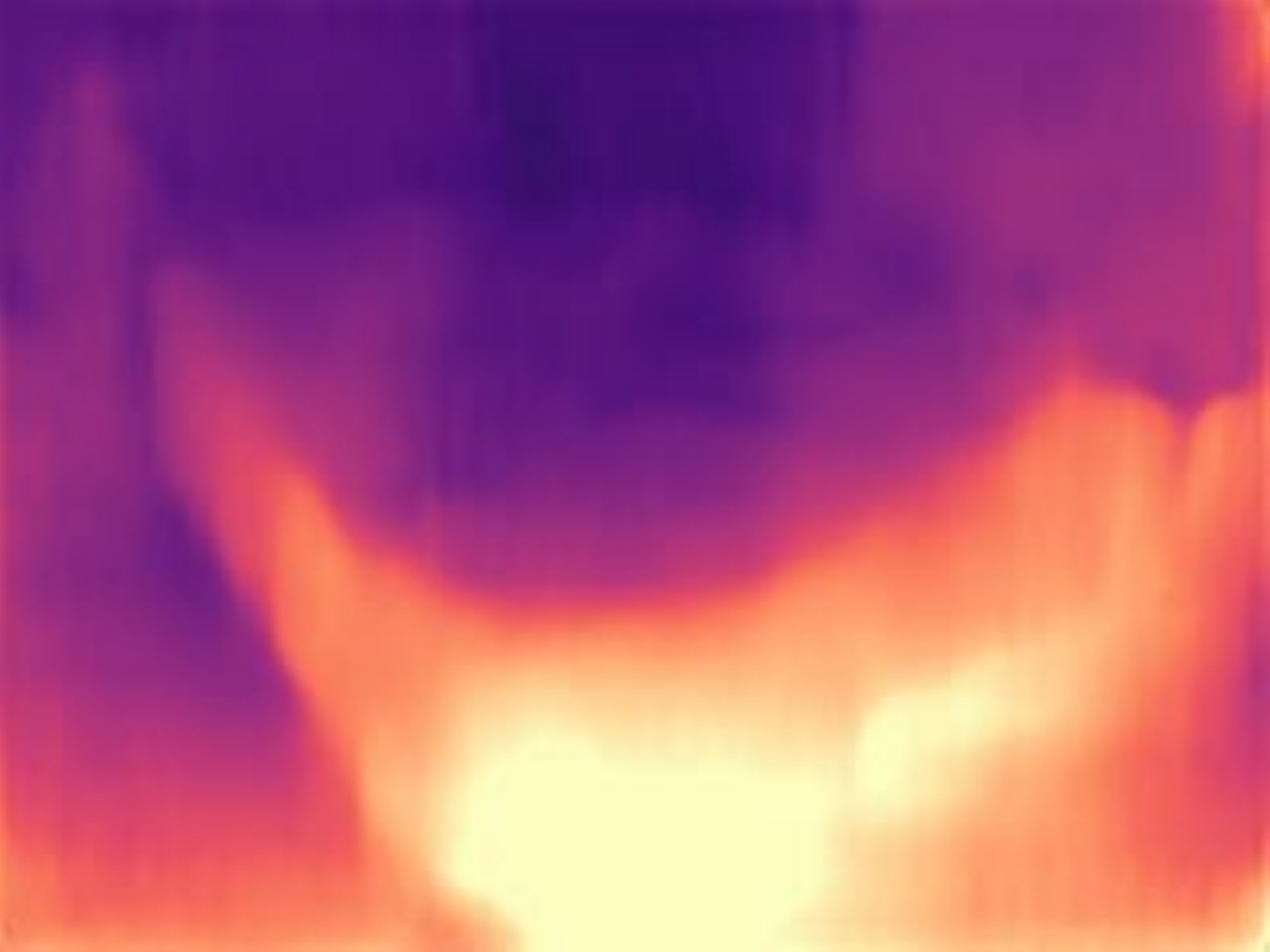} \qquad\qquad\quad & 
\includegraphics[width=\iw,height=\ih]{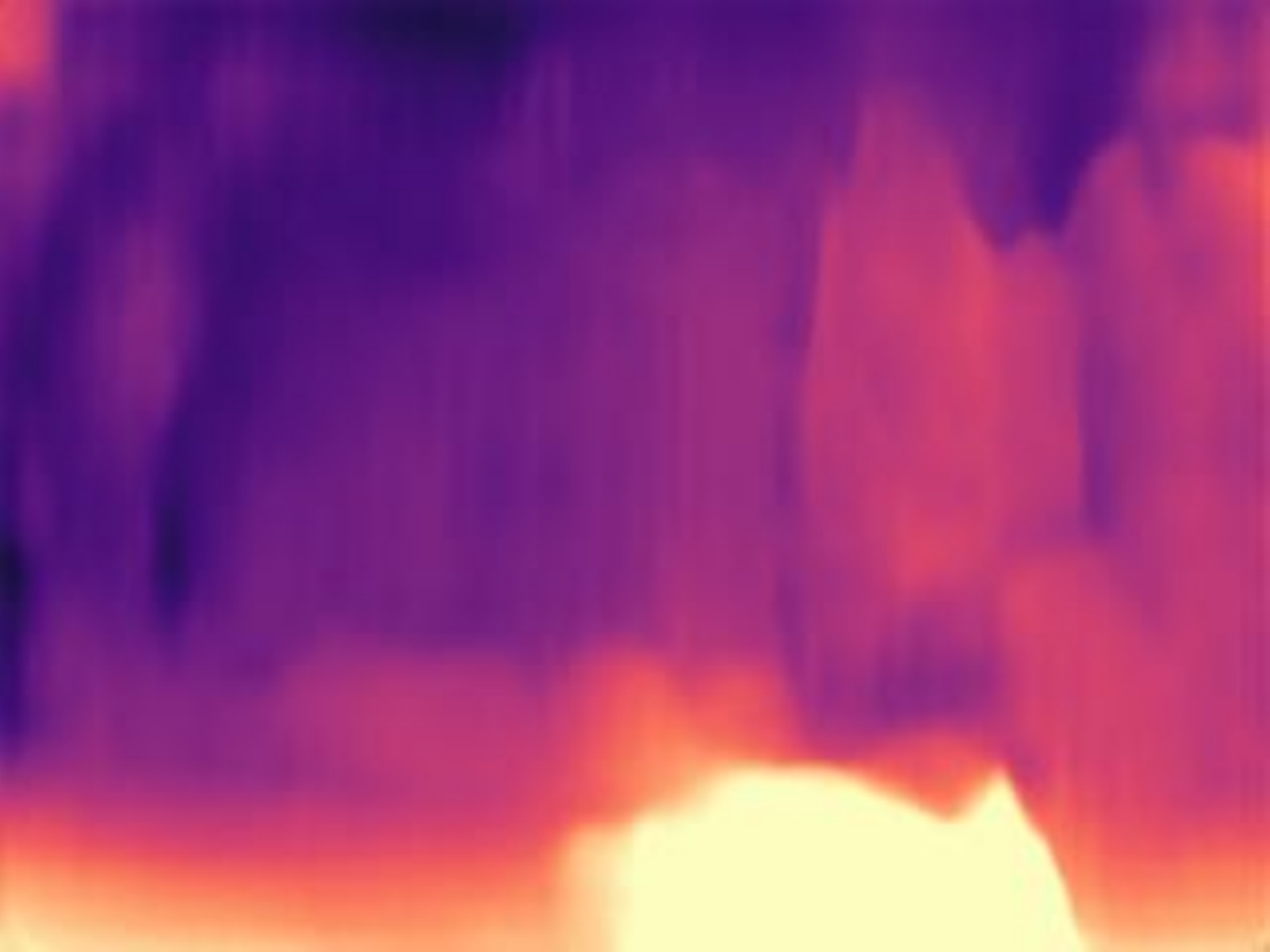} \qquad\qquad\quad & 
\includegraphics[width=\iw,height=\ih]{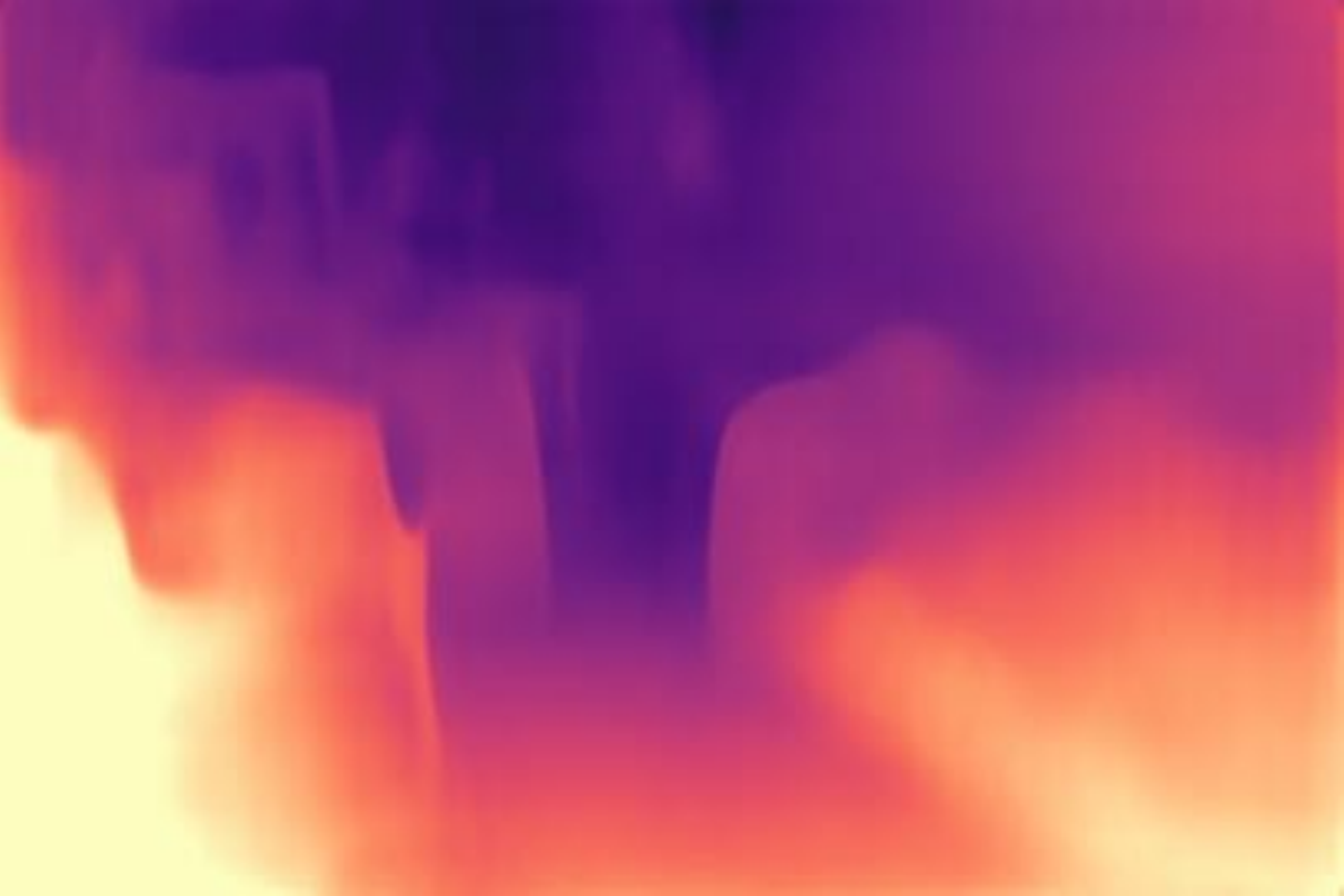} \qquad\qquad\quad & 
\includegraphics[width=\iw,height=\ih]{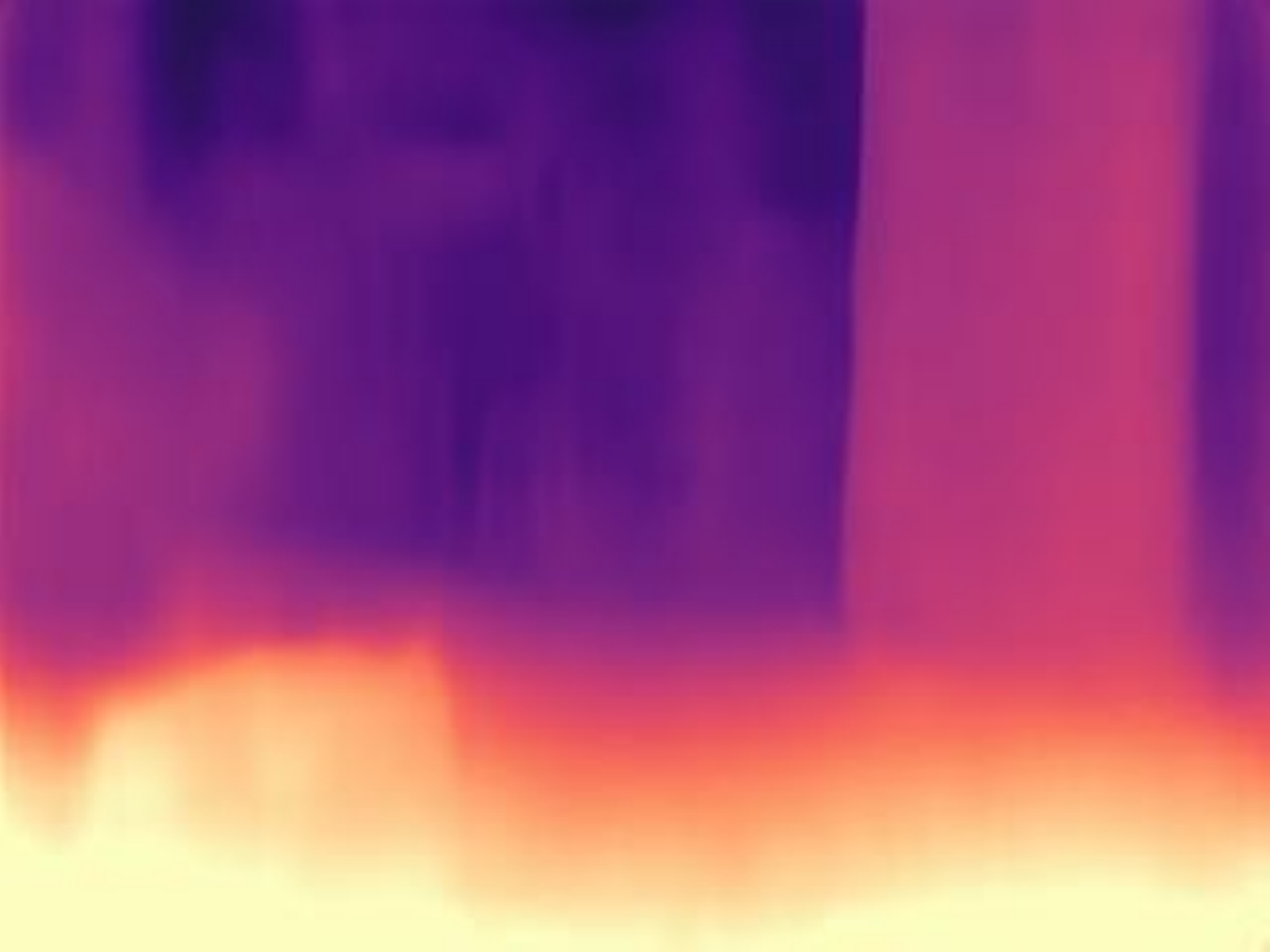}\\
\vspace{10mm}\\
\rotatebox[origin=c]{90}{\fontsize{\textw}{\texth} \selectfont MF-RegionViT\hspace{-270mm}}\hspace{24mm}
\includegraphics[width=\iw,height=\ih]{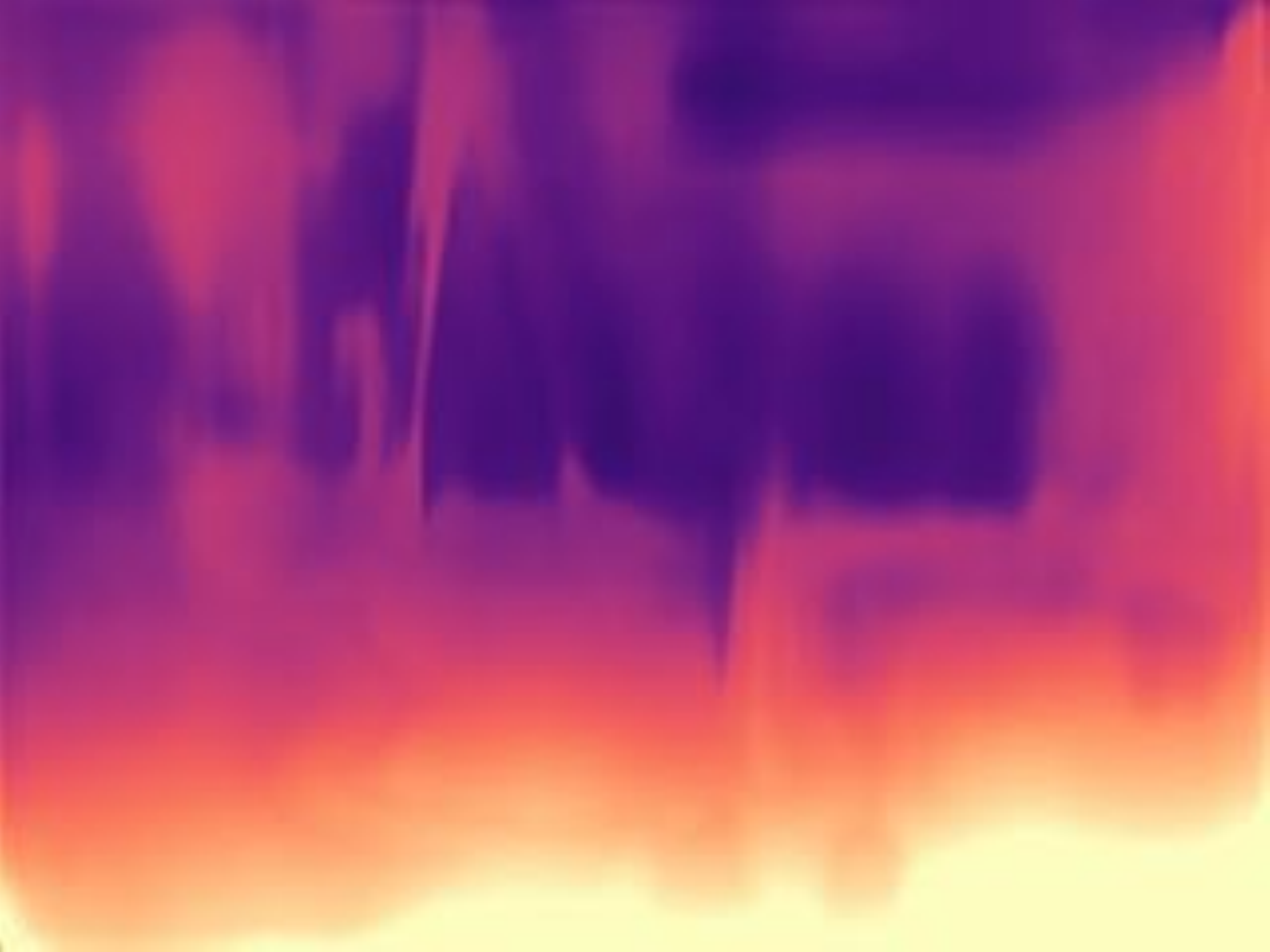} \qquad\qquad\quad & 
\includegraphics[width=\iw,height=\ih]{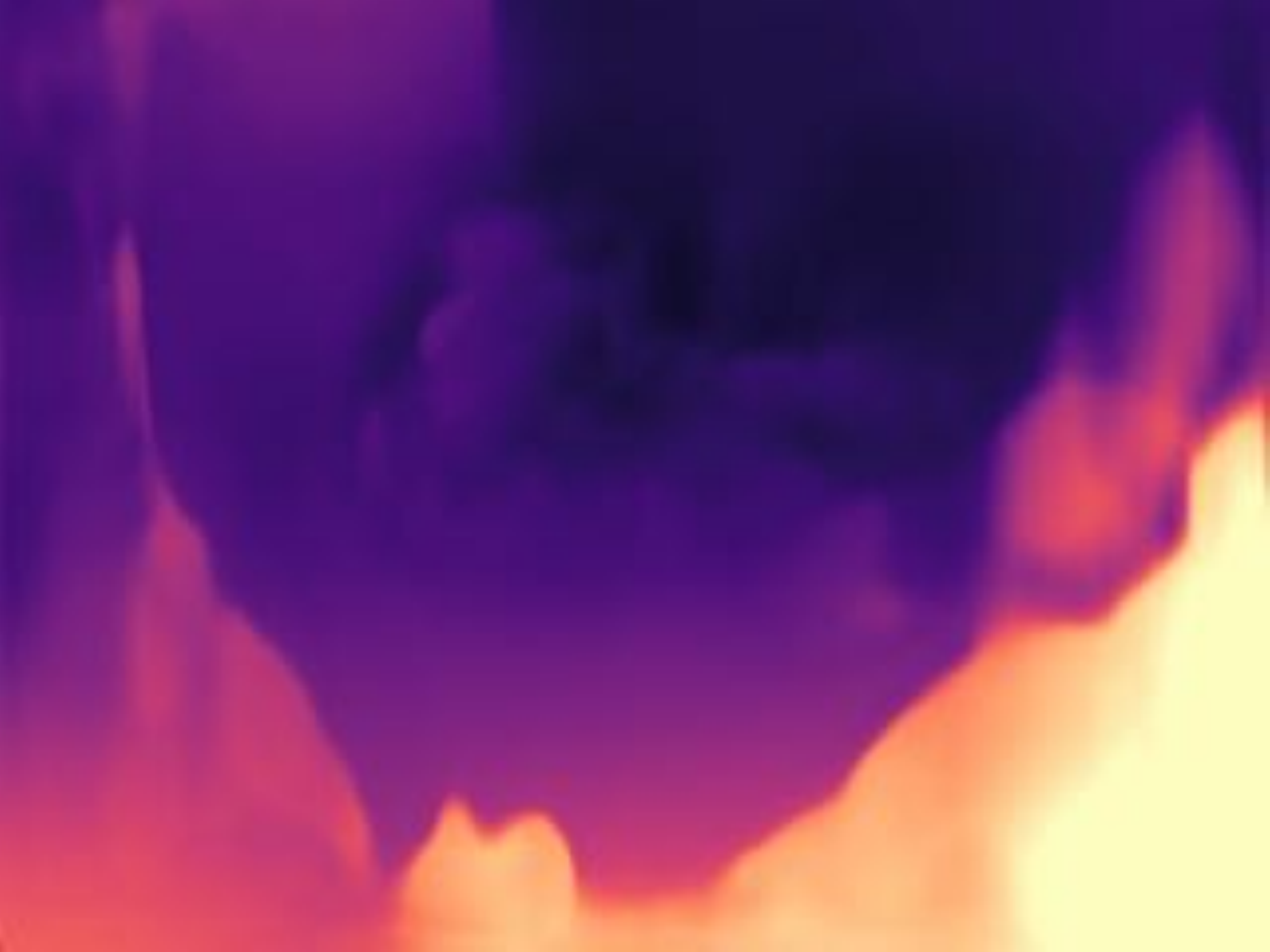} \qquad\qquad\quad & 
\includegraphics[width=\iw,height=\ih]{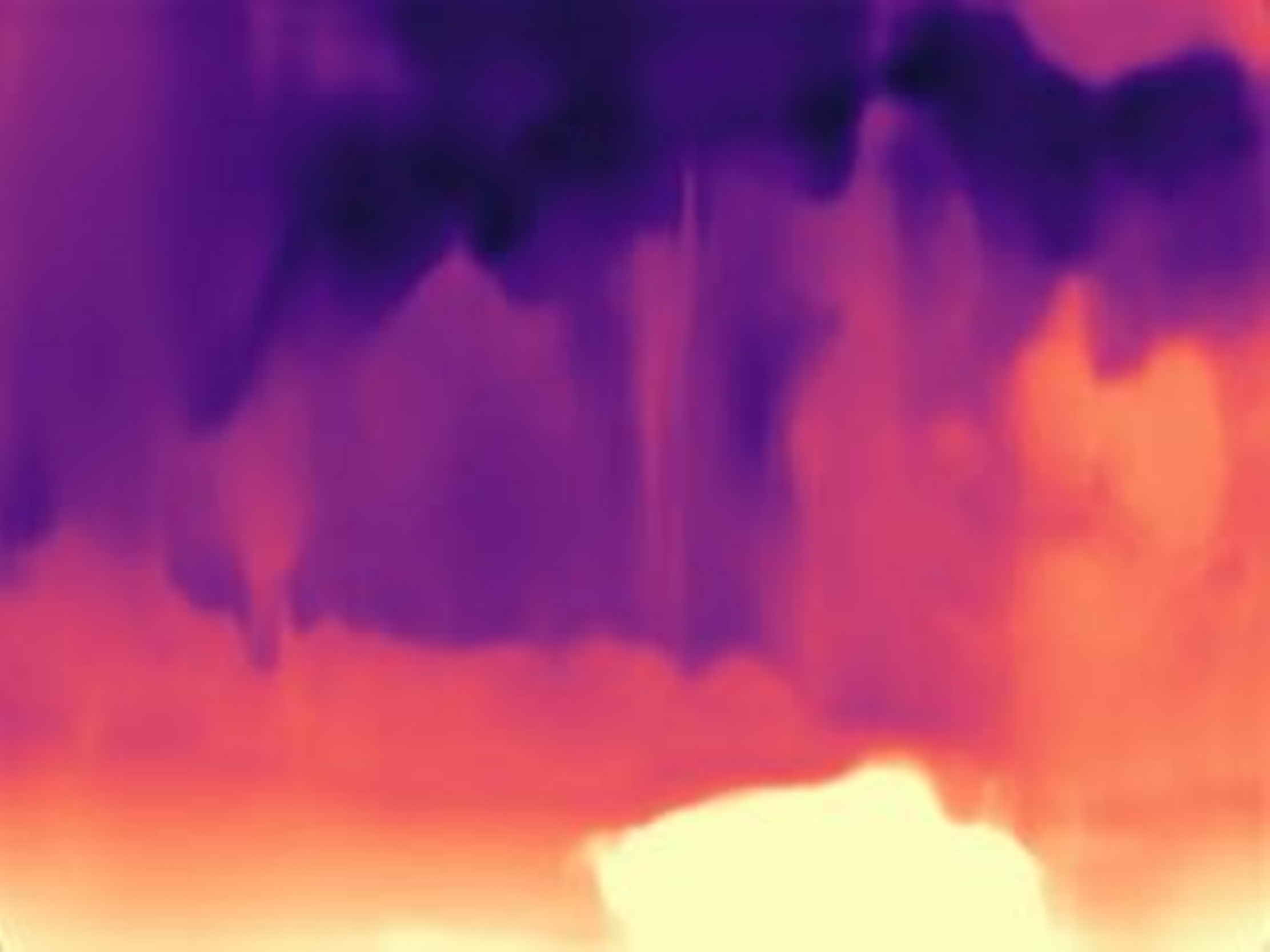} \qquad\qquad\quad & 
\includegraphics[width=\iw,height=\ih]{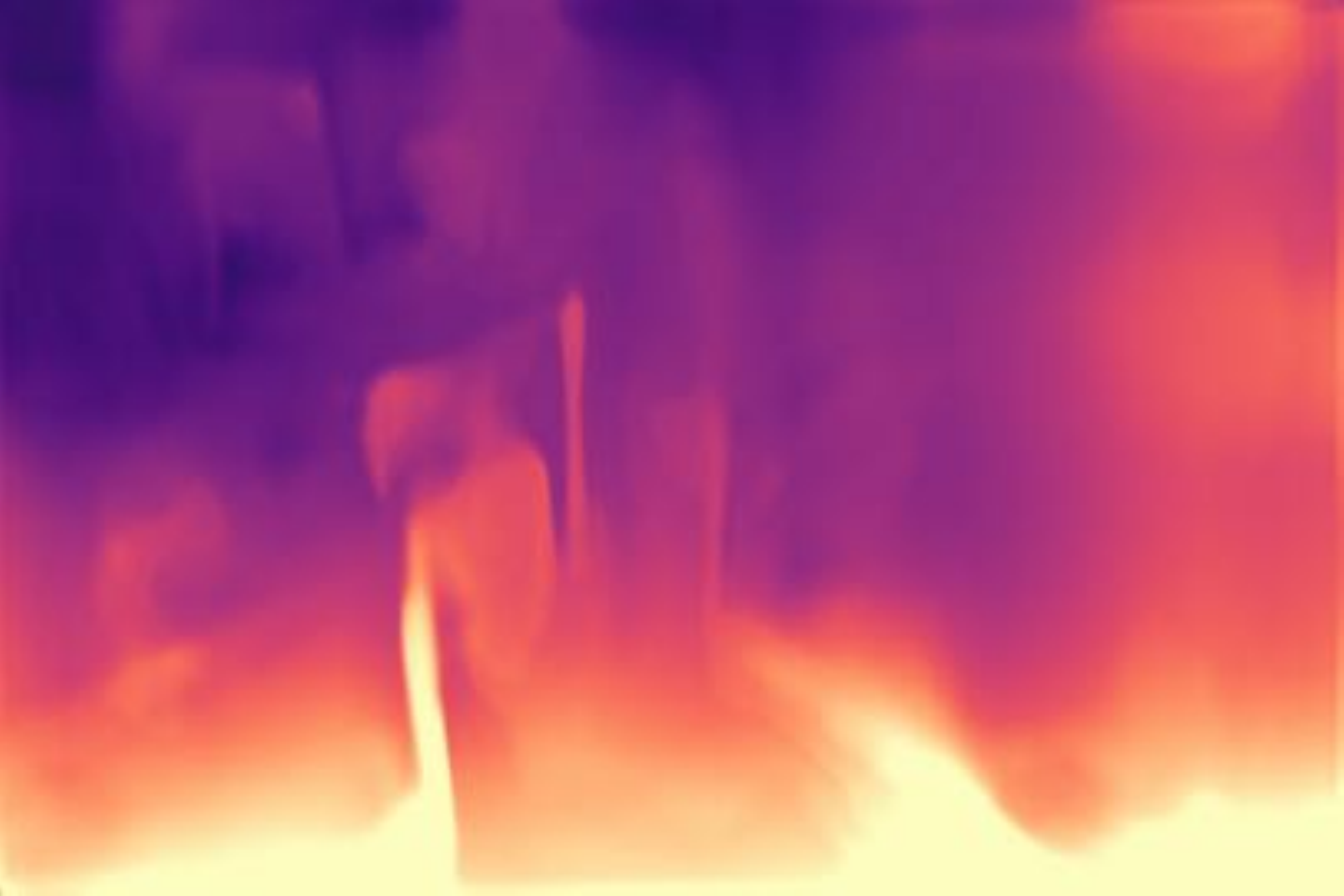} \qquad\qquad\quad & 
\includegraphics[width=\iw,height=\ih]{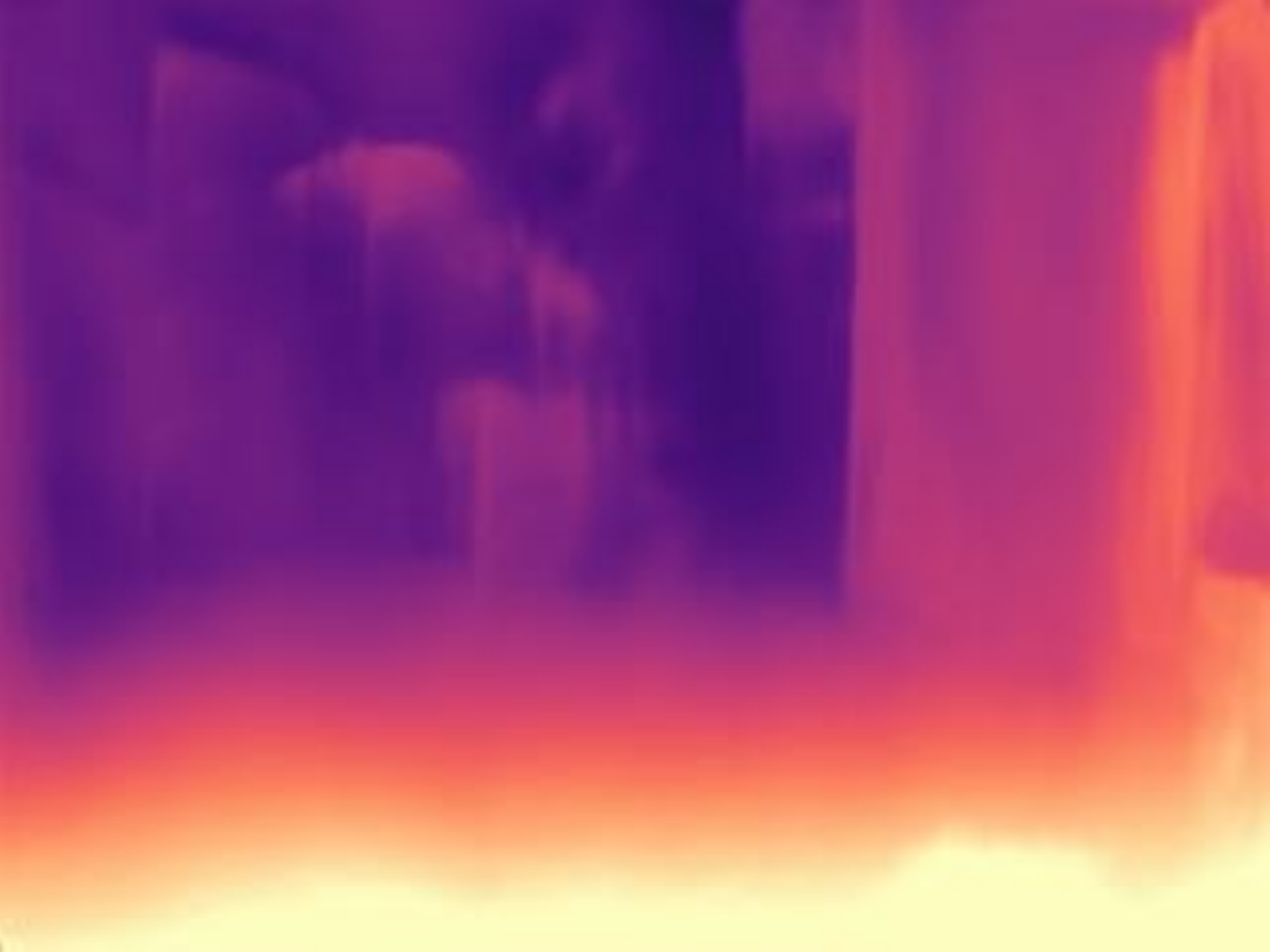}\\
\vspace{10mm}\\
\rotatebox[origin=c]{90}{\fontsize{\textw}{\texth} \selectfont MF-Twins\hspace{-270mm}}\hspace{24mm}
\includegraphics[width=\iw,height=\ih]{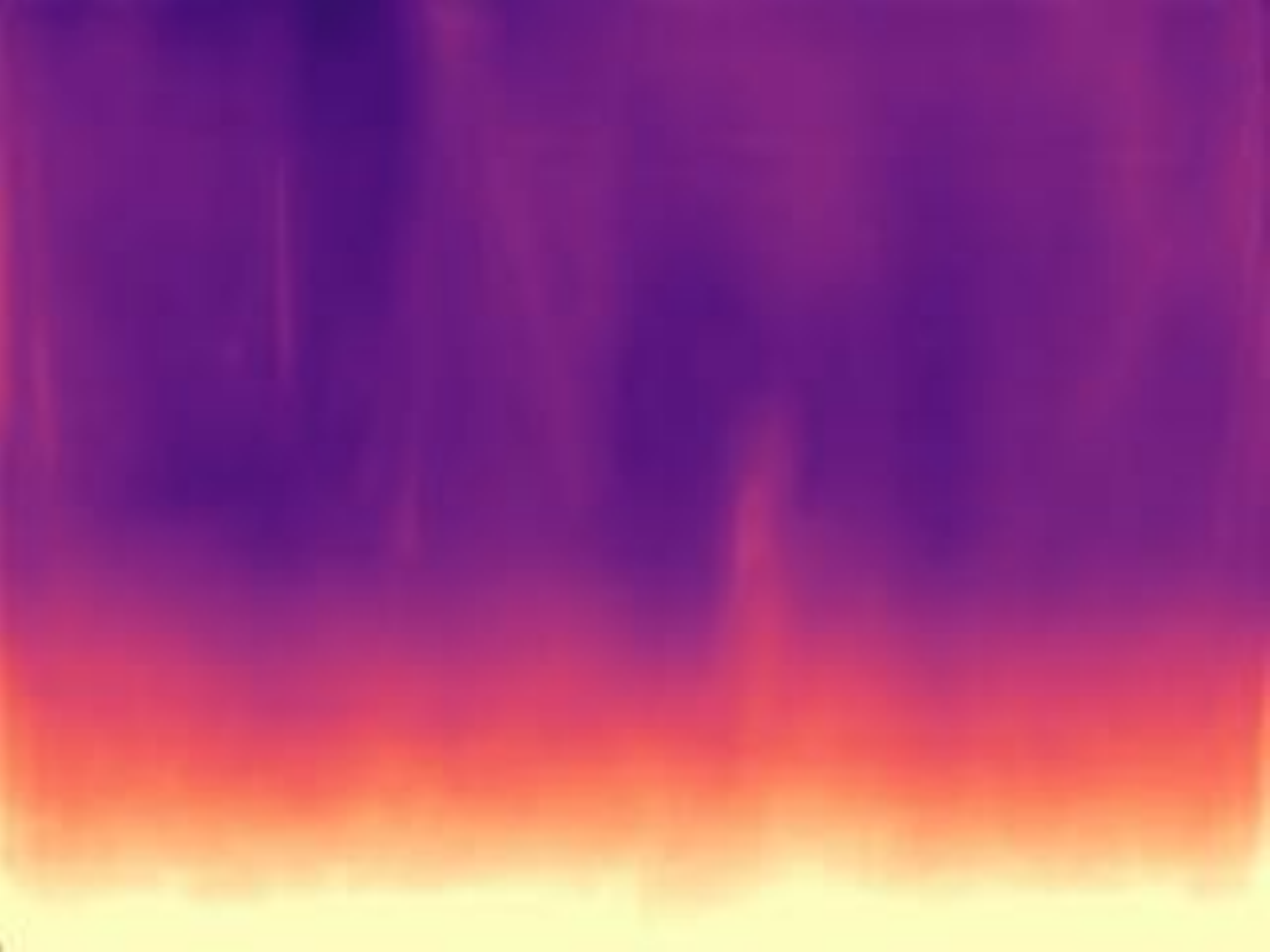} \qquad\qquad\quad & 
\includegraphics[width=\iw,height=\ih]{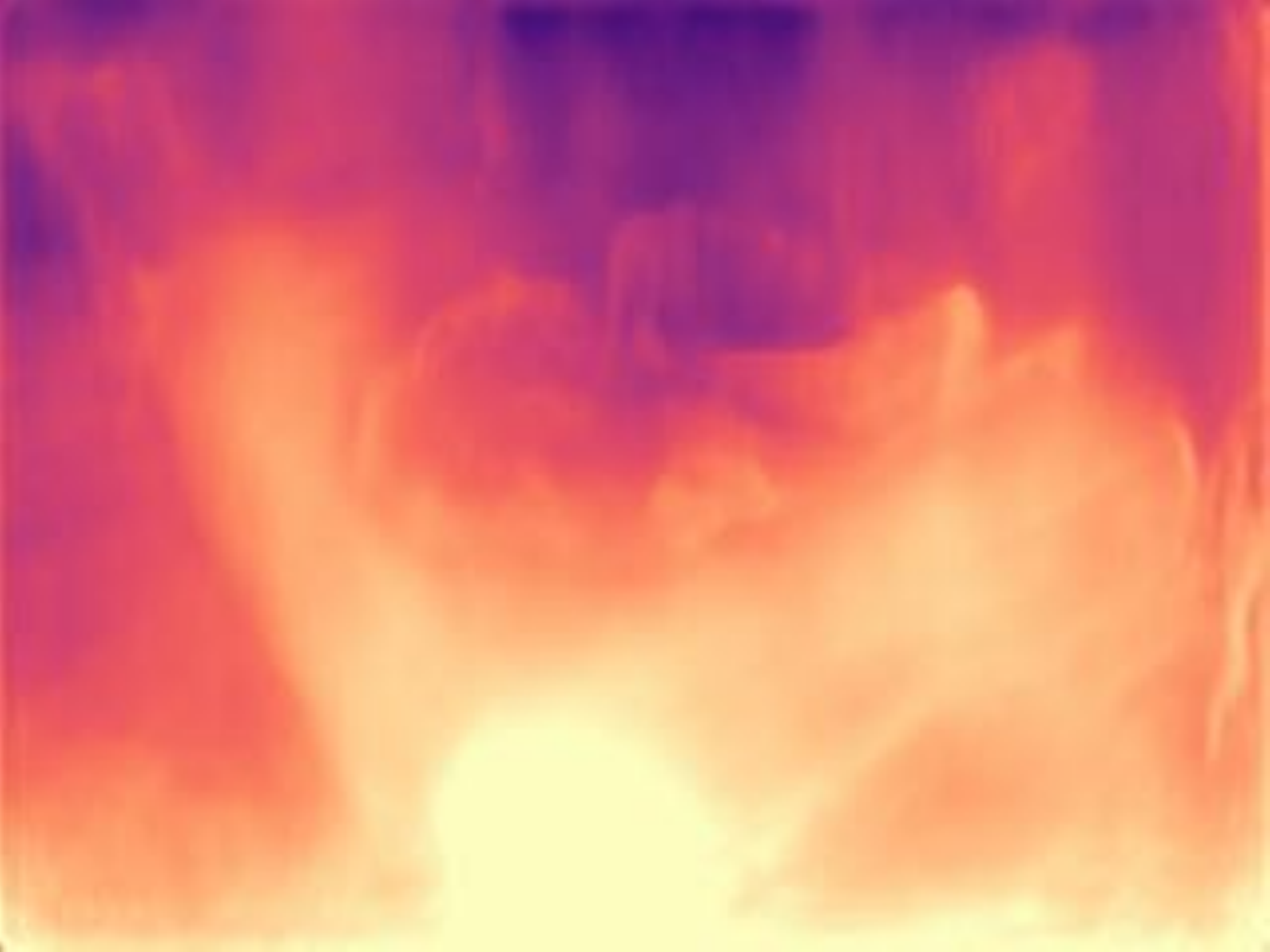} \qquad\qquad\quad & 
\includegraphics[width=\iw,height=\ih]{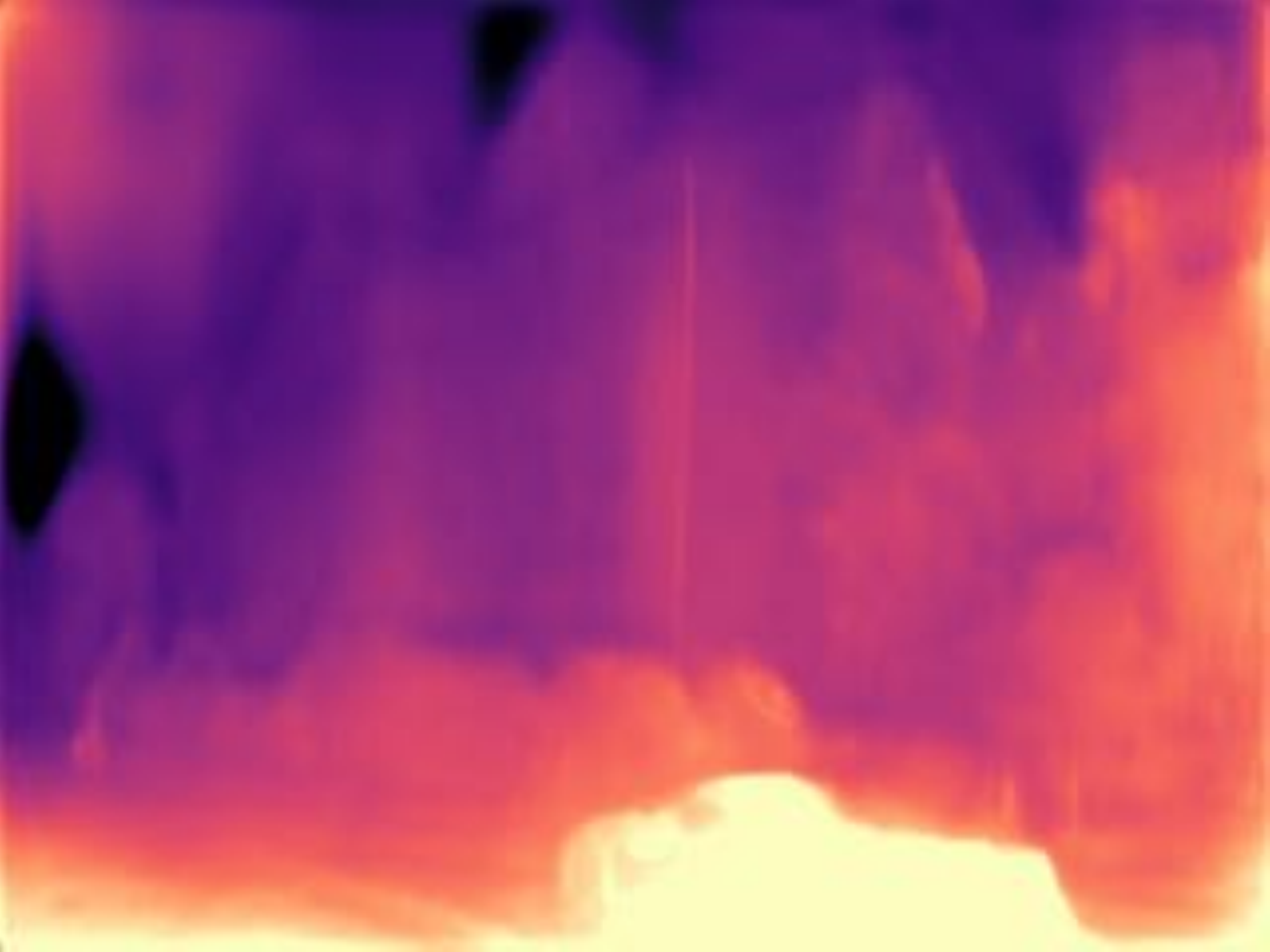} \qquad\qquad\quad & 
\includegraphics[width=\iw,height=\ih]{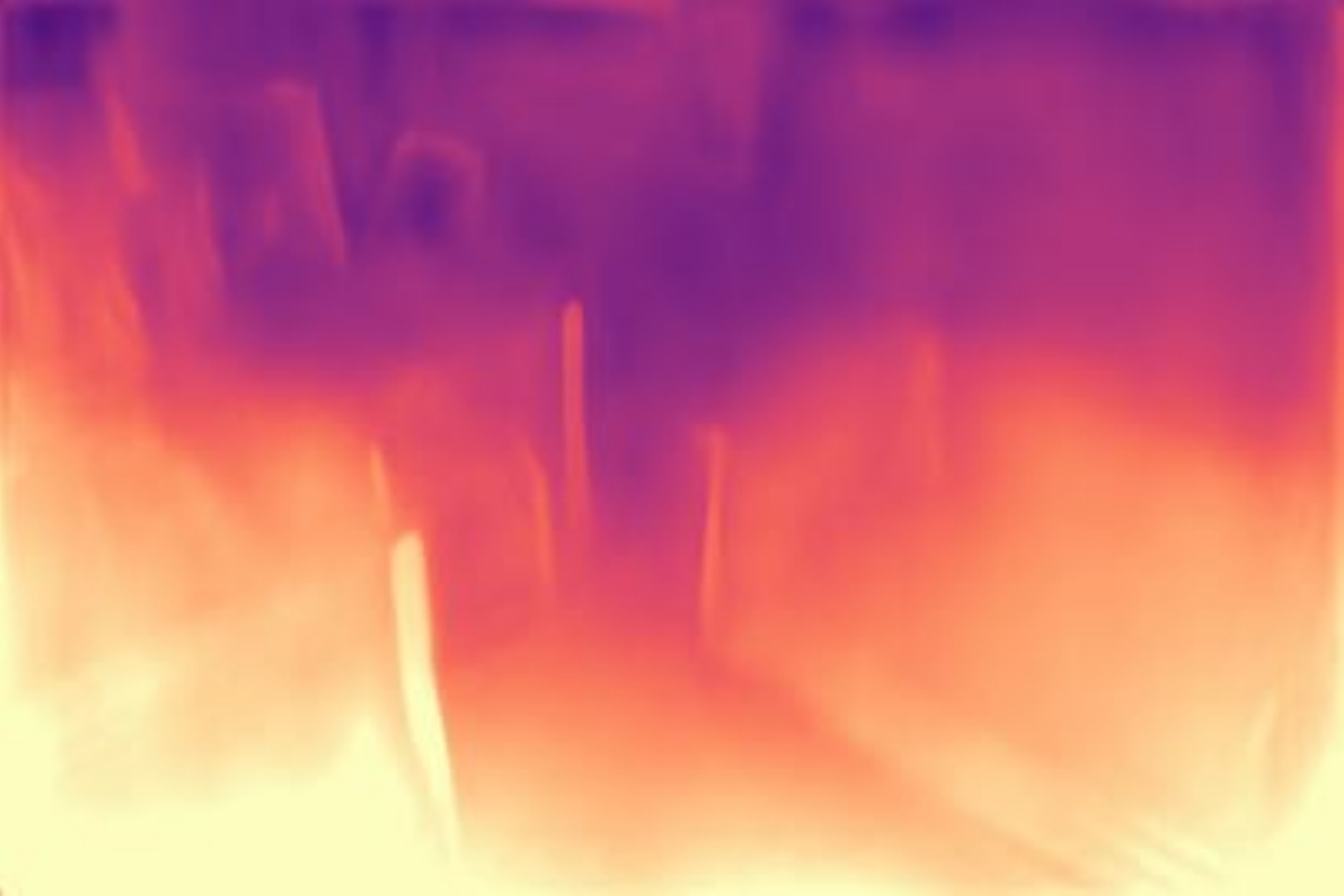} \qquad\qquad\quad & 
\includegraphics[width=\iw,height=\ih]{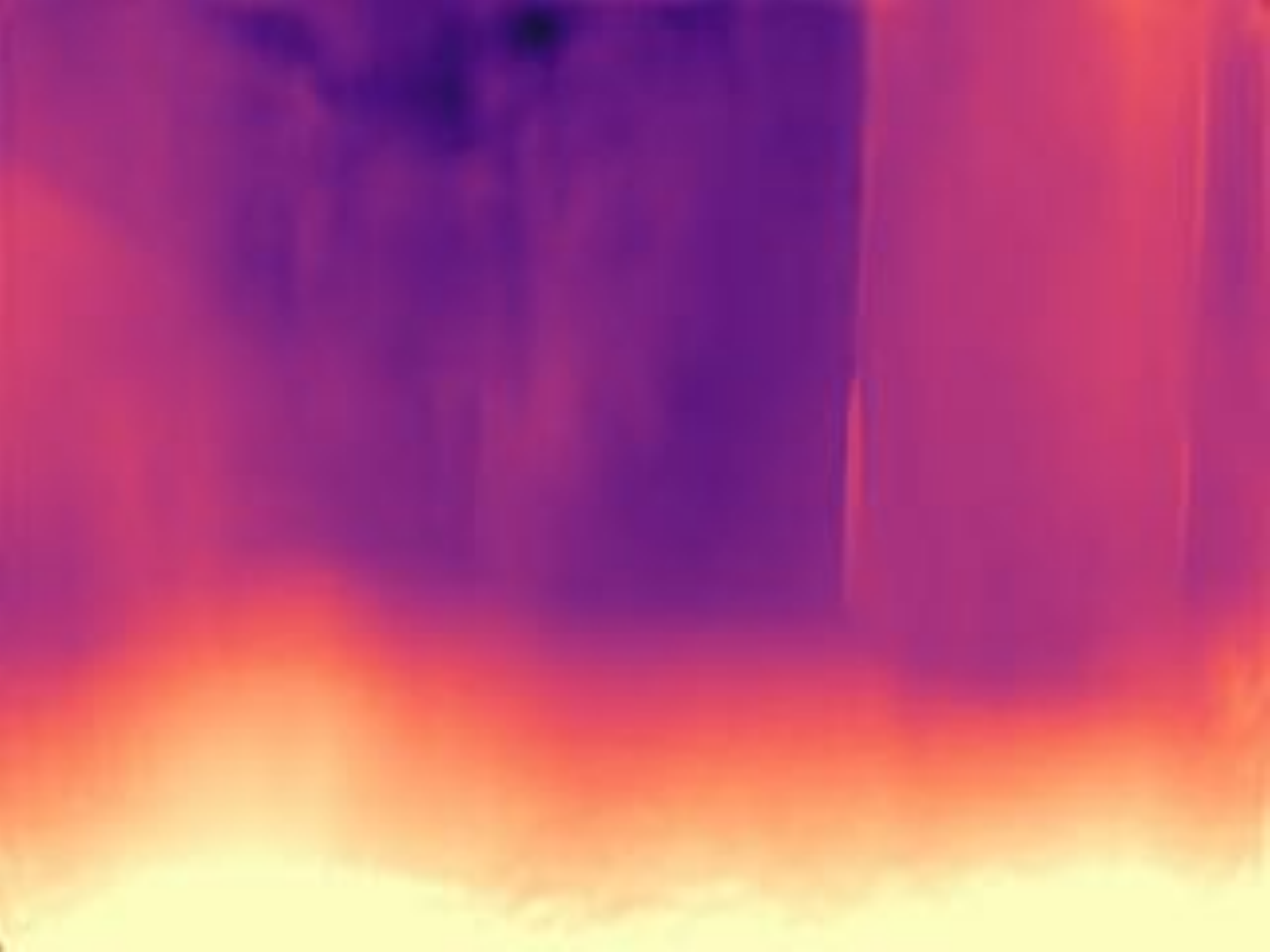}\\
\vspace{10mm}\\
\rotatebox[origin=c]{90}{\fontsize{\textw}{\texth} \selectfont MF-Ours\hspace{-270mm}}\hspace{24mm}
\includegraphics[width=\iw,height=\ih]{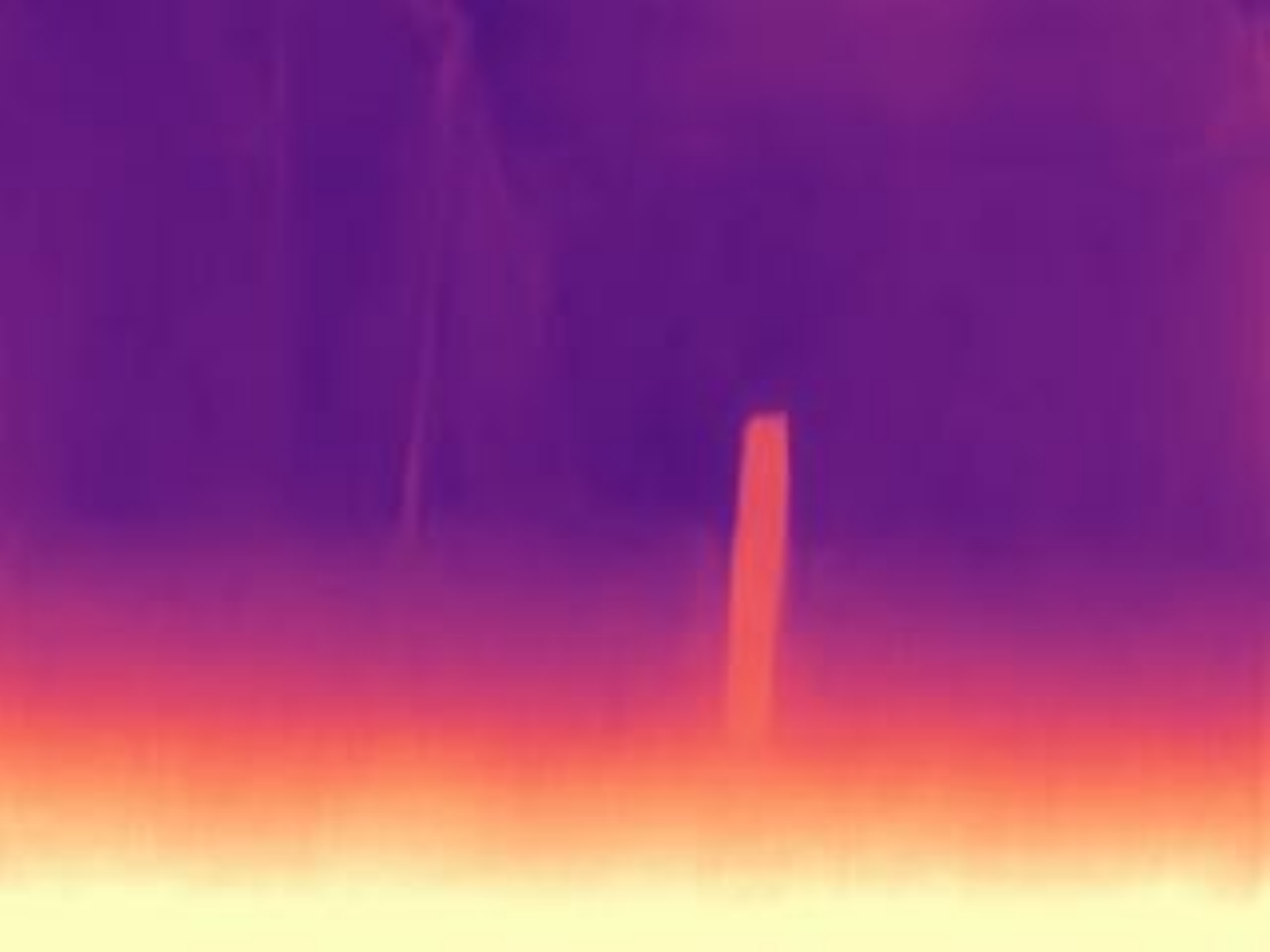} \qquad\qquad\quad & 
\includegraphics[width=\iw,height=\ih]{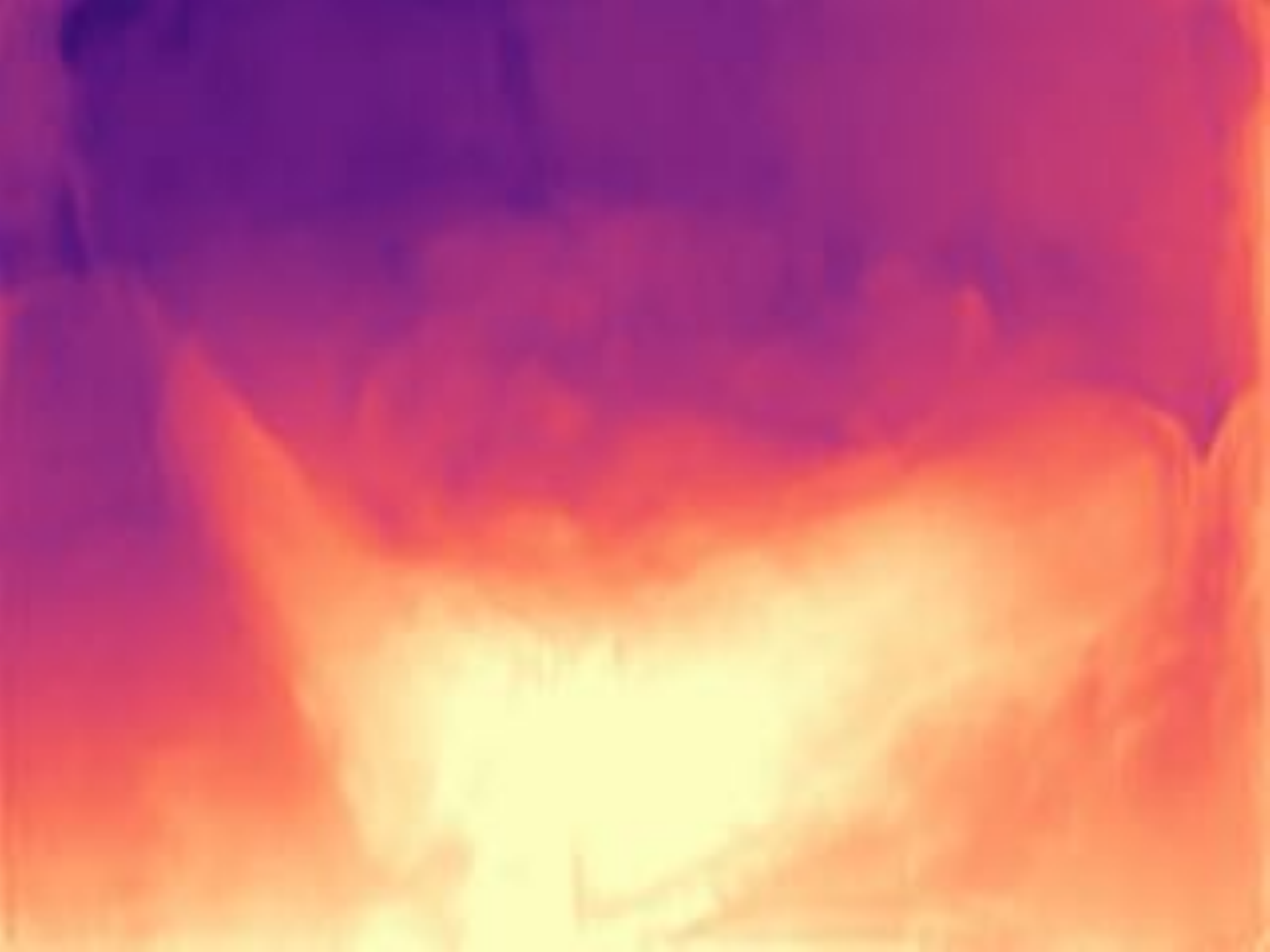} \qquad\qquad\quad & 
\includegraphics[width=\iw,height=\ih]{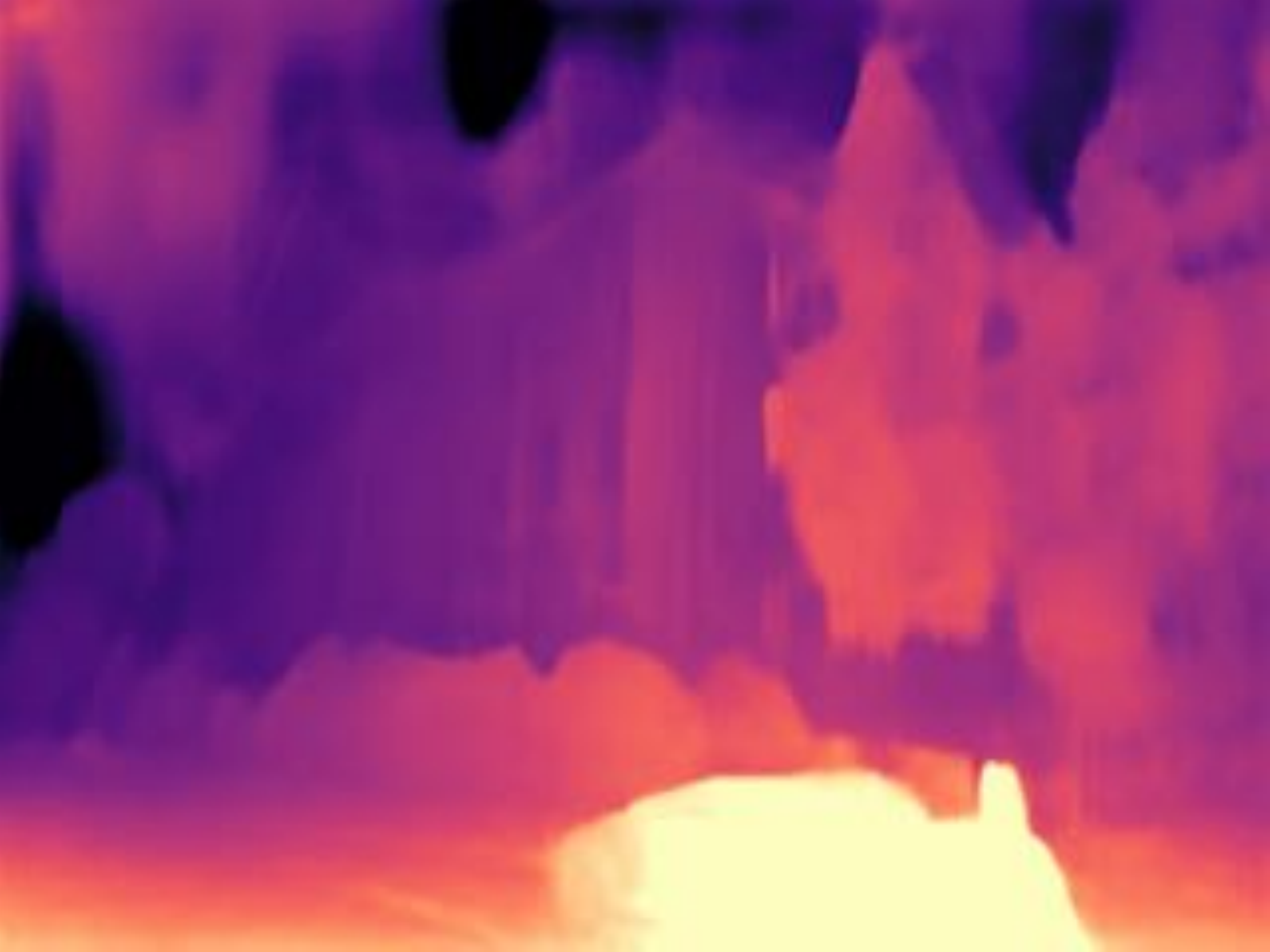} \qquad\qquad\quad & 
\includegraphics[width=\iw,height=\ih]{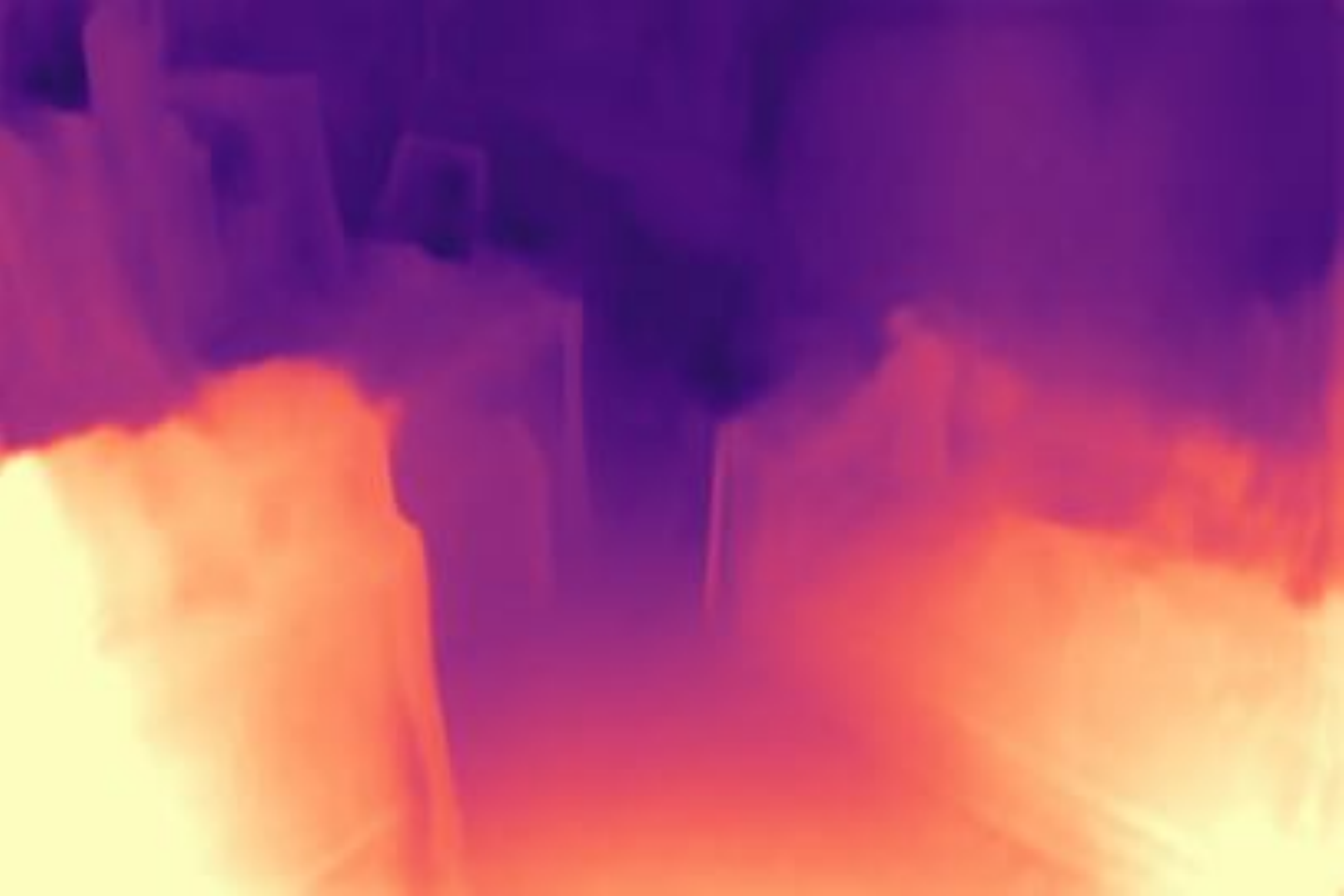} \qquad\qquad\quad & 
\includegraphics[width=\iw,height=\ih]{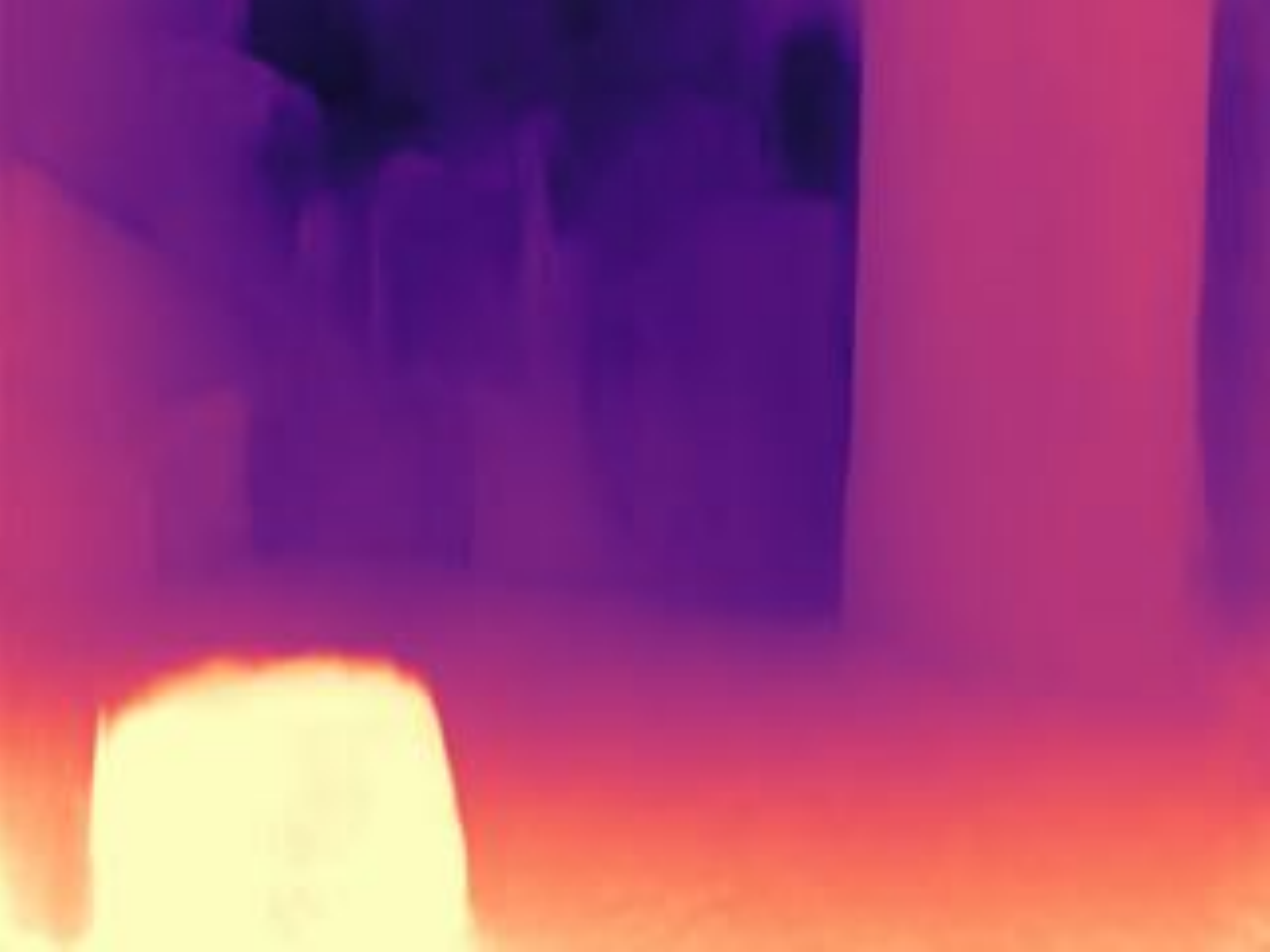}\\
\vspace{30mm}\\
\multicolumn{5}{c}{\fontsize{\w}{\h} \selectfont (b) Self-supervised Transformer-based methods } & 
\vspace{30mm}\\
\rotatebox[origin=c]{90}{\fontsize{\textw}{\texth} \selectfont BTS\hspace{-270mm}}\hspace{24mm}
\includegraphics[width=\iw,height=\ih]{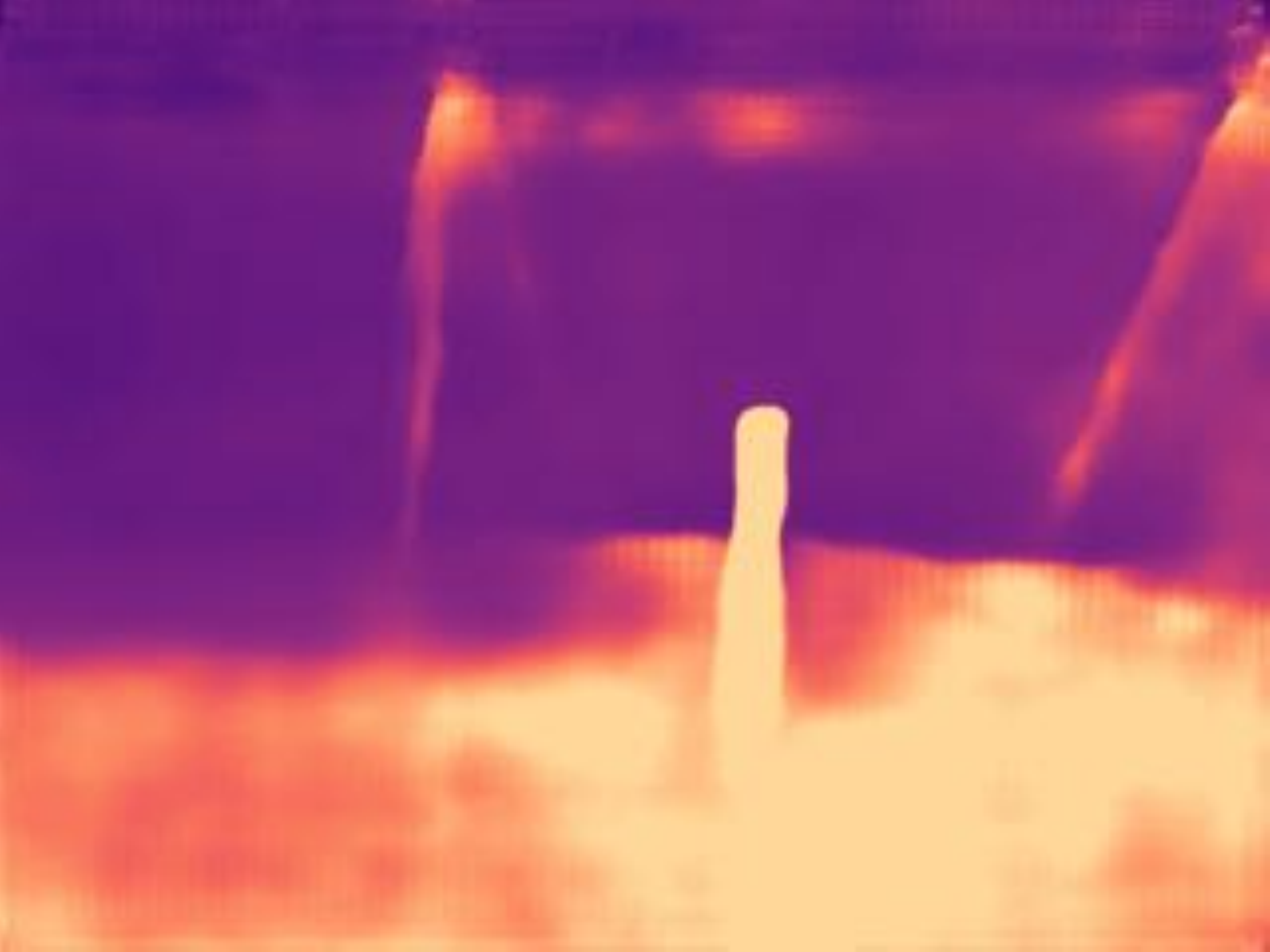} \qquad\qquad\quad & 
\includegraphics[width=\iw,height=\ih]{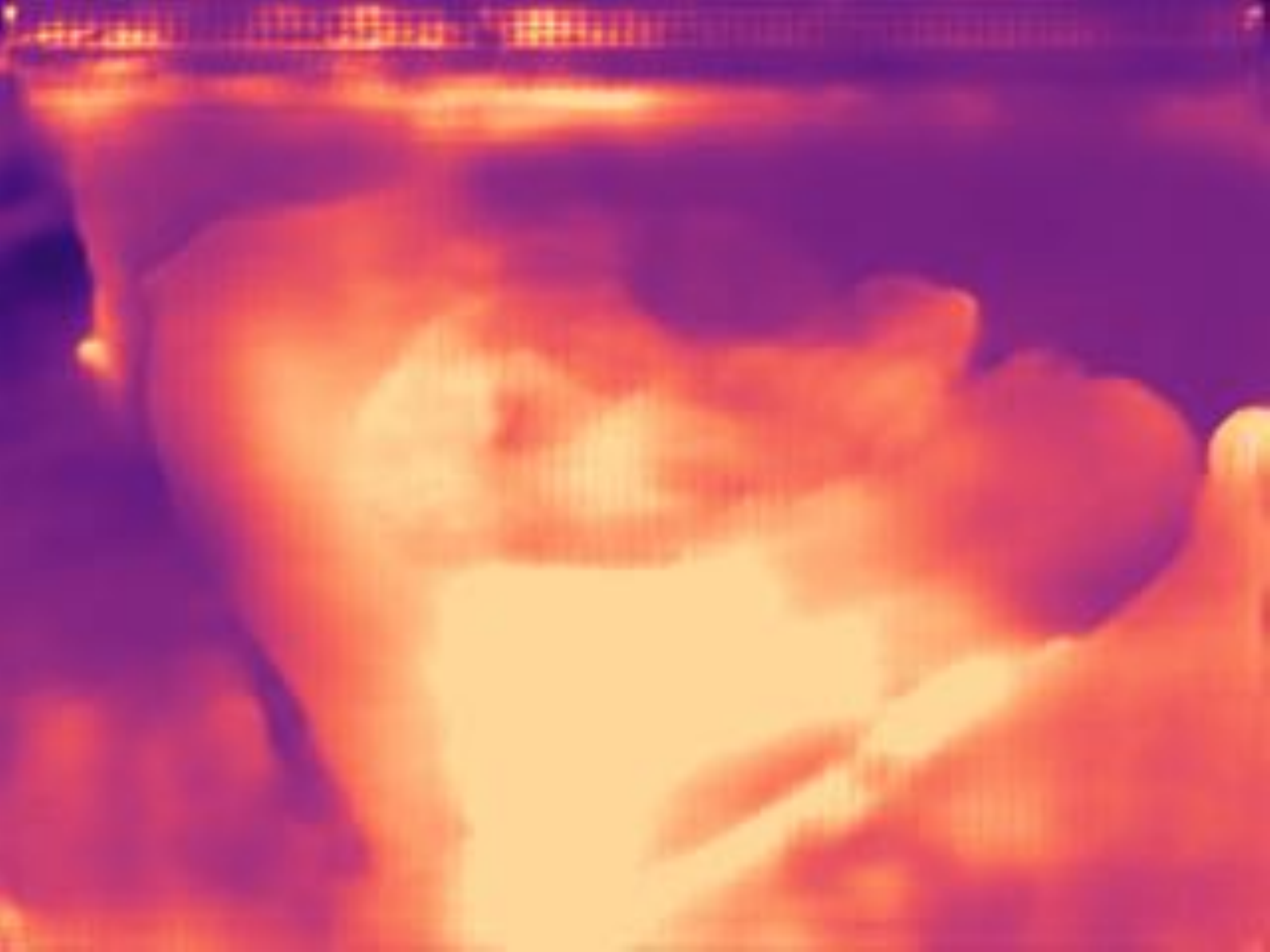} \qquad\qquad\quad & 
\includegraphics[width=\iw,height=\ih]{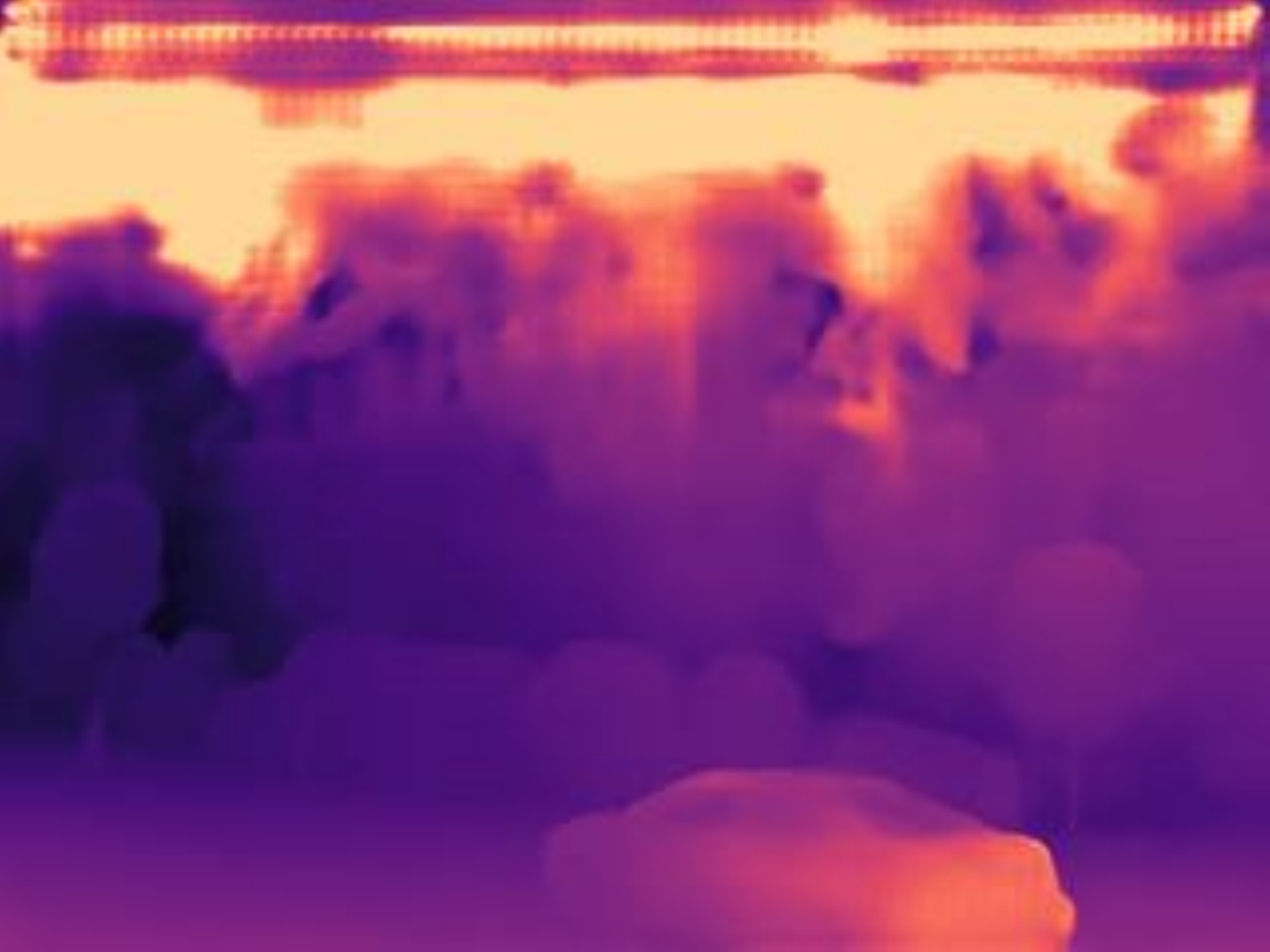} \qquad\qquad\quad & 
\includegraphics[width=\iw,height=\ih]{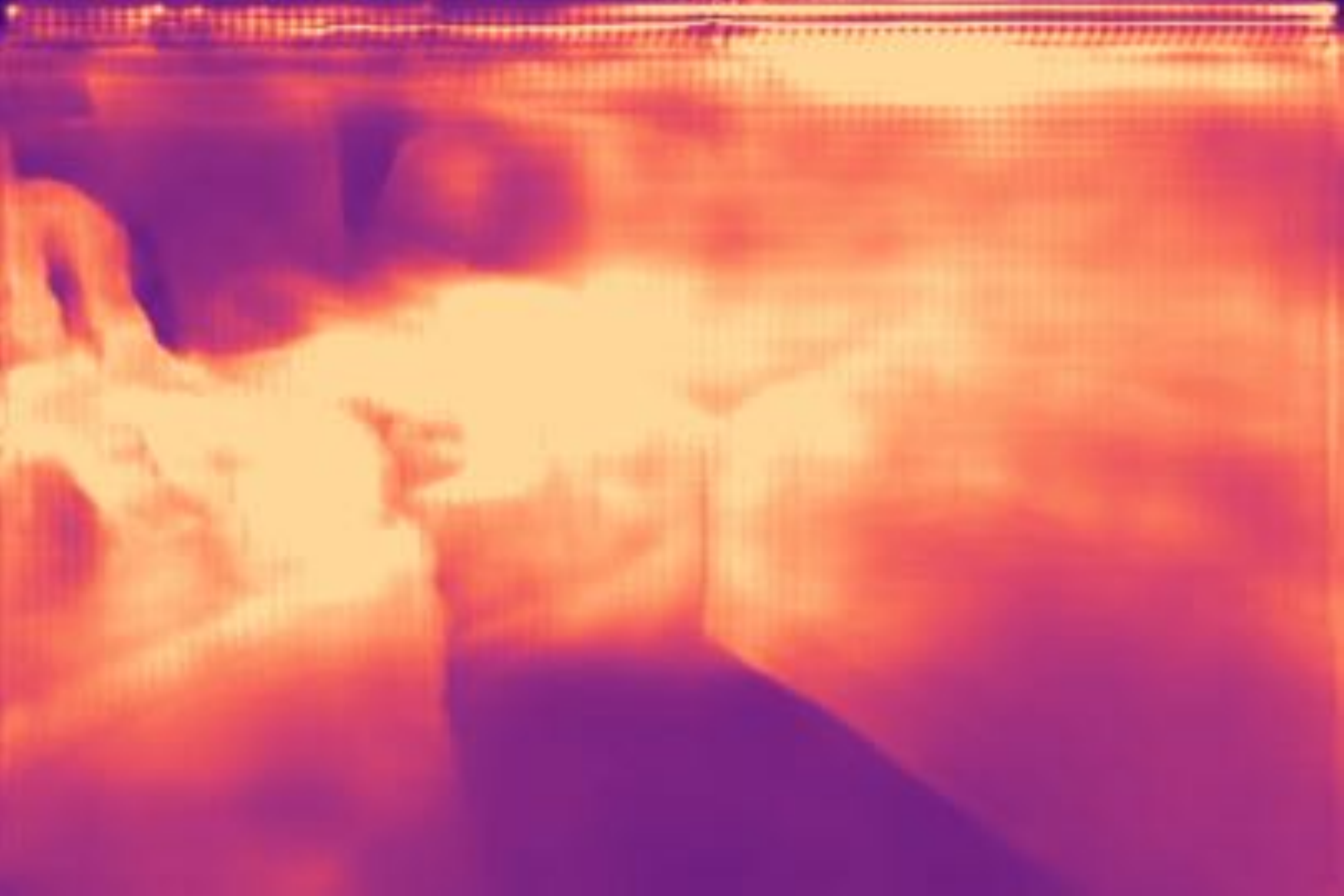} \qquad\qquad\quad & 
\includegraphics[width=\iw,height=\ih]{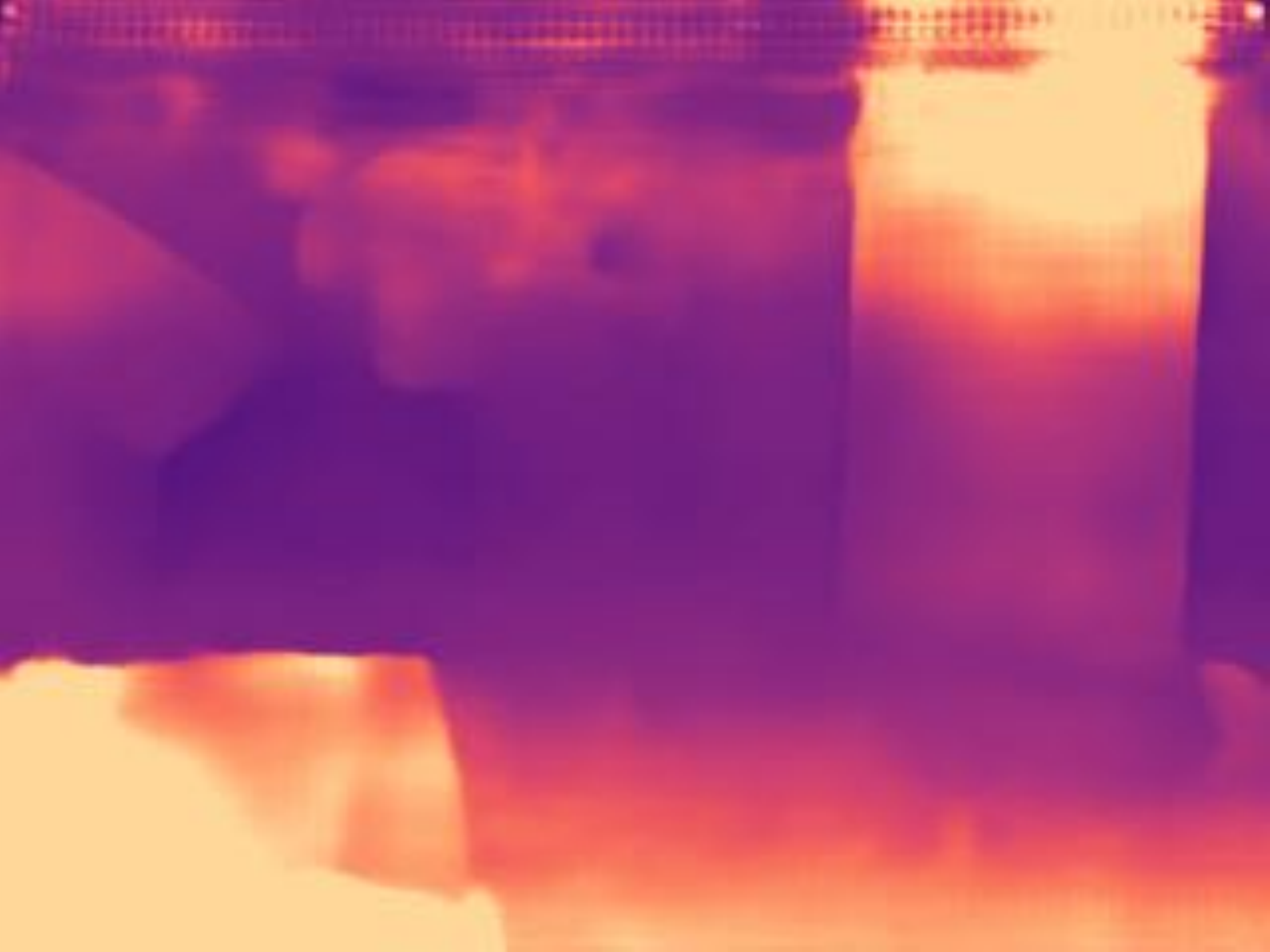}\\
\vspace{10mm}\\
\rotatebox[origin=c]{90}{\fontsize{\textw}{\texth} \selectfont Adabins\hspace{-270mm}}\hspace{24mm}
\includegraphics[width=\iw,height=\ih]{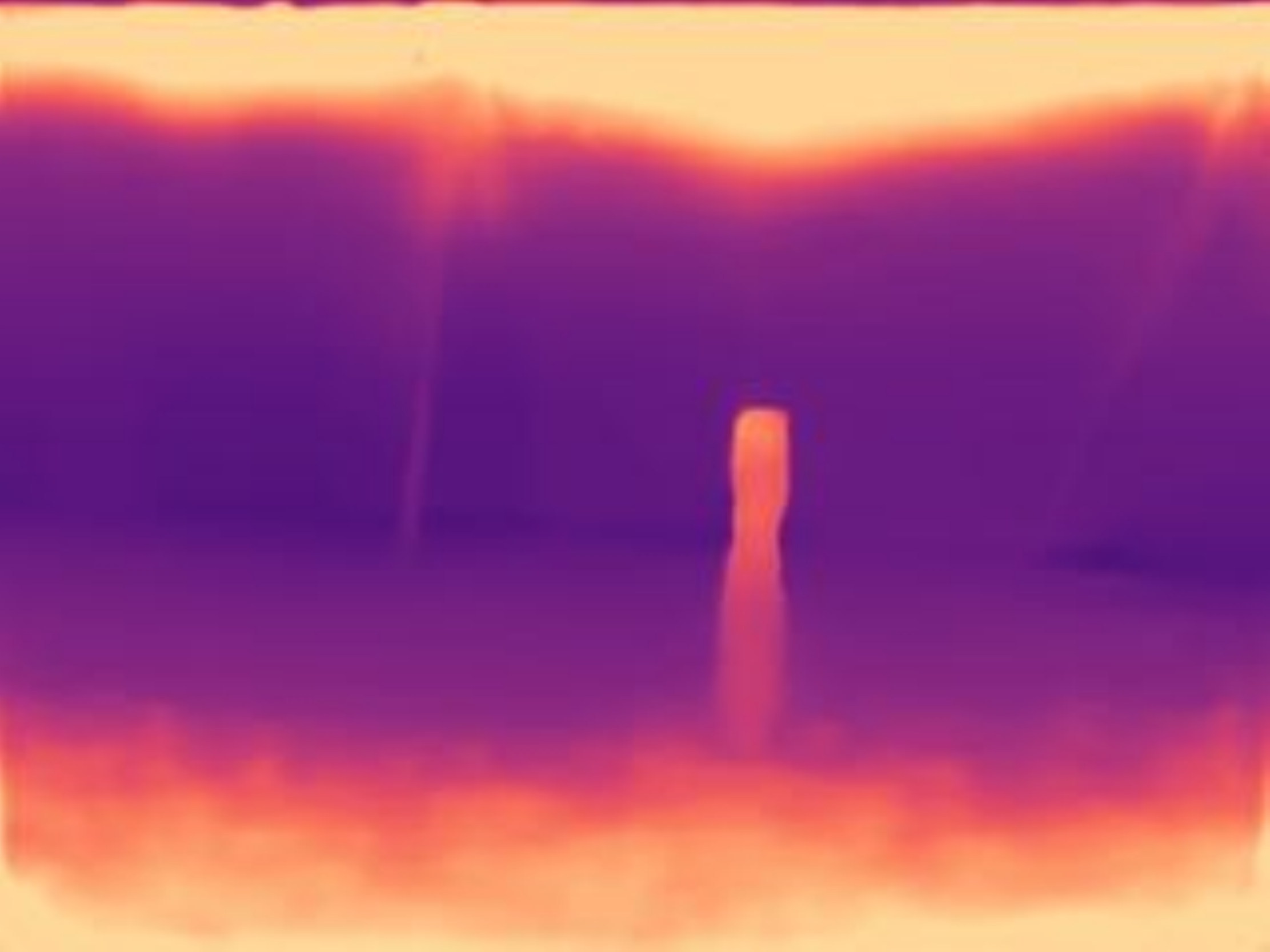} \qquad\qquad\quad & 
\includegraphics[width=\iw,height=\ih]{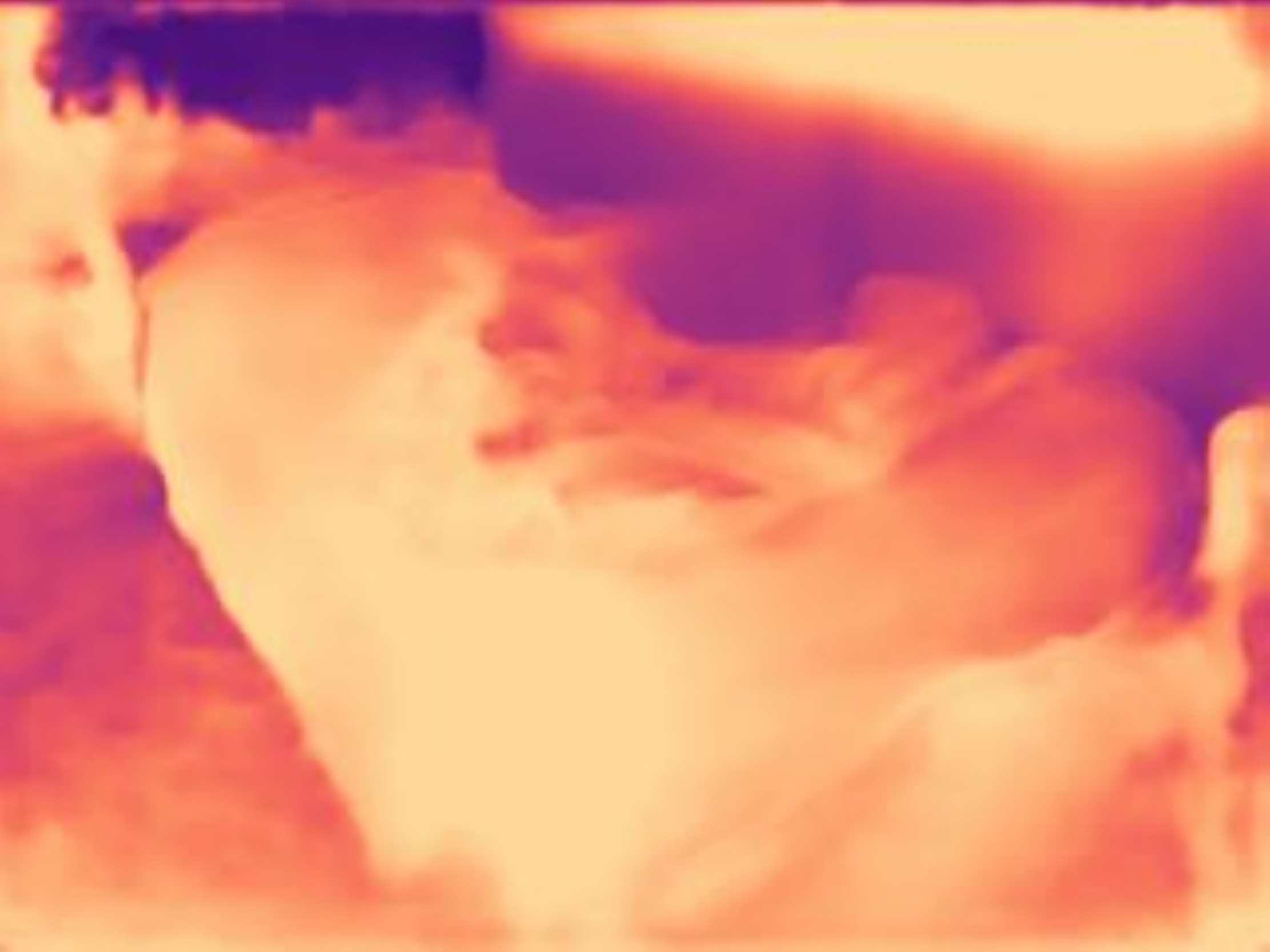} \qquad\qquad\quad & 
\includegraphics[width=\iw,height=\ih]{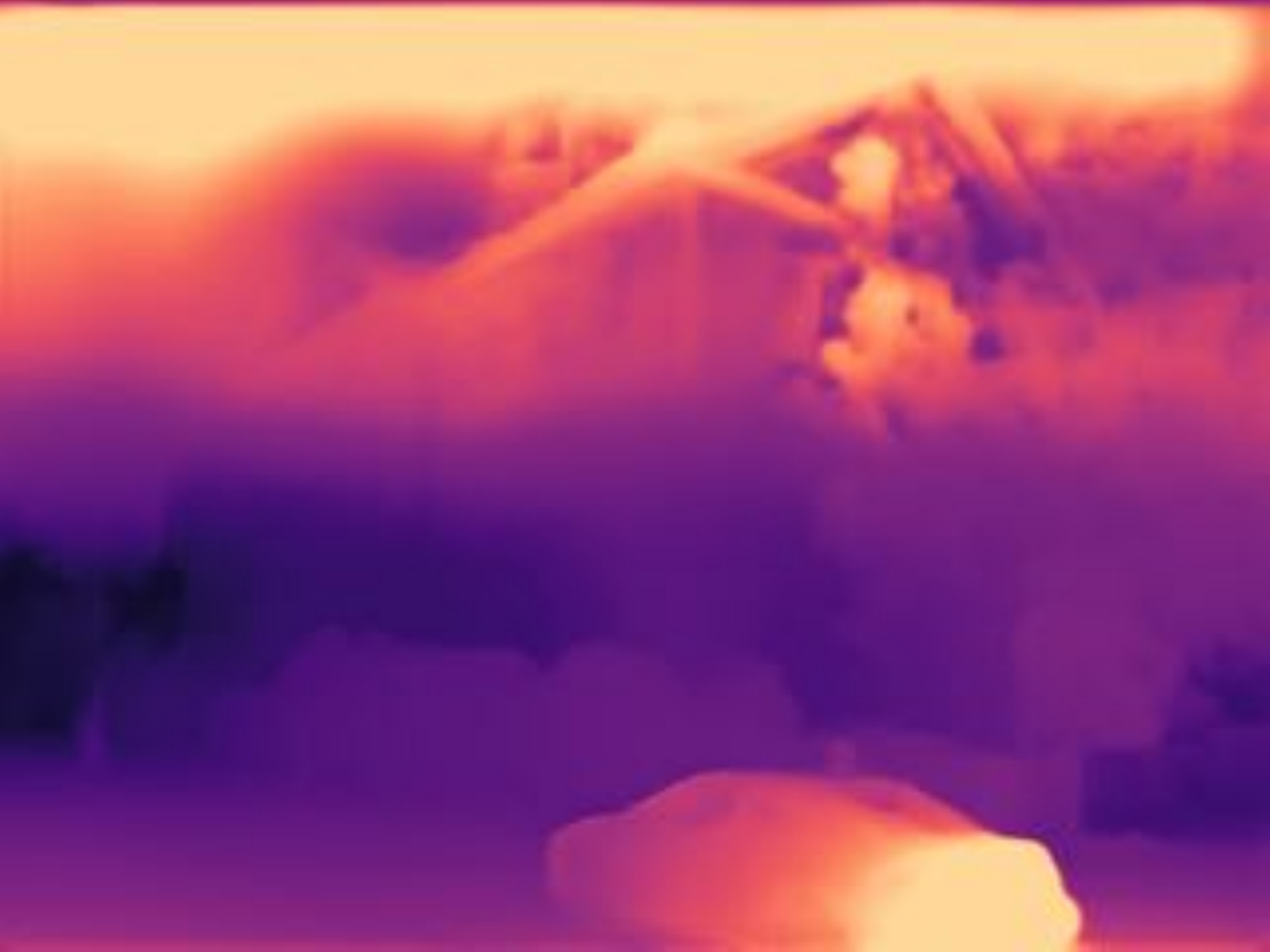} \qquad\qquad\quad & 
\includegraphics[width=\iw,height=\ih]{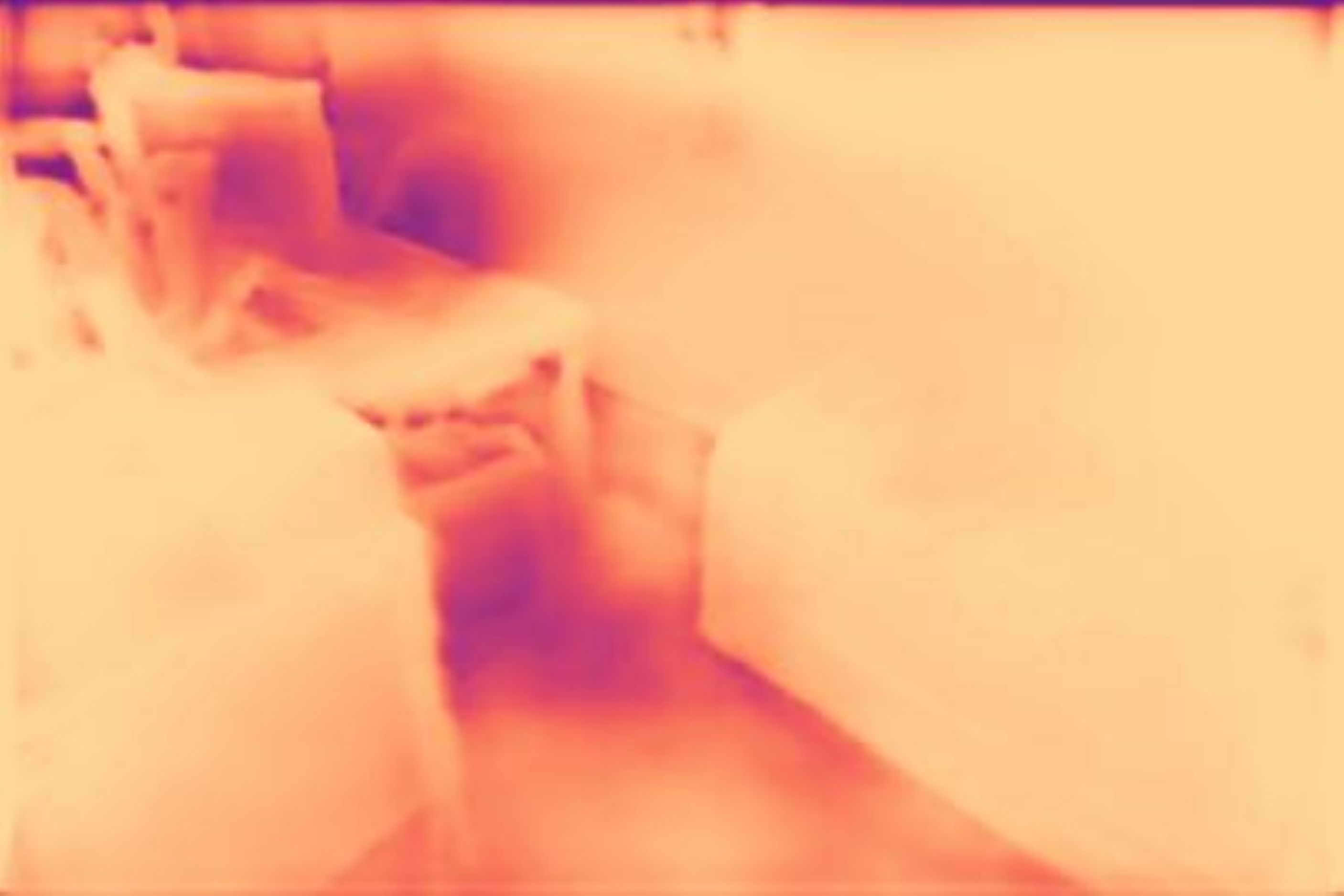} \qquad\qquad\quad & 
\includegraphics[width=\iw,height=\ih]{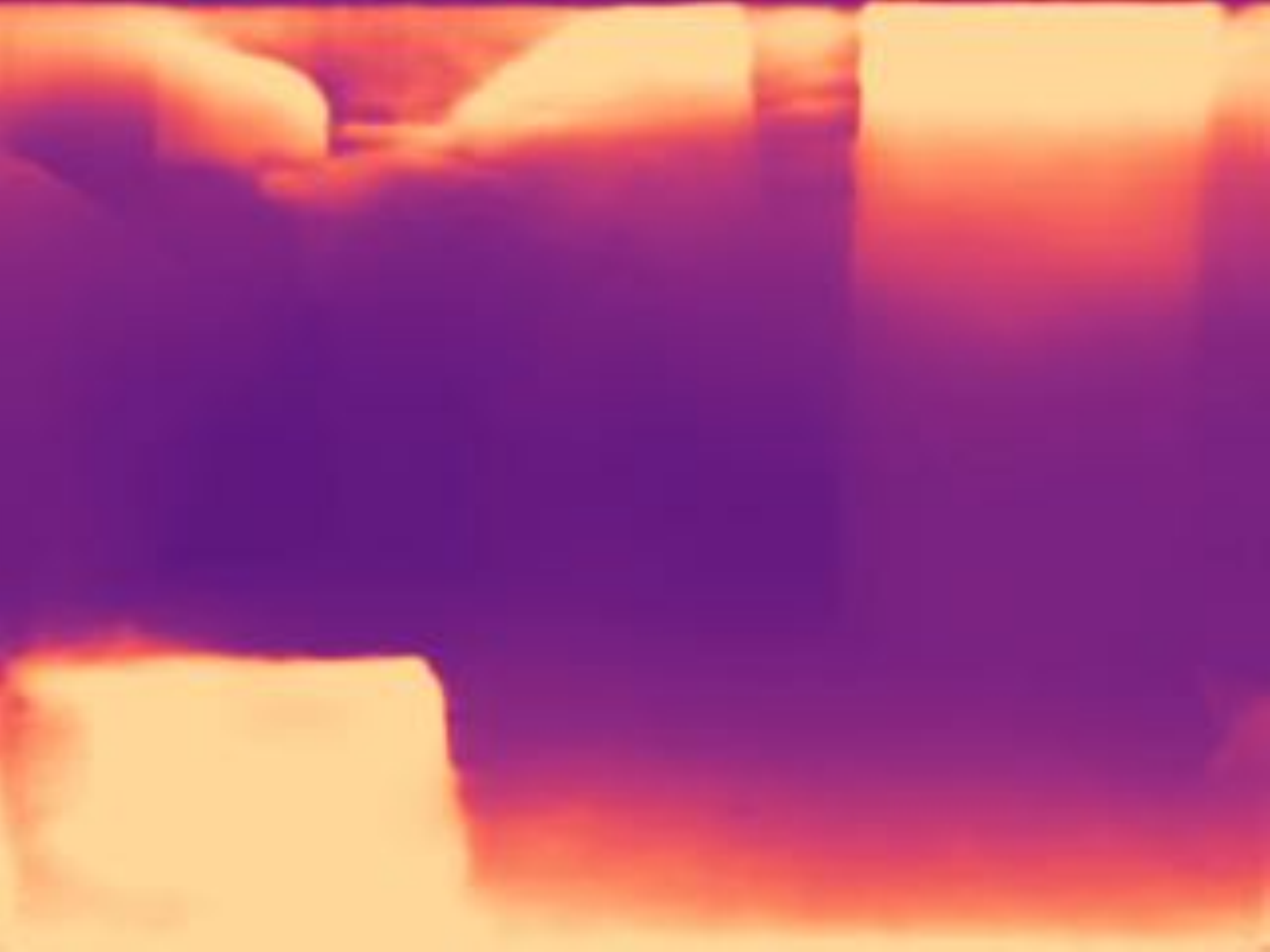}\\
\vspace{30mm}\\
\multicolumn{5}{c}{\fontsize{\w}{\h} \selectfont (c) Supervised CNN-based methods } & 
\vspace{30mm}\\
\rotatebox[origin=c]{90}{\fontsize{\textw}{\texth} \selectfont TransDepth\hspace{-270mm}}\hspace{24mm}
\includegraphics[width=\iw,height=\ih]{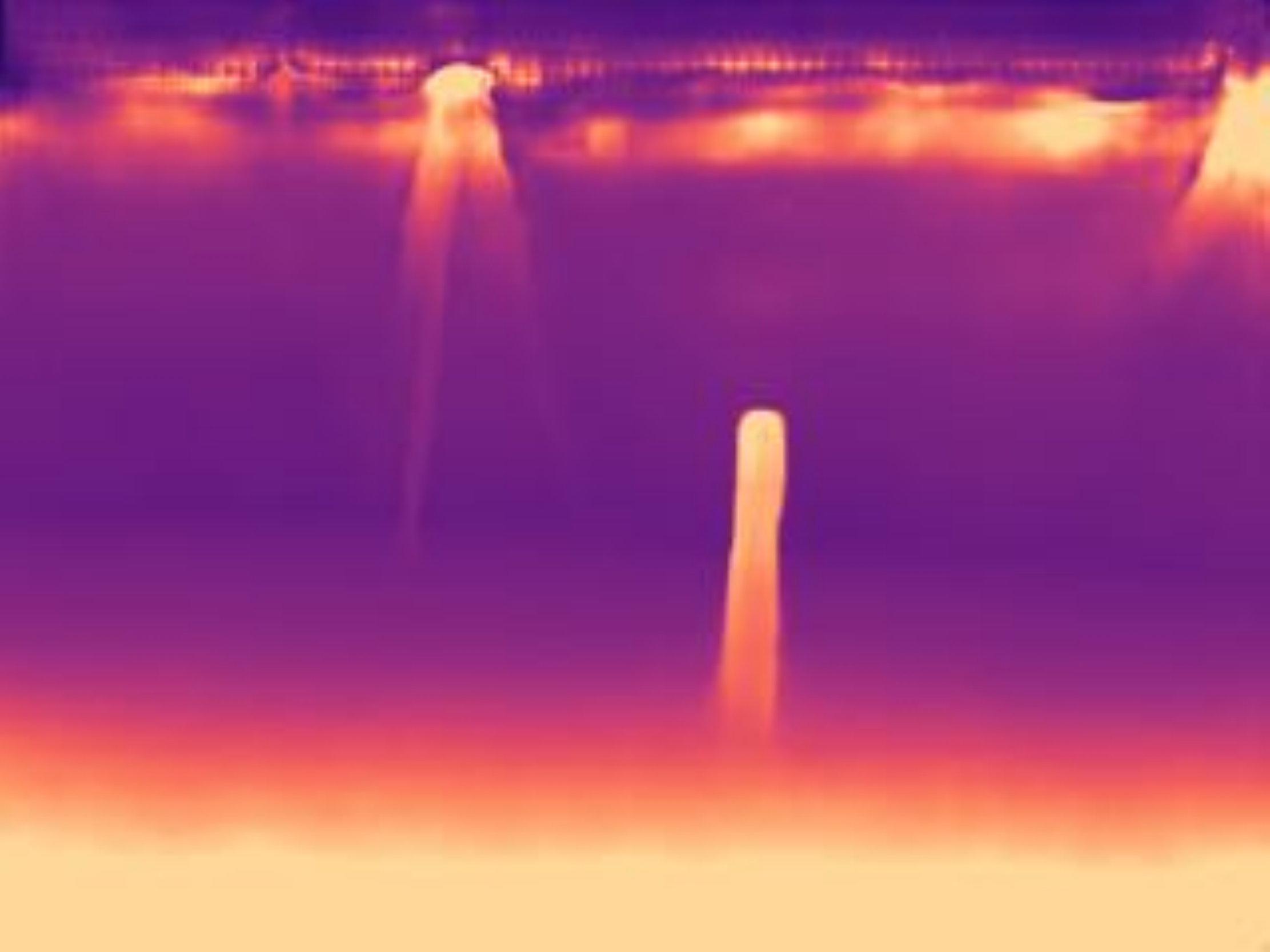} \qquad\qquad\quad & 
\includegraphics[width=\iw,height=\ih]{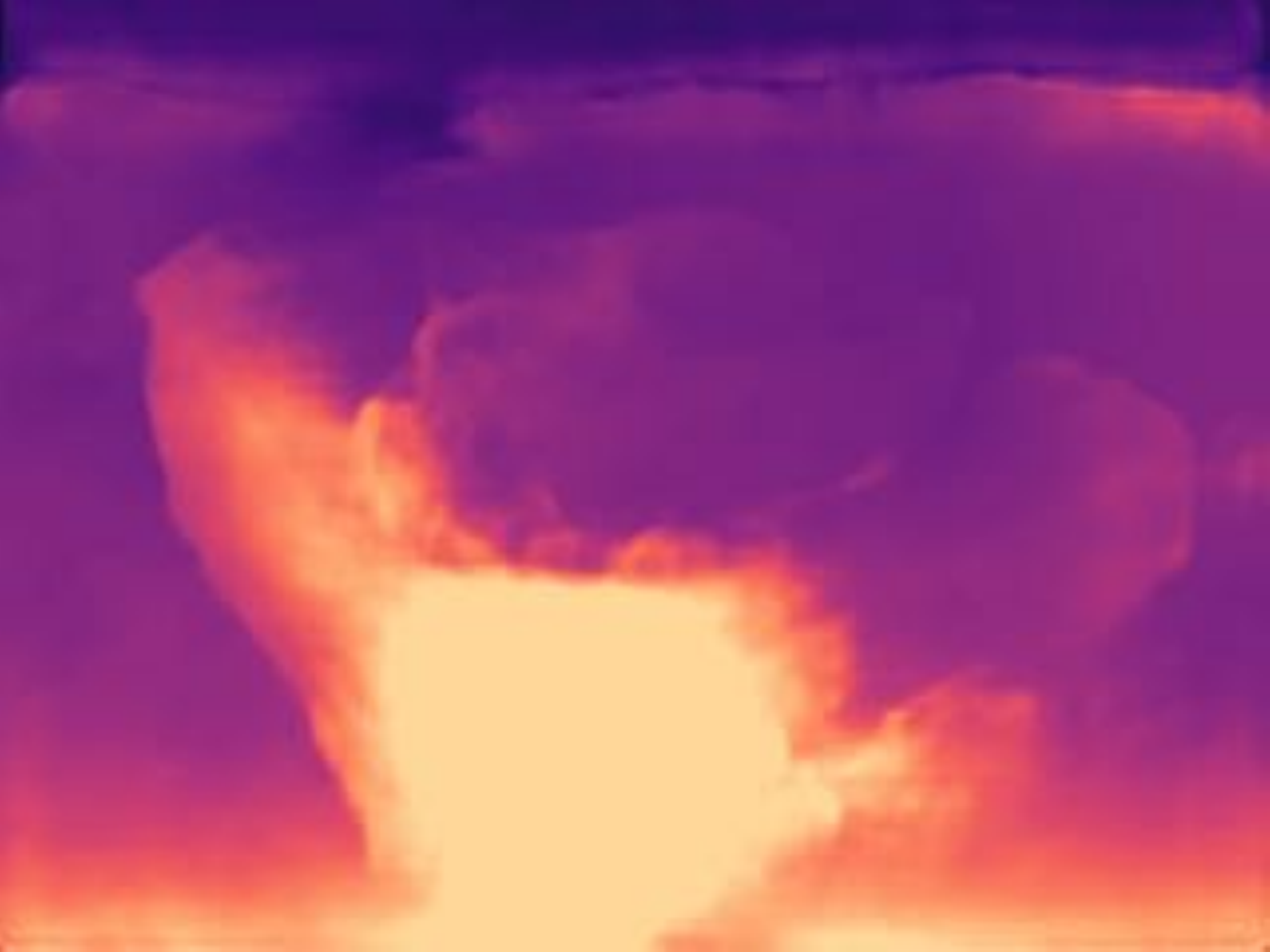} \qquad\qquad\quad & 
\includegraphics[width=\iw,height=\ih]{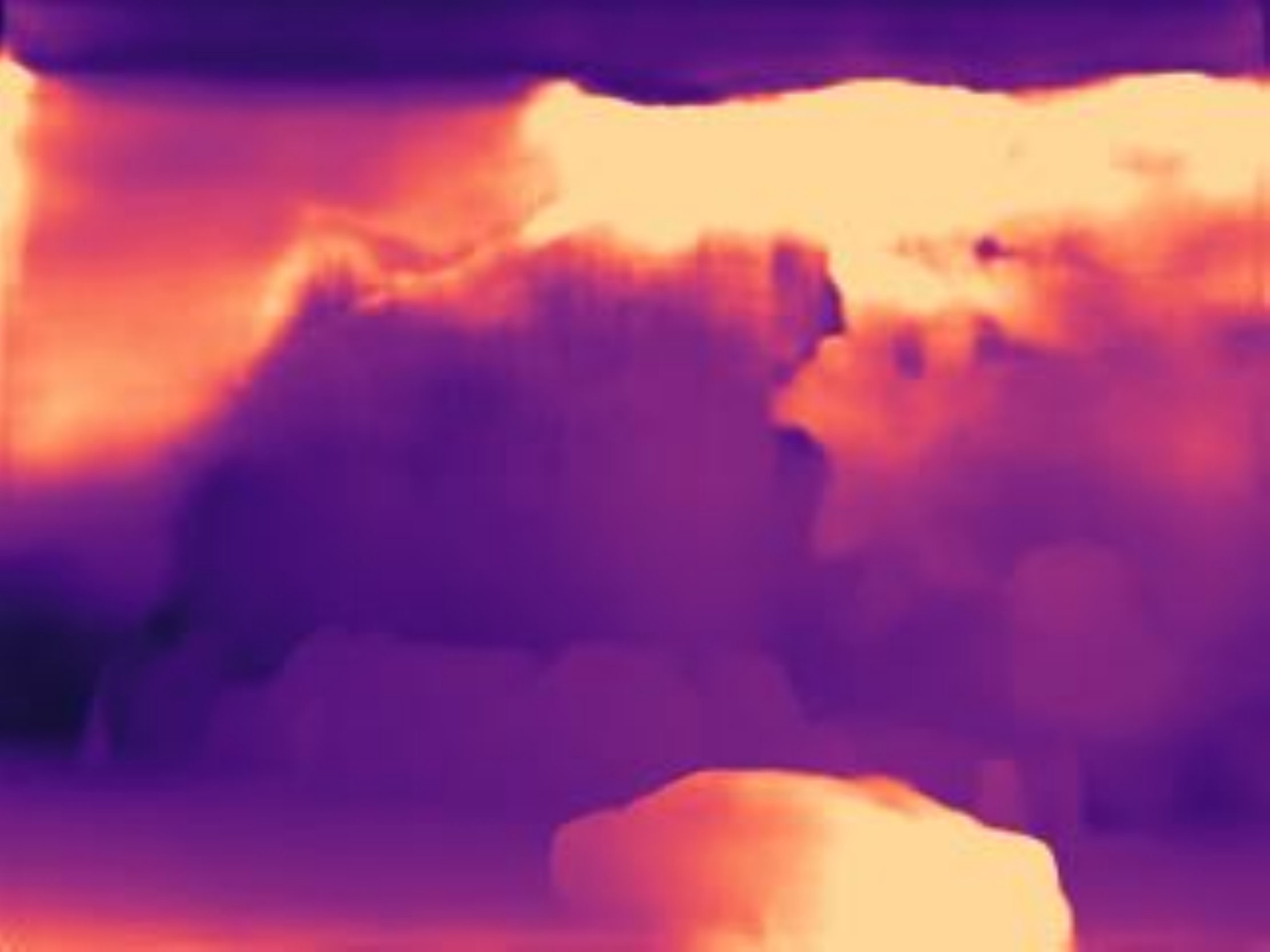} \qquad\qquad\quad & 
\includegraphics[width=\iw,height=\ih]{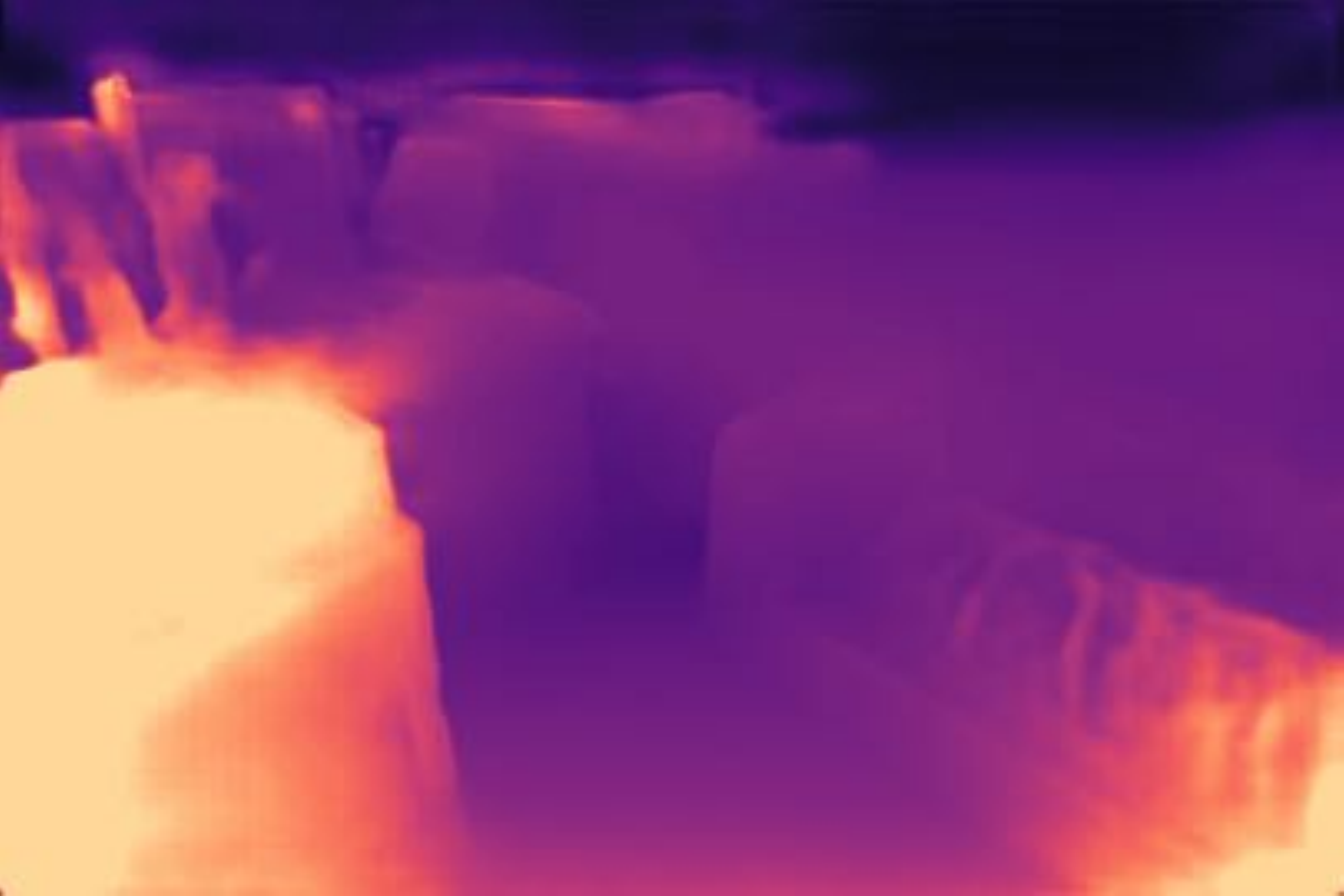} \qquad\qquad\quad & 
\includegraphics[width=\iw,height=\ih]{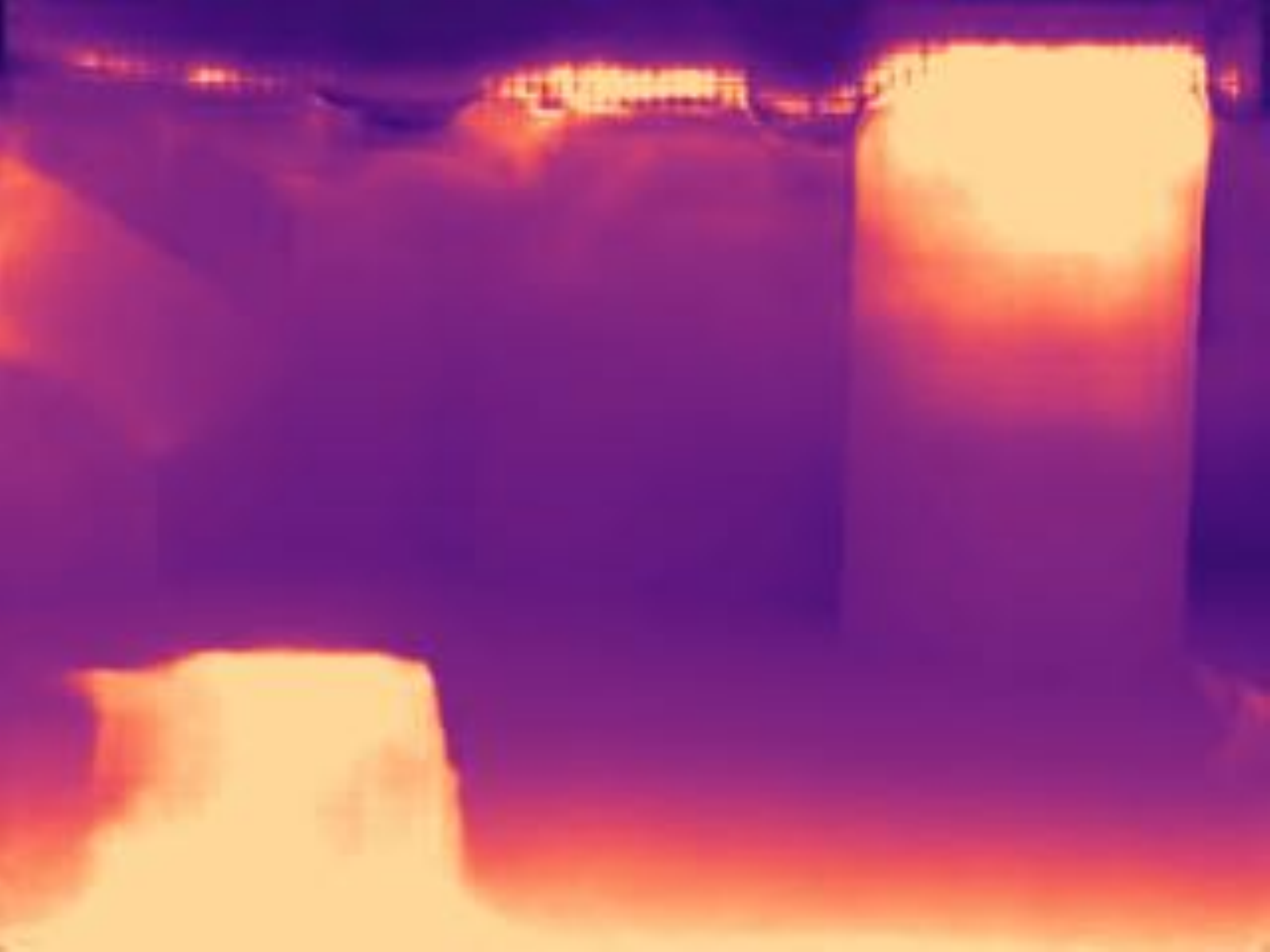}\\
\vspace{10mm}\\
\rotatebox[origin=c]{90}{\fontsize{\textw}{\texth} \selectfont DepthFormer\hspace{-270mm}}\hspace{24mm}
\includegraphics[width=\iw,height=\ih]{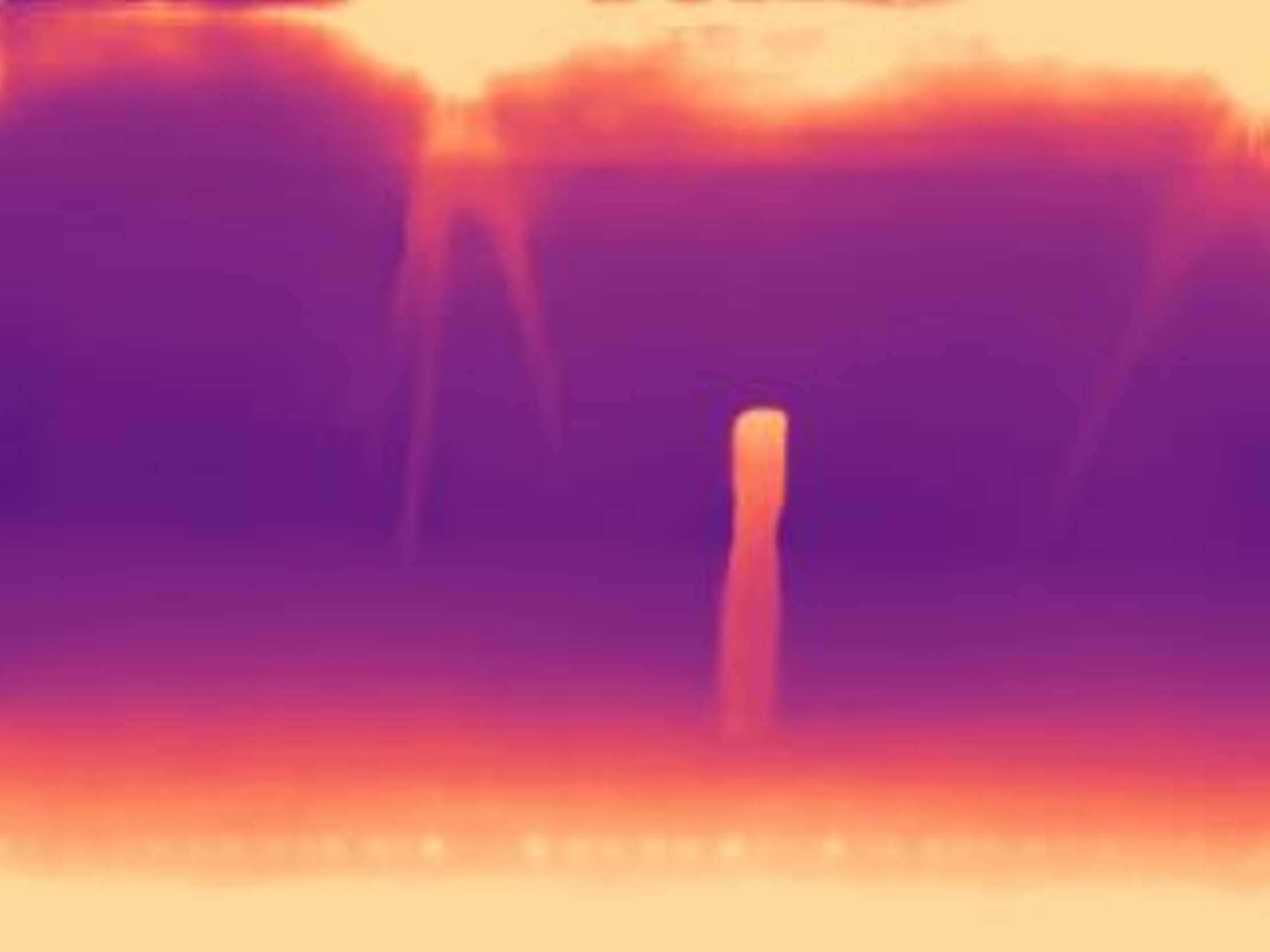} \qquad\qquad\quad & 
\includegraphics[width=\iw,height=\ih]{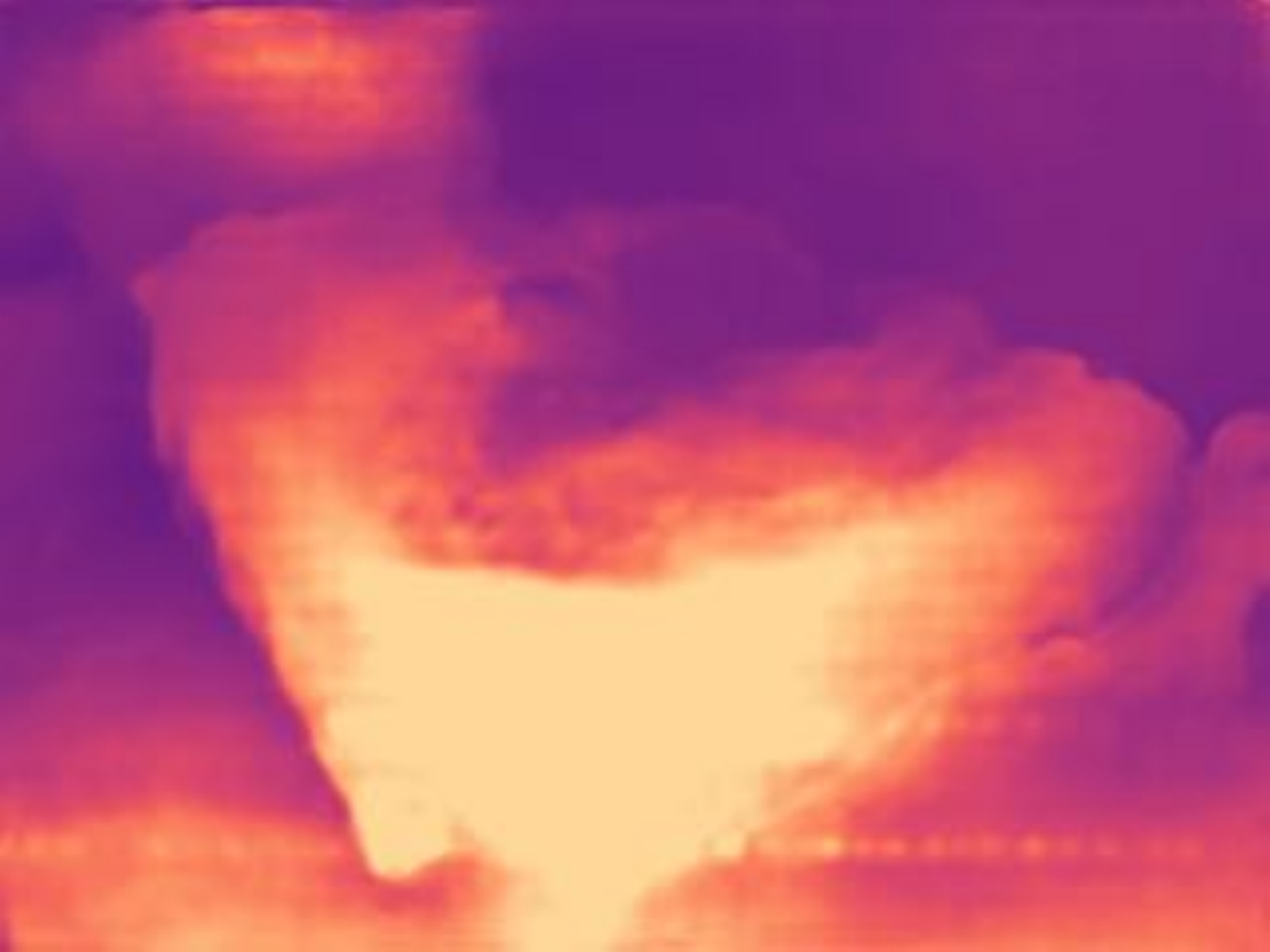} \qquad\qquad\quad & 
\includegraphics[width=\iw,height=\ih]{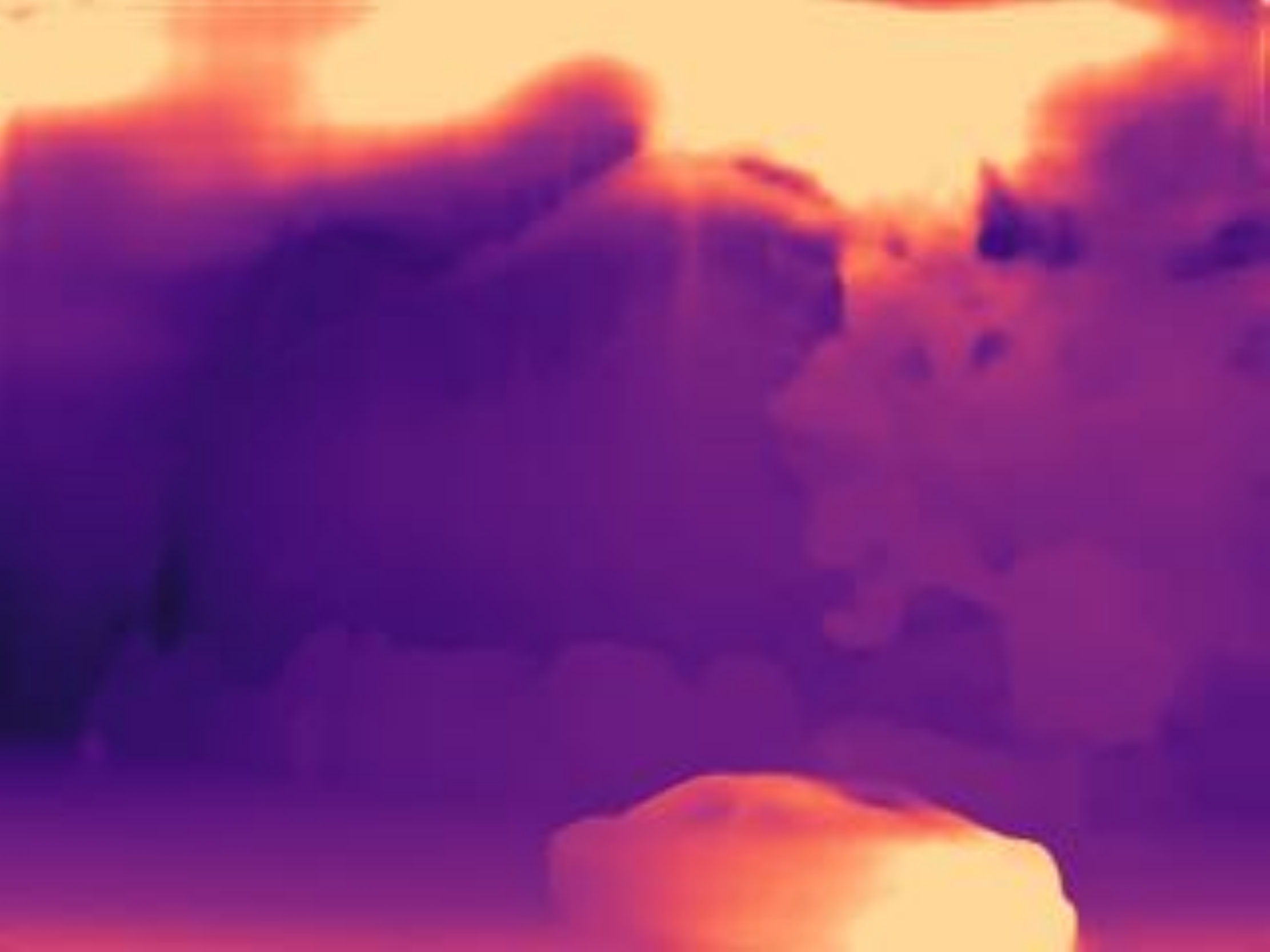} \qquad\qquad\quad & 
\includegraphics[width=\iw,height=\ih]{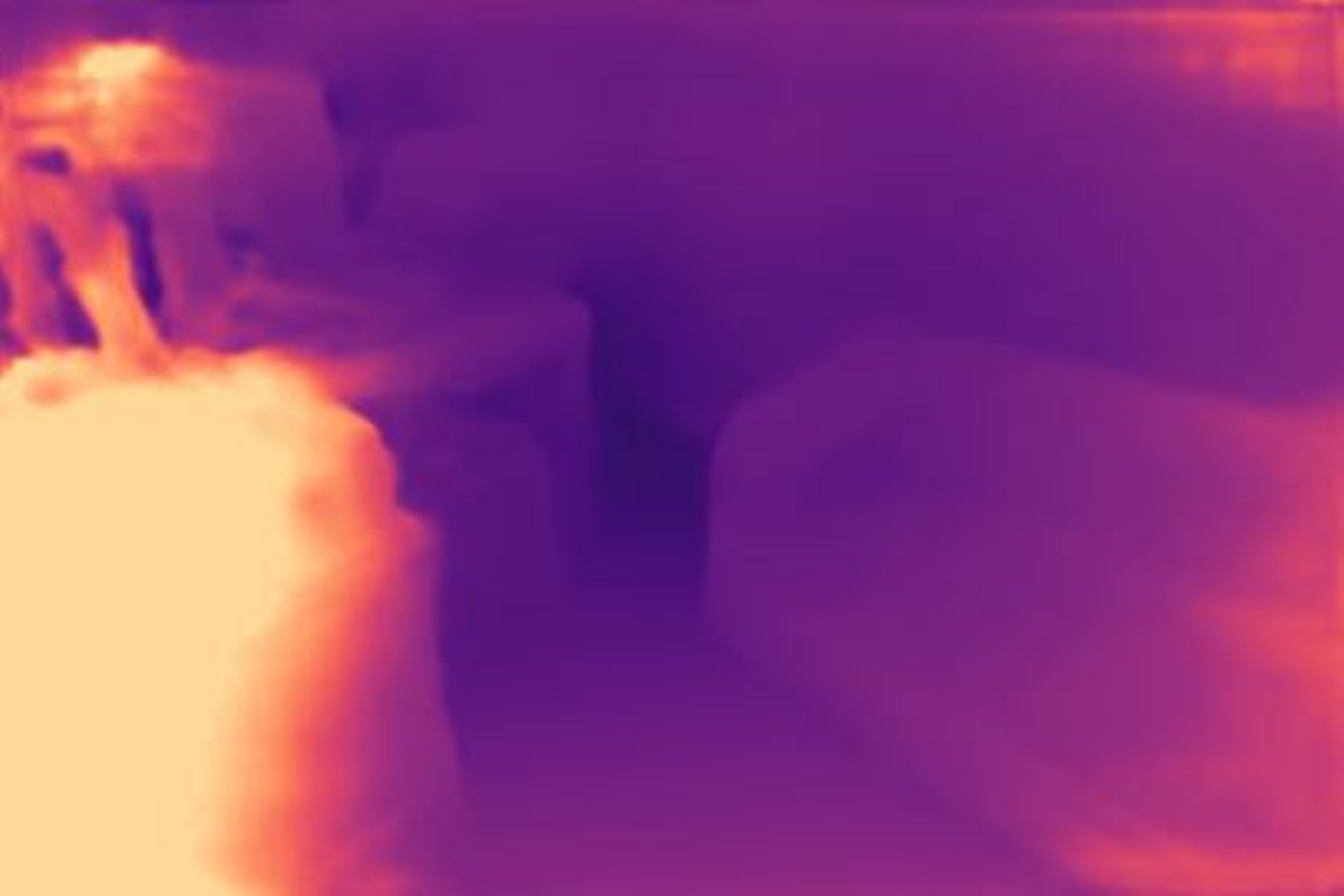} \qquad\qquad\quad & 
\includegraphics[width=\iw,height=\ih]{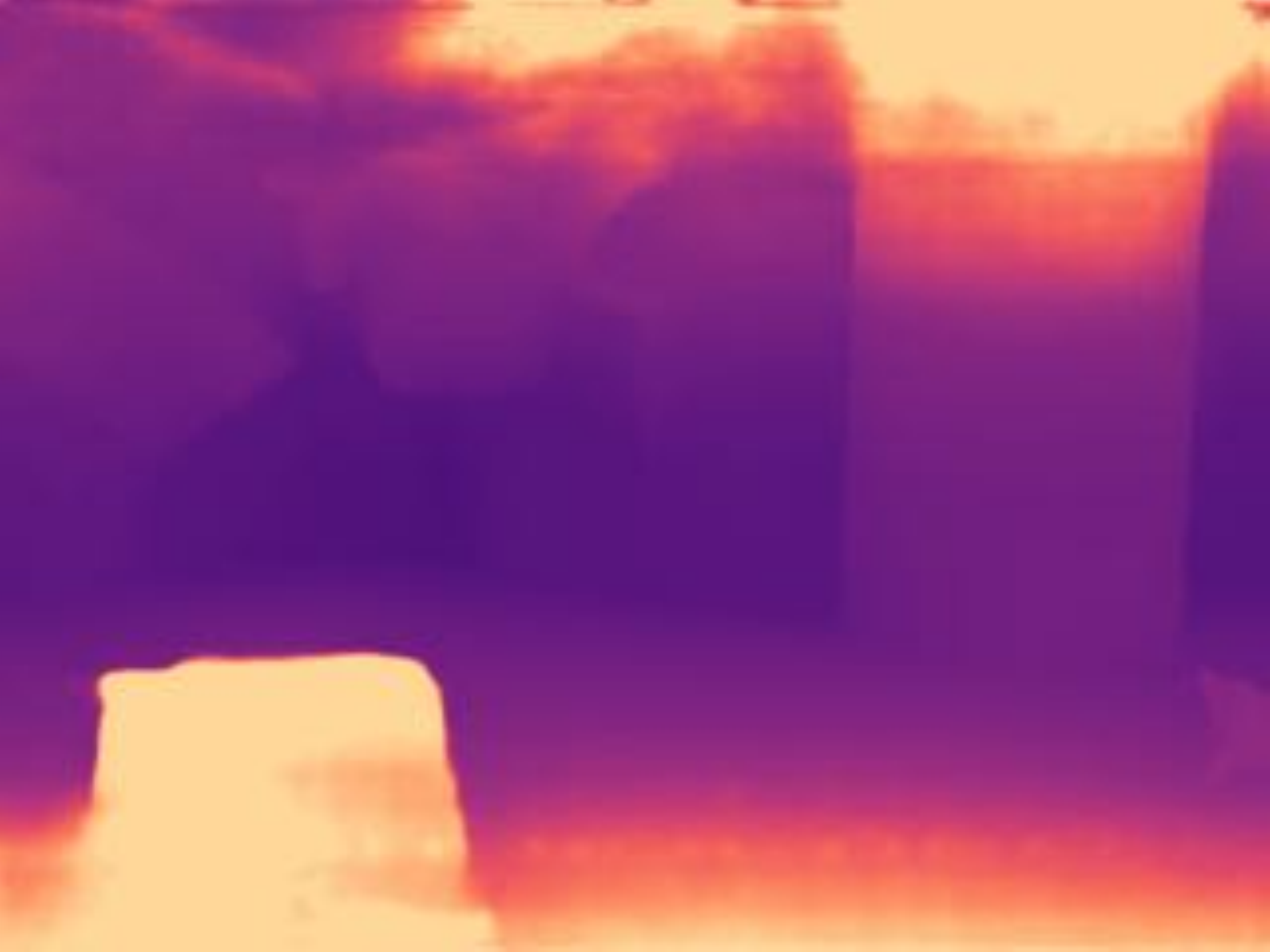}\\
\vspace{10mm}\\
\rotatebox[origin=c]{90}{\fontsize{\textw}{\texth} \selectfont GLPDepth\hspace{-270mm}}\hspace{24mm}
\includegraphics[width=\iw,height=\ih]{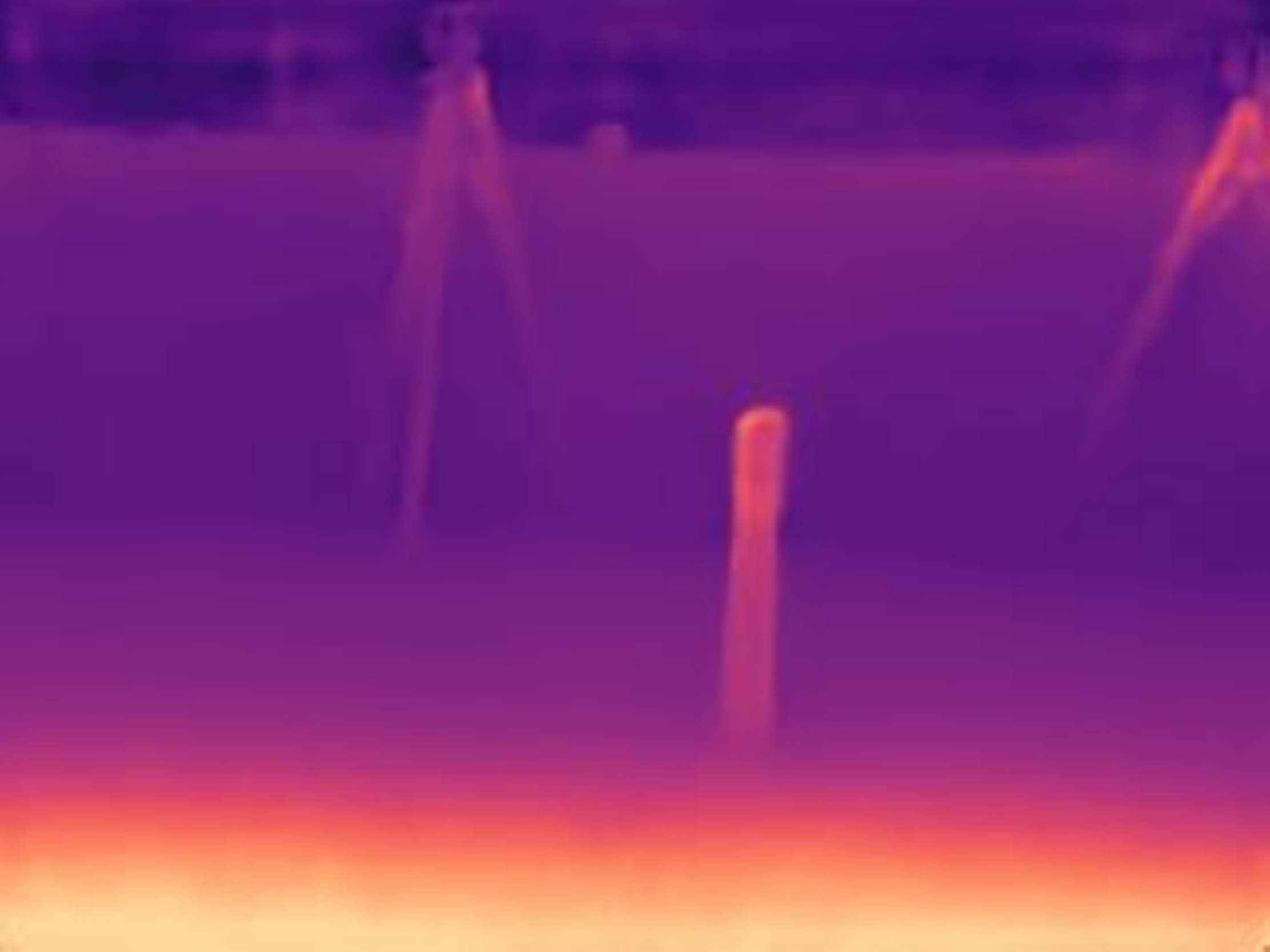} \qquad\qquad\quad & 
\includegraphics[width=\iw,height=\ih]{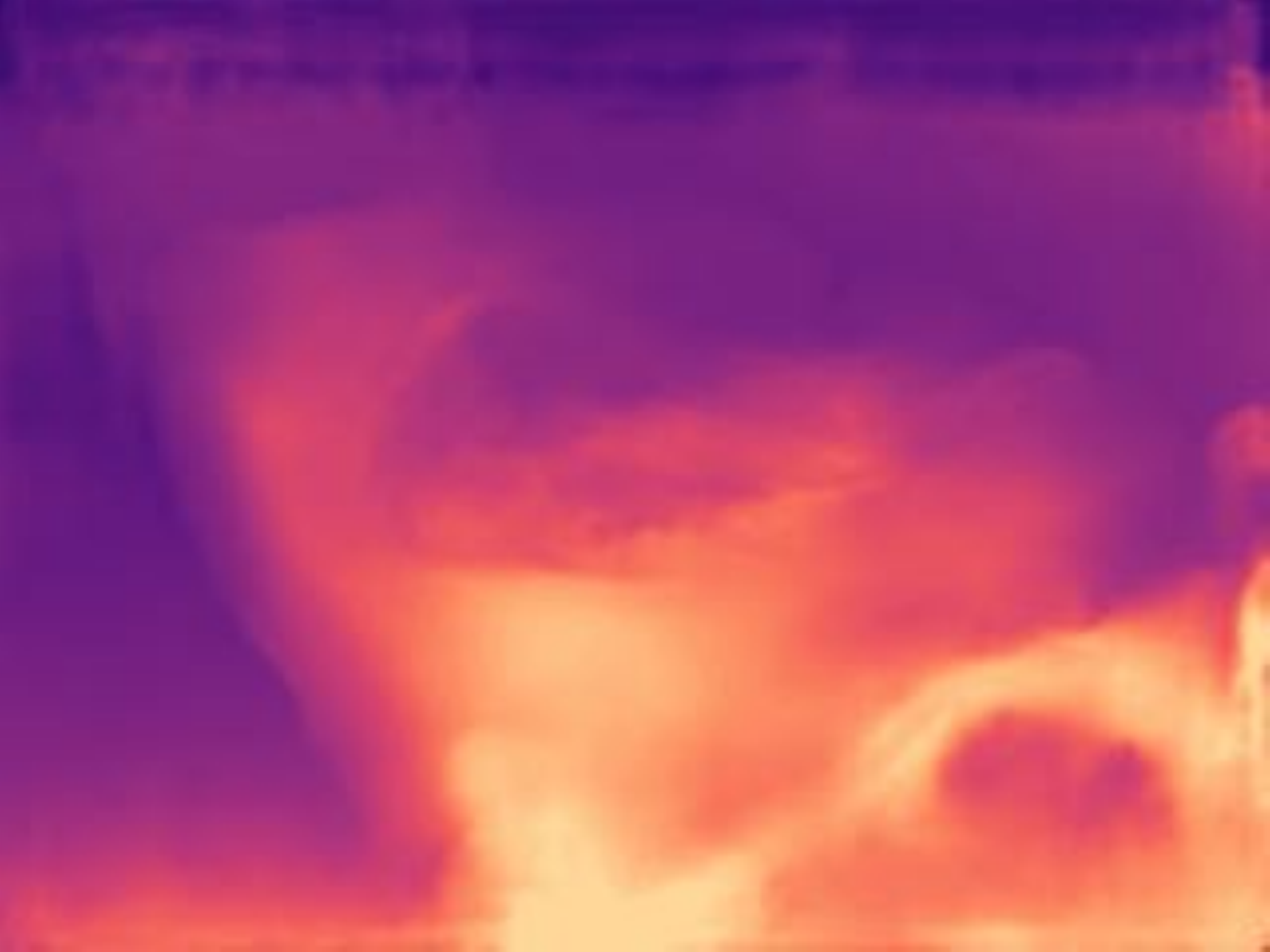} \qquad\qquad\quad & 
\includegraphics[width=\iw,height=\ih]{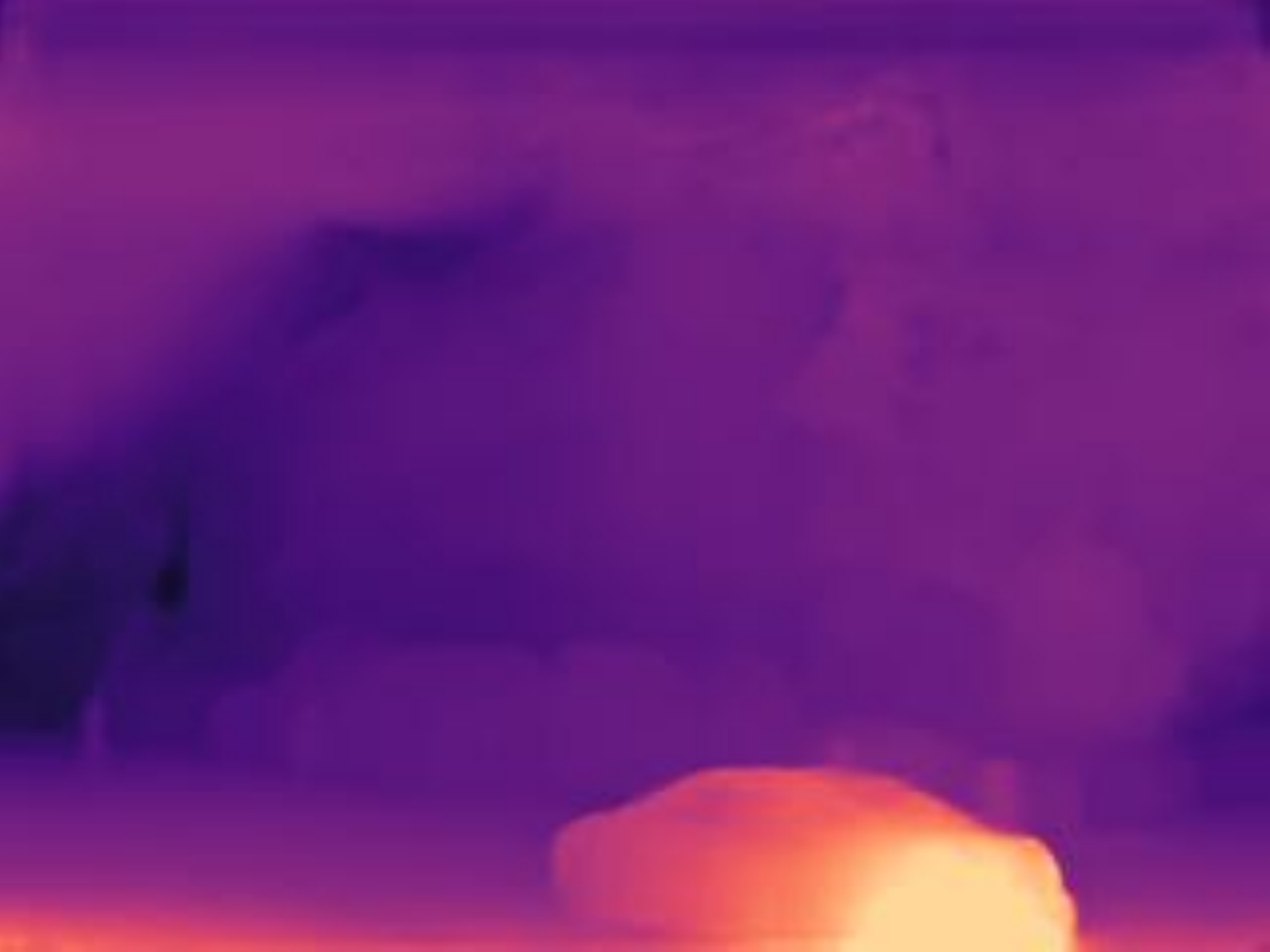} \qquad\qquad\quad & 
\includegraphics[width=\iw,height=\ih]{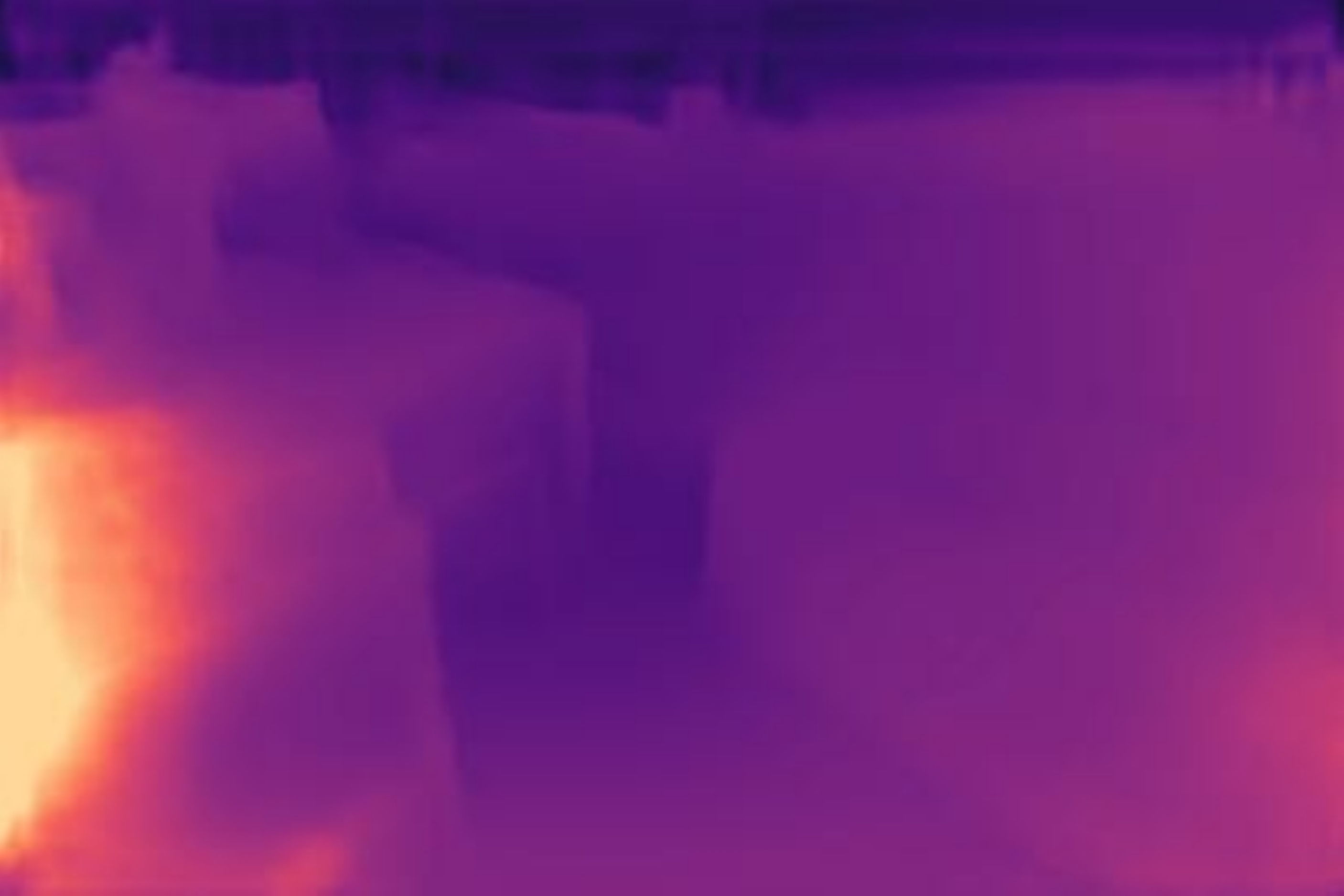} \qquad\qquad\quad & 
\includegraphics[width=\iw,height=\ih]{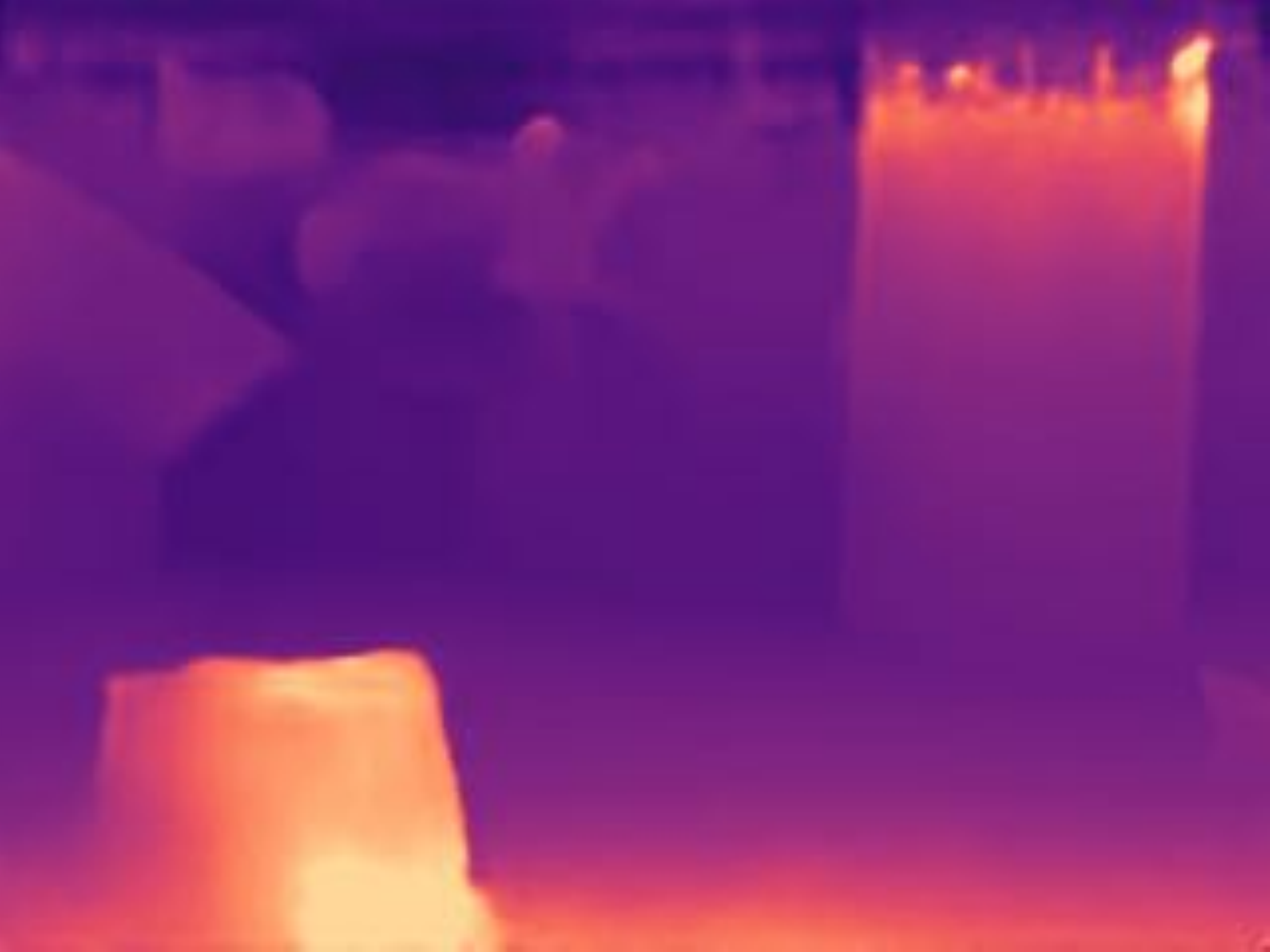}\\
\vspace{30mm}\\
\multicolumn{5}{c}{\fontsize{\w}{\h} \selectfont (d) Supervised Transformer-based methods } & 
\vspace{30mm}\\
\end{tabular}%
}
\caption{\textbf{Depth map results on out-of-distribution (RGBD, SUN3D, MVS, ETH3D, and Scenes11) datasets.}}
\label{figure_result_demons2}
\end{figure*}








\end{document}